\documentclass[10pt,twocolumn,letterpaper]{article}

\usepackage[accsupp]{axessibility} %

\usepackage{amssymb, amsmath, multirow}

\usepackage[svgnames,dvipsnames,x11names]{xcolor}

\usepackage{graphicx}
\usepackage{xparse}
\usepackage{expl3}

\usepackage{pgffor}
\usepackage{etoolbox}

\usepackage{placeins}

\usepackage{makecell}
\usepackage{array}

\usepackage{siunitx}
\usepackage{booktabs}
\usepackage{textcomp}

\usepackage{pgffor}
\usepackage{longtable}

\usepackage{mathtools}
\usepackage{float}
\usepackage{afterpage}

\usepackage[toc,page,header]{appendix}
\usepackage{minitoc}

\usepackage{amsthm}
\usepackage{thmtools}
\usepackage{thm-restate}

\declaretheorem[style=definition,name=Definition]{definitionplain}
\declaretheorem[style=remark]{remark}
\declaretheorem[style=definition,name=Example]{example}

\declaretheorem[style=plain,name=Corollary]{corollary}

\usepackage{needspace}
\usepackage{subcaption}

\def\ArxivMode{1}

\ifnum\ArxivMode=0
  \usepackage[review]{cvpr}      %
\else
  \usepackage[pagenumbers]{cvpr} %
\fi

\definecolor{darknavy}{RGB}{0,32,91}
\definecolor{richcobalt}{RGB}{0,71,171}
\definecolor{oxfordblue}{RGB}{0,33,71}
\definecolor{prussianblue}{RGB}{0,49,83}
\definecolor{ultramarine}{RGB}{18,10,143}
\definecolor{navy}{RGB}{0,0,128}
\definecolor{azure}{RGB}{0,83,156}
\definecolor{cobalt}{RGB}{0,71,171}
\definecolor{lightblue}{RGB}{173,216,230}
\definecolor{skyblue}{RGB}{135,206,235}
\definecolor{powderblue}{RGB}{176,224,230}

\definecolor{deepgoldenrod}{RGB}{184,134,11}
\definecolor{darkochre}{RGB}{204,119,34}
\definecolor{richbronze}{RGB}{205,127,50}
\definecolor{honey}{RGB}{197,145,0}
\definecolor{bronze}{RGB}{205,127,50}
\definecolor{ochre}{RGB}{204,119,34}
\definecolor{amber}{RGB}{255,191,0}
\definecolor{deepamber}{RGB}{255,191,0}
\definecolor{goldenrod}{RGB}{218,165,32}
\definecolor{deepsaffron}{RGB}{255,153,51}
\definecolor{lightyellow}{RGB}{255,255,224}
\definecolor{cream}{RGB}{255,253,208}
\definecolor{butter}{RGB}{255,248,179}

\definecolor{forestgreen}{RGB}{34,139,34}
\definecolor{huntergreen}{RGB}{53,94,59}
\definecolor{olive}{RGB}{85,107,47}
\definecolor{darkgreen}{RGB}{0,100,0}
\definecolor{evergreen}{RGB}{5,71,42}
\definecolor{pinetree}{RGB}{31,61,43}
\definecolor{emerald}{RGB}{46,139,87}
\definecolor{seagreen}{RGB}{46,139,87}
\definecolor{lightgreen}{RGB}{144,238,144}
\definecolor{mintgreen}{RGB}{152,255,152}
\definecolor{seafoam}{RGB}{159,226,191}

\definecolor{darkred}{RGB}{139,0,0}
\definecolor{maroon}{RGB}{128,0,0}
\definecolor{burgundy}{RGB}{128,0,32}
\definecolor{wine}{RGB}{114,47,55}
\definecolor{garnet}{RGB}{120,31,43}
\definecolor{crimson}{RGB}{139,0,0}
\definecolor{carmine}{RGB}{150,0,24}
\definecolor{rust}{RGB}{183,65,14}
\definecolor{lightred}{RGB}{255,102,102}
\definecolor{salmon}{RGB}{255,160,122}
\definecolor{coral}{RGB}{255,127,80}

\newcommand{\nightCite}[1]{%
  \ifnum\spacefactor>2000 \cite{#1}%
  \else\,\cite{#1}%
  \fi%
}

\newcommand{\nightCitet}[1]{%
  \ifnum\spacefactor>2000 \citet{#1}%
  \else\,\citet{#1}%
  \fi%
}
\newcommand{\ShortenedTo}[1]{henceforth #1}

\newcommand{\xpct}[1]{\mathbb E_{#1}}
\newcommand{\expected}[2]{\xpct{#1}\left[#2\right]}

\newcommand{\PLUS}{PLUS}

\newcommand{\ForwardAttn}{Forward Attention-Based Token Attribution}
\newcommand{\ensembleSep}{\:$\circ$\:}
\newcommand{\EVATwoCLIPLarge}{EVA2-CLIP-Large}
\newcommand{\RawAtt}{RawAtt}
\newcommand{\AttRollout}{Attention Rollout}

\newcommand{\attn}{\mathtt{Attn}}

\newcommand{\attngrad}{\mathtt{AttnGrad}}

\newcommand{\IxG}{%
  \ifmmode\text{Input}\!\times\!\text{Grad}\else Input\,$\times$\,Grad\fi%
}

\newcommand{\IxGShort}{%
  \ifmmode\operatorname{IxG}\else\textnormal{IxG}\fi%
}

\newcommand{\IG}{Integrated Gradients}
\newcommand{\IGShort}{%
  \ifmmode\operatorname{IG}\else\textnormal{IG}\fi%
}

\newcommand{\FullGradPLUS}{%
  \ifmmode\operatorname{FullGrad+}\else\textnormal{FullGrad+}\fi%
}
\newcommand{\FullGradPLUSLong}{FullGrad+\ensembleSep{}\PLUS{}}

\newcommand{\FairGrad}{LibraGrad}
\newcommand{\FairGradPrefix}{Libra}

\newcommand{\FairFullgrad}{\FairGradPrefix{} FullGrad}
\newcommand{\FairFullgradPLUS}{\FairGradPrefix{} FullGrad+}

\newcommand{\HiResCAM}{HiResCAM}

\newcommand{\GradCAMPLUS}{GradCAM+}
\newcommand{\GradCAMPLUSLong}{GradCAM\ensembleSep{}\PLUS{}}

\newcommand{\XGradCAMPLUS}{XGradCAM+}
\newcommand{\XGradCAMPLUSLong}{XGradCAM+\ensembleSep{}\PLUS{}}

\newcommand{\AttCAT}{AttCAT}
\newcommand{\GenAtt}{GenAtt}
\newcommand{\TokenTM}{TokenTM}

\newcommand{\relScore}{\mathtt{R}}

\newcommand{\LRPLong}{Layer-Wise Relevance Propagation (LRP)}

\newcommand{\AliLRPLong}{Conservative-LRP}
\newcommand{\AliLRP}{AliLRP}

\newcommand{\AttnLRP}{AttnLRP}

\newcommand{\DecompXLong}{DecompX-NoBias}
\newcommand{\DecompX}{DecompX}

\newcommand{\SRGLong}{Symmetric Relevance Gain (SRG)}
\newcommand{\SRGLongII}{Symmetric Relevance Gain, SRG}

\newcommand{\LIFLong}{Least-Influential-First Deletion (LIF)}
\newcommand{\LIFLongII}{Least-Influential-First Deletion, LIF}

\newcommand{\MIFLong}{Most-Influential-First Deletion (MIF)}

\newcommand{\SegmentationLong}{Segmentation Average Precision (AP)}

\newcommand{\OxfordPet}{Oxford-IIIT Pet}

\definecolor{mygray}{RGB}{200,200,200}

\newcommand{\spiderHRule}{%
    \color{mygray}%
    \noindent\hbox to \linewidth{\leaders\hbox to 15pt{\hss ... \hss}\hfil}\\%
    \color{black}%
  }
\newcommand{\inputWithLabel}[2]{%
  \begingroup%
  \let\oldlabel\label%
  \renewcommand{\label}[1]{\oldlabel{#2}}%
  \input{#1}%
  \let\label\oldlabel%
  \endgroup%
}

\newcommand{\inputTableStarWithLabel}[2]{%
  \begingroup%
  \renewenvironment{table}[1][]%
    {\begin{table*}[##1]}%
    {\end{table*}}%
  \let\oldlabel\label%
  \renewcommand{\label}[1]{\oldlabel{#2}}%
  \input{#1}%
  \let\label\oldlabel%
  \endgroup%
}

\newcommand{\inputTableSmallWithLabel}[2]{%
  \begingroup%
  \renewenvironment{table}[1][]%
    {\begin{table*}[##1]\small}%
    {\end{table*}}%
  \let\oldlabel\label%
  \renewcommand{\label}[1]{\oldlabel{#2}}%
  \input{#1}%
  \let\label\oldlabel%
  \endgroup%
}

\newcommand{\inputTableSmallWideWithLabel}[2]{%
  \begingroup%
  \let\oldlabel\label%
  \renewcommand{\label}[1]{\oldlabel{#2}}%
  \begingroup%
  \renewenvironment{table}[1][]%
    {\begin{table*}[##1]\small\setlength{\tabcolsep}{8pt}}%
    {\end{table*}}%
  \input{#1}%
  \endgroup%
  \let\label\oldlabel%
  \endgroup%
}

\newcommand{\inputTableSmallIIWithLabel}[2]{%
  \begingroup%
  \let\oldlabel\label%
  \renewcommand{\label}[1]{\oldlabel{#2}}%
  \begingroup%
  \renewenvironment{table}[1][]%
    {\begin{table*}[##1]\small\setlength{\tabcolsep}{4pt}}%
    {\end{table*}}%
  \input{#1}%
  \endgroup%
  \let\label\oldlabel%
  \endgroup%
}

\newcommand{\inputTableNSC}[1]{%
  \begingroup%

  \let\origtable\table
  \let\endorigtable\endtable

  \renewenvironment{table}[1][]%
    {\begin{origtable}[##1]\setlength{\tabcolsep}{1pt}}%
    {\end{origtable}}%
  \input{#1}%
  \endgroup%
}

\newcommand{\inputTableSS}[1]{%
  \begingroup%
  \let\origtable\table
  \let\endorigtable\endtable

  \renewenvironment{table}[1][]%
    {\begin{origtable}[##1]\small\setlength{\tabcolsep}{3pt}}%
    {\end{origtable}}%
  \input{#1}%
  \endgroup%
}

\newcommand{\inputTableSSWithLabel}[2]{%
  \begingroup%
  \let\oldlabel\label%
  \renewcommand{\label}[1]{\oldlabel{#2}}%
  \input{#1}%
  \let\label\oldlabel%
  \endgroup%
}

\let\originalincludegraphics\includegraphics
\ExplSyntaxOn
\NewDocumentCommand{\includeGraphicsNormalizedPath}{ o m }
 {
  \tl_set:Nn \l_tmpa_tl { #2 }

  \tl_trim_spaces:N \l_tmpa_tl

  \str_replace_all:Nnn \l_tmpa_tl { : } { _ }
  \str_replace_all:Nnn \l_tmpa_tl { & } { _ }

  \str_replace_all:Nnn \l_tmpa_tl { ~ } { _ }
  \str_replace_all:Nnn \l_tmpa_tl { \c_space_tl } { _ }

  \IfNoValueTF{#1}
    {\originalincludegraphics{\l_tmpa_tl}}
    {\originalincludegraphics[#1]{\l_tmpa_tl}}
 }
\ExplSyntaxOff

\newcommand{\CLIPRows}[4][h]{%
  \begin{figure}[#1]
    \centering
    \begingroup
    \def\do##1{%
      \includegraphics[width=\linewidth]{##1}
    }%
    \docsvlist{#2}%
    \endgroup
    \ifx\relax#3\relax
    \else
      \caption{#3}
    \fi
    \ifx\relax#4\relax\else
      \label{#4}
    \fi
  \end{figure}
}
\newcommand{\TableRows}[5][ht]{%
  \begin{table}[#1]
    #2
    \begingroup
    \def\do##1{%
  \begingroup%
  \renewenvironment{table}[1][]%
    {\begin{center}}%
    {\end{center}}%
  \input{##1}%
  \endgroup%
    }%
    \docsvlist{#3}%
    \endgroup
    \ifx\relax#4\relax\else
      \caption{#4}%
    \fi
    \ifx\relax#5\relax\else
      \label{#5}%
    \fi
  \end{table}
}

\newcommand{\TableRowsA}[4][ht]{%
  \TableRows[#1]{%
    \fontsize{11pt}{8.5pt}\selectfont
    \setlength{\tabcolsep}{4pt}
    \centering
  }{#2}{#3}{#4}%
}
\newcommand{\TableRowsB}[4][ht]{%
  \TableRows[#1]{%
    \fontsize{9.5pt}{8.5pt}\selectfont
    \setlength{\tabcolsep}{2pt}
    \centering
  }{#2}{#3}{#4}%
}
\newcommand{\nightParagraph}[1]{\paragraph{#1.}}

\definecolor{cvprblue}{rgb}{0.21,0.49,0.74}

\usepackage[pagebackref,     %
            breaklinks,       %
            colorlinks,       %
            allcolors=cvprblue, %
            bookmarksnumbered=true, %
            bookmarksopen=true,     %
            bookmarksdepth=3]{hyperref}

\def\paperID{4068} %
\def\confName{CVPR}
\def\confYear{2025}

\doparttoc %
\faketableofcontents %
\title{\FairGrad{}: Balancing Gradient Flow \\for Universally Better Vision Transformer Attributions}

\author{%
  Faridoun Mehri$^1$ \quad
  Mahdieh Soleymani Baghshah$^1$ \quad
  Mohammad Taher Pilehvar$^2$
  \vspace{0.3em} \\
  {\normalsize $^1$Sharif University of Technology, Iran} \quad
  {\normalsize $^2$Cardiff University, UK}
  \vspace{0.3em} \\
  {\tt\small f.meh16@student.sharif.edu}%
  \quad
  {\tt\small soleymani@sharif.edu} \quad
  {\tt\small pilehvarmt@cardiff.ac.uk}
}

\begin{document}
\maketitle
\begin{abstract}
  Why do gradient-based explanations struggle with Transformers, and how can we improve them?
  We identify gradient flow imbalances in Transformers that violate FullGrad-completeness, a critical property for attribution faithfulness that CNNs naturally possess. To address this issue, we introduce \FairGrad{}---a theoretically grounded post-hoc approach that corrects gradient imbalances through pruning and scaling of backward paths, without changing the forward pass or adding computational overhead.
  We evaluate \FairGrad{} using three metric families: Faithfulness, which quantifies prediction changes under perturbations of the most and least relevant features; Completeness Error, which measures attribution conservation relative to model outputs; and Segmentation AP, which assesses alignment with human perception.
  Extensive experiments across 8 architectures, 4 model sizes, and 4 datasets show that \FairGrad{} universally enhances gradient-based methods, outperforming existing white-box methods---including Transformer-specific approaches---across all metrics.
  We demonstrate superior qualitative results through two complementary evaluations: precise text-prompted region highlighting on CLIP models and accurate class discrimination between co-occurring animals on ImageNet-finetuned models---two settings on which existing methods often struggle.
  \FairGrad{} is effective even on the attention-free MLP-Mixer architecture, indicating potential for extension to other modern architectures.
\ifnum\ArxivMode=0
  Our code is freely available.
\else
  Our code is freely available at {\tt\small \url{https://github.com/NightMachinery/LibraGrad}}.
\fi
\end{abstract}

\section{Introduction}
\label{sec:intro}

\setlength{\textfloatsep}{0pt}
\setlength{\abovecaptionskip}{5pt}
\setlength{\belowcaptionskip}{10pt}

\begingroup%
  \begin{figure}[!t]
    \centering
    \includegraphics[width=\linewidth, trim=0 0pt 0 0]{
      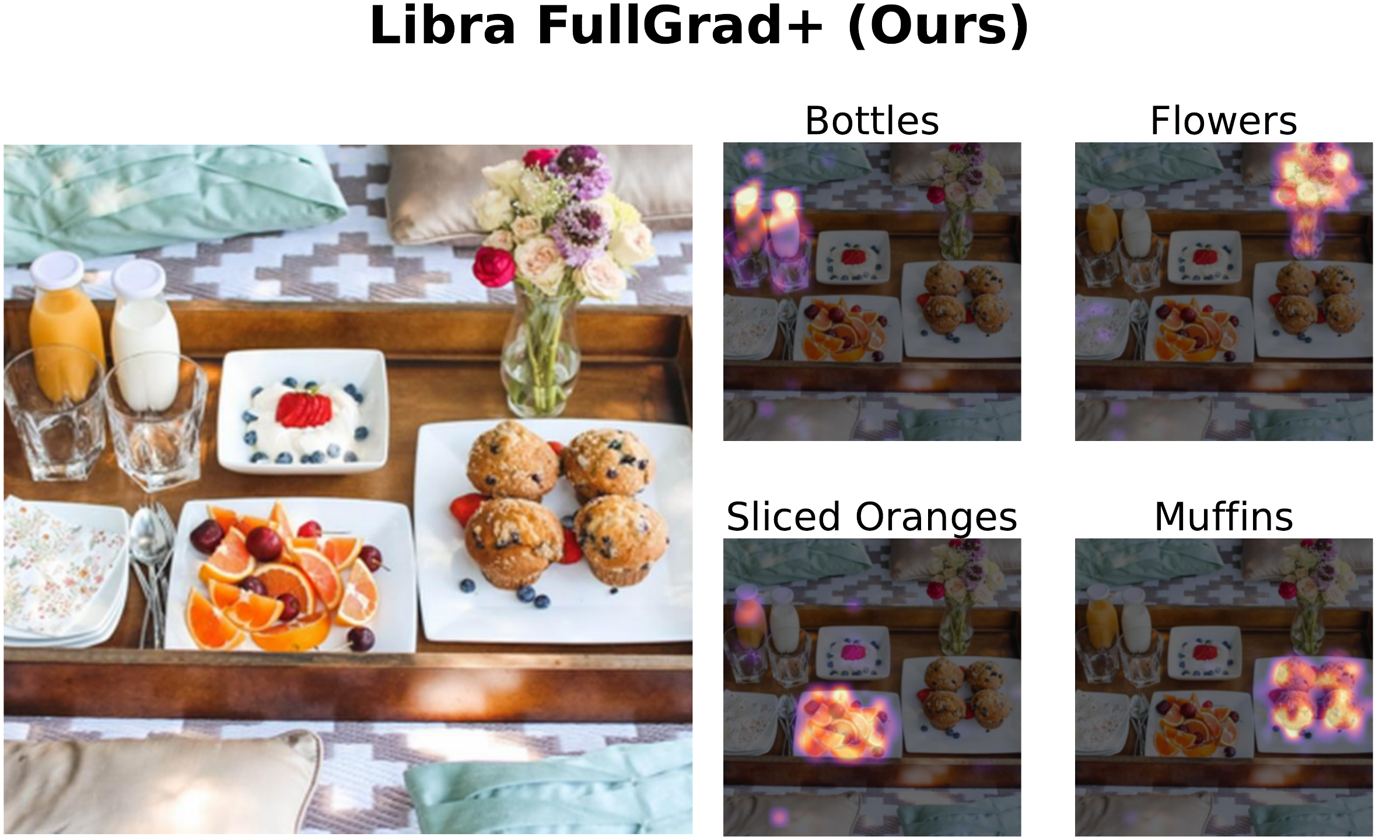
    }
    \par{}
    \vspace*{9pt}
    \includegraphics[width=0.99\linewidth, trim=0 0pt 0pt 0pt]{
      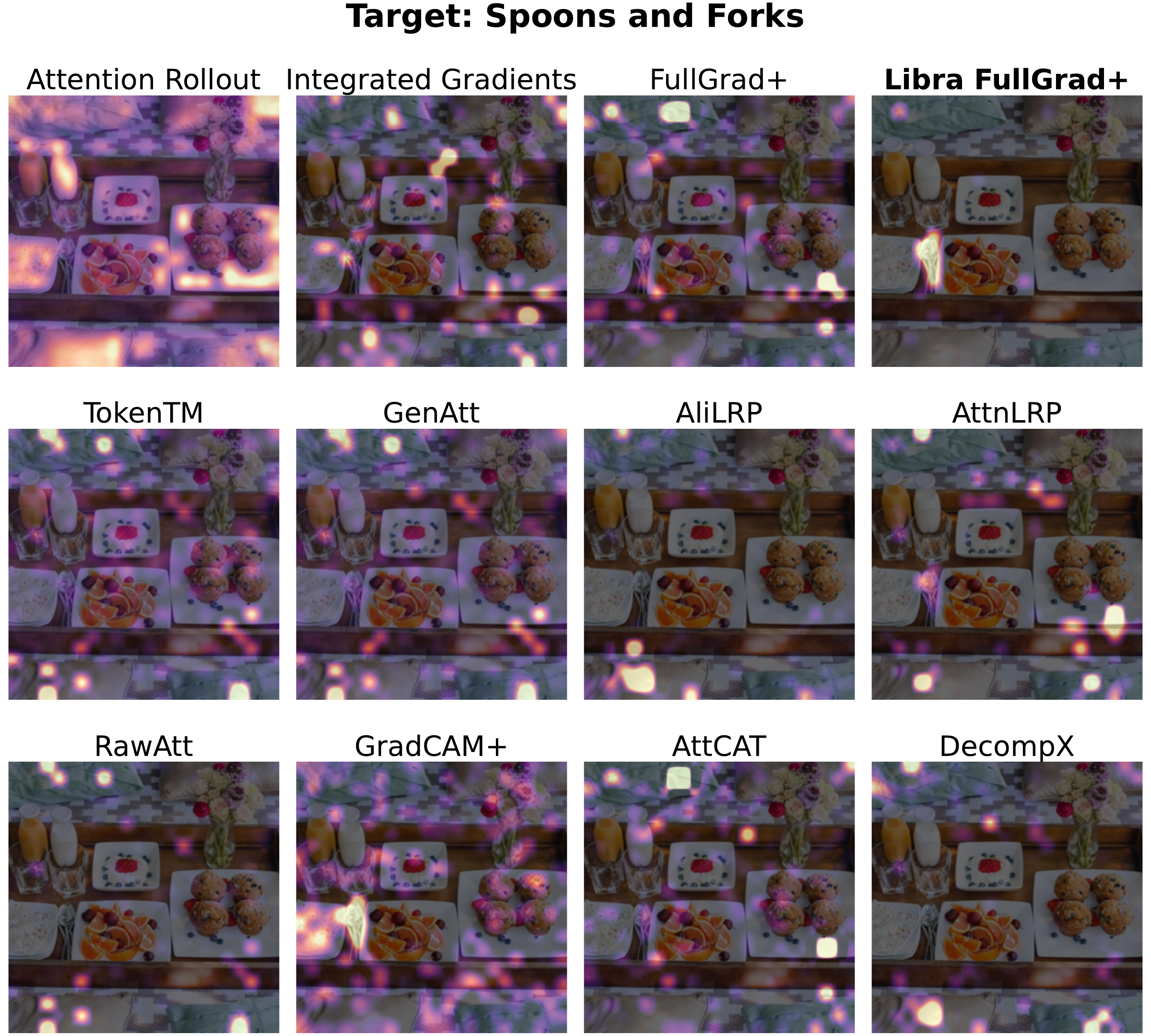
    }
    \caption{
      Qualitative comparison on \EVATwoCLIPLarge{}.
      Our proposed \FairFullgradPLUS{} generates
      prompt-specific
      attribution maps (top) and demonstrates improved localization compared to existing methods when explaining the model output for the ``spoons and forks'' prompt (bottom).
      For more qualitative examples, see Fig.~\ref{qual:FP_zele} and Appendix~\ref{apn:qual}.
    }
    \label{qual:FP}
  \end{figure}
\endgroup%

Understanding how deep learning models make decisions is crucial for deploying them in critical applications such as healthcare and autonomous driving. Input attribution methods, which quantify the influence of individual input features on a model's output\nightCite{Samek2021ExplainingDN, Binder2016LayerWiseRP, LYU2022TowardsFM, Madsen2021PosthocIF}, help us understand a model's decision for a single input and also serve as building blocks for advanced explanation techniques like CRAFT\nightCite{Fel2022CRAFTCR}.

In the field of CNN interpretability, gradient-based attribution techniques---particularly Integrated Gradients\nightCite{pmlr-v70-sundararajan17a} and FullGrad\nightCite{Srinivas2019FullGradientRF}---established a foundation for model explanation.
However, the architectural paradigm shift brought about by Vision Transformers (ViTs)\nightCite{vaswani-2017-attention, dosovitskiy-2021-image} has exposed limitations in these gradient-based methods, with attention-based attribution methods sometimes achieving more success. Hybrid methods, including \GenAtt{}\nightCite{chefer-2021-generic}, \TokenTM{}\nightCite{Wu2024TokenTM}, and \AttCAT{}\nightCite{qiang-2022-attcat}, attempt to bridge this gap by integrating gradient and attention-based approaches. Nonetheless, significant challenges persist: these methods lack theoretical foundations, struggle to distinguish between classes effectively, produce noisy attribution maps, and often work only with specific model architectures.

In this work, we identify the root cause of the failure of gradient-based methods: unbalanced gradient flow during backpropagation leads to unfaithful attribution scores. We demonstrate that while classical CNNs naturally preserve proper gradient flow through their locally affine operations, several components in modern Transformers
disrupt this property.

Our solution, \FairGrad{}, takes a different approach: instead of working around distorted gradients, it prevents the distortion from occurring in the first place by theoretically motivated pruning and scaling of backward paths, leaving the forward pass untouched.
Our comprehensive experiments across 8 architectures, 4 model sizes, and 4 datasets show that this not only improves all gradient-based attribution methods but also reveals that specialized attention-gradient hybrids are unnecessary---once gradients flow properly, the general-purpose \FairFullgradPLUS{} achieves superior or comparable performance.
We also extend \IG{} (\IGShort{}) \nightCite{pmlr-v70-sundararajan17a} and compose it with other gradient-based methods, and compare the universal improvement aspect of \FairGrad{} and \IGShort{}, showing \FairGrad{} vastly outperforms \IGShort{}. Furthermore, we theoretically prove that this is to be expected.

\section{Background and Related Work}
\label{sec:background}

Given a multi-output neural model, let $f: \mathbb{R}^n \rightarrow \mathbb{R}$ be a selected output function. For instance, if $\text{Model}(x) = (p_1, ..., p_k)$ represents class probabilities, we might choose $f(x) = p_i$ to analyze the model's prediction for the $i$-th class. An attribution method $A$ generates relevance scores $A(f)(x)_i$ for each feature $x_i$.

\subsection{Gradient-Based Attribution Methods}
\nightParagraph{\IxG{}} \IxGShort{}\nightCite{Shrikumar2016NotJA, Shrikumar2017LearningIF, Ancona2017TowardsBU} assigns feature relevance by
$\IxGShort{}(f)(x) = x \odot \nabla_x f(x)
$, where $\odot$ denotes element-wise multiplication.
\nightParagraph{FullGrad}
Expanding on \IxG{}, FullGrad\nightCite{Srinivas2019FullGradientRF} includes not only the input features but also the bias terms of each layer in the neural network. The FullGrad attribution map is calculated as:
\begin{equation*}
\text{FullGrad}(f)(x_{0}) = \IxGShort{}(f)(x_0) + \sum_{l=0}^{L-1} \sum_{b \in B_l} \IxGShort{}(f_{b})(b)
\end{equation*}
where $\IxGShort{}(f)(x_0)$ denotes the \IxG{} for the input $x_0$, and $\IxGShort{}(f_{b})(b)$ is the \IxG{} attribution map of the sub-network $f_{b}$ with a bias term $b$ from layer $l$ as the input.
Also, $f_b$ is the sub-network of $f$ starting from the bias term $b$ and going until the end of the model, whereas $B_l$ denotes the set of all bias terms in layer $l$.
\FullGradPLUSLong{} (\ShortenedTo{\FullGradPLUS{}})\nightCite{skipplus-cvprw24} is defined as follows:
\begin{equation*}
\begin{aligned}
  \FullGradPLUS{} & (f)(x_{0}) = \\
  & \sum_{l=0}^{L-1} \IxGShort{}(f_l)(x_l) + \sum_{l=0}^{L-1} \sum_{b \in B_l} \IxGShort{}(f_{b})(b)
\end{aligned}
\end{equation*}
where $\IxGShort{}(f_l)(x_l)$ is the \IxG{} attribution map of the sub-network $f_l$ with input $x_l$. \FullGradPLUS{} aggregates the input attribution maps of each layer along with the attribution maps of all bias terms in each layer.

\nightParagraph{\IG{}}
\IGShort{}\nightCite{pmlr-v70-sundararajan17a} computes attributions \wrt{} a baseline input $\bar{x}$ (\eg{}, zero):
\begin{equation*}
\IGShort{}(f)(x) = (x-\bar{x}) \odot \int_{\alpha=0}^1 \nabla_x f(\bar{x}+\alpha(x-\bar{x})) d\alpha
\end{equation*}

In practice, we approximate the integral using a 50-step Riemann summation.

\subsection{Other Attribution Methods}
In addition to the primary gradient-based methods above, we apply \FairGrad{} to several other general-purpose gradient methods, including \HiResCAM{}\nightCite{Draelos2020UseHI}, \GradCAMPLUSLong{} (\ShortenedTo{\GradCAMPLUS{}})\nightCite{selvaraju-2017-gradcam, skipplus-cvprw24,Jiang2021LayerCAMEH}, and \XGradCAMPLUSLong{} (\ShortenedTo{\XGradCAMPLUS{}})\nightCite{Fu2020AxiombasedGT, skipplus-cvprw24}. We further apply it to hybrid attention-gradient approaches specifically designed for Transformer architectures: \GenAtt{} (also known as GAE)\nightCite{chefer-2021-generic}, \TokenTM{}\nightCite{Wu2024TokenTM}, and \AttCAT{}\nightCite{qiang-2022-attcat}. To ensure a comprehensive evaluation, we also compare against attention-based attribution methods \RawAtt{}\nightCite{Caron2021EmergingPI, chefer-2021-transformer, Hao2020SelfAttentionAI}, \AttRollout{}\nightCite{abnar-2020-quantifying}, and \DecompXLong{} (\ShortenedTo{\DecompX{}})\nightCite{modarressi-2023-decompx}, as well as Transformer-specific \LRPLong{}-based\nightCite{Bach2015OnPE} techniques \AliLRPLong{} (\ShortenedTo{\AliLRP{}})\nightCite{pmlr-v162-ali22a} and \AttnLRP{}\nightCite{pmlr-v235-achtibat24a}. For a detailed overview of related work, see Appendix~\ref{apn:relatedWork}.

\section{Method}
\label{sec:method}

Understanding how input features contribute to a model's output is a central goal of attribution methods. For gradient-based attributions to be faithful, they must accurately reflect the influence of each input feature on the output.
This requires decomposing model outputs into input and bias contributions, formalized as:

\begin{definitionplain}%
A function $f$ is \textbf{FullGrad-complete} (or \textbf{FG-complete}) if, for all $x \in \mathbb{R}^n$,
\[
f(x) = J_x f \cdot x + \sum_{i} J_{b_i} f \cdot b_i,
\]
where $J_x f = \frac{\partial f}{\partial x} \in \mathbb{R}^{m \times n}$ is the Jacobian matrix of $f$ with respect to $x$, and $J_{b_i} f = \frac{\partial f}{\partial b_i} \in \mathbb{R}^{m \times d_i}$ are the Jacobian matrices of $f$ with respect to the bias terms $b_i$.
\end{definitionplain}

FG-completeness ensures that the sum of the attributions equals the model's output, leaving no unexplained residual. This is crucial for faithful interpretability, as it guarantees that all factors influencing the output are accounted for in the attribution scores, and no extraneous influence is attributed to the inputs.
In the following sections, we:

\begin{itemize}
    \item Establish that classical neural architectures are FG-complete, thereby explaining the historical success of gradient-based attribution on these models (\S\ref{subsec:classical}).
    \item Identify non-locally-affine layers in Transformers that break FG-completeness (\S\ref{subsec:transformer_layers}).
    \item Analyze how this causes gradient flow imbalance (\S\ref{subsec:gradient_imbalance}).
    \item Develop theoretical solutions to restore balanced gradients, introducing \emph{\FairGrad{}} (\S\ref{subsec:fairgrad_theory}).
    \item Present practical implementations of \FairGrad{} for common Transformer components (\S\ref{subsec:fairgrad_practice}).
\end{itemize}

Proofs of theorems and propositions are provided in Appendix~\ref{apn:theorems}.

\subsection{FG-Completeness of Classical Architectures}
\label{subsec:classical}

We begin by demonstrating that classical convolutional neural networks (CNNs) and multilayer perceptrons (MLPs) satisfy FG-completeness, which explains why gradient-based attribution methods are effective for these architectures. First, we introduce the concept of a locally affine function.

\begin{definitionplain}%
A function $f: \mathbb{R}^n \rightarrow \mathbb{R}^m$ is \textbf{locally affine} at a point $x_0 \in \mathbb{R}^n$ if there exists an open neighborhood $U \subset \mathbb{R}^n$ containing $x_0$, a matrix $W(x_0) \in \mathbb{R}^{m \times n}$, and a vector $b(x_0) \in \mathbb{R}^m$ such that
\[
f(x) = W(x_0) x + b(x_0), \quad \forall x \in U.
\]
\end{definitionplain}

Many activation functions used in neural networks, such as ReLU, are piecewise linear and therefore locally affine almost everywhere. Our next theorem shows that locally affine functions satisfy FG-completeness.

\ifdefined\localaffine
  \localaffine*
\else
  \begin{restatable}{theorem}{localaffine}
  \label{thm:local_affine_fg}
  Any locally affine function at $x_0$ is FG-complete in a neighborhood of $x_0$.
  \end{restatable}
\fi

Moreover, we can compose such functions and retain FG-completeness:

\ifdefined\compositionfinitefg
  \compositionfinitefg*
\else
  \begin{restatable}{theorem}{compositionfinitefg}
  \label{thm:composition_finite_fg}
  The composition of a finite number of FG-complete functions is FG-complete.
  \end{restatable}
\fi

Next, we show that FG-completeness is preserved under addition. This property is relevant for neural networks with residual connections, where the output of a layer is added to its input.
\ifdefined\additionfg
  \additionfg*
\else
  \begin{restatable}{theorem}{additionfg}
  \label{thm:addition_fg}
  Let $f_1, f_2$ be FG-complete functions. Then their sum $f = f_1 + f_2$ is FG-complete.
  \end{restatable}
\fi

We can now assert that classical neural network architectures are FG-complete:
\begin{corollary}%
\label{cor:classical_networks}
Classical neural networks employ several types of affine transformations $f(x) = Wx + b$%
:
\begin{enumerate}
    \item Linear: $W \in \mathbb{R}^{m \times n}$, $b \in \mathbb{R}^m$
    \item Convolutional: $W$ with spatial weight-sharing, $b$ broadcast per channel
    \item Pooling: AveragePool, Global-Average-Pool (special cases of Conv)
    \item BatchNorm (eval): $W = \text{diag}(\gamma/\sigma)$, $b = \beta - \mu\gamma/\sigma$
    \item LayerScale: $W = \text{diag}(\alpha)$, $b = \beta$
\end{enumerate}
Combined with piecewise-linear activations (Theorem~\ref{thm:local_affine_fg}) and skip connections (%
Theorem~\ref{thm:addition_fg}%
), these networks are FG-complete on $\mathbb{R}^n \setminus S$ (%
Theorem~\ref{thm:composition_finite_fg}%
), where $S$ denotes
the union of boundaries between linear regions
\end{corollary}

\subsection{Non-Locally-Affine Layers in Transformers}
\label{subsec:transformer_layers}

Despite the FG-completeness of classical architectures, modern Transformer models introduce several non-locally-affine operations that disrupt this property:

\begin{enumerate}
    \item \textbf{Gated Activations:} Functions like GELU and SiLU (Swish)\nightCite{Shazeer2020GLUVI} involve non-linear gating mechanisms.
    \item \textbf{Attention Mechanisms:} Self-attention and cross-attention layers perform weighted averaging based on nonlinear attention scores.
    \item \textbf{Multiplicative Feature Fusions:} Operations such as self-gating (\eg{}, SwiGLU\nightCite{Shazeer2020GLUVI}, MambaOut\nightCite{yu2024mambaout}) involve element-wise multiplication of different feedforward branches.
    \item \textbf{Normalizations:} LayerNorm divides by the standard deviation, introducing a division operation.
\end{enumerate}

These operations involve multiplicative (of which division is a special case) interactions and non-linear transformations that break the linearity required for FG-completeness, leading to imbalanced gradient flow and attribution failures, as we will discuss in the next section.

\subsection{Analysis of Gradient Flow Imbalance}
\label{subsec:gradient_imbalance}

We now analyze how each non-locally-affine operation affects gradient flow.
First, consider the element-wise multiplication of two FG-complete functions:

\ifdefined\naiveelemwise
  \naiveelemwise*
\else
  \begin{restatable}{proposition}{naiveelemwise}
  \label{prop:naive_elemwise}
  Let $f_1, f_2$ be FG-complete functions and let $f(x) = f_1(x) \odot f_2(x)$ be their element-wise product with Jacobians:
  \[
  J_x f = \text{diag}(f_2(x)) \cdot J_x f_1 + \text{diag}(f_1(x)) \cdot J_x f_2
  \]
  \[
  J_{b_i} f = \text{diag}(f_2(x)) \cdot J_{b_i} f_1 + \text{diag}(f_1(x)) \cdot J_{b_i} f_2
  \]
  Then $f$ is not FG-complete. Specifically:
  \[
  J_x f \cdot x + \sum_i J_{b_i} f \cdot b_i = 2f(x)
  \]
  \end{restatable}
\fi

So far, we've assumed both paths are FG-complete before multiplication. What happens when they're not? While each such case needs its own mathematical proof, multiplication tends to exacerbate any existing gradient flow imbalances rather than restore FG-completeness.
Two key examples illustrate this: division (a non-linear multiplicative operation), which we analyze next, and SiLU, which Proposition~\ref{prop:silu_not_fg} (in the Appendix) proves to lack FG-completeness.

\ifdefined\divisionfgzero
  \divisionfgzero*
\else
  \begin{restatable}{proposition}{divisionfgzero}
  \label{prop:division_fg_zero}
  Let $f_1, f_2$ be FG-complete functions with $f_2$ non-zero. FullGrad vanishes to exactly zero on their element-wise quotient $f(x) = f_1(x) \oslash f_2(x)$.
  \end{restatable}
\fi

Proposition~\ref{prop:division_fg_zero} demanded FG-completeness of both terms---a condition LayerNorm's denominator fails to satisfy.
Nevertheless, as we show next, this does \textit{not} spare LayerNorm from vanishing FullGrad attributions.

\ifdefined\layernormfg
  \layernormfg*
\else
  \begin{restatable}{proposition}{layernormfg}
  \label{thm:layernorm_fg}
  For the LayerNorm operation without affine parameters:
  \[
  \text{LN}(x)_i = \frac{x_i - \mu}{\sqrt{\sigma^2 + \varepsilon}},
  \]
  where $\mu = \frac{1}{N} \sum_{k=1}^N x_k$ and $\sigma^2 = \frac{1}{N} \sum_{k=1}^N (x_k - \mu)^2$,
  FullGrad approaches zero as $\varepsilon$ approaches zero:
  \[
  \lim_{\varepsilon \to 0} J_x \text{LN} \cdot x = 0.
  \]
  \end{restatable}
\fi

\subsection{\FairGrad{}: Theoretical Foundations}
\label{subsec:fairgrad_theory}

We now develop theoretical solutions to restore balanced gradient flow.

\ifdefined\elemwisefg
  \elemwisefg*
\else
  \begin{restatable}{theorem}{elemwisefg}
  \label{thm:elemwise_fg}
  Let $f_1, f_2$ be FG-complete functions. Then their element-wise product $f(x) = f_1(x) \odot f_2(x)$ is FG-complete when its Jacobians are defined with scaling coefficients $a, b \in \mathbb{R}$ where $a + b = 1$:
  \[
  J_x f = a[\text{diag}(f_2(x)) \cdot J_x f_1] + b[\text{diag}(f_1(x)) \cdot J_x f_2]
  \]
  \[
  J_{b_i} f = a[\text{diag}(f_2(x)) \cdot J_{b_i} f_1] + b[\text{diag}(f_1(x)) \cdot J_{b_i} f_2]
  \]
  \end{restatable}
\fi

\ifdefined\elemwisepruning
  \elemwisepruning*
\else
  \begin{restatable}{theorem}{elemwisepruning}
  \label{thm:elemwise_pruning}
  Let $f_1, f_2$ be arbitrary functions (not necessarily FG-complete), and let $f(x) = f_1(x) \odot f_2(x)$ be their element-wise product. Consider $f$ with scaled Jacobians as defined in Theorem~\ref{thm:elemwise_fg}. Then:
  \begin{enumerate}
    \item When $a = 0$, yielding $f(x) = [f_1(x)]_{\text{cst.}} \odot f_2(x)$ where $[\cdot]_{\text{cst.}}$ is the constant operator that zeroes gradients, $f$ is FG-complete if $f_2$ is FG-complete.
    \item By symmetry, when $b = 0$, $f$ is FG-complete if $f_1$ is FG-complete.
  \end{enumerate}
  \end{restatable}
\fi

When handling multiplicative interactions, we face a choice: ideally, we can scale gradients if both paths are FG-complete (Theorem~\ref{thm:elemwise_fg}), preserving information from both paths, or—when one path lacks FG-completeness—we can prune paths to restore FG-completeness by relying on just one FG-complete path (Theorem~\ref{thm:elemwise_pruning}).

\ifdefined\divisionfgcomplete
  \divisionfgcomplete*
\else
  \begin{restatable}{corollary}{divisionfgcomplete}
  \label{cor:division_fg_complete}
  Division can be made FG-complete by treating it as element-wise multiplication with a gradient-pruned non-linear reciprocal: $f(x) = f_1(x) \odot [1/f_2(x)]_{\text{cst.}}$
  which satisfies FG-completeness, by Theorem~\ref{thm:elemwise_pruning}.
  \end{restatable}
\fi

For division operations like those in LayerNorm, Corollary~\ref{cor:division_fg_complete} shows how treating the denominator as constant in the backward pass restores proper gradient flow.

These theoretical results suggest a general principle: balanced gradient flow can be achieved through strategic pruning and scaling of backward paths, without modifying the forward computation. Such pruning and scaling can be achieved using the following two gradient manipulation operators:
\nightParagraph{Constant Operator}
\label{def:const_op}
The constant operator $[\cdot]_{\text{cst.}}: \mathbb{R}^m \to \mathbb{R}^m$ satisfies:
\[
    [y]_{\text{cst.}} = y, \quad J_x [y]_{\text{cst.}} = 0
\]
\nightParagraph{SwapBackward}
\label{def:swap_backward}
The $\text{SwapBackward}: (f,g) \mapsto h$ operator, where $f,g,h: \mathbb{R}^n \rightarrow \mathbb{R}^m$, is defined by:
\[
    h(x) = f(x), \quad J_x h = J_x g
\]

Further theoretical insights about these operators, their computational complexity (unchanged compared to standard gradients, Table~\ref{tab:complexity}), and practical PyTorch implementations are available in Appendix~\ref{apn:grad_manip}.
\begin{table}[!t]
\centering
\begin{tabular}{l>{\arraybackslash}p{2.5cm}>{\arraybackslash}p{2.5cm}}
\toprule
\textbf{Method} & \textbf{Computation} & \textbf{Memory} \\
\cmidrule(r){1-1}
\cmidrule(l){2-3}
\IxG{} & ~~~~$\mathcal{O}(\text{1})$ & $\mathcal{O}(\sqrt{\text{Layers}})$ \\
Integrated Gradients & ~~~~$\mathcal{O}(\text{Steps})$ & $\mathcal{O}(\sqrt{\text{Layers}})$ \\
DecompX & ~~~~$\mathcal{O}(\text{Tokens})$ & $\mathcal{O}(\text{Tokens})$ \\
FullGrad+ & ~~~~$\mathcal{O}(\text{1})$ & $\mathcal{O}(\sqrt{\text{Layers}})$ \\
\textbf{Libra FullGrad+} & ~~~~$\mathcal{O}(\text{1})$ & $\mathcal{O}(\sqrt{\text{Layers}})$ \\
\bottomrule
\end{tabular}
\caption{Computational and memory complexities of attribution methods relative to one forward pass \cite{ chen2016trainingdeepnetssublinear,Srinivas2019FullGradientRF,pmlr-v70-sundararajan17a,pmlr-v235-achtibat24a,modarressi-2023-decompx}.}
\label{tab:complexity}
\end{table}

\subsection{\FairGrad{}: Practical Implementation}
\label{subsec:fairgrad_practice}

\paragraph{\FairGradPrefix{} Neural Operations.}
We now define FG-complete versions of common non-affine operations:

\textbf{\FairGradPrefix{} Attention:}
In attention mechanisms, we restrict gradient propagation to the value branch exclusively, rendering this operation locally affine and therefore FG-complete (Theorem~\ref{thm:local_affine_fg}):
\[
\text{\FairGradPrefix{}-Attention}(Q, K, V) = [\text{softmax}(QK^T)]_{\text{cst.}} \cdot V
\]

\textbf{\FairGradPrefix{} Gated Activation:} For gated activations like GELU and SiLU, we discard the non-linear gate's gradient:
\[
\text{\FairGradPrefix{}-GatedActivation}(x) = x \odot [\text{NonLinearGate}(x)]_{\text{cst.}}
\]

\textbf{\FairGradPrefix{} Self-Gating:} In self-gating operations like SwiGLU, the input flows through dual parallel feedforward paths ($f_1$, $f_2$) and reunifies via element-wise multiplication. To balance the gradient flow between branches, we scale each branch's gradient by $\frac{1}{2}$:
\[
\text{\FairGradPrefix{}-SelfGate}(x) = \text{SwapBackward}(f_1 \odot f_2, \tfrac{1}{2}(f_1 \odot f_2))(x)
\]

\textbf{\FairGradPrefix{} LayerNorm:} Using Corollary~\ref{cor:division_fg_complete} and the linearity of expectation ($\mu = \xpct{}[x]$):
\[
\text{\FairGradPrefix{}-LayerNorm}(x) = \frac{x - \mu}{[\sqrt{\sigma^2 + \varepsilon}]_{\text{cst.}}}
\]

\begin{corollary}%
\label{cor:fair_transformer}
A Transformer architecture attains FG-completeness when all non-linear components—specifically its attention mechanisms, activation functions, self-gating operations, and LayerNorms—are replaced with their \FairGradPrefix{} counterparts.
\end{corollary}

\paragraph{Universal Improvement.} While our theoretical discussion focuses on achieving FG-completeness, empirical results demonstrate that \FairGrad{}'s gradient balancing mechanism universally enhances gradient-based attribution methods. Intuitively, this is because standard gradient flow suffers from two fundamental flaws: it overemphasizes locally sensitive modules and assigns counterproductive negative signals to denominators in operations like LayerNorm.

\section{Experiments}
\label{sec:experiments}

\setlength{\belowcaptionskip}{0pt}
  \begingroup%
  \renewenvironment{figure}[1][]%
    {\begin{figure*}[#qual/FP_zele]}%
    {\end{figure*}}%
  \begingroup%
\def\zeleFPWidth{\linewidth}
\def\labelFont{\large} %

\begin{figure}[!t]
  \centering
  \begin{minipage}[c]{0.275\zeleFPWidth}
    \includegraphics[height=\linewidth]{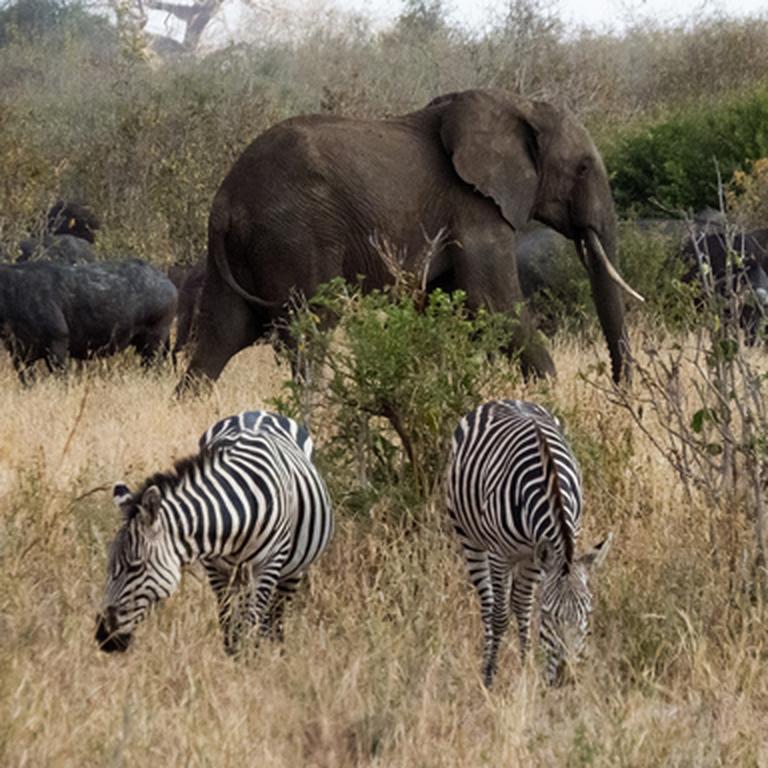}
  \end{minipage}%
  \hfill
  \begin{minipage}[c]{0.032\zeleFPWidth}
    \rotatebox{90}{\labelFont \textbf{Zebra\quad{}\quad{}\quad{}Elephant}}
  \end{minipage}%
  \begin{minipage}[c]{0.67\zeleFPWidth}
    \includegraphics[width=\linewidth]{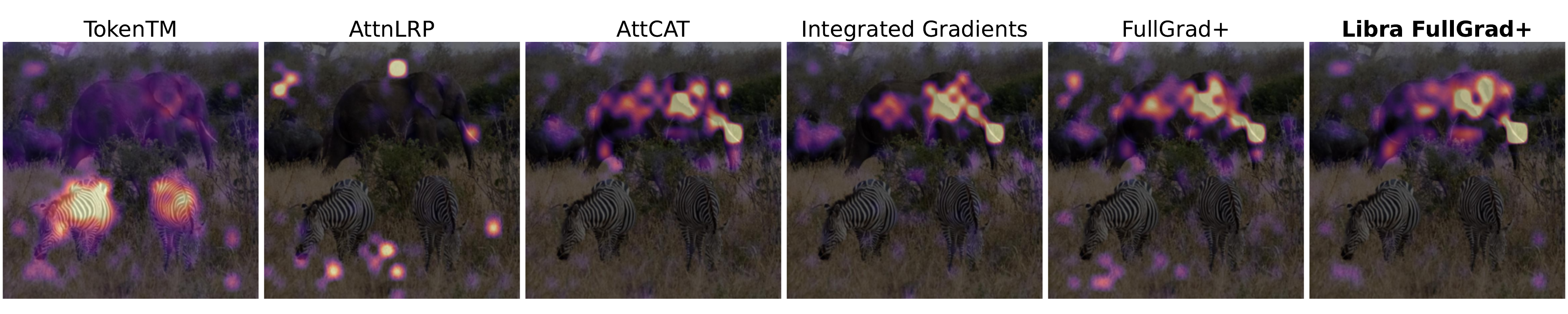}
    \\[2ex]
    \includegraphics[width=\linewidth]{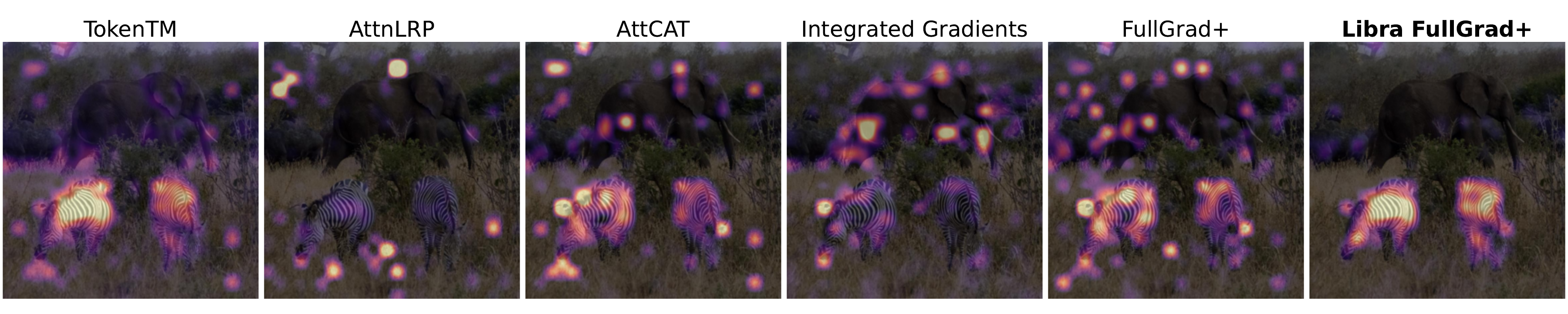}
  \end{minipage}
  \caption{%
    Cross-method comparison of class discriminativity on ViT-B.
    \Cf Fig.~\ref{qual:FP} and Appendix~\ref{apn:qual}.%
    }
  \label{qual:FP_zele}
\end{figure}
\endgroup%
  \endgroup%

\setlength{\belowcaptionskip}{10pt}
\inputTableSSWithLabel{files/ds2.tex}{main_tbl:across_ds_MIF}

We evaluate \FairGrad{} through three complementary metrics: Faithfulness, Completeness Error, and Segmentation.
For statistical validity, we report standard deviation upper bounds for all empirical results. In tables, we denote the best and second-best results in each column with bold and underline formatting, respectively.

\subsection{Experimental Setup}

Our evaluation spans two dimensions:

\begin{itemize}
    \item \textbf{Architectures:} Eight model families (ViT\nightCite{dosovitskiy-2021-image}, EVA2\nightCite{Sun2023EVACLIPIT, Fang2022EVAET, Fang2023EVA02AV}, BEiT2\nightCite{Peng2022BEiTVM, Bao2021BEiTBP}, FlexiViT\nightCite{Beyer2022FlexiViTOM}, SigLIP\footnote{SigLIP lacks a CLS token, making certain attention-based methods inapplicable.}\nightCite{Zhai2023SigmoidLF}, CLIP\nightCite{radford-2021-learning}, DeiT3\nightCite{Touvron2020TrainingDI, Touvron2022DeiTIR}, MLP-Mixer\nightCite{Tolstikhin2021MLPMixerAA}), using their largest\footnote{Huge for CLIP and DeiT3, large for others—except EVA2-S, chosen due to hardware constraints with larger EVA2 variants' input resolutions.} ImageNet-1k\nightCite{deng-2009-imagenet} finetuned variants.

    \item \textbf{Model Sizes:} All ViT variants: tiny (ViT-T), small (ViT-S), base (ViT-B), and large (ViT-L).
\end{itemize}

\nightParagraph{Faithfulness Metrics} 
We evaluate various attribution methods using faithfulness metrics, which quantify how accurately the attribution scores reflect the importance of input features in the model’s predictions.
These widely used metrics\nightCite{bluecher2024decoupling, skipplus-cvprw24, Wu2024TokenTM, modarressi-2023-decompx, ferrando-2022-measuring, chen-2020-generative, nguyen-2018-comparing} measure changes in model behavior as we progressively occlude input features in different orders. 
Here, we report the \MIFLong{} metric with predicted labels and accuracy measurement, which tracks performance degradation when occluding features by decreasing attribution importance.
Full details of this and related metrics (\LIFLongII{} and \SRGLongII{}) are provided in Appendix~\ref{apn:faithfulness}, with comprehensive results on all metrics available in Appendix~\ref{apn:quant}.

We evaluate all architectures on the ImageNet\nightCite{deng-2009-imagenet} dataset---the standard benchmark in the attribution literature\nightCite{chefer-2021-transformer, skipplus-cvprw24, Wu2024TokenTM, Xie2022ViTCXCE}.
On ViT-B, we also experiment with multiple other datasets: ImageNet-Hard\nightCite{taesiri2023zoom}%
, and following\nightCite{Covert2022LearningTE}, MURA (a medical X-ray dataset)\nightCite{Rajpurkar2017MURADT} and \OxfordPet{}\nightCite{parkhi12a}. ImageNet-Hard is a challenging dataset combining images from various existing ImageNet variants: ImageNet-V2\nightCite{Recht2019DoIC}, ImageNet-Sketch\nightCite{wang2019learning}, ImageNet-C\nightCite{hendrycks2019robustness}, ImageNet-R\nightCite{hendrycks2021many}, ImageNet-ReaL\nightCite{Beyer2020AreWD}, ImageNet-A\nightCite{hendrycks2021nae}, and ObjectNet\nightCite{Barbu2019ObjectNetAL}. 
We randomly select 1000 images from each dataset using a fixed seed.

\nightParagraph{Completeness Error}
We use Completeness Error to verify theoretical guarantees and validate implementation correctness:
\begin{equation}
\label{eq:ce}
\text{CE}(f, x, A) = \left\| f(x) - \sum_{i=1}^n A(f)(x)_i \right\|
\end{equation}
Lower CE values indicate better conservation of the model's output in the attribution scores. As this is just a sanity check, we use only 100 random images from the ImageNet dataset. See Appendix~\ref{apn:setup_CE} for further details.

\nightParagraph{Segmentation}
For segmentation, following\nightCite{skipplus-cvprw24}, we opt for ImageNet-S\nightCite{Gao2021LargeScaleUS}, which encompasses 919 distinct classes, using a random subset of 5000 images from the validation set.
Since segmentation masks provide ground truth annotations of object boundaries, they serve as an objective reference to evaluate how well feature attribution methods identify the truly relevant image regions that contribute to model predictions. See Appendix~\ref{apn:human_interp_eval} for further details.

\setlength{\belowcaptionskip}{0pt}
\inputTableStarWithLabel{files/tables_v1/across_models/sel1/seg/m_2.tex}{main_tbl:across_sel1_seg}

\subsection{Quantitative Results}
Our evaluations demonstrate that \FairGrad{} universally enhances gradient-based attribution methods across all tested models, architectures, and datasets (see Appendix~\ref{apn:quant} for comprehensive results).
Significant improvements are observed in both faithfulness and segmentation metrics (Tables~\ref{main_tbl:across_sel1_MIF} and \ref{main_tbl:across_sel1_seg}), and \FairFullgrad{} achieves optimal Completeness Error (Table~\ref{tab:across_models_completeness_error}).
These enhancements remain consistent across different model scales (Appendix~\ref{apn:across_model_sizes}) and datasets (Table~\ref{main_tbl:across_ds_MIF}, Appendix~\ref{apn:across_datasets}),
and extend to the attention-free MLP-Mixer (Appendix~\ref{per_model:MLP-Mixer-L}), validating that gradient flow imbalance, not attention mechanisms, is the core issue.

\nightParagraph{\IG{}} We also extend \IGShort{}\nightCite{pmlr-v70-sundararajan17a} and compose it with other gradient-based methods, and compare the universal improvement aspect of \FairGrad{} and \IGShort{} in Appendix~\ref{sec:ig_compositional}, showing that \FairGrad{} vastly outperforms \IGShort{}. 
Due to numerical instability, the practical approximation of \IGShort{} fails to meet its theoretical promise of completeness relative to the zero baseline (Table~\ref{tab:across_models_completeness_error}). Furthermore, we prove that the numerical instability observed is theoretically unavoidable for a fixed-step approximation (Proposition~\ref{thm:layernorm_ig} in the Appendix).

\nightParagraph{General-Purpose Methods Are Enough}
Once gradient flow is corrected, the general-purpose \FullGradPLUS{} outperforms Transformer-specific methods like GenAtt, TokenTM, and AttCAT across most metrics and models, with only a few exceptions where its performance remains competitive. This suggests that specialized architectures may not require specialized attribution methods when gradient flow is properly balanced.

\nightParagraph{Ablation Studies} Our ablation study (Table~\ref{tbl:MIF_m_ablation}) reveals three key insights: First, while gated activations theoretically break FG-completeness (Proposition~\ref{prop:silu_not_fg}), their practical impact is minimal as they often operate in saturated regimes. Second, LayerNorm's theoretically predicted vanishing attribution problem is empirically confirmed as the most significant factor. Finally, while bias terms are necessary for theoretical completeness, their practical impact is modest, suggesting that implementations can optionally omit them without severe consequences.
\subsection{Qualitative Analysis}
\label{sec:qual}
We evaluate \FairFullgradPLUS{} through two complementary scenarios: (1) text-prompted region attribution using CLIP models, demonstrating precise localization of prompted elements in complex scenes (Fig.~\ref{qual:FP}, Appendix~\ref{apn:qual_clip}), and (2) class discrimination on COCO\nightCite{Lin2014MicrosoftCC} images, showing accurate distinction between co-occurring animals (Fig.~\ref{qual:FP_zele}, Appendix~\ref{apn:qual_zele}). Both reinforce our quantitative findings that proper gradient flow enables general-purpose methods to outperform specialized approaches. Detailed protocols are in Appendix~\ref{apn:qual_setup}.

\section{Conclusion}
\label{sec:conclusion}
We introduced \FairGrad{}, correcting gradient flow imbalances via pruning and scaling backward paths.
FG-completeness, formalized here, ensures attributions decompose outputs faithfully.
We prove that while classical CNNs were naturally FG-complete (explaining their historical success with gradient-based methods), several operations in modern Transformers break this property. We provide both theoretical proofs for restoring FG-completeness and practical solutions that require no forward-pass modifications. Empirically, \FairGrad{} universally enhances gradient-based attributions across architectures, model sizes, and datasets, enabling general-purpose methods like \FullGradPLUS{} to outperform Transformer-specific approaches. This suggests that specialized architectures may not require specialized attribution methods when gradient flow is properly balanced. Our qualitative results further validate this insight.
Future work can explore compositions with other gradient-based methods, applications as a gradient regularizer,
and extensions to emerging architectural innovations.

\setlength{\belowcaptionskip}{5pt}
\begin{table*}[!th]
  \centering
  \small
  {\setlength{\tabcolsep}{5pt}
  \begingroup%
  \renewenvironment{table}[1][]%
    {\begin{center}}%
    {\end{center}}%
  \begin{table}[h]
\centering
\setlength{\tabcolsep}{5pt}
\begin{tabular}{lr@{}lr@{}lr@{}lr@{}lr@{}lr@{}lr@{}lr@{}l}
\toprule
Method & \multicolumn{2}{c}{ViT-L $\downarrow$} & \multicolumn{2}{c}{EVA2-S $\downarrow$} & \multicolumn{2}{c}{BEiT2-L $\downarrow$} & \multicolumn{2}{c}{FlexiViT-L $\downarrow$} & \multicolumn{2}{c}{SigLIP-L $\downarrow$} & \multicolumn{2}{c}{CLIP-H $\downarrow$} & \multicolumn{2}{c}{DeiT3-H $\downarrow$} & \multicolumn{2}{c}{\textit{Avg.} $\downarrow$} \\
\cmidrule(r){1-1}
\cmidrule(lr){2-15}
\cmidrule(l){16-17}
\IxG{}  & 13.6& \,\textcolor{gray}{±0.3} & 8.9& \,\textcolor{gray}{±~~0.2} & 9.0& \,\textcolor{gray}{±~~0.1} & 7.1& \,\textcolor{gray}{±0.1} & 9.3& \,\textcolor{gray}{±~~0.1} & \underline{1.3}& \,\textcolor{gray}{±0.0} & 8.6& \,\textcolor{gray}{±0.1} & 8.3& \,\textcolor{gray}{±~~0.2} \\
Integrated Gradients & \underline{8.5}& \,\textcolor{gray}{±1.5} & 4.8& \,\textcolor{gray}{±~~0.1} & \underline{6.7}& \,\textcolor{gray}{±~~0.1} & \underline{4.0}& \,\textcolor{gray}{±0.4} & \underline{5.1}& \,\textcolor{gray}{±~~0.2} & 8.2& \,\textcolor{gray}{±0.1} & 6.4& \,\textcolor{gray}{±0.5} & \underline{6.2}& \,\textcolor{gray}{±~~0.6} \\
DecompX & 11.3& \,\textcolor{gray}{±1.3} & 911.2& \,\textcolor{gray}{±33.7} & 199.2& \,\textcolor{gray}{±10.4} & 5.5& \,\textcolor{gray}{±0.5} & 242.1& \,\textcolor{gray}{±28.7} & 16.7& \,\textcolor{gray}{±0.8} & 7.7& \,\textcolor{gray}{±0.6} & 199.1& \,\textcolor{gray}{±17.2} \\
AliLRP & 29.5& \,\textcolor{gray}{±4.1} & 1233.1& \,\textcolor{gray}{±46.7} & 139.4& \,\textcolor{gray}{±~~6.2} & 7.8& \,\textcolor{gray}{±0.3} & 69.0& \,\textcolor{gray}{±~~8.8} & 15.4& \,\textcolor{gray}{±1.4} & 18.1& \,\textcolor{gray}{±0.7} & 216.1& \,\textcolor{gray}{±18.2} \\
AttnLRP & 11.0& \,\textcolor{gray}{±0.5} & \underline{2.2}& \,\textcolor{gray}{±~~0.2} & 38.2& \,\textcolor{gray}{±~~2.1} & 4.3& \,\textcolor{gray}{±0.3} & 30.4& \,\textcolor{gray}{±~~1.7} & 2.9& \,\textcolor{gray}{±0.2} & \underline{5.9}& \,\textcolor{gray}{±0.2} & 13.6& \,\textcolor{gray}{±~~1.0} \\
\midrule
FullGrad & 11.4& \,\textcolor{gray}{±0.7} & 9.5& \,\textcolor{gray}{±~~0.5} & 11.8& \,\textcolor{gray}{±~~0.5} & 19.8& \,\textcolor{gray}{±0.6} & 6.7& \,\textcolor{gray}{±~~0.4} & 7.3& \,\textcolor{gray}{±0.7} & 10.6& \,\textcolor{gray}{±0.3} & 11.0& \,\textcolor{gray}{±~~0.5} \\
\textbf{Libra FullGrad} & \textbf{0.0}& \,\textcolor{gray}{±0.0} & \textbf{0.0}& \,\textcolor{gray}{±~~0.0} & \textbf{0.0}& \,\textcolor{gray}{±~~0.0} & \textbf{0.0}& \,\textcolor{gray}{±0.0} & \textbf{0.0}& \,\textcolor{gray}{±~~0.0} & \textbf{0.0}& \,\textcolor{gray}{±0.0} & \textbf{0.0}& \,\textcolor{gray}{±0.0} & \textbf{0.0}& \,\textcolor{gray}{±~~0.0} \\
\bottomrule
\end{tabular}
\caption{Completeness Error (lower is better) across models for attribution methods. CE for \IGShort{} has been computed relative to the zero baseline. Methods without a theoretical basis for completeness (\eg, \AttRollout{}) are excluded, as their incompleteness is evident.}
\label{tab:across_models_completeness_error}
\end{table}
  \endgroup%

  }
  \begingroup%
  \renewenvironment{table}[1][]%
    {\begin{center}}%
    {\end{center}}%
  \begin{table}[h]
\centering
\setlength{\tabcolsep}{7pt}
\begin{tabular}{llllll}
\toprule
Method & \multicolumn{2}{c}{MIF Deletion (GT)} & \multicolumn{2}{c}{MIF Deletion (Predicted)} & \multicolumn{1}{c}{Segmentation} \\
\cmidrule(lr){2-3}
\cmidrule(lr){4-5}
\cmidrule(l){6-6}
  & \multicolumn{1}{c}{Accuracy} & \multicolumn{1}{c}{AOPC} & \multicolumn{1}{c}{Accuracy} & \multicolumn{1}{c}{AOPC} & \multicolumn{1}{c}{AP} \\
\midrule
\textbf{Libra FullGrad+} & \textbf{74.1} \textcolor{gray}{±0.1} & \textbf{45.5} \textcolor{gray}{±0.3} & \textbf{71.7} \textcolor{gray}{±0.1} & \textbf{50.5} \textcolor{gray}{±0.2} & \textbf{79.4} \textcolor{gray}{±0.3} \\
\quad No Att. & 68.0 \textcolor{gray}{±0.1} (\textcolor{coral}{~~-8.2\%}) & 40.8 \textcolor{gray}{±0.3} (\textcolor{coral}{-10.5\%}) & 65.2 \textcolor{gray}{±0.1} (\textcolor{coral}{~~-9.1\%}) & 45.5 \textcolor{gray}{±0.2} (\textcolor{coral}{-10.0\%}) & 72.2 \textcolor{gray}{±0.3} (\textcolor{coral}{~~-9.1\%}) \\
\quad No LN & 55.3 \textcolor{gray}{±0.1} (\textcolor{coral}{-25.3\%}) & 30.0 \textcolor{gray}{±0.3} (\textcolor{coral}{-34.2\%}) & 49.9 \textcolor{gray}{±0.1} (\textcolor{coral}{-30.4\%}) & 33.3 \textcolor{gray}{±0.2} (\textcolor{coral}{-34.1\%}) & 72.1 \textcolor{gray}{±0.3} (\textcolor{coral}{~~-9.2\%}) \\
\quad No Att. \& LN & 63.6 \textcolor{gray}{±0.1} (\textcolor{coral}{-14.1\%}) & 36.6 \textcolor{gray}{±0.2} (\textcolor{coral}{-19.7\%}) & 61.2 \textcolor{gray}{±0.1} (\textcolor{coral}{-14.7\%}) & 41.1 \textcolor{gray}{±0.2} (\textcolor{coral}{-18.6\%}) & 66.2 \textcolor{gray}{±0.3} (\textcolor{coral}{-16.7\%}) \\
\quad No Act. & \underline{74.0} \textcolor{gray}{±0.1} (\textcolor{coral}{~~-0.1\%}) & \underline{45.4} \textcolor{gray}{±0.3} (\textcolor{coral}{~~-0.3\%}) & \underline{71.6} \textcolor{gray}{±0.1} (\textcolor{coral}{~~-0.3\%}) & \underline{50.4} \textcolor{gray}{±0.2} (\textcolor{coral}{~~-0.4\%}) & \underline{79.3} \textcolor{gray}{±0.3} (\textcolor{coral}{~~-0.2\%}) \\
\quad No Gate & 69.8 \textcolor{gray}{±0.1} (\textcolor{coral}{~~-5.7\%}) & 41.9 \textcolor{gray}{±0.4} (\textcolor{coral}{~~-8.0\%}) & 67.0 \textcolor{gray}{±0.1} (\textcolor{coral}{~~-6.6\%}) & 46.7 \textcolor{gray}{±0.3} (\textcolor{coral}{~~-7.5\%}) & 71.1 \textcolor{gray}{±0.3} (\textcolor{coral}{-10.5\%}) \\
\quad No Bias & 73.9 \textcolor{gray}{±0.1} (\textcolor{coral}{~~-0.2\%}) & 45.3 \textcolor{gray}{±0.3} (\textcolor{coral}{~~-0.4\%}) & \underline{71.5} \textcolor{gray}{±0.1} (\textcolor{coral}{~~-0.3\%}) & 50.3 \textcolor{gray}{±0.2} (\textcolor{coral}{~~-0.4\%}) & 79.2 \textcolor{gray}{±0.3} (\textcolor{coral}{~~-0.3\%}) \\
Normal FullGrad+ & 50.9 \textcolor{gray}{±0.1} (\textcolor{coral}{-31.3\%}) & 25.7 \textcolor{gray}{±0.2} (\textcolor{coral}{-43.5\%}) & 48.0 \textcolor{gray}{±0.1} (\textcolor{coral}{-33.0\%}) & 30.0 \textcolor{gray}{±0.2} (\textcolor{coral}{-40.7\%}) & 51.5 \textcolor{gray}{±0.3} (\textcolor{coral}{-35.1\%}) \\
\bottomrule
\end{tabular}
\caption{Ablation study on the EVA2-S model showing the impact of removing individual components from LibraGrad. Abbreviations used: Att. (Attention), LN (LayerNorm), Act. (Gated Activation Functions), Gate (SwiGLU Self-Gating).}
\label{tbl:MIF_m_ablation}
\end{table}
  \endgroup%

  {\setlength{\tabcolsep}{8pt}
  \begingroup
    \let\oldlabel\label
    \renewcommand{\label}[1]{\oldlabel{main_tbl:across_sel1_MIF}}
  \begingroup%
  \renewenvironment{table}[1][]%
    {\begin{center}}%
    {\end{center}}%
  \begin{table}[h]
\centering
\begin{tabular}{lr@{}lr@{}lr@{}lr@{}lr@{}lr@{}lr@{}lr@{}l}
\toprule
Method & \multicolumn{2}{c}{ViT-L} & \multicolumn{2}{c}{EVA2-S} & \multicolumn{2}{c}{BEiT2-L} & \multicolumn{2}{c}{FlexiViT-L} & \multicolumn{2}{c}{SigLIP-L} & \multicolumn{2}{c}{CLIP-H} & \multicolumn{2}{c}{DeiT3-H} & \multicolumn{2}{c}{\textit{Avg.}} \\
\cmidrule(r){1-1}
\cmidrule(lr){2-15}
\cmidrule(l){16-17}
Random & 29.5& \,\textcolor{gray}{±0.1} & 21.2& \,\textcolor{gray}{±0.1} & 18.3& \,\textcolor{gray}{±0.1} & 19.2& \,\textcolor{gray}{±0.1} & 32.8& \,\textcolor{gray}{±0.1} & 28.0& \,\textcolor{gray}{±0.1} & 29.0& \,\textcolor{gray}{±0.1} & 25.4& \,\textcolor{gray}{±0.1} \\
RawAtt & 39.1& \,\textcolor{gray}{±0.1} & 50.8& \,\textcolor{gray}{±0.1} & 29.5& \,\textcolor{gray}{±0.1} & 41.7& \,\textcolor{gray}{±0.1} & -& & 42.5& \,\textcolor{gray}{±0.1} & 52.0& \,\textcolor{gray}{±0.1} & 42.6& \,\textcolor{gray}{±0.1} \\
Attention Rollout & 31.4& \,\textcolor{gray}{±0.1} & 41.1& \,\textcolor{gray}{±0.1} & 19.7& \,\textcolor{gray}{±0.1} & 23.2& \,\textcolor{gray}{±0.1} & -& & 41.3& \,\textcolor{gray}{±0.1} & 31.2& \,\textcolor{gray}{±0.1} & 31.3& \,\textcolor{gray}{±0.1} \\
AliLRP & 33.2& \,\textcolor{gray}{±0.1} & 48.0& \,\textcolor{gray}{±0.1} & 26.2& \,\textcolor{gray}{±0.1} & 24.9& \,\textcolor{gray}{±0.1} & 55.4& \,\textcolor{gray}{±0.1} & 34.4& \,\textcolor{gray}{±0.1} & 56.3& \,\textcolor{gray}{±0.1} & 39.8& \,\textcolor{gray}{±0.1} \\
AttnLRP & 41.8& \,\textcolor{gray}{±0.1} & 63.5& \,\textcolor{gray}{±0.1} & 37.7& \,\textcolor{gray}{±0.1} & 21.8& \,\textcolor{gray}{±0.1} & 62.2& \,\textcolor{gray}{±0.1} & 46.7& \,\textcolor{gray}{±0.1} & 40.7& \,\textcolor{gray}{±0.1} & 44.9& \,\textcolor{gray}{±0.1} \\
DecompX & 38.9& \,\textcolor{gray}{±0.1} & 46.8& \,\textcolor{gray}{±0.1} & 31.7& \,\textcolor{gray}{±0.1} & 35.5& \,\textcolor{gray}{±0.1} & 51.1& \,\textcolor{gray}{±0.1} & 42.4& \,\textcolor{gray}{±0.1} & 47.2& \,\textcolor{gray}{±0.1} & 42.0& \,\textcolor{gray}{±0.1} \\
Integrated Gradients & 35.9& \,\textcolor{gray}{±0.1} & 34.8& \,\textcolor{gray}{±0.1} & 23.2& \,\textcolor{gray}{±0.1} & 22.3& \,\textcolor{gray}{±0.1} & 44.0& \,\textcolor{gray}{±0.1} & 31.0& \,\textcolor{gray}{±0.1} & 33.2& \,\textcolor{gray}{±0.1} & 32.1& \,\textcolor{gray}{±0.1} \\
\midrule
\IxG{}  & 33.9& \,\textcolor{gray}{±0.1} & 32.3& \,\textcolor{gray}{±0.1} & 21.8& \,\textcolor{gray}{±0.1} & 19.9& \,\textcolor{gray}{±0.1} & 40.8& \,\textcolor{gray}{±0.1} & 31.4& \,\textcolor{gray}{±0.1} & 35.1& \,\textcolor{gray}{±0.1} & 30.7& \,\textcolor{gray}{±0.1} \\
\textbf{Libra \IxG{} } & 40.5& \,\textcolor{gray}{±0.1} & 64.1& \,\textcolor{gray}{±0.1} & 33.0& \,\textcolor{gray}{±0.1} & 36.4& \,\textcolor{gray}{±0.1} & 51.1& \,\textcolor{gray}{±0.1} & 43.1& \,\textcolor{gray}{±0.1} & 47.7& \,\textcolor{gray}{±0.1} & 45.1& \,\textcolor{gray}{±0.1} \\
\midrule
AttCAT & 44.8& \,\textcolor{gray}{±0.1} & 54.1& \,\textcolor{gray}{±0.1} & 33.9& \,\textcolor{gray}{±0.1} & 41.9& \,\textcolor{gray}{±0.1} & 45.9& \,\textcolor{gray}{±0.1} & 39.0& \,\textcolor{gray}{±0.1} & 44.0& \,\textcolor{gray}{±0.1} & 43.4& \,\textcolor{gray}{±0.1} \\
\textbf{Libra AttCAT} & \underline{61.3}& \,\textcolor{gray}{±0.1} & \underline{69.5}& \,\textcolor{gray}{±0.1} & \underline{48.9}& \,\textcolor{gray}{±0.1} & \underline{58.4}& \,\textcolor{gray}{±0.1} & \textbf{77.4}& \,\textcolor{gray}{±0.1} & \underline{58.5}& \,\textcolor{gray}{±0.1} & \underline{70.5}& \,\textcolor{gray}{±0.1} & \underline{63.5}& \,\textcolor{gray}{±0.1} \\
\midrule
GenAtt & 51.8& \,\textcolor{gray}{±0.1} & 40.7& \,\textcolor{gray}{±0.1} & 30.8& \,\textcolor{gray}{±0.1} & 53.0& \,\textcolor{gray}{±0.1} & -& & 51.0& \,\textcolor{gray}{±0.1} & 64.6& \,\textcolor{gray}{±0.1} & 48.7& \,\textcolor{gray}{±0.1} \\
\textbf{Libra GenAtt} & 55.4& \,\textcolor{gray}{±0.1} & 42.1& \,\textcolor{gray}{±0.1} & 32.9& \,\textcolor{gray}{±0.1} & 54.1& \,\textcolor{gray}{±0.1} & -& & 58.1& \,\textcolor{gray}{±0.1} & 66.5& \,\textcolor{gray}{±0.1} & 51.5& \,\textcolor{gray}{±0.1} \\
\midrule
TokenTM & 50.0& \,\textcolor{gray}{±0.1} & 44.7& \,\textcolor{gray}{±0.1} & 39.6& \,\textcolor{gray}{±0.1} & 49.3& \,\textcolor{gray}{±0.1} & -& & 51.9& \,\textcolor{gray}{±0.1} & 63.3& \,\textcolor{gray}{±0.1} & 49.8& \,\textcolor{gray}{±0.1} \\
\textbf{Libra TokenTM} & 52.5& \,\textcolor{gray}{±0.1} & 46.0& \,\textcolor{gray}{±0.1} & 38.3& \,\textcolor{gray}{±0.1} & 51.0& \,\textcolor{gray}{±0.1} & -& & 57.4& \,\textcolor{gray}{±0.1} & 65.2& \,\textcolor{gray}{±0.1} & 51.7& \,\textcolor{gray}{±0.1} \\
\midrule
GradCAM+ & 48.6& \,\textcolor{gray}{±0.1} & 47.1& \,\textcolor{gray}{±0.1} & 33.4& \,\textcolor{gray}{±0.1} & 28.7& \,\textcolor{gray}{±0.1} & 43.5& \,\textcolor{gray}{±0.1} & 33.0& \,\textcolor{gray}{±0.1} & 44.5& \,\textcolor{gray}{±0.1} & 39.8& \,\textcolor{gray}{±0.1} \\
\textbf{Libra GradCAM+} & 56.5& \,\textcolor{gray}{±0.1} & 67.0& \,\textcolor{gray}{±0.1} & 37.5& \,\textcolor{gray}{±0.1} & 33.7& \,\textcolor{gray}{±0.1} & 47.4& \,\textcolor{gray}{±0.1} & 36.2& \,\textcolor{gray}{±0.1} & 48.7& \,\textcolor{gray}{±0.1} & 46.7& \,\textcolor{gray}{±0.1} \\
\midrule
HiResCAM & 25.7& \,\textcolor{gray}{±0.1} & 59.1& \,\textcolor{gray}{±0.1} & 35.8& \,\textcolor{gray}{±0.1} & 23.8& \,\textcolor{gray}{±0.1} & 31.4& \,\textcolor{gray}{±0.1} & 37.6& \,\textcolor{gray}{±0.1} & 25.8& \,\textcolor{gray}{±0.1} & 34.2& \,\textcolor{gray}{±0.1} \\
\textbf{Libra HiResCAM} & 49.0& \,\textcolor{gray}{±0.1} & 62.6& \,\textcolor{gray}{±0.1} & 37.2& \,\textcolor{gray}{±0.1} & 56.5& \,\textcolor{gray}{±0.1} & 46.1& \,\textcolor{gray}{±0.1} & 48.9& \,\textcolor{gray}{±0.1} & 53.8& \,\textcolor{gray}{±0.1} & 50.6& \,\textcolor{gray}{±0.1} \\
\midrule
XGradCAM+ & 45.9& \,\textcolor{gray}{±0.1} & 50.2& \,\textcolor{gray}{±0.1} & 30.6& \,\textcolor{gray}{±0.1} & 26.6& \,\textcolor{gray}{±0.1} & 51.4& \,\textcolor{gray}{±0.1} & 39.4& \,\textcolor{gray}{±0.1} & 45.1& \,\textcolor{gray}{±0.1} & 41.3& \,\textcolor{gray}{±0.1} \\
\textbf{Libra XGradCAM+} & 58.8& \,\textcolor{gray}{±0.1} & 69.3& \,\textcolor{gray}{±0.1} & 45.6& \,\textcolor{gray}{±0.1} & 44.3& \,\textcolor{gray}{±0.1} & 63.6& \,\textcolor{gray}{±0.1} & 57.7& \,\textcolor{gray}{±0.1} & 66.1& \,\textcolor{gray}{±0.1} & 57.9& \,\textcolor{gray}{±0.1} \\
\midrule
FullGrad+ & 45.1& \,\textcolor{gray}{±0.1} & 48.0& \,\textcolor{gray}{±0.1} & 29.0& \,\textcolor{gray}{±0.1} & 38.9& \,\textcolor{gray}{±0.1} & 43.6& \,\textcolor{gray}{±0.1} & 37.6& \,\textcolor{gray}{±0.1} & 41.9& \,\textcolor{gray}{±0.1} & 40.6& \,\textcolor{gray}{±0.1} \\
\textbf{Libra FullGrad+} & \textbf{62.4}& \,\textcolor{gray}{±0.1} & \textbf{71.7}& \,\textcolor{gray}{±0.1} & \textbf{50.0}& \,\textcolor{gray}{±0.1} & \textbf{59.1}& \,\textcolor{gray}{±0.1} & \underline{73.5}& \,\textcolor{gray}{±0.1} & \textbf{61.1}& \,\textcolor{gray}{±0.1} & \textbf{71.5}& \,\textcolor{gray}{±0.1} & \textbf{64.2}& \,\textcolor{gray}{±0.1} \\
\bottomrule
\end{tabular}
\caption{Most-Influential-First Deletion (MIF) Accuracy evaluated using predicted labels across multiple models.}
\label{across_models_sel1_MIF_1p_accuracy_m_2}
\end{table}
  \endgroup%

    \let\label\oldlabel
  \endgroup
  }
\end{table*}

\FloatBarrier{}
{
    \small
    \bibliographystyle{ieeenat_fullname}
    \bibliography{main}

\begin{thebibliography}{94}
\providecommand{\natexlab}[1]{#1}
\providecommand{\url}[1]{\texttt{#1}}
\expandafter\ifx\csname urlstyle\endcsname\relax
  \providecommand{\doi}[1]{doi: #1}\else
  \providecommand{\doi}{doi: \begingroup \urlstyle{rm}\Url}\fi

\bibitem[Abnar and Zuidema(2020)]{abnar-2020-quantifying}
Samira Abnar and Willem Zuidema.
\newblock Quantifying attention flow in transformers.
\newblock In \emph{Proceedings of the 58th Annual Meeting of the Association
  for Computational Linguistics}, pages 4190--4197, Online, 2020. Association
  for Computational Linguistics.

\bibitem[Achtibat et~al.(2024)Achtibat, Hatefi, Dreyer, Jain, Wiegand,
  Lapuschkin, and Samek]{pmlr-v235-achtibat24a}
Reduan Achtibat, Sayed Mohammad~Vakilzadeh Hatefi, Maximilian Dreyer, Aakriti
  Jain, Thomas Wiegand, Sebastian Lapuschkin, and Wojciech Samek.
\newblock {A}ttn{LRP}: Attention-aware layer-wise relevance propagation for
  transformers.
\newblock In \emph{Proceedings of the 41st International Conference on Machine
  Learning}, pages 135--168. PMLR, 2024.

\bibitem[Ali et~al.(2022)Ali, Schnake, Eberle, Montavon, M{\"u}ller, and
  Wolf]{pmlr-v162-ali22a}
Ameen Ali, Thomas Schnake, Oliver Eberle, Gr{\'e}goire Montavon, Klaus-Robert
  M{\"u}ller, and Lior Wolf.
\newblock {XAI} for transformers: Better explanations through conservative
  propagation.
\newblock In \emph{Proceedings of the 39th International Conference on Machine
  Learning}, pages 435--451. PMLR, 2022.

\bibitem[Ancona et~al.(2017)Ancona, Ceolini, {\"O}ztireli, and
  Gross]{Ancona2017TowardsBU}
Marco Ancona, Enea Ceolini, Cengiz {\"O}ztireli, and Markus~H. Gross.
\newblock Towards better understanding of gradient-based attribution methods
  for deep neural networks.
\newblock In \emph{International Conference on Learning Representations}, 2017.

\bibitem[Anders et~al.(2020)Anders, Neumann, Marinc, Samek, M{\"u}ller, and
  Lapuschkin]{Anders2020XAIFA}
Christopher~J. Anders, David Neumann, Talmaj Marinc, Wojciech Samek,
  Klaus-Robert M{\"u}ller, and Sebastian Lapuschkin.
\newblock Xai for analyzing and unlearning spurious correlations in imagenet.
\newblock 2020.

\bibitem[Bach et~al.(2015)Bach, Binder, Montavon, Klauschen, M{\"u}ller, and
  Samek]{Bach2015OnPE}
Sebastian Bach, Alexander Binder, Gr{\'e}goire Montavon, Frederick Klauschen,
  Klaus-Robert M{\"u}ller, and Wojciech Samek.
\newblock On pixel-wise explanations for non-linear classifier decisions by
  layer-wise relevance propagation.
\newblock \emph{PLoS ONE}, 10, 2015.

\bibitem[Bao et~al.(2021)Bao, Dong, and Wei]{Bao2021BEiTBP}
Hangbo Bao, Li Dong, and Furu Wei.
\newblock Beit: Bert pre-training of image transformers.
\newblock \emph{ArXiv}, abs/2106.08254, 2021.

\bibitem[Barbu et~al.(2019)Barbu, Mayo, Alverio, Luo, Wang, Gutfreund,
  Tenenbaum, and Katz]{Barbu2019ObjectNetAL}
Andrei Barbu, David Mayo, Julian Alverio, William Luo, Christopher Wang, Dan
  Gutfreund, Joshua~B. Tenenbaum, and Boris Katz.
\newblock Objectnet: A large-scale bias-controlled dataset for pushing the
  limits of object recognition models.
\newblock In \emph{Neural Information Processing Systems}, 2019.

\bibitem[Becking et~al.(2021)Becking, Dreyer, Samek, M{\"u}ller, and
  Lapuschkin]{Becking2021ECQxEQ}
Daniel Becking, Maximilian Dreyer, Wojciech Samek, Karsten M{\"u}ller, and
  Sebastian Lapuschkin.
\newblock Ecqx: Explainability-driven quantization for low-bit and sparse dnns.
\newblock \emph{ArXiv}, abs/2109.04236, 2021.

\bibitem[Beyer et~al.(2020)Beyer, H'enaff, Kolesnikov, Zhai, and van~den
  Oord]{Beyer2020AreWD}
Lucas Beyer, Olivier~J. H'enaff, Alexander Kolesnikov, Xiaohua Zhai, and
  A{\"a}ron van~den Oord.
\newblock Are we done with imagenet?
\newblock \emph{ArXiv}, abs/2006.07159, 2020.

\bibitem[Beyer et~al.(2022)Beyer, Izmailov, Kolesnikov, Caron, Kornblith, Zhai,
  Minderer, Tschannen, Alabdulmohsin, and Pavetic]{Beyer2022FlexiViTOM}
Lucas Beyer, Pavel Izmailov, Alexander Kolesnikov, Mathilde Caron, Simon
  Kornblith, Xiaohua Zhai, Matthias Minderer, Michael Tschannen, Ibrahim~M.
  Alabdulmohsin, and Filip Pavetic.
\newblock Flexivit: One model for all patch sizes.
\newblock \emph{2023 IEEE/CVF Conference on Computer Vision and Pattern
  Recognition (CVPR)}, pages 14496--14506, 2022.

\bibitem[Binder et~al.(2016)Binder, Bach, Montavon, M{\"u}ller, and
  Samek]{Binder2016LayerWiseRP}
Alexander Binder, Sebastian Bach, Gr{\'e}goire Montavon, Klaus-Robert
  M{\"u}ller, and Wojciech Samek.
\newblock Layer-wise relevance propagation for deep neural network
  architectures.
\newblock 2016.

\bibitem[Blücher et~al.(2024)Blücher, Vielhaben, and
  Strodthoff]{bluecher2024decoupling}
Stefan Blücher, Johanna Vielhaben, and Nils Strodthoff.
\newblock Decoupling pixel flipping and occlusion strategy for consistent xai
  benchmarks.
\newblock \emph{Transactions on Machine Learning Research}, 2024.

\bibitem[Brunner et~al.(2020)Brunner, Liu, Pascual, Richter, Ciaramita, and
  Wattenhofer]{brunner-2020-identifiability}
Gino Brunner, Yang Liu, Damian Pascual, Oliver Richter, Massimiliano Ciaramita,
  and Roger Wattenhofer.
\newblock On identifiability in transformers.
\newblock In \emph{International Conference on Learning Representations}, 2020.

\bibitem[Caron et~al.(2021)Caron, Touvron, Misra, J'egou, Mairal, Bojanowski,
  and Joulin]{Caron2021EmergingPI}
Mathilde Caron, Hugo Touvron, Ishan Misra, Herv'e J'egou, Julien Mairal, Piotr
  Bojanowski, and Armand Joulin.
\newblock Emerging properties in self-supervised vision transformers.
\newblock \emph{2021 IEEE/CVF International Conference on Computer Vision
  (ICCV)}, pages 9630--9640, 2021.

\bibitem[Chefer et~al.(2021{\natexlab{a}})Chefer, Gur, and
  Wolf]{chefer-2021-generic}
Hila Chefer, Shir Gur, and Lior Wolf.
\newblock Generic attention-model explainability for interpreting bi-modal and
  encoder-decoder transformers.
\newblock In \emph{Proceedings of the IEEE/CVF International Conference on
  Computer Vision (ICCV)}, pages 397--406, 2021{\natexlab{a}}.

\bibitem[Chefer et~al.(2021{\natexlab{b}})Chefer, Gur, and
  Wolf]{chefer-2021-transformer}
Hila Chefer, Shir Gur, and Lior Wolf.
\newblock Transformer interpretability beyond attention visualization.
\newblock In \emph{Proceedings of the IEEE/CVF Conference on Computer Vision
  and Pattern Recognition (CVPR)}, pages 782--791, 2021{\natexlab{b}}.

\bibitem[Chefer et~al.(2022)Chefer, Schwartz, and Wolf]{Chefer2022OptimizingRM}
Hila Chefer, Idan Schwartz, and Lior Wolf.
\newblock Optimizing relevance maps of vision transformers improves robustness.
\newblock \emph{ArXiv}, abs/2206.01161, 2022.

\bibitem[Chefer et~al.(2023)Chefer, Alaluf, Vinker, Wolf, and
  Cohen-Or]{Chefer2023AttendandExciteAS}
Hila Chefer, Yuval Alaluf, Yael Vinker, Lior Wolf, and Daniel Cohen-Or.
\newblock Attend-and-excite: Attention-based semantic guidance for
  text-to-image diffusion models.
\newblock \emph{ArXiv}, abs/2301.13826, 2023.

\bibitem[Chen et~al.(2020)Chen, Radford, Child, Wu, Jun, Luan, and
  Sutskever]{chen-2020-generative}
Mark Chen, Alec Radford, Rewon Child, Jeffrey Wu, Heewoo Jun, David Luan, and
  Ilya Sutskever.
\newblock Generative pretraining from pixels.
\newblock In \emph{Proceedings of the 37th International Conference on Machine
  Learning}, pages 1691--1703. PMLR, 2020.

\bibitem[Chen et~al.(2016)Chen, Xu, Zhang, and
  Guestrin]{chen2016trainingdeepnetssublinear}
Tianqi Chen, Bing Xu, Chiyuan Zhang, and Carlos Guestrin.
\newblock Training deep nets with sublinear memory cost, 2016.

\bibitem[Covert et~al.(2022)Covert, Kim, and Lee]{Covert2022LearningTE}
Ian Covert, Chanwoo Kim, and Su-In Lee.
\newblock Learning to estimate shapley values with vision transformers.
\newblock \emph{ArXiv}, abs/2206.05282, 2022.

\bibitem[Deb et~al.(2023)Deb, Deiseroth, Weinbach, Schramowski, and
  Kersting]{deb-2023-atman}
Mayukh Deb, Bj{\"{o}}rn Deiseroth, Samuel Weinbach, Patrick Schramowski, and
  Kristian Kersting.
\newblock Atman: Understanding transformer predictions through memory efficient
  attention manipulation.
\newblock \emph{CoRR}, abs/2301.08110, 2023.

\bibitem[Deng et~al.(2009)Deng, Dong, Socher, Li, Li, and
  Fei-Fei]{deng-2009-imagenet}
Jia Deng, Wei Dong, Richard Socher, Li-Jia Li, Kai Li, and Li Fei-Fei.
\newblock Imagenet: A large-scale hierarchical image database.
\newblock In \emph{2009 IEEE Conference on Computer Vision and Pattern
  Recognition}, pages 248--255, 2009.

\bibitem[Dosovitskiy et~al.(2021)Dosovitskiy, Beyer, Kolesnikov, Weissenborn,
  Zhai, Unterthiner, Dehghani, Minderer, Heigold, Gelly, Uszkoreit, and
  Houlsby]{dosovitskiy-2021-image}
Alexey Dosovitskiy, Lucas Beyer, Alexander Kolesnikov, Dirk Weissenborn,
  Xiaohua Zhai, Thomas Unterthiner, Mostafa Dehghani, Matthias Minderer, Georg
  Heigold, Sylvain Gelly, Jakob Uszkoreit, and Neil Houlsby.
\newblock An image is worth 16x16 words: Transformers for image recognition at
  scale.
\newblock In \emph{International Conference on Learning Representations}, 2021.

\bibitem[Draelos and Carin(2020)]{Draelos2020UseHI}
Rachel~Lea Draelos and Lawrence Carin.
\newblock Use hirescam instead of grad-cam for faithful explanations of
  convolutional neural networks.
\newblock 2020.

\bibitem[Ede et~al.(2022)Ede, Baghdadlian, Weber, Nguyen, Zanca, Samek, and
  Lapuschkin]{Ede2022ExplainTN}
Sami Ede, Serop Baghdadlian, Leander Weber, An~Thai Nguyen, Dario Zanca,
  Wojciech Samek, and Sebastian Lapuschkin.
\newblock Explain to not forget: Defending against catastrophic forgetting with
  xai.
\newblock In \emph{International Cross-Domain Conference on Machine Learning
  and Knowledge Extraction}, 2022.

\bibitem[Fang et~al.(2022)Fang, Wang, Xie, Sun, Wu, Wang, Huang, Wang, and
  Cao]{Fang2022EVAET}
Yuxin Fang, Wen Wang, Binhui Xie, Quan-Sen Sun, Ledell~Yu Wu, Xinggang Wang,
  Tiejun Huang, Xinlong Wang, and Yue Cao.
\newblock Eva: Exploring the limits of masked visual representation learning at
  scale.
\newblock \emph{2023 IEEE/CVF Conference on Computer Vision and Pattern
  Recognition (CVPR)}, pages 19358--19369, 2022.

\bibitem[Fang et~al.(2023)Fang, Sun, Wang, Huang, Wang, and
  Cao]{Fang2023EVA02AV}
Yuxin Fang, Quan Sun, Xinggang Wang, Tiejun Huang, Xinlong Wang, and Yue Cao.
\newblock Eva-02: A visual representation for neon genesis.
\newblock \emph{ArXiv}, abs/2303.11331, 2023.

\bibitem[Fayyaz et~al.(2021)Fayyaz, Koohpayegani, Jafari, Sengupta, Joze,
  Sommerlade, Pirsiavash, and Gall]{Fayyaz2021AdaptiveTS}
Mohsen Fayyaz, Soroush~Abbasi Koohpayegani, Farnoush~Rezaei Jafari, Sunando
  Sengupta, Hamid Reza~Vaezi Joze, Eric Sommerlade, Hamed Pirsiavash, and
  Juergen Gall.
\newblock Adaptive token sampling for efficient vision transformers.
\newblock In \emph{European Conference on Computer Vision}, 2021.

\bibitem[Fel et~al.(2022)Fel, Picard, B{\'e}thune, Boissin, Vigouroux, Colin,
  Cadene, and Serre]{Fel2022CRAFTCR}
Thomas Fel, Agustin Picard, Louis B{\'e}thune, Thibaut Boissin, David
  Vigouroux, Julien Colin, R'emi Cadene, and Thomas Serre.
\newblock Craft: Concept recursive activation factorization for explainability.
\newblock \emph{2023 IEEE/CVF Conference on Computer Vision and Pattern
  Recognition (CVPR)}, pages 2711--2721, 2022.

\bibitem[Ferrando et~al.(2022)Ferrando, G{\'a}llego, and
  Costa-juss{\`a}]{ferrando-2022-measuring}
Javier Ferrando, Gerard~I. G{\'a}llego, and Marta~R. Costa-juss{\`a}.
\newblock Measuring the mixing of contextual information in the transformer.
\newblock In \emph{Proceedings of the 2022 Conference on Empirical Methods in
  Natural Language Processing}, pages 8698--8714, Abu Dhabi, United Arab
  Emirates, 2022. Association for Computational Linguistics.

\bibitem[Fu et~al.(2020)Fu, Hu, Dong, Guo, Gao, and Li]{Fu2020AxiombasedGT}
Ruigang Fu, Qingyong Hu, Xiaohu Dong, Yulan Guo, Yinghui Gao, and Biao Li.
\newblock Axiom-based grad-cam: Towards accurate visualization and explanation
  of cnns.
\newblock \emph{ArXiv}, abs/2008.02312, 2020.

\bibitem[Gao et~al.(2021)Gao, Li, Yang, Cheng, Han, and
  Torr]{Gao2021LargeScaleUS}
Shangqi Gao, Zhong-Yu Li, Ming-Hsuan Yang, Mingg-Ming Cheng, Junwei Han, and
  Philip H.~S. Torr.
\newblock Large-scale unsupervised semantic segmentation.
\newblock \emph{IEEE Transactions on Pattern Analysis and Machine
  Intelligence}, 45:\penalty0 7457--7476, 2021.

\bibitem[Hao et~al.(2020)Hao, Dong, Wei, and Xu]{Hao2020SelfAttentionAI}
Yaru Hao, Li Dong, Furu Wei, and Ke Xu.
\newblock Self-attention attribution: Interpreting information interactions
  inside transformer.
\newblock In \emph{AAAI Conference on Artificial Intelligence}, 2020.

\bibitem[Hendrycks and Dietterich(2019)]{hendrycks2019robustness}
Dan Hendrycks and Thomas Dietterich.
\newblock Benchmarking neural network robustness to common corruptions and
  perturbations.
\newblock \emph{Proceedings of the International Conference on Learning
  Representations}, 2019.

\bibitem[Hendrycks et~al.(2021{\natexlab{a}})Hendrycks, Basart, Mu, Kadavath,
  Wang, Dorundo, Desai, Zhu, Parajuli, Guo, Song, Steinhardt, and
  Gilmer]{hendrycks2021many}
Dan Hendrycks, Steven Basart, Norman Mu, Saurav Kadavath, Frank Wang, Evan
  Dorundo, Rahul Desai, Tyler Zhu, Samyak Parajuli, Mike Guo, Dawn Song, Jacob
  Steinhardt, and Justin Gilmer.
\newblock The many faces of robustness: A critical analysis of
  out-of-distribution generalization.
\newblock \emph{ICCV}, 2021{\natexlab{a}}.

\bibitem[Hendrycks et~al.(2021{\natexlab{b}})Hendrycks, Zhao, Basart,
  Steinhardt, and Song]{hendrycks2021nae}
Dan Hendrycks, Kevin Zhao, Steven Basart, Jacob Steinhardt, and Dawn Song.
\newblock Natural adversarial examples.
\newblock \emph{CVPR}, 2021{\natexlab{b}}.

\bibitem[Huang and Kong(2022)]{Huang2022TransferableAA}
Yi Huang and Adams Wai-Kin Kong.
\newblock Transferable adversarial attack based on integrated gradients.
\newblock \emph{ArXiv}, abs/2205.13152, 2022.

\bibitem[Iwana et~al.(2019)Iwana, Kuroki, and Uchida]{Iwana2019ExplainingCN}
Brian~Kenji Iwana, Ryohei Kuroki, and Seiichi Uchida.
\newblock Explaining convolutional neural networks using softmax gradient
  layer-wise relevance propagation.
\newblock \emph{2019 IEEE/CVF International Conference on Computer Vision
  Workshop (ICCVW)}, pages 4176--4185, 2019.

\bibitem[Jiang et~al.(2021)Jiang, Zhang, Hou, Cheng, and
  Wei]{Jiang2021LayerCAMEH}
Peng-Tao Jiang, Chang-Bin Zhang, Qibin Hou, Ming-Ming Cheng, and Yunchao Wei.
\newblock Layercam: Exploring hierarchical class activation maps for
  localization.
\newblock \emph{IEEE Transactions on Image Processing}, 30:\penalty0
  5875--5888, 2021.

\bibitem[Kim et~al.(2023)Kim, Lee, Kim, Ha, and Zhu]{Kim2023DenseTG}
Yunji Kim, Jiyoung Lee, Jin-Hwa Kim, Jung-Woo Ha, and Jun-Yan Zhu.
\newblock Dense text-to-image generation with attention modulation.
\newblock \emph{ArXiv}, abs/2308.12964, 2023.

\bibitem[Kindermans et~al.(2016)Kindermans, Sch{\"{u}}tt, M{\"{u}}ller, and
  D{\"{a}}hne]{kindermans-2016-investigating}
Pieter{-}Jan Kindermans, Kristof Sch{\"{u}}tt, Klaus{-}Robert M{\"{u}}ller, and
  Sven D{\"{a}}hne.
\newblock Investigating the influence of noise and distractors on the
  interpretation of neural networks.
\newblock \emph{CoRR}, abs/1611.07270, 2016.

\bibitem[Kobayashi et~al.(2020)Kobayashi, Kuribayashi, Yokoi, and
  Inui]{kobayashi-2020-attention}
Goro Kobayashi, Tatsuki Kuribayashi, Sho Yokoi, and Kentaro Inui.
\newblock Attention is not only a weight: Analyzing transformers with vector
  norms.
\newblock In \emph{Proceedings of the 2020 Conference on Empirical Methods in
  Natural Language Processing (EMNLP)}, pages 7057--7075, Online, 2020.
  Association for Computational Linguistics.

\bibitem[Kobayashi et~al.(2021)Kobayashi, Kuribayashi, Yokoi, and
  Inui]{kobayashi-2021-incorporating}
Goro Kobayashi, Tatsuki Kuribayashi, Sho Yokoi, and Kentaro Inui.
\newblock {I}ncorporating {R}esidual and {N}ormalization {L}ayers into
  {A}nalysis of {M}asked {L}anguage {M}odels.
\newblock In \emph{Proceedings of the 2021 Conference on Empirical Methods in
  Natural Language Processing}, pages 4547--4568, Online and Punta Cana,
  Dominican Republic, 2021. Association for Computational Linguistics.

\bibitem[Lin et~al.(2014)Lin, Maire, Belongie, Hays, Perona, Ramanan,
  Doll{\'a}r, and Zitnick]{Lin2014MicrosoftCC}
Tsung-Yi Lin, Michael Maire, Serge~J. Belongie, James Hays, Pietro Perona, Deva
  Ramanan, Piotr Doll{\'a}r, and C.~Lawrence Zitnick.
\newblock Microsoft coco: Common objects in context.
\newblock In \emph{European Conference on Computer Vision}, 2014.

\bibitem[LYU et~al.(2022)LYU, Apidianaki, and Callison-Burch]{LYU2022TowardsFM}
QING LYU, Marianna Apidianaki, and Chris Callison-Burch.
\newblock Towards faithful model explanation in nlp: A survey.
\newblock \emph{ArXiv}, abs/2209.11326, 2022.

\bibitem[Madsen et~al.(2021)Madsen, Reddy, and Chandar]{Madsen2021PosthocIF}
Andreas Madsen, Siva Reddy, and A.~P.~Sarath Chandar.
\newblock Post-hoc interpretability for neural nlp: A survey.
\newblock \emph{ACM Computing Surveys}, 55:\penalty0 1 -- 42, 2021.

\bibitem[Mehri et~al.(2024)Mehri, Fayyaz, Baghshah, and
  Pilehvar]{skipplus-cvprw24}
Faridoun Mehri, Mohsen Fayyaz, Mahdieh~Soleymani Baghshah, and Mohammad~Taher
  Pilehvar.
\newblock Skip{PLUS}: Skip the first few layers to better explain vision
  transformers.
\newblock \emph{2024 IEEE/CVF Conference on Computer Vision and Pattern
  Recognition Workshops (CVPRW)}, pages 204--215, 2024.

\bibitem[Modarressi et~al.(2022{\natexlab{a}})Modarressi, Fayyaz, Yaghoobzadeh,
  and Pilehvar]{modarressi-2022-globenc}
Ali Modarressi, Mohsen Fayyaz, Yadollah Yaghoobzadeh, and Mohammad~Taher
  Pilehvar.
\newblock {G}lob{E}nc: Quantifying global token attribution by incorporating
  the whole encoder layer in transformers.
\newblock In \emph{Proceedings of the 2022 Conference of the North American
  Chapter of the Association for Computational Linguistics: Human Language
  Technologies}, pages 258--271, Seattle, United States, 2022{\natexlab{a}}.
  Association for Computational Linguistics.

\bibitem[Modarressi et~al.(2022{\natexlab{b}})Modarressi, Mohebbi, and
  Pilehvar]{Modarressi2022AdapLeRSU}
A. Modarressi, Hosein Mohebbi, and Mohammad~Taher Pilehvar.
\newblock Adapler: Speeding up inference by adaptive length reduction.
\newblock In \emph{Annual Meeting of the Association for Computational
  Linguistics}, 2022{\natexlab{b}}.

\bibitem[Modarressi et~al.(2023)Modarressi, Fayyaz, Aghazadeh, Yaghoobzadeh,
  and Pilehvar]{modarressi-2023-decompx}
Ali Modarressi, Mohsen Fayyaz, Ehsan Aghazadeh, Yadollah Yaghoobzadeh, and
  Mohammad~Taher Pilehvar.
\newblock {D}ecomp{X}: Explaining transformers decisions by propagating token
  decomposition.
\newblock In \emph{Proceedings of the 61st Annual Meeting of the Association
  for Computational Linguistics (Volume 1: Long Papers)}, pages 2649--2664,
  Toronto, Canada, 2023. Association for Computational Linguistics.

\bibitem[Montavon et~al.(2017)Montavon, Lapuschkin, Binder, Samek, and
  M\"{u}ller]{montavon-2017-explaining}
Gr\'{e}goire Montavon, Sebastian Lapuschkin, Alexander Binder, Wojciech Samek,
  and Klaus-Robert M\"{u}ller.
\newblock Explaining nonlinear classification decisions with deep taylor
  decomposition.
\newblock \emph{Pattern Recogn.}, 65\penalty0 (C):\penalty0 211–222, 2017.

\bibitem[Nguyen(2018)]{nguyen-2018-comparing}
Dong Nguyen.
\newblock Comparing automatic and human evaluation of local explanations for
  text classification.
\newblock In \emph{Proceedings of the 2018 Conference of the North {A}merican
  Chapter of the Association for Computational Linguistics: Human Language
  Technologies, Volume 1 (Long Papers)}, pages 1069--1078, New Orleans,
  Louisiana, 2018. Association for Computational Linguistics.

\bibitem[Noohdani et~al.(2024)Noohdani, Hosseini, Parast, Araghi, and
  Baghshah]{Noohdani2024DecomposeandComposeAC}
Fahimeh~Hosseini Noohdani, Parsa Hosseini, Arian~Yazdan Parast,
  Hamidreza~Yaghoubi Araghi, and Mahdieh~Soleymani Baghshah.
\newblock Decompose-and-compose: A compositional approach to mitigating
  spurious correlation.
\newblock In \emph{Proceedings of the IEEE/CVF Conference on Computer Vision
  and Pattern Recognition (CVPR)}, 2024.

\bibitem[Novello et~al.(2022)Novello, Fel, and Vigouroux]{Novello2022MakingSO}
Paul Novello, Thomas Fel, and David Vigouroux.
\newblock Making sense of dependence: Efficient black-box explanations using
  dependence measure.
\newblock \emph{ArXiv}, abs/2206.06219, 2022.

\bibitem[Paiss et~al.(2022)Paiss, Chefer, and Wolf]{paiss-2022-token}
Roni Paiss, Hila Chefer, and Lior Wolf.
\newblock No token left behind: Explainability-aided image classification
  and generation.
\newblock In \emph{Computer Vision -- ECCV 2022}, pages 334--350, Cham, 2022.
  Springer Nature Switzerland.

\bibitem[Parkhi et~al.(2012)Parkhi, Vedaldi, Zisserman, and Jawahar]{parkhi12a}
Omkar~M. Parkhi, Andrea Vedaldi, Andrew Zisserman, and C.~V. Jawahar.
\newblock Cats and dogs.
\newblock In \emph{IEEE Conference on Computer Vision and Pattern Recognition},
  2012.

\bibitem[Peng et~al.(2022)Peng, Dong, Bao, Ye, and Wei]{Peng2022BEiTVM}
Zhiliang Peng, Li Dong, Hangbo Bao, Qixiang Ye, and Furu Wei.
\newblock Beit v2: Masked image modeling with vector-quantized visual
  tokenizers.
\newblock \emph{ArXiv}, abs/2208.06366, 2022.

\bibitem[Petsiuk et~al.(2018)Petsiuk, Das, and Saenko]{Petsiuk2018RISERI}
Vitali Petsiuk, Abir Das, and Kate Saenko.
\newblock Rise: Randomized input sampling for explanation of black-box models.
\newblock \emph{ArXiv}, abs/1806.07421, 2018.

\bibitem[Qiang et~al.(2022)Qiang, Pan, Li, Li, Jang, and
  Zhu]{qiang-2022-attcat}
Yao Qiang, Deng Pan, Chengyin Li, Xin Li, Rhongho Jang, and Dongxiao Zhu.
\newblock Att{CAT}: Explaining transformers via attentive class activation
  tokens.
\newblock In \emph{Advances in Neural Information Processing Systems}, 2022.

\bibitem[Radford et~al.(2021)Radford, Kim, Hallacy, Ramesh, Goh, Agarwal,
  Sastry, Askell, Mishkin, Clark, Krueger, and
  Sutskever]{radford-2021-learning}
Alec Radford, Jong~Wook Kim, Chris Hallacy, Aditya Ramesh, Gabriel Goh,
  Sandhini Agarwal, Girish Sastry, Amanda Askell, Pamela Mishkin, Jack Clark,
  Gretchen Krueger, and Ilya Sutskever.
\newblock Learning transferable visual models from natural language
  supervision.
\newblock In \emph{Proceedings of the 38th International Conference on Machine
  Learning}, pages 8748--8763. PMLR, 2021.

\bibitem[Rajpurkar et~al.(2017)Rajpurkar, Irvin, Bagul, Ding, Duan, Mehta,
  Yang, Zhu, Laird, Ball, Langlotz, Shpanskaya, Lungren, and
  Ng]{Rajpurkar2017MURADT}
Pranav Rajpurkar, Jeremy~A. Irvin, Aarti Bagul, Daisy~Yi Ding, Tony Duan,
  Hershel Mehta, Brandon Yang, Kaylie Zhu, Dillon Laird, Robyn~L. Ball, C.
  Langlotz, Katie~S. Shpanskaya, Matthew~P. Lungren, and A. Ng.
\newblock Mura dataset: Towards radiologist-level abnormality detection in
  musculoskeletal radiographs.
\newblock \emph{ArXiv}, abs/1712.06957, 2017.

\bibitem[Recht et~al.(2019)Recht, Roelofs, Schmidt, and Shankar]{Recht2019DoIC}
Benjamin Recht, Rebecca Roelofs, Ludwig Schmidt, and Vaishaal Shankar.
\newblock Do imagenet classifiers generalize to imagenet?
\newblock In \emph{International Conference on Machine Learning}, 2019.

\bibitem[Ribeiro et~al.(2016)Ribeiro, Singh, and Guestrin]{Ribeiro2016WhySI}
Marco~Tulio Ribeiro, Sameer Singh, and Carlos Guestrin.
\newblock “why should i trust you?”: Explaining the predictions of any
  classifier.
\newblock \emph{Proceedings of the 22nd ACM SIGKDD International Conference on
  Knowledge Discovery and Data Mining}, 2016.

\bibitem[Samek et~al.(2021)Samek, Montavon, Lapuschkin, Anders, and
  M{\"u}ller]{Samek2021ExplainingDN}
Wojciech Samek, Gr{\'e}goire Montavon, Sebastian Lapuschkin, Christopher~J.
  Anders, and Klaus-Robert M{\"u}ller.
\newblock Explaining deep neural networks and beyond: A review of methods and
  applications.
\newblock \emph{Proceedings of the IEEE}, 109:\penalty0 247--278, 2021.

\bibitem[Selvaraju et~al.(2017)Selvaraju, Cogswell, Das, Vedantam, Parikh, and
  Batra]{selvaraju-2017-gradcam}
Ramprasaath~R. Selvaraju, Michael Cogswell, Abhishek Das, Ramakrishna Vedantam,
  Devi Parikh, and Dhruv Batra.
\newblock Grad-cam: Visual explanations from deep networks via gradient-based
  localization.
\newblock In \emph{2017 IEEE International Conference on Computer Vision
  (ICCV)}, pages 618--626, 2017.

\bibitem[Selvaraju et~al.(2019)Selvaraju, Lee, Shen, Jin, Batra, and
  Parikh]{Selvaraju2019TakingAH}
Ramprasaath~R. Selvaraju, Stefan Lee, Yilin Shen, Hongxia Jin, Dhruv Batra, and
  Devi Parikh.
\newblock Taking a hint: Leveraging explanations to make vision and language
  models more grounded.
\newblock \emph{2019 IEEE/CVF International Conference on Computer Vision
  (ICCV)}, pages 2591--2600, 2019.

\bibitem[Shazeer(2020)]{Shazeer2020GLUVI}
Noam~M. Shazeer.
\newblock Glu variants improve transformer.
\newblock \emph{ArXiv}, abs/2002.05202, 2020.

\bibitem[Shi et~al.(2023)Shi, Zhang, Zheng, and Wang]{Shi2023PAMIPI}
Wei Shi, Wentao Zhang, Weishi Zheng, and Ruixuan Wang.
\newblock Pami: partition input and aggregate outputs for model interpretation.
\newblock \emph{ArXiv}, abs/2302.03318, 2023.

\bibitem[Shrikumar et~al.(2016)Shrikumar, Greenside, Shcherbina, and
  Kundaje]{Shrikumar2016NotJA}
Avanti Shrikumar, Peyton Greenside, Anna Shcherbina, and Anshul Kundaje.
\newblock Not just a black box: Learning important features through propagating
  activation differences.
\newblock \emph{ArXiv}, abs/1605.01713, 2016.

\bibitem[Shrikumar et~al.(2017)Shrikumar, Greenside, and
  Kundaje]{Shrikumar2017LearningIF}
Avanti Shrikumar, Peyton Greenside, and Anshul Kundaje.
\newblock Learning important features through propagating activation
  differences.
\newblock In \emph{International Conference on Machine Learning}, 2017.

\bibitem[Simonyan et~al.(2014)Simonyan, Vedaldi, and
  Zisserman]{simonyan-2014-deep}
Karen Simonyan, Andrea Vedaldi, and Andrew Zisserman.
\newblock Deep inside convolutional networks: Visualising image classification
  models and saliency maps.
\newblock In \emph{2nd International Conference on Learning Representations,
  {ICLR} 2014, Banff, AB, Canada, April 14-16, 2014, Workshop Track
  Proceedings}, 2014.

\bibitem[Springenberg et~al.(2014)Springenberg, Dosovitskiy, Brox, and
  Riedmiller]{Springenberg2014StrivingFS}
Jost~Tobias Springenberg, Alexey Dosovitskiy, Thomas Brox, and Martin~A.
  Riedmiller.
\newblock Striving for simplicity: The all convolutional net.
\newblock \emph{CoRR}, abs/1412.6806, 2014.

\bibitem[Srinivas and Fleuret(2019)]{Srinivas2019FullGradientRF}
Suraj Srinivas and François Fleuret.
\newblock Full-gradient representation for neural network visualization.
\newblock In \emph{Neural Information Processing Systems}, 2019.

\bibitem[Sun et~al.(2023)Sun, Fang, Wu, Wang, and Cao]{Sun2023EVACLIPIT}
Quan Sun, Yuxin Fang, Ledell~Yu Wu, Xinlong Wang, and Yue Cao.
\newblock Eva-clip: Improved training techniques for clip at scale.
\newblock \emph{ArXiv}, abs/2303.15389, 2023.

\bibitem[Sundararajan et~al.(2017)Sundararajan, Taly, and
  Yan]{pmlr-v70-sundararajan17a}
Mukund Sundararajan, Ankur Taly, and Qiqi Yan.
\newblock Axiomatic attribution for deep networks.
\newblock In \emph{Proceedings of the 34th International Conference on Machine
  Learning}, pages 3319--3328. PMLR, 2017.

\bibitem[Taesiri et~al.(2023)Taesiri, Nguyen, Habchi, Bezemer, and
  Nguyen]{taesiri2023zoom}
Mohammad~Reza Taesiri, Giang Nguyen, Sarra Habchi, Cor-Paul Bezemer, and Anh
  Nguyen.
\newblock Imagenet-hard: The hardest images remaining from a study of the power
  of zoom and spatial biases in image classification.
\newblock 2023.

\bibitem[Tolstikhin et~al.(2021)Tolstikhin, Houlsby, Kolesnikov, Beyer, Zhai,
  Unterthiner, Yung, Keysers, Uszkoreit, Lucic, and
  Dosovitskiy]{Tolstikhin2021MLPMixerAA}
Ilya~O. Tolstikhin, Neil Houlsby, Alexander Kolesnikov, Lucas Beyer, Xiaohua
  Zhai, Thomas Unterthiner, Jessica Yung, Daniel Keysers, Jakob Uszkoreit,
  Mario Lucic, and Alexey Dosovitskiy.
\newblock Mlp-mixer: An all-mlp architecture for vision.
\newblock In \emph{Neural Information Processing Systems}, 2021.

\bibitem[Touvron et~al.(2020)Touvron, Cord, Douze, Massa, Sablayrolles, and
  J'egou]{Touvron2020TrainingDI}
Hugo Touvron, Matthieu Cord, Matthijs Douze, Francisco Massa, Alexandre
  Sablayrolles, and Herv'e J'egou.
\newblock Training data-efficient image transformers \& distillation through
  attention.
\newblock \emph{ArXiv}, abs/2012.12877, 2020.

\bibitem[Touvron et~al.(2022)Touvron, Cord, and J'egou]{Touvron2022DeiTIR}
Hugo Touvron, Matthieu Cord, and Herv'e J'egou.
\newblock Deit iii: Revenge of the vit.
\newblock In \emph{European Conference on Computer Vision}, 2022.

\bibitem[Vaswani et~al.(2017)Vaswani, Shazeer, Parmar, Uszkoreit, Jones, Gomez,
  Kaiser, and Polosukhin]{vaswani-2017-attention}
Ashish Vaswani, Noam Shazeer, Niki Parmar, Jakob Uszkoreit, Llion Jones,
  Aidan~N Gomez, \L~ukasz Kaiser, and Illia Polosukhin.
\newblock Attention is all you need.
\newblock In \emph{Advances in Neural Information Processing Systems}. Curran
  Associates, Inc., 2017.

\bibitem[Voita et~al.(2019)Voita, Talbot, Moiseev, Sennrich, and
  Titov]{voita-2019-analyzing}
Elena Voita, David Talbot, Fedor Moiseev, Rico Sennrich, and Ivan Titov.
\newblock Analyzing multi-head self-attention: Specialized heads do the heavy
  lifting, the rest can be pruned.
\newblock In \emph{Proceedings of the 57th Annual Meeting of the Association
  for Computational Linguistics}, pages 5797--5808, Florence, Italy, 2019.
  Association for Computational Linguistics.

\bibitem[Wang et~al.(2019{\natexlab{a}})Wang, Ge, Lipton, and
  Xing]{wang2019learning}
Haohan Wang, Songwei Ge, Zachary Lipton, and Eric~P Xing.
\newblock Learning robust global representations by penalizing local predictive
  power.
\newblock In \emph{Advances in Neural Information Processing Systems}, pages
  10506--10518, 2019{\natexlab{a}}.

\bibitem[Wang et~al.(2019{\natexlab{b}})Wang, Wang, Du, Yang, Zhang, Ding,
  Mardziel, and Hu]{Wang2019ScoreCAMSV}
Haofan Wang, Zifan Wang, Mengnan Du, Fan Yang, Zijian Zhang, Sirui Ding,
  Piotr~(Peter) Mardziel, and Xia Hu.
\newblock Score-cam: Score-weighted visual explanations for convolutional
  neural networks.
\newblock \emph{2020 IEEE/CVF Conference on Computer Vision and Pattern
  Recognition Workshops (CVPRW)}, pages 111--119, 2019{\natexlab{b}}.

\bibitem[Weber et~al.(2022)Weber, Lapuschkin, Binder, and
  Samek]{Weber2022BeyondEO}
Leander Weber, Sebastian Lapuschkin, Alexander Binder, and Wojciech Samek.
\newblock Beyond explaining: Opportunities and challenges of xai-based model
  improvement.
\newblock \emph{Inf. Fusion}, 92:\penalty0 154--176, 2022.

\bibitem[Wu et~al.(2024)Wu, Duan, Kang, Tang, and Yan]{Wu2024TokenTM}
Junyi Wu, Bin Duan, Weitai Kang, Hao Tang, and Yan Yan.
\newblock Token transformation matters: Towards faithful post-hoc explanation
  for vision transformer.
\newblock \emph{2024 IEEE/CVF Conference on Computer Vision and Pattern
  Recognition (CVPR)}, pages 10926--10935, 2024.

\bibitem[Wu et~al.(2020)Wu, Su, Chen, Zhao, King, Lyu, and
  Tai]{Wu2020BoostingTT}
Weibin Wu, Yuxin Su, Xixian Chen, Shenglin Zhao, Irwin King, Michael~R. Lyu,
  and Yu-Wing Tai.
\newblock Boosting the transferability of adversarial samples via attention.
\newblock \emph{2020 IEEE/CVF Conference on Computer Vision and Pattern
  Recognition (CVPR)}, pages 1158--1167, 2020.

\bibitem[Xie et~al.(2022)Xie, hui Li, Cao, and L.Zhang]{Xie2022ViTCXCE}
Weiyan Xie, Xiao hui Li, Caleb~Chen Cao, and Nevin L.Zhang.
\newblock Vit-cx: Causal explanation of vision transformers.
\newblock In \emph{International Joint Conference on Artificial Intelligence},
  2022.

\bibitem[Yang et~al.(2019)Yang, Chen, Hsieh, ling Wang, and
  Jordan]{Yang2019MLLOODA}
Puyudi Yang, Jianbo Chen, Cho-Jui Hsieh, Jane ling Wang, and Michael~I. Jordan.
\newblock Ml-loo: Detecting adversarial examples with feature attribution.
\newblock In \emph{AAAI Conference on Artificial Intelligence}, 2019.

\bibitem[Yu and Wang(2024)]{yu2024mambaout}
Weihao Yu and Xinchao Wang.
\newblock Mambaout: Do we really need mamba for vision?
\newblock \emph{arXiv preprint arXiv:2405.07992}, 2024.

\bibitem[Zhai et~al.(2023)Zhai, Mustafa, Kolesnikov, and
  Beyer]{Zhai2023SigmoidLF}
Xiaohua Zhai, Basil Mustafa, Alexander Kolesnikov, and Lucas Beyer.
\newblock Sigmoid loss for language image pre-training.
\newblock \emph{ArXiv}, abs/2303.15343, 2023.

\bibitem[Zhang et~al.(2016)Zhang, Lin, Brandt, Shen, and
  Sclaroff]{Zhang2016TopDownNA}
Jianming Zhang, Zhe~L. Lin, Jonathan Brandt, Xiaohui Shen, and Stan Sclaroff.
\newblock Top-down neural attention by excitation backprop.
\newblock \emph{International Journal of Computer Vision}, 126:\penalty0
  1084--1102, 2016.

\bibitem[Zhang et~al.(2022)Zhang, Wu, tse Huang, Huang, Wang, Su, and
  Lyu]{Zhang2022ImprovingAT}
Jianping Zhang, Weibin Wu, Jen tse Huang, Yizhan Huang, Wenxuan Wang, Yuxin Su,
  and Michael~R. Lyu.
\newblock Improving adversarial transferability via neuron attribution-based
  attacks.
\newblock \emph{2022 IEEE/CVF Conference on Computer Vision and Pattern
  Recognition (CVPR)}, pages 14973--14982, 2022.

\end{thebibliography}
}

\setlength{\abovecaptionskip}{5pt}
\setlength{\belowcaptionskip}{0pt}

\onecolumn
\appendix
\part{}

\setcounter{page}{1}
\begingroup
\centering
\Large
\textbf{\thetitle}\\
\vspace{0.5em}Supplementary Material \\
\vspace{1.0em}
\endgroup

\pdfbookmark[0]{Appendix}{appendix}

\parttoc %

\clearpage{}
\section{Method: Further Details}
\subsection{Gradient Manipulation Operators}
\label{apn:grad_manip}
  \begingroup%
  \def\label#sec/gradient_manip_operators{}%
  \nightParagraph{Constant Operator}
\label{def:const_op}
The constant operator $[\cdot]_{\text{cst.}}: \mathbb{R}^m \to \mathbb{R}^m$ satisfies:
\[
    [y]_{\text{cst.}} = y, \quad J_x [y]_{\text{cst.}} = 0
\]
\nightParagraph{SwapBackward}
\label{def:swap_backward}
The $\text{SwapBackward}: (f,g) \mapsto h$ operator, where $f,g,h: \mathbb{R}^n \rightarrow \mathbb{R}^m$, is defined by:
\[
    h(x) = f(x), \quad J_x h = J_x g
\]
  \endgroup%

\begin{remark}[Duality]
These operators are dual: the constant operator can be implemented via SwapBackward by scaling to zero:
\[
[y]_{\text{cst.}} \equiv \text{SwapBackward}(y, 0)
\]
while SwapBackward can be constructed from the constant operator:
\[
\text{SwapBackward}(f, g)(x) = [f(x)]_{\text{cst.}} + (g(x) - [g(x)]_{\text{cst.}})
\]
\end{remark}

\begin{remark}[PyTorch Implementation]
In PyTorch, the constant operator can be implemented using \texttt{detach}:
\[
[y]_{\text{cst.}} \equiv y.\text{detach}()
\]
For SwapBackward, we have two equivalent implementations:
\begin{enumerate}
    \item Via duality: $\text{SwapBackward}(f,g)(x) = f(x).\text{detach}() + (g(x) - g(x).\text{detach}())$
    \item Via custom backward: Define an \texttt{autograd.Function} that returns $f(x)$ in \texttt{forward} and propagates gradients as if it were $g(x)$ in \texttt{backward}
\end{enumerate}
Both implementations yield identical gradients, though the latter may be more computationally efficient, while the former may be easier to implement.
\end{remark}
\begin{remark}[Computational Efficiency]
Both core operations of \FairGrad{} preserve or improve efficiency—constant operators reduce computation through pruning, while SwapBackward maintains original complexity regardless of implementation. See Table~\ref{tab:complexity} for comparative analysis.
\end{remark}

\subsection{Theorems}
\label{apn:theorems}

\subsubsection{FullGrad-Completeness of Affine Functions}
\label{apn:affine}

\begin{definitionplain}%
A function $f : \mathbb{R}^n \rightarrow \mathbb{R}^m$ is \textbf{affine} if it can be expressed as $f(x) = Wx + b$ for some matrix $W \in \mathbb{R}^{m \times n}$ and vector $b \in \mathbb{R}^m$.
\end{definitionplain}

\ifdefined\affinecomplete
  \affinecomplete*
\else
  \begin{restatable}{theorem}{affinecomplete}
  \label{thm:affine_fg}
  Any affine function $f: \mathbb{R}^n \rightarrow \mathbb{R}^m$ is FG-complete.
  \end{restatable}
\fi

\begin{proof}
Let $f(x) = Wx + b$ be an affine function. The Jacobians are:
\[
J_x f = W, \quad J_b f = I,
\]
where $I$ is the identity matrix. By direct computation:
\[
J_x f \cdot x + J_b f \cdot b = Wx + b = f(x),
\]
proving FG-completeness.
\end{proof}

\subsubsection{FullGrad-Completeness of Locally Affine Functions}
\label{apn:local_affine}

\begin{definitionplain}%
A function $f: \mathbb{R}^n \rightarrow \mathbb{R}^m$ is \textbf{locally affine} at a point $x_0 \in \mathbb{R}^n$ if there exists an open neighborhood $U \subset \mathbb{R}^n$ containing $x_0$, a matrix $W(x_0) \in \mathbb{R}^{m \times n}$, and a vector $b(x_0) \in \mathbb{R}^m$ such that
\[
f(x) = W(x_0) x + b(x_0), \quad \forall x \in U.
\]
\end{definitionplain}

\begin{example}%
Consider the ReLU function $\text{ReLU}: \mathbb{R} \rightarrow \mathbb{R}$ defined by $\text{ReLU}(x) = \max(0, x)$. The ReLU function is locally affine at every point $x_0 \neq 0$:
\begin{itemize}
    \item For $x_0 > 0$: $\text{ReLU}(x) = x$ in a neighborhood, so $W(x_0) = 1$, $b(x_0) = 0$
    \item For $x_0 < 0$: $\text{ReLU}(x) = 0$ in a neighborhood, so $W(x_0) = 0$, $b(x_0) = 0$
\end{itemize}
\end{example}

\ifdefined\localaffine
  \localaffine*
\else
  \begin{restatable}{theorem}{localaffine}
  \label{thm:local_affine_fg}
  Any locally affine function at $x_0$ is FG-complete in a neighborhood of $x_0$.
  \end{restatable}
\fi

\begin{proof}
Let $f$ be locally affine at $x_0$. By definition, there exists an open neighborhood $U$ of $x_0$ and matrices $W(x_0)$, $b(x_0)$ such that for all $x \in U$:
\[
f(x) = W(x_0)x + b(x_0)
\]
This is an affine function in $U$, and thus by Theorem~\ref{thm:affine_fg}, it is FG-complete in $U$.
\end{proof}

\subsubsection{FullGrad-Completeness of Composition of Two Functions}
\label{apn:composition_two}

\ifdefined\compositiontwofg
  \compositiontwofg*
\else
  \begin{restatable}{theorem}{compositiontwofg}
  \label{thm:composition_two_fg}
  Let $f_1, f_2$ be FG-complete functions. Then their composition $f = f_2 \circ f_1$ is also FG-complete.
  \end{restatable}
\fi

\begin{proof}
Let $y = f_1(x)$. By FG-completeness of $f_1$ and $f_2$:
\[
f_1(x) = J_x f_1 \cdot x + \sum_i J_{b_i} f_1 \cdot b_i
\]
\[
f_2(y) = J_y f_2 \cdot y + \sum_j J_{c_j} f_2 \cdot c_j
\]
where $b_i$ and $c_j$ are bias terms in $f_1$ and $f_2$ respectively.

For the composition $f = f_2 \circ f_1$, by the chain rule:
\[
J_x f = J_y f_2 \cdot J_x f_1
\]
For bias terms $b_i$ in $f_1$:
\[
J_{b_i} f = J_y f_2 \cdot J_{b_i} f_1
\]
For bias terms $c_j$ in $f_2$:
\[
J_{c_j} f = J_{c_j} f_2
\]

Therefore:
\begin{align*}
J_x f \cdot x + \sum_i J_{b_i} f \cdot b_i + \sum_j J_{c_j} f \cdot c_j 
&= J_y f_2 \cdot J_x f_1 \cdot x + \sum_i J_y f_2 \cdot J_{b_i} f_1 \cdot b_i + \sum_j J_{c_j} f_2 \cdot c_j \\
&= J_y f_2 \cdot (J_x f_1 \cdot x + \sum_i J_{b_i} f_1 \cdot b_i) + \sum_j J_{c_j} f_2 \cdot c_j \\
&= J_y f_2 \cdot f_1(x) + \sum_j J_{c_j} f_2 \cdot c_j \\
&= J_y f_2 \cdot y + \sum_j J_{c_j} f_2 \cdot c_j \\
&= f_2(y) = f_2(f_1(x)) = f(x)
\end{align*}
proving the FG-completeness of the composition.
\end{proof}

\subsubsection{FullGrad-Completeness of Finite Function Compositions}
\label{apn:composition_finite}

\ifdefined\compositionfinitefg
  \compositionfinitefg*
\else
  \begin{restatable}{theorem}{compositionfinitefg}
  \label{thm:composition_finite_fg}
  The composition of a finite number of FG-complete functions is FG-complete.
  \end{restatable}
\fi

\begin{proof}
Let $f = f_k \circ \cdots \circ f_1$ be a composition of $k$ FG-complete functions. We prove the result by induction on $k$.

\textbf{Base case ($k=1$):} A single FG-complete function is FG-complete by definition.

\textbf{Inductive hypothesis:} Assume the composition of $n$ FG-complete functions is FG-complete.

\textbf{Inductive step:} Consider a composition of $n+1$ FG-complete functions:
\[
g = f_{n+1} \circ f_n \circ \cdots \circ f_1
\]
Let $h = f_n \circ \cdots \circ f_1$. By the inductive hypothesis, $h$ is FG-complete. Then $g = f_{n+1} \circ h$ is a composition of two FG-complete functions, which is FG-complete by Theorem~\ref{thm:composition_two_fg}.

By induction, the composition of any finite number of FG-complete functions is FG-complete.
\end{proof}

\begin{corollary}%
\label{cor:local_affine_composition}
The composition of a finite number of locally affine functions at $x_0$ is FG-complete in a neighborhood of $x_0$.
\end{corollary}

\subsubsection{FullGrad-Completeness of Function Addition}
\label{apn:addition}

\ifdefined\additionfg
  \additionfg*
\else
  \begin{restatable}{theorem}{additionfg}
  \label{thm:addition_fg}
  Let $f_1, f_2$ be FG-complete functions. Then their sum $f = f_1 + f_2$ is FG-complete.
  \end{restatable}
\fi

\begin{proof}
Since $f_1$ and $f_2$ are FG-complete, we have:
\[
f_1(x) = J_x f_1 \cdot x + \sum_i J_{b_i} f_1 \cdot b_i
\]
\[
f_2(x) = J_x f_2 \cdot x + \sum_j J_{c_j} f_2 \cdot c_j
\]
Then, for their sum $f(x) = f_1(x) + f_2(x)$, the Jacobians are:
\[
J_x f = J_x f_1 + J_x f_2
\]
\[
J_{b_i} f = J_{b_i} f_1, \quad J_{c_j} f = J_{c_j} f_2
\]
Therefore:
\begin{align*}
J_x f \cdot x + \sum_i J_{b_i} f \cdot b_i + \sum_j J_{c_j} f \cdot c_j &= (J_x f_1 + J_x f_2) \cdot x + \sum_i J_{b_i} f_1 \cdot b_i + \sum_j J_{c_j} f_2 \cdot c_j \\
&= [J_x f_1 \cdot x + \sum_i J_{b_i} f_1 \cdot b_i] + [J_x f_2 \cdot x + \sum_j J_{c_j} f_2 \cdot c_j] \\
&= f_1(x) + f_2(x) \\
&= f(x)
\end{align*}
Thus, $f$ is FG-complete.
\end{proof}

\begin{corollary}%
\label{cor:residual_fg}
Let $f$ be FG-complete. Then the residual connection defined by $g(x) = x + f(x)$ is FG-complete.
\end{corollary}

\subsubsection{Gradient Flow in Element-Wise Multiplication}
\label{apn:elemwise}

We first show that the naive approach to element-wise multiplication is not FG-complete.

\ifdefined\naiveelemwise
  \naiveelemwise*
\else
  \begin{restatable}{proposition}{naiveelemwise}
  \label{prop:naive_elemwise}
  Let $f_1, f_2$ be FG-complete functions and let $f(x) = f_1(x) \odot f_2(x)$ be their element-wise product with Jacobians:
  \[
  J_x f = \text{diag}(f_2(x)) \cdot J_x f_1 + \text{diag}(f_1(x)) \cdot J_x f_2
  \]
  \[
  J_{b_i} f = \text{diag}(f_2(x)) \cdot J_{b_i} f_1 + \text{diag}(f_1(x)) \cdot J_{b_i} f_2
  \]
  Then $f$ is not FG-complete. Specifically:
  \[
  J_x f \cdot x + \sum_i J_{b_i} f \cdot b_i = 2f(x)
  \]
  \end{restatable}
\fi

\begin{proof}
Since $f_1$ and $f_2$ are FG-complete:
\[
f_1(x) = J_x f_1 \cdot x + \sum_i J_{b_i} f_1 \cdot b_i
\]
\[
f_2(x) = J_x f_2 \cdot x + \sum_i J_{b_i} f_2 \cdot b_i
\]

Computing $J_x f \cdot x + \sum_i J_{b_i} f \cdot b_i$ with the standard Jacobians:
\begin{align*}
&[\text{diag}(f_2(x)) \cdot J_x f_1 + \text{diag}(f_1(x)) \cdot J_x f_2] \cdot x + \\
&\sum_i [\text{diag}(f_2(x)) \cdot J_{b_i} f_1 + \text{diag}(f_1(x)) \cdot J_{b_i} f_2] \cdot b_i \\
&= \text{diag}(f_2(x)) \cdot (J_x f_1 \cdot x + \sum_i J_{b_i} f_1 \cdot b_i) + \\
&\phantom{=\ } \text{diag}(f_1(x)) \cdot (J_x f_2 \cdot x + \sum_i J_{b_i} f_2 \cdot b_i) \\
&= \text{diag}(f_2(x)) \cdot f_1(x) + \text{diag}(f_1(x)) \cdot f_2(x) \\
&= f_2(x) \odot f_1(x) + f_1(x) \odot f_2(x) = 2f(x)
\end{align*}

Therefore, the naive element-wise product yields twice the desired output in the FG-completeness equation, making it not FG-complete.
\end{proof}

However, by properly scaling the Jacobian terms, we can achieve FG-completeness:

\ifdefined\elemwisefg
  \elemwisefg*
\else
  \begin{restatable}{theorem}{elemwisefg}
  \label{thm:elemwise_fg}
  Let $f_1, f_2$ be FG-complete functions. Then their element-wise product $f(x) = f_1(x) \odot f_2(x)$ is FG-complete when its Jacobians are defined with scaling coefficients $a, b \in \mathbb{R}$ where $a + b = 1$:
  \[
  J_x f = a[\text{diag}(f_2(x)) \cdot J_x f_1] + b[\text{diag}(f_1(x)) \cdot J_x f_2]
  \]
  \[
  J_{b_i} f = a[\text{diag}(f_2(x)) \cdot J_{b_i} f_1] + b[\text{diag}(f_1(x)) \cdot J_{b_i} f_2]
  \]
  \end{restatable}
\fi

\begin{proof}
The proof follows the same structure as Proposition~\ref{prop:naive_elemwise}, but with scaled Jacobians:
\begin{align*}
&[a\text{diag}(f_2(x)) \cdot J_x f_1 + b\text{diag}(f_1(x)) \cdot J_x f_2] \cdot x + \\
&\sum_i [a\text{diag}(f_2(x)) \cdot J_{b_i} f_1 + b\text{diag}(f_1(x)) \cdot J_{b_i} f_2] \cdot b_i \\
&= a\text{diag}(f_2(x)) \cdot (J_x f_1 \cdot x + \sum_i J_{b_i} f_1 \cdot b_i) + \\
&\phantom{=\ } b\text{diag}(f_1(x)) \cdot (J_x f_2 \cdot x + \sum_i J_{b_i} f_2 \cdot b_i) \\
&= a\text{diag}(f_2(x)) \cdot f_1(x) + b\text{diag}(f_1(x)) \cdot f_2(x) \\
&= (a+b)(f_1(x) \odot f_2(x)) = f_1(x) \odot f_2(x) = f(x)
\end{align*}

where the last equality follows from $a + b = 1$, proving the FG-completeness of $f$ with the scaled Jacobian definitions.
\end{proof}

\ifdefined\elemwisepruning
  \elemwisepruning*
\else
  \begin{restatable}{theorem}{elemwisepruning}
  \label{thm:elemwise_pruning}
  Let $f_1, f_2$ be arbitrary functions (not necessarily FG-complete), and let $f(x) = f_1(x) \odot f_2(x)$ be their element-wise product. Consider $f$ with scaled Jacobians as defined in Theorem~\ref{thm:elemwise_fg}. Then:
  \begin{enumerate}
    \item When $a = 0$, yielding $f(x) = [f_1(x)]_{\text{cst.}} \odot f_2(x)$ where $[\cdot]_{\text{cst.}}$ is the constant operator that zeroes gradients, $f$ is FG-complete if $f_2$ is FG-complete.
    \item By symmetry, when $b = 0$, $f$ is FG-complete if $f_1$ is FG-complete.
  \end{enumerate}
  \end{restatable}
\fi

\begin{proof}
Let $a = 0$ (thus $b = 1$). If $f_2$ is FG-complete:
\begin{align*}
&[\text{diag}(f_1(x)) \cdot J_x f_2] \cdot x + \sum_i [\text{diag}(f_1(x)) \cdot J_{b_i} f_2] \cdot b_i \\
&= \text{diag}(f_1(x)) \cdot (J_x f_2 \cdot x + \sum_i J_{b_i} f_2 \cdot b_i) \\
&= \text{diag}(f_1(x)) \cdot f_2(x) \\
&= f_1(x) \odot f_2(x) = f(x)
\end{align*}
proving the FG-completeness of $f$.
\end{proof}

\subsubsection{Non-FG-Completeness of SiLU Activation}
\label{apn:silu}

\ifdefined\silunotfg
  \silunotfg*
\else
  \begin{restatable}{proposition}{silunotfg}
  \label{prop:silu_not_fg}
  The SiLU activation function $\text{SiLU}(x) = x \cdot \sigma(x)$, where $\sigma(x) = \frac{1}{1 + e^{-x}}$ is the sigmoid function, is not FG-complete. Specifically, there exists $x \in \mathbb{R}$ such that:
  \[
  J_x \text{SiLU} \cdot x \neq \text{SiLU}(x)
  \]
  \end{restatable}
\fi

\begin{proof}
The Jacobian of SiLU is:
\[
J_x \text{SiLU} = \sigma(x) + x\sigma'(x)
\]
where $\sigma'(x) = \sigma(x)(1-\sigma(x))$ is the derivative of the sigmoid function.

Therefore:
\begin{align*}
J_x \text{SiLU} \cdot x &= x\sigma(x) + x^2\sigma'(x) \\
&= x\sigma(x) + x^2\sigma(x)(1-\sigma(x)) \\
&= x\sigma(x)\left(1 + x(1-\sigma(x))\right) \\
&= \text{SiLU}(x)\left(1 + x(1-\sigma(x))\right) \\
&= \text{SiLU}(x)\left(1 + x - x\sigma(x)\right) \\
&= \text{SiLU}(x)\left(1 + x - \text{SiLU}(x)\right)
\end{align*}

For $J_x \text{SiLU} \cdot x = \text{SiLU}(x)$, we require:
\[
\text{SiLU}(x)\left(1 + x - \text{SiLU}(x)\right) = \text{SiLU}(x)
\]
Subtracting $\text{SiLU}(x)$ from both sides:
\[
\text{SiLU}(x)\left(1 + x - \text{SiLU}(x)\right) - \text{SiLU}(x) = 0
\]
Simplifying:
\[
\text{SiLU}(x)\left((1 + x - \text{SiLU}(x)) - 1\right) = 0
\]
\[
\text{SiLU}(x)\left(x - \text{SiLU}(x)\right) = 0
\]
Thus, we require either $\text{SiLU}(x) = 0$, or $x = \text{SiLU}(x)$:
\begin{itemize}
    \item $\text{SiLU}(x) = 0$, which happens when $x = 0$, or when $\sigma(x) = 0$, requiring $x \to -\infty$, leading to $\text{SiLU}(x) = x \cdot 0 = 0$.
    \item $x = \text{SiLU}(x)$, which occurs when $\sigma(x) = 1$, requiring $x \to \infty$. %
\end{itemize}

For all other values of $x$, we have $J_x \text{SiLU} \cdot x \neq \text{SiLU}(x)$. For example, at $x = 1$:
\[
\text{SiLU}(1) = 1 \cdot \sigma(1) \approx 0.731
\]
\[
J_x \text{SiLU} \cdot x = \text{SiLU}(1)\left(1 + 1 - \text{SiLU}(1)\right) \approx 0.731 \times (1 + 1 - 0.731) \approx 0.731 \times 1.269 \approx 0.928 \neq 0.731
\]
proving that SiLU is not FG-complete.

\end{proof}

\subsubsection{Gradient Flow in Division}
\label{apn:division}

\ifdefined\divisionfgzero
  \divisionfgzero*
\else
  \begin{restatable}{proposition}{divisionfgzero}
  \label{prop:division_fg_zero}
  Let $f_1, f_2$ be FG-complete functions with $f_2$ non-zero. FullGrad vanishes to exactly zero on their element-wise quotient $f(x) = f_1(x) \oslash f_2(x)$.
  \end{restatable}
\fi

\begin{proof}
Since \( f_1 \) and \( f_2 \) are FG-complete, we have:
\begin{align*}
f_1(x) &= J_x f_1 \cdot x + \sum_i J_{b_i^{(1)}} f_1 \cdot b_i^{(1)}, \\
f_2(x) &= J_x f_2 \cdot x + \sum_j J_{b_j^{(2)}} f_2 \cdot b_j^{(2)}.
\end{align*}

The Jacobian of \( f \) with respect to \( x \) is:
\begin{align*}
J_x f &= \operatorname{diag}\left( \frac{1}{f_2(x)} \right) J_x f_1 - \operatorname{diag}\left( \frac{f_1(x)}{f_2(x)^2} \right) J_x f_2, \label{eq:Jx_f}
\end{align*}
where \( \operatorname{diag}(v) \) denotes a diagonal matrix with vector \( v \) on the diagonal and the fractions denote element-wise division.

Similarly, the Jacobians with respect to the biases are:
\begin{align*}
J_{b_i^{(1)}} f &= \operatorname{diag}\left( \frac{1}{f_2(x)} \right) J_{b_i^{(1)}} f_1, \\
J_{b_j^{(2)}} f &= -\operatorname{diag}\left( \frac{f_1(x)}{f_2(x)^2} \right) J_{b_j^{(2)}} f_2.
\end{align*}

Now, compute the FullGrad attributions of $f$:
\begin{align*}
J_x f \cdot x &+ \sum_i J_{b_i^{(1)}} f \cdot b_i^{(1)} + \sum_j J_{b_j^{(2)}} f \cdot b_j^{(2)} \nonumber \\
&= \left[ \operatorname{diag}\left( \frac{1}{f_2(x)} \right) J_x f_1 - \operatorname{diag}\left( \frac{f_1(x)}{f_2(x)^2} \right) J_x f_2 \right] \cdot x \nonumber \\
&\quad + \sum_i \operatorname{diag}\left( \frac{1}{f_2(x)} \right) J_{b_i^{(1)}} f_1 \cdot b_i^{(1)} - \sum_j \operatorname{diag}\left( \frac{f_1(x)}{f_2(x)^2} \right) J_{b_j^{(2)}} f_2 \cdot b_j^{(2)} \nonumber \\
&= \operatorname{diag}\left( \frac{1}{f_2(x)} \right) \left( J_x f_1 \cdot x + \sum_i J_{b_i^{(1)}} f_1 \cdot b_i^{(1)} \right) \nonumber \\
&\quad - \operatorname{diag}\left( \frac{f_1(x)}{f_2(x)^2} \right) \left( J_x f_2 \cdot x + \sum_j J_{b_j^{(2)}} f_2 \cdot b_j^{(2)} \right) \\
&= \operatorname{diag}\left( \frac{1}{f_2(x)} \right) \left( J_x f_1 \cdot x + \sum_i J_{b_i^{(1)}} f_1 \cdot b_i^{(1)} \right) \nonumber \\
&\quad - \operatorname{diag}\left( \frac{f_1(x)}{f_2(x)^2} \right) \left( J_x f_2 \cdot x + \sum_j J_{b_j^{(2)}} f_2 \cdot b_j^{(2)} \right) \\
&= \operatorname{diag}\left( \frac{1}{f_2(x)} \right) f_1(x) - \operatorname{diag}\left( \frac{f_1(x)}{f_2(x)^{2}} \right) f_2(x) \\
&= \frac{f_1(x)}{f_2(x)} - \frac{f_1(x)}{f_2(x)} = f(x) - f(x) = 0.
\end{align*}
\end{proof}

\ifdefined\divisionfgcomplete
  \divisionfgcomplete*
\else
  \begin{restatable}{corollary}{divisionfgcomplete}
  \label{cor:division_fg_complete}
  Division can be made FG-complete by treating it as element-wise multiplication with a gradient-pruned non-linear reciprocal: $f(x) = f_1(x) \odot [1/f_2(x)]_{\text{cst.}}$
  which satisfies FG-completeness, by Theorem~\ref{thm:elemwise_pruning}.
  \end{restatable}
\fi

\subsubsection{How Does FullGrad Behave on LayerNorm?}
\label{apn:layernorm}

\ifdefined\layernormfg
  \layernormfg*
\else
  \begin{restatable}{proposition}{layernormfg}
  \label{thm:layernorm_fg}
  For the LayerNorm operation without affine parameters:
  \[
  \text{LN}(x)_i = \frac{x_i - \mu}{\sqrt{\sigma^2 + \varepsilon}},
  \]
  where $\mu = \frac{1}{N} \sum_{k=1}^N x_k$ and $\sigma^2 = \frac{1}{N} \sum_{k=1}^N (x_k - \mu)^2$,
  FullGrad approaches zero as $\varepsilon$ approaches zero:
  \[
  \lim_{\varepsilon \to 0} J_x \text{LN} \cdot x = 0.
  \]
  \end{restatable}
\fi

\begin{proof}
Let $x \in \mathbb{R}^N$. We decompose LayerNorm into two operations:
\begin{enumerate}
    \item Centering: $y = x - \mu \mathbf{1}$, where $\mathbf{1}$ is the vector of ones
    \item Scaling: $z = y/s$, where $s = \sqrt{\sigma^2 + \varepsilon}$
\end{enumerate}

The Jacobian of centering is:
\[
(J_x y)_{ij} = \delta_{ij} - \frac{1}{N}
\]
which gives $(J_x y \cdot x)_i = x_i - \mu = y_i$.

The Jacobian of scaling is:
\[
(J_y z)_{ij} = \frac{\delta_{ij}}{s} - \frac{y_i y_j}{N s^3}
\]

By the chain rule:
\[
J_x \text{LN} \cdot x = J_y z \cdot J_x y \cdot x = J_y z \cdot y
\]

Computing $(J_y z \cdot y)_i$:
\begin{align*}
(J_y z \cdot y)_i &= \sum_{j=1}^N \left(\frac{\delta_{ij}}{s} - \frac{y_i y_j}{N s^3}\right) y_j \\
&= \frac{y_i}{s} - \frac{y_i}{N s^3} \sum_{j=1}^N y_j^2 \\
&= \frac{y_i}{s} - \frac{y_i \sigma^2}{s^3} \\
&= \frac{y_i}{s} - \frac{y_i(s^2 - \varepsilon)}{s^3} \\
&= y_i \cdot \frac{\varepsilon}{s^3}
\end{align*}

Since $s = \sqrt{\sigma^2 + \varepsilon} \geq \sqrt{\sigma^2}$ for all $\varepsilon > 0$, and $y_i$ is independent of $\varepsilon$, we have for each component $i$:
\[
\lim_{\varepsilon \to 0} (J_x \text{LN} \cdot x)_i = \lim_{\varepsilon \to 0} y_i \cdot \frac{\varepsilon}{s^3} = 0,
\]
completing the proof.
\end{proof}

\subsubsection{Non-Viability of \IG{} on LayerNorm}
\begin{restatable}{proposition}{layernormig}
\label{thm:layernorm_ig}
For the LayerNorm operation without affine parameters as defined in Proposition~\ref{thm:layernorm_fg}, Integrated Gradients with a zero baseline approaches zero when approximated using an $n$-step (with $n$ fixed) Riemann summation as $\varepsilon$ approaches zero.
\end{restatable}
\begin{proof}
For any baseline $\bar{x}$, Integrated Gradients can be written as:
\[
\text{IG}(x, \bar{x}) = \int_0^1 J_x \text{LN}(\bar{x} + \alpha(x - \bar{x})) \cdot (x - \bar{x}) \, d\alpha
\]
Using an $n$-step Riemann sum approximation:
\[
\text{IG}(x, \bar{x}) \approx \frac{1}{n} \sum_{k=1}^n J_x \text{LN}(\bar{x} + \frac{k}{n}(x - \bar{x})) \cdot (x - \bar{x})
\]
Setting $\bar{x} = 0$:
\[
\text{IG}(x, 0) \approx \frac{1}{n} \sum_{k=1}^n J_x \text{LN}(\frac{k}{n}x) \cdot x
\]
From Proposition~\ref{thm:layernorm_fg}, we know that for any input $x'$:
\[
\lim_{\varepsilon \to 0} J_x \text{LN}(x') \cdot x' = 0
\]
For each step $k$ in the Riemann sum, let $x_k = \frac{k}{n}x$. We can exchange the limit with the finite sum:
\begin{align*}
\lim_{\varepsilon \to 0} \text{IG}(x, 0)
&\approx \lim_{\varepsilon \to 0} \frac{1}{n} \sum_{k=1}^n J_x \text{LN}(\frac{k}{n}x) \cdot x \\
&= \frac{1}{n} \sum_{k=1}^n \lim_{\varepsilon \to 0} J_x \text{LN}(x_k) \cdot x \\
&= \frac{1}{n} \sum_{k=1}^n \lim_{\varepsilon \to 0} \frac{k}{n} J_{x_k} \text{LN}(x_k) \cdot x \\
&= \frac{1}{n} \sum_{k=1}^n \lim_{\varepsilon \to 0} J_{x_k} \text{LN}(x_k) \cdot x_{k} \\
&= \frac{1}{n} \sum_{k=1}^n 0 \\
&= 0
\end{align*}
where we applied Proposition~\ref{thm:layernorm_fg} to $x_k$.
\end{proof}

\clearpage{}
\section{Detailed Experimental Setup}
\label{apn:setup}
\subsection{Empirical Completeness Evaluation}
\label{apn:setup_CE}
  \begingroup%
  \def\label#sec/CE_intro{}%
  Consider an attribution method $A$ that assigns relevance scores $A(f)(x)_i$ to each input feature $x_i$ relative to model $f$ (see \S\ref{sec:background} for notation). The Completeness Error (CE) is defined as:

\begin{equation}
\label{eq:ce}
\text{CE}(f, x, A) = \left\| f(x) - \sum_{i=1}^n A(f)(x)_i \right\|
\end{equation}

Lower CE values indicate better conservation of the model's output in the attribution scores.
  \endgroup%

We say $A$ is complete on a given architecture $f$ when $\text{CE} = 0$. While our theoretical analysis proves that Transformers exhibit FG-completeness under our modifications, we perform empirical validation to: (1) verify the theoretical guarantees, (2) validate implementation correctness, and (3) demonstrate how prior methods fail to achieve completeness. As this is just a sanity check, we use only 100 random images from the ImageNet dataset\nightCite{deng-2009-imagenet}, and set the attribution target to the predicted logit of the model.

\subsection{Faithfulness Metrics}
\label{apn:faithfulness}
We evaluate attribution methods through faithfulness metrics that quantify how well attribution scores reflect the true importance of input features to model predictions. These widely used metrics\nightCite{bluecher2024decoupling, skipplus-cvprw24, Wu2024TokenTM, modarressi-2023-decompx, ferrando-2022-measuring, chen-2020-generative, nguyen-2018-comparing} measure changes in model behavior as we progressively occlude input features in different orders.

For a given feature ordering $\pi$ and occlusion fraction $s/n$ (where $n$ is the total number of features), we compute the area under curve:

\begin{equation}
    \text{AUC}[\pi] = \frac{1}{n} \sum_{s=0}^{n} v^{\text{perf}}(x_{\Pi(s)})
\end{equation}

where $\Pi(s)$ represents keeping only the first $s$ features according to ordering $\pi$, and $v^{\text{perf}}(x_{\Pi(s)})$ measures model performance on this partially occluded input. This can be either classification accuracy (more robust to outliers) or the change in predicted probability for the target class (called AOPC, more granular). Both measures can use either ground truth or predicted target classes.

The \MIFLong~metric measures performance degradation when occluding features in order of decreasing attribution scores:

\begin{equation}
    \text{MIF}[\phi] = \text{AUC}[\pi^\phi]
\end{equation}

where $\pi^\phi$ orders features by decreasing attribution values. Since lower MIF scores indicate better attributions (faster performance degradation), we normalize it as:

\begin{equation}
    \text{MIF}_\text{norm}[\phi] = 100 - \text{MIF}[\phi]
\end{equation}

The \LIFLong~metric measures performance when occluding features in order of increasing attribution scores:

\begin{equation}
    \text{LIF}[\phi] = \text{AUC}[(\pi^\phi)^r]
\end{equation}

where $(\pi^\phi)^r$ is the reverse ordering. LIF can be interpreted as a counterfactual metric - features with the most negative attribution scores often contribute to competing classes, so their removal can actually increase the target class probability. Since higher LIF scores already indicate better attributions (slower degradation when removing negative contributors), it requires no normalization.

The \SRGLong~measure\nightCite{bluecher2024decoupling} is defined as the average of both metrics:

\begin{equation}
    \text{SRG}[\phi] = \frac{\text{LIF}[\phi] + \text{MIF}_\text{norm}[\phi]}{2}
\end{equation}

In this work, we primarily focus on MIF with predicted labels and accuracy measurement, as our goal is to identify positive feature contributions to model predictions rather than counterfactual explanations. We report comprehensive results using both accuracy and AOPC metrics for MIF, LIF and SRG using both ground truth and predicted labels in Appendix~\ref{apn:quant}.

\subsubsection{True Token Masking}
\label{apn:trueTokenMasking}
Instead of simply overlaying a color mask, we choose to completely exclude the masked patches from the model's input (for models that support token exclusion)\nightCite{Covert2022LearningTE, skipplus-cvprw24}. At the same time, we preserve accurate positional encodings for the unmasked patches. We term this strategy \textit{True Token Masking}. The conventional method of using the color black (or simply zeroing the tokens in text-based Transformers) for patch masking encounters several issues:
\begin{itemize}
  \item If a patch is predominantly black, painting it black does not effectively eliminate its informational content. For instance, a black drawing on a white background would remain mostly unchanged.
  \item Patches might serve computational functions, such as acting as a scratchpad for the model's internal processes. Masking these with black does not prevent the model from using them for such purposes.
  \item Introducing a black mask can create artifacts in the image, potentially leading to out-of-distribution data, which affects the model's performance.
\end{itemize}

\subsection{Human Interpretability Evaluation}
\label{apn:human_interp_eval}
Although lacking a strong theoretical justification, human interpretability evaluations serve as effective sanity checks and provide a quantitative measure that aligns with intuitive inferences drawn from qualitative examples of attribution methods. Following the zero-shot segmentation setup proposed by \cite{chefer-2021-transformer, Wu2024TokenTM, skipplus-cvprw24}, we report the Average Precision (AP) metric. This evaluation requires a dataset with ground truth labels for the target class. Notably, AP is invariant to shift and scale transformations, mirroring the properties of our faithfulness metrics.

\clearpage{}
\subsection{Qualitative Evaluation}
\label{apn:qual_setup}
Our qualitative evaluation comprises two complementary scenarios, each designed to assess different aspects of attribution quality:

\nightParagraph{Text-Prompted Attribution on CLIP}
CLIP models are trained to output similarity scores between image-text pairs, enabling flexible zero-shot queries through natural language prompts.
Our first evaluation scenario uses the text-image similarity scores output by CLIP models as attribution targets. For each test image, we systematically probe different regions and concepts using targeted text prompts, enabling a detailed assessment of each attribution method's ability to locate described elements within complex scenes.

\nightParagraph{Multi-Class Discrimination} Using ImageNet-finetuned models, we evaluate class discriminativity on carefully selected images from the COCO 2017 training set\nightCite{Lin2014MicrosoftCC}. We specifically focus on images containing both zebras and elephants within the same frame, with both animals clearly visible and not significantly occluded. Given the rarity of such co-occurrences, our evaluation encompasses all available instances. The attribution target is set to the output class probabilities of ``Zebra'' and ``African Elephant''. This choice is motivated by several factors:
\begin{itemize}
    \item Prior work\nightCite{Iwana2019ExplainingCN, skipplus-cvprw24} has established these animals as effective test cases for attribution evaluation.
    \item ImageNet has a single class for zebras and three classes for elephants, which is in contrast to most other animals that can have tens of different fine-grained ImageNet classes.
    \item They co-occur in nature.
    \item Their distinct visual characteristics help verify that attributions are truly class-specific rather than merely highlighting salient regions.
\end{itemize}

\nightParagraph{Method Selection}
We showcase three categories of attribution methods: fundamental gradient-based approaches (\IG{} and \FullGradPLUS{}), our proposed \FairFullgradPLUS{}, and contemporary Transformer-specific methods (\AttCAT{}, \AttnLRP{}, and \TokenTM{}). The latter group was selected based on strong performance on quantitative metrics. Between \TokenTM{} and \GenAtt{}, which generate nearly identical attribution maps, we employ \TokenTM{} as the more recent formulation.

\subsubsection{Qualitative Visualization Method}
\label{apn:qual_method}
To visualize attribution maps:

\begin{enumerate}
    \item \textbf{Negative Value Removal:} We first apply ReLU to remove negative attribution scores, as we focus on positive feature contributions.

    \item \textbf{Robust Scaling:} Rather than using absolute maximum values which can be sensitive to outliers, we compute the 99th percentile of the attribution scores. We then scale the values by dividing by this robust maximum.

    \item \textbf{Spatial Upsampling:} The token-level attribution map is upsampled to the original image resolution using bicubic interpolation.

    \item \textbf{Range Normalization:} Finally, we clamp values to $[0,1]$.
\end{enumerate}

\clearpage{}
\section{Qualitative Results}
\label{apn:qual}

Following the evaluation protocol in Appendix~\ref{apn:qual_setup}, we present a comprehensive qualitative analysis below.

\subsection{Text-Prompted Qualitative Examples on \EVATwoCLIPLarge{}}
\label{apn:qual_clip}
Our first evaluation scenario uses \EVATwoCLIPLarge{}'s text-image similarity scores as attribution targets. For each test image, we systematically probe different regions and concepts using targeted text prompts, enabling a detailed assessment of each attribution method's ability to locate described elements within complex scenes.

\begingroup

\CLIPRows[!b]{%
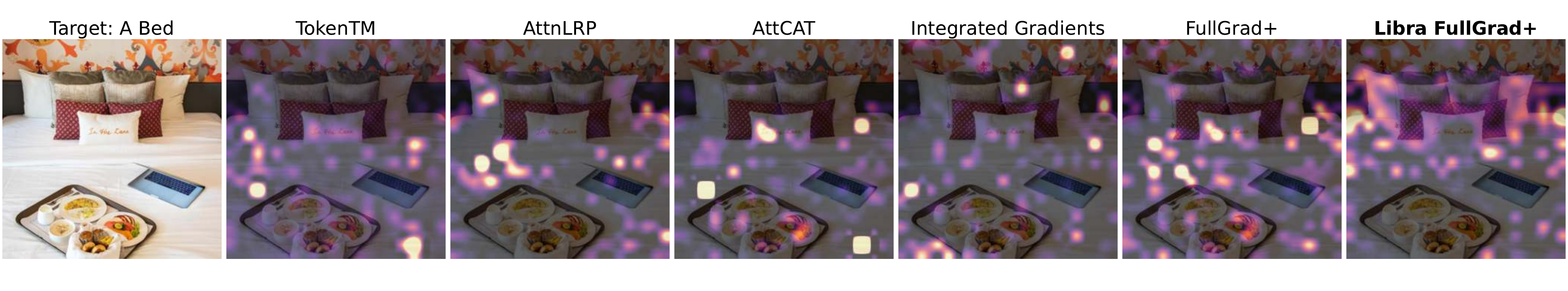,%
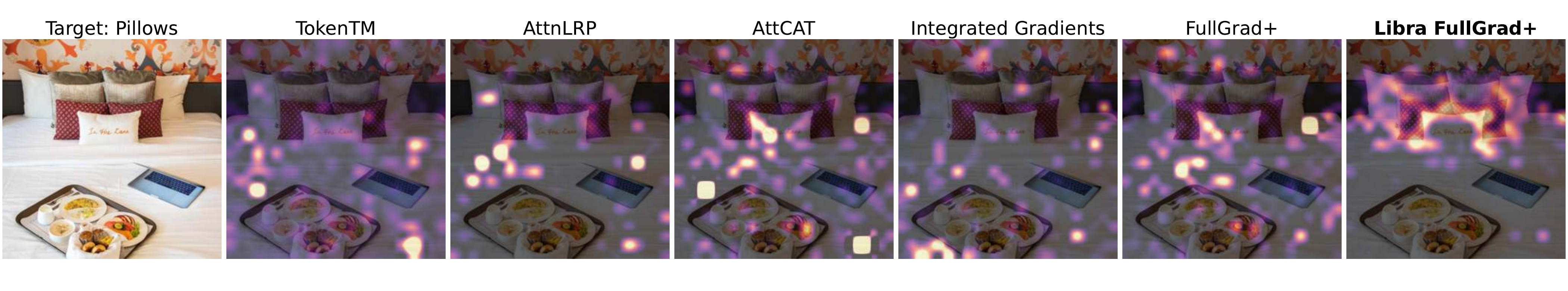,%
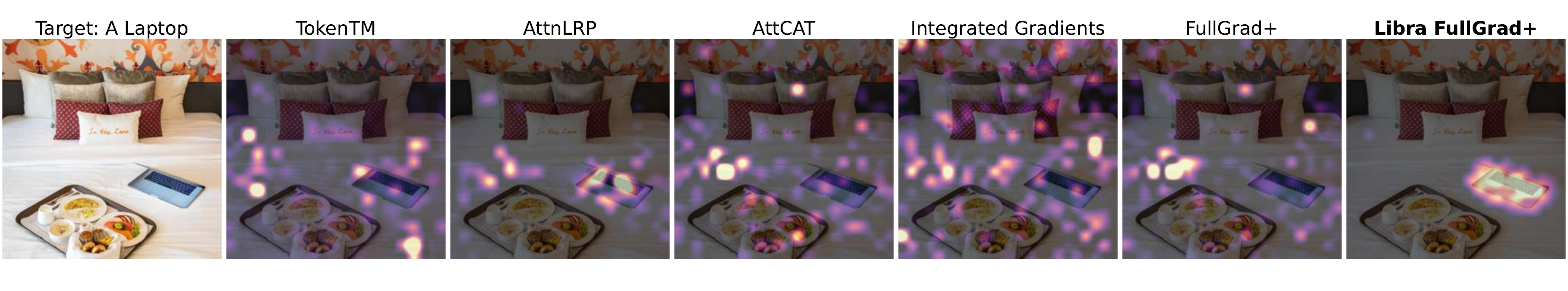,%
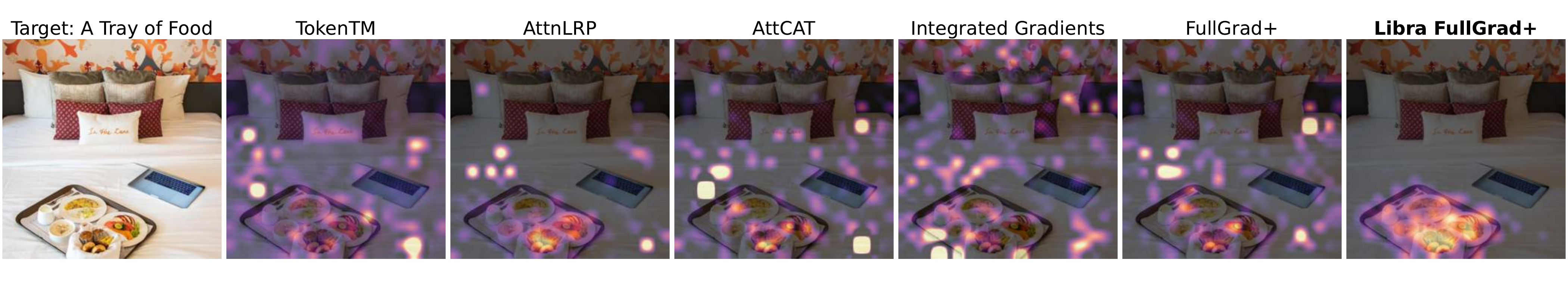,%
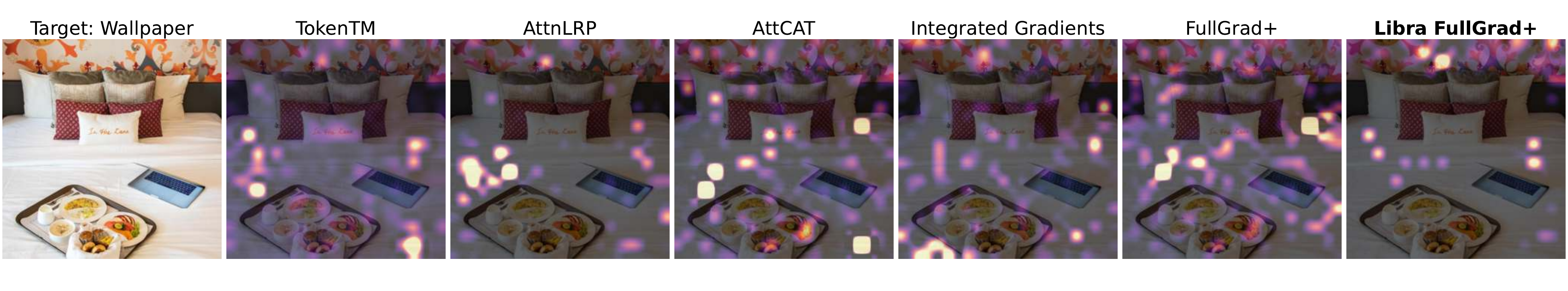,%
}{}{}

\begin{figure}[h]
    \centering
    \includegraphics[width=\linewidth]{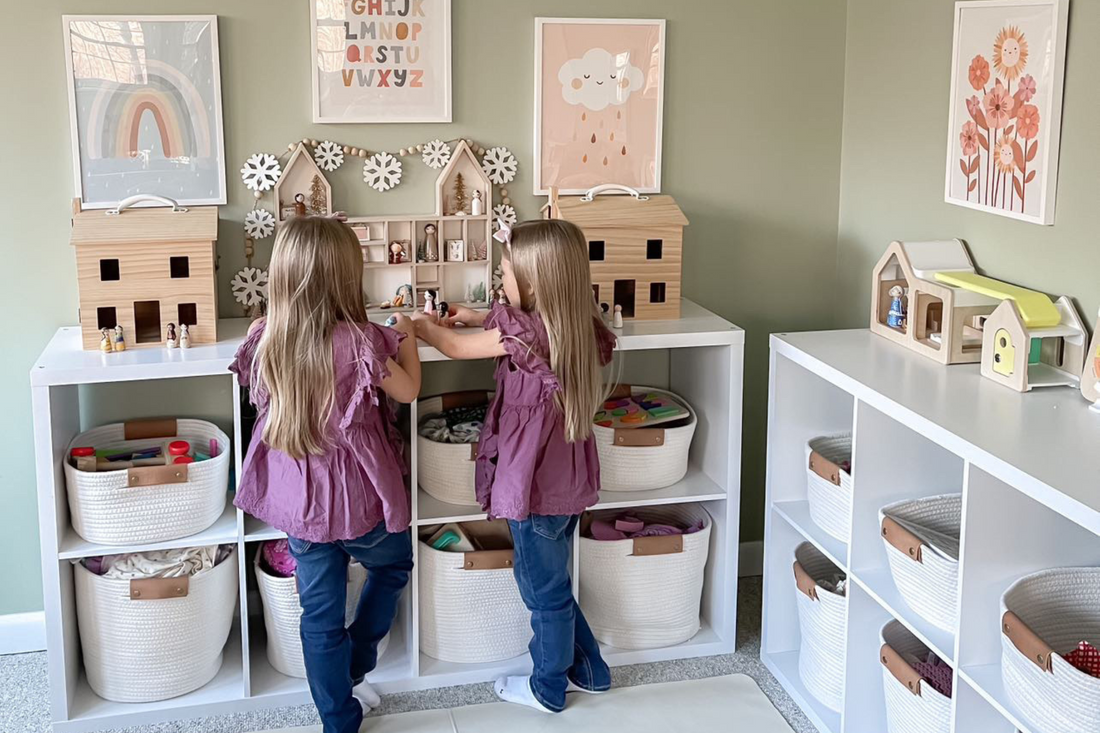}
\end{figure}
\CLIPRows[!b]{%
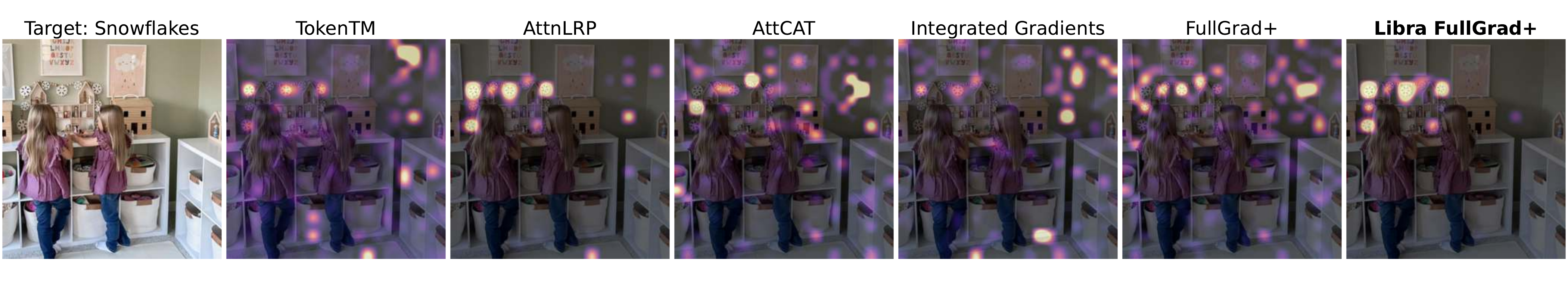,%
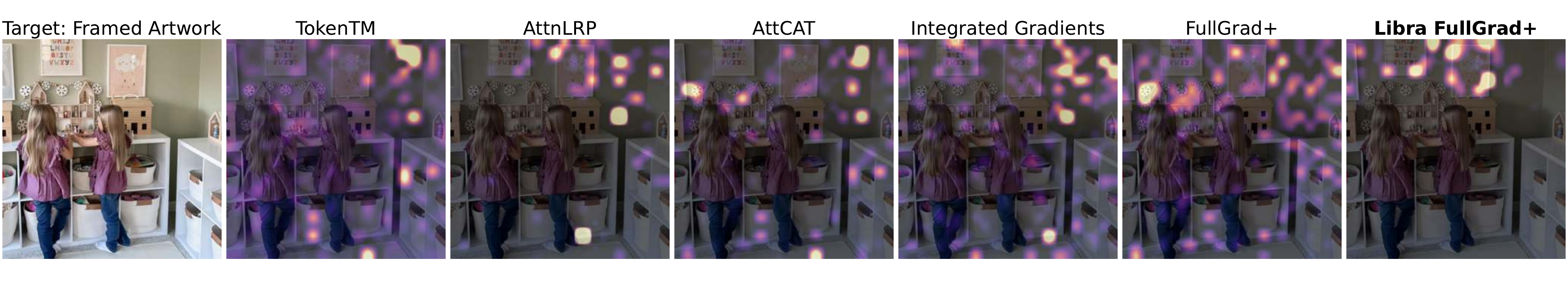,%
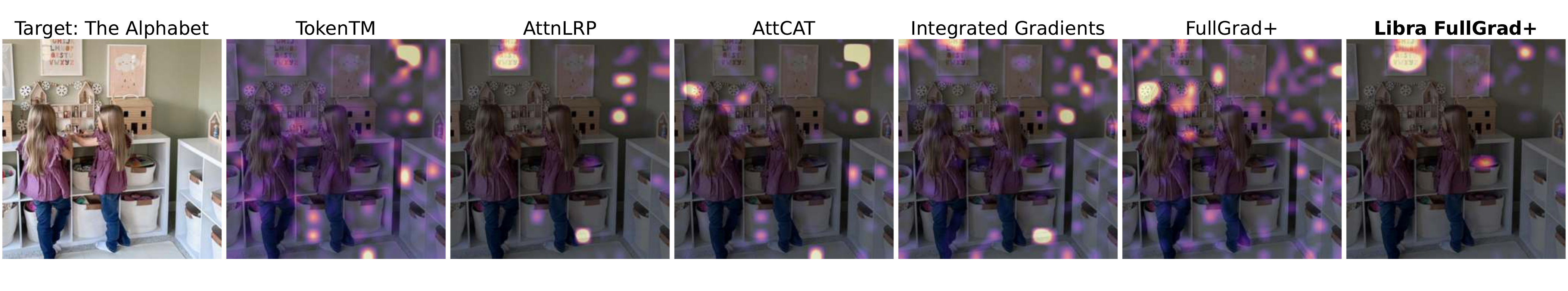,%
}{}{}
\CLIPRows[!b]{%
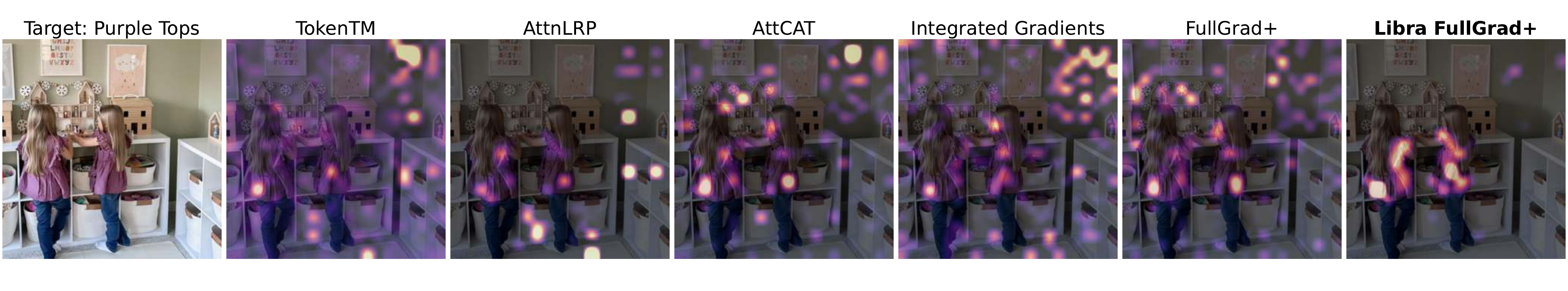,%
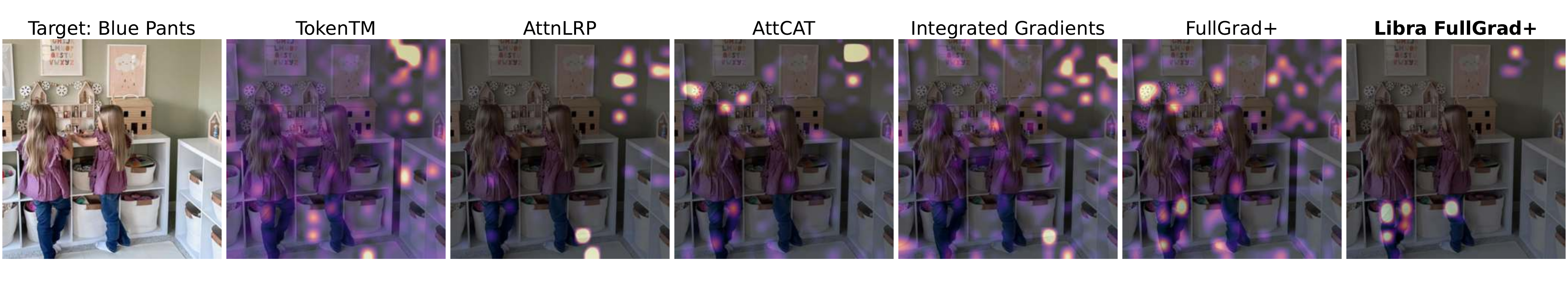,%
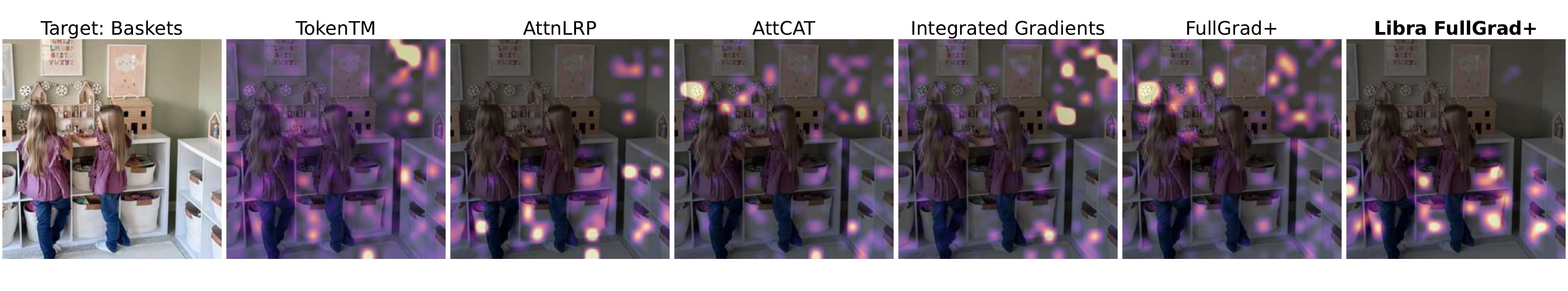,%
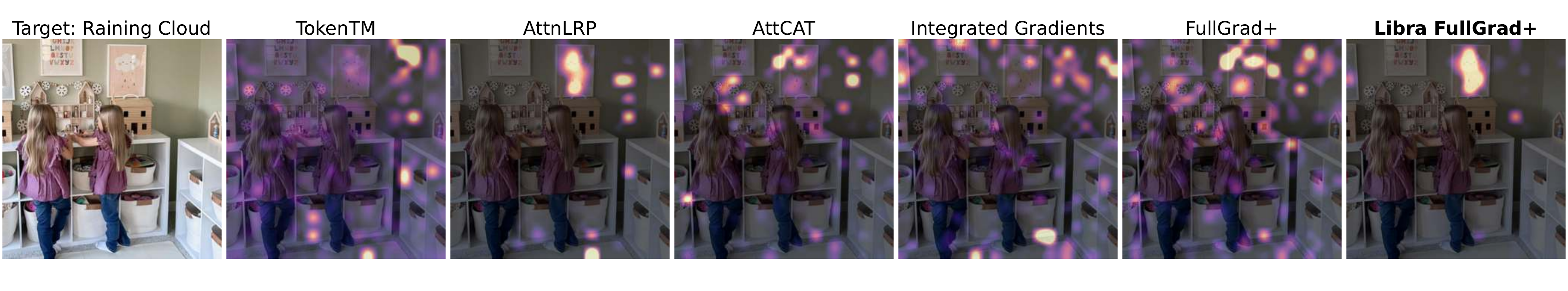,%
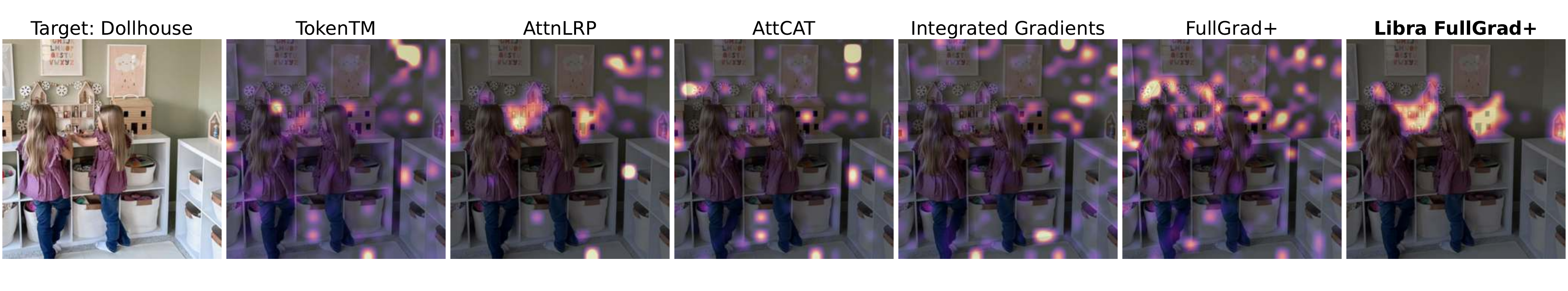,%
}{}{}

\begin{figure}[h]
    \centering
    \includegraphics[width=\linewidth]{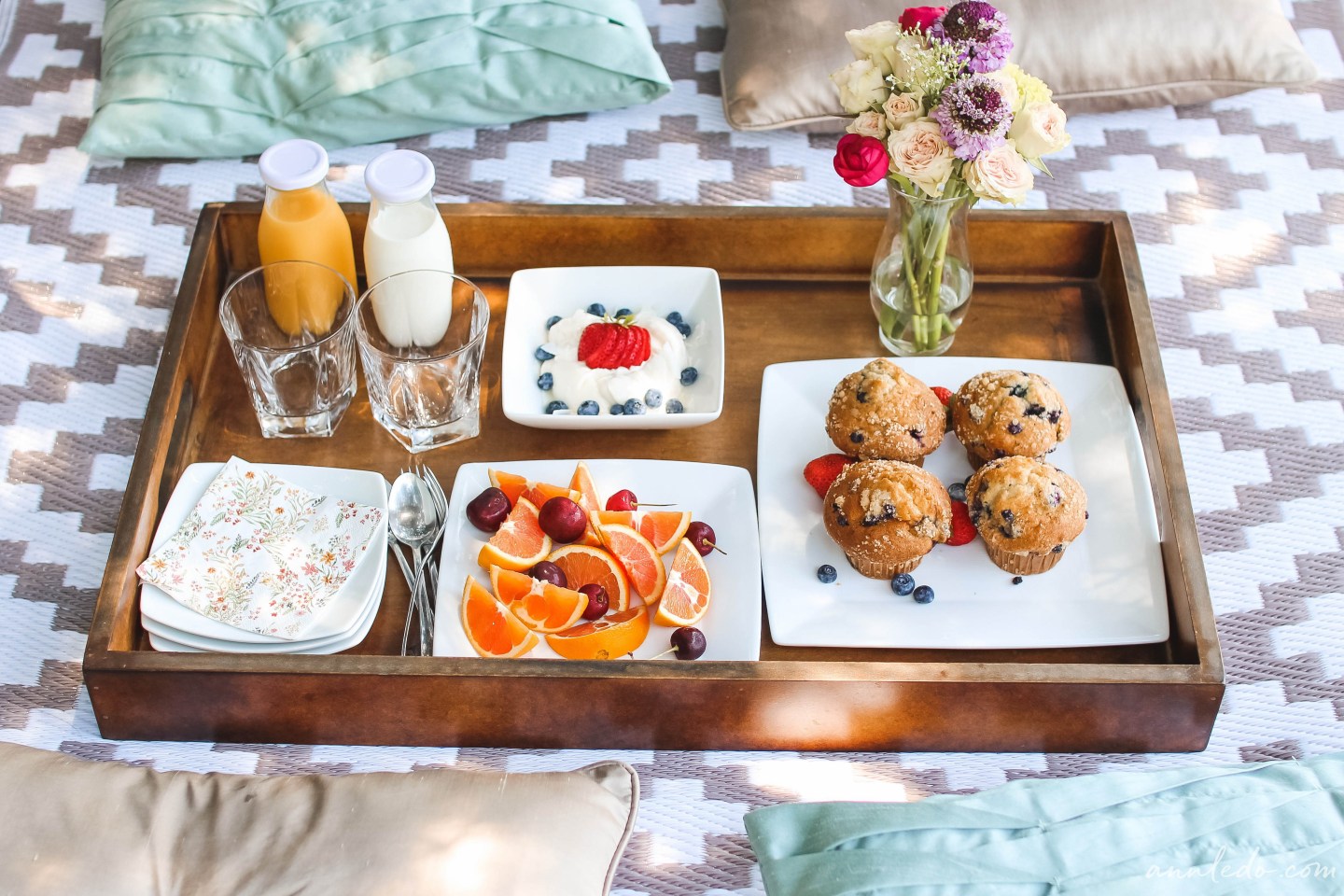}
\end{figure}
\CLIPRows[!b]{%
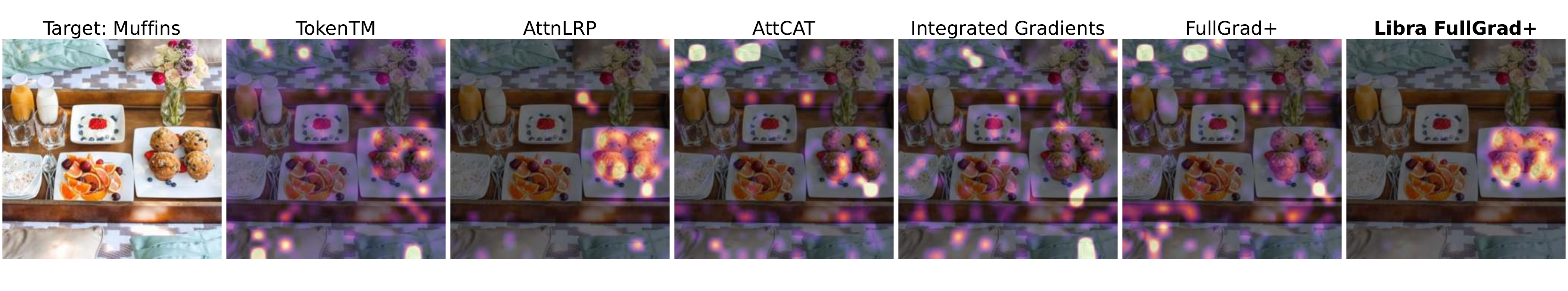,%
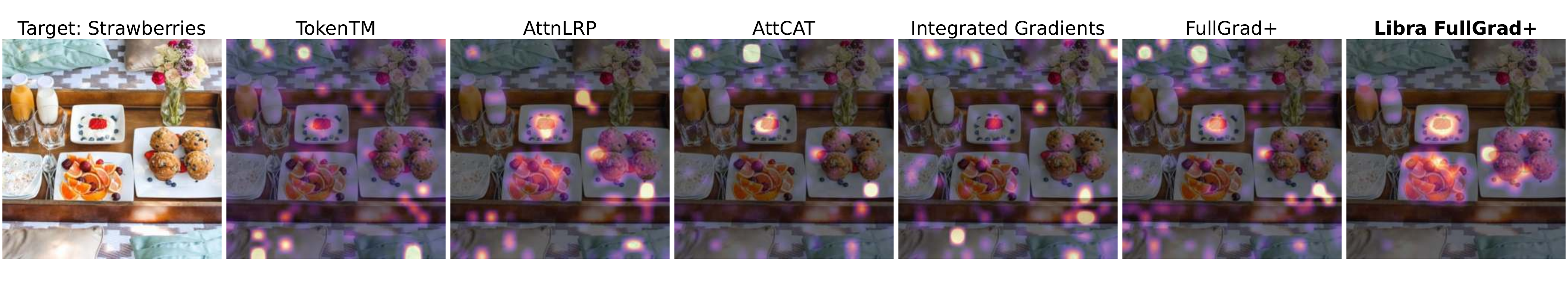,%
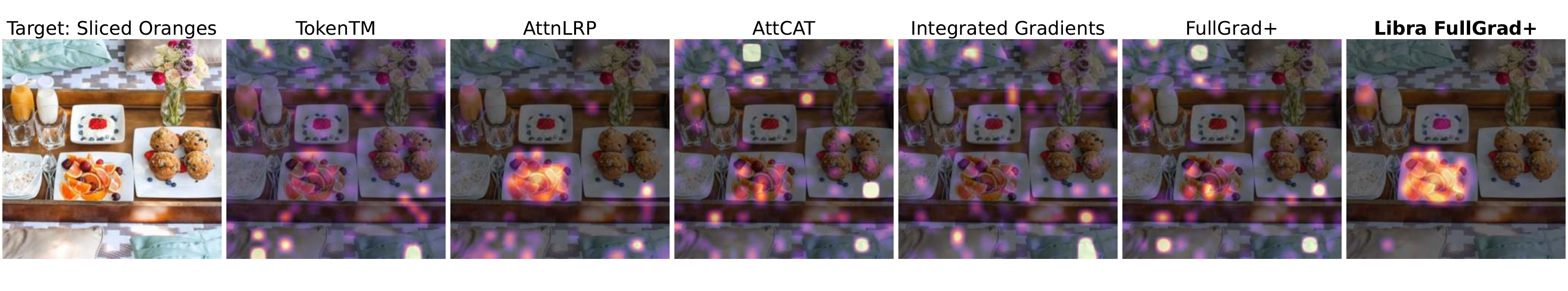,%
}{}{}

\CLIPRows[!b]{%
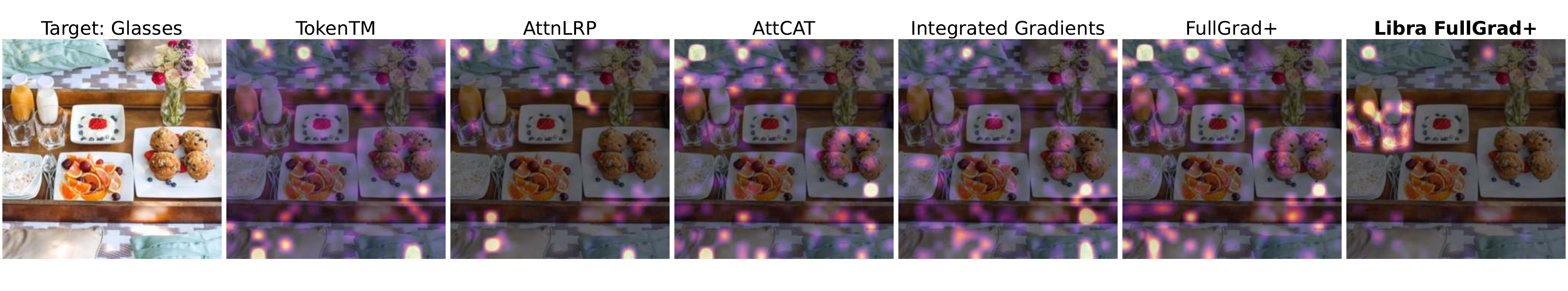,%
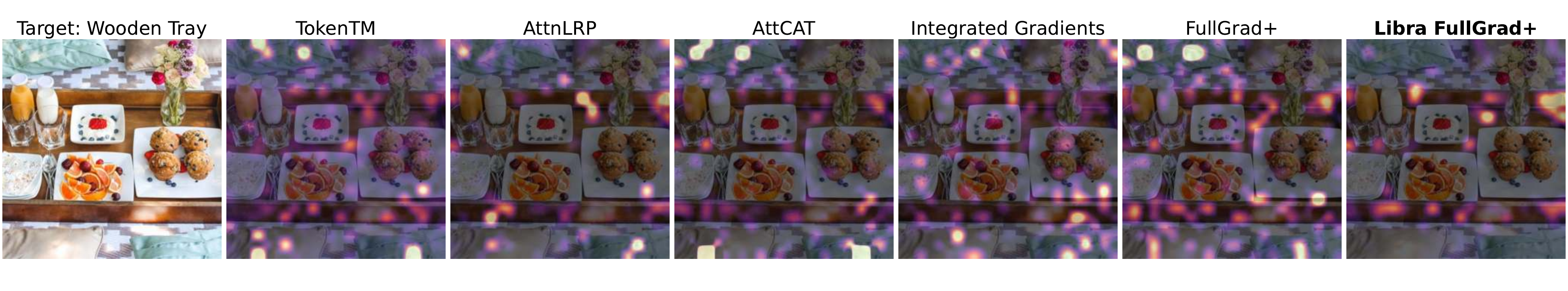,%
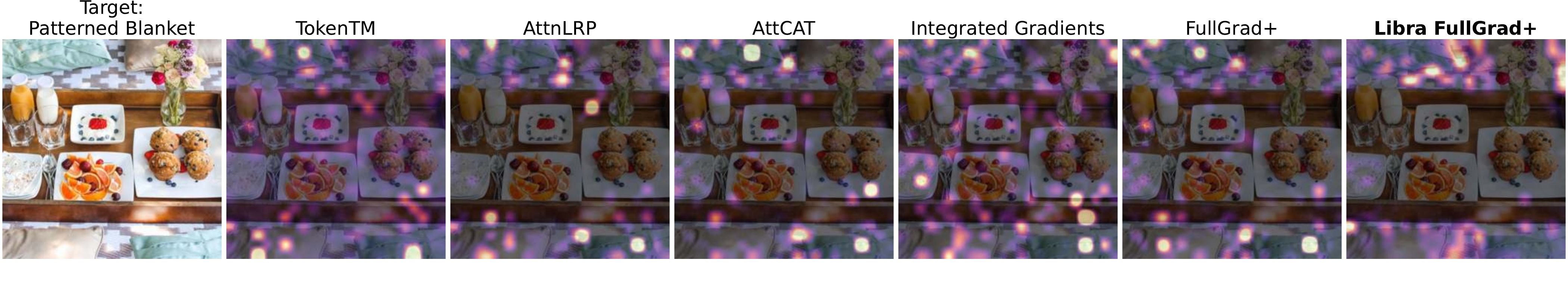,%
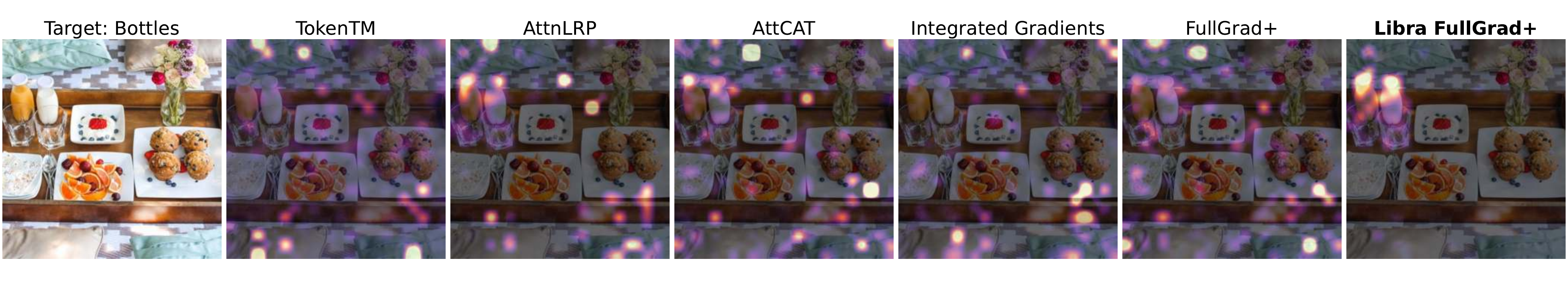,%
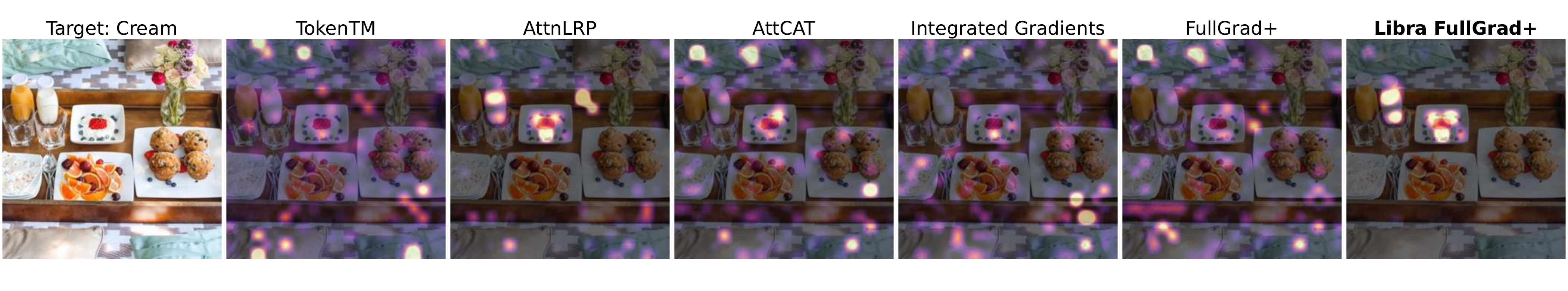,%
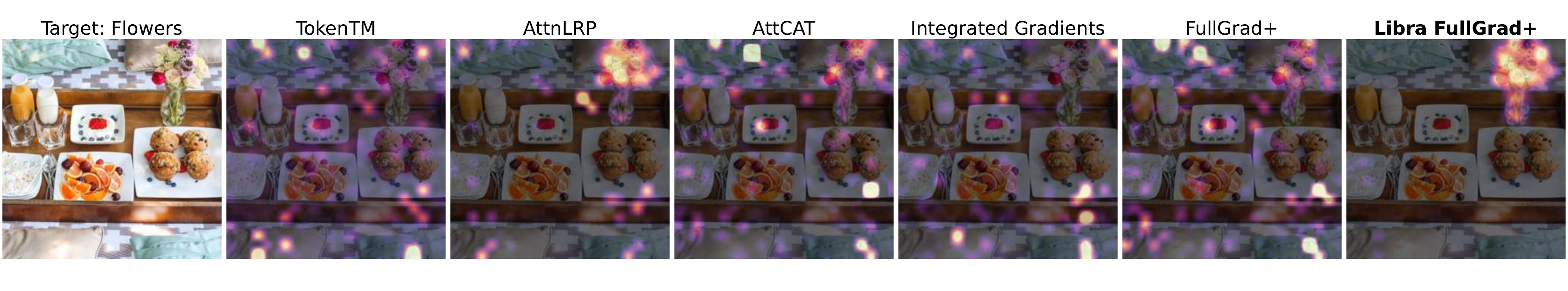,%
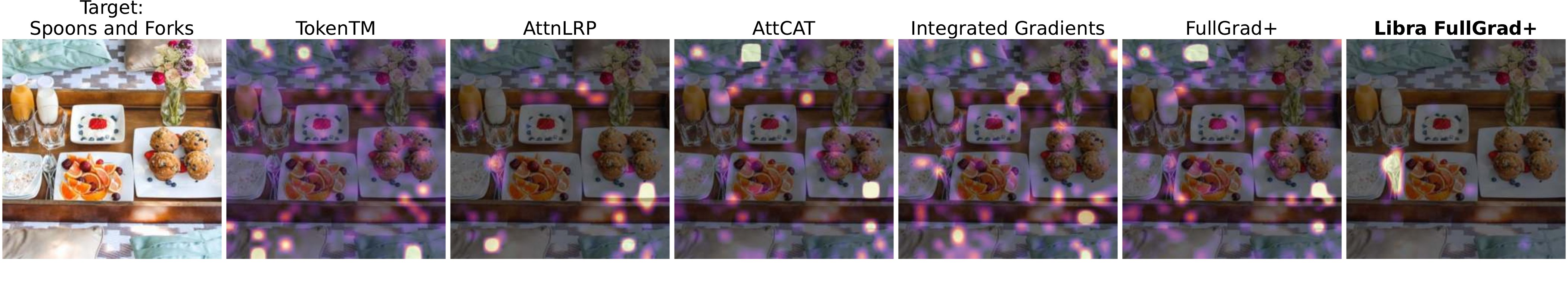,%
}{}{}

\CLIPRows[!b]{%
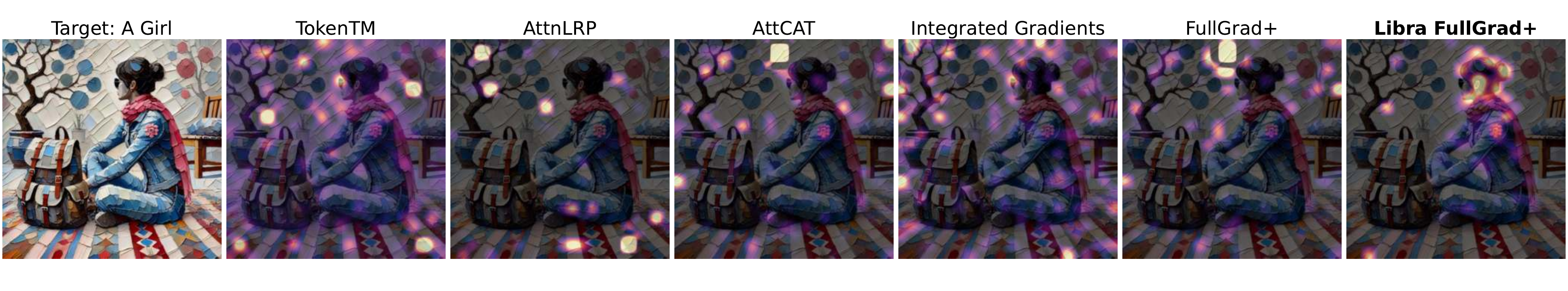,%
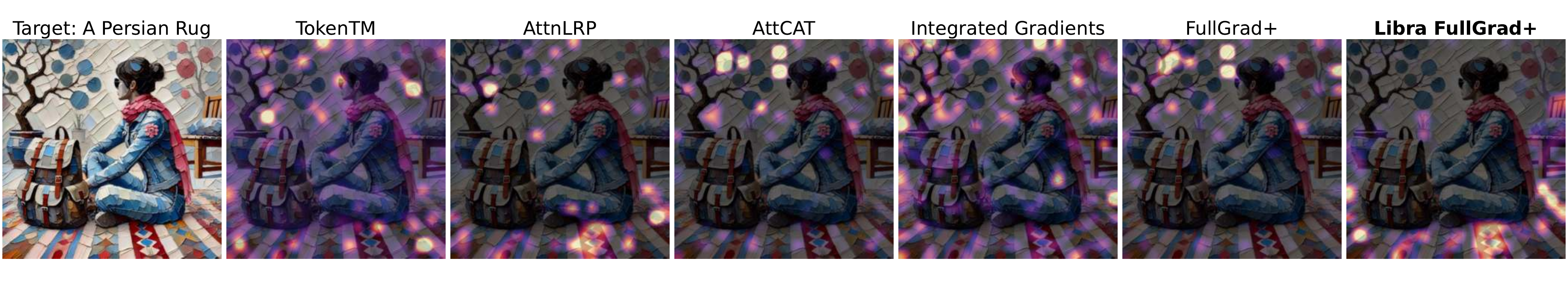,%
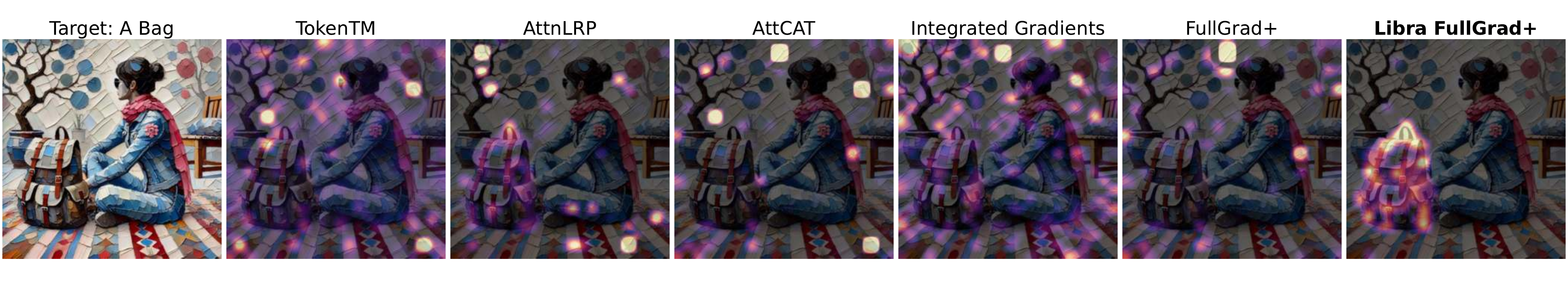,%
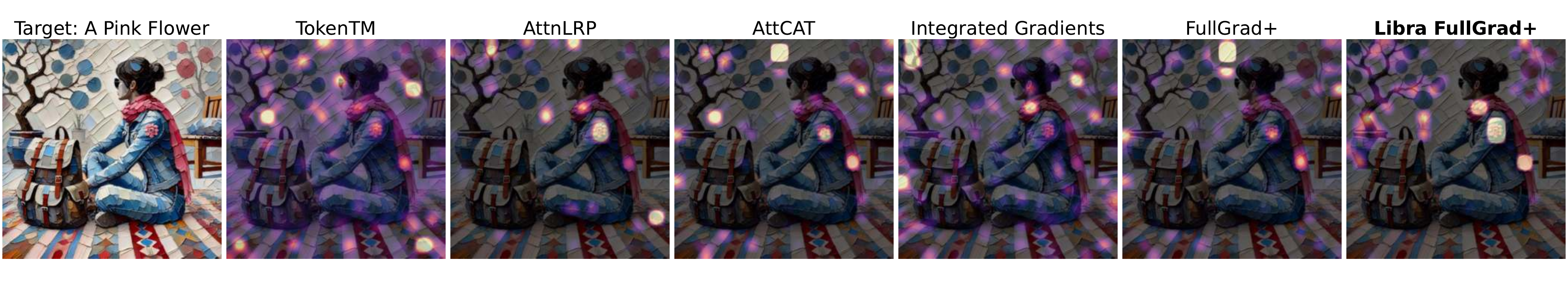,%
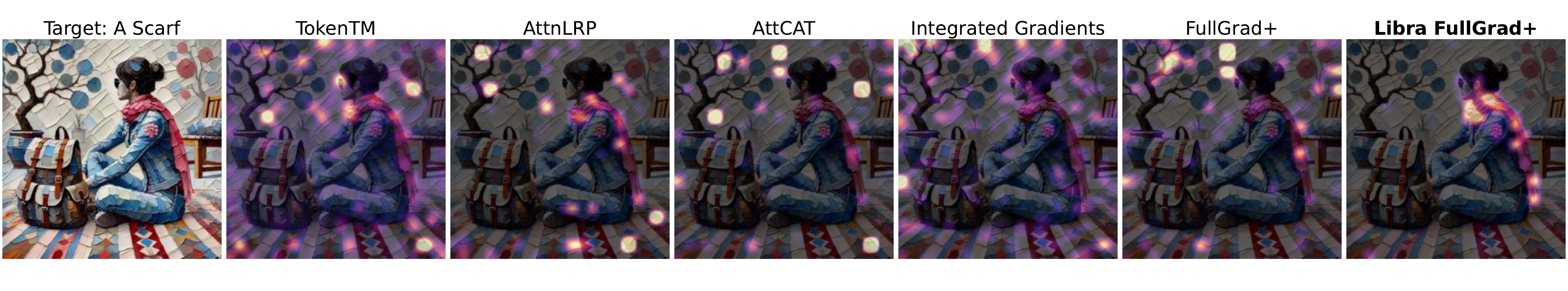,%
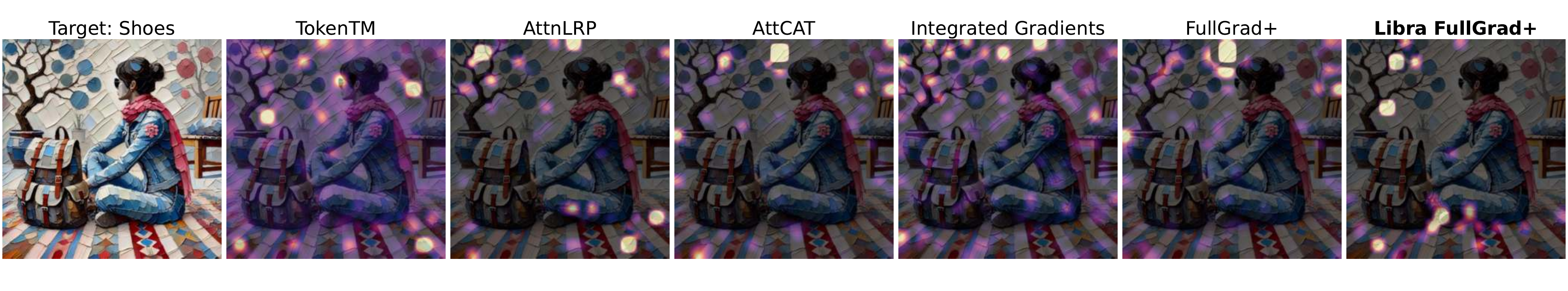,%
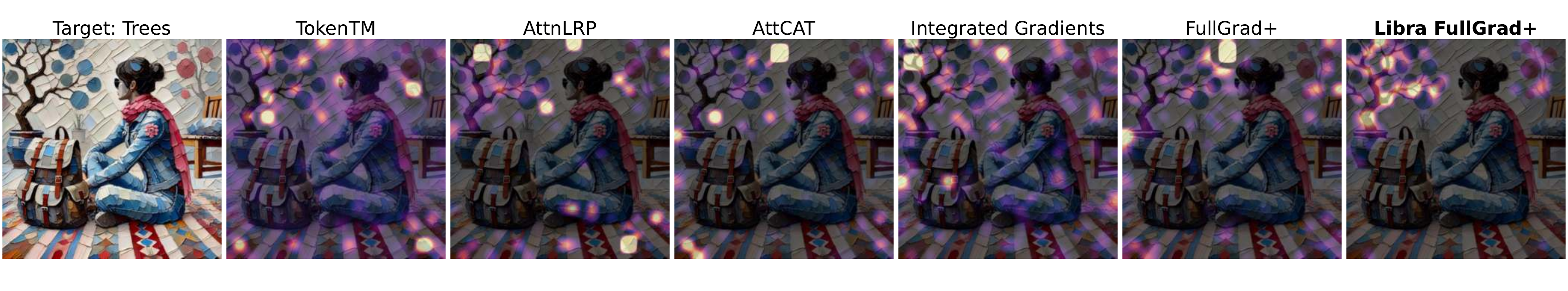,%
}{}{}

\CLIPRows[!b]{%
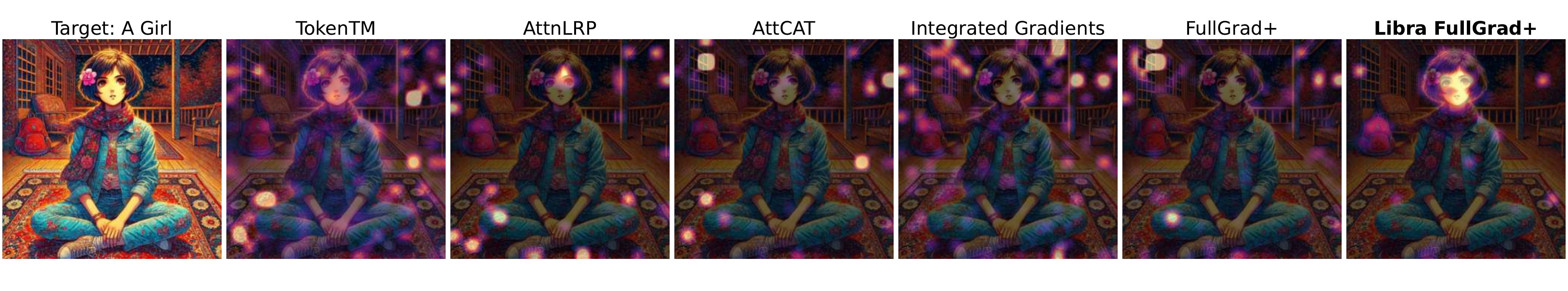,%
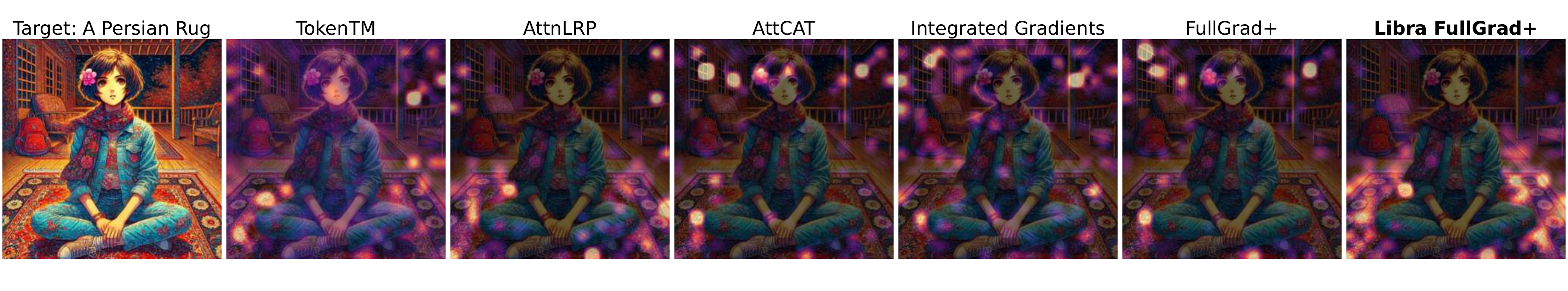,%
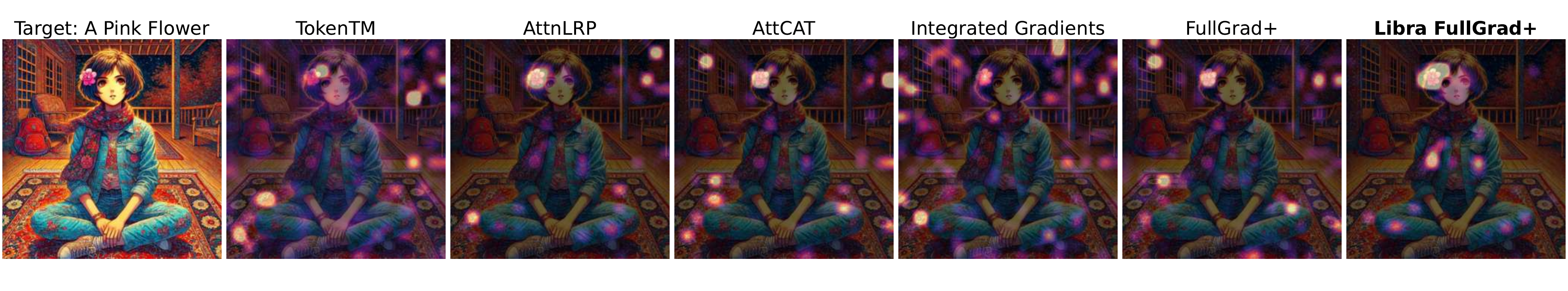,%
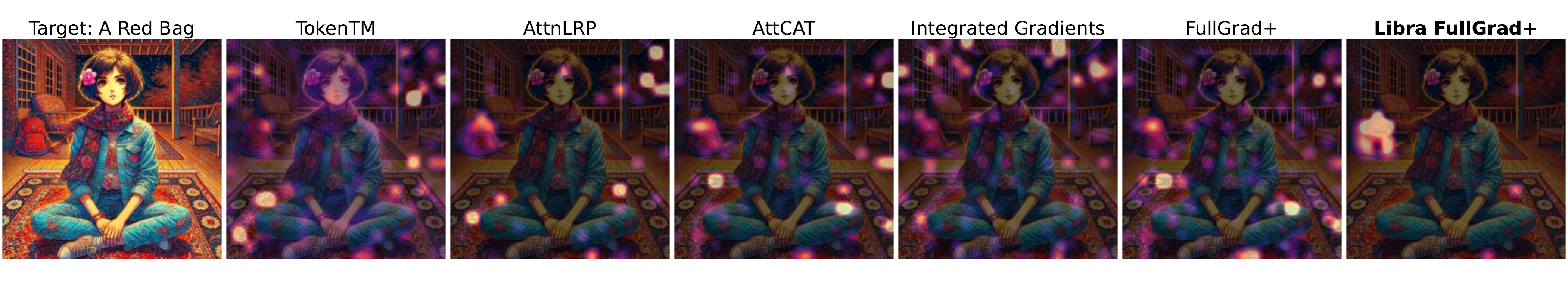,%
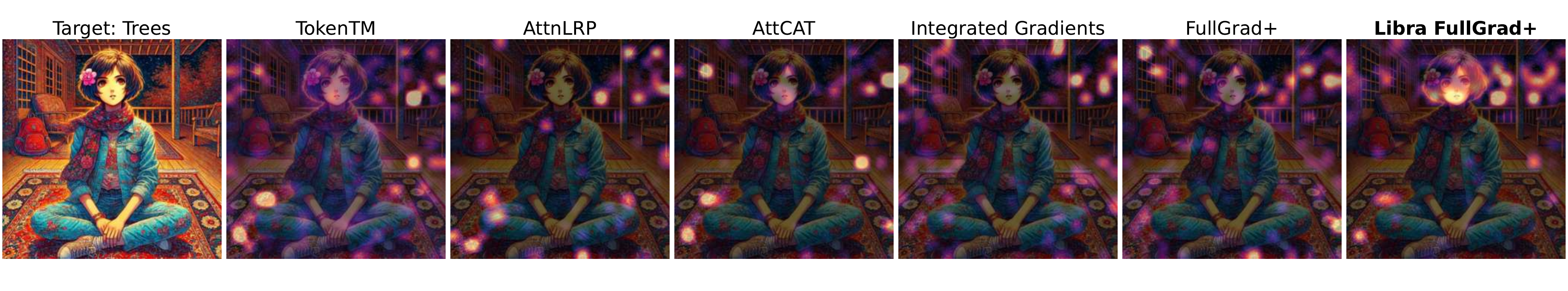,%
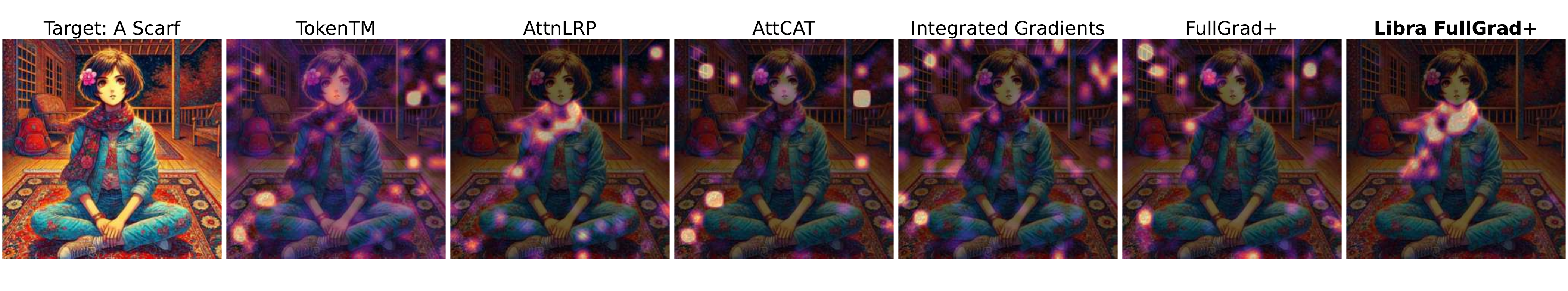,%
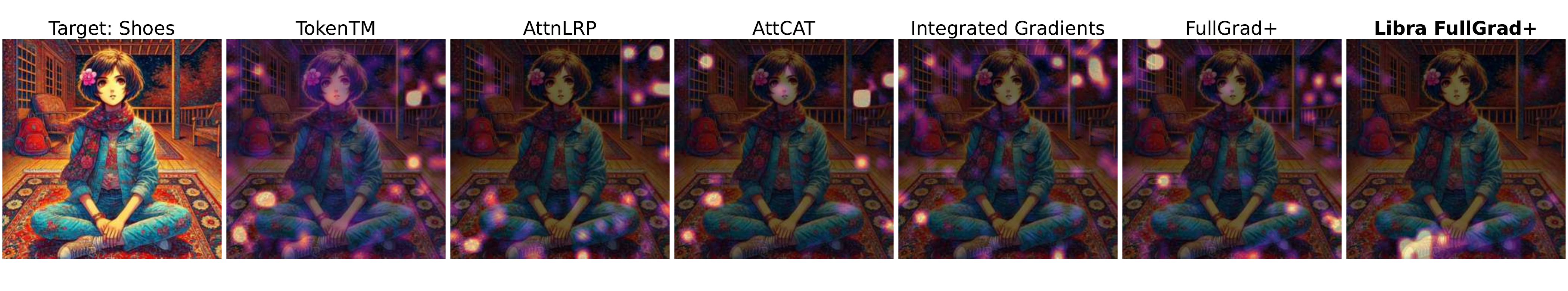,%
}{}{}

\CLIPRows[!b]{%
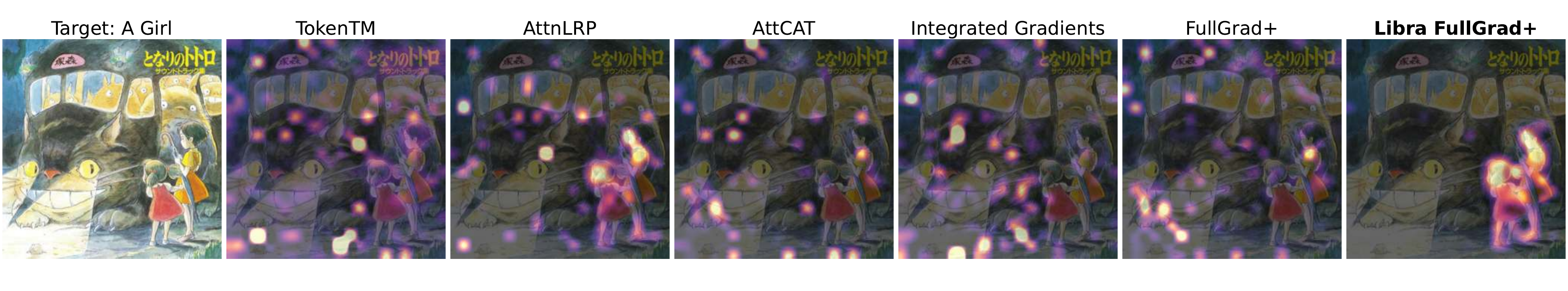,%
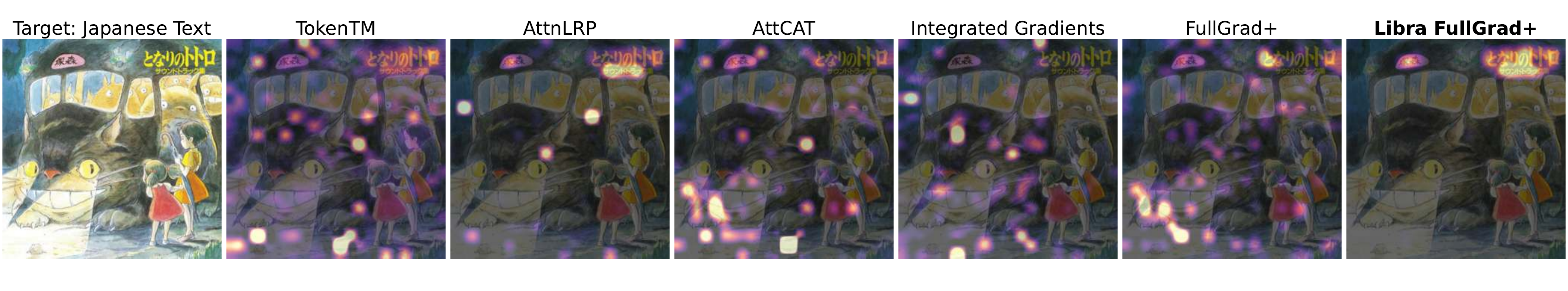,%
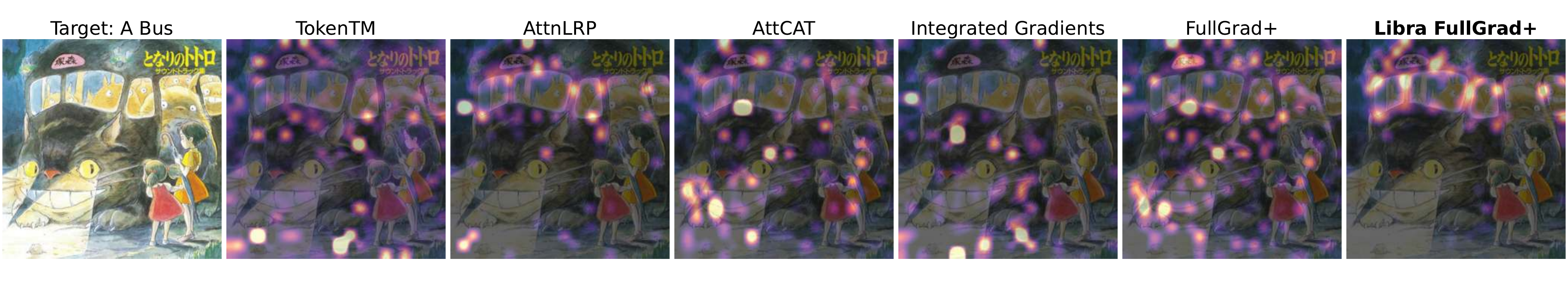,%
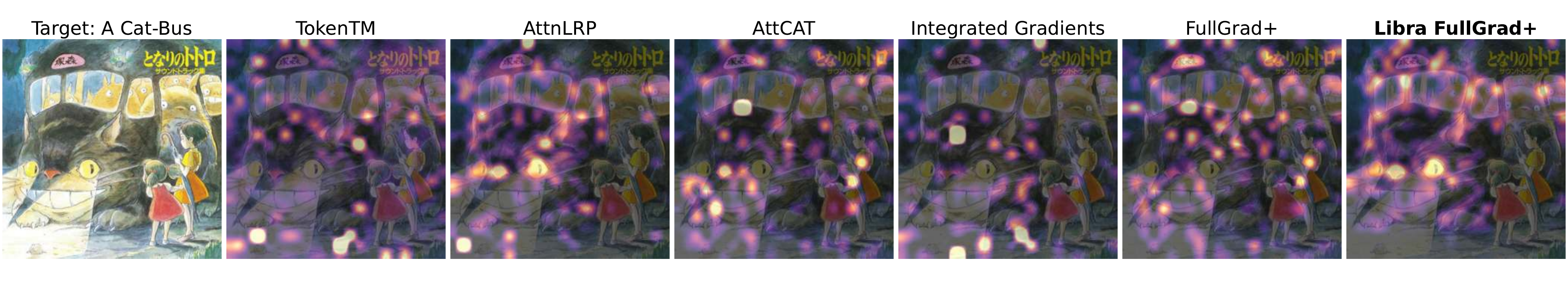,%
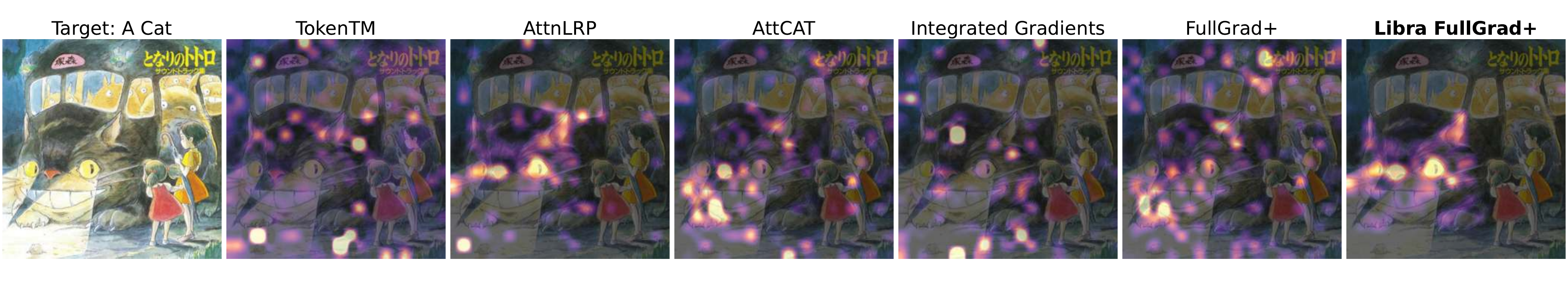,%
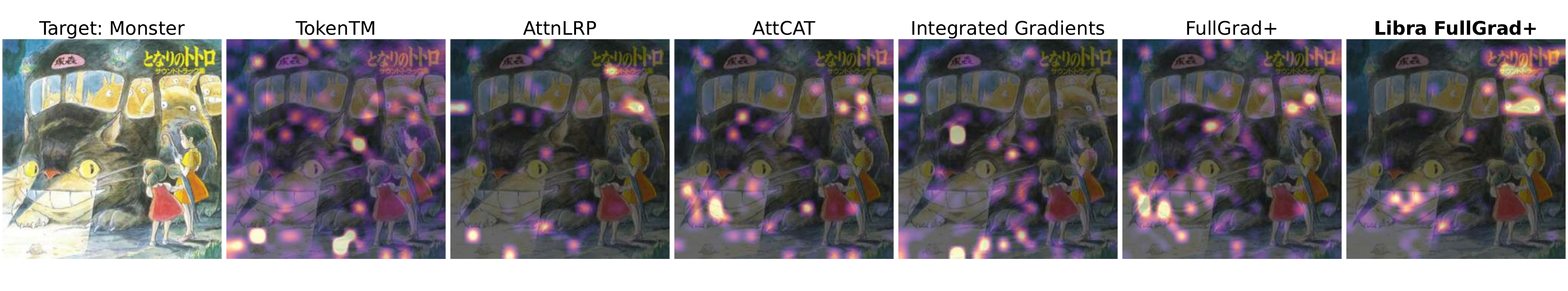,%
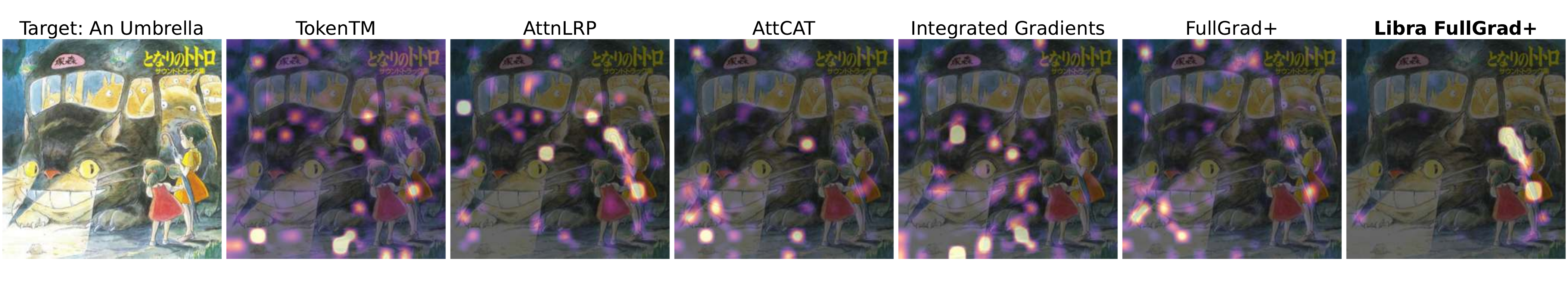,%
}{}{}

\begin{figure}[h]
    \centering
    \includegraphics[width=\linewidth]{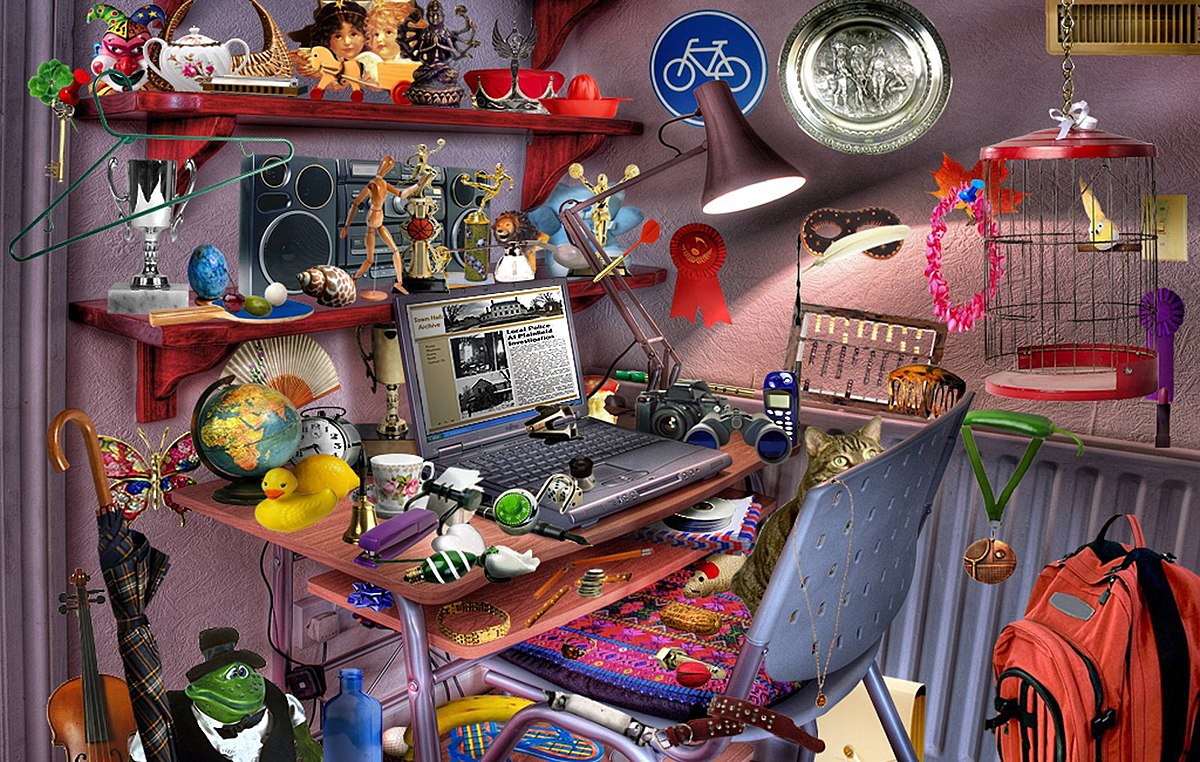}
\end{figure}
\CLIPRows[!b]{%
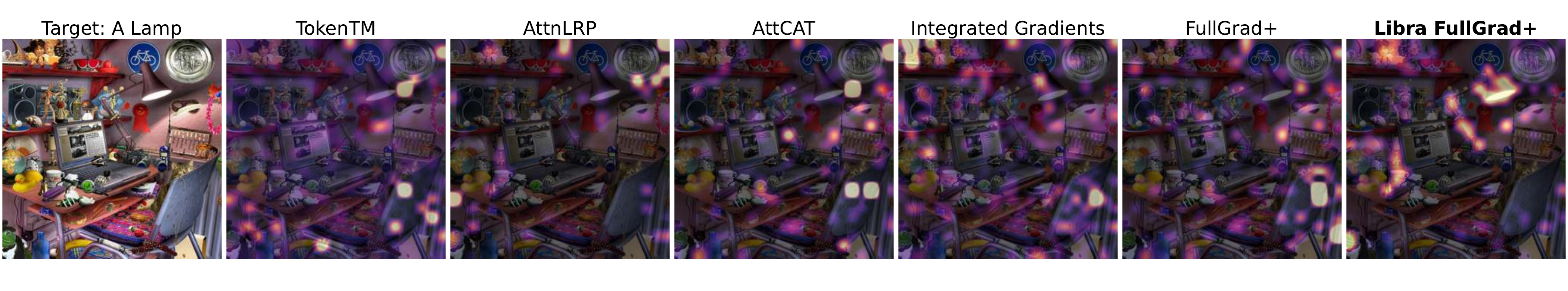,%
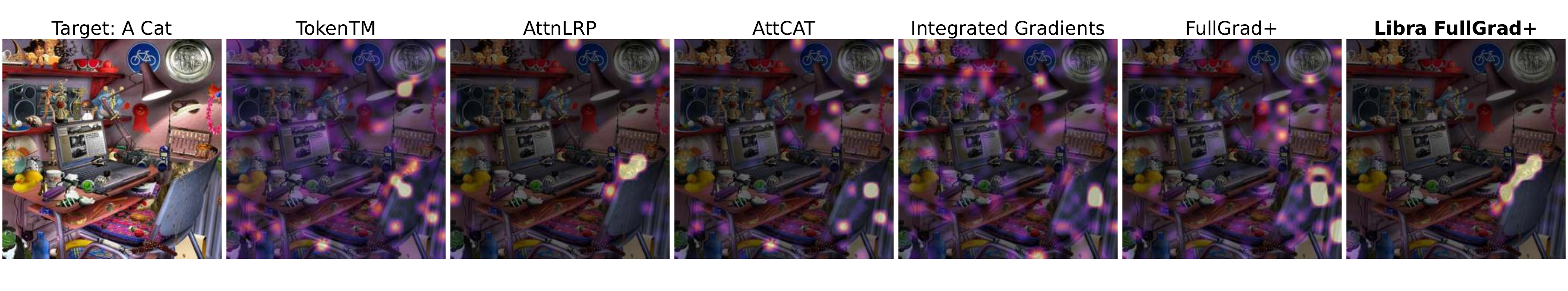,%
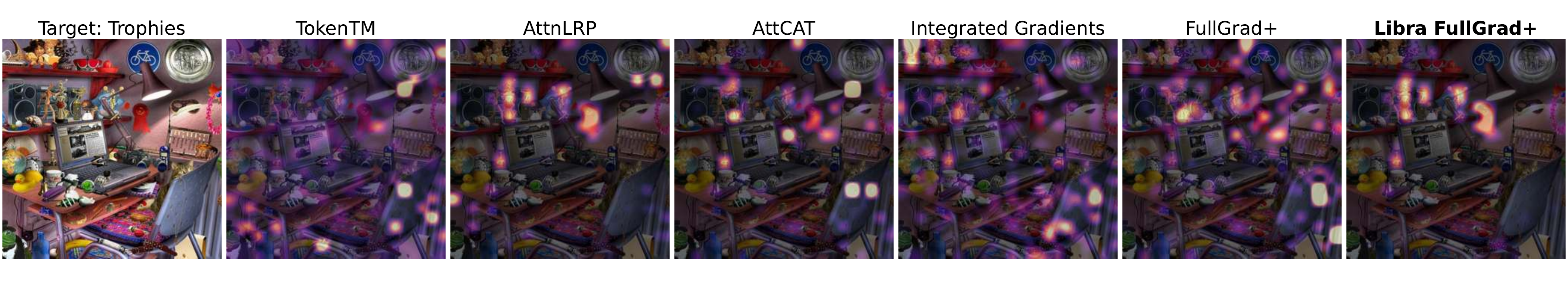,%
}{}{}

\CLIPRows[!b]{%
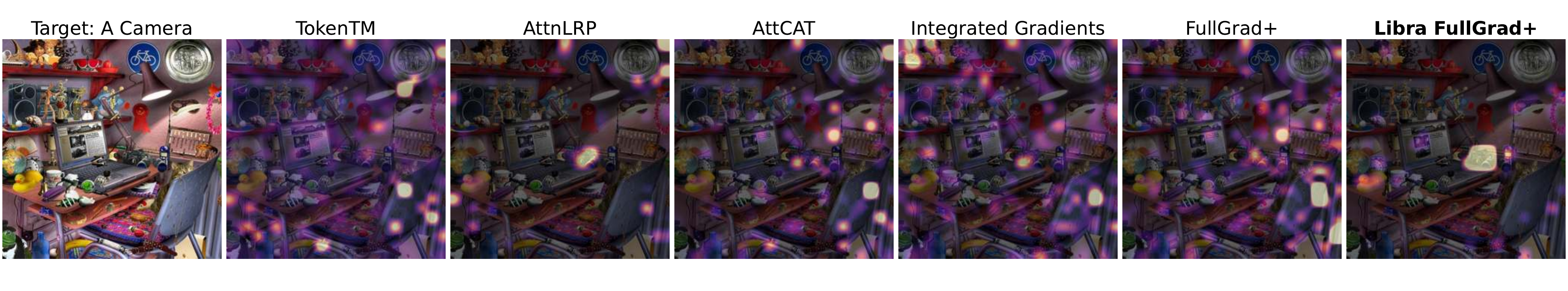,%
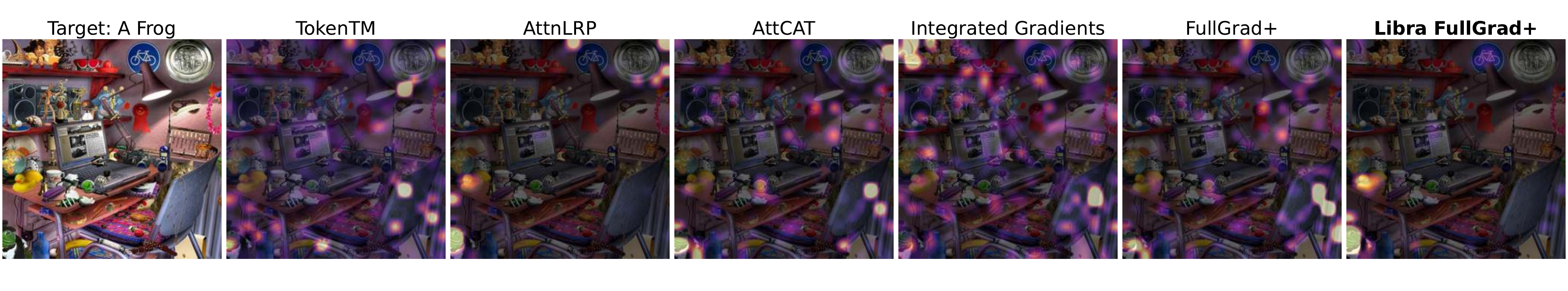,%
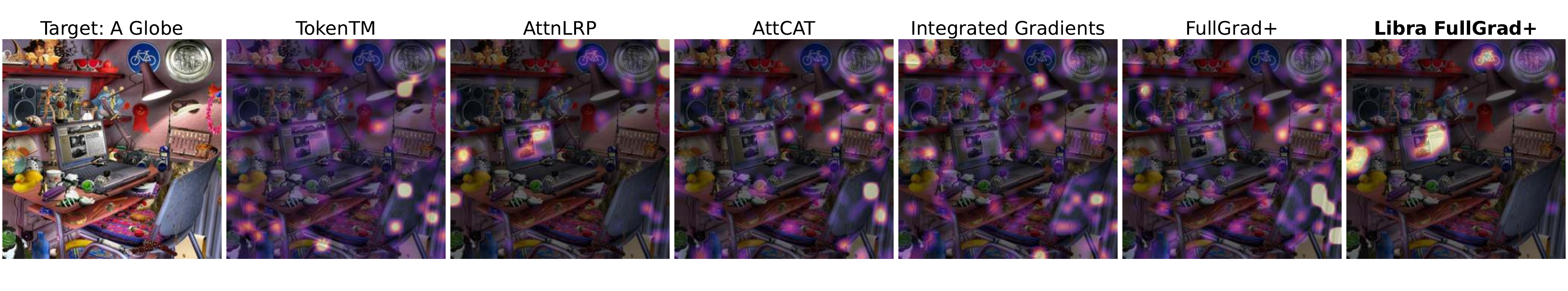,%
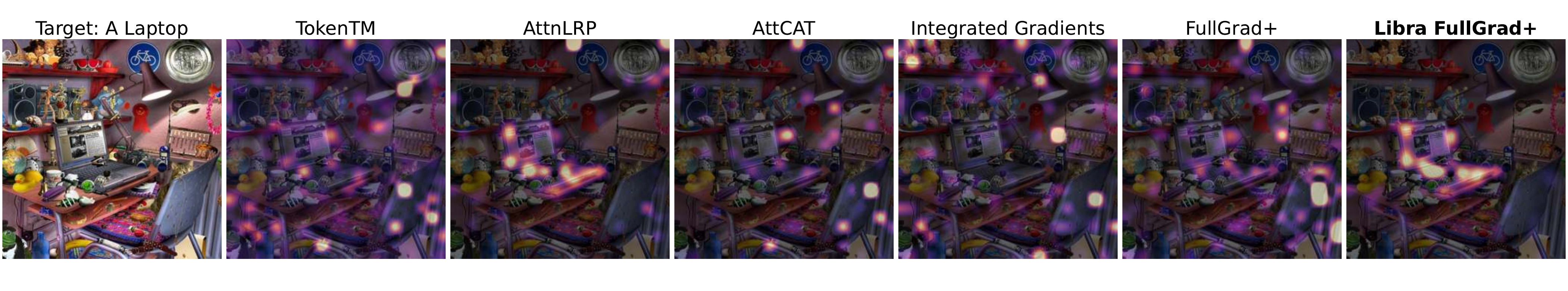,%
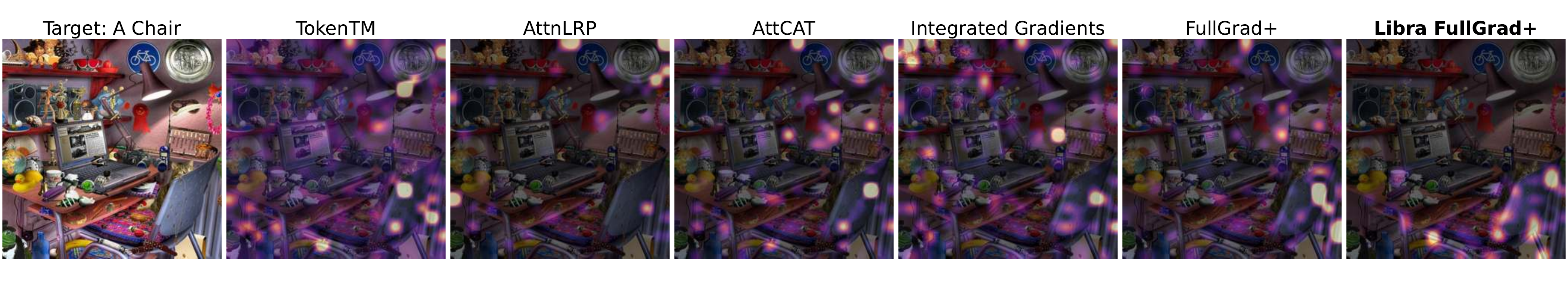,%
}{}{}

\CLIPRows[!b]{%
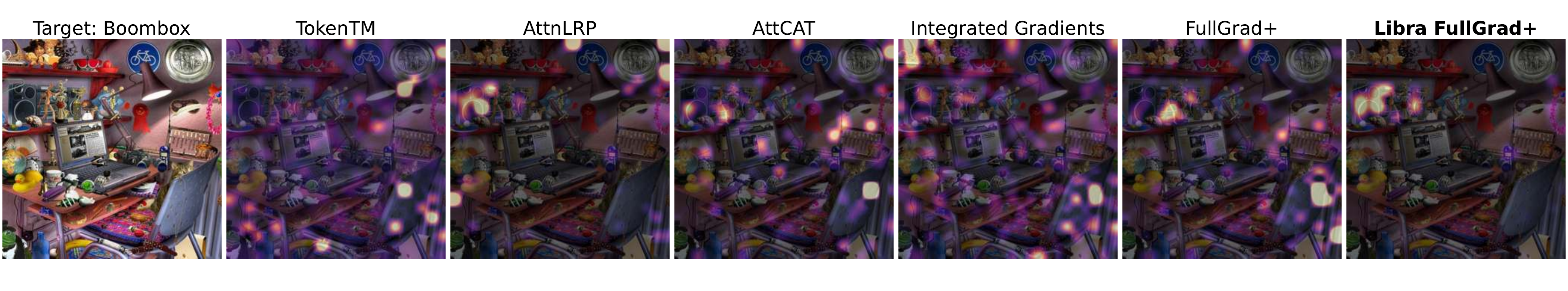,%
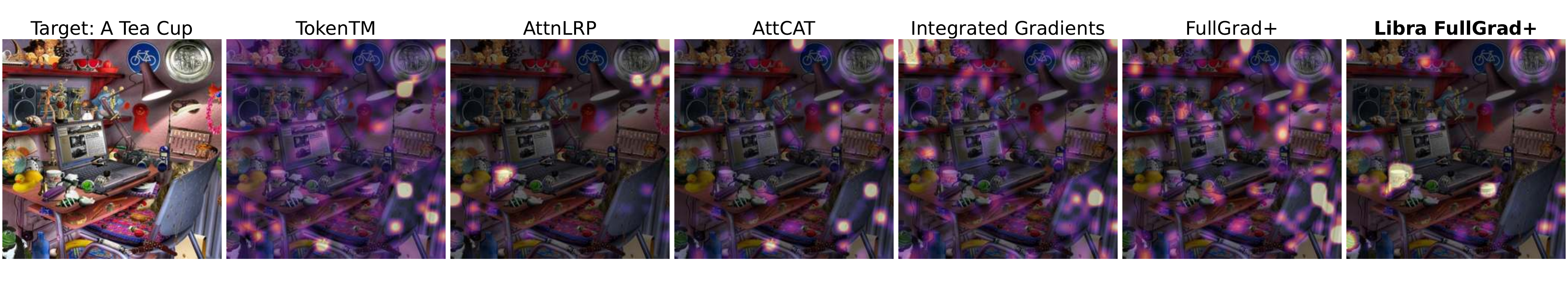,%
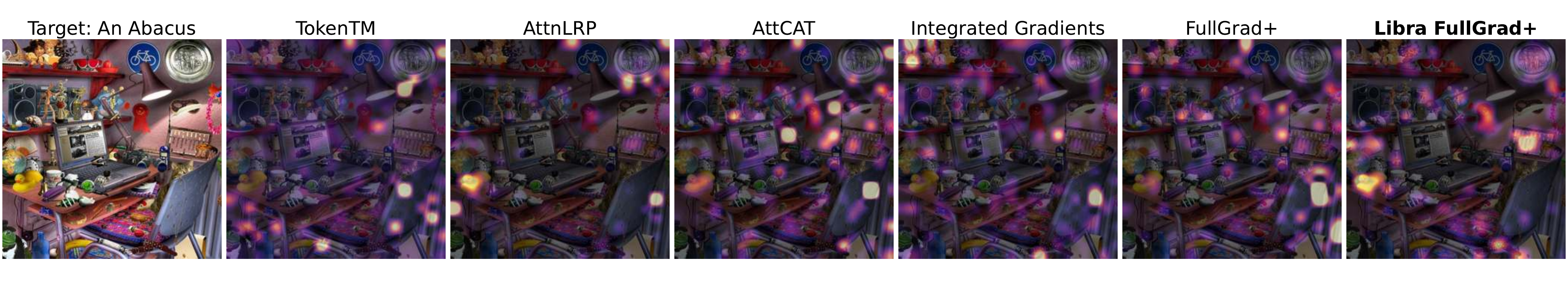,%
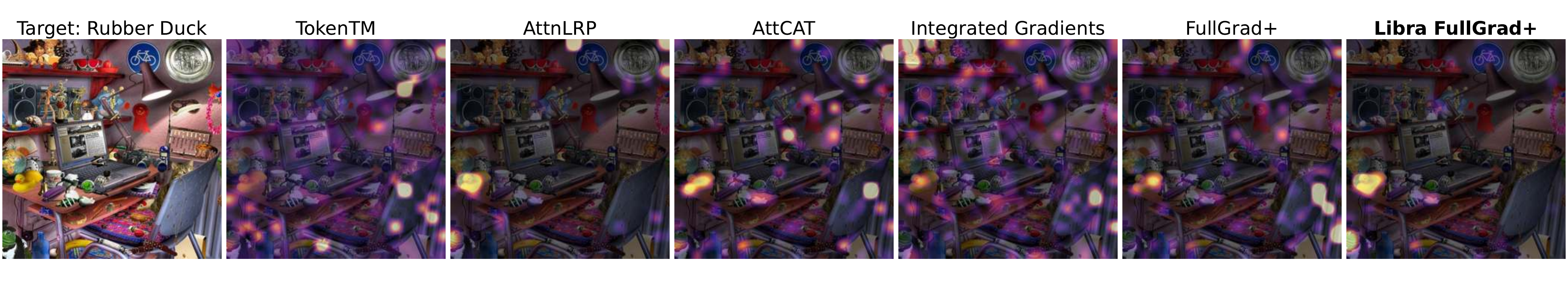,%
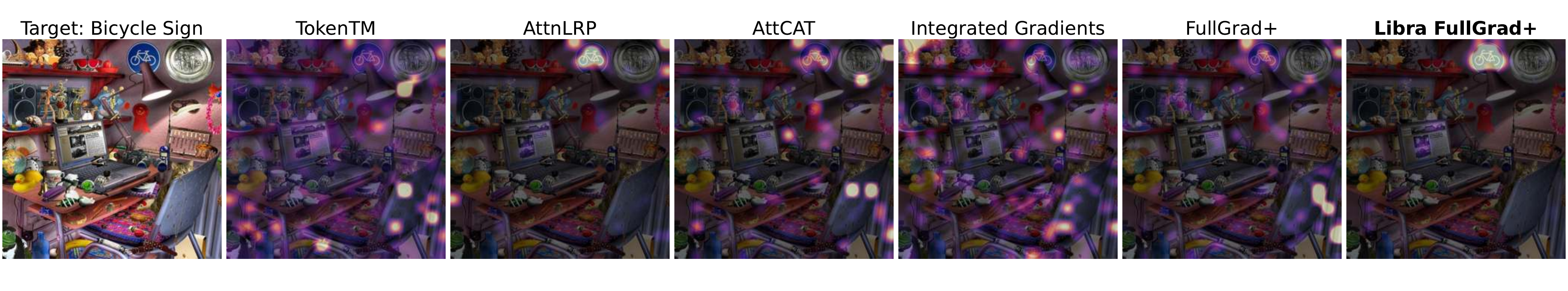,%
}{}{}

\begin{figure}[h]
    \centering
    \includegraphics[width=\linewidth]{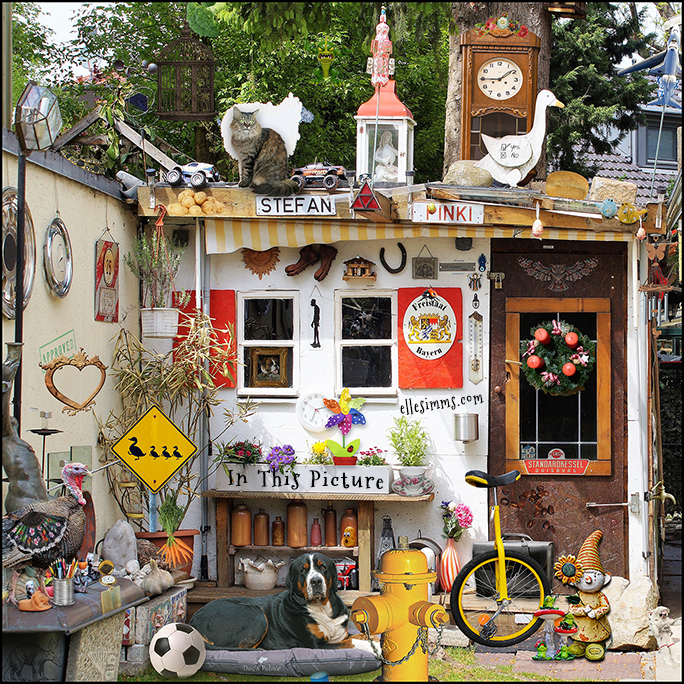}
\end{figure}
\CLIPRows[!b]{%
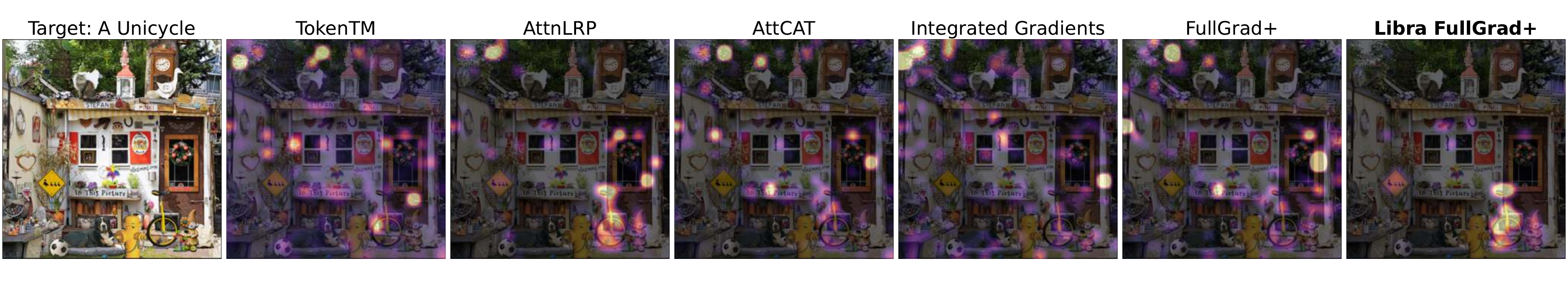,%
}{}{}

\CLIPRows[!b]{%
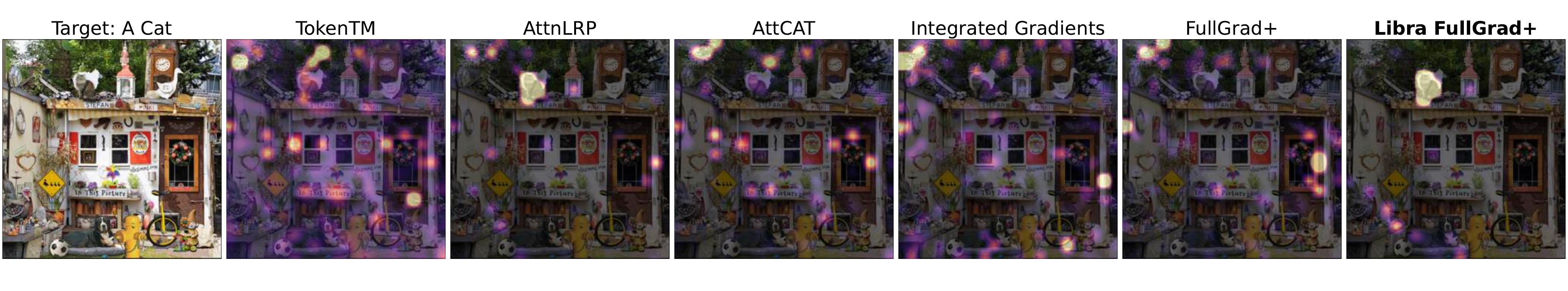,%
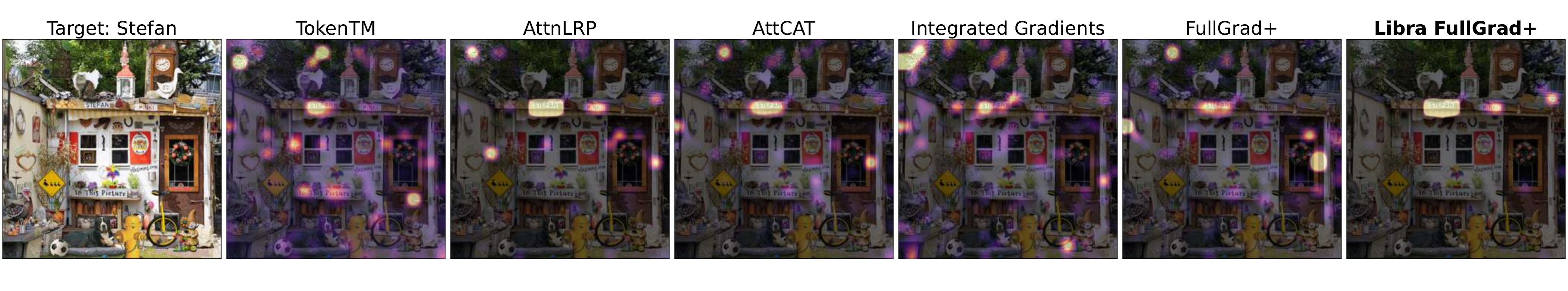,%
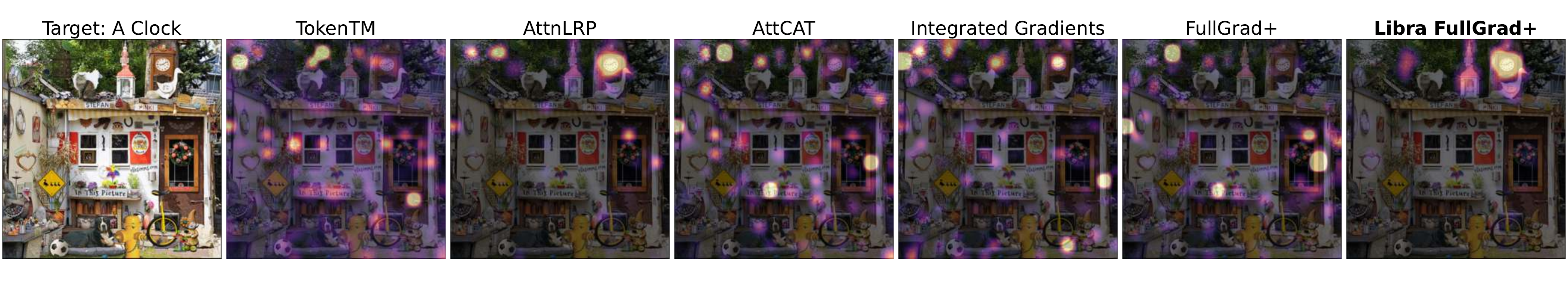,%
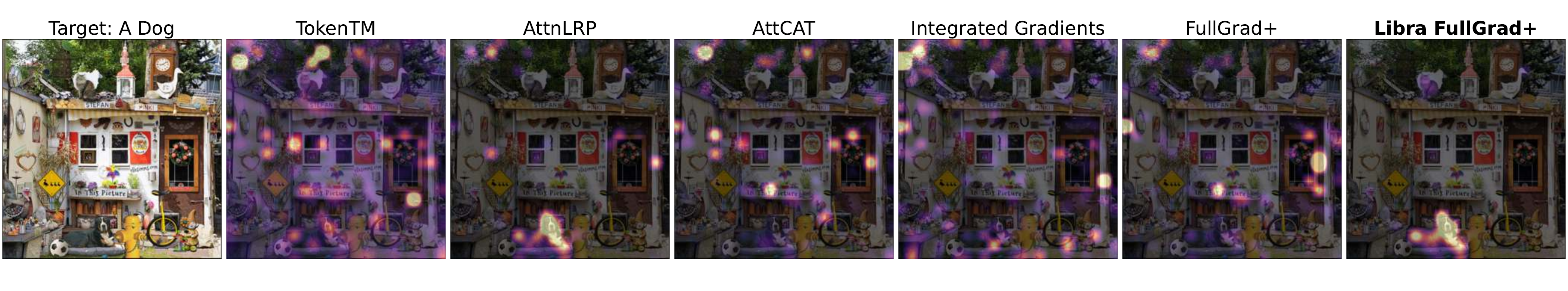,%
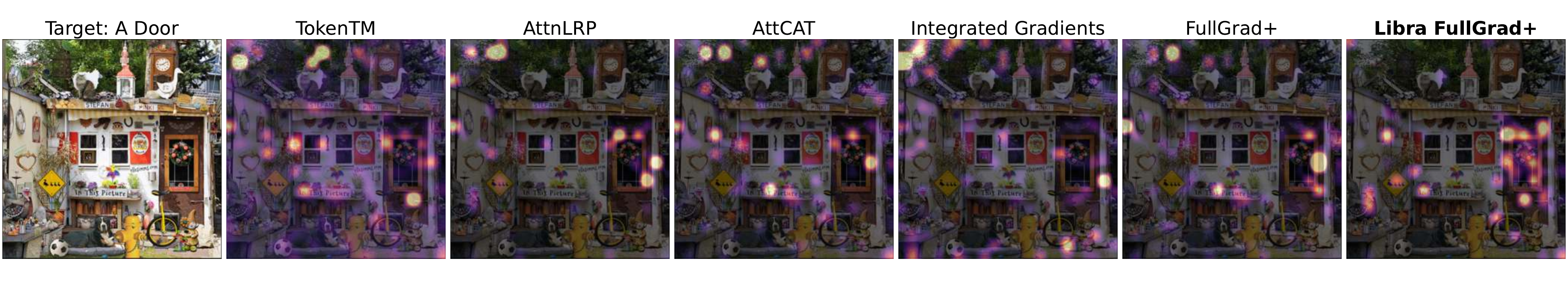,%
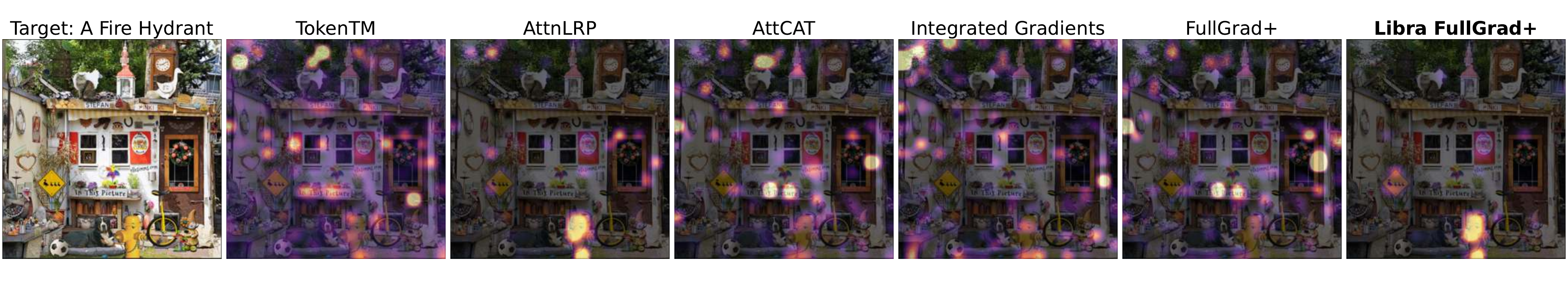,%
}{}{}

\CLIPRows[!b]{%
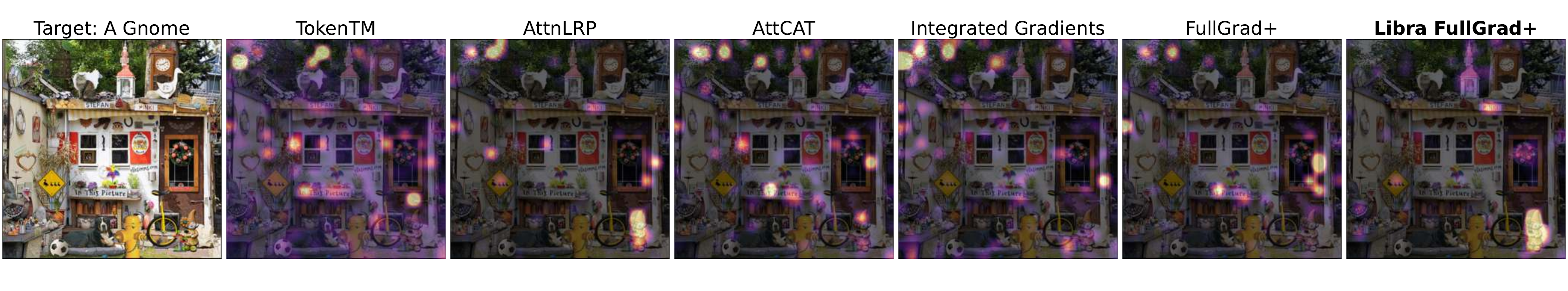,%
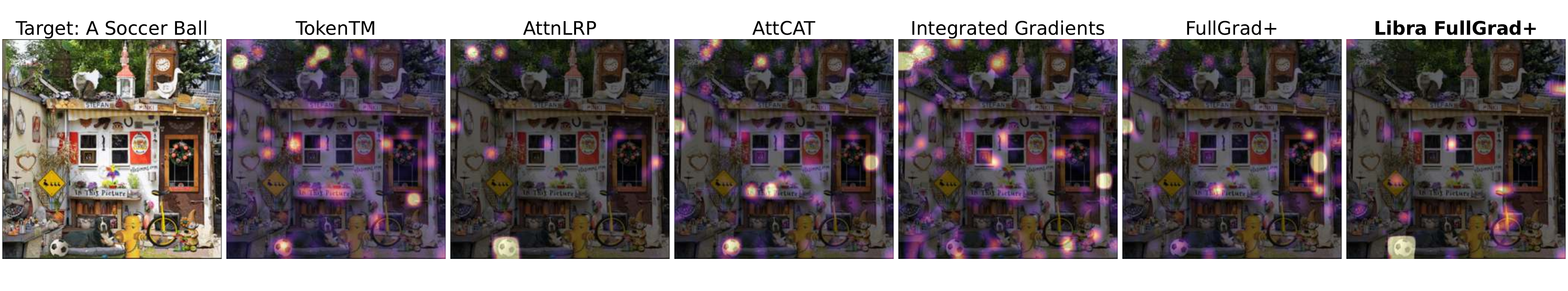,%
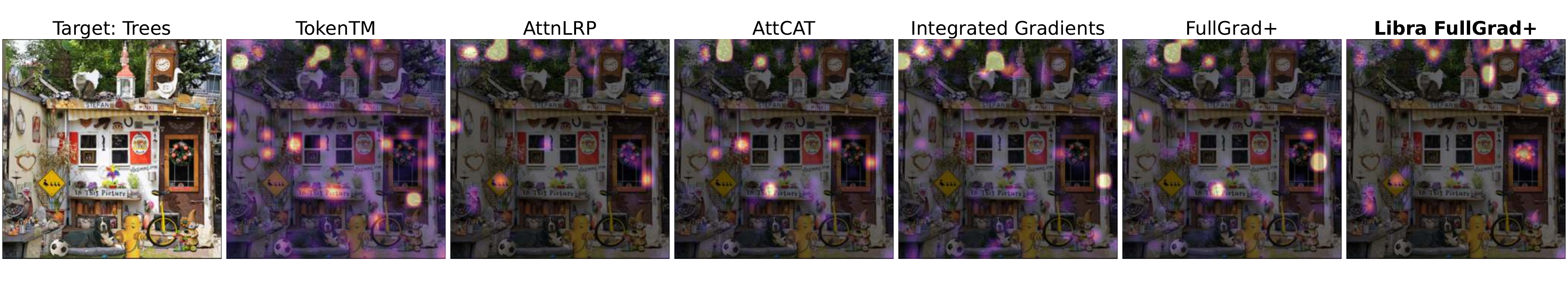,%
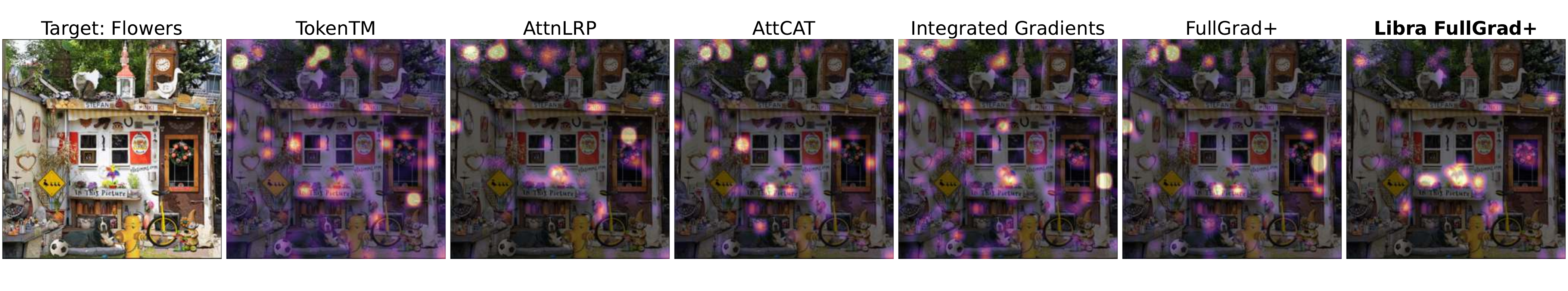,%
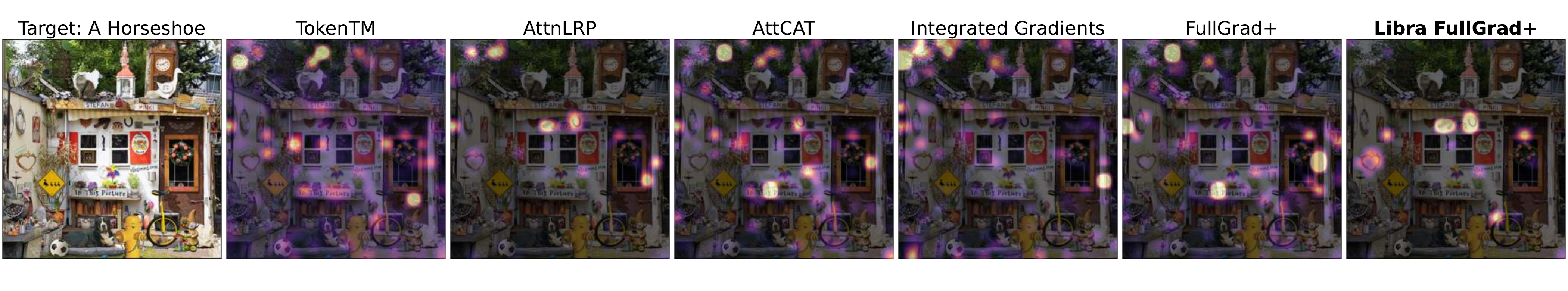,%
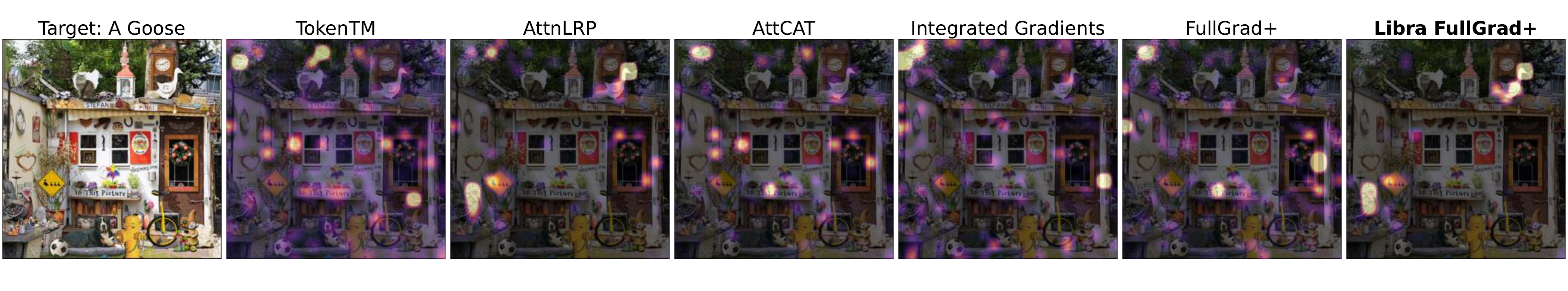,%
}{}{}

\begin{figure}[h]
    \centering
    \includegraphics[width=\linewidth]{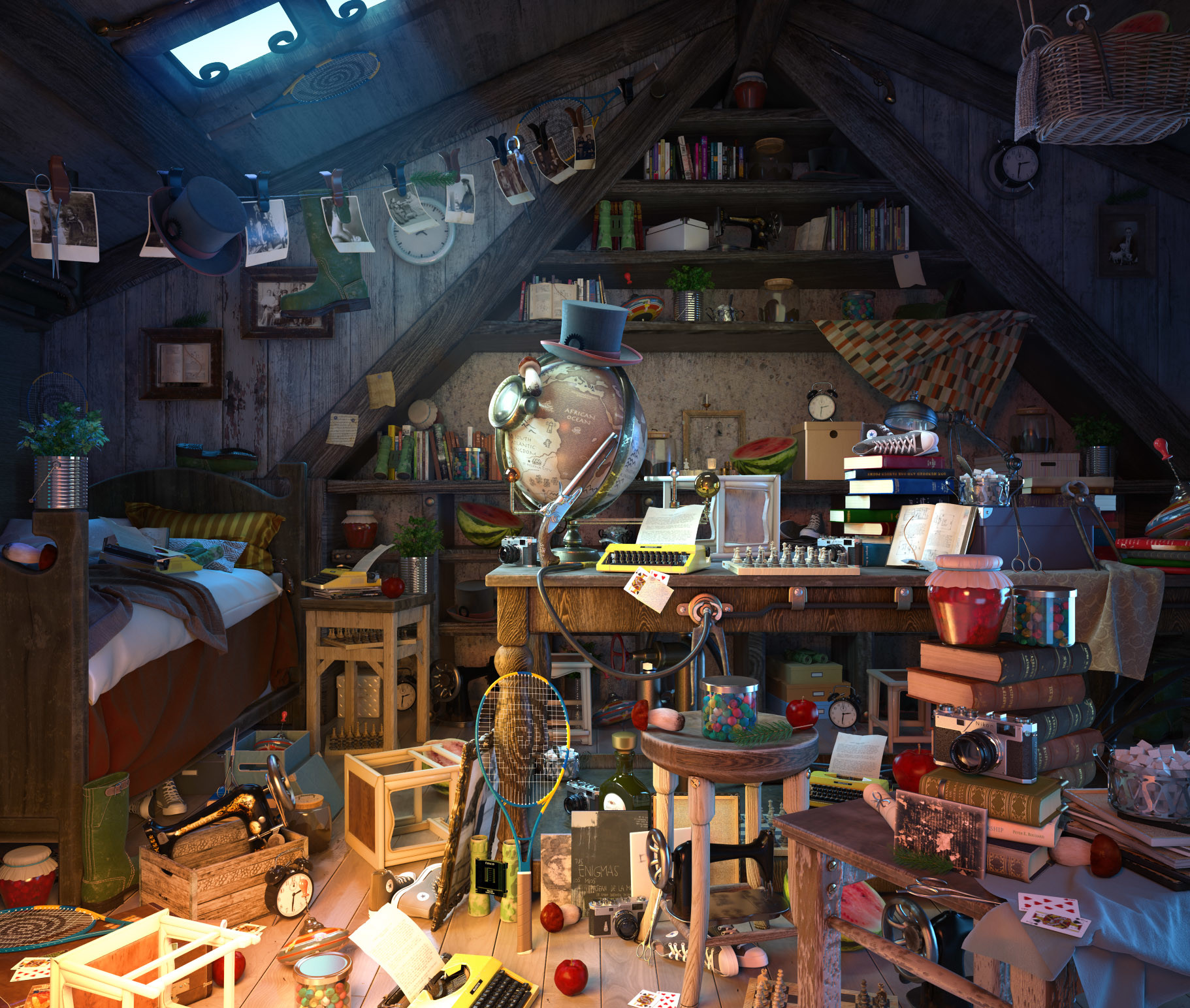}
\end{figure}
\CLIPRows[!b]{%
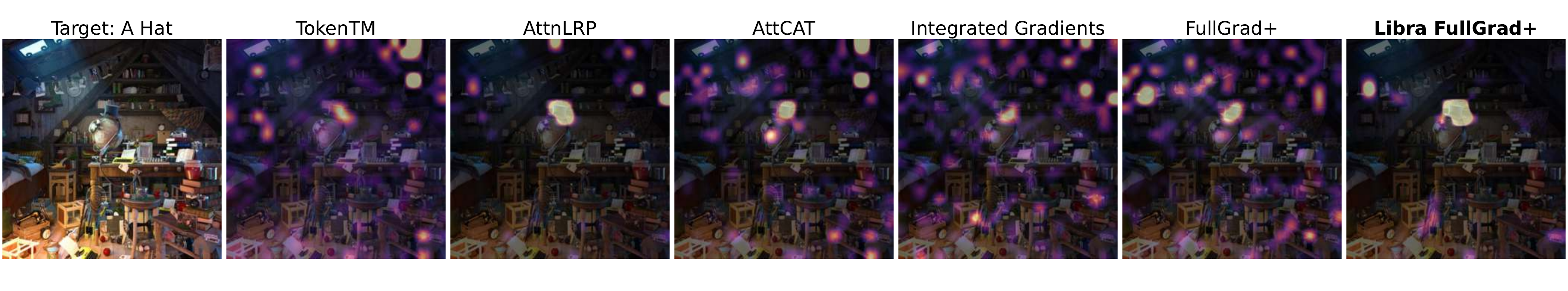,%
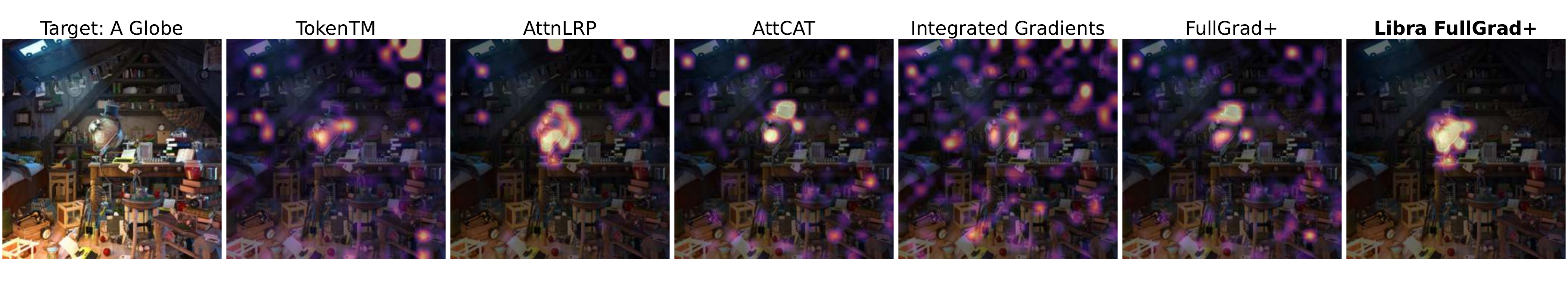,%
}{}{}

\CLIPRows[!b]{%
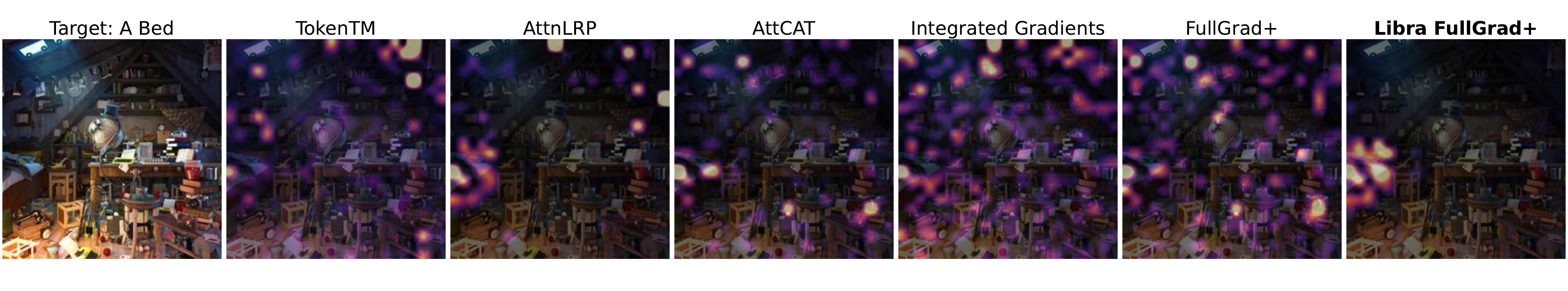,%
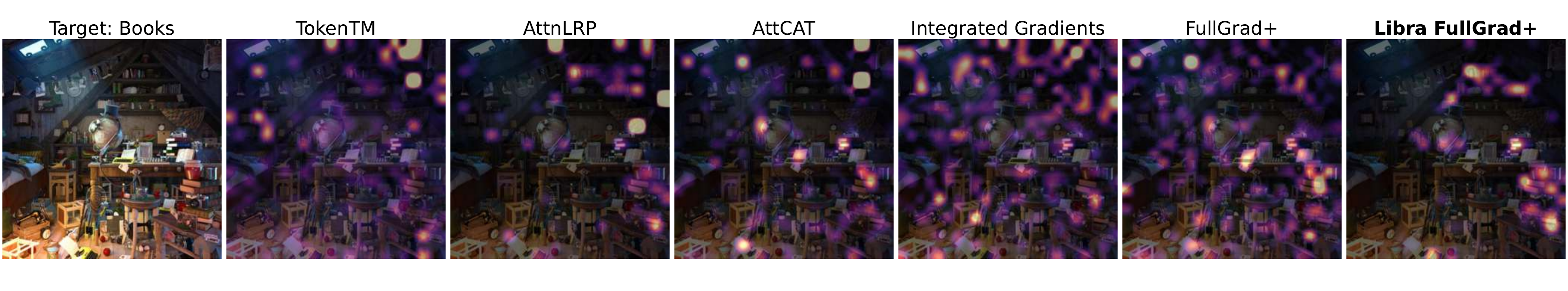,%
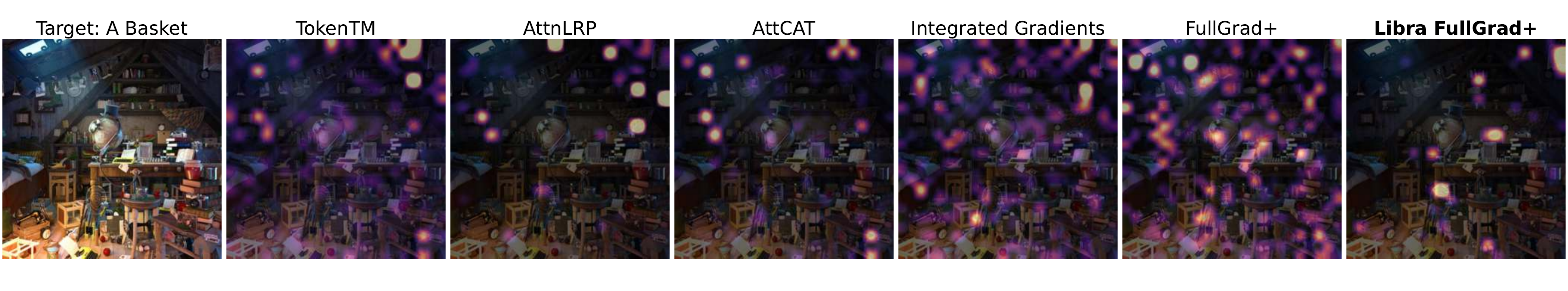,%
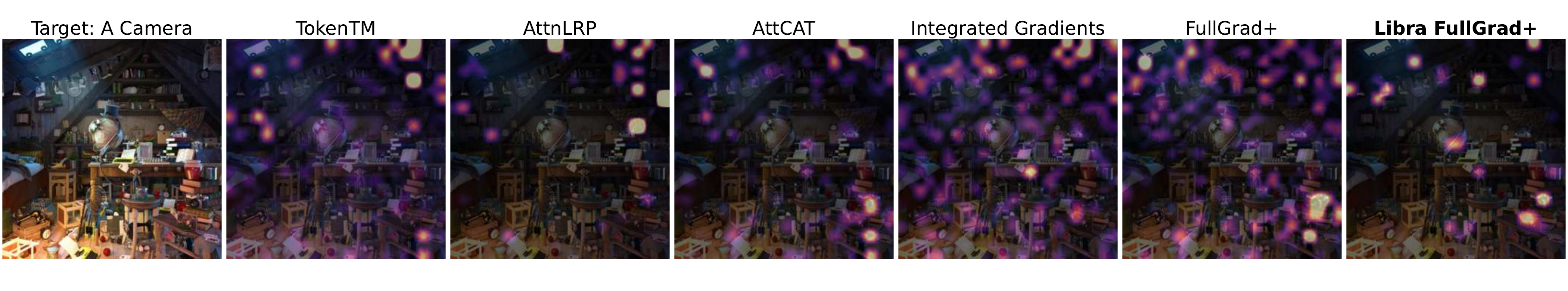,%
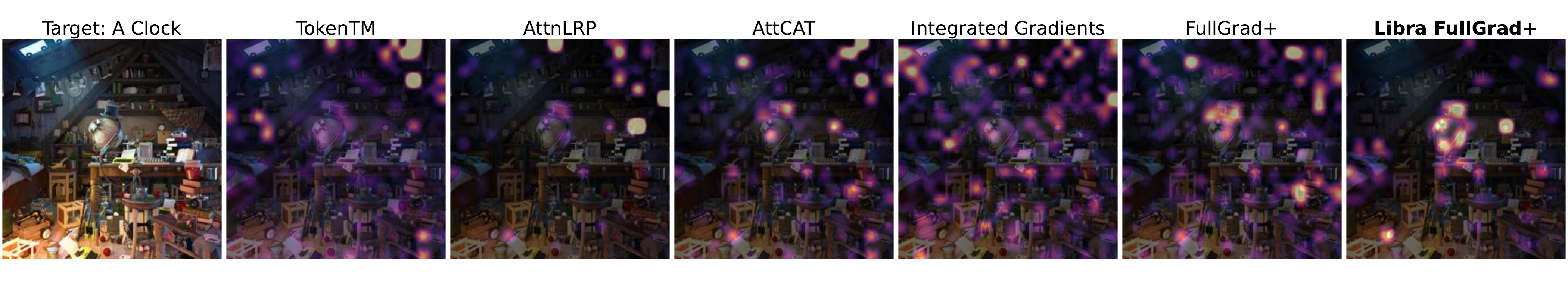,%
}{}{}

\CLIPRows[!b]{%
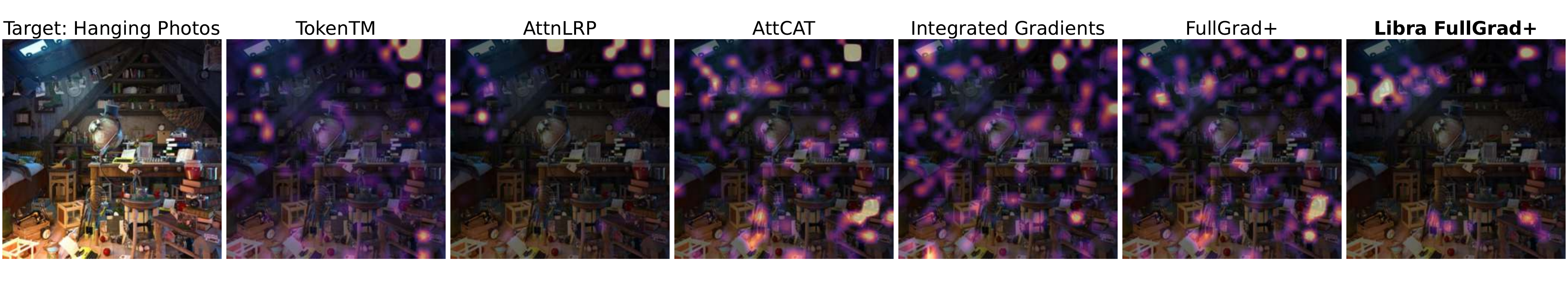,%
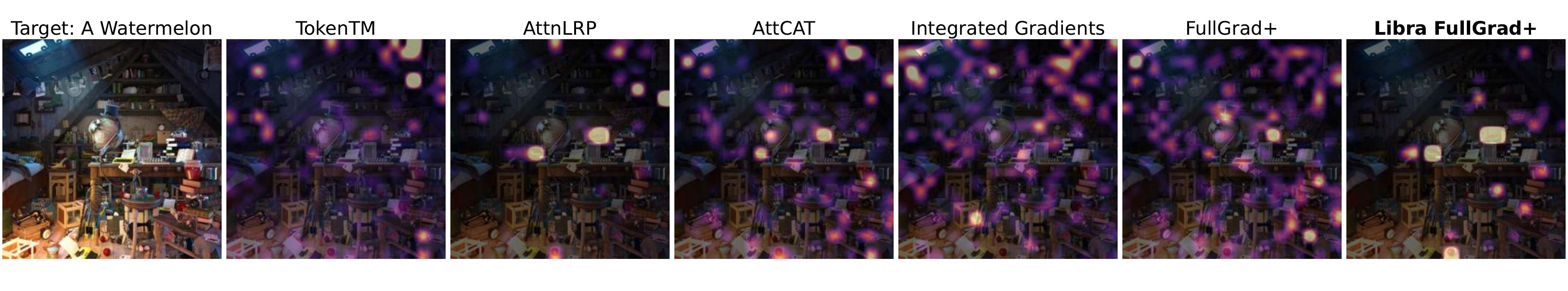,%
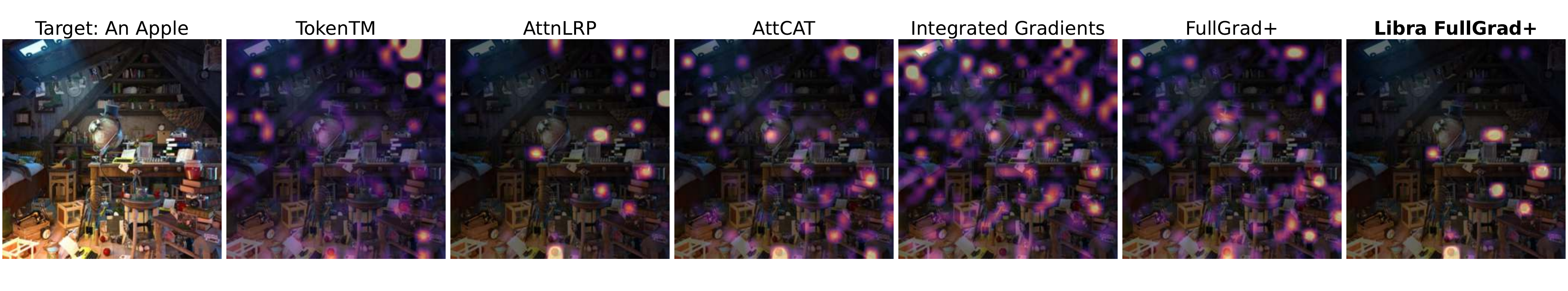,%
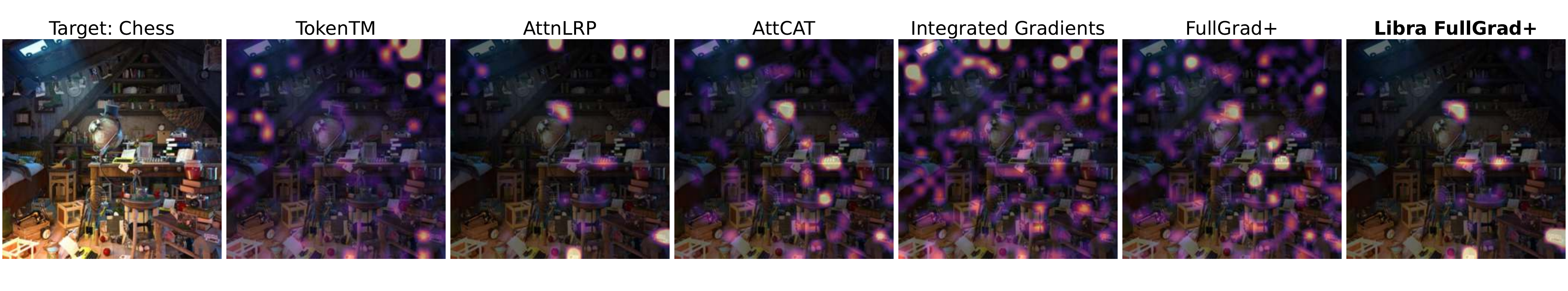,%
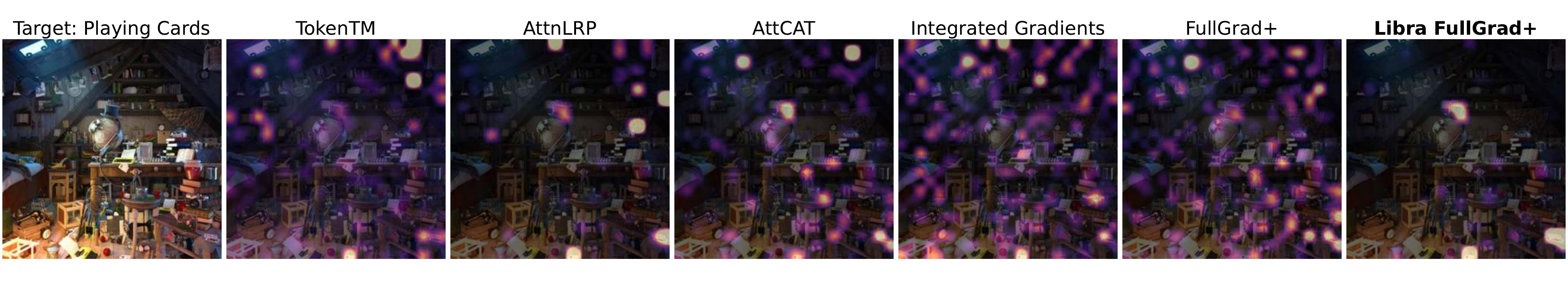,%
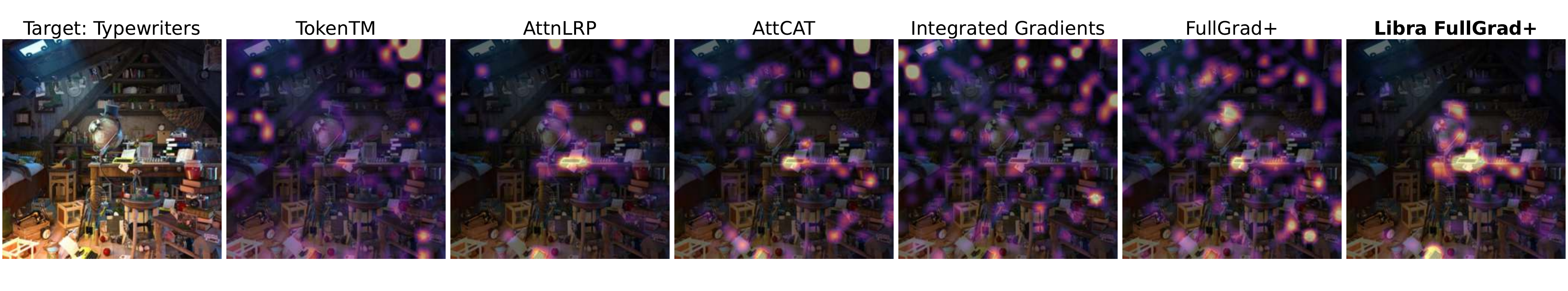,%
}{}{}

\CLIPRows[!b]{%
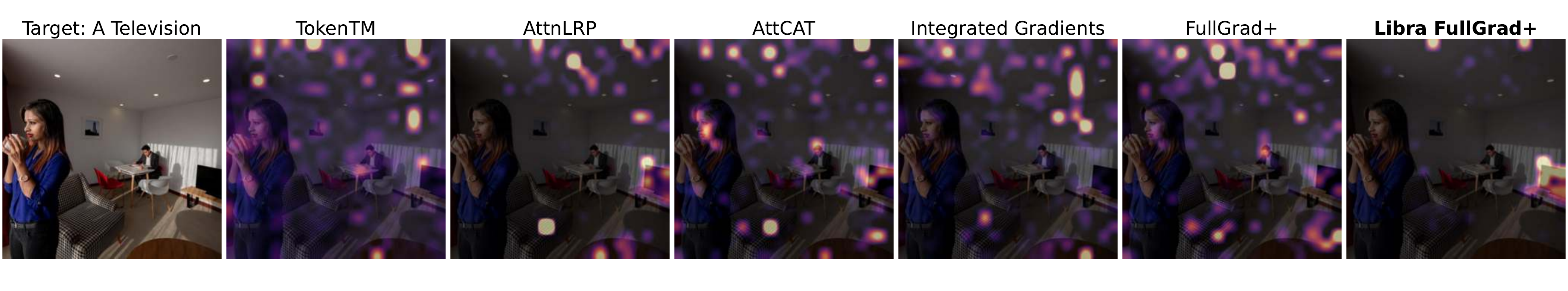,%
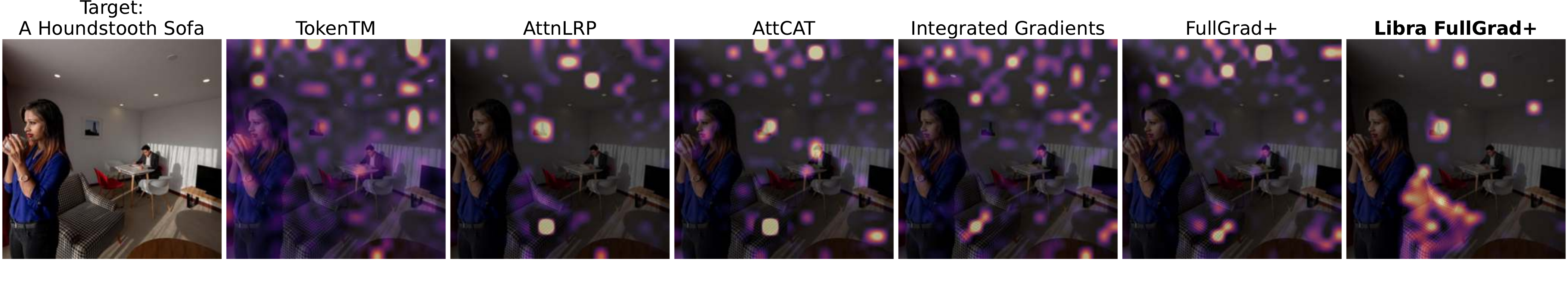,%
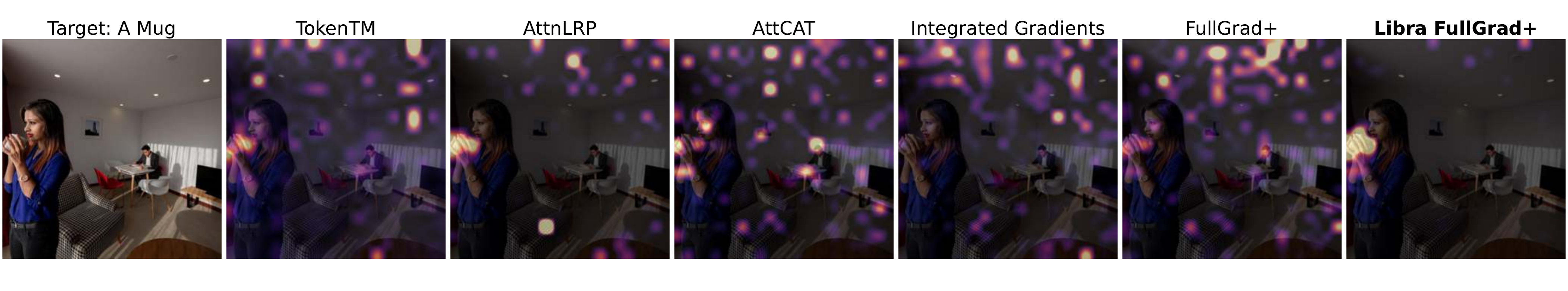,%
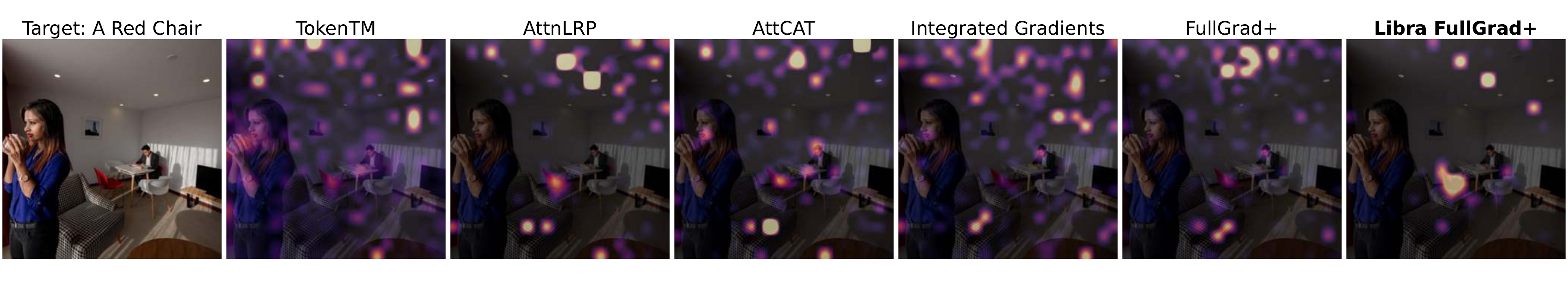,%
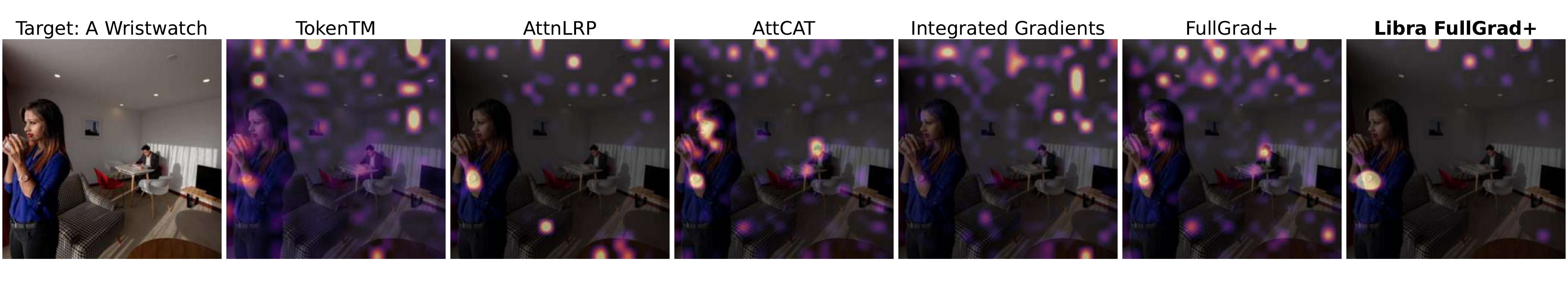,%
}{}{}

\CLIPRows[!b]{%
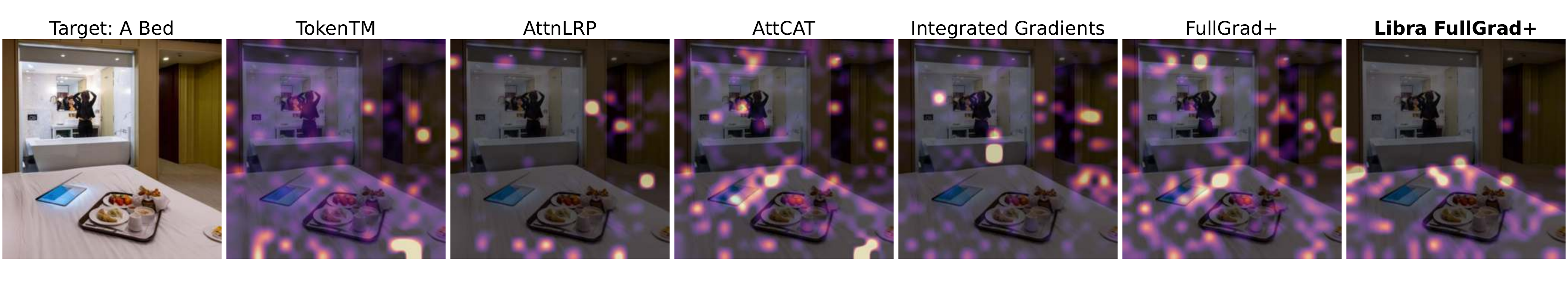,%
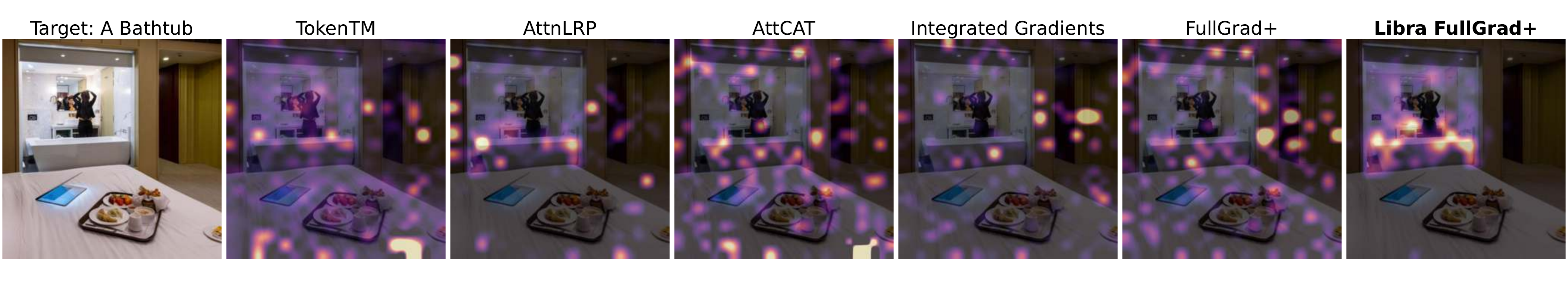,%
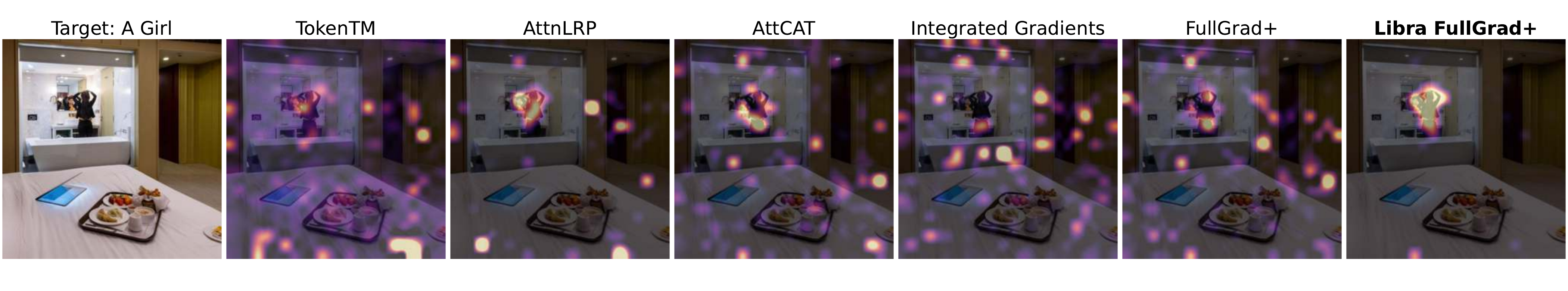,%
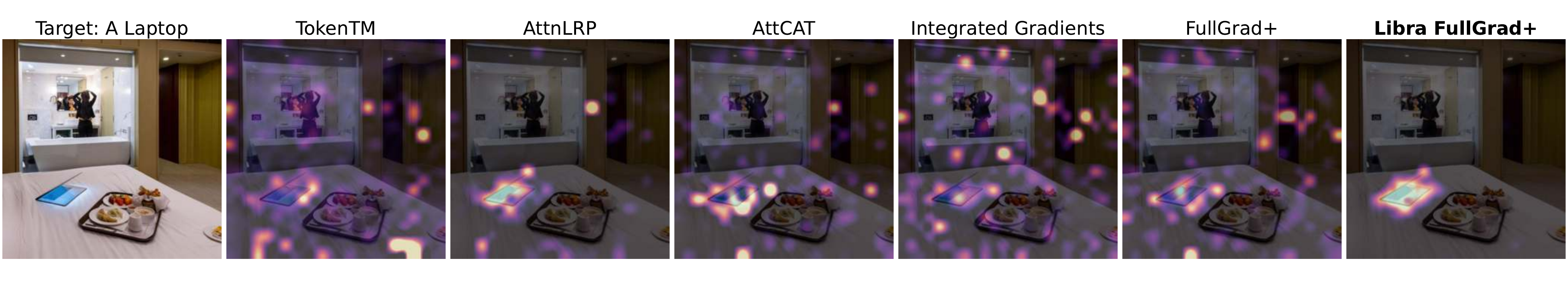,%
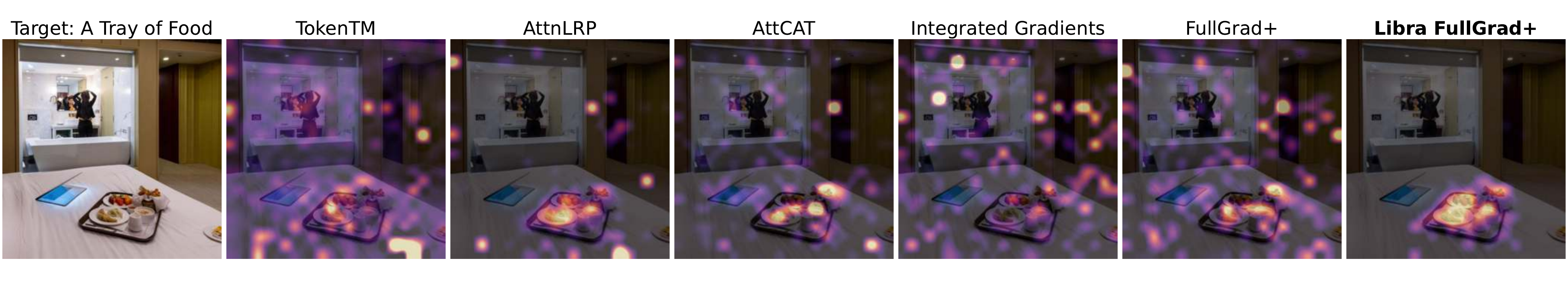,%
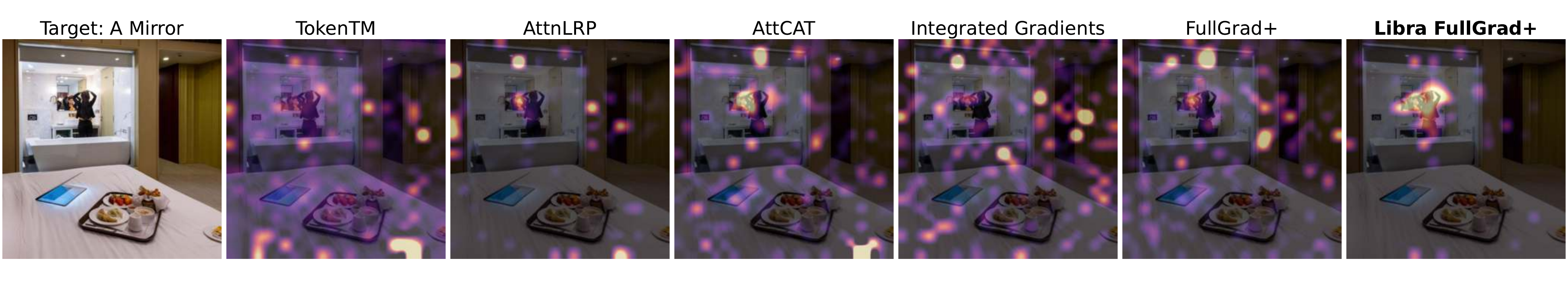,%
}{}{}

\CLIPRows[!b]{%
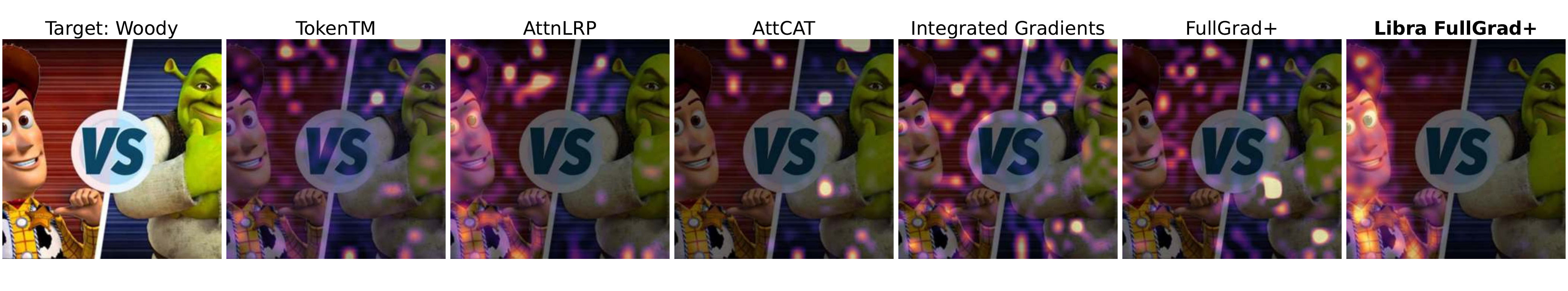,%
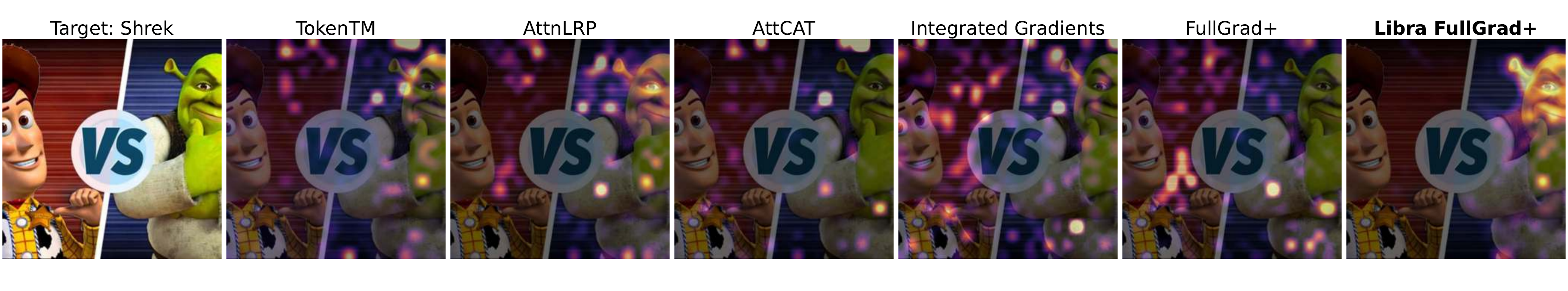,%
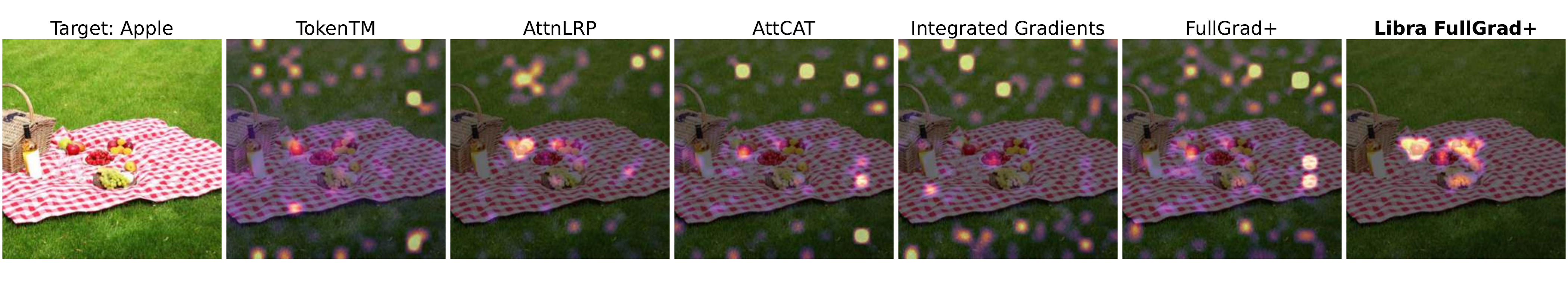,%
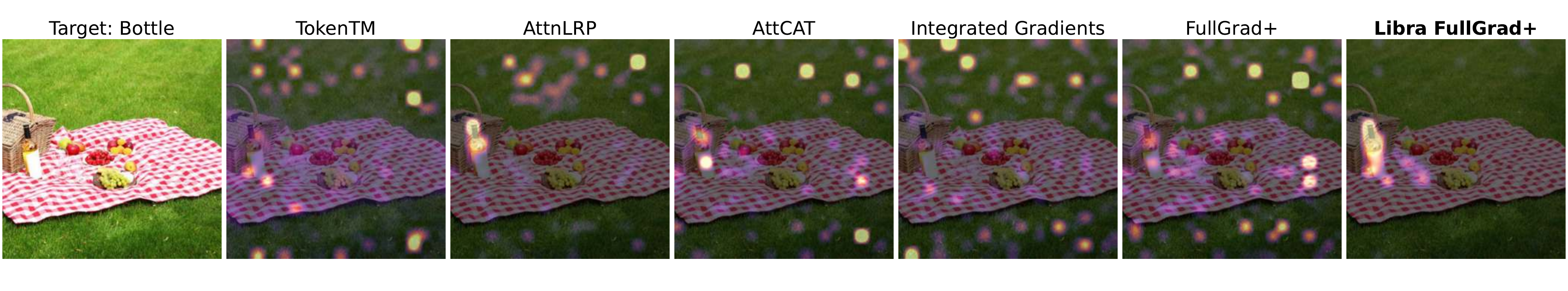,%
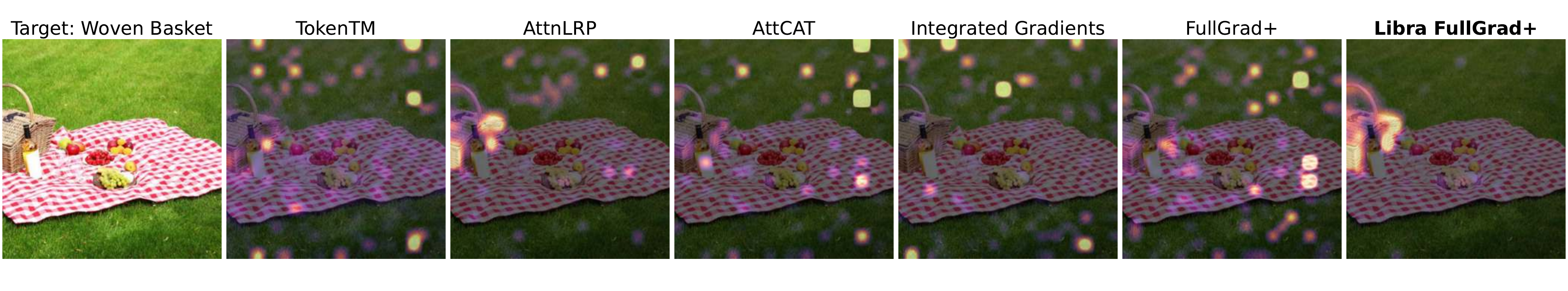,%
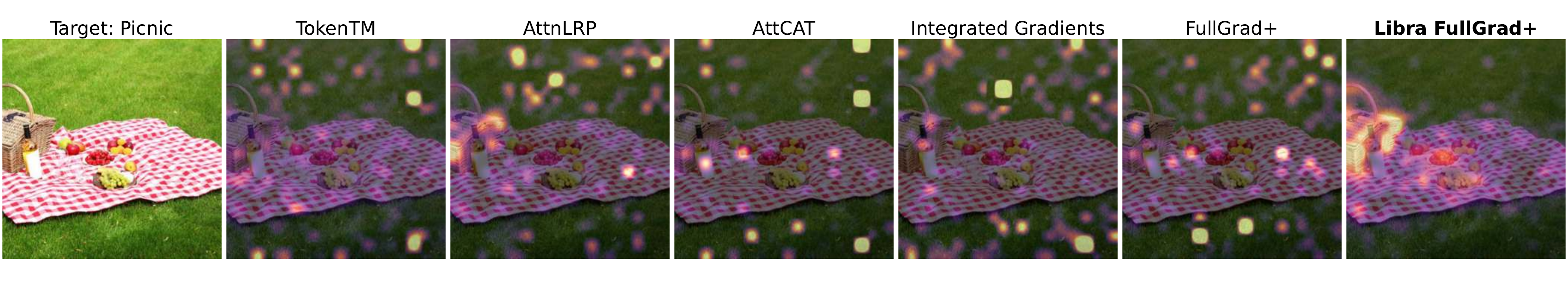,%
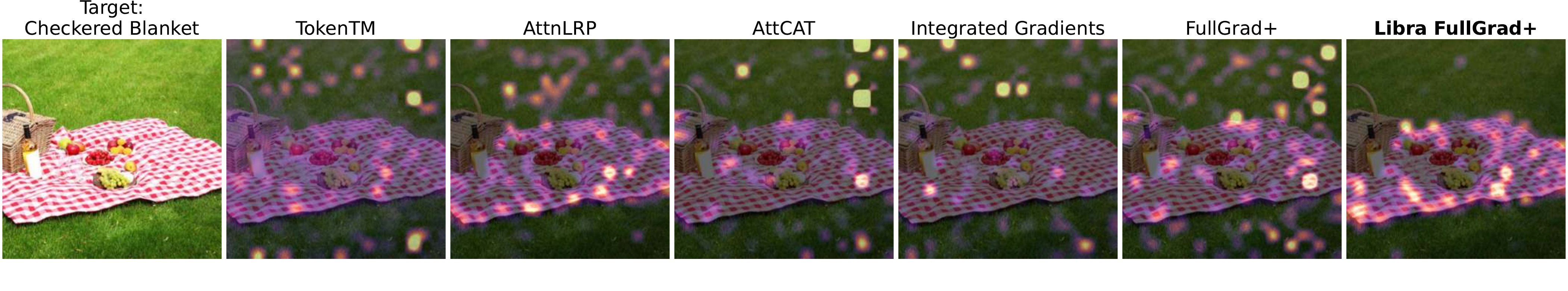,%
}{}{}

\CLIPRows[!b]{%
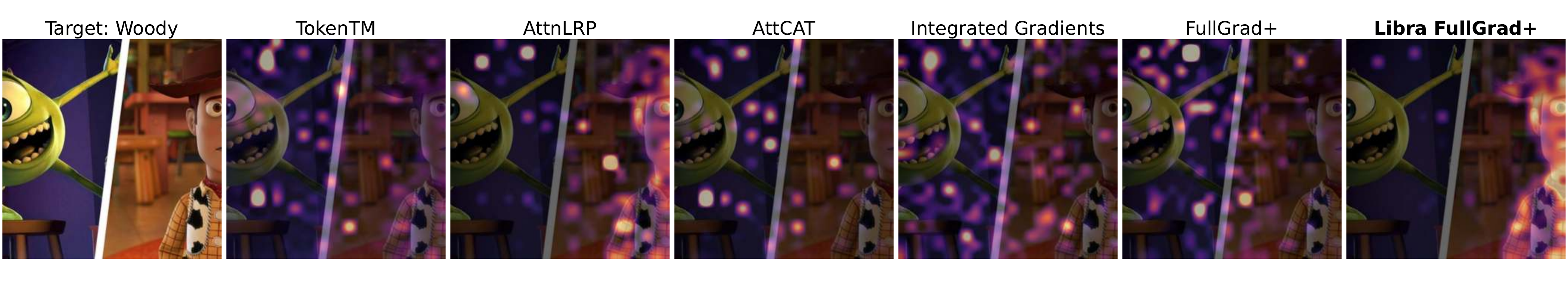,%
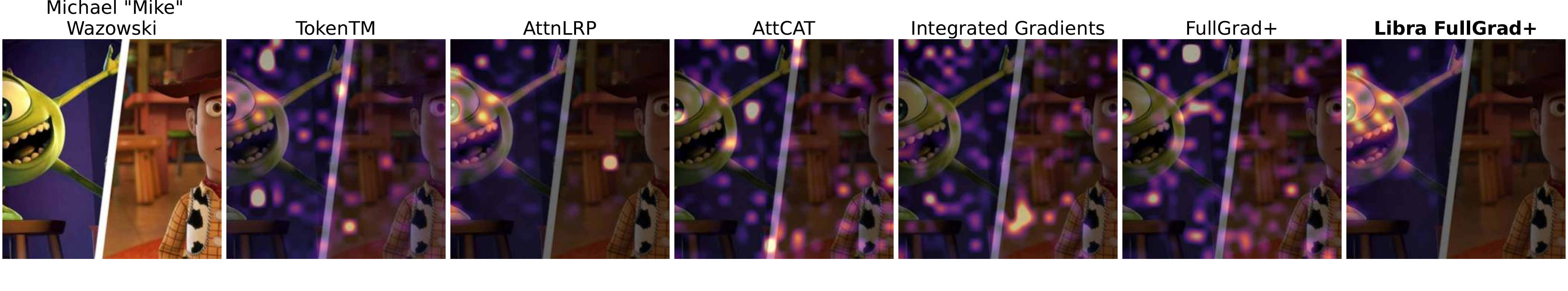,%
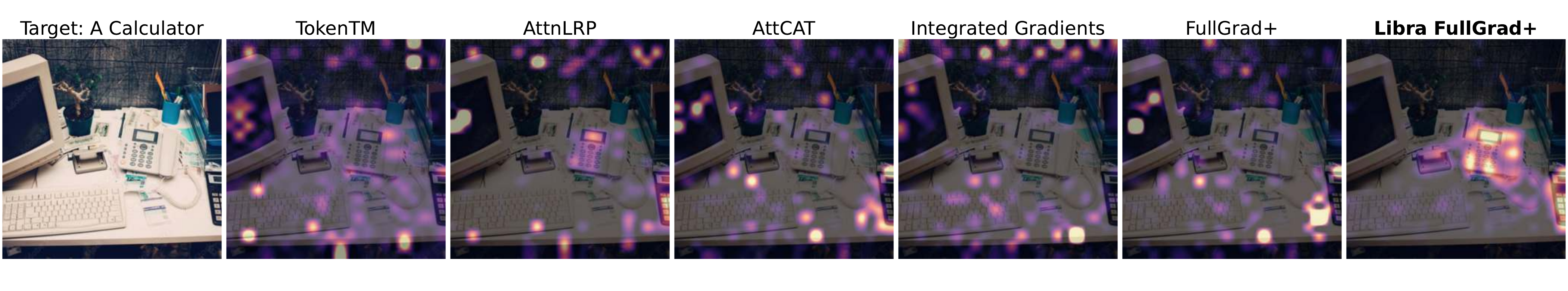,%
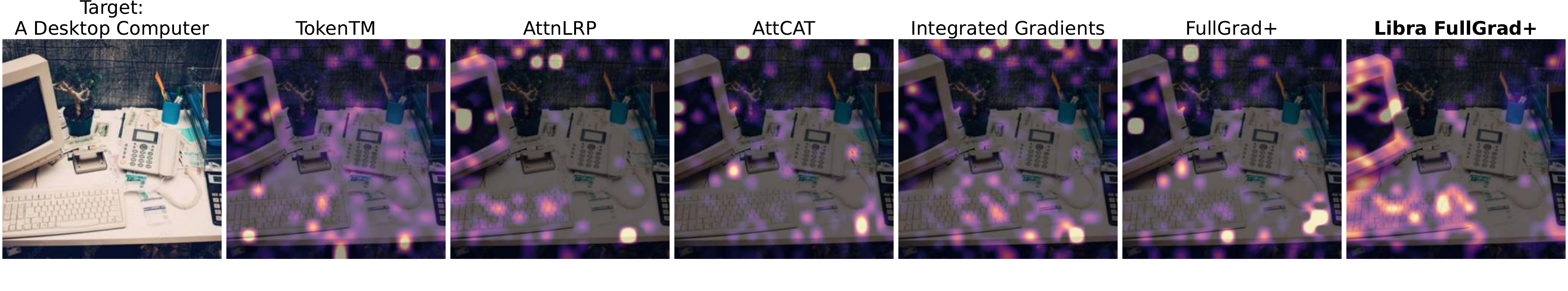,%
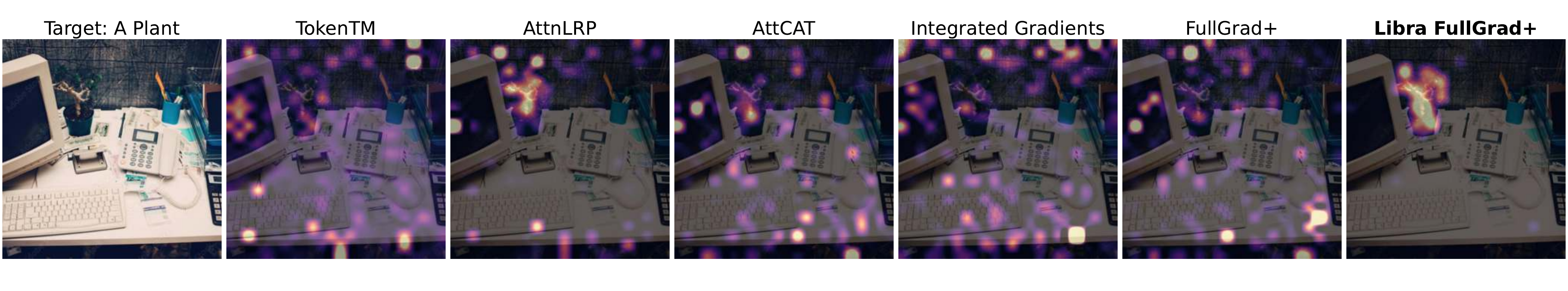,%
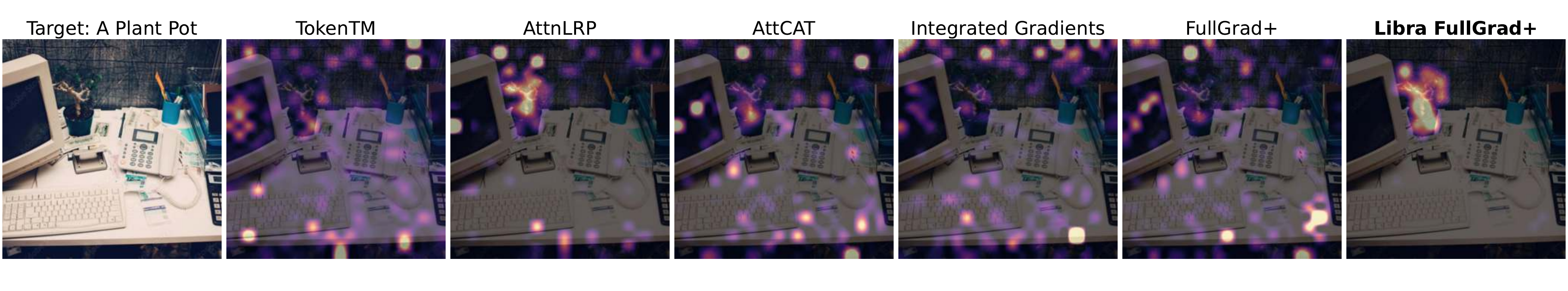,%
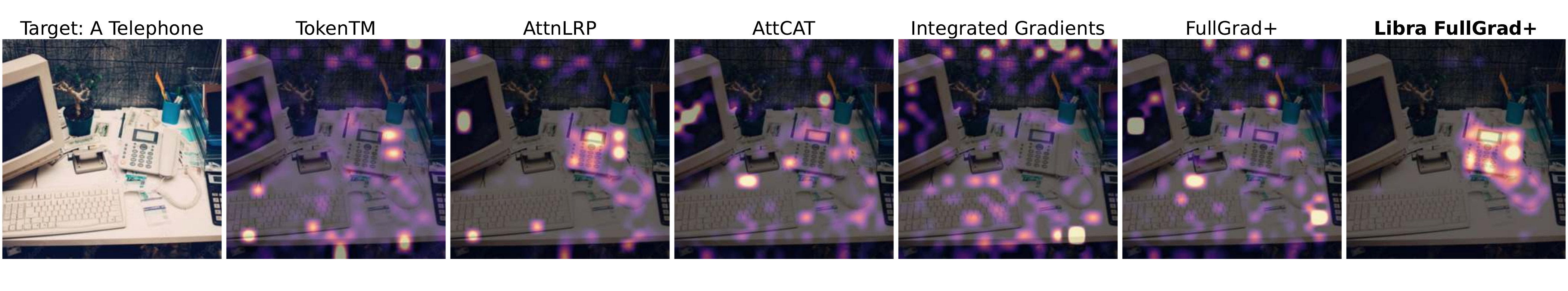,%
}{}{}

\CLIPRows[!b]{%
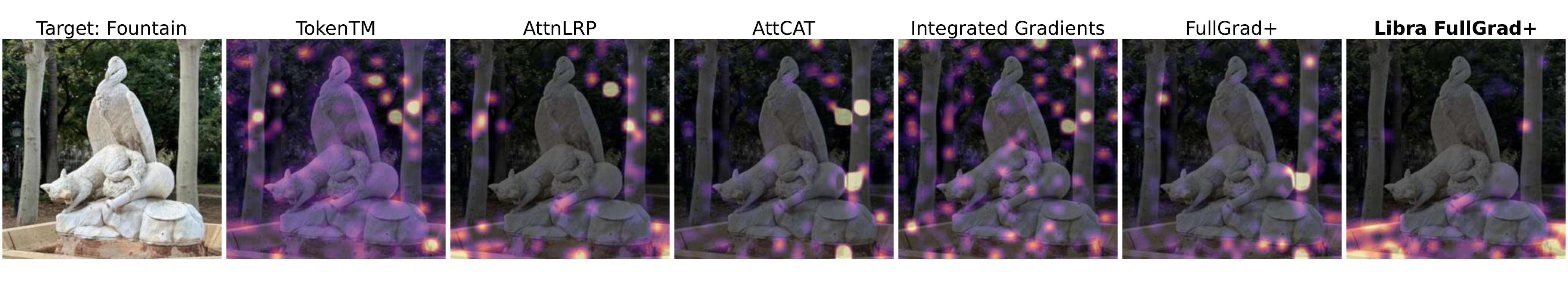,%
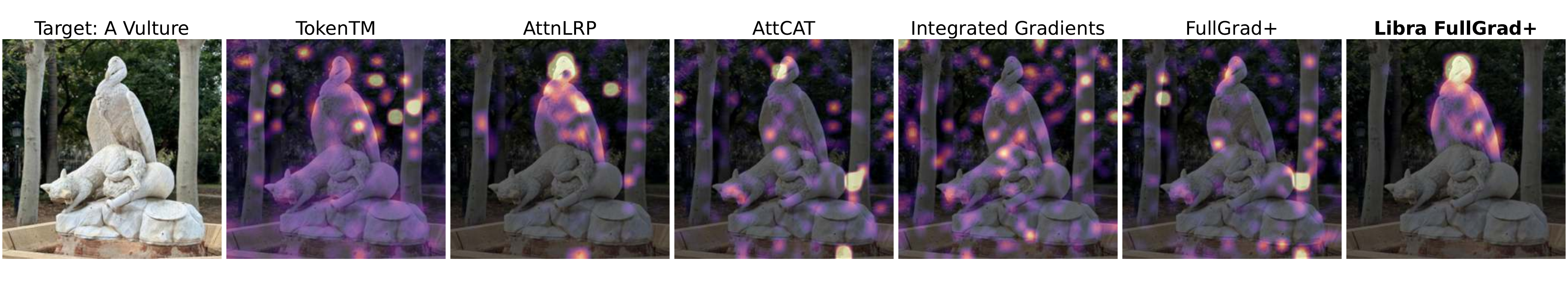,%
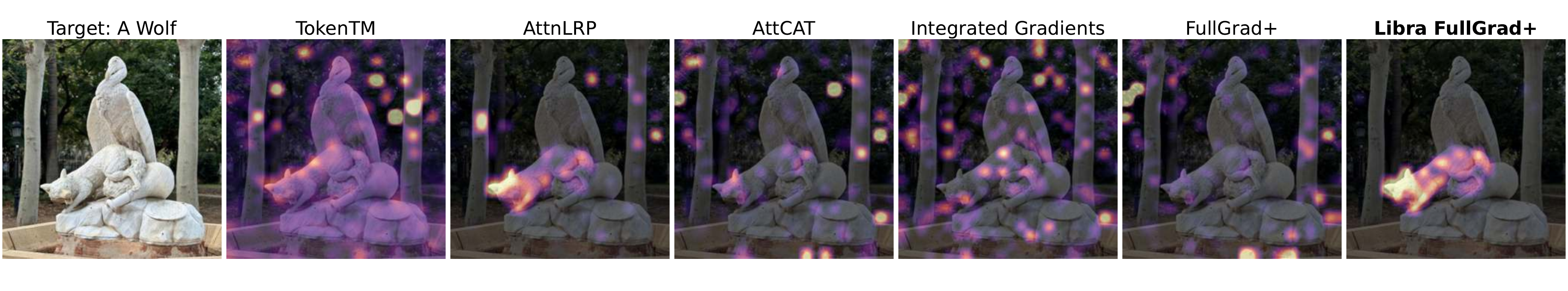,%
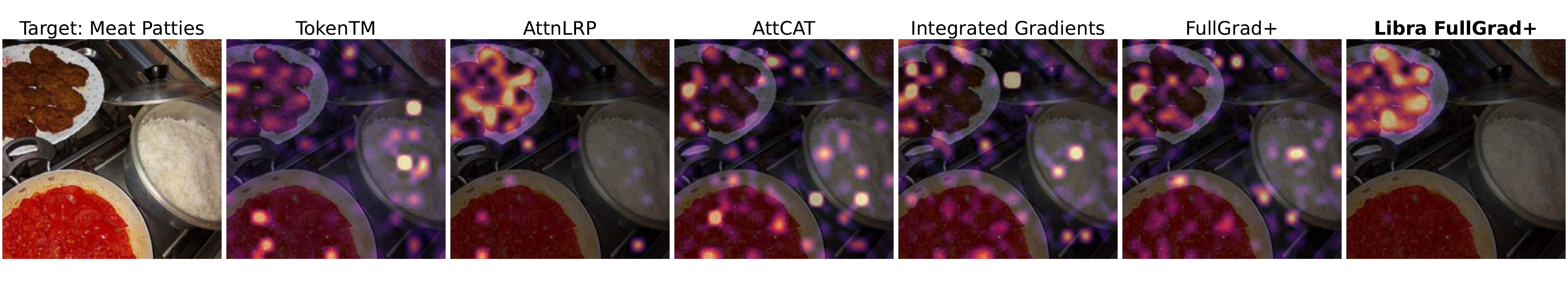,%
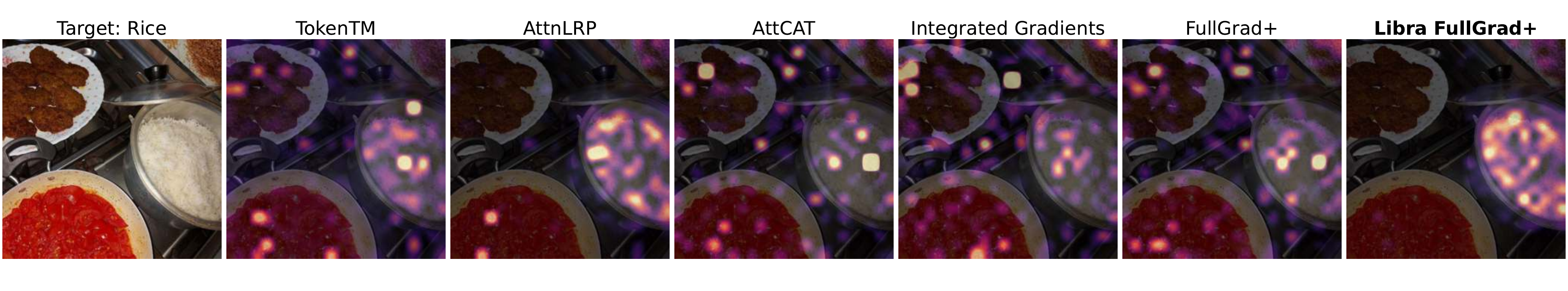,%
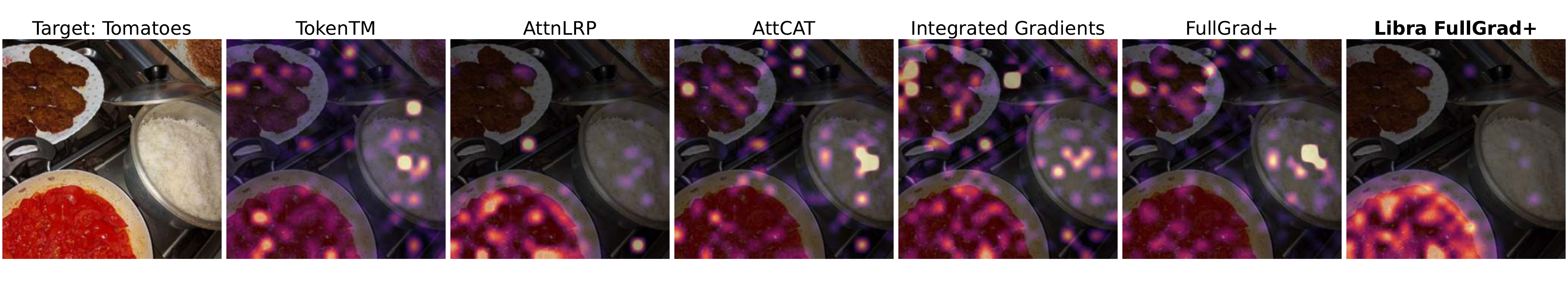,%
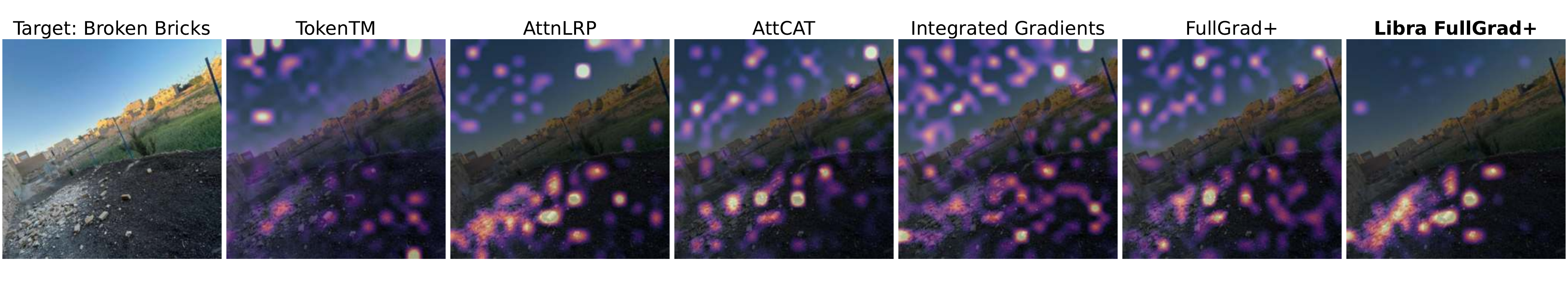,%
}{}{}

\CLIPRows[!b]{%
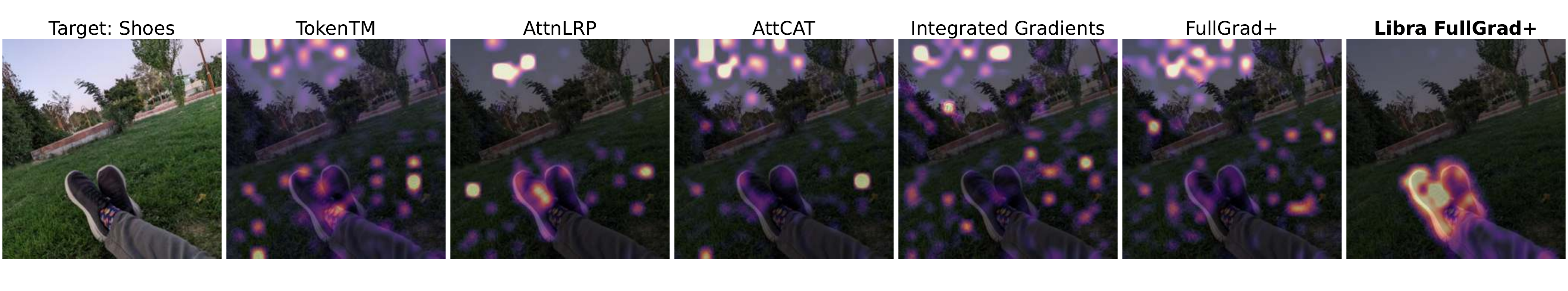,%
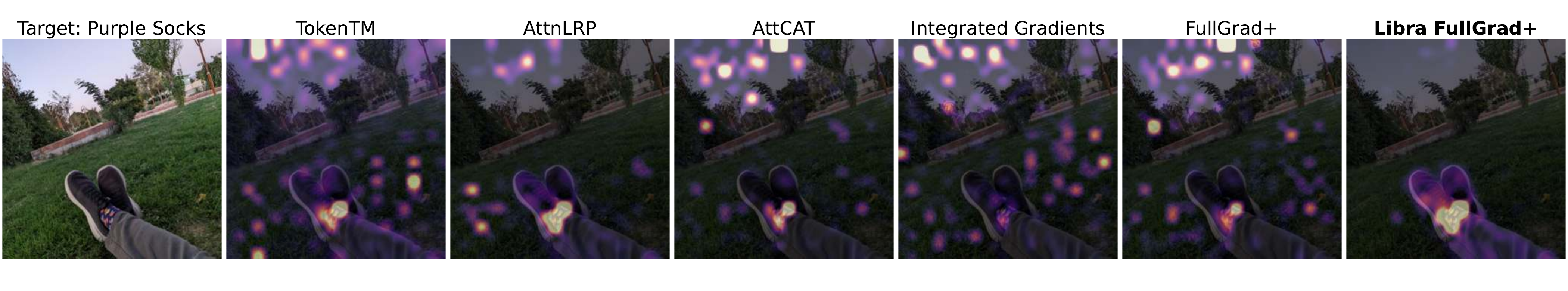,%
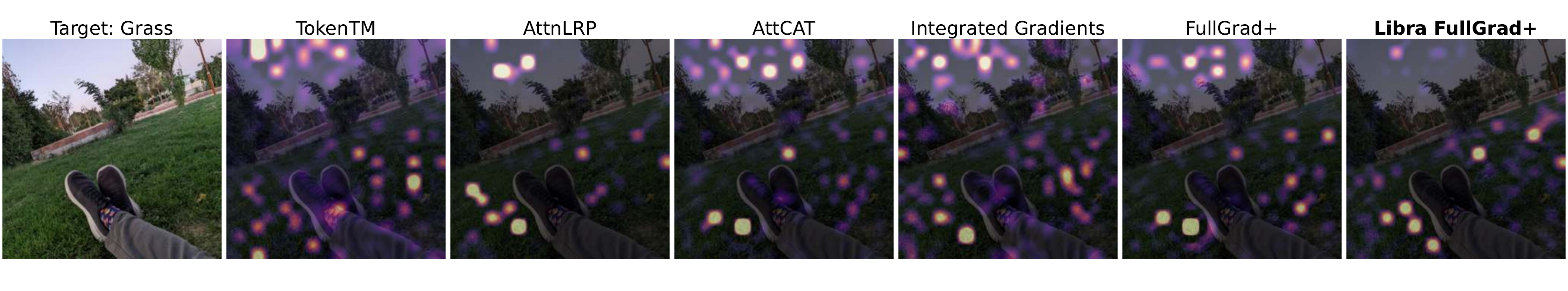,%
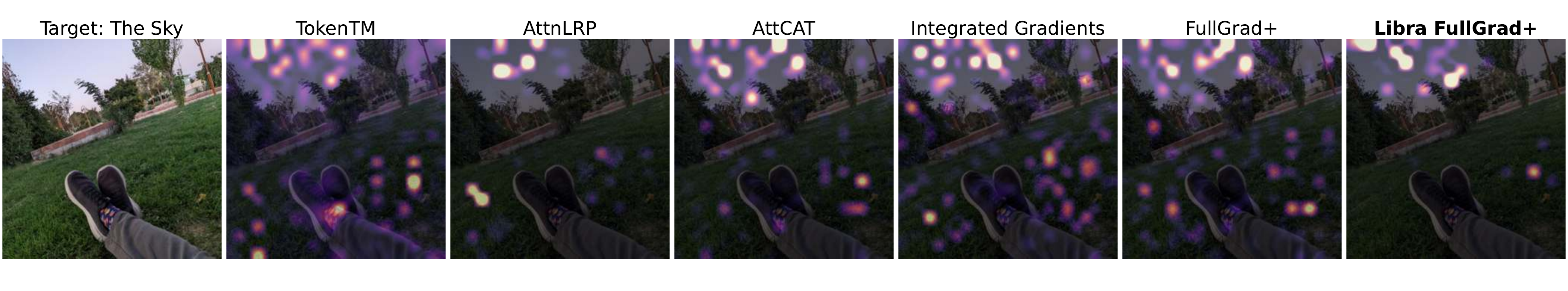,%
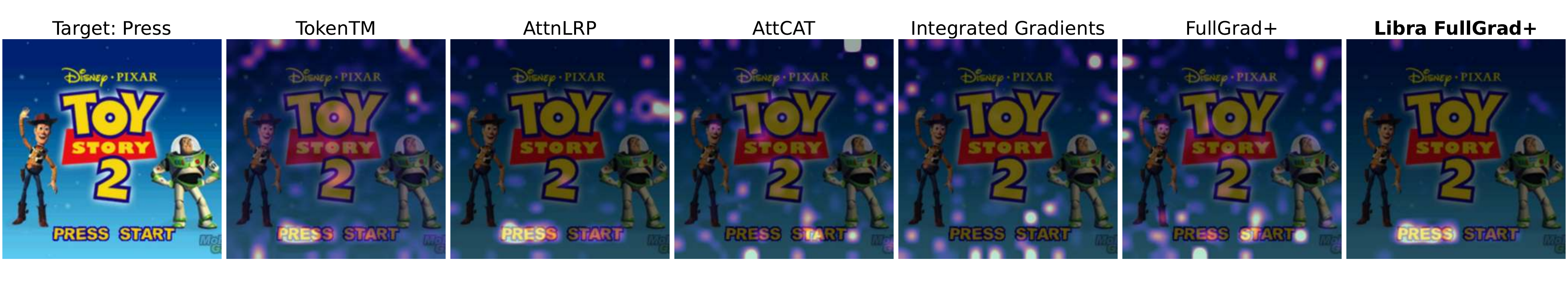,%
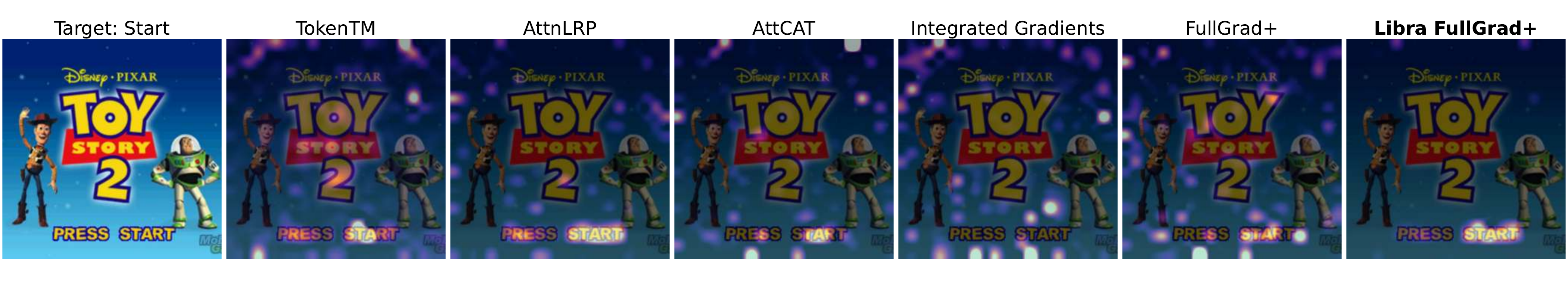,%
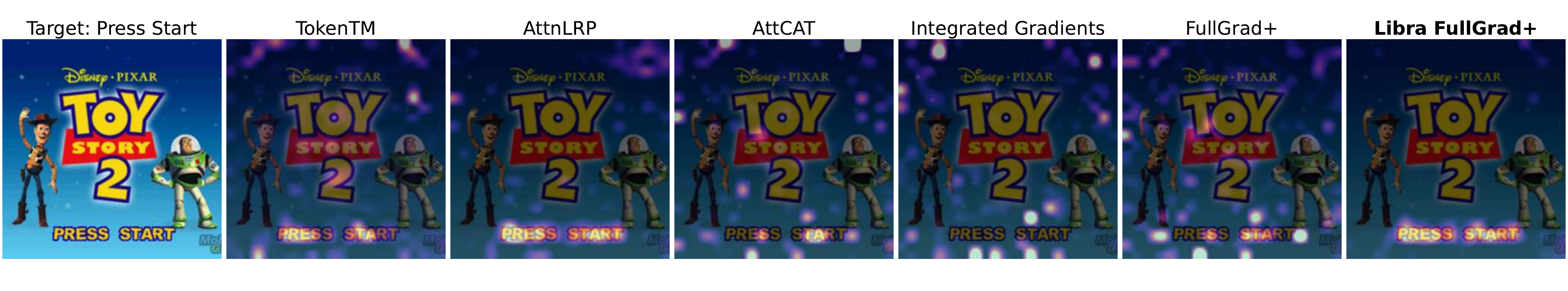,%
}{}{}

\CLIPRows[!b]{%
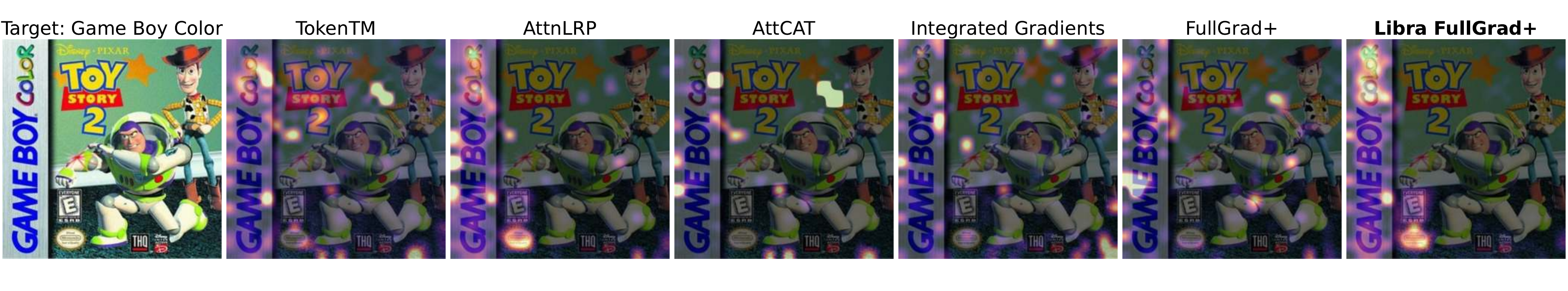,%
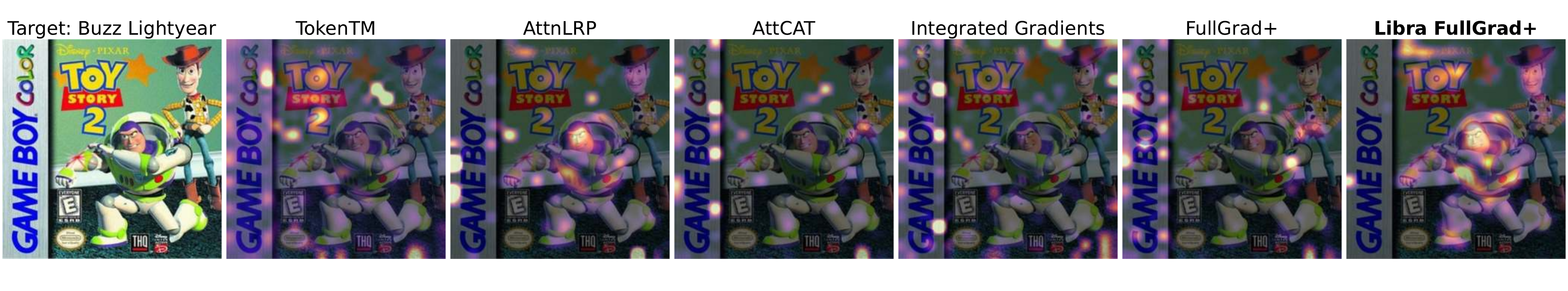,%
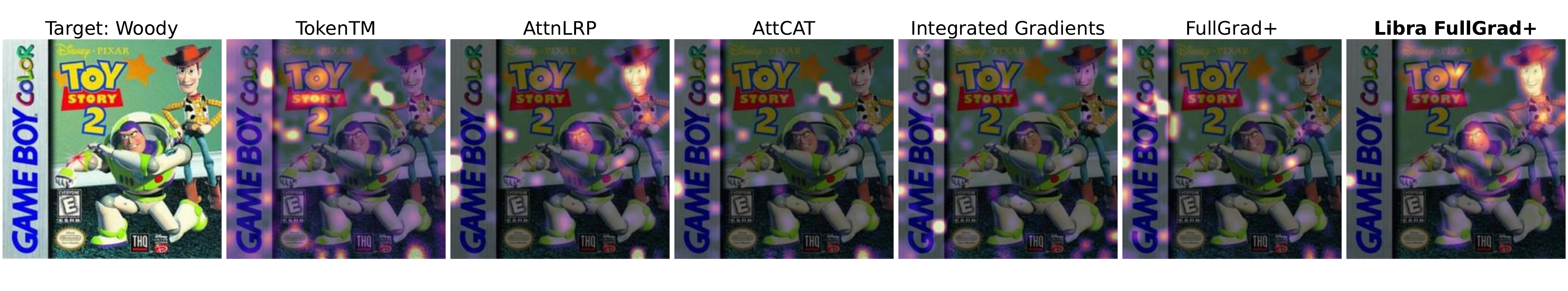,%
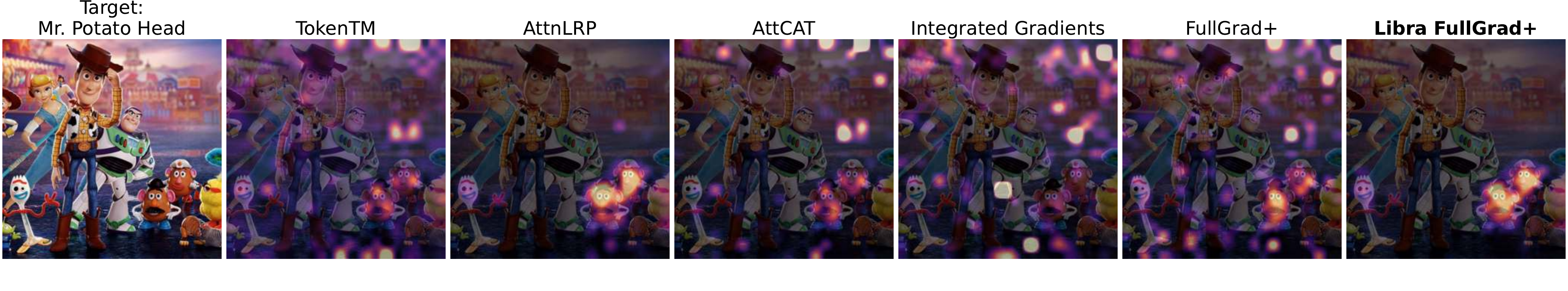,%
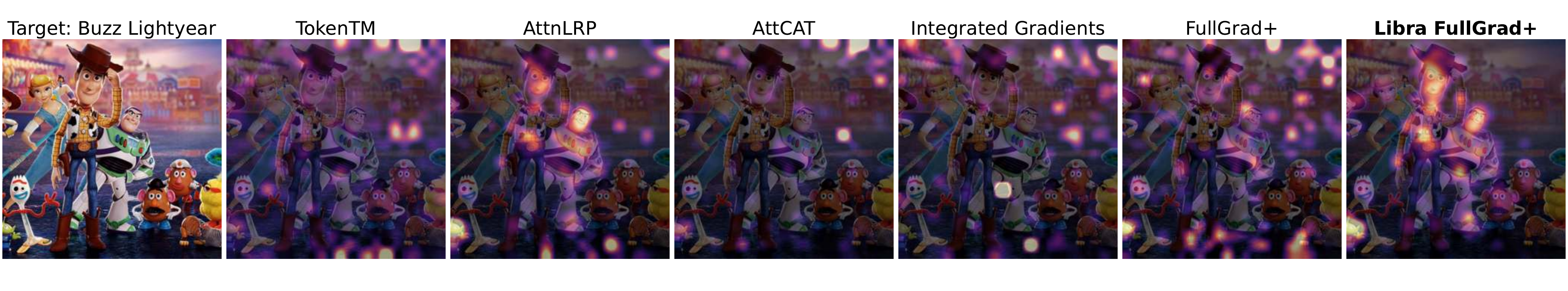,%
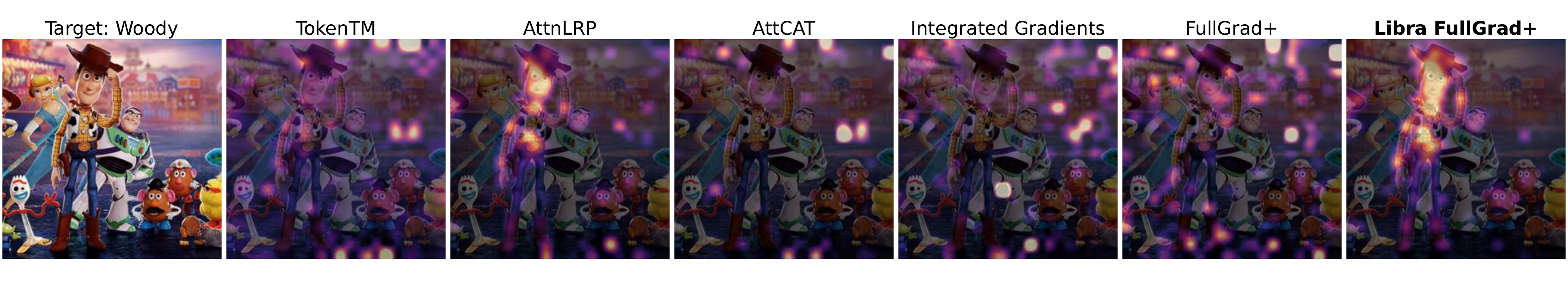,%
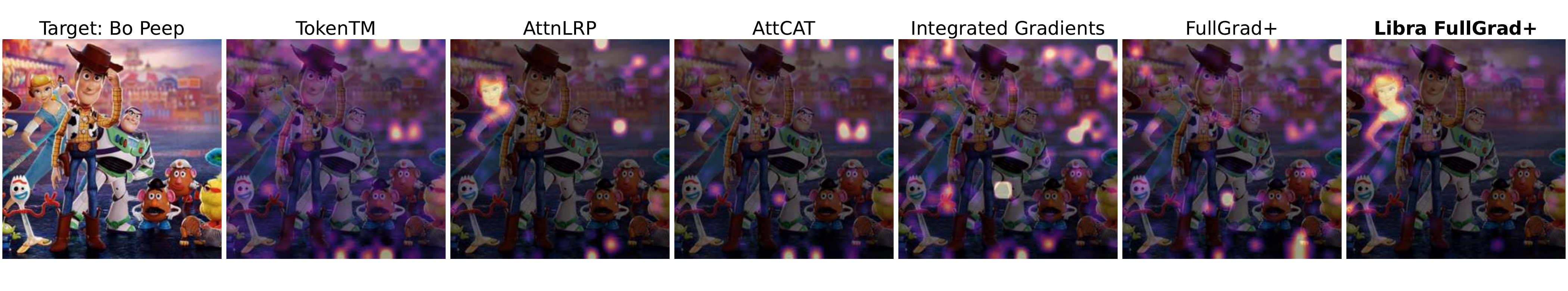,%
}{}{}

\CLIPRows[!b]{%
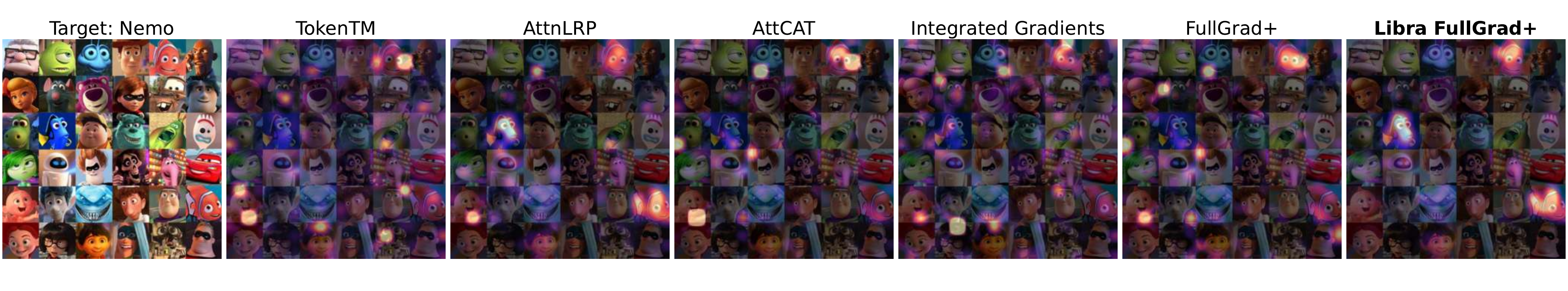,%
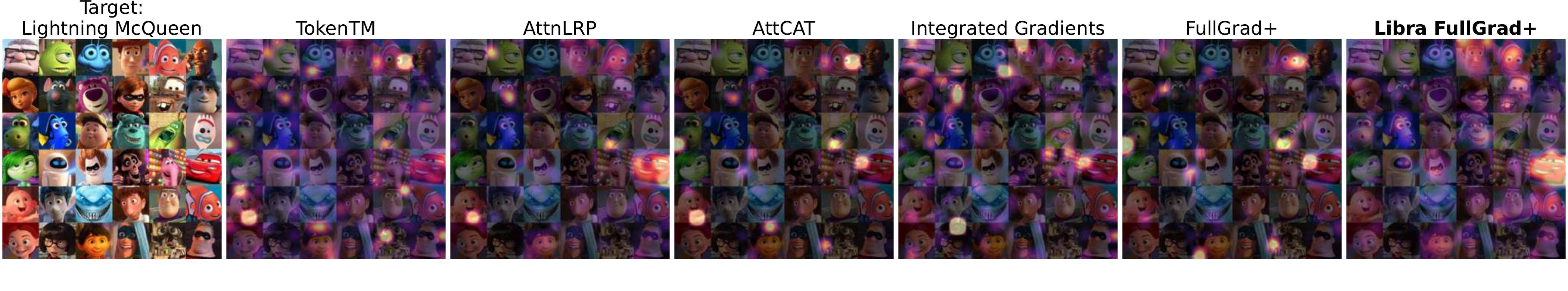,%
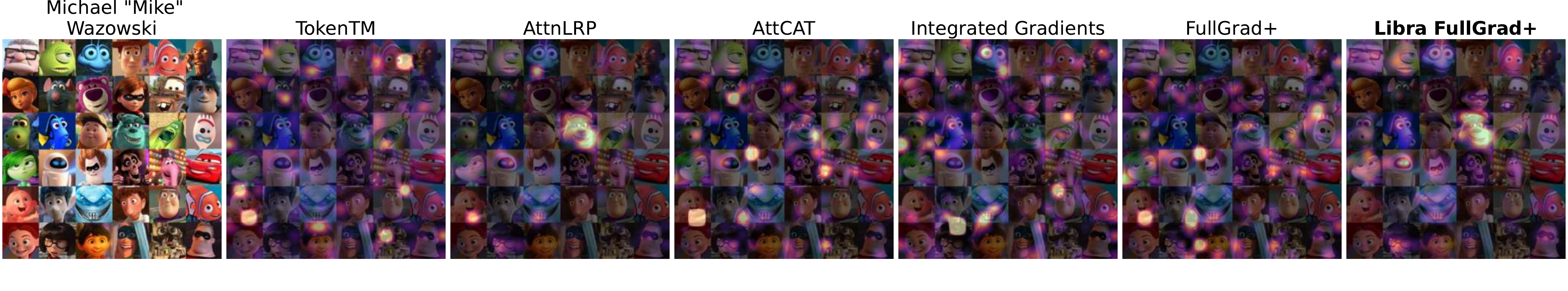,%
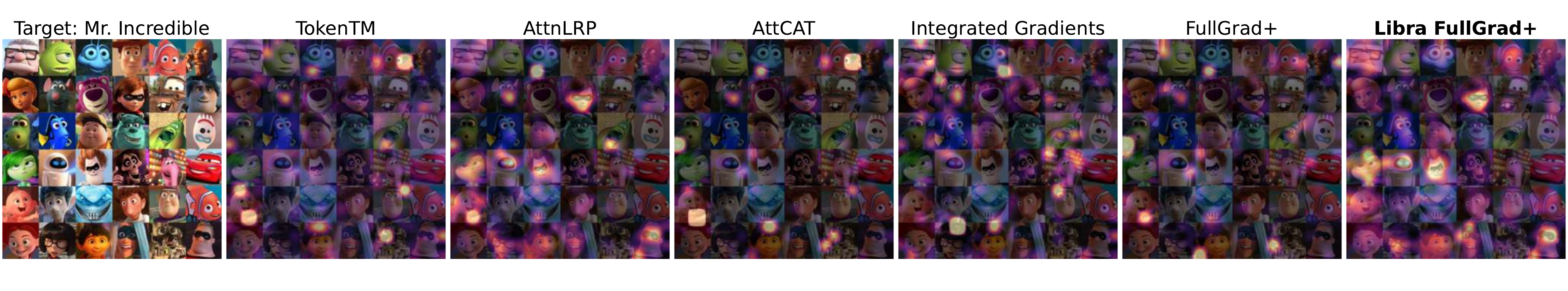,%
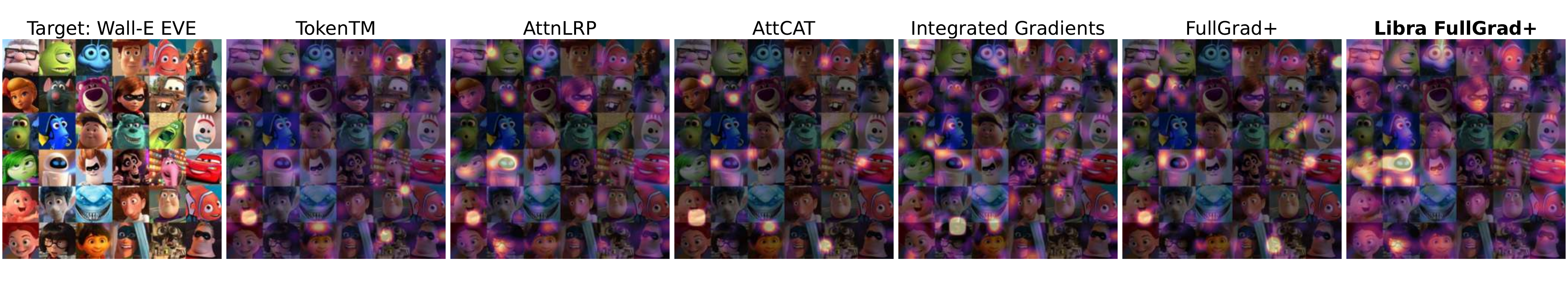,%
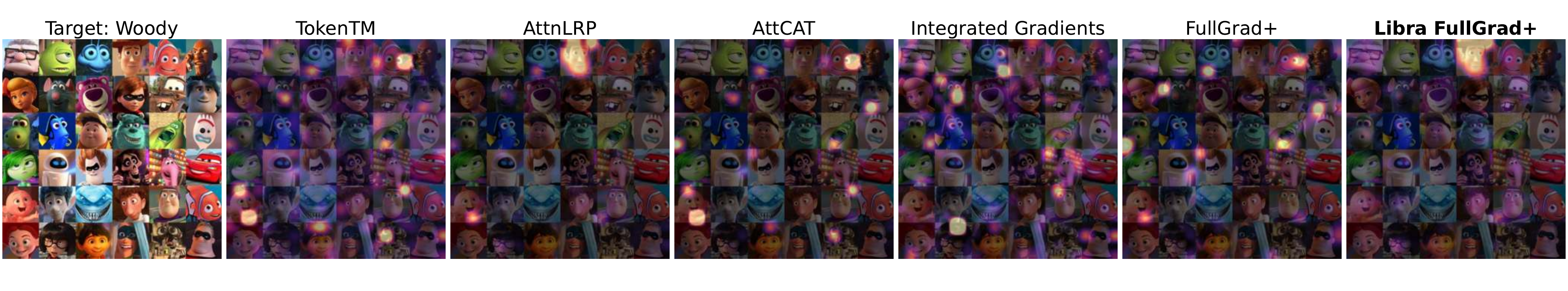,%
}{}{}

\CLIPRows[!b]{%
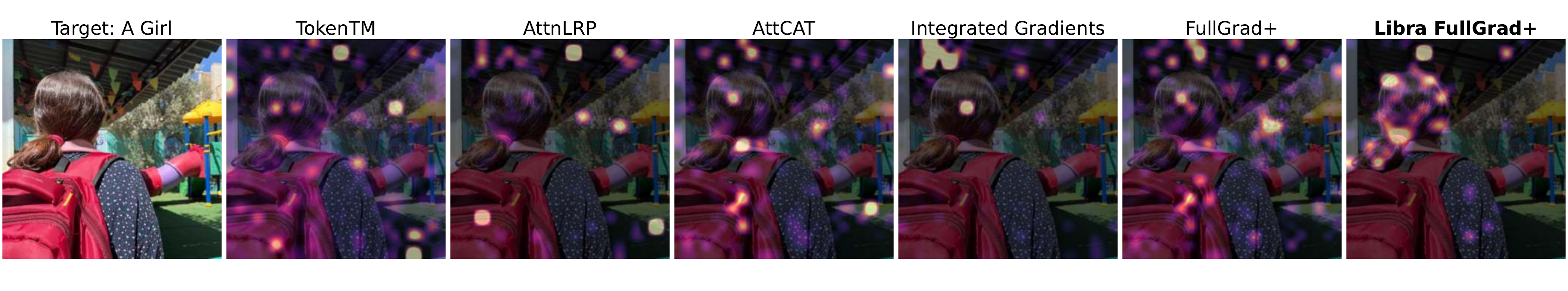,%
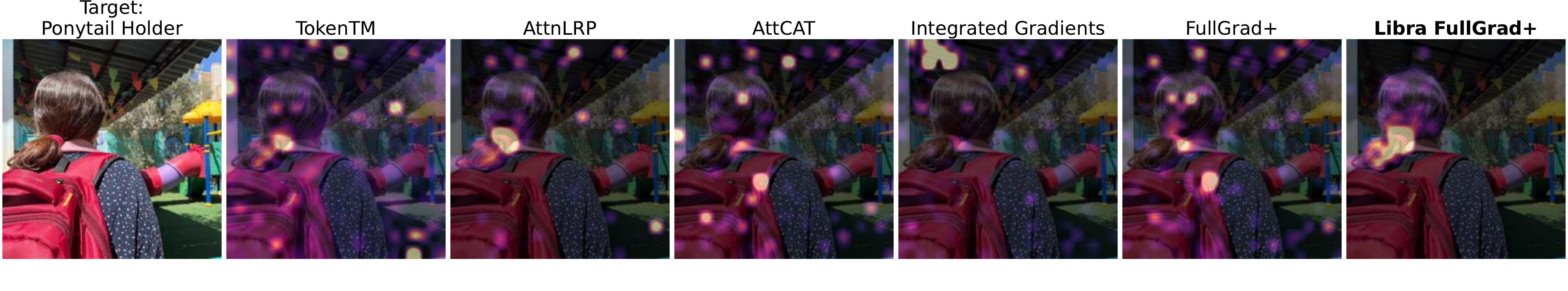,%
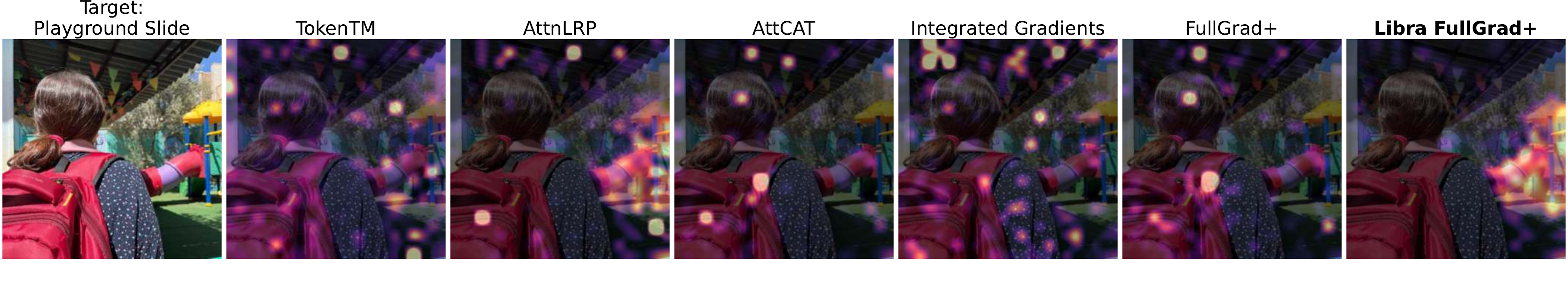,%
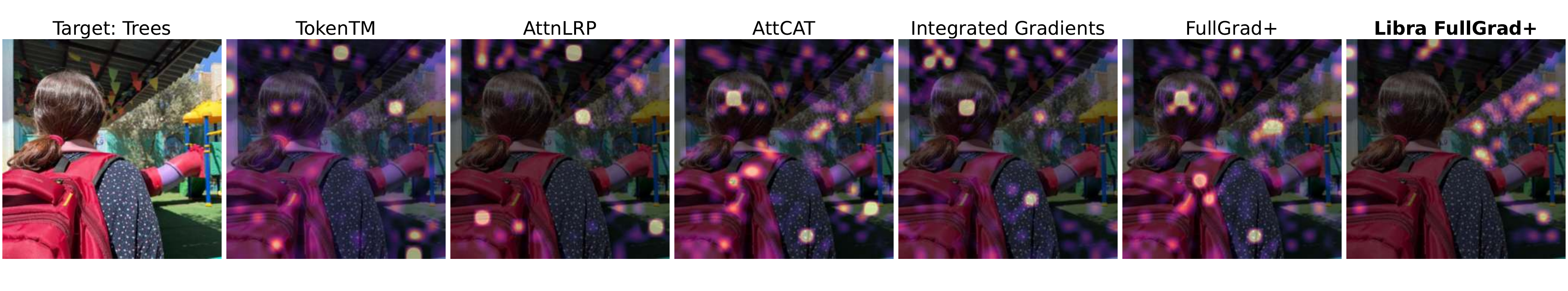,%
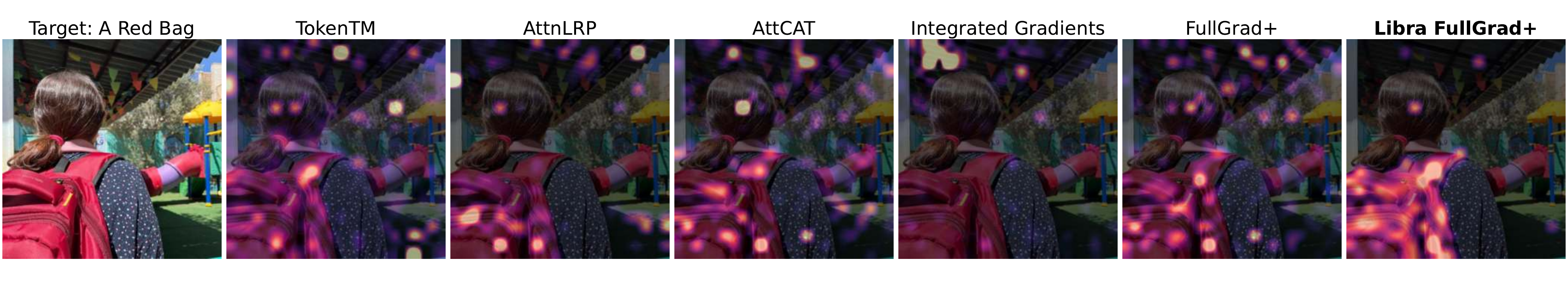,%
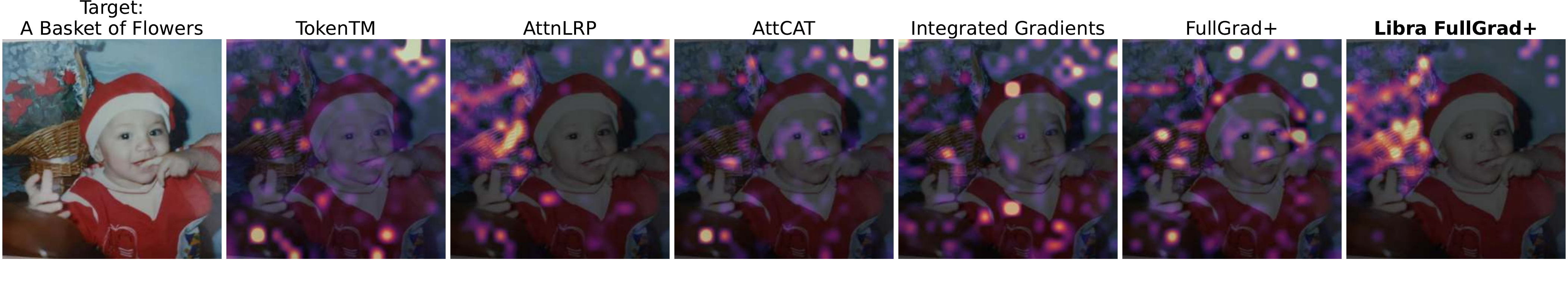,%
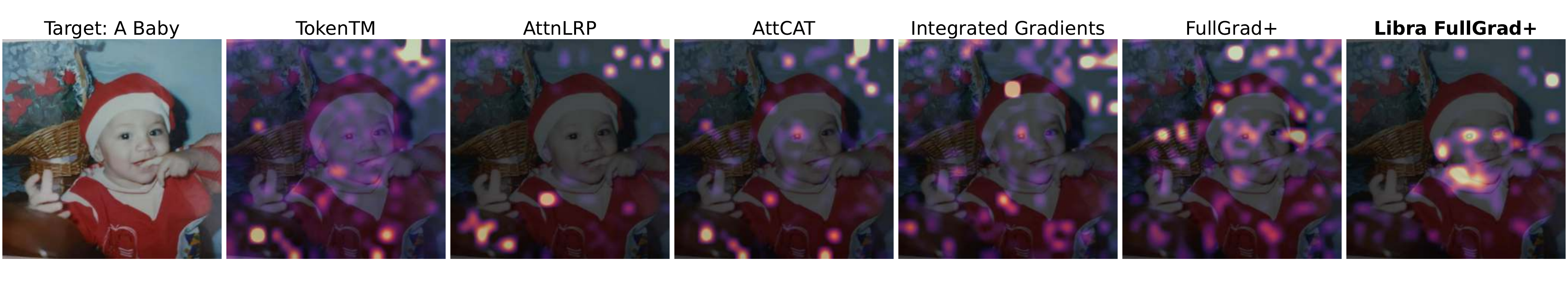,%
}{}{}

\CLIPRows[!b]{%
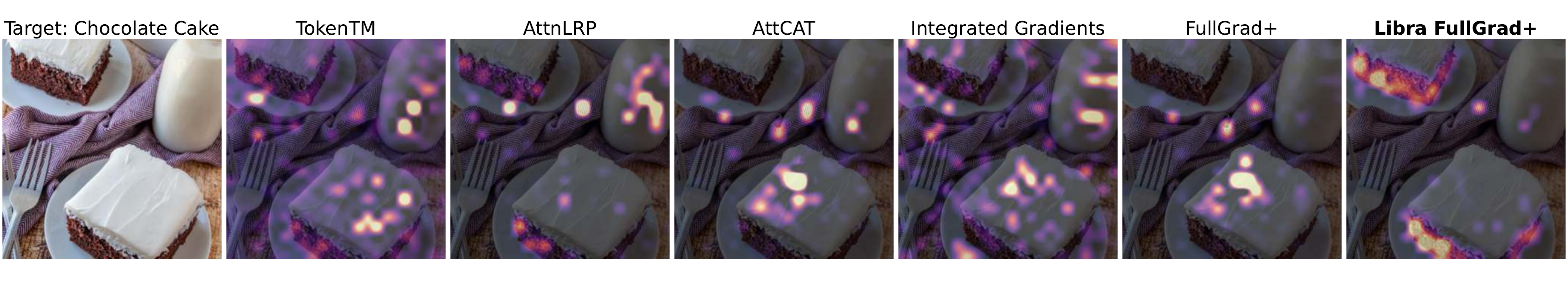,%
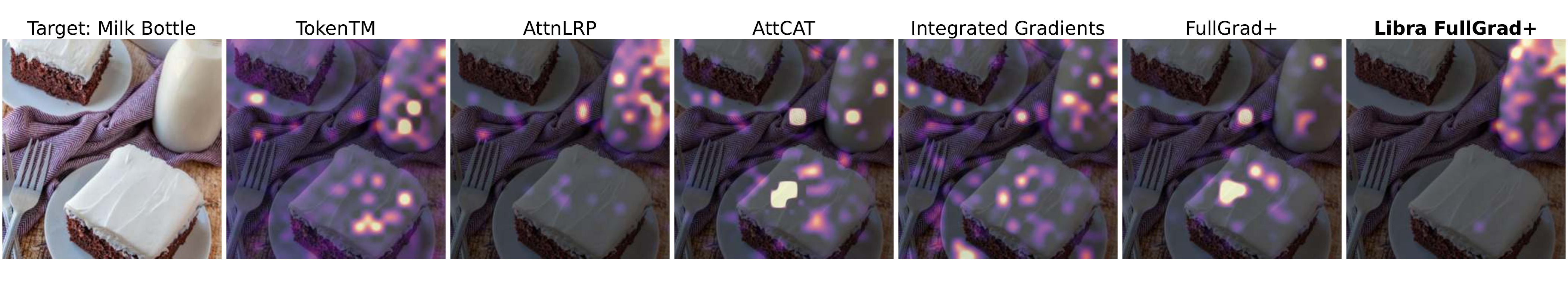,%
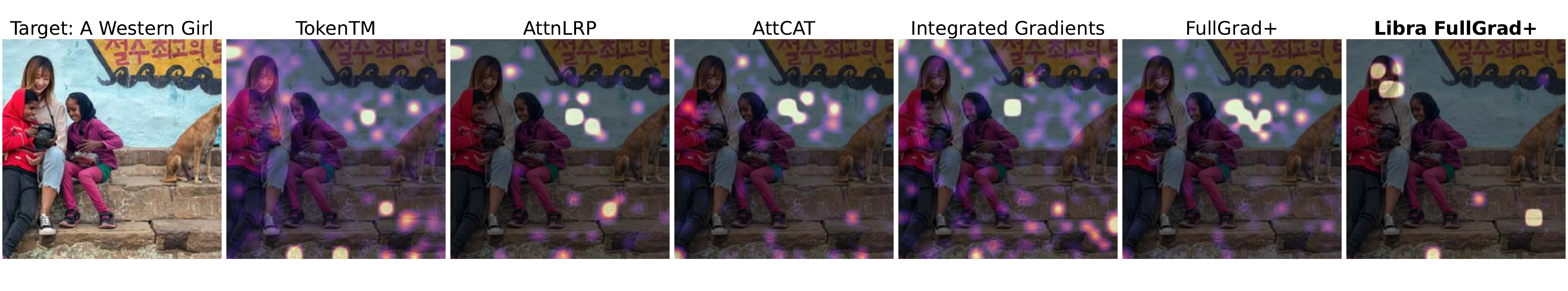,%
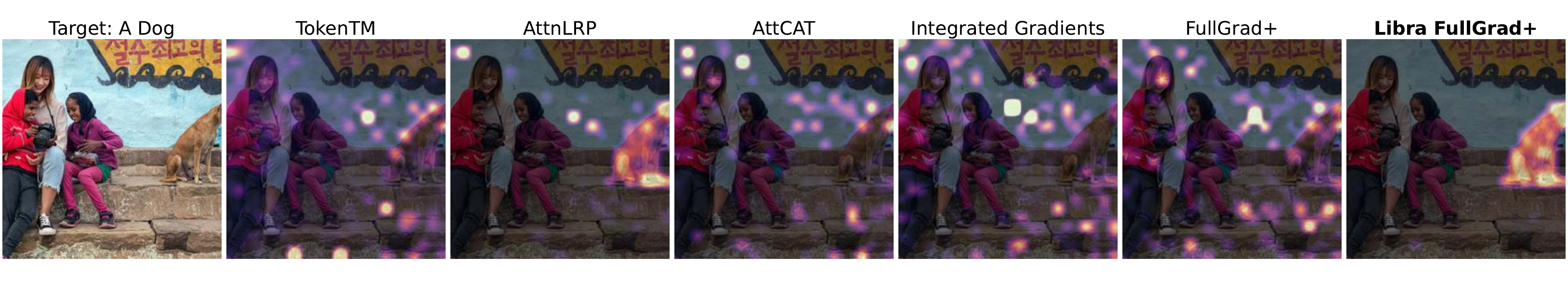,%
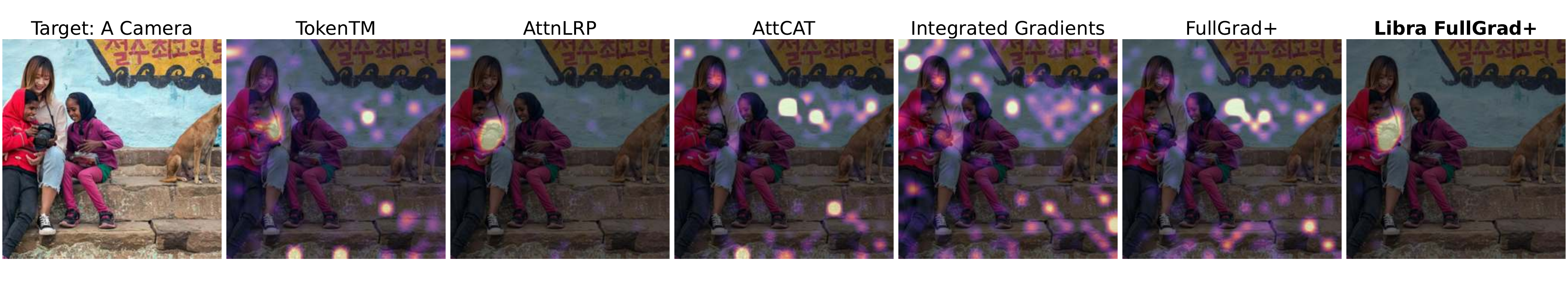,%
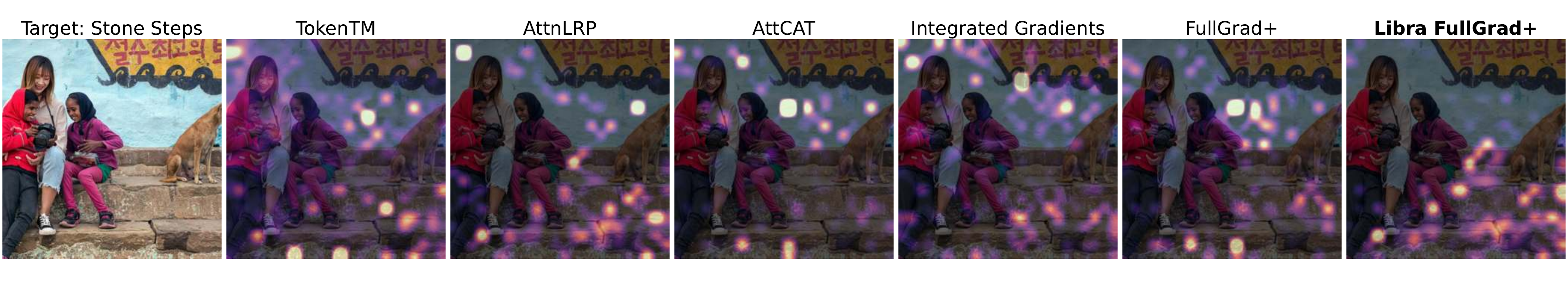,%
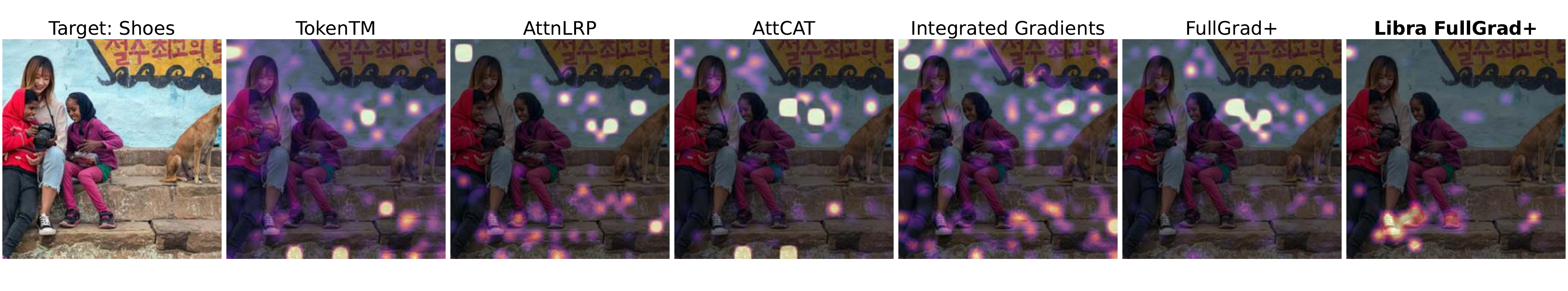,%
}{}{}

\CLIPRows[!b]{%
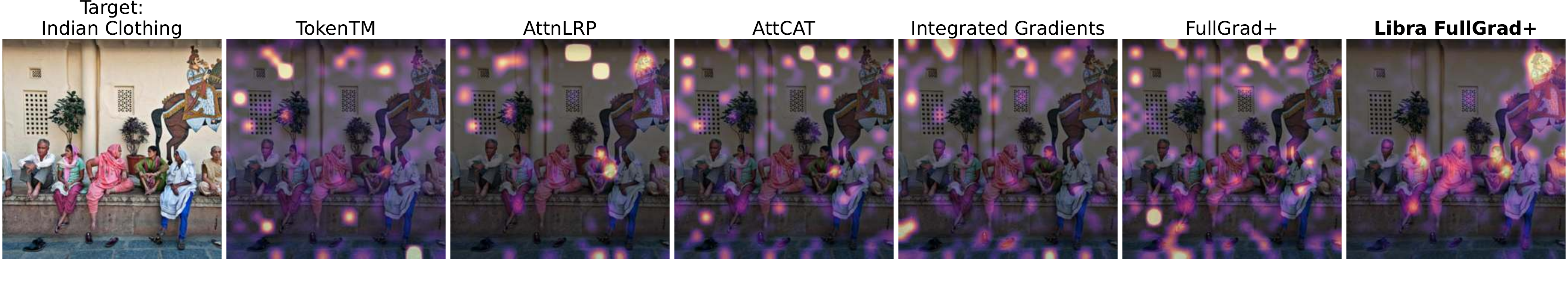,%
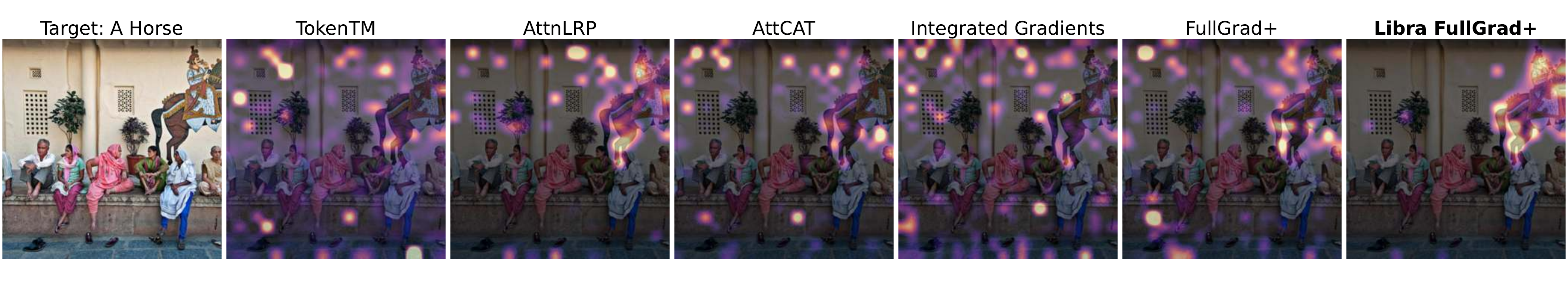,%
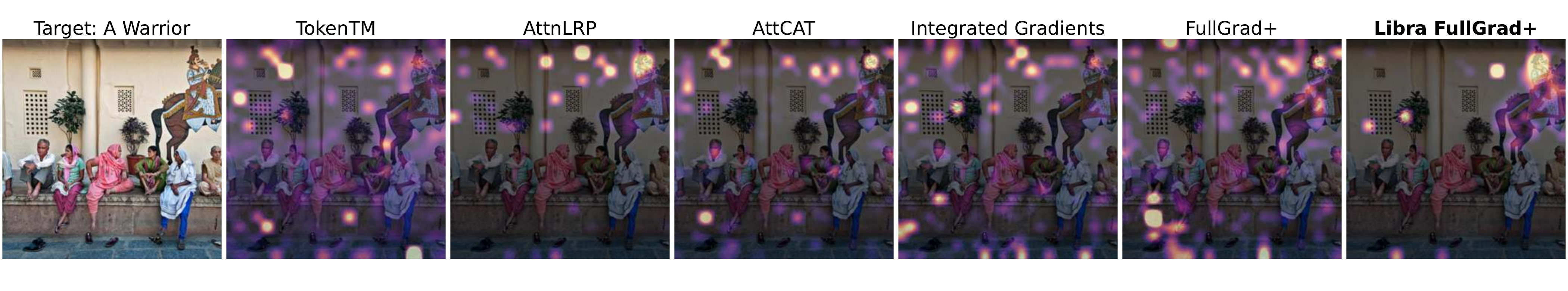,%
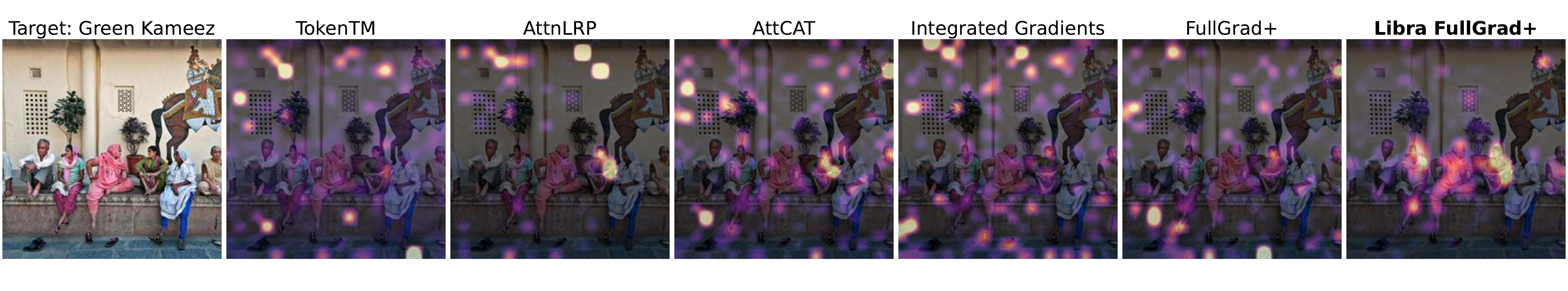,%
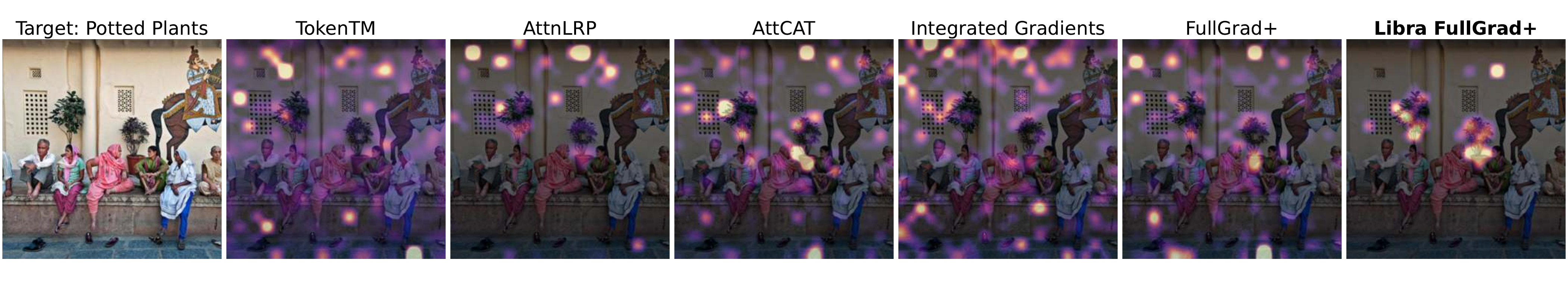,%
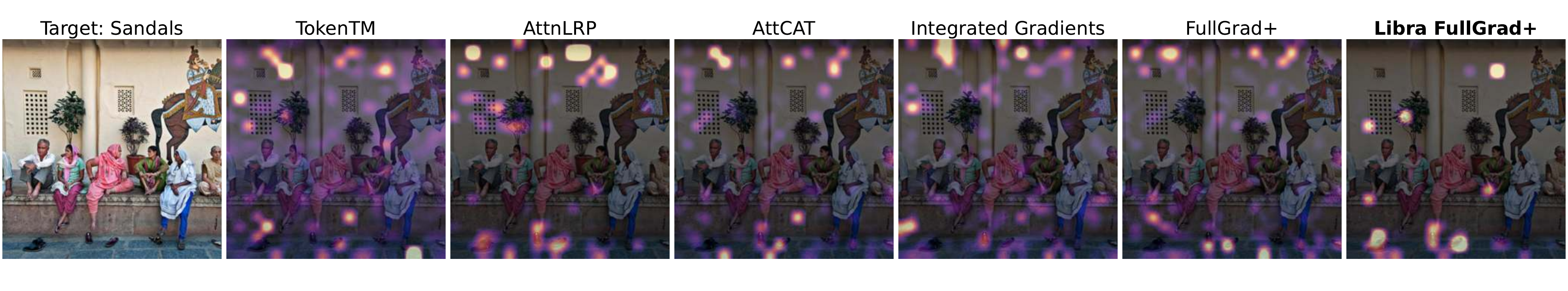,%
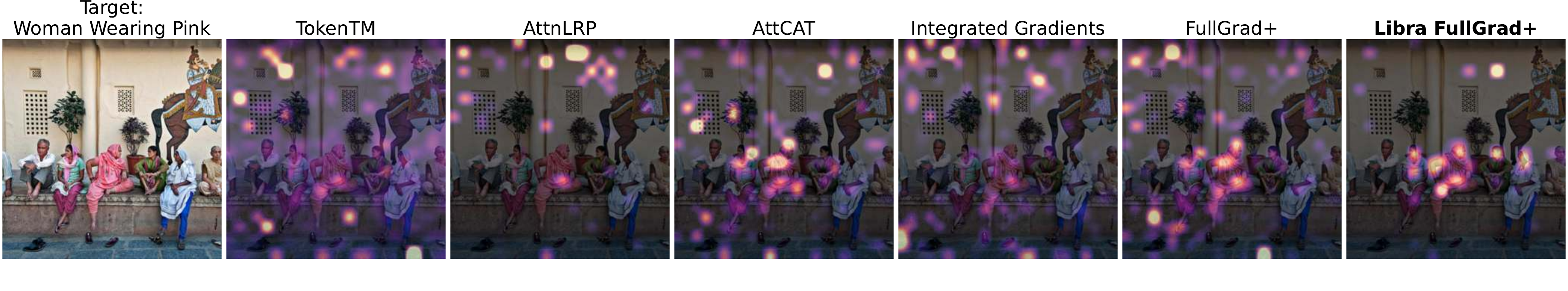,%
}{}{}

\CLIPRows[!b]{%
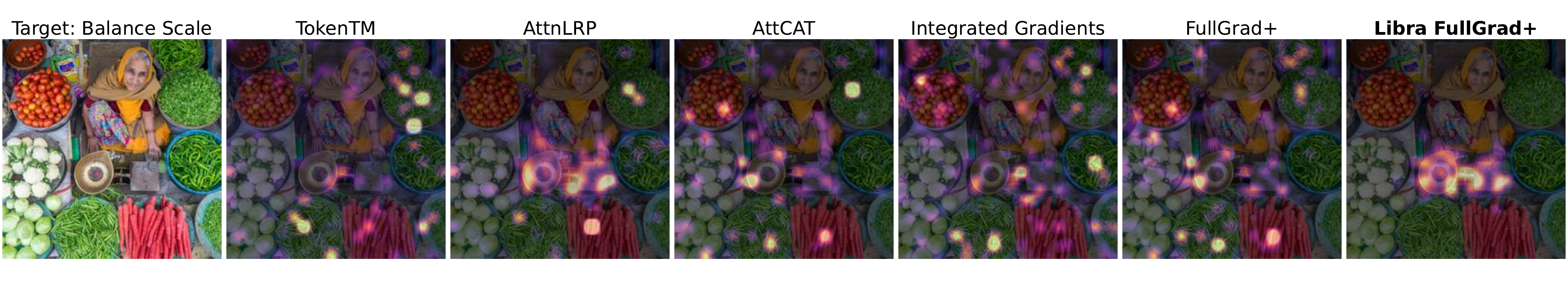,%
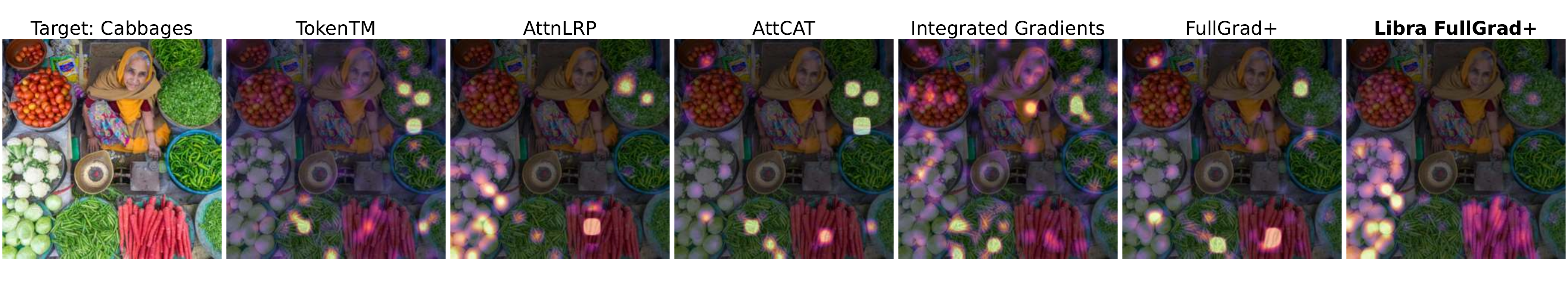,%
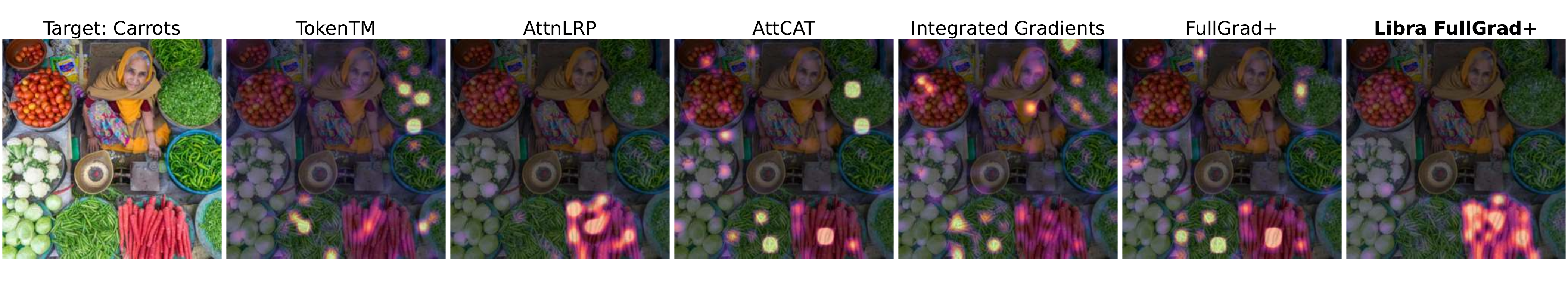,%
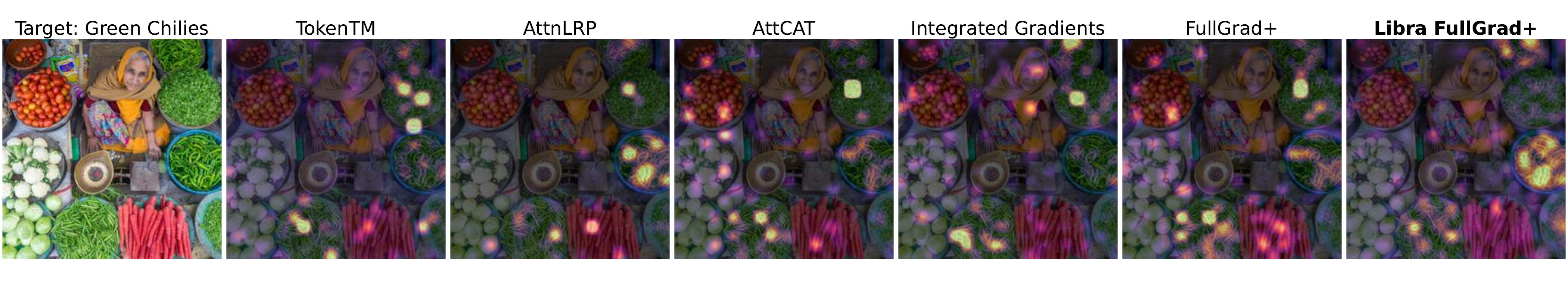,%
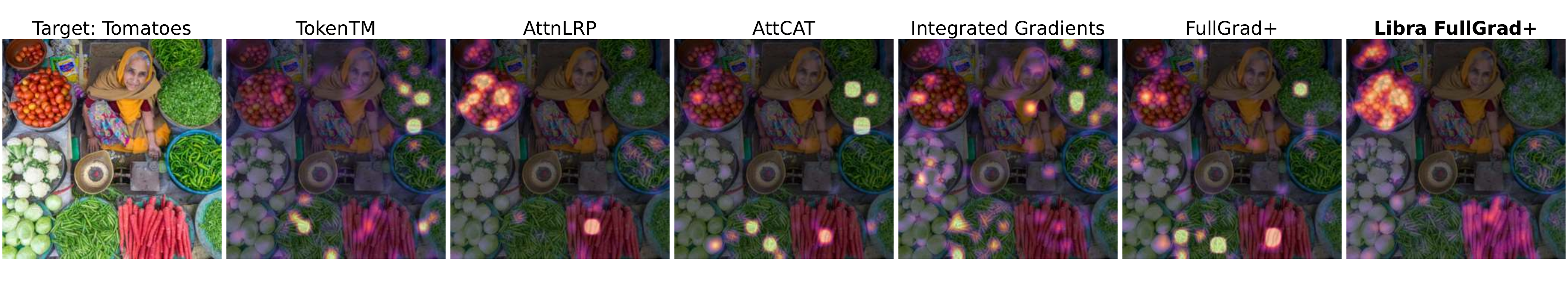,%
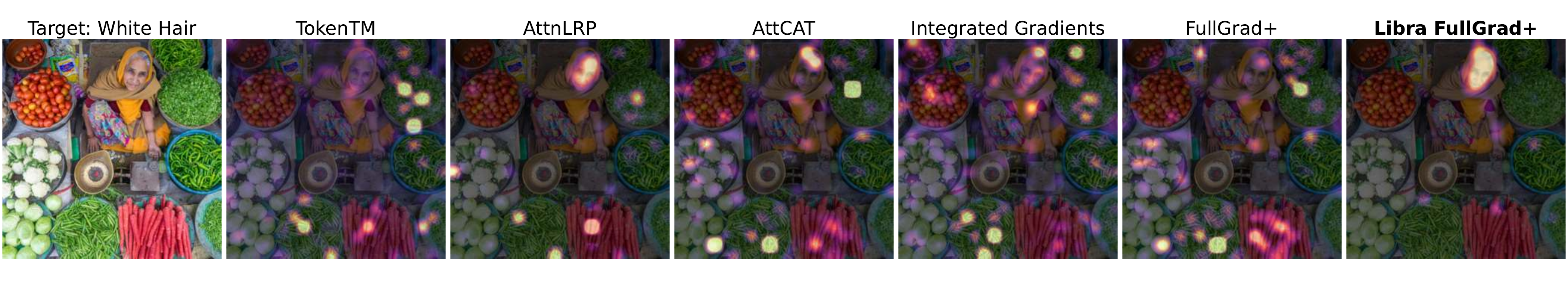,%
}{}{}

\CLIPRows[!b]{%
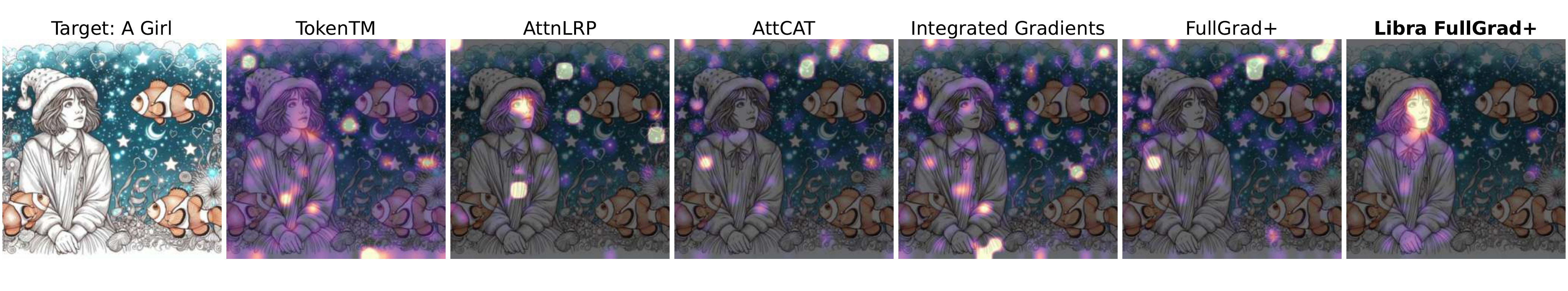,%
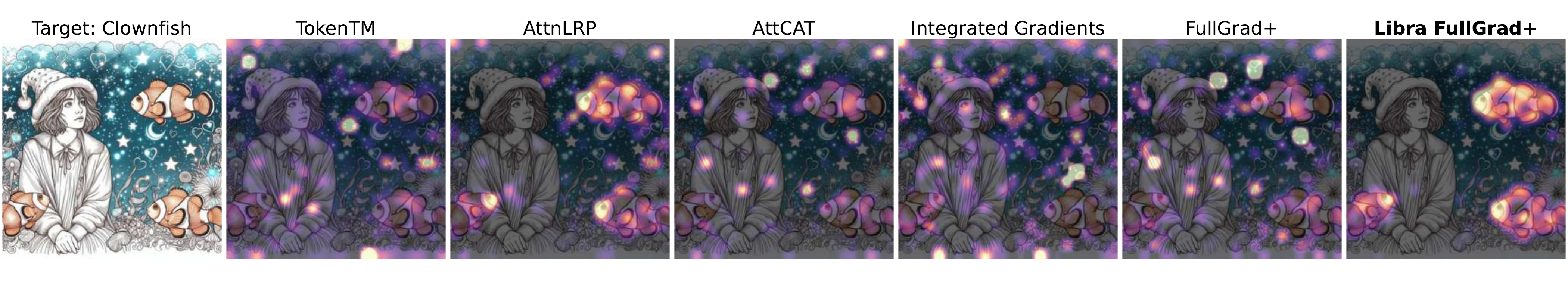,%
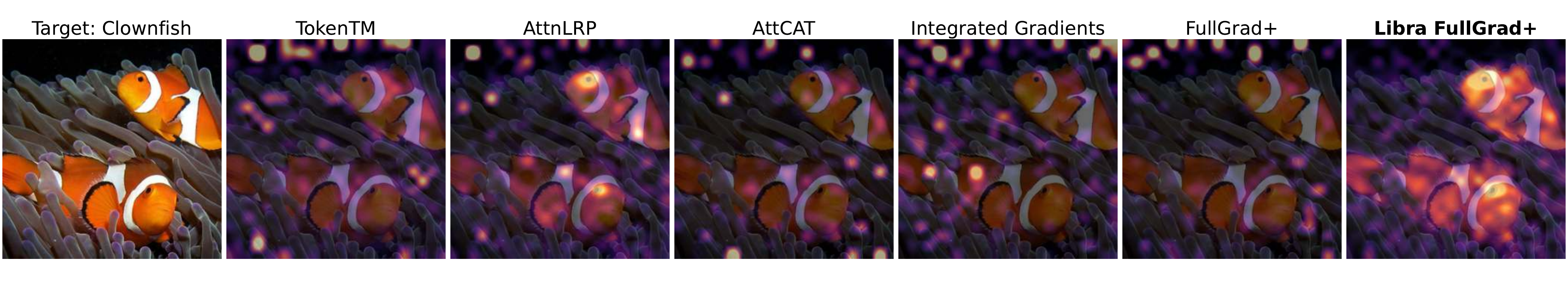,%
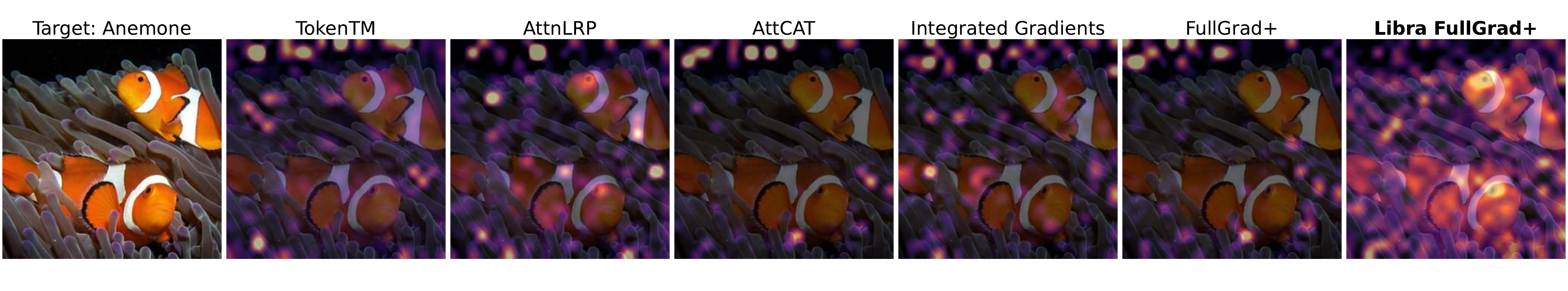,%
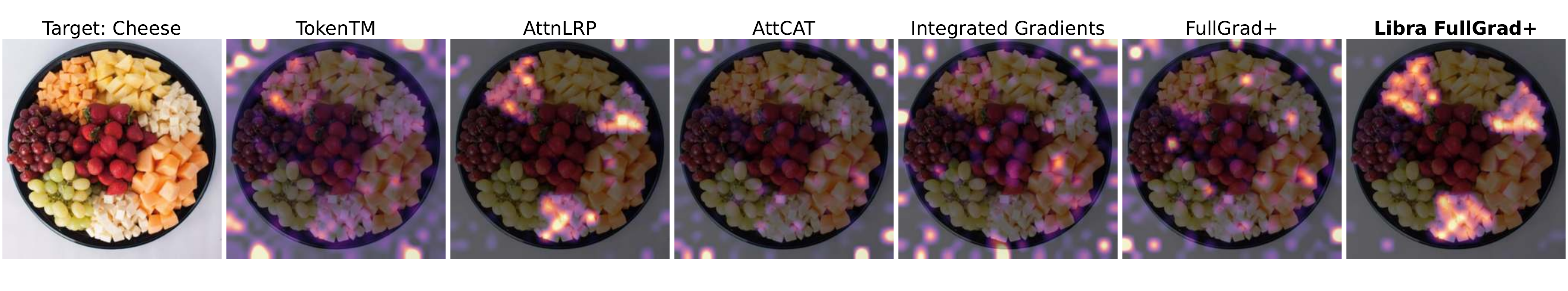,%
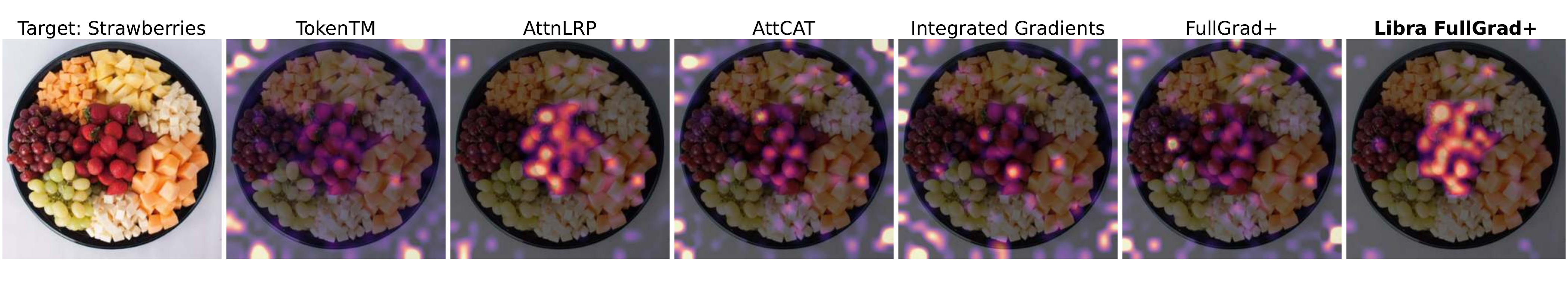,%
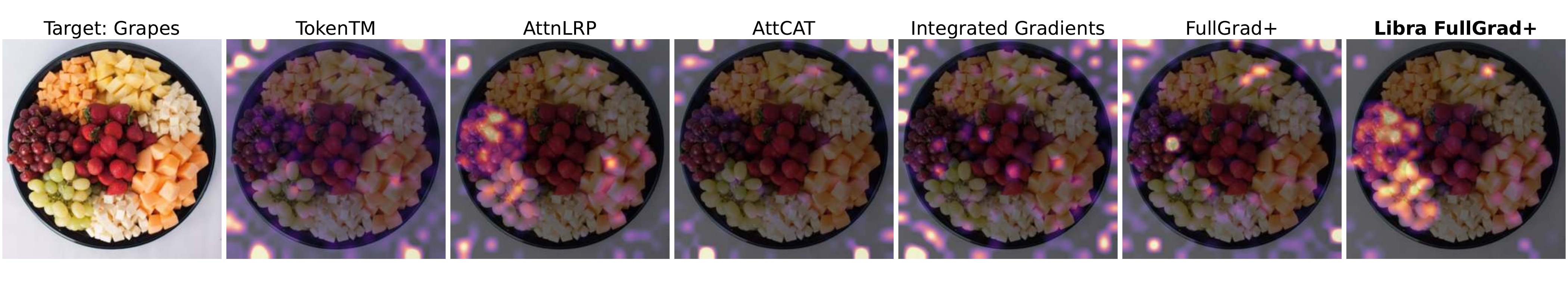,%
}{}{}

\CLIPRows[!b]{%
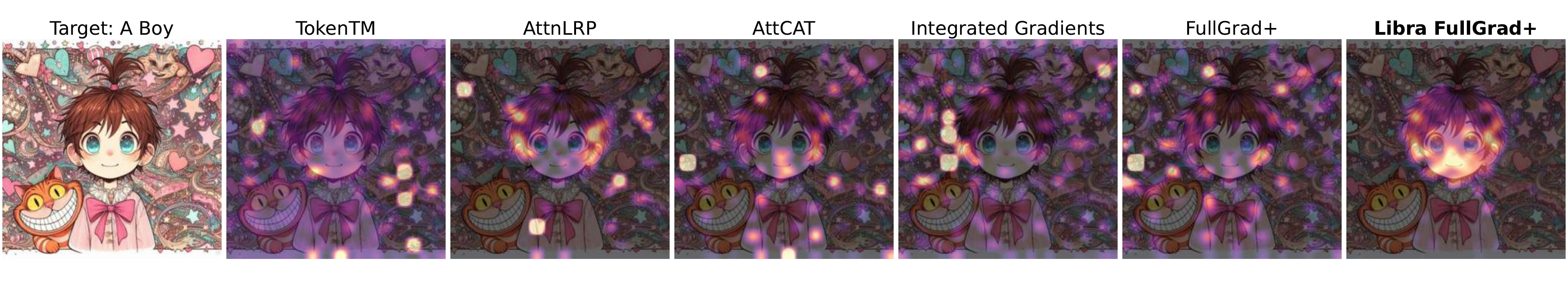,%
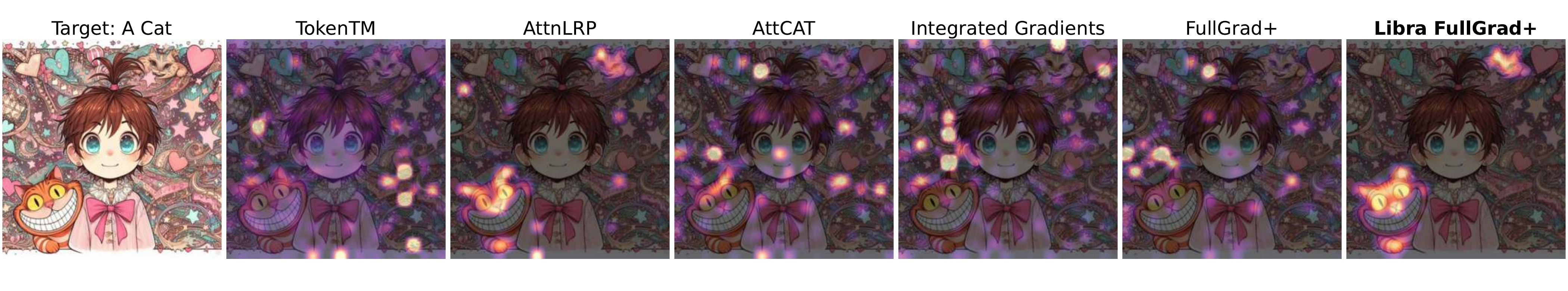,%
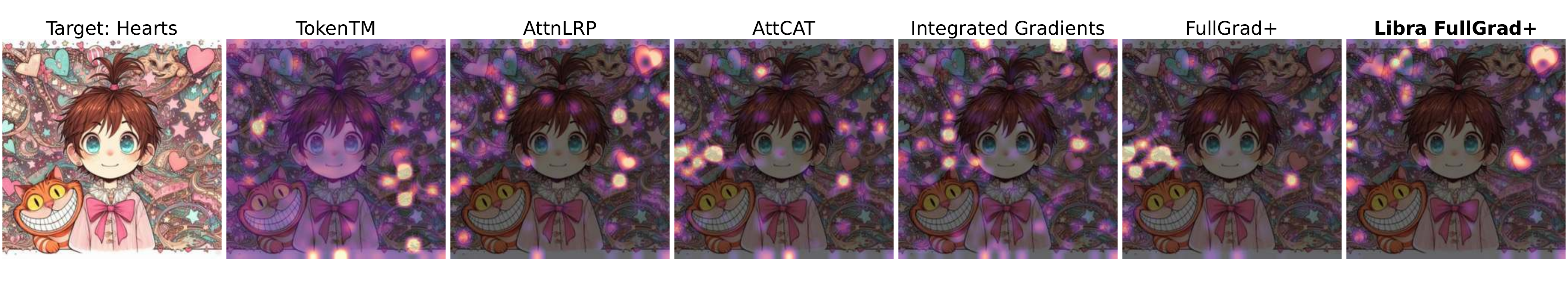,%
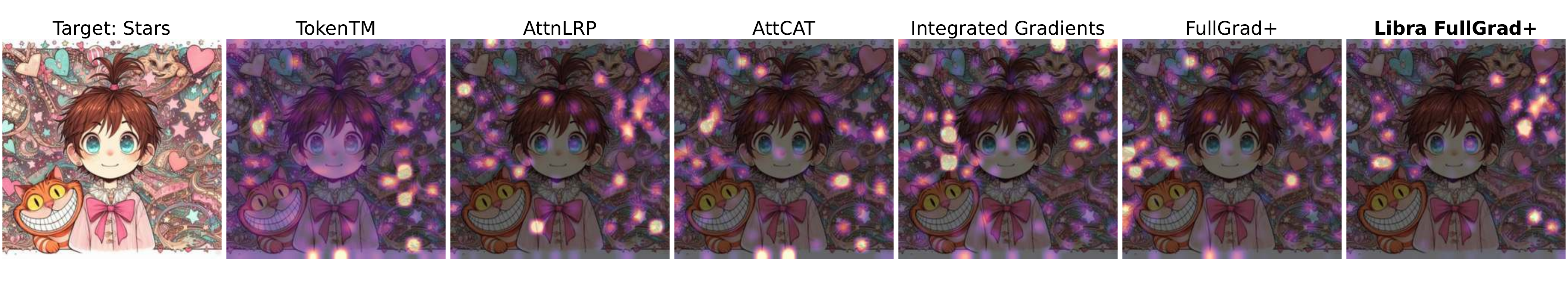,%
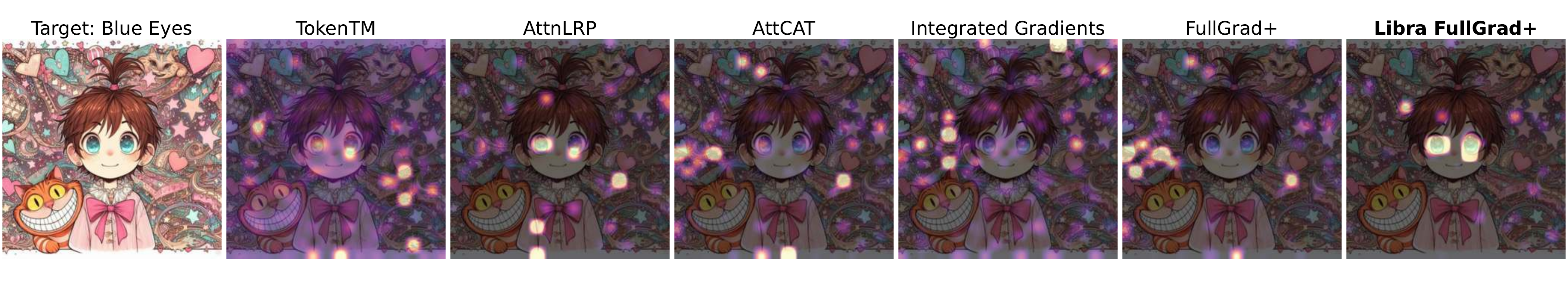,%
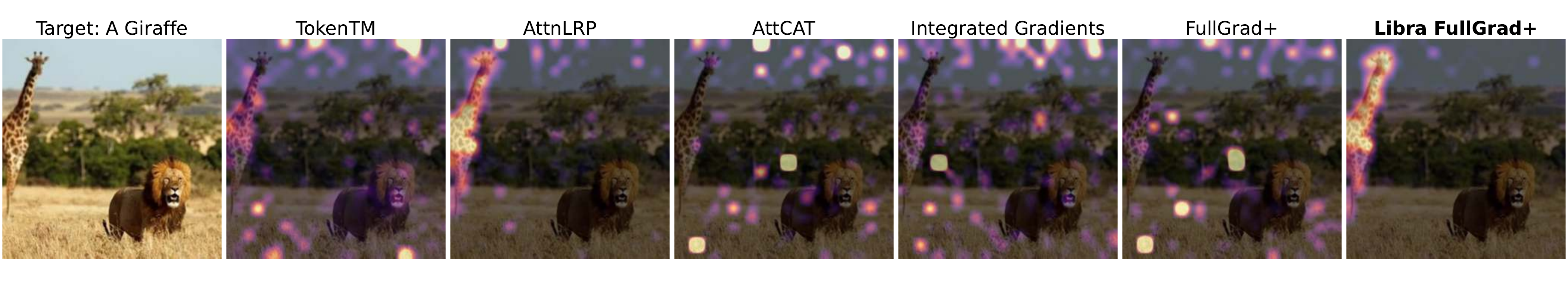,%
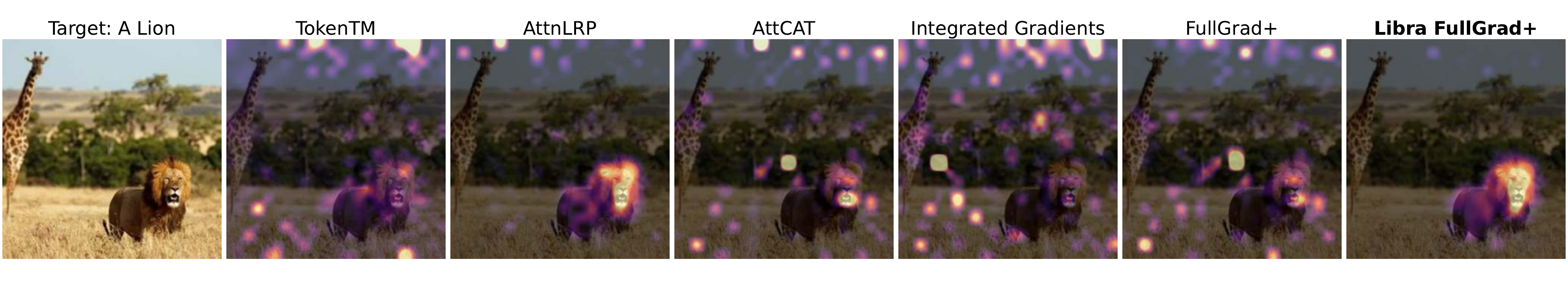,%
}{}{}

\CLIPRows[!b]{%
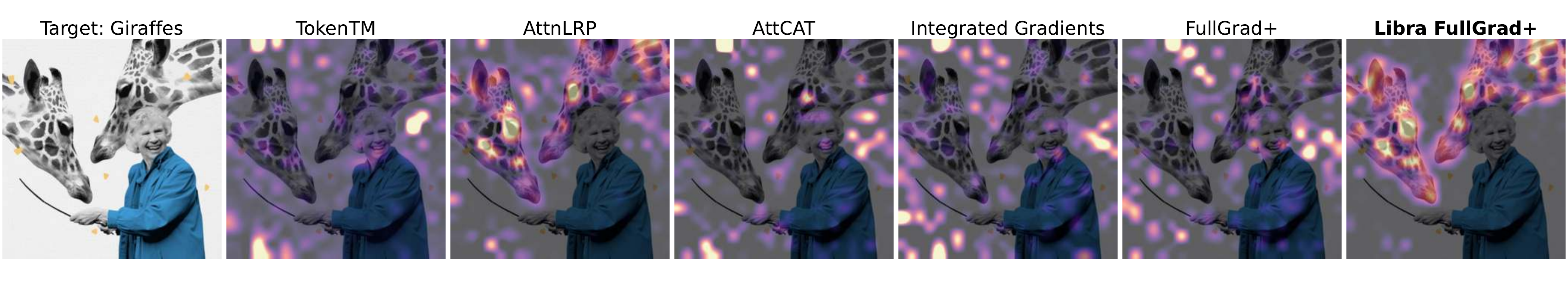,%
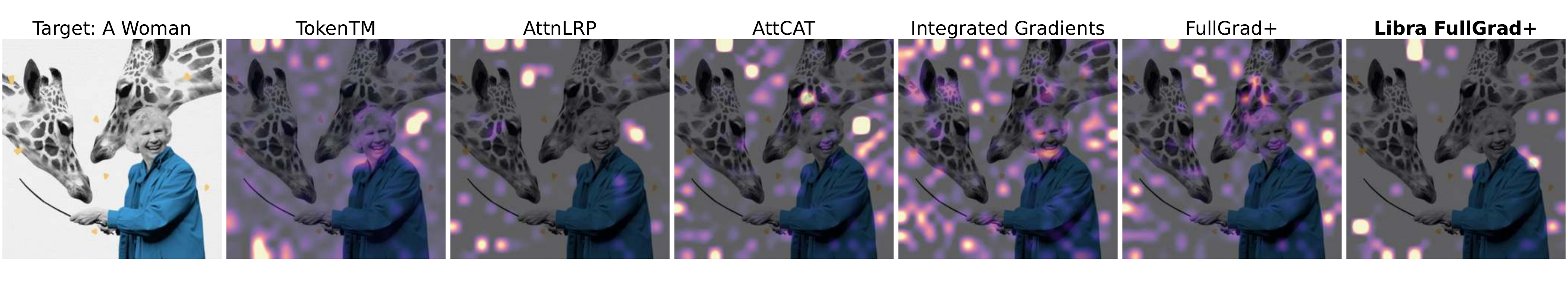,%
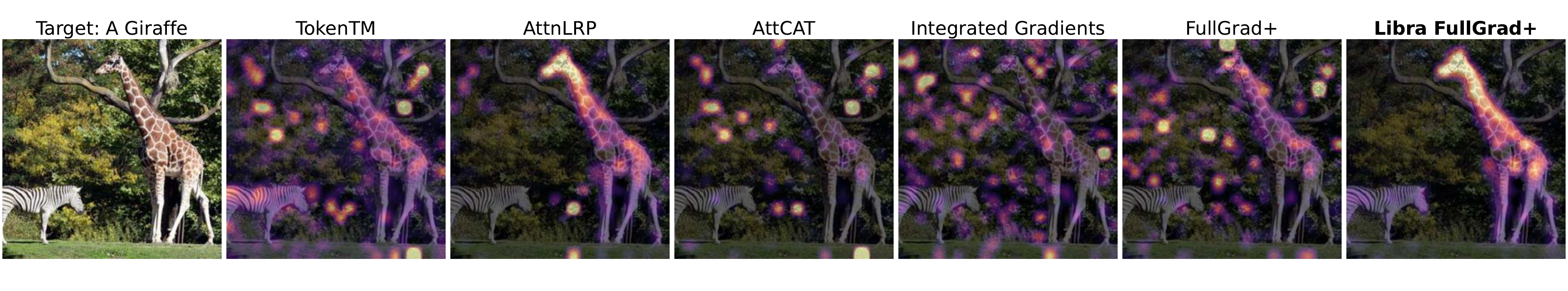,%
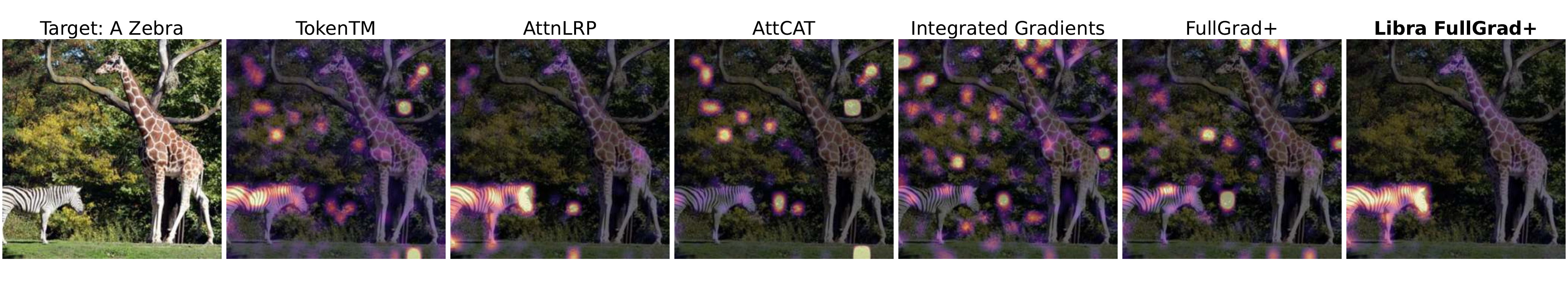,%
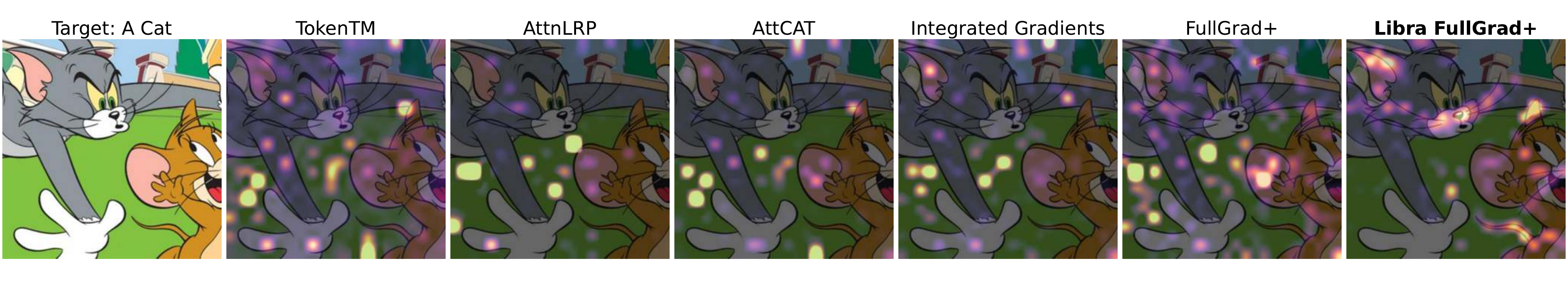,%
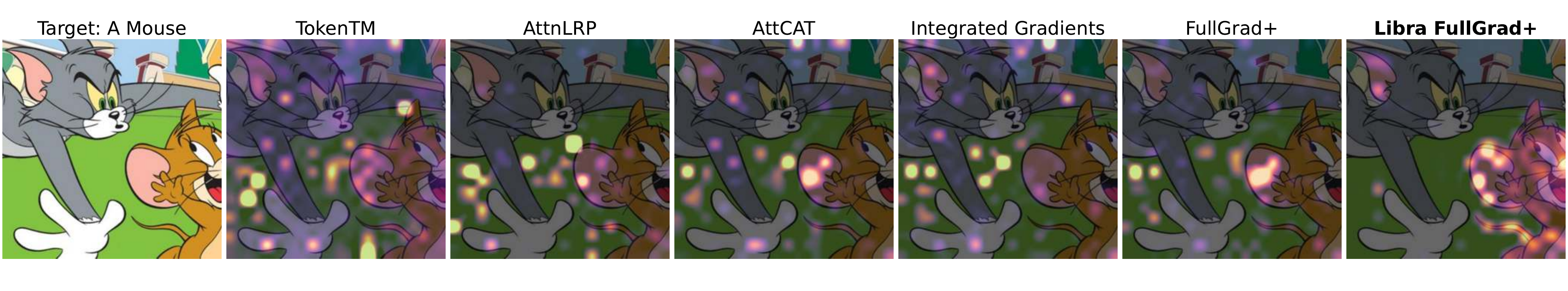,%
}{}{}

\CLIPRows[!b]{%
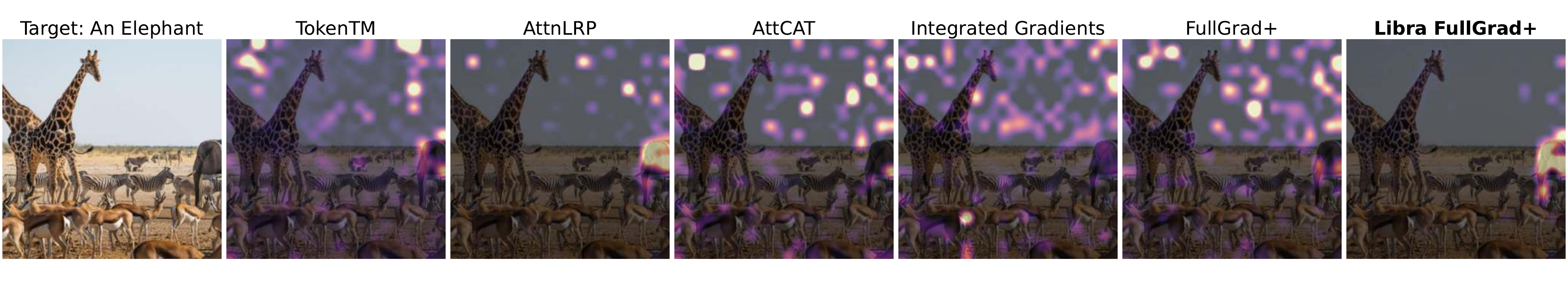,%
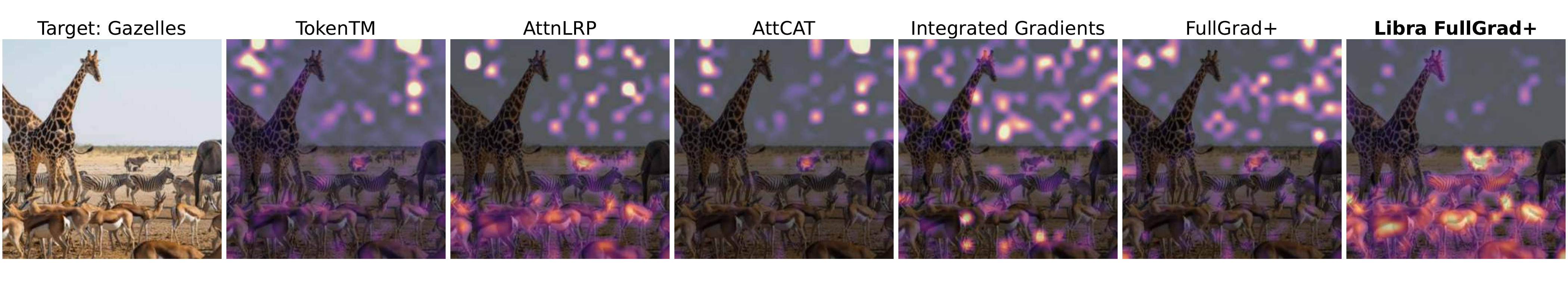,%
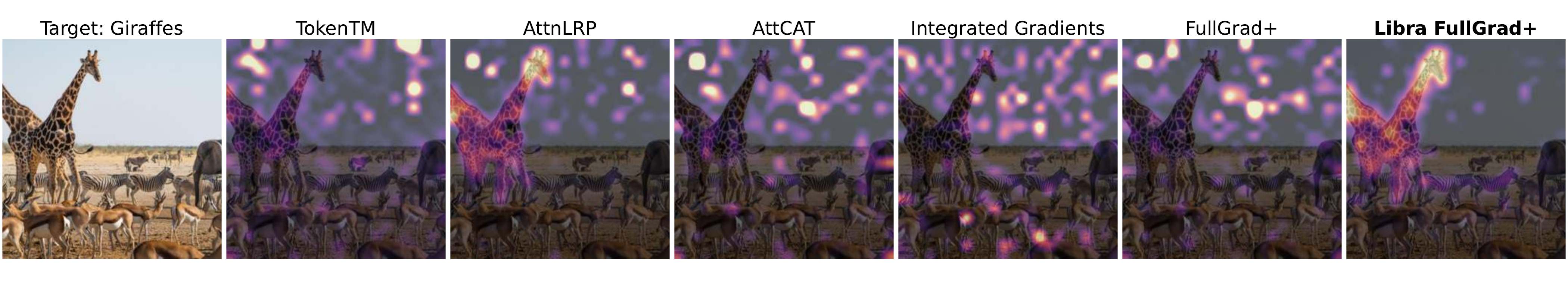,%
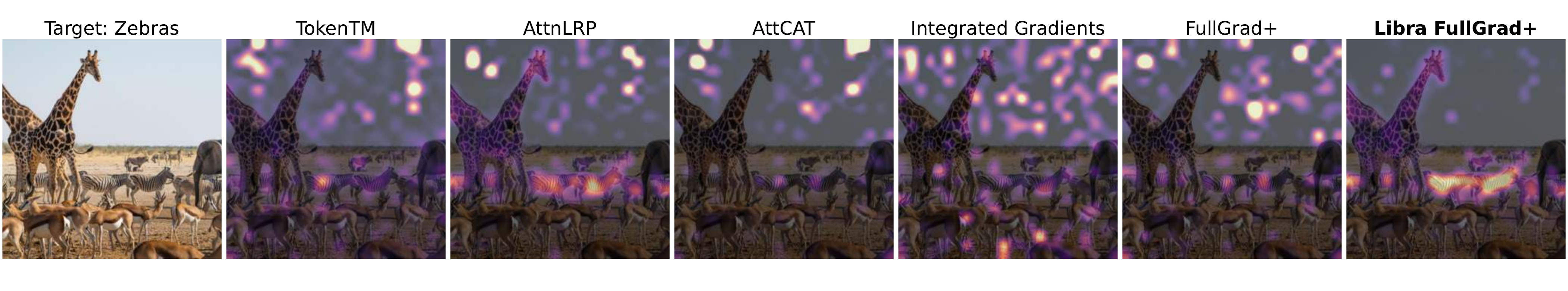,%
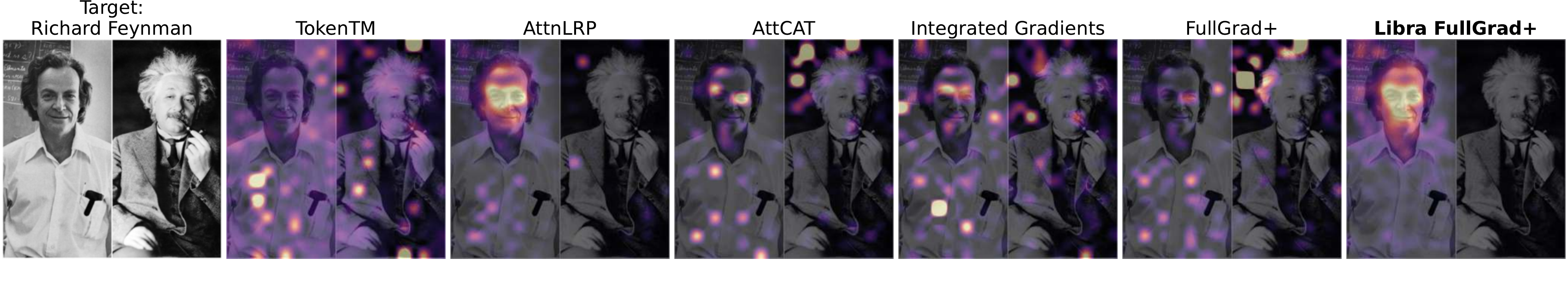,%
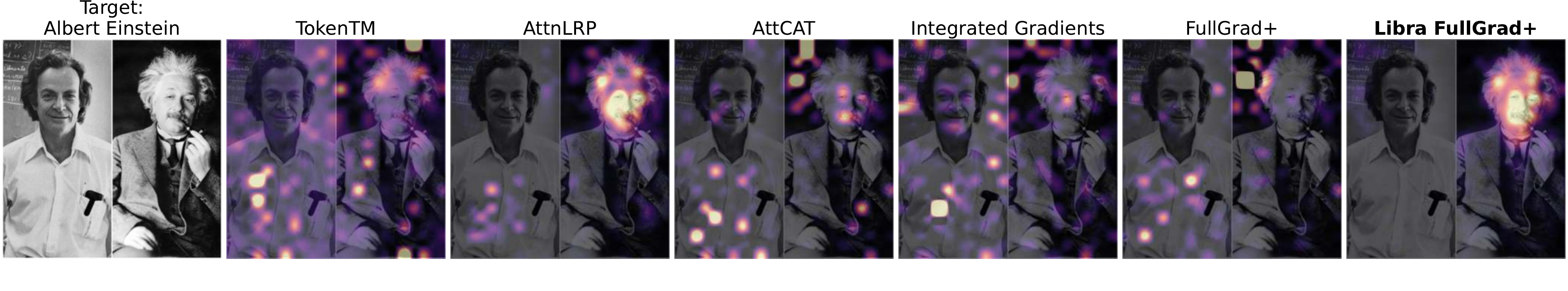,%
}{}{}

\CLIPRows[!b]{%
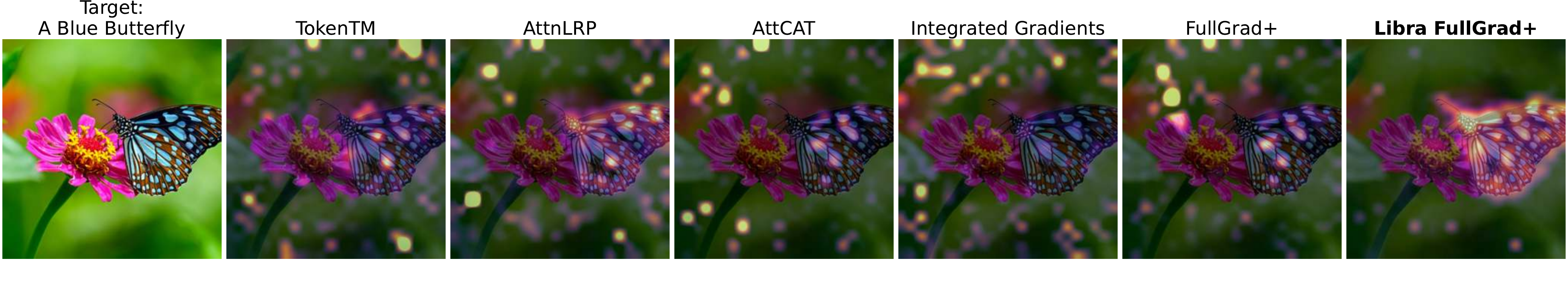,%
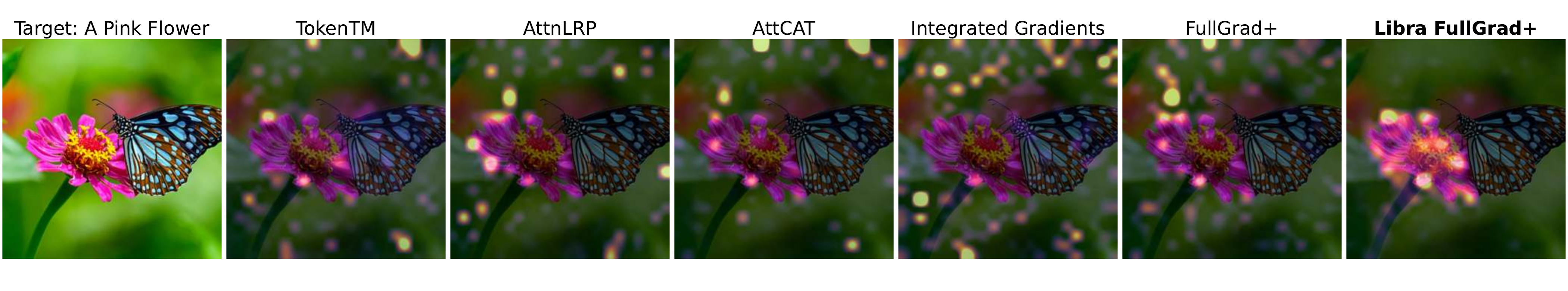,%
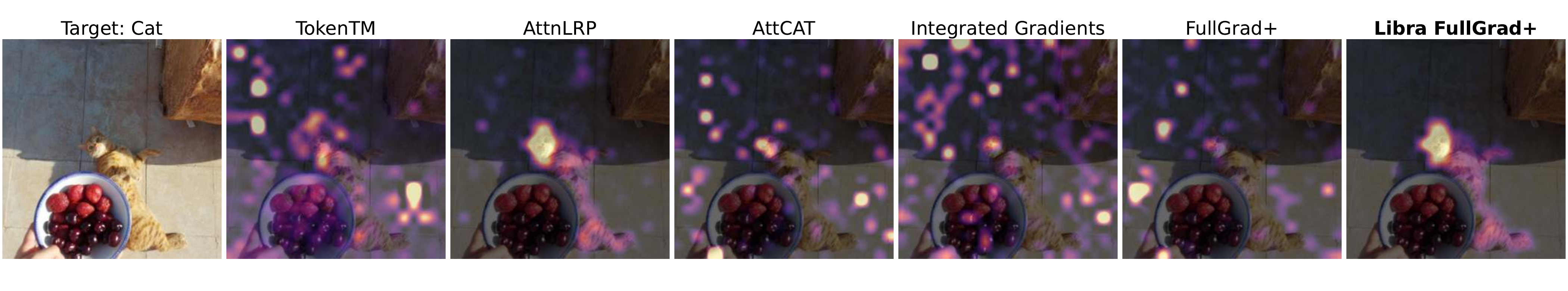,%
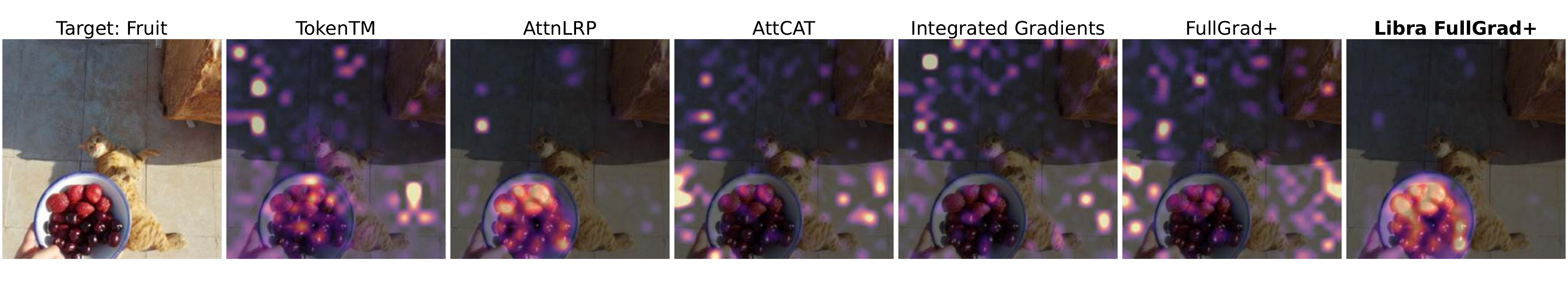,%
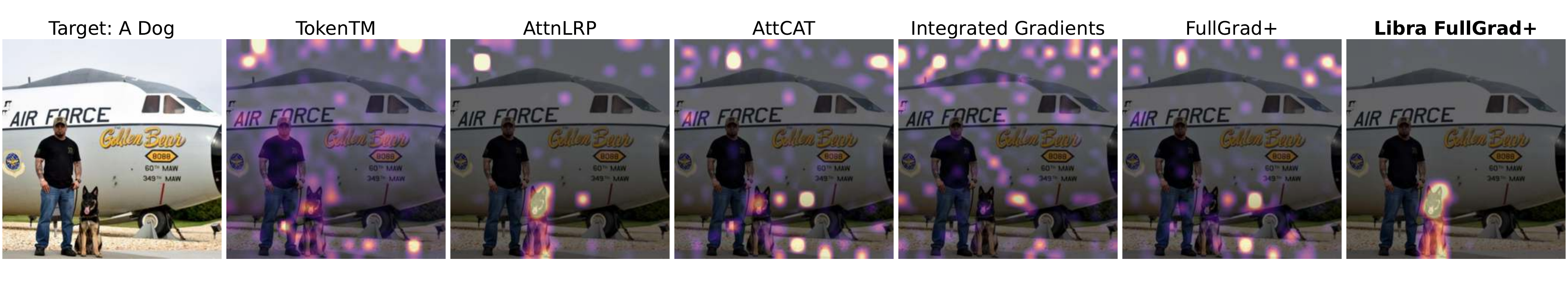,%
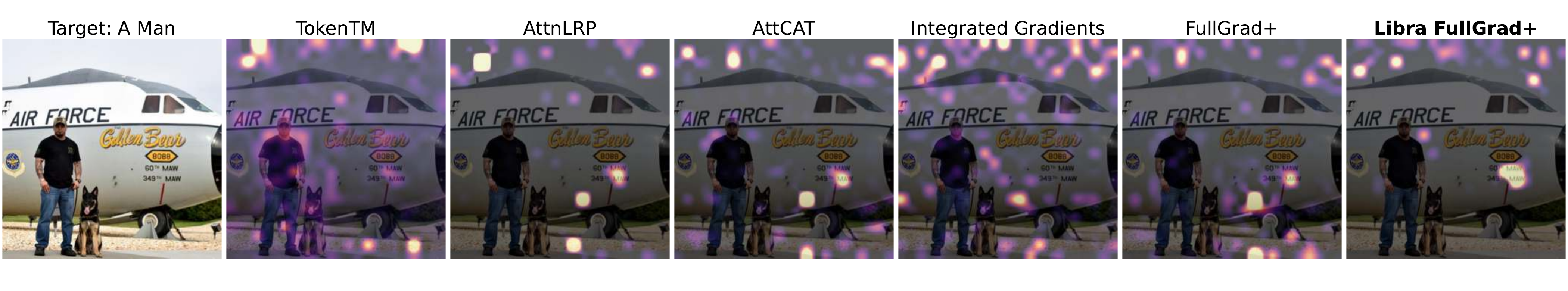,%
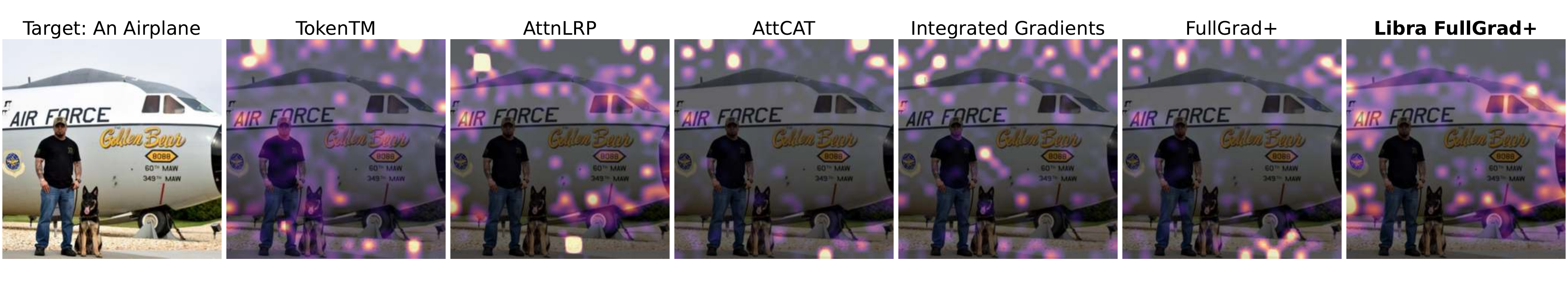,%
}{}{}

\CLIPRows[!b]{%
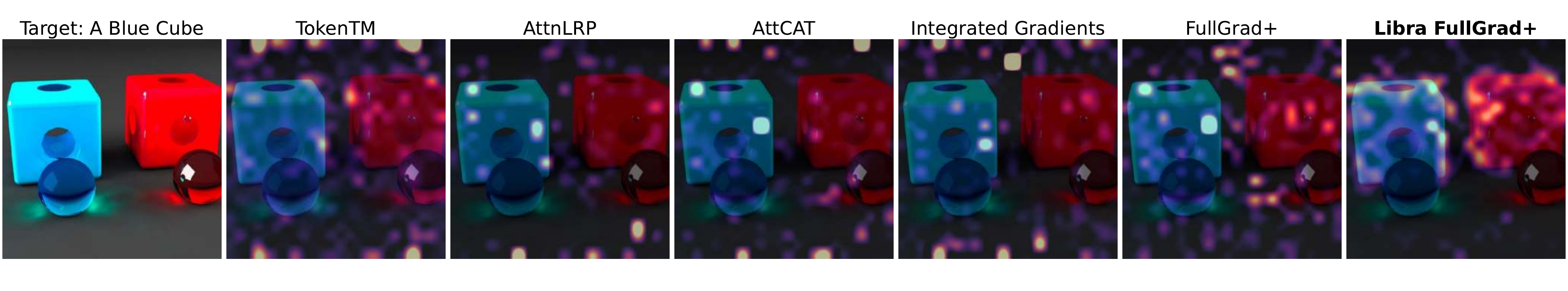,%
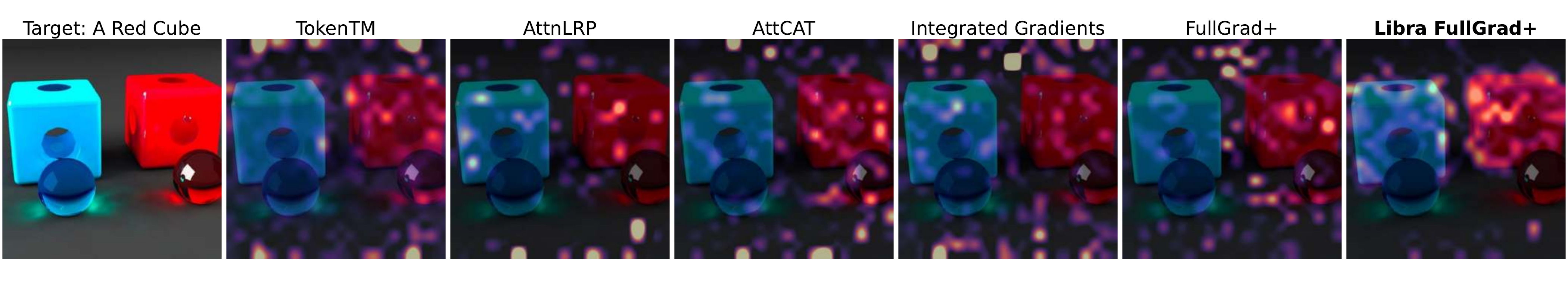,%
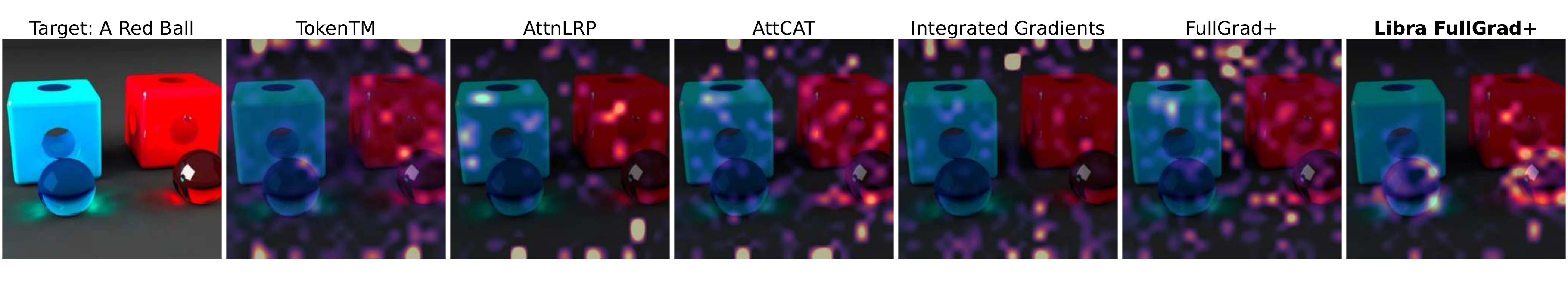,%
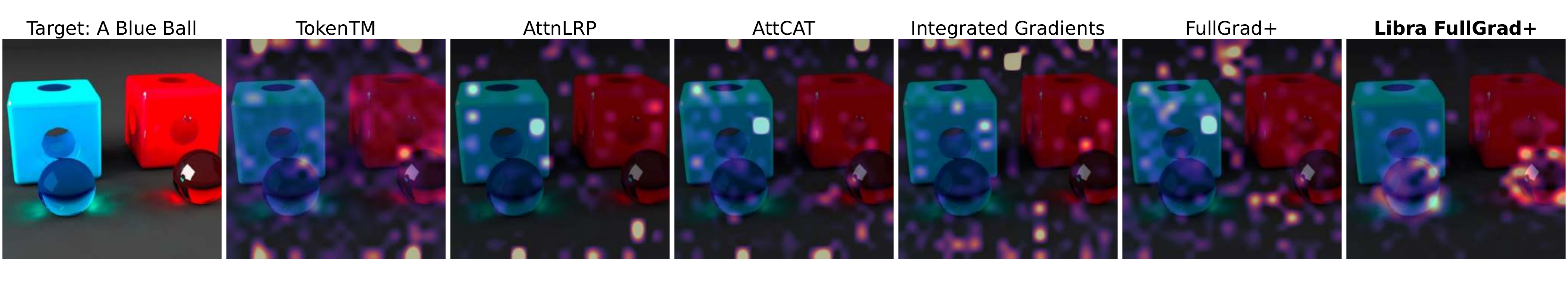,%
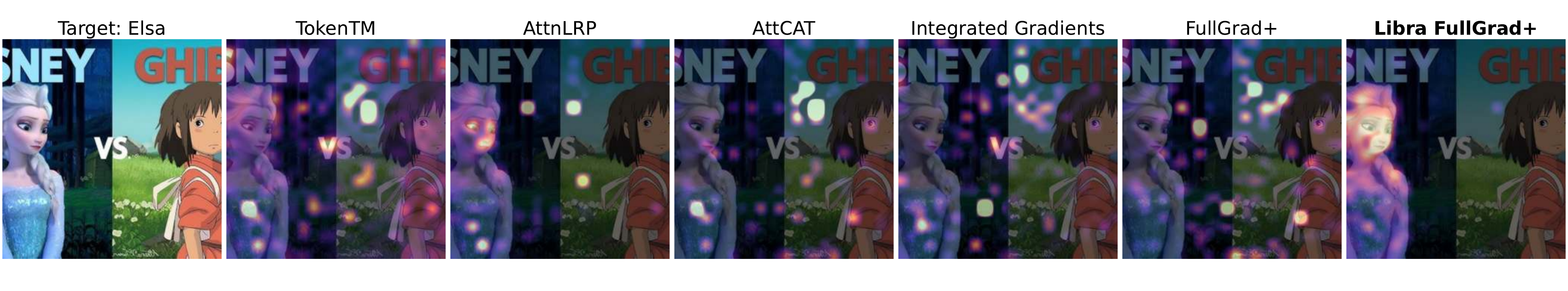,%
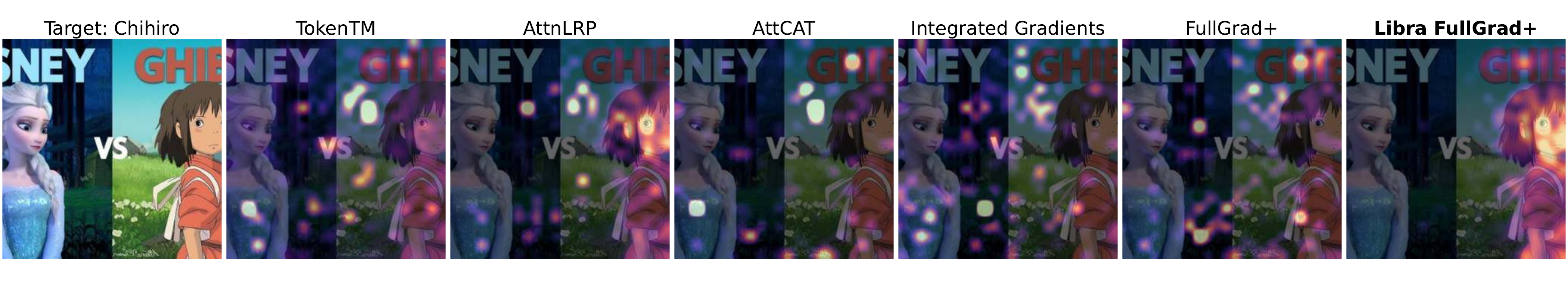,%
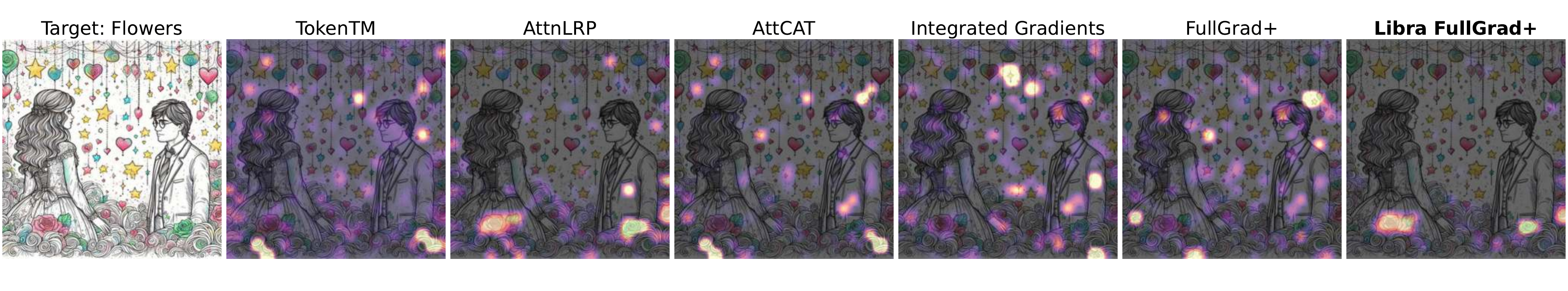,%
}{}{}

\CLIPRows[!b]{%
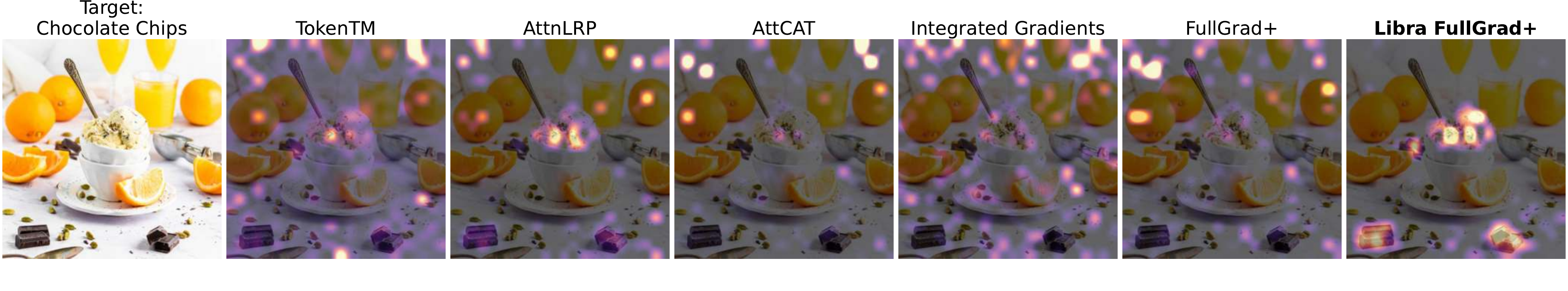,%
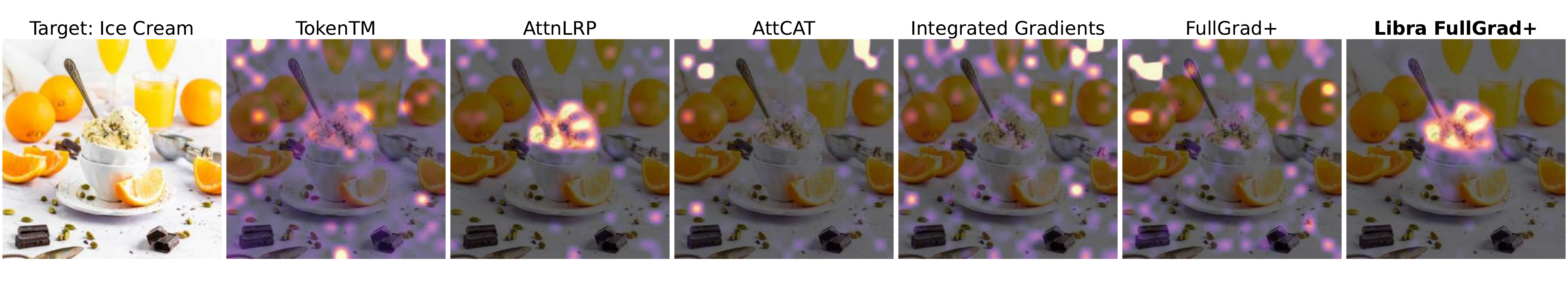,%
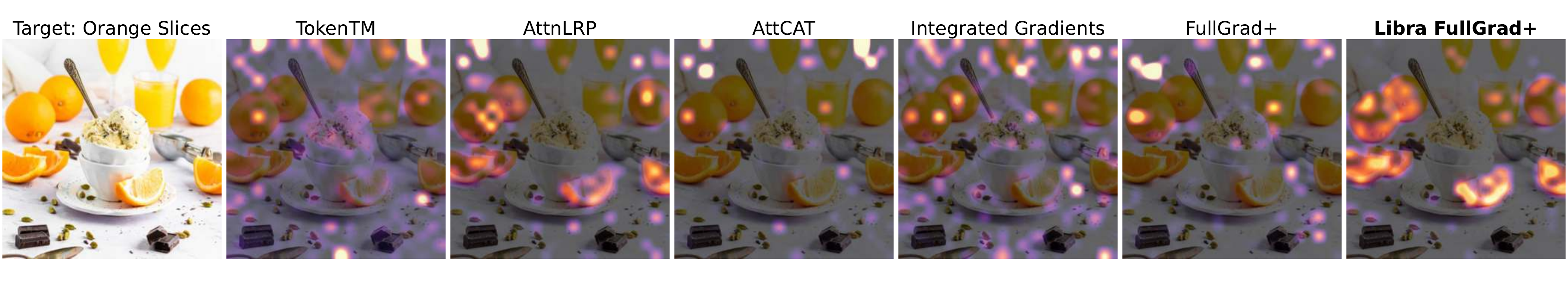,%
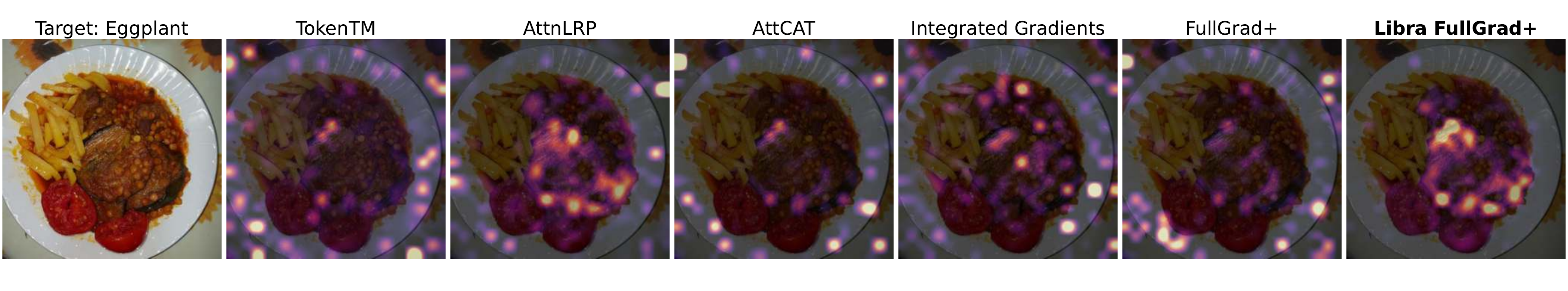,%
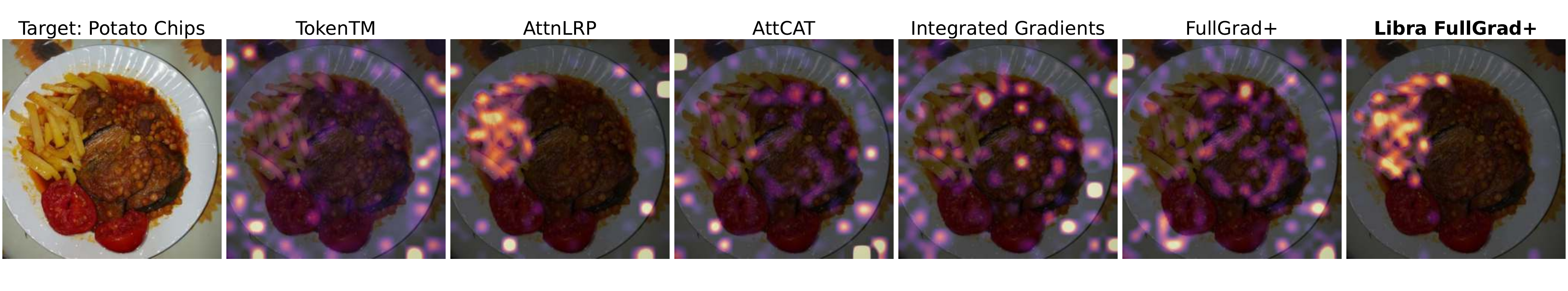,%
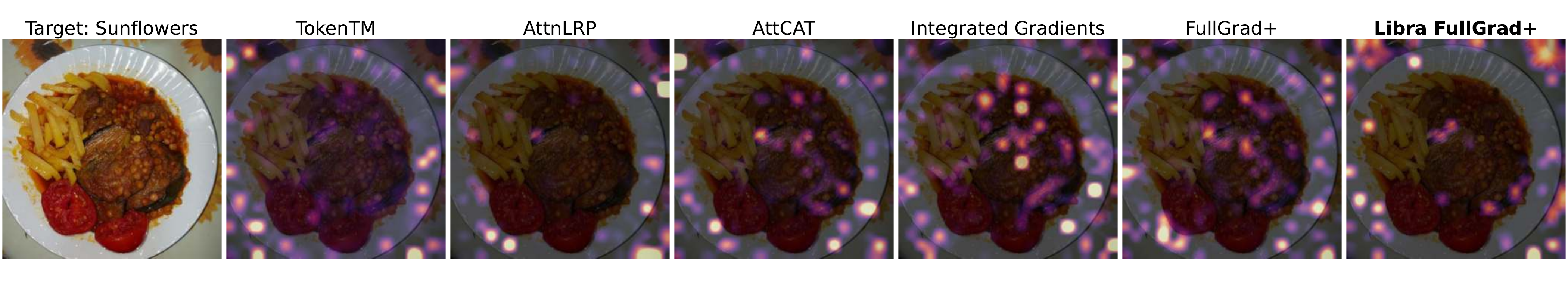,%
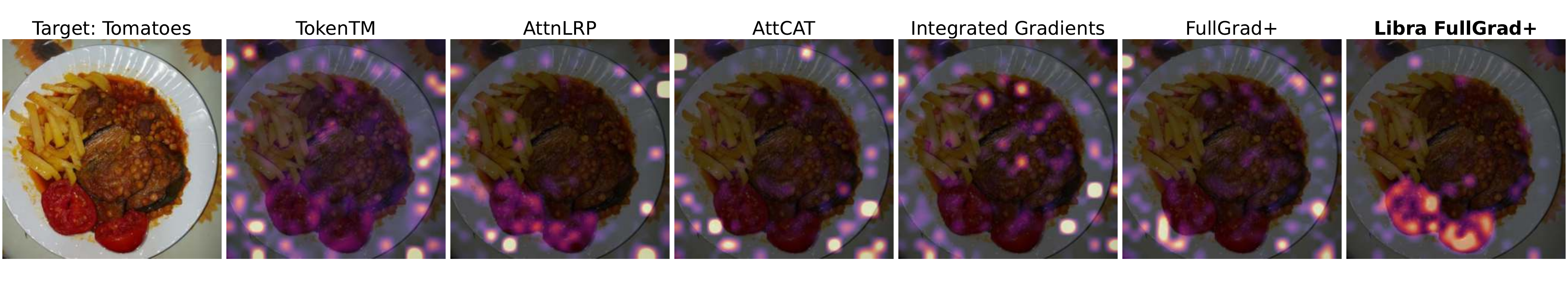,%
}{}{}

\CLIPRows[!b]{%
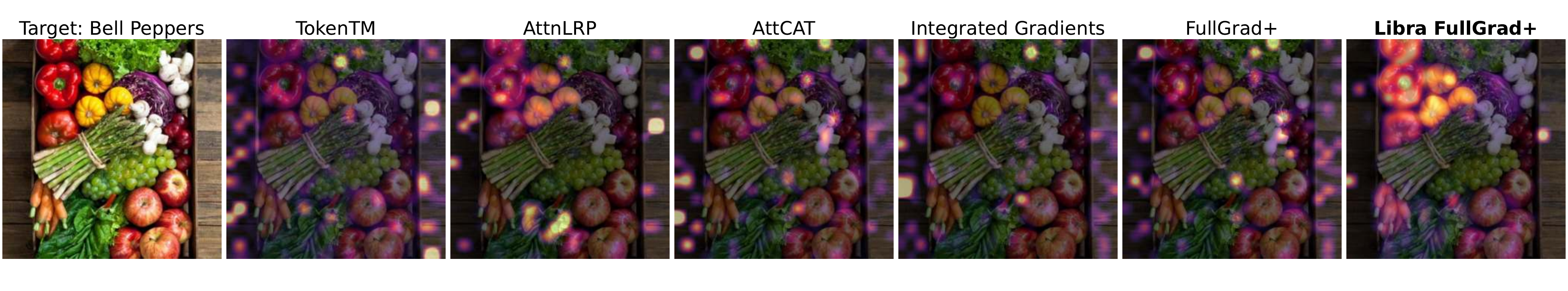,%
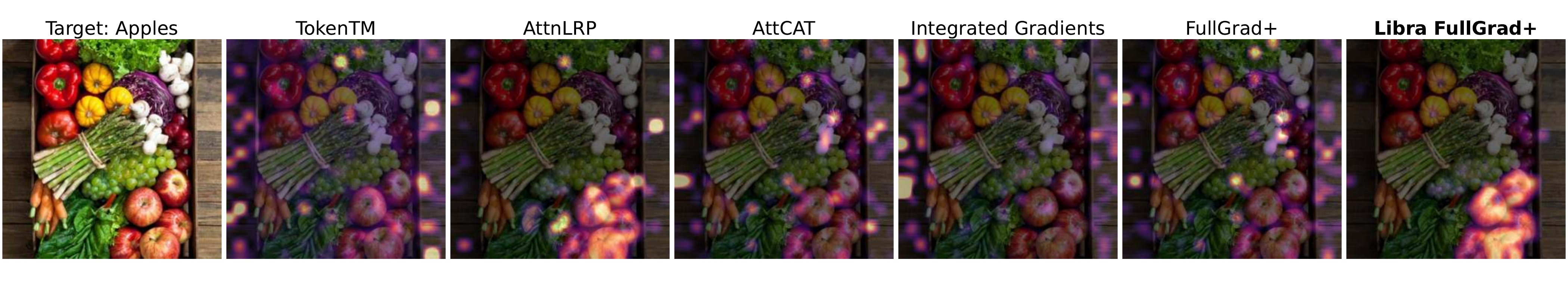,%
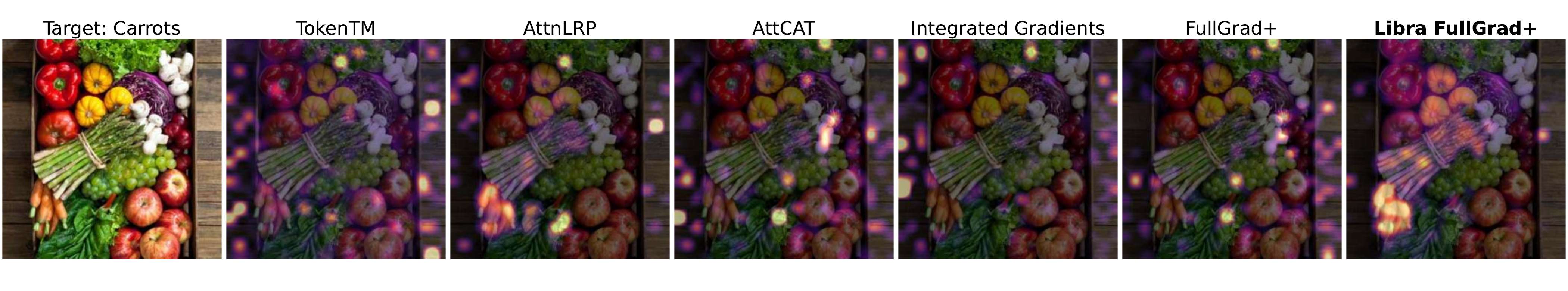,%
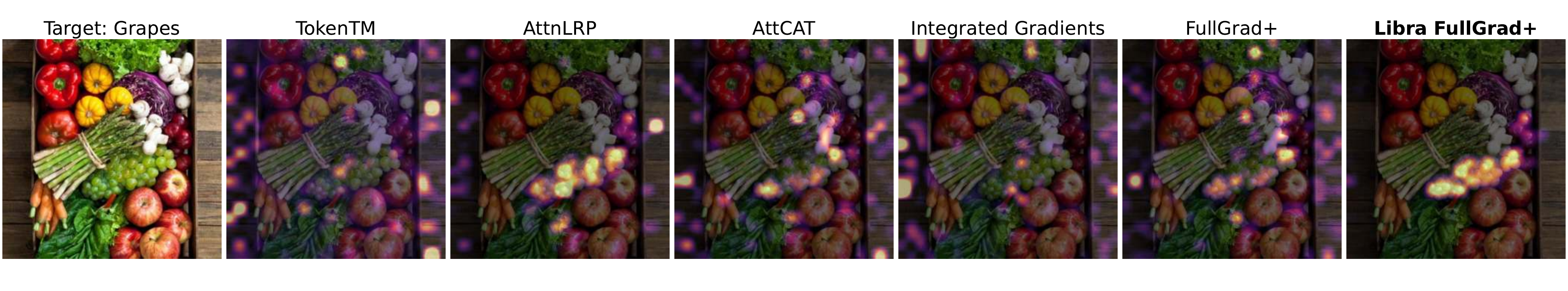,%
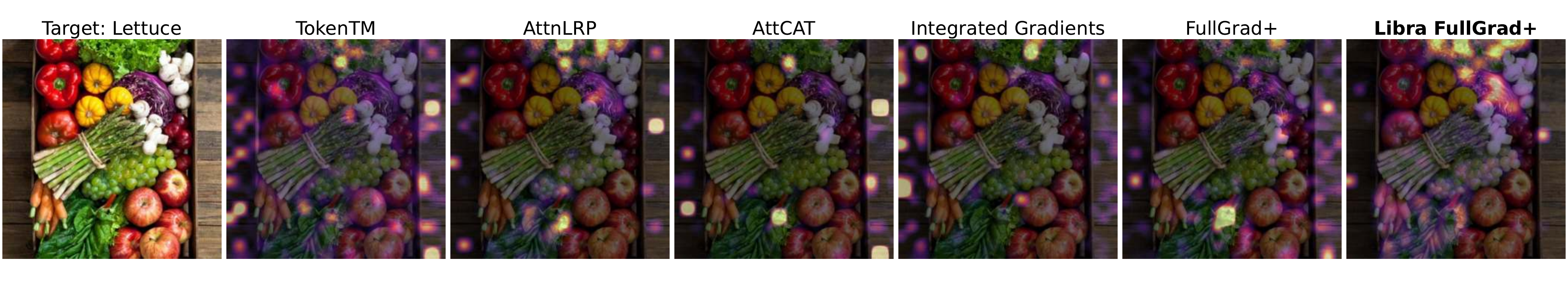,%
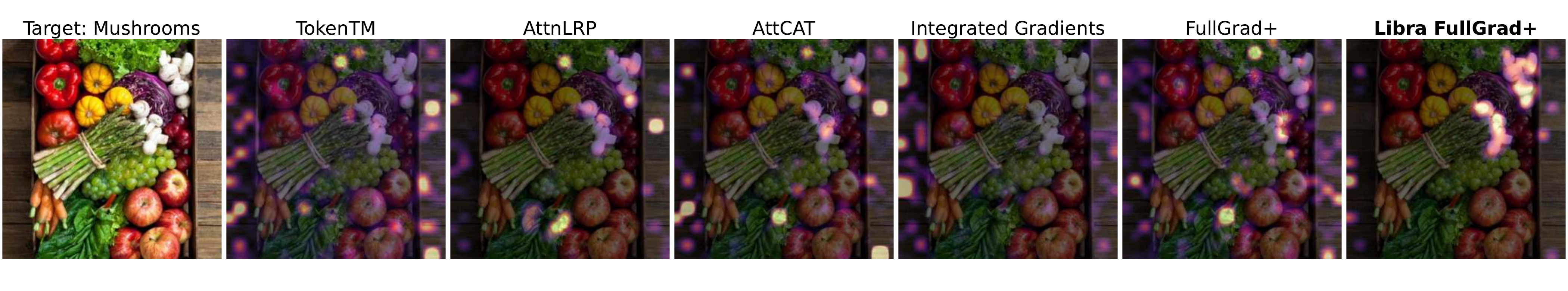,%
}{}{}

\CLIPRows[!b]{%
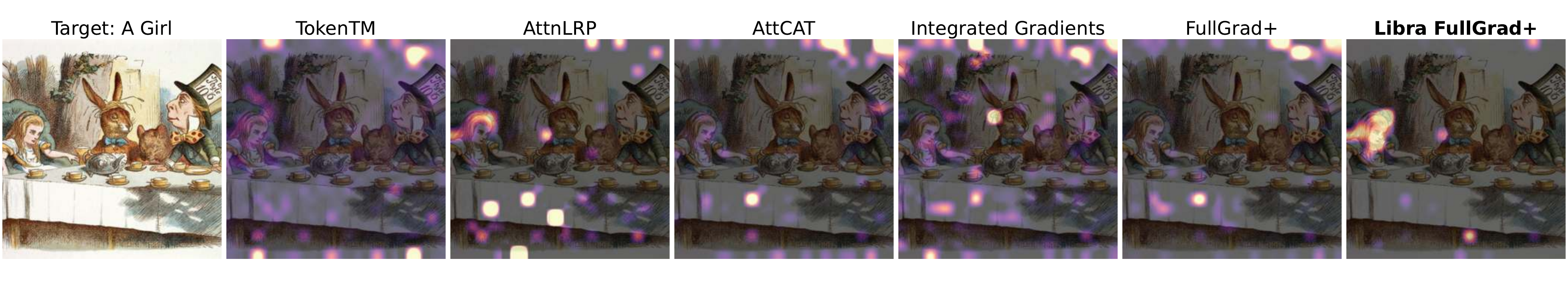,%
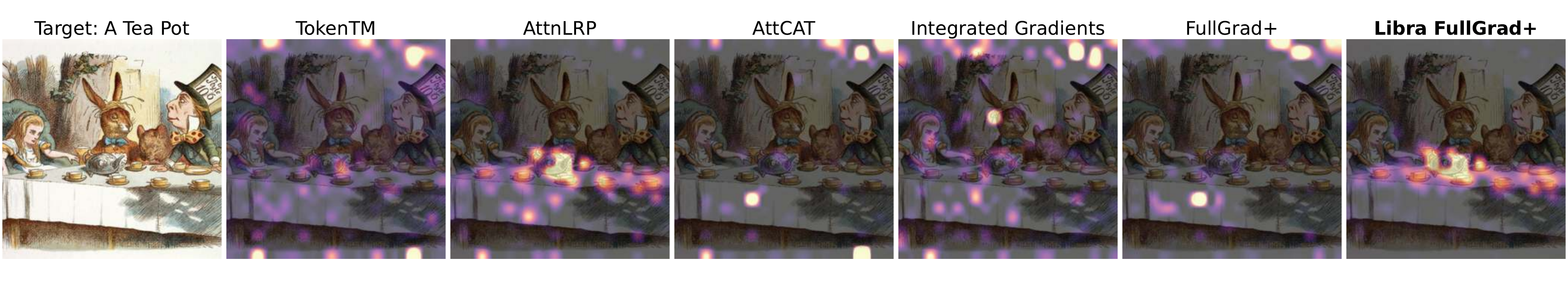,%
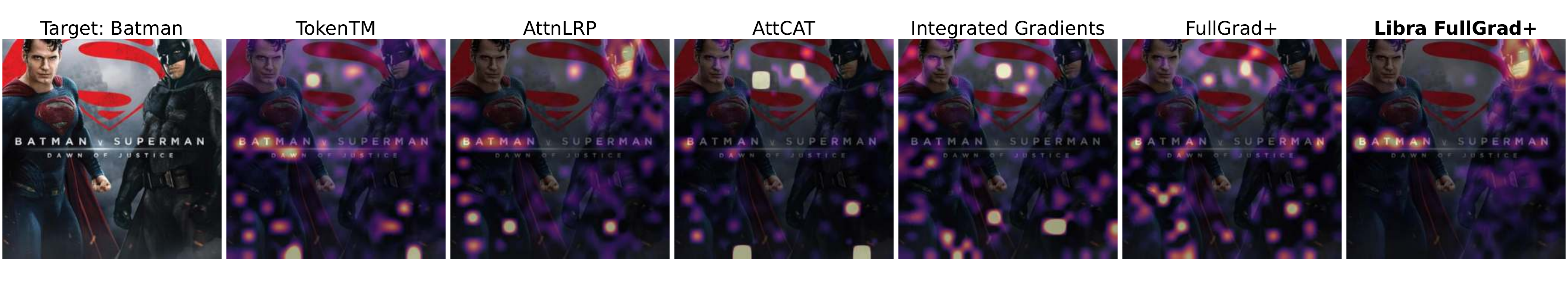,%
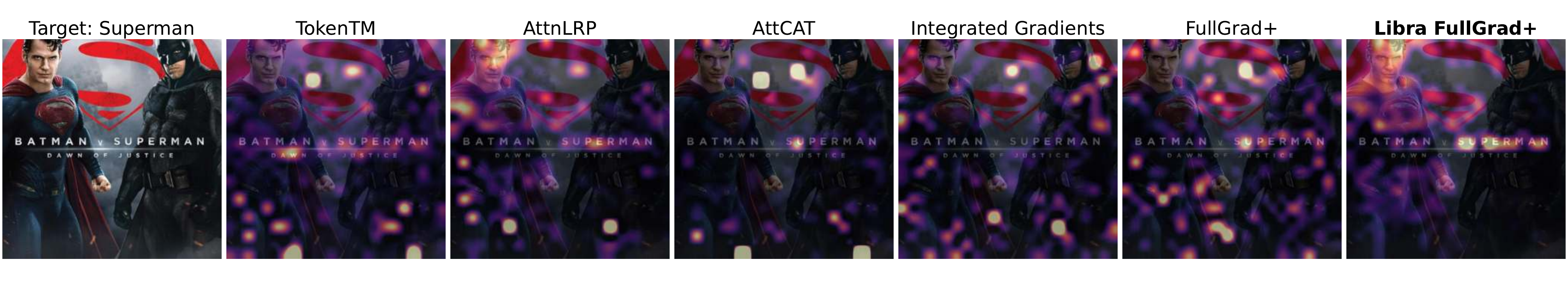,%
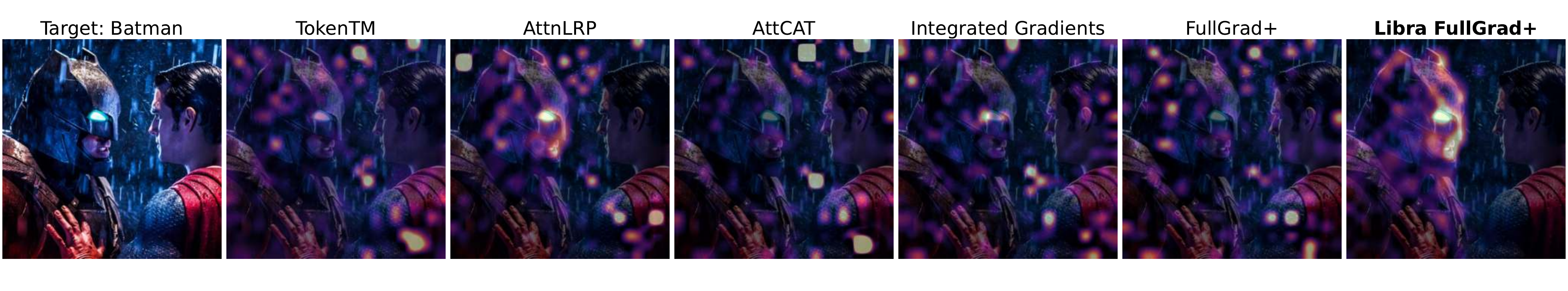,%
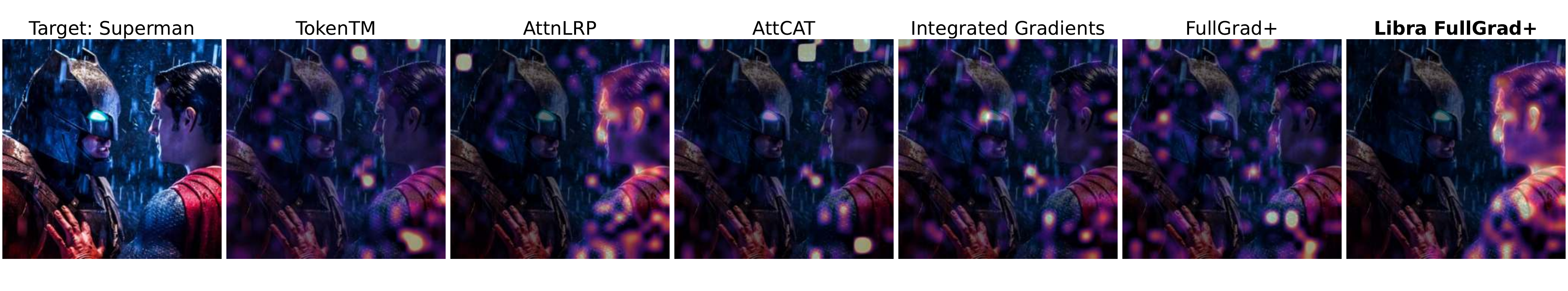,%
}{}{}

\CLIPRows[!b]{%
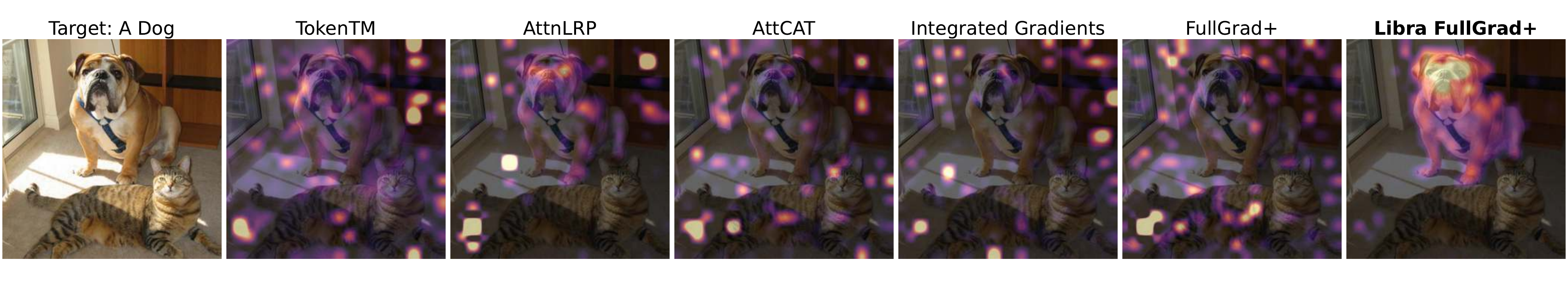,%
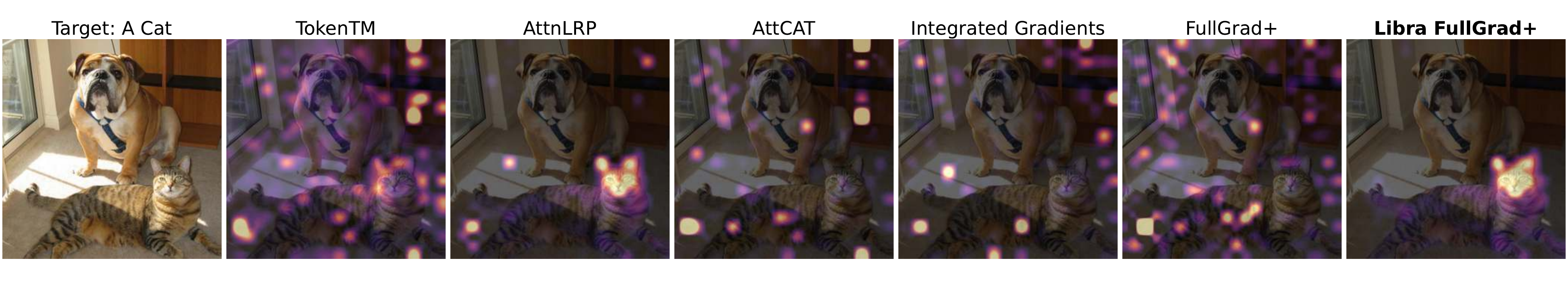,%
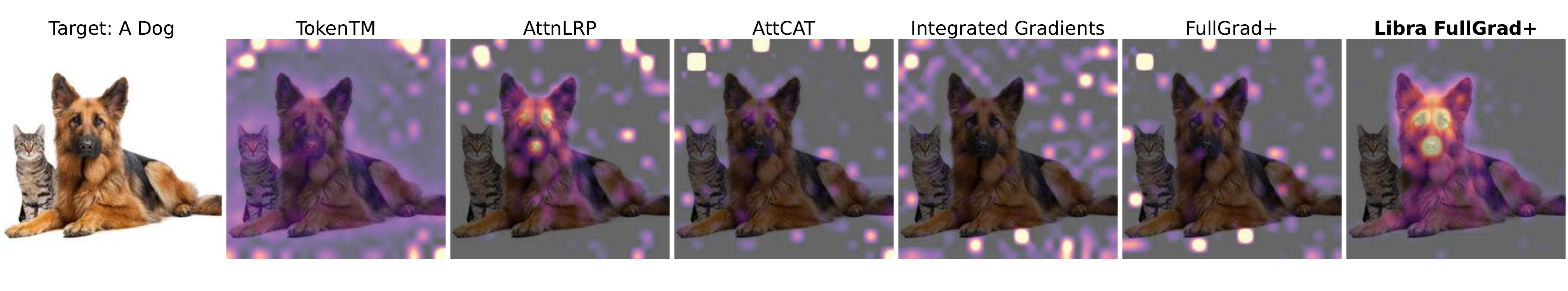,%
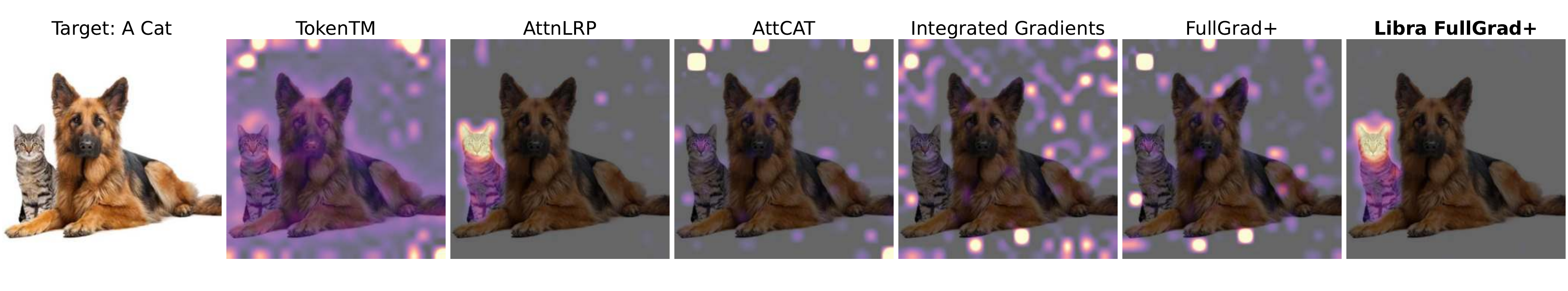,%
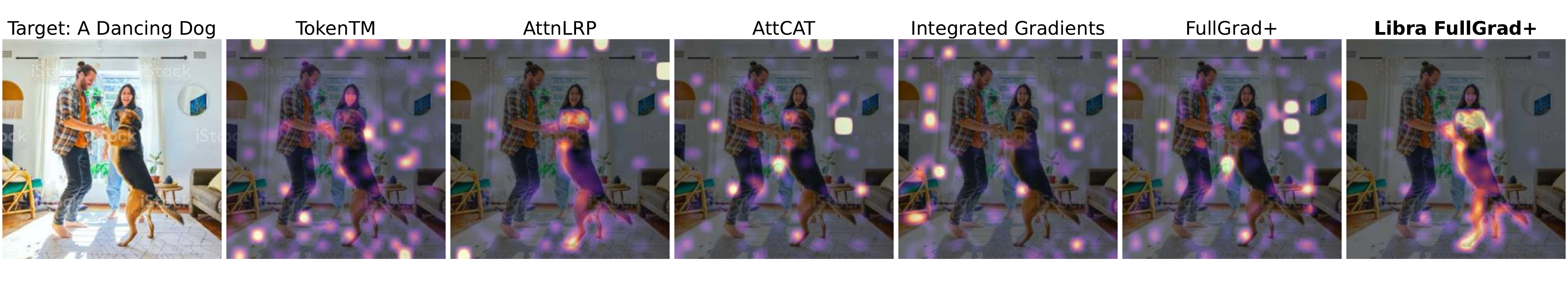,%
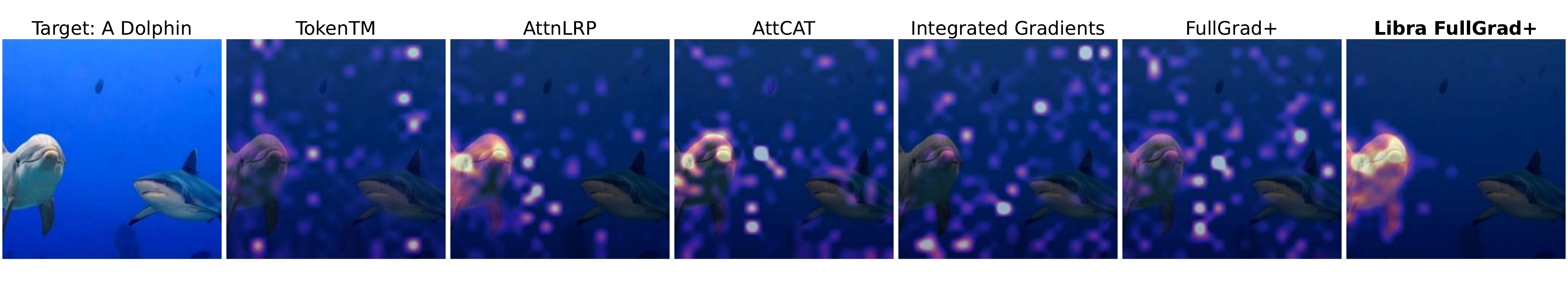,%
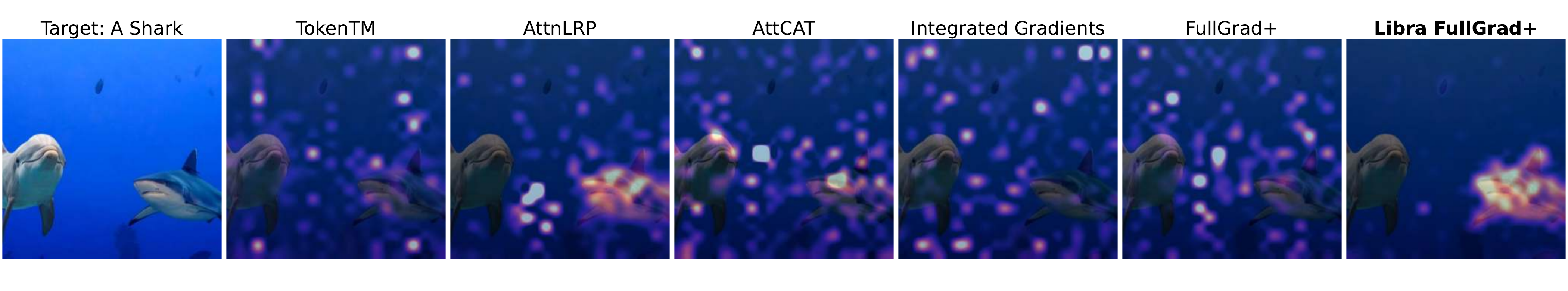,%
}{}{}

\CLIPRows[!b]{%
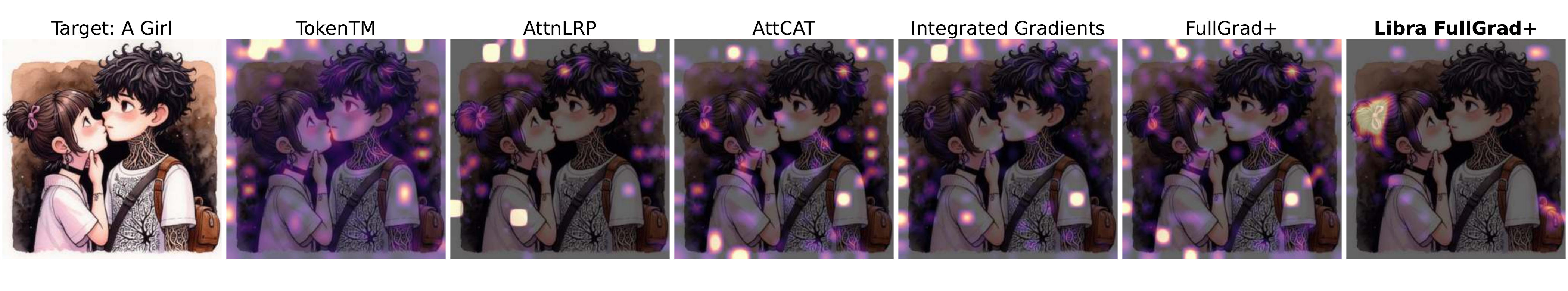,%
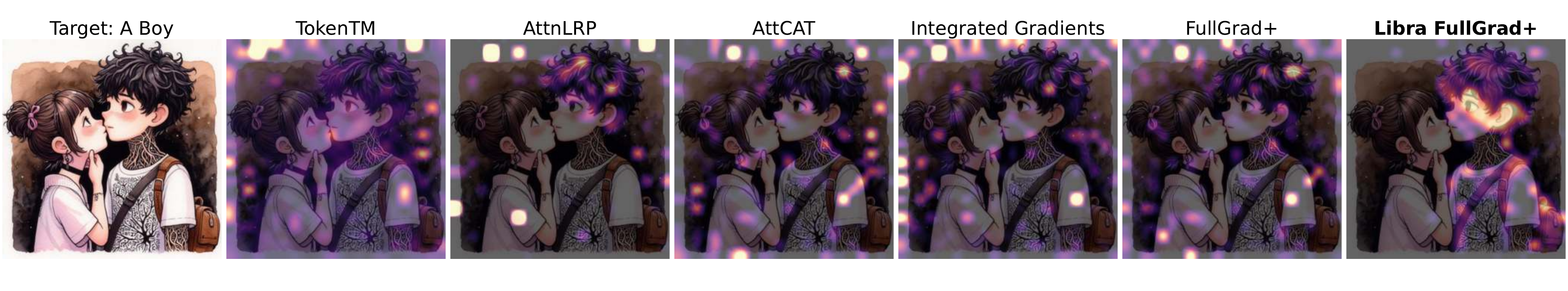,%
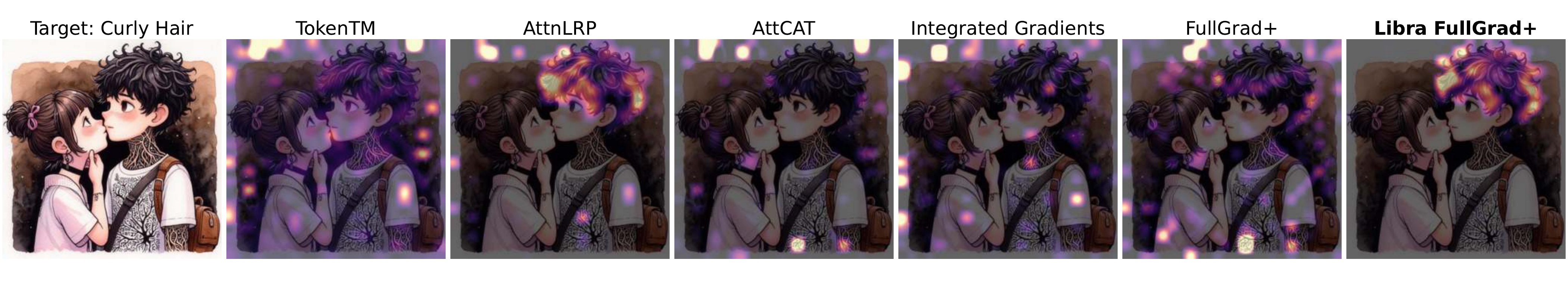,%
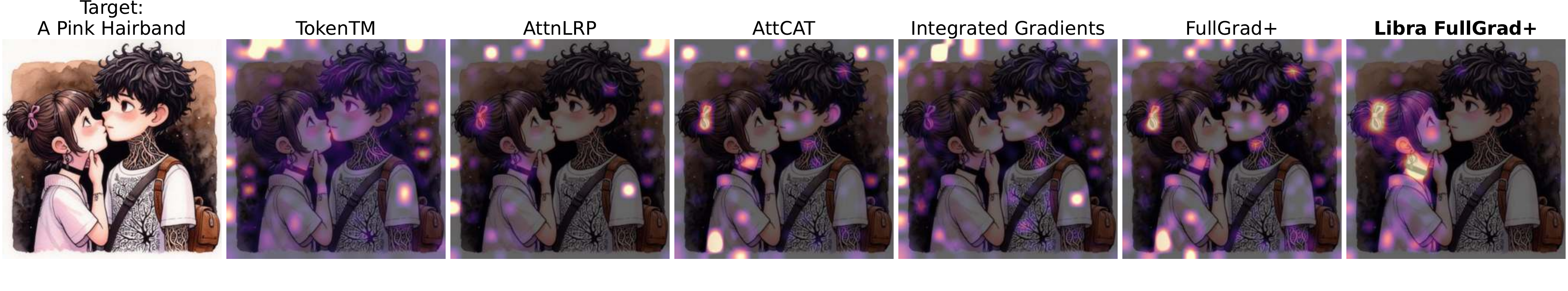,%
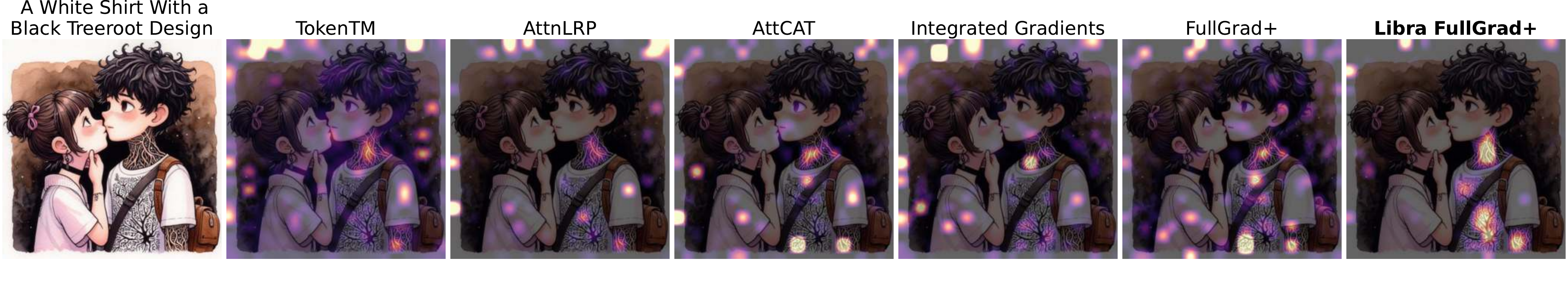,%
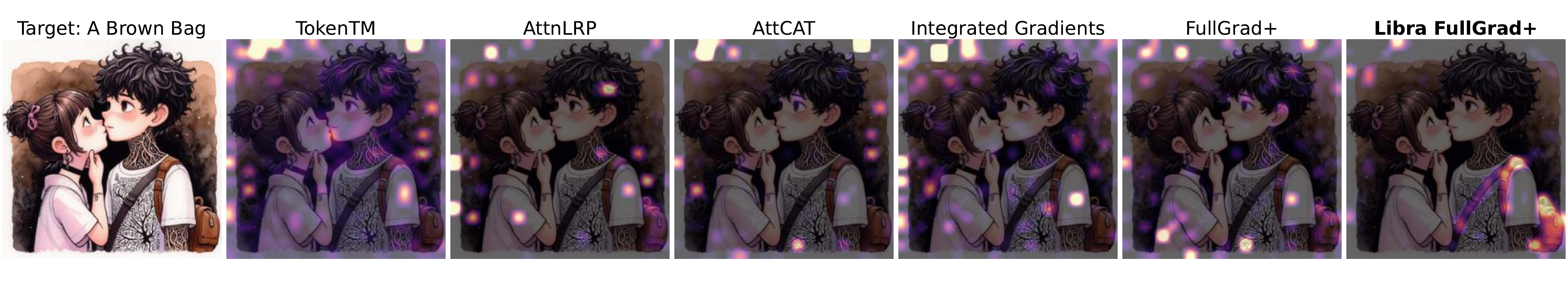,%
}{}{}

\endgroup

\clearpage{}
\subsection{A Comparative Study of Elephant-Zebra Multi-Class Attribution on COCO}
\label{apn:qual_zele}
Following Appendix~\ref{apn:qual_setup}, we assess attribution methods' ability to generate class-discriminative explanations on ImageNet-finetuned models, focusing on challenging scenes containing co-occurring elephants and zebras.

\newcommand{\ZeleAll}[1]{%
\CLIPRows[!b]{%
#1/zele_279208_elephant.pdf,%
#1/zele_279208_zebra.pdf,%
#1/zele_200133_elephant.pdf,%
#1/zele_200133_zebra.pdf,%
#1/zele_251577_elephant.pdf,%
#1/zele_251577_zebra.pdf,%
}{}{}
\CLIPRows[!b]{%
#1/zele_278763_elephant.pdf,%
#1/zele_278763_zebra.pdf,%
#1/zele_176226_elephant.pdf,%
#1/zele_176226_zebra.pdf,%
#1/zele_31569_elephant.pdf,%
#1/zele_31569_zebra.pdf,%
}{}{}
\CLIPRows[!b]{%
#1/zele_526534_elephant.pdf,%
#1/zele_526534_zebra.pdf,%
#1/zele_560326_elephant.pdf,%
#1/zele_560326_zebra.pdf,%
#1/zele_62263_elephant.pdf,%
#1/zele_62263_zebra.pdf,%
#1/zele_98949_elephant.pdf,%
#1/zele_98949_zebra.pdf,%
}{}{}
}

\subsubsection{Elephant-Zebra Qualitative Comparison on ViT-B}
\ZeleAll{files/qual_v5/vit_base_patch16_384.augreg_in1k/FairGrad/zele_3}

\clearpage{}
\subsubsection{Elephant-Zebra Qualitative Comparison on BEiT2-L}
\ZeleAll{files/qual_v5/beitv2_large_patch16_224.in1k_ft_in22k_in1k/FairGrad/zele_3}

\FloatBarrier{}

\clearpage{}
\FloatBarrier

\clearpage{}
\section{Quantitative Results}
\label{apn:quant}
\subsection{Comparison of Compositions With \FairGrad{} Versus \IG{}}
\label{sec:ig_compositional}

  \begingroup%
  \renewenvironment{table}[1][]%
    {\begin{table*}[#files/tables_v1/vit_large_patch16_224.augreg_in21k_ft_in1k/MIF/m_IG.tex]\small\setlength{\tabcolsep}{4pt}}%
    {\end{table*}}%
  \begin{table}[h]
\centering
\begin{tabular}{lccccc}
\toprule
Method & \multicolumn{2}{c}{MIF Deletion (GT)} & \multicolumn{2}{c}{MIF Deletion (Predicted)} & Segmentation \\
  & Accuracy & AOPC & Accuracy & AOPC & AP \\
\cmidrule(r){1-1}
\cmidrule(lr){2-3}
\cmidrule(lr){4-5}
\cmidrule(l){6-6}
Random & 36.9 \textcolor{gray}{±0.1} & 14.1 \textcolor{gray}{±0.2} & 29.5 \textcolor{gray}{±0.1} & 15.8 \textcolor{gray}{±0.2} & 42.0 \textcolor{gray}{±0.4} \\
RawAtt & 45.4 \textcolor{gray}{±0.1} & 22.9 \textcolor{gray}{±0.3} & 39.1 \textcolor{gray}{±0.1} & 25.3 \textcolor{gray}{±0.2} & 40.2 \textcolor{gray}{±0.4} \\
Attention Rollout & 39.0 \textcolor{gray}{±0.1} & 16.5 \textcolor{gray}{±0.3} & 31.4 \textcolor{gray}{±0.1} & 18.3 \textcolor{gray}{±0.3} & 39.9 \textcolor{gray}{±0.3} \\
AliLRP & 39.8 \textcolor{gray}{±0.1} & 17.2 \textcolor{gray}{±0.3} & 33.2 \textcolor{gray}{±0.1} & 19.2 \textcolor{gray}{±0.2} & 42.7 \textcolor{gray}{±0.4} \\
AttnLRP & 47.1 \textcolor{gray}{±0.1} & 24.8 \textcolor{gray}{±0.3} & 41.8 \textcolor{gray}{±0.1} & 27.6 \textcolor{gray}{±0.3} & 47.2 \textcolor{gray}{±0.3} \\
DecompX & 44.4 \textcolor{gray}{±0.1} & 22.6 \textcolor{gray}{±0.3} & 38.9 \textcolor{gray}{±0.1} & 25.3 \textcolor{gray}{±0.3} & 54.2 \textcolor{gray}{±0.3} \\
TokenTM & 54.9 \textcolor{gray}{±0.1} & 31.8 \textcolor{gray}{±0.3} & 50.0 \textcolor{gray}{±0.1} & 34.9 \textcolor{gray}{±0.3} & 50.0 \textcolor{gray}{±0.3} \\
\midrule
\IxG{}  & 40.1 \textcolor{gray}{±0.1} & 17.5 \textcolor{gray}{±0.3} & 33.9 \textcolor{gray}{±0.1} & 19.6 \textcolor{gray}{±0.2} & 43.6 \textcolor{gray}{±0.4} \\
Int. Gradients & 46.3 \textcolor{gray}{±0.1} (\textcolor{blue}{+15.4\%}) & 23.1 \textcolor{gray}{±0.3} (\textcolor{blue}{+32.1\%}) & 35.9 \textcolor{gray}{±0.1} (\textcolor{blue}{+6.1\%}) & 21.9 \textcolor{gray}{±0.2} (\textcolor{blue}{+11.6\%}) & 46.6 \textcolor{gray}{±0.3} (\textcolor{blue}{+6.9\%}) \\
\textbf{Libra \IxG{} } & 45.9 \textcolor{gray}{±0.1} (\textcolor{blue}{+14.4\%}) & 23.4 \textcolor{gray}{±0.3} (\textcolor{blue}{+33.5\%}) & 40.5 \textcolor{gray}{±0.1} (\textcolor{blue}{+19.6\%}) & 26.1 \textcolor{gray}{±0.3} (\textcolor{blue}{+33.1\%}) & 53.6 \textcolor{gray}{±0.3} (\textcolor{blue}{+22.9\%}) \\
\midrule
AttCAT & 48.7 \textcolor{gray}{±0.1} & 25.7 \textcolor{gray}{±0.3} & 44.8 \textcolor{gray}{±0.1} & 29.0 \textcolor{gray}{±0.3} & 44.9 \textcolor{gray}{±0.3} \\
Int. AttCAT & 53.4 \textcolor{gray}{±0.1} (\textcolor{blue}{+9.7\%}) & 29.3 \textcolor{gray}{±0.3} (\textcolor{blue}{+13.9\%}) & 43.2 \textcolor{gray}{±0.1} (\textcolor{coral}{-3.6\%}) & 27.7 \textcolor{gray}{±0.3} (\textcolor{coral}{-4.2\%}) & 50.3 \textcolor{gray}{±0.3} (\textcolor{blue}{+12.1\%}) \\
\textbf{Libra AttCAT} & \underline{64.7} \textcolor{gray}{±0.1} (\textcolor{blue}{+33.0\%}) & \underline{40.5} \textcolor{gray}{±0.3} (\textcolor{blue}{+57.3\%}) & \underline{61.3} \textcolor{gray}{±0.1} (\textcolor{blue}{+36.9\%}) & \underline{44.5} \textcolor{gray}{±0.3} (\textcolor{blue}{+53.6\%}) & 53.3 \textcolor{gray}{±0.3} (\textcolor{blue}{+18.8\%}) \\
\midrule
GenAtt & 56.4 \textcolor{gray}{±0.1} & 33.2 \textcolor{gray}{±0.3} & 51.8 \textcolor{gray}{±0.1} & 36.5 \textcolor{gray}{±0.3} & 50.9 \textcolor{gray}{±0.3} \\
Int. GenAtt & 52.7 \textcolor{gray}{±0.1} (\textcolor{coral}{-6.6\%}) & 29.3 \textcolor{gray}{±0.4} (\textcolor{coral}{-11.9\%}) & 43.6 \textcolor{gray}{±0.1} (\textcolor{coral}{-15.9\%}) & 28.6 \textcolor{gray}{±0.3} (\textcolor{coral}{-21.5\%}) & 49.1 \textcolor{gray}{±0.3} (\textcolor{coral}{-3.6\%}) \\
\textbf{Libra GenAtt} & 59.7 \textcolor{gray}{±0.1} (\textcolor{blue}{+5.9\%}) & 36.2 \textcolor{gray}{±0.3} (\textcolor{blue}{+8.9\%}) & 55.4 \textcolor{gray}{±0.1} (\textcolor{blue}{+6.8\%}) & 39.6 \textcolor{gray}{±0.3} (\textcolor{blue}{+8.7\%}) & 58.6 \textcolor{gray}{±0.3} (\textcolor{blue}{+15.1\%}) \\
\midrule
TokenTM & 54.9 \textcolor{gray}{±0.1} & 31.8 \textcolor{gray}{±0.3} & 50.0 \textcolor{gray}{±0.1} & 34.9 \textcolor{gray}{±0.3} & 50.0 \textcolor{gray}{±0.3} \\
Int. TokenTM & 53.3 \textcolor{gray}{±0.1} (\textcolor{coral}{-2.8\%}) & 30.3 \textcolor{gray}{±0.3} (\textcolor{coral}{-4.9\%}) & 46.4 \textcolor{gray}{±0.1} (\textcolor{coral}{-7.2\%}) & 31.7 \textcolor{gray}{±0.3} (\textcolor{coral}{-9.3\%}) & 49.5 \textcolor{gray}{±0.3} (\textcolor{coral}{-0.9\%}) \\
\textbf{Libra TokenTM} & 57.3 \textcolor{gray}{±0.1} (\textcolor{blue}{+4.5\%}) & 34.2 \textcolor{gray}{±0.3} (\textcolor{blue}{+7.4\%}) & 52.5 \textcolor{gray}{±0.1} (\textcolor{blue}{+5.0\%}) & 37.4 \textcolor{gray}{±0.3} (\textcolor{blue}{+7.1\%}) & 53.9 \textcolor{gray}{±0.3} (\textcolor{blue}{+7.9\%}) \\
\midrule
GradCAM+ & 53.4 \textcolor{gray}{±0.1} & 30.0 \textcolor{gray}{±0.3} & 48.6 \textcolor{gray}{±0.1} & 33.0 \textcolor{gray}{±0.2} & 52.1 \textcolor{gray}{±0.4} \\
Int. GradCAM+ & 47.9 \textcolor{gray}{±0.1} (\textcolor{coral}{-10.3\%}) & 24.1 \textcolor{gray}{±0.2} (\textcolor{coral}{-19.8\%}) & 41.4 \textcolor{gray}{±0.1} (\textcolor{coral}{-14.7\%}) & 25.8 \textcolor{gray}{±0.3} (\textcolor{coral}{-21.7\%}) & 50.0 \textcolor{gray}{±0.4} (\textcolor{coral}{-4.0\%}) \\
\textbf{Libra GradCAM+} & 60.9 \textcolor{gray}{±0.1} (\textcolor{blue}{+14.0\%}) & 36.7 \textcolor{gray}{±0.3} (\textcolor{blue}{+22.0\%}) & 56.5 \textcolor{gray}{±0.1} (\textcolor{blue}{+16.2\%}) & 40.1 \textcolor{gray}{±0.3} (\textcolor{blue}{+21.8\%}) & 60.2 \textcolor{gray}{±0.4} (\textcolor{blue}{+15.5\%}) \\
\midrule
HiResCAM & 32.7 \textcolor{gray}{±0.1} & 10.6 \textcolor{gray}{±0.2} & 25.7 \textcolor{gray}{±0.1} & 12.2 \textcolor{gray}{±0.2} & 38.5 \textcolor{gray}{±0.4} \\
Int. HiResCAM & 31.2 \textcolor{gray}{±0.1} (\textcolor{coral}{-4.5\%}) & 9.1 \textcolor{gray}{±0.3} (\textcolor{coral}{-14.0\%}) & 26.4 \textcolor{gray}{±0.1} (\textcolor{blue}{+2.8\%}) & 12.4 \textcolor{gray}{±0.2} (\textcolor{blue}{+1.2\%}) & 38.4 \textcolor{gray}{±0.4} (\textcolor{coral}{-0.2\%}) \\
\textbf{Libra HiResCAM} & 54.0 \textcolor{gray}{±0.1} (\textcolor{blue}{+65.2\%}) & 30.2 \textcolor{gray}{±0.3} (\textcolor{blue}{+186.3\%}) & 49.0 \textcolor{gray}{±0.1} (\textcolor{blue}{+90.7\%}) & 33.2 \textcolor{gray}{±0.3} (\textcolor{blue}{+171.8\%}) & 48.0 \textcolor{gray}{±0.3} (\textcolor{blue}{+24.8\%}) \\
\midrule
XGradCAM+ & 50.9 \textcolor{gray}{±0.1} & 27.7 \textcolor{gray}{±0.3} & 45.9 \textcolor{gray}{±0.1} & 30.5 \textcolor{gray}{±0.3} & 46.9 \textcolor{gray}{±0.4} \\
Int. XGradCAM+ & 48.4 \textcolor{gray}{±0.1} (\textcolor{coral}{-4.9\%}) & 24.7 \textcolor{gray}{±0.2} (\textcolor{coral}{-10.7\%}) & 40.2 \textcolor{gray}{±0.1} (\textcolor{coral}{-12.3\%}) & 25.2 \textcolor{gray}{±0.3} (\textcolor{coral}{-17.6\%}) & 48.0 \textcolor{gray}{±0.4} (\textcolor{blue}{+2.4\%}) \\
\textbf{Libra XGradCAM+} & 63.0 \textcolor{gray}{±0.1} (\textcolor{blue}{+23.6\%}) & 38.6 \textcolor{gray}{±0.3} (\textcolor{blue}{+39.2\%}) & 58.8 \textcolor{gray}{±0.1} (\textcolor{blue}{+28.1\%}) & 42.2 \textcolor{gray}{±0.3} (\textcolor{blue}{+38.3\%}) & \underline{60.3} \textcolor{gray}{±0.4} (\textcolor{blue}{+28.6\%}) \\
\midrule
FullGrad+ & 49.1 \textcolor{gray}{±0.1} & 25.8 \textcolor{gray}{±0.3} & 45.1 \textcolor{gray}{±0.1} & 28.9 \textcolor{gray}{±0.3} & 44.2 \textcolor{gray}{±0.3} \\
Int. FullGrad+ & 52.5 \textcolor{gray}{±0.1} (\textcolor{blue}{+7.0\%}) & 28.3 \textcolor{gray}{±0.3} (\textcolor{blue}{+9.5\%}) & 42.1 \textcolor{gray}{±0.1} (\textcolor{coral}{-6.6\%}) & 26.6 \textcolor{gray}{±0.3} (\textcolor{coral}{-7.9\%}) & 49.1 \textcolor{gray}{±0.3} (\textcolor{blue}{+11.2\%}) \\
\textbf{Libra FullGrad+} & \textbf{65.5} \textcolor{gray}{±0.1} (\textcolor{blue}{+33.5\%}) & \textbf{41.2} \textcolor{gray}{±0.3} (\textcolor{blue}{+59.5\%}) & \textbf{62.4} \textcolor{gray}{±0.1} (\textcolor{blue}{+38.5\%}) & \textbf{45.3} \textcolor{gray}{±0.3} (\textcolor{blue}{+56.5\%}) & \textbf{64.5} \textcolor{gray}{±0.3} (\textcolor{blue}{+46.0\%}) \\
\bottomrule
\end{tabular}
\caption{Comparison of gradient-based attribution methods and their compositions with LibraGrad and Integrated Gradients (Int. Gradients, IG) on the ViT-L model. Metrics reported are faithfulness (Most-Influential-First Deletion, MIF) and Segmentation Average Precision (AP). The results demonstrate that composing with LibraGrad universally enhances the performance of existing methods more effectively than composing with IG.}
\label{tbl:MIF_m_IG}
\end{table}
  \endgroup%

  \begingroup%
  \renewenvironment{table}[1][]%
    {\begin{table*}[#files/tables_v1/vit_large_patch16_224.augreg_in21k_ft_in1k/LIF/m_IG.tex]}%
    {\end{table*}}%
  \begin{table}[h]
\centering
\begin{tabular}{lcccc}
\toprule
Method & \multicolumn{2}{c}{LIF Deletion (GT)} & \multicolumn{2}{c}{LIF Deletion (Predicted)} \\
  & Accuracy & AOPC & Accuracy & AOPC \\
\cmidrule(r){1-1}
\cmidrule(lr){2-3}
\cmidrule(l){4-5}
Random & 62.9 \textcolor{gray}{±0.1} & 85.4 \textcolor{gray}{±0.2} & 70.2 \textcolor{gray}{±0.1} & 83.7 \textcolor{gray}{±0.2} \\
RawAtt & 60.3 \textcolor{gray}{±0.1} & 83.3 \textcolor{gray}{±0.2} & 67.6 \textcolor{gray}{±0.1} & 81.5 \textcolor{gray}{±0.1} \\
Attention Rollout & 61.9 \textcolor{gray}{±0.1} & 84.1 \textcolor{gray}{±0.2} & 68.3 \textcolor{gray}{±0.1} & 81.9 \textcolor{gray}{±0.2} \\
AliLRP & 65.4 \textcolor{gray}{±0.1} & 87.7 \textcolor{gray}{±0.2} & 72.5 \textcolor{gray}{±0.1} & 85.9 \textcolor{gray}{±0.2} \\
AttnLRP & 70.3 \textcolor{gray}{±0.1} & 92.9 \textcolor{gray}{±0.2} & 77.6 \textcolor{gray}{±0.1} & 91.3 \textcolor{gray}{±0.2} \\
DecompX & 68.8 \textcolor{gray}{±0.1} & 91.0 \textcolor{gray}{±0.2} & 75.8 \textcolor{gray}{±0.1} & 89.3 \textcolor{gray}{±0.2} \\
TokenTM & 68.9 \textcolor{gray}{±0.1} & 91.6 \textcolor{gray}{±0.2} & 77.3 \textcolor{gray}{±0.1} & 90.3 \textcolor{gray}{±0.2} \\
\midrule
\IxG{}  & 65.8 \textcolor{gray}{±0.1} & 88.4 \textcolor{gray}{±0.2} & 72.8 \textcolor{gray}{±0.1} & 86.7 \textcolor{gray}{±0.1} \\
Int. Gradients & 71.1 \textcolor{gray}{±0.1} (\textcolor{blue}{+8.1\%}) & 93.3 \textcolor{gray}{±0.2} (\textcolor{blue}{+5.5\%}) & 73.5 \textcolor{gray}{±0.1} (\textcolor{blue}{+0.9\%}) & 88.4 \textcolor{gray}{±0.2} (\textcolor{blue}{+1.9\%}) \\
\textbf{Libra \IxG{} } & 70.1 \textcolor{gray}{±0.1} (\textcolor{blue}{+6.6\%}) & 92.0 \textcolor{gray}{±0.2} (\textcolor{blue}{+4.0\%}) & 76.7 \textcolor{gray}{±0.1} (\textcolor{blue}{+5.4\%}) & 90.2 \textcolor{gray}{±0.2} (\textcolor{blue}{+4.0\%}) \\
\midrule
AttCAT & 71.8 \textcolor{gray}{±0.1} & 94.3 \textcolor{gray}{±0.2} & 77.5 \textcolor{gray}{±0.1} & 92.6 \textcolor{gray}{±0.2} \\
Int. AttCAT & 75.2 \textcolor{gray}{±0.1} (\textcolor{blue}{+4.8\%}) & 97.5 \textcolor{gray}{±0.2} (\textcolor{blue}{+3.5\%}) & 76.6 \textcolor{gray}{±0.1} (\textcolor{coral}{-1.1\%}) & 92.2 \textcolor{gray}{±0.2} (\textcolor{coral}{-0.5\%}) \\
\textbf{Libra AttCAT} & \underline{76.3} \textcolor{gray}{±0.1} (\textcolor{blue}{+6.2\%}) & \underline{98.5} \textcolor{gray}{±0.2} (\textcolor{blue}{+4.5\%}) & \underline{82.2} \textcolor{gray}{±0.1} (\textcolor{blue}{+6.1\%}) & \underline{97.1} \textcolor{gray}{±0.2} (\textcolor{blue}{+4.8\%}) \\
\midrule
GenAtt & 70.0 \textcolor{gray}{±0.1} & 92.8 \textcolor{gray}{±0.2} & 78.2 \textcolor{gray}{±0.1} & 91.5 \textcolor{gray}{±0.2} \\
Int. GenAtt & 69.3 \textcolor{gray}{±0.1} (\textcolor{coral}{-1.0\%}) & 91.7 \textcolor{gray}{±0.2} (\textcolor{coral}{-1.1\%}) & 74.6 \textcolor{gray}{±0.1} (\textcolor{coral}{-4.5\%}) & 88.0 \textcolor{gray}{±0.2} (\textcolor{coral}{-3.8\%}) \\
\textbf{Libra GenAtt} & 70.9 \textcolor{gray}{±0.1} (\textcolor{blue}{+1.3\%}) & 93.2 \textcolor{gray}{±0.2} (\textcolor{blue}{+0.5\%}) & 78.8 \textcolor{gray}{±0.1} (\textcolor{blue}{+0.7\%}) & 92.0 \textcolor{gray}{±0.2} (\textcolor{blue}{+0.5\%}) \\
\midrule
TokenTM & 68.9 \textcolor{gray}{±0.1} & 91.6 \textcolor{gray}{±0.2} & 77.3 \textcolor{gray}{±0.1} & 90.3 \textcolor{gray}{±0.2} \\
Int. TokenTM & 69.0 \textcolor{gray}{±0.1} (\textcolor{blue}{+0.2\%}) & 91.5 \textcolor{gray}{±0.2} (\textcolor{coral}{-0.1\%}) & 76.1 \textcolor{gray}{±0.1} (\textcolor{coral}{-1.5\%}) & 89.0 \textcolor{gray}{±0.2} (\textcolor{coral}{-1.4\%}) \\
\textbf{Libra TokenTM} & 69.4 \textcolor{gray}{±0.1} (\textcolor{blue}{+0.8\%}) & 92.1 \textcolor{gray}{±0.2} (\textcolor{blue}{+0.5\%}) & 77.8 \textcolor{gray}{±0.1} (\textcolor{blue}{+0.7\%}) & 90.8 \textcolor{gray}{±0.2} (\textcolor{blue}{+0.6\%}) \\
\midrule
GradCAM+ & 70.5 \textcolor{gray}{±0.1} & 92.9 \textcolor{gray}{±0.2} & 76.8 \textcolor{gray}{±0.1} & 91.0 \textcolor{gray}{±0.2} \\
Int. GradCAM+ & 69.0 \textcolor{gray}{±0.1} (\textcolor{coral}{-2.2\%}) & 91.0 \textcolor{gray}{±0.2} (\textcolor{coral}{-2.1\%}) & 73.2 \textcolor{gray}{±0.1} (\textcolor{coral}{-4.7\%}) & 87.6 \textcolor{gray}{±0.2} (\textcolor{coral}{-3.7\%}) \\
\textbf{Libra GradCAM+} & 72.6 \textcolor{gray}{±0.1} (\textcolor{blue}{+2.9\%}) & 94.4 \textcolor{gray}{±0.2} (\textcolor{blue}{+1.6\%}) & 79.1 \textcolor{gray}{±0.1} (\textcolor{blue}{+3.0\%}) & 92.7 \textcolor{gray}{±0.2} (\textcolor{blue}{+1.8\%}) \\
\midrule
HiResCAM & 53.6 \textcolor{gray}{±0.1} & 76.7 \textcolor{gray}{±0.2} & 59.3 \textcolor{gray}{±0.1} & 74.2 \textcolor{gray}{±0.3} \\
Int. HiResCAM & 50.7 \textcolor{gray}{±0.1} (\textcolor{coral}{-5.5\%}) & 74.3 \textcolor{gray}{±0.3} (\textcolor{coral}{-3.2\%}) & 60.4 \textcolor{gray}{±0.1} (\textcolor{blue}{+1.9\%}) & 75.0 \textcolor{gray}{±0.3} (\textcolor{blue}{+1.0\%}) \\
\textbf{Libra HiResCAM} & 67.4 \textcolor{gray}{±0.1} (\textcolor{blue}{+25.7\%}) & 90.0 \textcolor{gray}{±0.2} (\textcolor{blue}{+17.3\%}) & 73.8 \textcolor{gray}{±0.1} (\textcolor{blue}{+24.4\%}) & 88.0 \textcolor{gray}{±0.2} (\textcolor{blue}{+18.6\%}) \\
\midrule
XGradCAM+ & 69.5 \textcolor{gray}{±0.1} & 92.1 \textcolor{gray}{±0.2} & 75.7 \textcolor{gray}{±0.1} & 90.1 \textcolor{gray}{±0.2} \\
Int. XGradCAM+ & 69.1 \textcolor{gray}{±0.1} (\textcolor{coral}{-0.6\%}) & 91.1 \textcolor{gray}{±0.2} (\textcolor{coral}{-1.0\%}) & 72.2 \textcolor{gray}{±0.1} (\textcolor{coral}{-4.7\%}) & 86.8 \textcolor{gray}{±0.2} (\textcolor{coral}{-3.7\%}) \\
\textbf{Libra XGradCAM+} & 73.5 \textcolor{gray}{±0.1} (\textcolor{blue}{+5.7\%}) & 95.3 \textcolor{gray}{±0.2} (\textcolor{blue}{+3.5\%}) & 80.0 \textcolor{gray}{±0.1} (\textcolor{blue}{+5.6\%}) & 93.7 \textcolor{gray}{±0.2} (\textcolor{blue}{+3.9\%}) \\
\midrule
FullGrad+ & 71.5 \textcolor{gray}{±0.1} & 93.8 \textcolor{gray}{±0.2} & 76.8 \textcolor{gray}{±0.1} & 91.8 \textcolor{gray}{±0.2} \\
Int. FullGrad+ & 74.8 \textcolor{gray}{±0.1} (\textcolor{blue}{+4.7\%}) & 97.1 \textcolor{gray}{±0.2} (\textcolor{blue}{+3.5\%}) & 76.0 \textcolor{gray}{±0.1} (\textcolor{coral}{-1.0\%}) & 91.5 \textcolor{gray}{±0.2} (\textcolor{coral}{-0.4\%}) \\
\textbf{Libra FullGrad+} & \textbf{76.8} \textcolor{gray}{±0.1} (\textcolor{blue}{+7.5\%}) & \textbf{98.9} \textcolor{gray}{±0.2} (\textcolor{blue}{+5.4\%}) & \textbf{82.6} \textcolor{gray}{±0.1} (\textcolor{blue}{+7.6\%}) & \textbf{97.4} \textcolor{gray}{±0.2} (\textcolor{blue}{+6.0\%}) \\
\bottomrule
\end{tabular}
\caption{Comparison of gradient-based attribution methods and their compositions with LibraGrad and IG on the ViT-L model.}
\label{tbl:LIF_m_IG}
\end{table}
  \endgroup%

  \begingroup%
  \renewenvironment{table}[1][]%
    {\begin{table*}[#files/tables_v1/vit_large_patch16_224.augreg_in21k_ft_in1k/SRG/m_IG.tex]}%
    {\end{table*}}%
  \begin{table}[h]
\centering
\begin{tabular}{lcccc}
\toprule
Method & \multicolumn{2}{c}{SRG (GT)} & \multicolumn{2}{c}{SRG (Predicted)} \\
  & Accuracy & AOPC & Accuracy & AOPC \\
\cmidrule(r){1-1}
\cmidrule(lr){2-3}
\cmidrule(l){4-5}
Random & 49.9 \textcolor{gray}{±0.1} & 49.7 \textcolor{gray}{±0.2} & 49.8 \textcolor{gray}{±0.1} & 49.8 \textcolor{gray}{±0.2} \\
RawAtt & 52.9 \textcolor{gray}{±0.1} & 53.1 \textcolor{gray}{±0.2} & 53.3 \textcolor{gray}{±0.1} & 53.4 \textcolor{gray}{±0.2} \\
Attention Rollout & 50.4 \textcolor{gray}{±0.1} & 50.3 \textcolor{gray}{±0.3} & 49.9 \textcolor{gray}{±0.1} & 50.1 \textcolor{gray}{±0.2} \\
AliLRP & 52.6 \textcolor{gray}{±0.1} & 52.4 \textcolor{gray}{±0.2} & 52.8 \textcolor{gray}{±0.1} & 52.5 \textcolor{gray}{±0.2} \\
AttnLRP & 58.7 \textcolor{gray}{±0.1} & 58.8 \textcolor{gray}{±0.3} & 59.7 \textcolor{gray}{±0.1} & 59.5 \textcolor{gray}{±0.2} \\
DecompX & 56.6 \textcolor{gray}{±0.1} & 56.8 \textcolor{gray}{±0.3} & 57.4 \textcolor{gray}{±0.1} & 57.3 \textcolor{gray}{±0.2} \\
TokenTM & 61.9 \textcolor{gray}{±0.1} & 61.7 \textcolor{gray}{±0.3} & 63.6 \textcolor{gray}{±0.1} & 62.6 \textcolor{gray}{±0.2} \\
\midrule
\IxG{}  & 53.0 \textcolor{gray}{±0.1} & 53.0 \textcolor{gray}{±0.2} & 53.3 \textcolor{gray}{±0.1} & 53.2 \textcolor{gray}{±0.2} \\
Int. Gradients & 58.7 \textcolor{gray}{±0.1} (\textcolor{blue}{+10.9\%}) & 58.2 \textcolor{gray}{±0.3} (\textcolor{blue}{+9.9\%}) & 54.7 \textcolor{gray}{±0.1} (\textcolor{blue}{+2.6\%}) & 55.1 \textcolor{gray}{±0.2} (\textcolor{blue}{+3.7\%}) \\
\textbf{Libra \IxG{} } & 58.0 \textcolor{gray}{±0.1} (\textcolor{blue}{+9.5\%}) & 57.7 \textcolor{gray}{±0.3} (\textcolor{blue}{+8.9\%}) & 58.6 \textcolor{gray}{±0.1} (\textcolor{blue}{+9.9\%}) & 58.2 \textcolor{gray}{±0.2} (\textcolor{blue}{+9.4\%}) \\
\midrule
AttCAT & 60.2 \textcolor{gray}{±0.1} & 60.0 \textcolor{gray}{±0.2} & 61.2 \textcolor{gray}{±0.1} & 60.8 \textcolor{gray}{±0.2} \\
Int. AttCAT & 64.3 \textcolor{gray}{±0.1} (\textcolor{blue}{+6.8\%}) & 63.4 \textcolor{gray}{±0.2} (\textcolor{blue}{+5.7\%}) & 59.9 \textcolor{gray}{±0.1} (\textcolor{coral}{-2.0\%}) & 60.0 \textcolor{gray}{±0.2} (\textcolor{coral}{-1.4\%}) \\
\textbf{Libra AttCAT} & \underline{70.5} \textcolor{gray}{±0.1} (\textcolor{blue}{+17.0\%}) & \underline{69.5} \textcolor{gray}{±0.3} (\textcolor{blue}{+15.8\%}) & \underline{71.8} \textcolor{gray}{±0.1} (\textcolor{blue}{+17.4\%}) & \underline{70.8} \textcolor{gray}{±0.2} (\textcolor{blue}{+16.4\%}) \\
\midrule
GenAtt & 63.2 \textcolor{gray}{±0.1} & 63.0 \textcolor{gray}{±0.2} & 65.0 \textcolor{gray}{±0.1} & 64.0 \textcolor{gray}{±0.2} \\
Int. GenAtt & 61.0 \textcolor{gray}{±0.1} (\textcolor{coral}{-3.5\%}) & 60.5 \textcolor{gray}{±0.3} (\textcolor{coral}{-4.0\%}) & 59.1 \textcolor{gray}{±0.1} (\textcolor{coral}{-9.1\%}) & 58.3 \textcolor{gray}{±0.3} (\textcolor{coral}{-8.9\%}) \\
\textbf{Libra GenAtt} & 65.3 \textcolor{gray}{±0.1} (\textcolor{blue}{+3.3\%}) & 64.7 \textcolor{gray}{±0.3} (\textcolor{blue}{+2.7\%}) & 67.1 \textcolor{gray}{±0.1} (\textcolor{blue}{+3.2\%}) & 65.8 \textcolor{gray}{±0.3} (\textcolor{blue}{+2.8\%}) \\
\midrule
TokenTM & 61.9 \textcolor{gray}{±0.1} & 61.7 \textcolor{gray}{±0.3} & 63.6 \textcolor{gray}{±0.1} & 62.6 \textcolor{gray}{±0.2} \\
Int. TokenTM & 61.2 \textcolor{gray}{±0.1} (\textcolor{coral}{-1.1\%}) & 60.9 \textcolor{gray}{±0.3} (\textcolor{coral}{-1.3\%}) & 61.2 \textcolor{gray}{±0.1} (\textcolor{coral}{-3.7\%}) & 60.3 \textcolor{gray}{±0.2} (\textcolor{coral}{-3.6\%}) \\
\textbf{Libra TokenTM} & 63.4 \textcolor{gray}{±0.1} (\textcolor{blue}{+2.4\%}) & 63.1 \textcolor{gray}{±0.3} (\textcolor{blue}{+2.3\%}) & 65.2 \textcolor{gray}{±0.1} (\textcolor{blue}{+2.4\%}) & 64.1 \textcolor{gray}{±0.3} (\textcolor{blue}{+2.4\%}) \\
\midrule
GradCAM+ & 62.0 \textcolor{gray}{±0.1} & 61.5 \textcolor{gray}{±0.3} & 62.7 \textcolor{gray}{±0.1} & 62.0 \textcolor{gray}{±0.2} \\
Int. GradCAM+ & 58.5 \textcolor{gray}{±0.1} (\textcolor{coral}{-5.7\%}) & 57.5 \textcolor{gray}{±0.2} (\textcolor{coral}{-6.4\%}) & 57.3 \textcolor{gray}{±0.1} (\textcolor{coral}{-8.6\%}) & 56.7 \textcolor{gray}{±0.3} (\textcolor{coral}{-8.5\%}) \\
\textbf{Libra GradCAM+} & 66.7 \textcolor{gray}{±0.1} (\textcolor{blue}{+7.7\%}) & 65.5 \textcolor{gray}{±0.3} (\textcolor{blue}{+6.6\%}) & 67.8 \textcolor{gray}{±0.1} (\textcolor{blue}{+8.1\%}) & 66.4 \textcolor{gray}{±0.2} (\textcolor{blue}{+7.2\%}) \\
\midrule
HiResCAM & 43.2 \textcolor{gray}{±0.1} & 43.6 \textcolor{gray}{±0.2} & 42.5 \textcolor{gray}{±0.1} & 43.2 \textcolor{gray}{±0.2} \\
Int. HiResCAM & 41.0 \textcolor{gray}{±0.1} (\textcolor{coral}{-5.1\%}) & 41.7 \textcolor{gray}{±0.3} (\textcolor{coral}{-4.5\%}) & 43.4 \textcolor{gray}{±0.1} (\textcolor{blue}{+2.2\%}) & 43.7 \textcolor{gray}{±0.3} (\textcolor{blue}{+1.0\%}) \\
\textbf{Libra HiResCAM} & 60.7 \textcolor{gray}{±0.1} (\textcolor{blue}{+40.7\%}) & 60.1 \textcolor{gray}{±0.2} (\textcolor{blue}{+37.7\%}) & 61.4 \textcolor{gray}{±0.1} (\textcolor{blue}{+44.4\%}) & 60.6 \textcolor{gray}{±0.2} (\textcolor{blue}{+40.3\%}) \\
\midrule
XGradCAM+ & 60.2 \textcolor{gray}{±0.1} & 59.9 \textcolor{gray}{±0.3} & 60.8 \textcolor{gray}{±0.1} & 60.3 \textcolor{gray}{±0.2} \\
Int. XGradCAM+ & 58.8 \textcolor{gray}{±0.1} (\textcolor{coral}{-2.4\%}) & 57.9 \textcolor{gray}{±0.2} (\textcolor{coral}{-3.3\%}) & 56.2 \textcolor{gray}{±0.1} (\textcolor{coral}{-7.5\%}) & 56.0 \textcolor{gray}{±0.3} (\textcolor{coral}{-7.2\%}) \\
\textbf{Libra XGradCAM+} & 68.2 \textcolor{gray}{±0.1} (\textcolor{blue}{+13.3\%}) & 66.9 \textcolor{gray}{±0.3} (\textcolor{blue}{+11.8\%}) & 69.4 \textcolor{gray}{±0.1} (\textcolor{blue}{+14.1\%}) & 68.0 \textcolor{gray}{±0.3} (\textcolor{blue}{+12.6\%}) \\
\midrule
FullGrad+ & 60.3 \textcolor{gray}{±0.1} & 59.8 \textcolor{gray}{±0.2} & 60.9 \textcolor{gray}{±0.1} & 60.4 \textcolor{gray}{±0.2} \\
Int. FullGrad+ & 63.7 \textcolor{gray}{±0.1} (\textcolor{blue}{+5.6\%}) & 62.7 \textcolor{gray}{±0.2} (\textcolor{blue}{+4.8\%}) & 59.1 \textcolor{gray}{±0.1} (\textcolor{coral}{-3.1\%}) & 59.0 \textcolor{gray}{±0.2} (\textcolor{coral}{-2.2\%}) \\
\textbf{Libra FullGrad+} & \textbf{71.2} \textcolor{gray}{±0.1} (\textcolor{blue}{+18.1\%}) & \textbf{70.0} \textcolor{gray}{±0.3} (\textcolor{blue}{+17.1\%}) & \textbf{72.5} \textcolor{gray}{±0.1} (\textcolor{blue}{+19.0\%}) & \textbf{71.3} \textcolor{gray}{±0.2} (\textcolor{blue}{+18.1\%}) \\
\bottomrule
\end{tabular}
\caption{Comparison of gradient-based attribution methods and their compositions with LibraGrad and IG on the ViT-L model.}
\label{tbl:SRG_m_IG}
\end{table}
  \endgroup%

\clearpage{}
\subsection{Across Models}
\label{apn:across_models}
\label{apn:across_models_mif}
\TableRowsA[h]{%
  files/tables_v1/across_models/sel1/MIF_1p_accuracy/m_2.tex,%
  files/tables_v1/across_models/sel1/MIF_gt_accuracy/m_2.tex,%
}{}{}

\TableRowsA{%
  files/tables_v1/across_models/sel1/MIF_1p_AOPC/m_2.tex,%
  files/tables_v1/across_models/sel1/MIF_gt_AOPC/m_2.tex,%
}{}{}

\clearpage{}

\label{apn:across_models_lif}

\TableRowsA[h]{%
  files/tables_v1/across_models/sel1/LIF_1p_accuracy/m_2.tex,%
  files/tables_v1/across_models/sel1/LIF_gt_accuracy/m_2.tex,%
}{}{}

\TableRowsA{%
  files/tables_v1/across_models/sel1/LIF_1p_AOPC/m_2.tex,%
  files/tables_v1/across_models/sel1/LIF_gt_AOPC/m_2.tex,%
}{}{}

\clearpage{}

\label{apn:across_models_srg}

\TableRowsA[h]{%
  files/tables_v1/across_models/sel1/SRG_1p_accuracy/m_2.tex,%
  files/tables_v1/across_models/sel1/SRG_gt_accuracy/m_2.tex,%
}{}{}

\TableRowsA{%
  files/tables_v1/across_models/sel1/SRG_1p_AOPC/m_2.tex,%
  files/tables_v1/across_models/sel1/SRG_gt_AOPC/m_2.tex,%
}{}{}

\clearpage{}

\subsubsection{\SegmentationLong}
\label{apn:across_models_seg}

\begin{table*}[h]
  \centering
  \small
  {%
  \begingroup%
  \renewenvironment{table}[1][]%
    {\begin{center}}%
    {\end{center}}%
  \begin{table}[!t]
\centering
\begin{tabular}{lr@{}lr@{}lr@{}lr@{}lr@{}lr@{}lr@{}lr@{}l}
\toprule
Method & \multicolumn{2}{c}{ViT-L} & \multicolumn{2}{c}{EVA2-S} & \multicolumn{2}{c}{BEiT2-L} & \multicolumn{2}{c}{FlexiViT-L} & \multicolumn{2}{c}{SigLIP-L} & \multicolumn{2}{c}{CLIP-H} & \multicolumn{2}{c}{DeiT3-H} & \multicolumn{2}{c}{\textit{Avg.}} \\
\cmidrule(r){1-1}
\cmidrule(lr){2-15}
\cmidrule(l){16-17}
Random & 42.0& \,\textcolor{gray}{±0.4} & 37.7& \,\textcolor{gray}{±0.3} & 39.8& \,\textcolor{gray}{±0.4} & 39.8& \,\textcolor{gray}{±0.4} & 33.0& \,\textcolor{gray}{±0.3} & 37.8& \,\textcolor{gray}{±0.3} & 37.8& \,\textcolor{gray}{±0.3} & 38.3& \,\textcolor{gray}{±0.3} \\
RawAtt & 40.2& \,\textcolor{gray}{±0.4} & 59.0& \,\textcolor{gray}{±0.3} & 47.6& \,\textcolor{gray}{±0.3} & 49.8& \,\textcolor{gray}{±0.3} & -& & 41.6& \,\textcolor{gray}{±0.3} & 49.7& \,\textcolor{gray}{±0.3} & 48.0& \,\textcolor{gray}{±0.3} \\
Attention Rollout & 39.9& \,\textcolor{gray}{±0.3} & 45.3& \,\textcolor{gray}{±0.3} & 42.2& \,\textcolor{gray}{±0.3} & 42.2& \,\textcolor{gray}{±0.3} & -& & 51.7& \,\textcolor{gray}{±0.4} & 34.1& \,\textcolor{gray}{±0.3} & 42.6& \,\textcolor{gray}{±0.3} \\
AliLRP & 42.7& \,\textcolor{gray}{±0.4} & 58.7& \,\textcolor{gray}{±0.3} & 43.9& \,\textcolor{gray}{±0.3} & 49.6& \,\textcolor{gray}{±0.3} & 33.5& \,\textcolor{gray}{±0.3} & 38.1& \,\textcolor{gray}{±0.3} & 52.2& \,\textcolor{gray}{±0.3} & 45.5& \,\textcolor{gray}{±0.3} \\
AttnLRP & 47.2& \,\textcolor{gray}{±0.3} & 73.1& \,\textcolor{gray}{±0.2} & 66.0& \,\textcolor{gray}{±0.3} & 43.4& \,\textcolor{gray}{±0.4} & 36.0& \,\textcolor{gray}{±0.3} & 50.9& \,\textcolor{gray}{±0.3} & 36.0& \,\textcolor{gray}{±0.3} & 50.4& \,\textcolor{gray}{±0.3} \\
DecompX & 54.2& \,\textcolor{gray}{±0.3} & 60.0& \,\textcolor{gray}{±0.3} & 55.6& \,\textcolor{gray}{±0.3} & 59.2& \,\textcolor{gray}{±0.3} & 40.5& \,\textcolor{gray}{±0.3} & 55.0& \,\textcolor{gray}{±0.3} & 49.5& \,\textcolor{gray}{±0.3} & 53.4& \,\textcolor{gray}{±0.3} \\
Integrated Gradients & 46.6& \,\textcolor{gray}{±0.3} & 51.2& \,\textcolor{gray}{±0.3} & 46.7& \,\textcolor{gray}{±0.3} & 41.3& \,\textcolor{gray}{±0.4} & 41.6& \,\textcolor{gray}{±0.3} & 36.9& \,\textcolor{gray}{±0.3} & 38.9& \,\textcolor{gray}{±0.3} & 43.3& \,\textcolor{gray}{±0.3} \\
\midrule
\IxG{}  & 43.6& \,\textcolor{gray}{±0.4} & 42.5& \,\textcolor{gray}{±0.3} & 39.6& \,\textcolor{gray}{±0.4} & 41.4& \,\textcolor{gray}{±0.4} & 35.5& \,\textcolor{gray}{±0.3} & 36.8& \,\textcolor{gray}{±0.3} & 39.6& \,\textcolor{gray}{±0.3} & 39.9& \,\textcolor{gray}{±0.3} \\
\textbf{Libra \IxG{} } & 53.6& \,\textcolor{gray}{±0.3} & 72.1& \,\textcolor{gray}{±0.3} & 54.8& \,\textcolor{gray}{±0.3} & 60.4& \,\textcolor{gray}{±0.3} & 39.9& \,\textcolor{gray}{±0.3} & 54.2& \,\textcolor{gray}{±0.3} & 49.0& \,\textcolor{gray}{±0.3} & 54.8& \,\textcolor{gray}{±0.3} \\
\midrule
AttCAT & 44.9& \,\textcolor{gray}{±0.3} & 58.9& \,\textcolor{gray}{±0.3} & 52.2& \,\textcolor{gray}{±0.3} & 45.1& \,\textcolor{gray}{±0.3} & 37.6& \,\textcolor{gray}{±0.3} & 38.9& \,\textcolor{gray}{±0.3} & 41.7& \,\textcolor{gray}{±0.3} & 45.6& \,\textcolor{gray}{±0.3} \\
\textbf{Libra AttCAT} & 53.3& \,\textcolor{gray}{±0.3} & 75.1& \,\textcolor{gray}{±0.3} & 65.5& \,\textcolor{gray}{±0.3} & 74.4& \,\textcolor{gray}{±0.3} & 46.8& \,\textcolor{gray}{±0.3} & 61.7& \,\textcolor{gray}{±0.3} & 60.1& \,\textcolor{gray}{±0.3} & 62.4& \,\textcolor{gray}{±0.3} \\
\midrule
GenAtt & 50.9& \,\textcolor{gray}{±0.3} & 42.3& \,\textcolor{gray}{±0.3} & 47.9& \,\textcolor{gray}{±0.3} & 75.1& \,\textcolor{gray}{±0.2} & -& & 55.9& \,\textcolor{gray}{±0.3} & 66.2& \,\textcolor{gray}{±0.2} & 56.4& \,\textcolor{gray}{±0.3} \\
\textbf{Libra GenAtt} & 58.6& \,\textcolor{gray}{±0.3} & 44.3& \,\textcolor{gray}{±0.3} & 48.8& \,\textcolor{gray}{±0.3} & \underline{79.4}& \,\textcolor{gray}{±0.2} & -& & \textbf{76.2}& \,\textcolor{gray}{±0.2} & \textbf{76.5}& \,\textcolor{gray}{±0.2} & 64.0& \,\textcolor{gray}{±0.3} \\
\midrule
TokenTM & 50.0& \,\textcolor{gray}{±0.3} & 45.5& \,\textcolor{gray}{±0.3} & 56.0& \,\textcolor{gray}{±0.3} & 72.2& \,\textcolor{gray}{±0.2} & -& & 58.6& \,\textcolor{gray}{±0.3} & 61.7& \,\textcolor{gray}{±0.2} & 57.3& \,\textcolor{gray}{±0.3} \\
\textbf{Libra TokenTM} & 53.9& \,\textcolor{gray}{±0.3} & 46.7& \,\textcolor{gray}{±0.3} & 54.2& \,\textcolor{gray}{±0.3} & 76.2& \,\textcolor{gray}{±0.2} & -& & 71.5& \,\textcolor{gray}{±0.3} & 70.8& \,\textcolor{gray}{±0.2} & 62.2& \,\textcolor{gray}{±0.3} \\
\midrule
GradCAM+ & 52.1& \,\textcolor{gray}{±0.4} & 49.3& \,\textcolor{gray}{±0.4} & 53.5& \,\textcolor{gray}{±0.4} & 40.5& \,\textcolor{gray}{±0.4} & 44.3& \,\textcolor{gray}{±0.4} & 43.0& \,\textcolor{gray}{±0.4} & 60.3& \,\textcolor{gray}{±0.4} & 49.0& \,\textcolor{gray}{±0.4} \\
\textbf{Libra GradCAM+} & 60.2& \,\textcolor{gray}{±0.4} & \underline{79.8}& \,\textcolor{gray}{±0.3} & \underline{69.4}& \,\textcolor{gray}{±0.4} & 50.2& \,\textcolor{gray}{±0.4} & 41.7& \,\textcolor{gray}{±0.3} & 47.4& \,\textcolor{gray}{±0.4} & 46.7& \,\textcolor{gray}{±0.4} & 56.5& \,\textcolor{gray}{±0.4} \\
\midrule
HiResCAM & 38.5& \,\textcolor{gray}{±0.4} & 73.2& \,\textcolor{gray}{±0.3} & 60.8& \,\textcolor{gray}{±0.3} & 43.7& \,\textcolor{gray}{±0.3} & 36.3& \,\textcolor{gray}{±0.3} & 45.9& \,\textcolor{gray}{±0.3} & 41.3& \,\textcolor{gray}{±0.3} & 48.5& \,\textcolor{gray}{±0.3} \\
\textbf{Libra HiResCAM} & 48.0& \,\textcolor{gray}{±0.3} & 76.5& \,\textcolor{gray}{±0.3} & 69.0& \,\textcolor{gray}{±0.3} & \textbf{81.6}& \,\textcolor{gray}{±0.3} & \underline{47.5}& \,\textcolor{gray}{±0.3} & 56.8& \,\textcolor{gray}{±0.3} & \underline{76.3}& \,\textcolor{gray}{±0.3} & \underline{65.1}& \,\textcolor{gray}{±0.3} \\
\midrule
XGradCAM+ & 46.9& \,\textcolor{gray}{±0.4} & 55.2& \,\textcolor{gray}{±0.4} & 49.0& \,\textcolor{gray}{±0.4} & 38.5& \,\textcolor{gray}{±0.4} & 43.0& \,\textcolor{gray}{±0.3} & 47.7& \,\textcolor{gray}{±0.4} & 48.9& \,\textcolor{gray}{±0.4} & 47.0& \,\textcolor{gray}{±0.4} \\
\textbf{Libra XGradCAM+} & \underline{60.3}& \,\textcolor{gray}{±0.4} & \textbf{82.7}& \,\textcolor{gray}{±0.3} & \textbf{71.4}& \,\textcolor{gray}{±0.3} & 63.3& \,\textcolor{gray}{±0.4} & 44.3& \,\textcolor{gray}{±0.4} & \underline{73.3}& \,\textcolor{gray}{±0.3} & 59.4& \,\textcolor{gray}{±0.3} & 65.0& \,\textcolor{gray}{±0.3} \\
\midrule
FullGrad+ & 44.2& \,\textcolor{gray}{±0.3} & 51.5& \,\textcolor{gray}{±0.3} & 47.4& \,\textcolor{gray}{±0.3} & 44.1& \,\textcolor{gray}{±0.3} & 37.7& \,\textcolor{gray}{±0.3} & 38.5& \,\textcolor{gray}{±0.3} & 40.6& \,\textcolor{gray}{±0.3} & 43.4& \,\textcolor{gray}{±0.3} \\
\textbf{Libra FullGrad+} & \textbf{64.5}& \,\textcolor{gray}{±0.3} & 79.4& \,\textcolor{gray}{±0.3} & 67.9& \,\textcolor{gray}{±0.3} & 75.1& \,\textcolor{gray}{±0.3} & \textbf{51.7}& \,\textcolor{gray}{±0.3} & 71.5& \,\textcolor{gray}{±0.3} & 65.1& \,\textcolor{gray}{±0.3} & \textbf{67.9}& \,\textcolor{gray}{±0.3} \\
\bottomrule
\end{tabular}
\caption{Segmentation AP for different methods (and their \FairGradPrefix{} enhancements) across multiple models.}
\label{across_models_sel1_seg_m_2}
\end{table}
  \endgroup%

  }
\end{table*}
\clearpage{}

\subsection{Across Model Sizes}
\label{apn:across_model_sizes}

\label{apn:across_model_sizes_mif}

\TableRowsB[h]{%
  files/tables_v1/across_models/depth/MIF_1p_accuracy/m_2.tex,%
  files/tables_v1/across_models/depth/MIF_gt_accuracy/m_2.tex,%
}{}{}

\TableRowsB{%
  files/tables_v1/across_models/depth/MIF_1p_AOPC/m_2.tex,%
  files/tables_v1/across_models/depth/MIF_gt_AOPC/m_2.tex,%
}{}{}

\clearpage{}

\label{apn:across_model_sizes_lif}

\TableRowsB[h]{%
  files/tables_v1/across_models/depth/LIF_1p_accuracy/m_2.tex,%
  files/tables_v1/across_models/depth/LIF_gt_accuracy/m_2.tex,%
}{}{}

\TableRowsB{%
  files/tables_v1/across_models/depth/LIF_1p_AOPC/m_2.tex,%
  files/tables_v1/across_models/depth/LIF_gt_AOPC/m_2.tex,%
}{}{}

\clearpage{}

\label{apn:across_model_sizes_srg}

\TableRowsB[h]{%
  files/tables_v1/across_models/depth/SRG_1p_accuracy/m_2.tex,%
  files/tables_v1/across_models/depth/SRG_gt_accuracy/m_2.tex,%
}{}{}

\TableRowsB{%
  files/tables_v1/across_models/depth/SRG_1p_AOPC/m_2.tex,%
  files/tables_v1/across_models/depth/SRG_gt_AOPC/m_2.tex,%
}{}{}

\clearpage{}

\subsubsection{\SegmentationLong}
\label{apn:across_model_sizes_seg}

  \begingroup%
  \renewenvironment{table}[1][]%
    {\begin{table*}[#files/tables_v1/across_models/depth/seg/m_2.tex]\small\setlength{\tabcolsep}{4pt}}%
    {\end{table*}}%
  \begin{table}[h]
\centering
\begin{tabular}{lccccc}
\toprule
Method & ViT-Tiny & ViT-Small & ViT-Base & ViT-Large & \textit{Avg.} \\
\cmidrule(r){1-1}
\cmidrule(lr){2-5}
\cmidrule(l){6-6}
Random & 42.0 \textcolor{gray}{±0.4} & 41.9 \textcolor{gray}{±0.4} & 41.9 \textcolor{gray}{±0.4} & 42.0 \textcolor{gray}{±0.4} & 41.9 \textcolor{gray}{±0.4} \\
RawAtt & 60.2 \textcolor{gray}{±0.3} & 57.8 \textcolor{gray}{±0.3} & 46.9 \textcolor{gray}{±0.3} & 40.2 \textcolor{gray}{±0.4} & 51.3 \textcolor{gray}{±0.3} \\
Attention Rollout & 61.2 \textcolor{gray}{±0.4} & 47.1 \textcolor{gray}{±0.3} & 45.3 \textcolor{gray}{±0.3} & 39.9 \textcolor{gray}{±0.3} & 48.3 \textcolor{gray}{±0.3} \\
AliLRP & 54.5 \textcolor{gray}{±0.3} & 42.5 \textcolor{gray}{±0.4} & 43.8 \textcolor{gray}{±0.4} & 42.7 \textcolor{gray}{±0.4} & 45.9 \textcolor{gray}{±0.3} \\
AttnLRP & 59.7 \textcolor{gray}{±0.3} & 46.2 \textcolor{gray}{±0.3} & 42.0 \textcolor{gray}{±0.4} & 47.2 \textcolor{gray}{±0.3} & 48.8 \textcolor{gray}{±0.3} \\
DecompX & 60.0 \textcolor{gray}{±0.3} & 47.7 \textcolor{gray}{±0.3} & 44.3 \textcolor{gray}{±0.3} & 54.2 \textcolor{gray}{±0.3} & 51.6 \textcolor{gray}{±0.3} \\
Integrated Gradients & 52.4 \textcolor{gray}{±0.3} & 51.7 \textcolor{gray}{±0.3} & 47.5 \textcolor{gray}{±0.3} & 46.6 \textcolor{gray}{±0.3} & 49.6 \textcolor{gray}{±0.3} \\
\midrule
\IxG{}  & 50.6 \textcolor{gray}{±0.3} & 48.5 \textcolor{gray}{±0.3} & 44.8 \textcolor{gray}{±0.3} & 43.6 \textcolor{gray}{±0.4} & 46.9 \textcolor{gray}{±0.3} \\
\textbf{Libra \IxG{} } & 57.1 \textcolor{gray}{±0.3} (\textcolor{blue}{+12.8\%}) & 46.0 \textcolor{gray}{±0.3} (\textcolor{coral}{-5.1\%}) & 44.4 \textcolor{gray}{±0.3} (\textcolor{coral}{-0.9\%}) & 53.6 \textcolor{gray}{±0.3} (\textcolor{blue}{+22.9\%}) & 50.3 \textcolor{gray}{±0.3} (\textcolor{blue}{+7.3\%}) \\
\midrule
AttCAT & 54.7 \textcolor{gray}{±0.3} & 49.8 \textcolor{gray}{±0.3} & 44.5 \textcolor{gray}{±0.3} & 44.9 \textcolor{gray}{±0.3} & 48.5 \textcolor{gray}{±0.3} \\
\textbf{Libra AttCAT} & 61.1 \textcolor{gray}{±0.3} (\textcolor{blue}{+11.7\%}) & 56.0 \textcolor{gray}{±0.3} (\textcolor{blue}{+12.4\%}) & 61.5 \textcolor{gray}{±0.3} (\textcolor{blue}{+38.3\%}) & 53.3 \textcolor{gray}{±0.3} (\textcolor{blue}{+18.8\%}) & 58.0 \textcolor{gray}{±0.3} (\textcolor{blue}{+19.6\%}) \\
\midrule
GenAtt & 71.1 \textcolor{gray}{±0.3} & 65.9 \textcolor{gray}{±0.2} & 71.0 \textcolor{gray}{±0.2} & 50.9 \textcolor{gray}{±0.3} & 64.7 \textcolor{gray}{±0.3} \\
\textbf{Libra GenAtt} & \textbf{75.0} \textcolor{gray}{±0.3} (\textcolor{blue}{+5.5\%}) & \underline{71.0} \textcolor{gray}{±0.3} (\textcolor{blue}{+7.7\%}) & \textbf{77.5} \textcolor{gray}{±0.2} (\textcolor{blue}{+9.2\%}) & 58.6 \textcolor{gray}{±0.3} (\textcolor{blue}{+15.1\%}) & \textbf{70.5} \textcolor{gray}{±0.3} (\textcolor{blue}{+9.0\%}) \\
\midrule
TokenTM & 70.8 \textcolor{gray}{±0.3} & 68.2 \textcolor{gray}{±0.2} & 70.2 \textcolor{gray}{±0.2} & 50.0 \textcolor{gray}{±0.3} & 64.8 \textcolor{gray}{±0.3} \\
\textbf{Libra TokenTM} & \underline{73.7} \textcolor{gray}{±0.3} (\textcolor{blue}{+4.1\%}) & \textbf{71.4} \textcolor{gray}{±0.2} (\textcolor{blue}{+4.7\%}) & 73.9 \textcolor{gray}{±0.2} (\textcolor{blue}{+5.2\%}) & 53.9 \textcolor{gray}{±0.3} (\textcolor{blue}{+7.9\%}) & \underline{68.2} \textcolor{gray}{±0.3} (\textcolor{blue}{+5.3\%}) \\
\midrule
GradCAM+ & 48.4 \textcolor{gray}{±0.4} & 46.4 \textcolor{gray}{±0.4} & 50.2 \textcolor{gray}{±0.4} & 52.1 \textcolor{gray}{±0.4} & 49.3 \textcolor{gray}{±0.4} \\
\textbf{Libra GradCAM+} & 56.3 \textcolor{gray}{±0.4} (\textcolor{blue}{+16.4\%}) & 60.7 \textcolor{gray}{±0.4} (\textcolor{blue}{+30.8\%}) & 72.1 \textcolor{gray}{±0.3} (\textcolor{blue}{+43.6\%}) & 60.2 \textcolor{gray}{±0.4} (\textcolor{blue}{+15.5\%}) & 62.3 \textcolor{gray}{±0.4} (\textcolor{blue}{+26.5\%}) \\
\midrule
HiResCAM & 50.6 \textcolor{gray}{±0.4} & 48.4 \textcolor{gray}{±0.4} & 59.0 \textcolor{gray}{±0.3} & 38.5 \textcolor{gray}{±0.4} & 49.1 \textcolor{gray}{±0.4} \\
\textbf{Libra HiResCAM} & 63.8 \textcolor{gray}{±0.3} (\textcolor{blue}{+26.1\%}) & 69.4 \textcolor{gray}{±0.3} (\textcolor{blue}{+43.2\%}) & 72.6 \textcolor{gray}{±0.3} (\textcolor{blue}{+23.1\%}) & 48.0 \textcolor{gray}{±0.3} (\textcolor{blue}{+24.8\%}) & 63.4 \textcolor{gray}{±0.3} (\textcolor{blue}{+29.1\%}) \\
\midrule
XGradCAM+ & 48.8 \textcolor{gray}{±0.4} & 45.4 \textcolor{gray}{±0.4} & 41.0 \textcolor{gray}{±0.4} & 46.9 \textcolor{gray}{±0.4} & 45.5 \textcolor{gray}{±0.4} \\
\textbf{Libra XGradCAM+} & 61.4 \textcolor{gray}{±0.4} (\textcolor{blue}{+26.0\%}) & 62.3 \textcolor{gray}{±0.4} (\textcolor{blue}{+37.2\%}) & \underline{75.0} \textcolor{gray}{±0.3} (\textcolor{blue}{+82.8\%}) & \underline{60.3} \textcolor{gray}{±0.4} (\textcolor{blue}{+28.6\%}) & 64.7 \textcolor{gray}{±0.4} (\textcolor{blue}{+42.3\%}) \\
\midrule
FullGrad+ & 53.2 \textcolor{gray}{±0.3} & 50.0 \textcolor{gray}{±0.3} & 45.2 \textcolor{gray}{±0.3} & 44.2 \textcolor{gray}{±0.3} & 48.1 \textcolor{gray}{±0.3} \\
\textbf{Libra FullGrad+} & 65.0 \textcolor{gray}{±0.3} (\textcolor{blue}{+22.2\%}) & 59.6 \textcolor{gray}{±0.3} (\textcolor{blue}{+19.2\%}) & 65.5 \textcolor{gray}{±0.3} (\textcolor{blue}{+44.8\%}) & \textbf{64.5} \textcolor{gray}{±0.3} (\textcolor{blue}{+46.0\%}) & 63.6 \textcolor{gray}{±0.3} (\textcolor{blue}{+32.2\%}) \\
\bottomrule
\end{tabular}
\caption{How Segmentation AP varies with different model sizes.}
\label{across_models_depth_seg_m_2}
\end{table}
  \endgroup%

\clearpage{}

\subsection{Across Datasets}
\label{apn:across_datasets}

\label{apn:across_datasets_mif}

\TableRowsB[h]{%
  files/tables_v1/across_models/ds/MIF_1p_accuracy/m_2.tex,%
  files/tables_v1/across_models/ds/MIF_gt_accuracy/m_2.tex,%
}{}{}

\TableRowsB{%
  files/tables_v1/across_models/ds/MIF_1p_AOPC/m_2.tex,%
  files/tables_v1/across_models/ds/MIF_gt_AOPC/m_2.tex,%
}{}{}

\clearpage{}

\label{apn:across_datasets_lif}

\TableRowsB[h]{%
  files/tables_v1/across_models/ds/LIF_1p_accuracy/m_2.tex,%
  files/tables_v1/across_models/ds/LIF_gt_accuracy/m_2.tex,%
}{}{}

\TableRowsB{%
  files/tables_v1/across_models/ds/LIF_1p_AOPC/m_2.tex,%
  files/tables_v1/across_models/ds/LIF_gt_AOPC/m_2.tex,%
}{}{}

\clearpage{}

\label{apn:across_datasets_srg}

\TableRowsB[h]{%
  files/tables_v1/across_models/ds/SRG_1p_accuracy/m_2.tex,%
  files/tables_v1/across_models/ds/SRG_gt_accuracy/m_2.tex,%
}{}{}

\TableRowsB{%
  files/tables_v1/across_models/ds/SRG_1p_AOPC/m_2.tex,%
  files/tables_v1/across_models/ds/SRG_gt_AOPC/m_2.tex,%
}{}{}

\clearpage{}

\subsection{Results Per Model}
\label{apn:results_per_model}

\FloatBarrier
\clearpage
\subsubsection{MLP-Mixer-L}
\label{per_model:MLP-Mixer-L}
Since MLP-Mixer is an attention-free architecture, certain attribution methods couldn't be applied and were omitted.
\begin{table}[ht]
  \fontsize{9.5pt}{8.5pt}\selectfont

  \setlength{\tabcolsep}{2pt}
  \centering
  \begingroup%
  \renewenvironment{table}[1][]%
    {\begin{center}}%
    {\end{center}}%
  \begin{table}[h]
\centering
\begin{tabular}{lccccc}
\toprule
Method & \multicolumn{2}{c}{MIF Deletion (GT)} & \multicolumn{2}{c}{MIF Deletion (Predicted)} & Segmentation \\
  & Accuracy & AOPC & Accuracy & AOPC & AP \\
\cmidrule(r){1-1}
\cmidrule(lr){2-3}
\cmidrule(lr){4-5}
\cmidrule(l){6-6}
Random & 48.7 \textcolor{gray}{±0.1} & 20.3 \textcolor{gray}{±0.3} & 42.0 \textcolor{gray}{±0.1} & 25.8 \textcolor{gray}{±0.2} & 43.2 \textcolor{gray}{±0.4} \\
AliLRP & 64.6 \textcolor{gray}{±0.1} & 33.9 \textcolor{gray}{±0.3} & 60.2 \textcolor{gray}{±0.1} & 41.0 \textcolor{gray}{±0.2} & 58.6 \textcolor{gray}{±0.3} \\
DecompX & 66.0 \textcolor{gray}{±0.1} & 35.6 \textcolor{gray}{±0.3} & 61.8 \textcolor{gray}{±0.1} & 42.8 \textcolor{gray}{±0.2} & 59.6 \textcolor{gray}{±0.3} \\
Integrated Gradients & 62.2 \textcolor{gray}{±0.1} & 30.8 \textcolor{gray}{±0.2} & 53.0 \textcolor{gray}{±0.1} & 34.7 \textcolor{gray}{±0.2} & 54.3 \textcolor{gray}{±0.3} \\
\midrule
\IxG{}  & 59.0 \textcolor{gray}{±0.1} & 28.8 \textcolor{gray}{±0.2} & 54.5 \textcolor{gray}{±0.1} & 35.3 \textcolor{gray}{±0.2} & 52.3 \textcolor{gray}{±0.3} \\
\textbf{Libra \IxG{} } & \textbf{79.6} \textcolor{gray}{±0.1} (\textcolor{blue}{+34.9\%}) & \textbf{43.6} \textcolor{gray}{±0.3} (\textcolor{blue}{+51.5\%}) & \textbf{77.0} \textcolor{gray}{±0.1} (\textcolor{blue}{+41.3\%}) & \textbf{51.4} \textcolor{gray}{±0.2} (\textcolor{blue}{+45.5\%}) & \underline{68.1} \textcolor{gray}{±0.3} (\textcolor{blue}{+30.2\%}) \\
\midrule
GradCAM+ & 62.2 \textcolor{gray}{±0.1} & 31.1 \textcolor{gray}{±0.3} & 57.7 \textcolor{gray}{±0.1} & 37.9 \textcolor{gray}{±0.2} & 52.2 \textcolor{gray}{±0.4} \\
\textbf{Libra GradCAM+} & 66.3 \textcolor{gray}{±0.1} (\textcolor{blue}{+6.6\%}) & 34.4 \textcolor{gray}{±0.3} (\textcolor{blue}{+10.6\%}) & 61.9 \textcolor{gray}{±0.1} (\textcolor{blue}{+7.2\%}) & 41.3 \textcolor{gray}{±0.2} (\textcolor{blue}{+9.1\%}) & 57.8 \textcolor{gray}{±0.3} (\textcolor{blue}{+10.9\%}) \\
\midrule
HiResCAM & 54.2 \textcolor{gray}{±0.1} & 25.3 \textcolor{gray}{±0.3} & 48.2 \textcolor{gray}{±0.1} & 31.3 \textcolor{gray}{±0.2} & 47.4 \textcolor{gray}{±0.4} \\
\textbf{Libra HiResCAM} & 55.0 \textcolor{gray}{±0.1} (\textcolor{blue}{+1.4\%}) & 26.1 \textcolor{gray}{±0.3} (\textcolor{blue}{+3.4\%}) & 48.9 \textcolor{gray}{±0.1} (\textcolor{blue}{+1.4\%}) & 32.1 \textcolor{gray}{±0.2} (\textcolor{blue}{+2.8\%}) & 50.5 \textcolor{gray}{±0.3} (\textcolor{blue}{+6.5\%}) \\
\midrule
XGradCAM+ & 62.8 \textcolor{gray}{±0.1} & 31.7 \textcolor{gray}{±0.3} & 58.3 \textcolor{gray}{±0.1} & 38.4 \textcolor{gray}{±0.2} & 53.3 \textcolor{gray}{±0.4} \\
\textbf{Libra XGradCAM+} & 69.1 \textcolor{gray}{±0.1} (\textcolor{blue}{+10.2\%}) & 36.4 \textcolor{gray}{±0.3} (\textcolor{blue}{+15.0\%}) & 65.1 \textcolor{gray}{±0.1} (\textcolor{blue}{+11.5\%}) & 43.5 \textcolor{gray}{±0.2} (\textcolor{blue}{+13.3\%}) & 62.8 \textcolor{gray}{±0.3} (\textcolor{blue}{+17.7\%}) \\
\midrule
FullGrad+ & 64.0 \textcolor{gray}{±0.1} & 31.7 \textcolor{gray}{±0.3} & 60.2 \textcolor{gray}{±0.1} & 38.6 \textcolor{gray}{±0.3} & 53.3 \textcolor{gray}{±0.3} \\
\textbf{Libra FullGrad+} & \underline{76.0} \textcolor{gray}{±0.1} (\textcolor{blue}{+18.8\%}) & \underline{41.3} \textcolor{gray}{±0.3} (\textcolor{blue}{+30.1\%}) & \underline{73.1} \textcolor{gray}{±0.1} (\textcolor{blue}{+21.5\%}) & \underline{48.9} \textcolor{gray}{±0.2} (\textcolor{blue}{+26.4\%}) & \textbf{70.1} \textcolor{gray}{±0.3} (\textcolor{blue}{+31.4\%}) \\
\bottomrule
\end{tabular}
\caption{Comparison of attribution methods and their LibraGrad-enhanced versions on the MLP-Mixer-L model. We report faithfulness metrics using Most-Influential-First Deletion, MIF with ground-truth (GT) and predicted labels, including Accuracy and Area Over Perturbation Curve (AOPC) and Segmentation Average Precision (AP). The results demonstrate that composing existing methods with LibraGrad significantly enhances their performance across all metrics.}
\label{tbl:MIF_m_2}
\end{table}
  \endgroup%

\end{table}

\begin{table}[ht]
  \fontsize{11pt}{8.5pt}\selectfont

  \setlength{\tabcolsep}{4pt}
  \centering

  \begingroup%
  \renewenvironment{table}[1][]%
    {\begin{center}}%
    {\end{center}}%
  \begin{table}[h]
\centering
\begin{tabular}{lcccc}
\toprule
Method & \multicolumn{2}{c}{LIF Deletion (GT)} & \multicolumn{2}{c}{LIF Deletion (Predicted)} \\
  & Accuracy & AOPC & Accuracy & AOPC \\
\cmidrule(r){1-1}
\cmidrule(lr){2-3}
\cmidrule(l){4-5}
Random & 51.1 \textcolor{gray}{±0.1} & 79.3 \textcolor{gray}{±0.2} & 57.6 \textcolor{gray}{±0.1} & 73.6 \textcolor{gray}{±0.2} \\
AliLRP & 66.6 \textcolor{gray}{±0.1} & 89.5 \textcolor{gray}{±0.2} & 74.3 \textcolor{gray}{±0.1} & 84.6 \textcolor{gray}{±0.2} \\
DecompX & 66.3 \textcolor{gray}{±0.1} & 89.8 \textcolor{gray}{±0.2} & 74.1 \textcolor{gray}{±0.1} & 84.9 \textcolor{gray}{±0.2} \\
Integrated Gradients & 65.3 \textcolor{gray}{±0.1} & 88.7 \textcolor{gray}{±0.2} & 67.2 \textcolor{gray}{±0.1} & 80.2 \textcolor{gray}{±0.2} \\
\midrule
\IxG{}  & 61.8 \textcolor{gray}{±0.1} & 85.4 \textcolor{gray}{±0.3} & 69.0 \textcolor{gray}{±0.1} & 80.2 \textcolor{gray}{±0.2} \\
\textbf{Libra \IxG{} } & \textbf{74.2} \textcolor{gray}{±0.1} (\textcolor{blue}{+19.9\%}) & \textbf{94.5} \textcolor{gray}{±0.2} (\textcolor{blue}{+10.6\%}) & \textbf{81.6} \textcolor{gray}{±0.1} (\textcolor{blue}{+18.2\%}) & \textbf{90.0} \textcolor{gray}{±0.2} (\textcolor{blue}{+12.2\%}) \\
\midrule
GradCAM+ & 63.7 \textcolor{gray}{±0.1} & 87.2 \textcolor{gray}{±0.2} & 70.6 \textcolor{gray}{±0.1} & 81.8 \textcolor{gray}{±0.2} \\
\textbf{Libra GradCAM+} & 66.6 \textcolor{gray}{±0.1} (\textcolor{blue}{+4.6\%}) & 89.4 \textcolor{gray}{±0.2} (\textcolor{blue}{+2.6\%}) & 73.7 \textcolor{gray}{±0.1} (\textcolor{blue}{+4.5\%}) & 84.3 \textcolor{gray}{±0.2} (\textcolor{blue}{+3.0\%}) \\
\midrule
HiResCAM & 56.6 \textcolor{gray}{±0.1} & 82.9 \textcolor{gray}{±0.2} & 63.9 \textcolor{gray}{±0.1} & 77.6 \textcolor{gray}{±0.2} \\
\textbf{Libra HiResCAM} & 57.4 \textcolor{gray}{±0.1} (\textcolor{blue}{+1.4\%}) & 83.0 \textcolor{gray}{±0.2} (\textcolor{blue}{+0.2\%}) & 64.4 \textcolor{gray}{±0.1} (\textcolor{blue}{+0.8\%}) & 77.5 \textcolor{gray}{±0.2} (\textcolor{coral}{-0.1\%}) \\
\midrule
XGradCAM+ & 64.3 \textcolor{gray}{±0.1} & 87.3 \textcolor{gray}{±0.2} & 71.3 \textcolor{gray}{±0.1} & 82.1 \textcolor{gray}{±0.2} \\
\textbf{Libra XGradCAM+} & 67.8 \textcolor{gray}{±0.1} (\textcolor{blue}{+5.5\%}) & 90.0 \textcolor{gray}{±0.2} (\textcolor{blue}{+3.0\%}) & 74.8 \textcolor{gray}{±0.1} (\textcolor{blue}{+5.0\%}) & 85.0 \textcolor{gray}{±0.2} (\textcolor{blue}{+3.5\%}) \\
\midrule
FullGrad+ & 66.4 \textcolor{gray}{±0.1} & 88.7 \textcolor{gray}{±0.3} & 73.4 \textcolor{gray}{±0.1} & 83.8 \textcolor{gray}{±0.2} \\
\textbf{Libra FullGrad+} & \underline{72.6} \textcolor{gray}{±0.1} (\textcolor{blue}{+9.4\%}) & \underline{91.1} \textcolor{gray}{±0.2} (\textcolor{blue}{+2.6\%}) & \underline{80.1} \textcolor{gray}{±0.1} (\textcolor{blue}{+9.2\%}) & \underline{86.3} \textcolor{gray}{±0.2} (\textcolor{blue}{+3.0\%}) \\
\bottomrule
\end{tabular}
\caption{Comparison of attribution methods and their LibraGrad-enhanced versions on the MLP-Mixer-L model.}
\label{tbl:LIF_m_2}
\end{table}
  \endgroup%

  \begingroup%
  \renewenvironment{table}[1][]%
    {\begin{center}}%
    {\end{center}}%
  \begin{table}[h]
\centering
\begin{tabular}{lcccc}
\toprule
Method & \multicolumn{2}{c}{SRG (GT)} & \multicolumn{2}{c}{SRG (Predicted)} \\
  & Accuracy & AOPC & Accuracy & AOPC \\
\cmidrule(r){1-1}
\cmidrule(lr){2-3}
\cmidrule(l){4-5}
Random & 49.9 \textcolor{gray}{±0.1} & 49.8 \textcolor{gray}{±0.2} & 49.8 \textcolor{gray}{±0.1} & 49.7 \textcolor{gray}{±0.2} \\
AliLRP & 65.6 \textcolor{gray}{±0.1} & 61.7 \textcolor{gray}{±0.3} & 67.3 \textcolor{gray}{±0.1} & 62.8 \textcolor{gray}{±0.2} \\
DecompX & 66.2 \textcolor{gray}{±0.1} & 62.7 \textcolor{gray}{±0.3} & 68.0 \textcolor{gray}{±0.1} & 63.8 \textcolor{gray}{±0.2} \\
Integrated Gradients & 63.8 \textcolor{gray}{±0.1} & 59.7 \textcolor{gray}{±0.2} & 60.1 \textcolor{gray}{±0.1} & 57.4 \textcolor{gray}{±0.2} \\
\midrule
\IxG{}  & 60.4 \textcolor{gray}{±0.1} & 57.1 \textcolor{gray}{±0.2} & 61.8 \textcolor{gray}{±0.1} & 57.8 \textcolor{gray}{±0.2} \\
\textbf{Libra \IxG{} } & \textbf{76.9} \textcolor{gray}{±0.1} (\textcolor{blue}{+27.2\%}) & \textbf{69.0} \textcolor{gray}{±0.2} (\textcolor{blue}{+20.9\%}) & \textbf{79.3} \textcolor{gray}{±0.1} (\textcolor{blue}{+28.4\%}) & \textbf{70.7} \textcolor{gray}{±0.2} (\textcolor{blue}{+22.4\%}) \\
\midrule
GradCAM+ & 62.9 \textcolor{gray}{±0.1} & 59.1 \textcolor{gray}{±0.2} & 64.1 \textcolor{gray}{±0.1} & 59.8 \textcolor{gray}{±0.2} \\
\textbf{Libra GradCAM+} & 66.5 \textcolor{gray}{±0.1} (\textcolor{blue}{+5.6\%}) & 61.9 \textcolor{gray}{±0.3} (\textcolor{blue}{+4.7\%}) & 67.8 \textcolor{gray}{±0.1} (\textcolor{blue}{+5.7\%}) & 62.8 \textcolor{gray}{±0.2} (\textcolor{blue}{+4.9\%}) \\
\midrule
HiResCAM & 55.4 \textcolor{gray}{±0.1} & 54.1 \textcolor{gray}{±0.3} & 56.1 \textcolor{gray}{±0.1} & 54.4 \textcolor{gray}{±0.2} \\
\textbf{Libra HiResCAM} & 56.2 \textcolor{gray}{±0.1} (\textcolor{blue}{+1.4\%}) & 54.6 \textcolor{gray}{±0.3} (\textcolor{blue}{+0.9\%}) & 56.7 \textcolor{gray}{±0.1} (\textcolor{blue}{+1.0\%}) & 54.8 \textcolor{gray}{±0.2} (\textcolor{blue}{+0.7\%}) \\
\midrule
XGradCAM+ & 63.5 \textcolor{gray}{±0.1} & 59.5 \textcolor{gray}{±0.2} & 64.8 \textcolor{gray}{±0.1} & 60.2 \textcolor{gray}{±0.2} \\
\textbf{Libra XGradCAM+} & 68.5 \textcolor{gray}{±0.1} (\textcolor{blue}{+7.8\%}) & 63.2 \textcolor{gray}{±0.3} (\textcolor{blue}{+6.2\%}) & 69.9 \textcolor{gray}{±0.1} (\textcolor{blue}{+7.9\%}) & 64.2 \textcolor{gray}{±0.2} (\textcolor{blue}{+6.6\%}) \\
\midrule
FullGrad+ & 65.2 \textcolor{gray}{±0.1} & 60.2 \textcolor{gray}{±0.3} & 66.8 \textcolor{gray}{±0.1} & 61.2 \textcolor{gray}{±0.2} \\
\textbf{Libra FullGrad+} & \underline{74.3} \textcolor{gray}{±0.1} (\textcolor{blue}{+14.0\%}) & \underline{66.2} \textcolor{gray}{±0.3} (\textcolor{blue}{+9.8\%}) & \underline{76.6} \textcolor{gray}{±0.1} (\textcolor{blue}{+14.7\%}) & \underline{67.6} \textcolor{gray}{±0.2} (\textcolor{blue}{+10.4\%}) \\
\bottomrule
\end{tabular}
\caption{Comparison of attribution methods and their LibraGrad-enhanced versions on the MLP-Mixer-L model.}
\label{tbl:SRG_m_2}
\end{table}
  \endgroup%

\end{table}

\FloatBarrier
\clearpage
\subsubsection{ViT-T}
\label{per_model:ViT-T}

\begin{table}[ht]
  \fontsize{9.5pt}{8.5pt}\selectfont

  \setlength{\tabcolsep}{2pt}
  \centering
  \begingroup%
  \renewenvironment{table}[1][]%
    {\begin{center}}%
    {\end{center}}%
  \begin{table}[h]
\centering
\begin{tabular}{lccccc}
\toprule
Method & \multicolumn{2}{c}{MIF Deletion (GT)} & \multicolumn{2}{c}{MIF Deletion (Predicted)} & Segmentation \\
  & Accuracy & AOPC & Accuracy & AOPC & AP \\
\cmidrule(r){1-1}
\cmidrule(lr){2-3}
\cmidrule(lr){4-5}
\cmidrule(l){6-6}
Random & 50.1 \textcolor{gray}{±0.1} & 17.0 \textcolor{gray}{±0.2} & 40.5 \textcolor{gray}{±0.1} & 20.7 \textcolor{gray}{±0.2} & 42.0 \textcolor{gray}{±0.4} \\
RawAtt & 74.0 \textcolor{gray}{±0.1} & 38.6 \textcolor{gray}{±0.3} & 69.5 \textcolor{gray}{±0.1} & 44.8 \textcolor{gray}{±0.3} & 60.2 \textcolor{gray}{±0.3} \\
Attention Rollout & 68.7 \textcolor{gray}{±0.1} & 33.6 \textcolor{gray}{±0.3} & 64.1 \textcolor{gray}{±0.1} & 39.8 \textcolor{gray}{±0.3} & 61.2 \textcolor{gray}{±0.4} \\
AliLRP & 68.9 \textcolor{gray}{±0.1} & 33.4 \textcolor{gray}{±0.3} & 64.4 \textcolor{gray}{±0.1} & 39.3 \textcolor{gray}{±0.2} & 54.5 \textcolor{gray}{±0.3} \\
AttnLRP & 73.4 \textcolor{gray}{±0.1} & 37.8 \textcolor{gray}{±0.3} & 69.7 \textcolor{gray}{±0.1} & 44.3 \textcolor{gray}{±0.3} & 59.7 \textcolor{gray}{±0.3} \\
DecompX & 74.0 \textcolor{gray}{±0.1} & 38.2 \textcolor{gray}{±0.3} & 70.4 \textcolor{gray}{±0.1} & 44.8 \textcolor{gray}{±0.3} & 60.0 \textcolor{gray}{±0.3} \\
Integrated Gradients & 69.7 \textcolor{gray}{±0.1} & 32.8 \textcolor{gray}{±0.3} & 57.1 \textcolor{gray}{±0.1} & 33.3 \textcolor{gray}{±0.2} & 52.4 \textcolor{gray}{±0.3} \\
\midrule
\IxG{}  & 61.1 \textcolor{gray}{±0.1} & 26.3 \textcolor{gray}{±0.3} & 55.6 \textcolor{gray}{±0.1} & 31.8 \textcolor{gray}{±0.2} & 50.6 \textcolor{gray}{±0.3} \\
\textbf{Libra \IxG{} } & 74.5 \textcolor{gray}{±0.1} (\textcolor{blue}{+22.0\%}) & 37.5 \textcolor{gray}{±0.3} (\textcolor{blue}{+42.6\%}) & 70.8 \textcolor{gray}{±0.1} (\textcolor{blue}{+27.2\%}) & 44.0 \textcolor{gray}{±0.3} (\textcolor{blue}{+38.3\%}) & 57.1 \textcolor{gray}{±0.3} (\textcolor{blue}{+12.8\%}) \\
\midrule
AttCAT & 72.6 \textcolor{gray}{±0.1} & 35.6 \textcolor{gray}{±0.3} & 69.3 \textcolor{gray}{±0.1} & 42.0 \textcolor{gray}{±0.3} & 54.7 \textcolor{gray}{±0.3} \\
\textbf{Libra AttCAT} & \underline{83.6} \textcolor{gray}{±0.1} (\textcolor{blue}{+15.2\%}) & \underline{45.0} \textcolor{gray}{±0.3} (\textcolor{blue}{+26.6\%}) & \underline{81.0} \textcolor{gray}{±0.1} (\textcolor{blue}{+16.7\%}) & \underline{52.1} \textcolor{gray}{±0.2} (\textcolor{blue}{+24.1\%}) & 61.1 \textcolor{gray}{±0.3} (\textcolor{blue}{+11.7\%}) \\
\midrule
GenAtt & 80.4 \textcolor{gray}{±0.1} & 42.7 \textcolor{gray}{±0.3} & 77.1 \textcolor{gray}{±0.1} & 49.4 \textcolor{gray}{±0.3} & 71.1 \textcolor{gray}{±0.3} \\
\textbf{Libra GenAtt} & 81.6 \textcolor{gray}{±0.1} (\textcolor{blue}{+1.4\%}) & 43.6 \textcolor{gray}{±0.3} (\textcolor{blue}{+2.3\%}) & 78.4 \textcolor{gray}{±0.1} (\textcolor{blue}{+1.7\%}) & 50.5 \textcolor{gray}{±0.2} (\textcolor{blue}{+2.2\%}) & \textbf{75.0} \textcolor{gray}{±0.3} (\textcolor{blue}{+5.5\%}) \\
\midrule
TokenTM & 78.8 \textcolor{gray}{±0.1} & 41.8 \textcolor{gray}{±0.3} & 75.0 \textcolor{gray}{±0.1} & 48.3 \textcolor{gray}{±0.3} & 70.8 \textcolor{gray}{±0.3} \\
\textbf{Libra TokenTM} & 79.9 \textcolor{gray}{±0.1} (\textcolor{blue}{+1.4\%}) & 42.6 \textcolor{gray}{±0.3} (\textcolor{blue}{+2.0\%}) & 76.2 \textcolor{gray}{±0.1} (\textcolor{blue}{+1.6\%}) & 49.2 \textcolor{gray}{±0.3} (\textcolor{blue}{+1.9\%}) & \underline{73.7} \textcolor{gray}{±0.3} (\textcolor{blue}{+4.1\%}) \\
\midrule
GradCAM+ & 70.5 \textcolor{gray}{±0.1} & 34.1 \textcolor{gray}{±0.3} & 66.2 \textcolor{gray}{±0.1} & 40.1 \textcolor{gray}{±0.2} & 48.4 \textcolor{gray}{±0.4} \\
\textbf{Libra GradCAM+} & 76.8 \textcolor{gray}{±0.1} (\textcolor{blue}{+8.9\%}) & 39.9 \textcolor{gray}{±0.3} (\textcolor{blue}{+16.8\%}) & 72.9 \textcolor{gray}{±0.1} (\textcolor{blue}{+10.1\%}) & 46.4 \textcolor{gray}{±0.2} (\textcolor{blue}{+15.7\%}) & 56.3 \textcolor{gray}{±0.4} (\textcolor{blue}{+16.4\%}) \\
\midrule
HiResCAM & 48.0 \textcolor{gray}{±0.1} & 15.8 \textcolor{gray}{±0.3} & 39.0 \textcolor{gray}{±0.1} & 19.5 \textcolor{gray}{±0.3} & 50.6 \textcolor{gray}{±0.4} \\
\textbf{Libra HiResCAM} & 74.1 \textcolor{gray}{±0.1} (\textcolor{blue}{+54.3\%}) & 37.8 \textcolor{gray}{±0.3} (\textcolor{blue}{+138.5\%}) & 69.9 \textcolor{gray}{±0.1} (\textcolor{blue}{+79.1\%}) & 44.0 \textcolor{gray}{±0.2} (\textcolor{blue}{+125.6\%}) & 63.8 \textcolor{gray}{±0.3} (\textcolor{blue}{+26.1\%}) \\
\midrule
XGradCAM+ & 71.7 \textcolor{gray}{±0.1} & 35.1 \textcolor{gray}{±0.3} & 67.5 \textcolor{gray}{±0.1} & 41.2 \textcolor{gray}{±0.2} & 48.8 \textcolor{gray}{±0.4} \\
\textbf{Libra XGradCAM+} & 80.6 \textcolor{gray}{±0.1} (\textcolor{blue}{+12.4\%}) & 42.7 \textcolor{gray}{±0.3} (\textcolor{blue}{+21.7\%}) & 77.0 \textcolor{gray}{±0.1} (\textcolor{blue}{+14.1\%}) & 49.5 \textcolor{gray}{±0.2} (\textcolor{blue}{+20.1\%}) & 61.4 \textcolor{gray}{±0.4} (\textcolor{blue}{+26.0\%}) \\
\midrule
FullGrad+ & 69.8 \textcolor{gray}{±0.1} & 33.1 \textcolor{gray}{±0.3} & 65.9 \textcolor{gray}{±0.1} & 39.2 \textcolor{gray}{±0.3} & 53.2 \textcolor{gray}{±0.3} \\
\textbf{Libra FullGrad+} & \textbf{84.2} \textcolor{gray}{±0.1} (\textcolor{blue}{+20.8\%}) & \textbf{45.6} \textcolor{gray}{±0.3} (\textcolor{blue}{+37.8\%}) & \textbf{81.7} \textcolor{gray}{±0.1} (\textcolor{blue}{+24.0\%}) & \textbf{52.7} \textcolor{gray}{±0.2} (\textcolor{blue}{+34.7\%}) & 65.0 \textcolor{gray}{±0.3} (\textcolor{blue}{+22.2\%}) \\
\bottomrule
\end{tabular}
\caption{Comparison of attribution methods and their LibraGrad-enhanced versions on the ViT-T model. We report faithfulness metrics using Most-Influential-First Deletion, MIF with ground-truth (GT) and predicted labels, including Accuracy and Area Over Perturbation Curve (AOPC) and Segmentation Average Precision (AP). The results demonstrate that composing existing methods with LibraGrad significantly enhances their performance across all metrics.}
\label{tbl:MIF_m_2}
\end{table}
  \endgroup%

\end{table}

\begin{table}[ht]
  \fontsize{11pt}{8.5pt}\selectfont

  \setlength{\tabcolsep}{4pt}
  \centering

  \begingroup%
  \renewenvironment{table}[1][]%
    {\begin{center}}%
    {\end{center}}%
  \begin{table}[h]
\centering
\begin{tabular}{lcccc}
\toprule
Method & \multicolumn{2}{c}{LIF Deletion (GT)} & \multicolumn{2}{c}{LIF Deletion (Predicted)} \\
  & Accuracy & AOPC & Accuracy & AOPC \\
\cmidrule(r){1-1}
\cmidrule(lr){2-3}
\cmidrule(l){4-5}
Random & 49.2 \textcolor{gray}{±0.1} & 82.8 \textcolor{gray}{±0.2} & 58.6 \textcolor{gray}{±0.1} & 79.0 \textcolor{gray}{±0.2} \\
RawAtt & 55.2 \textcolor{gray}{±0.1} & 88.1 \textcolor{gray}{±0.2} & 67.3 \textcolor{gray}{±0.1} & 85.6 \textcolor{gray}{±0.2} \\
Attention Rollout & 54.4 \textcolor{gray}{±0.1} & 87.2 \textcolor{gray}{±0.2} & 65.4 \textcolor{gray}{±0.1} & 84.3 \textcolor{gray}{±0.2} \\
AliLRP & 63.1 \textcolor{gray}{±0.1} & 95.0 \textcolor{gray}{±0.3} & 73.0 \textcolor{gray}{±0.1} & 92.6 \textcolor{gray}{±0.2} \\
AttnLRP & 63.2 \textcolor{gray}{±0.1} & 95.6 \textcolor{gray}{±0.2} & 74.3 \textcolor{gray}{±0.1} & 93.2 \textcolor{gray}{±0.2} \\
DecompX & 63.3 \textcolor{gray}{±0.1} & 95.4 \textcolor{gray}{±0.2} & 74.8 \textcolor{gray}{±0.1} & 93.1 \textcolor{gray}{±0.2} \\
Integrated Gradients & 64.0 \textcolor{gray}{±0.1} & 96.3 \textcolor{gray}{±0.2} & 66.9 \textcolor{gray}{±0.1} & 88.8 \textcolor{gray}{±0.2} \\
\midrule
\IxG{}  & 58.6 \textcolor{gray}{±0.1} & 90.8 \textcolor{gray}{±0.2} & 69.0 \textcolor{gray}{±0.1} & 88.3 \textcolor{gray}{±0.2} \\
\textbf{Libra \IxG{} } & 65.2 \textcolor{gray}{±0.1} (\textcolor{blue}{+11.3\%}) & 96.4 \textcolor{gray}{±0.2} (\textcolor{blue}{+6.2\%}) & 74.8 \textcolor{gray}{±0.1} (\textcolor{blue}{+8.4\%}) & 93.9 \textcolor{gray}{±0.2} (\textcolor{blue}{+6.3\%}) \\
\midrule
AttCAT & 66.5 \textcolor{gray}{±0.1} & 98.0 \textcolor{gray}{±0.2} & 74.8 \textcolor{gray}{±0.1} & 95.3 \textcolor{gray}{±0.2} \\
\textbf{Libra AttCAT} & \underline{69.5} \textcolor{gray}{±0.1} (\textcolor{blue}{+4.5\%}) & \underline{100.8} \textcolor{gray}{±0.2} (\textcolor{blue}{+2.9\%}) & \underline{77.9} \textcolor{gray}{±0.1} (\textcolor{blue}{+4.2\%}) & \underline{98.1} \textcolor{gray}{±0.2} (\textcolor{blue}{+2.9\%}) \\
\midrule
GenAtt & 63.6 \textcolor{gray}{±0.1} & 95.4 \textcolor{gray}{±0.2} & 76.2 \textcolor{gray}{±0.1} & 93.5 \textcolor{gray}{±0.2} \\
\textbf{Libra GenAtt} & 62.3 \textcolor{gray}{±0.1} (\textcolor{coral}{-2.0\%}) & 94.3 \textcolor{gray}{±0.2} (\textcolor{coral}{-1.1\%}) & 74.6 \textcolor{gray}{±0.1} (\textcolor{coral}{-2.1\%}) & 92.2 \textcolor{gray}{±0.2} (\textcolor{coral}{-1.4\%}) \\
\midrule
TokenTM & 61.2 \textcolor{gray}{±0.1} & 93.4 \textcolor{gray}{±0.2} & 74.2 \textcolor{gray}{±0.1} & 91.3 \textcolor{gray}{±0.2} \\
\textbf{Libra TokenTM} & 60.8 \textcolor{gray}{±0.1} (\textcolor{coral}{-0.6\%}) & 92.7 \textcolor{gray}{±0.2} (\textcolor{coral}{-0.7\%}) & 73.7 \textcolor{gray}{±0.1} (\textcolor{coral}{-0.6\%}) & 90.5 \textcolor{gray}{±0.2} (\textcolor{coral}{-0.9\%}) \\
\midrule
GradCAM+ & 57.9 \textcolor{gray}{±0.1} & 88.6 \textcolor{gray}{±0.2} & 65.1 \textcolor{gray}{±0.1} & 84.3 \textcolor{gray}{±0.2} \\
\textbf{Libra GradCAM+} & 61.9 \textcolor{gray}{±0.1} (\textcolor{blue}{+7.0\%}) & 93.3 \textcolor{gray}{±0.2} (\textcolor{blue}{+5.2\%}) & 70.2 \textcolor{gray}{±0.1} (\textcolor{blue}{+7.8\%}) & 89.6 \textcolor{gray}{±0.2} (\textcolor{blue}{+6.2\%}) \\
\midrule
HiResCAM & 42.4 \textcolor{gray}{±0.1} & 76.6 \textcolor{gray}{±0.3} & 48.3 \textcolor{gray}{±0.1} & 71.3 \textcolor{gray}{±0.2} \\
\textbf{Libra HiResCAM} & 60.0 \textcolor{gray}{±0.1} (\textcolor{blue}{+41.5\%}) & 90.1 \textcolor{gray}{±0.2} (\textcolor{blue}{+17.7\%}) & 68.0 \textcolor{gray}{±0.1} (\textcolor{blue}{+40.8\%}) & 86.2 \textcolor{gray}{±0.2} (\textcolor{blue}{+20.8\%}) \\
\midrule
XGradCAM+ & 59.5 \textcolor{gray}{±0.1} & 90.5 \textcolor{gray}{±0.3} & 66.7 \textcolor{gray}{±0.1} & 86.4 \textcolor{gray}{±0.2} \\
\textbf{Libra XGradCAM+} & 64.4 \textcolor{gray}{±0.1} (\textcolor{blue}{+8.3\%}) & 95.5 \textcolor{gray}{±0.2} (\textcolor{blue}{+5.6\%}) & 72.8 \textcolor{gray}{±0.1} (\textcolor{blue}{+9.0\%}) & 91.9 \textcolor{gray}{±0.2} (\textcolor{blue}{+6.3\%}) \\
\midrule
FullGrad+ & 64.5 \textcolor{gray}{±0.1} & 96.3 \textcolor{gray}{±0.2} & 73.4 \textcolor{gray}{±0.1} & 93.5 \textcolor{gray}{±0.2} \\
\textbf{Libra FullGrad+} & \textbf{70.2} \textcolor{gray}{±0.1} (\textcolor{blue}{+8.8\%}) & \textbf{101.8} \textcolor{gray}{±0.2} (\textcolor{blue}{+5.7\%}) & \textbf{78.8} \textcolor{gray}{±0.1} (\textcolor{blue}{+7.3\%}) & \textbf{99.2} \textcolor{gray}{±0.2} (\textcolor{blue}{+6.1\%}) \\
\bottomrule
\end{tabular}
\caption{Comparison of attribution methods and their LibraGrad-enhanced versions on the ViT-T model.}
\label{tbl:LIF_m_2}
\end{table}
  \endgroup%

  \begingroup%
  \renewenvironment{table}[1][]%
    {\begin{center}}%
    {\end{center}}%
  \begin{table}[h]
\centering
\begin{tabular}{lcccc}
\toprule
Method & \multicolumn{2}{c}{SRG (GT)} & \multicolumn{2}{c}{SRG (Predicted)} \\
  & Accuracy & AOPC & Accuracy & AOPC \\
\cmidrule(r){1-1}
\cmidrule(lr){2-3}
\cmidrule(l){4-5}
Random & 49.6 \textcolor{gray}{±0.1} & 49.9 \textcolor{gray}{±0.2} & 49.6 \textcolor{gray}{±0.1} & 49.9 \textcolor{gray}{±0.2} \\
RawAtt & 64.6 \textcolor{gray}{±0.1} & 63.3 \textcolor{gray}{±0.3} & 68.4 \textcolor{gray}{±0.1} & 65.2 \textcolor{gray}{±0.2} \\
Attention Rollout & 61.6 \textcolor{gray}{±0.1} & 60.4 \textcolor{gray}{±0.3} & 64.8 \textcolor{gray}{±0.1} & 62.1 \textcolor{gray}{±0.2} \\
AliLRP & 66.0 \textcolor{gray}{±0.1} & 64.2 \textcolor{gray}{±0.3} & 68.7 \textcolor{gray}{±0.1} & 66.0 \textcolor{gray}{±0.2} \\
AttnLRP & 68.3 \textcolor{gray}{±0.1} & 66.7 \textcolor{gray}{±0.3} & 72.0 \textcolor{gray}{±0.1} & 68.8 \textcolor{gray}{±0.2} \\
DecompX & 68.6 \textcolor{gray}{±0.1} & 66.8 \textcolor{gray}{±0.3} & 72.6 \textcolor{gray}{±0.1} & 69.0 \textcolor{gray}{±0.2} \\
Integrated Gradients & 66.8 \textcolor{gray}{±0.1} & 64.6 \textcolor{gray}{±0.3} & 62.0 \textcolor{gray}{±0.1} & 61.1 \textcolor{gray}{±0.2} \\
\midrule
\IxG{}  & 59.8 \textcolor{gray}{±0.1} & 58.5 \textcolor{gray}{±0.2} & 62.3 \textcolor{gray}{±0.1} & 60.0 \textcolor{gray}{±0.2} \\
\textbf{Libra \IxG{} } & 69.9 \textcolor{gray}{±0.1} (\textcolor{blue}{+16.8\%}) & 67.0 \textcolor{gray}{±0.3} (\textcolor{blue}{+14.4\%}) & 72.8 \textcolor{gray}{±0.1} (\textcolor{blue}{+16.8\%}) & 68.9 \textcolor{gray}{±0.2} (\textcolor{blue}{+14.8\%}) \\
\midrule
AttCAT & 69.5 \textcolor{gray}{±0.1} & 66.8 \textcolor{gray}{±0.3} & 72.0 \textcolor{gray}{±0.1} & 68.6 \textcolor{gray}{±0.2} \\
\textbf{Libra AttCAT} & \underline{76.5} \textcolor{gray}{±0.1} (\textcolor{blue}{+10.1\%}) & \underline{72.9} \textcolor{gray}{±0.2} (\textcolor{blue}{+9.2\%}) & \underline{79.4} \textcolor{gray}{±0.1} (\textcolor{blue}{+10.2\%}) & \underline{75.1} \textcolor{gray}{±0.2} (\textcolor{blue}{+9.4\%}) \\
\midrule
GenAtt & 72.0 \textcolor{gray}{±0.1} & 69.1 \textcolor{gray}{±0.2} & 76.6 \textcolor{gray}{±0.1} & 71.5 \textcolor{gray}{±0.2} \\
\textbf{Libra GenAtt} & 71.9 \textcolor{gray}{±0.1} (\textcolor{coral}{-0.1\%}) & 69.0 \textcolor{gray}{±0.2} (\textcolor{coral}{-0.1\%}) & 76.5 \textcolor{gray}{±0.1} (\textcolor{coral}{-0.2\%}) & 71.4 \textcolor{gray}{±0.2} (\textcolor{coral}{-0.1\%}) \\
\midrule
TokenTM & 70.0 \textcolor{gray}{±0.1} & 67.6 \textcolor{gray}{±0.2} & 74.6 \textcolor{gray}{±0.1} & 69.8 \textcolor{gray}{±0.2} \\
\textbf{Libra TokenTM} & 70.3 \textcolor{gray}{±0.1} (\textcolor{blue}{+0.5\%}) & 67.7 \textcolor{gray}{±0.2} (\textcolor{blue}{+0.1\%}) & 75.0 \textcolor{gray}{±0.1} (\textcolor{blue}{+0.5\%}) & 69.9 \textcolor{gray}{±0.2} (\textcolor{blue}{+0.1\%}) \\
\midrule
GradCAM+ & 64.2 \textcolor{gray}{±0.1} & 61.4 \textcolor{gray}{±0.3} & 65.7 \textcolor{gray}{±0.1} & 62.2 \textcolor{gray}{±0.2} \\
\textbf{Libra GradCAM+} & 69.3 \textcolor{gray}{±0.1} (\textcolor{blue}{+8.0\%}) & 66.6 \textcolor{gray}{±0.3} (\textcolor{blue}{+8.4\%}) & 71.5 \textcolor{gray}{±0.1} (\textcolor{blue}{+9.0\%}) & 68.0 \textcolor{gray}{±0.2} (\textcolor{blue}{+9.3\%}) \\
\midrule
HiResCAM & 45.2 \textcolor{gray}{±0.1} & 46.2 \textcolor{gray}{±0.3} & 43.7 \textcolor{gray}{±0.1} & 45.4 \textcolor{gray}{±0.2} \\
\textbf{Libra HiResCAM} & 67.0 \textcolor{gray}{±0.1} (\textcolor{blue}{+48.3\%}) & 64.0 \textcolor{gray}{±0.2} (\textcolor{blue}{+38.4\%}) & 68.9 \textcolor{gray}{±0.1} (\textcolor{blue}{+57.9\%}) & 65.1 \textcolor{gray}{±0.2} (\textcolor{blue}{+43.3\%}) \\
\midrule
XGradCAM+ & 65.6 \textcolor{gray}{±0.1} & 62.8 \textcolor{gray}{±0.3} & 67.1 \textcolor{gray}{±0.1} & 63.8 \textcolor{gray}{±0.2} \\
\textbf{Libra XGradCAM+} & 72.5 \textcolor{gray}{±0.1} (\textcolor{blue}{+10.5\%}) & 69.1 \textcolor{gray}{±0.2} (\textcolor{blue}{+10.1\%}) & 74.9 \textcolor{gray}{±0.1} (\textcolor{blue}{+11.6\%}) & 70.7 \textcolor{gray}{±0.2} (\textcolor{blue}{+10.8\%}) \\
\midrule
FullGrad+ & 67.1 \textcolor{gray}{±0.1} & 64.7 \textcolor{gray}{±0.3} & 69.7 \textcolor{gray}{±0.1} & 66.3 \textcolor{gray}{±0.2} \\
\textbf{Libra FullGrad+} & \textbf{77.2} \textcolor{gray}{±0.1} (\textcolor{blue}{+15.0\%}) & \textbf{73.7} \textcolor{gray}{±0.2} (\textcolor{blue}{+13.9\%}) & \textbf{80.3} \textcolor{gray}{±0.1} (\textcolor{blue}{+15.2\%}) & \textbf{76.0} \textcolor{gray}{±0.2} (\textcolor{blue}{+14.5\%}) \\
\bottomrule
\end{tabular}
\caption{Comparison of attribution methods and their LibraGrad-enhanced versions on the ViT-T model.}
\label{tbl:SRG_m_2}
\end{table}
  \endgroup%

\end{table}

\FloatBarrier
\clearpage
\subsubsection{ViT-S}
\label{per_model:ViT-S}

\begin{table}[ht]
  \fontsize{9.5pt}{8.5pt}\selectfont

  \setlength{\tabcolsep}{2pt}
  \centering
  \begingroup%
  \renewenvironment{table}[1][]%
    {\begin{center}}%
    {\end{center}}%
  \begin{table}[h]
\centering
\begin{tabular}{lccccc}
\toprule
Method & \multicolumn{2}{c}{MIF Deletion (GT)} & \multicolumn{2}{c}{MIF Deletion (Predicted)} & Segmentation \\
  & Accuracy & AOPC & Accuracy & AOPC & AP \\
\cmidrule(r){1-1}
\cmidrule(lr){2-3}
\cmidrule(lr){4-5}
\cmidrule(l){6-6}
Random & 41.8 \textcolor{gray}{±0.1} & 15.8 \textcolor{gray}{±0.3} & 33.8 \textcolor{gray}{±0.1} & 18.6 \textcolor{gray}{±0.2} & 41.9 \textcolor{gray}{±0.4} \\
RawAtt & 63.1 \textcolor{gray}{±0.1} & 36.5 \textcolor{gray}{±0.3} & 58.7 \textcolor{gray}{±0.1} & 41.2 \textcolor{gray}{±0.3} & 57.8 \textcolor{gray}{±0.3} \\
Attention Rollout & 51.2 \textcolor{gray}{±0.1} & 25.1 \textcolor{gray}{±0.4} & 45.1 \textcolor{gray}{±0.1} & 28.8 \textcolor{gray}{±0.2} & 47.1 \textcolor{gray}{±0.3} \\
AliLRP & 48.9 \textcolor{gray}{±0.1} & 22.8 \textcolor{gray}{±0.3} & 42.3 \textcolor{gray}{±0.1} & 26.2 \textcolor{gray}{±0.3} & 42.5 \textcolor{gray}{±0.4} \\
AttnLRP & 57.7 \textcolor{gray}{±0.1} & 30.9 \textcolor{gray}{±0.3} & 52.4 \textcolor{gray}{±0.1} & 35.2 \textcolor{gray}{±0.2} & 46.2 \textcolor{gray}{±0.3} \\
DecompX & 56.0 \textcolor{gray}{±0.1} & 29.5 \textcolor{gray}{±0.3} & 50.4 \textcolor{gray}{±0.1} & 33.6 \textcolor{gray}{±0.2} & 47.7 \textcolor{gray}{±0.3} \\
Integrated Gradients & 56.9 \textcolor{gray}{±0.1} & 29.1 \textcolor{gray}{±0.3} & 46.0 \textcolor{gray}{±0.1} & 29.3 \textcolor{gray}{±0.3} & 51.7 \textcolor{gray}{±0.3} \\
\midrule
\IxG{}  & 47.9 \textcolor{gray}{±0.1} & 21.6 \textcolor{gray}{±0.3} & 41.8 \textcolor{gray}{±0.1} & 25.0 \textcolor{gray}{±0.3} & 48.5 \textcolor{gray}{±0.3} \\
\textbf{Libra \IxG{} } & 54.9 \textcolor{gray}{±0.1} (\textcolor{blue}{+14.7\%}) & 28.2 \textcolor{gray}{±0.3} (\textcolor{blue}{+30.4\%}) & 49.3 \textcolor{gray}{±0.1} (\textcolor{blue}{+18.0\%}) & 32.2 \textcolor{gray}{±0.2} (\textcolor{blue}{+28.5\%}) & 46.0 \textcolor{gray}{±0.3} (\textcolor{coral}{-5.1\%}) \\
\midrule
AttCAT & 62.1 \textcolor{gray}{±0.1} & 33.4 \textcolor{gray}{±0.3} & 58.9 \textcolor{gray}{±0.1} & 38.2 \textcolor{gray}{±0.3} & 49.8 \textcolor{gray}{±0.3} \\
\textbf{Libra AttCAT} & \textbf{73.6} \textcolor{gray}{±0.1} (\textcolor{blue}{+18.5\%}) & \textbf{43.9} \textcolor{gray}{±0.3} (\textcolor{blue}{+31.4\%}) & \textbf{70.3} \textcolor{gray}{±0.1} (\textcolor{blue}{+19.3\%}) & \textbf{48.9} \textcolor{gray}{±0.3} (\textcolor{blue}{+28.0\%}) & 56.0 \textcolor{gray}{±0.3} (\textcolor{blue}{+12.4\%}) \\
\midrule
GenAtt & 69.7 \textcolor{gray}{±0.1} & 41.3 \textcolor{gray}{±0.3} & 66.3 \textcolor{gray}{±0.1} & 46.3 \textcolor{gray}{±0.3} & 65.9 \textcolor{gray}{±0.2} \\
\textbf{Libra GenAtt} & 71.7 \textcolor{gray}{±0.1} (\textcolor{blue}{+2.9\%}) & 43.2 \textcolor{gray}{±0.3} (\textcolor{blue}{+4.6\%}) & 68.2 \textcolor{gray}{±0.1} (\textcolor{blue}{+2.9\%}) & 48.2 \textcolor{gray}{±0.3} (\textcolor{blue}{+4.2\%}) & \underline{71.0} \textcolor{gray}{±0.3} (\textcolor{blue}{+7.7\%}) \\
\midrule
TokenTM & 68.9 \textcolor{gray}{±0.1} & 40.8 \textcolor{gray}{±0.3} & 65.2 \textcolor{gray}{±0.1} & 45.9 \textcolor{gray}{±0.3} & 68.2 \textcolor{gray}{±0.2} \\
\textbf{Libra TokenTM} & 70.3 \textcolor{gray}{±0.1} (\textcolor{blue}{+2.1\%}) & 42.2 \textcolor{gray}{±0.3} (\textcolor{blue}{+3.4\%}) & 66.5 \textcolor{gray}{±0.1} (\textcolor{blue}{+2.0\%}) & 47.3 \textcolor{gray}{±0.3} (\textcolor{blue}{+3.0\%}) & \textbf{71.4} \textcolor{gray}{±0.2} (\textcolor{blue}{+4.7\%}) \\
\midrule
GradCAM+ & 59.9 \textcolor{gray}{±0.1} & 31.5 \textcolor{gray}{±0.3} & 55.5 \textcolor{gray}{±0.1} & 35.8 \textcolor{gray}{±0.3} & 46.4 \textcolor{gray}{±0.4} \\
\textbf{Libra GradCAM+} & 70.2 \textcolor{gray}{±0.1} (\textcolor{blue}{+17.0\%}) & 41.2 \textcolor{gray}{±0.3} (\textcolor{blue}{+30.7\%}) & 66.5 \textcolor{gray}{±0.1} (\textcolor{blue}{+19.7\%}) & 46.1 \textcolor{gray}{±0.3} (\textcolor{blue}{+28.7\%}) & 60.7 \textcolor{gray}{±0.4} (\textcolor{blue}{+30.8\%}) \\
\midrule
HiResCAM & 38.4 \textcolor{gray}{±0.1} & 13.1 \textcolor{gray}{±0.2} & 29.5 \textcolor{gray}{±0.1} & 15.3 \textcolor{gray}{±0.2} & 48.4 \textcolor{gray}{±0.4} \\
\textbf{Libra HiResCAM} & 67.4 \textcolor{gray}{±0.1} (\textcolor{blue}{+75.5\%}) & 39.6 \textcolor{gray}{±0.3} (\textcolor{blue}{+202.6\%}) & 63.4 \textcolor{gray}{±0.1} (\textcolor{blue}{+114.7\%}) & 44.4 \textcolor{gray}{±0.2} (\textcolor{blue}{+190.6\%}) & 69.4 \textcolor{gray}{±0.3} (\textcolor{blue}{+43.2\%}) \\
\midrule
XGradCAM+ & 60.3 \textcolor{gray}{±0.1} & 31.9 \textcolor{gray}{±0.4} & 55.9 \textcolor{gray}{±0.1} & 36.2 \textcolor{gray}{±0.3} & 45.4 \textcolor{gray}{±0.4} \\
\textbf{Libra XGradCAM+} & 72.1 \textcolor{gray}{±0.1} (\textcolor{blue}{+19.5\%}) & 42.8 \textcolor{gray}{±0.3} (\textcolor{blue}{+34.1\%}) & 68.5 \textcolor{gray}{±0.1} (\textcolor{blue}{+22.4\%}) & 47.8 \textcolor{gray}{±0.3} (\textcolor{blue}{+32.0\%}) & 62.3 \textcolor{gray}{±0.4} (\textcolor{blue}{+37.2\%}) \\
\midrule
FullGrad+ & 59.6 \textcolor{gray}{±0.1} & 31.5 \textcolor{gray}{±0.3} & 55.8 \textcolor{gray}{±0.1} & 36.1 \textcolor{gray}{±0.3} & 50.0 \textcolor{gray}{±0.3} \\
\textbf{Libra FullGrad+} & \underline{73.5} \textcolor{gray}{±0.1} (\textcolor{blue}{+23.3\%}) & \underline{43.8} \textcolor{gray}{±0.3} (\textcolor{blue}{+39.0\%}) & \underline{70.1} \textcolor{gray}{±0.1} (\textcolor{blue}{+25.8\%}) & \textbf{48.9} \textcolor{gray}{±0.3} (\textcolor{blue}{+35.3\%}) & 59.6 \textcolor{gray}{±0.3} (\textcolor{blue}{+19.2\%}) \\
\bottomrule
\end{tabular}
\caption{Comparison of attribution methods and their LibraGrad-enhanced versions on the ViT-S model. We report faithfulness metrics using Most-Influential-First Deletion, MIF with ground-truth (GT) and predicted labels, including Accuracy and Area Over Perturbation Curve (AOPC) and Segmentation Average Precision (AP). The results demonstrate that composing existing methods with LibraGrad significantly enhances their performance across all metrics.}
\label{tbl:MIF_m_2}
\end{table}
  \endgroup%

\end{table}

\begin{table}[ht]
  \fontsize{11pt}{8.5pt}\selectfont

  \setlength{\tabcolsep}{4pt}
  \centering

  \begingroup%
  \renewenvironment{table}[1][]%
    {\begin{center}}%
    {\end{center}}%
  \begin{table}[h]
\centering
\begin{tabular}{lcccc}
\toprule
Method & \multicolumn{2}{c}{LIF Deletion (GT)} & \multicolumn{2}{c}{LIF Deletion (Predicted)} \\
  & Accuracy & AOPC & Accuracy & AOPC \\
\cmidrule(r){1-1}
\cmidrule(lr){2-3}
\cmidrule(l){4-5}
Random & 57.7 \textcolor{gray}{±0.1} & 84.1 \textcolor{gray}{±0.2} & 66.5 \textcolor{gray}{±0.1} & 81.8 \textcolor{gray}{±0.2} \\
RawAtt & 63.9 \textcolor{gray}{±0.1} & 89.4 \textcolor{gray}{±0.2} & 72.8 \textcolor{gray}{±0.1} & 87.2 \textcolor{gray}{±0.2} \\
Attention Rollout & 58.6 \textcolor{gray}{±0.1} & 84.8 \textcolor{gray}{±0.2} & 67.3 \textcolor{gray}{±0.1} & 82.4 \textcolor{gray}{±0.3} \\
AliLRP & 62.5 \textcolor{gray}{±0.1} & 88.2 \textcolor{gray}{±0.2} & 70.6 \textcolor{gray}{±0.1} & 85.8 \textcolor{gray}{±0.2} \\
AttnLRP & 68.4 \textcolor{gray}{±0.1} & 94.6 \textcolor{gray}{±0.2} & 77.3 \textcolor{gray}{±0.1} & 92.8 \textcolor{gray}{±0.2} \\
DecompX & 66.5 \textcolor{gray}{±0.1} & 92.2 \textcolor{gray}{±0.2} & 75.3 \textcolor{gray}{±0.1} & 90.3 \textcolor{gray}{±0.2} \\
Integrated Gradients & 69.6 \textcolor{gray}{±0.1} & 95.4 \textcolor{gray}{±0.2} & 73.9 \textcolor{gray}{±0.1} & 90.1 \textcolor{gray}{±0.2} \\
\midrule
\IxG{}  & 64.2 \textcolor{gray}{±0.1} & 90.0 \textcolor{gray}{±0.3} & 72.1 \textcolor{gray}{±0.1} & 87.8 \textcolor{gray}{±0.3} \\
\textbf{Libra \IxG{} } & 66.4 \textcolor{gray}{±0.1} (\textcolor{blue}{+3.4\%}) & 92.0 \textcolor{gray}{±0.2} (\textcolor{blue}{+2.1\%}) & 74.3 \textcolor{gray}{±0.1} (\textcolor{blue}{+3.0\%}) & 89.6 \textcolor{gray}{±0.2} (\textcolor{blue}{+2.1\%}) \\
\midrule
AttCAT & 71.9 \textcolor{gray}{±0.1} & 97.7 \textcolor{gray}{±0.2} & 78.5 \textcolor{gray}{±0.1} & 95.6 \textcolor{gray}{±0.2} \\
\textbf{Libra AttCAT} & \underline{74.2} \textcolor{gray}{±0.1} (\textcolor{blue}{+3.3\%}) & \underline{100.0} \textcolor{gray}{±0.2} (\textcolor{blue}{+2.4\%}) & \textbf{81.0} \textcolor{gray}{±0.1} (\textcolor{blue}{+3.2\%}) & \underline{97.8} \textcolor{gray}{±0.2} (\textcolor{blue}{+2.3\%}) \\
\midrule
GenAtt & 69.6 \textcolor{gray}{±0.1} & 94.7 \textcolor{gray}{±0.2} & 79.1 \textcolor{gray}{±0.1} & 92.9 \textcolor{gray}{±0.2} \\
\textbf{Libra GenAtt} & 69.5 \textcolor{gray}{±0.1} (\textcolor{coral}{-0.2\%}) & 94.5 \textcolor{gray}{±0.2} (\textcolor{coral}{-0.2\%}) & 79.0 \textcolor{gray}{±0.1} (\textcolor{coral}{-0.1\%}) & 92.7 \textcolor{gray}{±0.2} (\textcolor{coral}{-0.2\%}) \\
\midrule
TokenTM & 67.4 \textcolor{gray}{±0.1} & 92.7 \textcolor{gray}{±0.2} & 77.2 \textcolor{gray}{±0.1} & 90.8 \textcolor{gray}{±0.2} \\
\textbf{Libra TokenTM} & 67.4 \textcolor{gray}{±0.1} (\textcolor{coral}{-0.1\%}) & 92.8 \textcolor{gray}{±0.2} (\textcolor{blue}{+0.1\%}) & 77.1 \textcolor{gray}{±0.1} (\textcolor{coral}{-0.1\%}) & 90.8 \textcolor{gray}{±0.2} (+0.0\%) \\
\midrule
GradCAM+ & 65.0 \textcolor{gray}{±0.1} & 90.5 \textcolor{gray}{±0.2} & 71.9 \textcolor{gray}{±0.1} & 88.1 \textcolor{gray}{±0.2} \\
\textbf{Libra GradCAM+} & 70.7 \textcolor{gray}{±0.1} (\textcolor{blue}{+8.9\%}) & 96.0 \textcolor{gray}{±0.2} (\textcolor{blue}{+6.1\%}) & 78.0 \textcolor{gray}{±0.1} (\textcolor{blue}{+8.4\%}) & 93.7 \textcolor{gray}{±0.2} (\textcolor{blue}{+6.4\%}) \\
\midrule
HiResCAM & 55.8 \textcolor{gray}{±0.1} & 81.8 \textcolor{gray}{±0.3} & 62.8 \textcolor{gray}{±0.1} & 78.7 \textcolor{gray}{±0.3} \\
\textbf{Libra HiResCAM} & 68.5 \textcolor{gray}{±0.1} (\textcolor{blue}{+22.6\%}) & 93.3 \textcolor{gray}{±0.2} (\textcolor{blue}{+14.1\%}) & 76.1 \textcolor{gray}{±0.1} (\textcolor{blue}{+21.2\%}) & 91.1 \textcolor{gray}{±0.2} (\textcolor{blue}{+15.7\%}) \\
\midrule
XGradCAM+ & 66.3 \textcolor{gray}{±0.1} & 91.8 \textcolor{gray}{±0.2} & 73.5 \textcolor{gray}{±0.1} & 89.5 \textcolor{gray}{±0.2} \\
\textbf{Libra XGradCAM+} & 71.4 \textcolor{gray}{±0.1} (\textcolor{blue}{+7.7\%}) & 96.5 \textcolor{gray}{±0.2} (\textcolor{blue}{+5.1\%}) & 78.5 \textcolor{gray}{±0.1} (\textcolor{blue}{+6.8\%}) & 94.1 \textcolor{gray}{±0.2} (\textcolor{blue}{+5.2\%}) \\
\midrule
FullGrad+ & 70.3 \textcolor{gray}{±0.1} & 96.3 \textcolor{gray}{±0.2} & 77.6 \textcolor{gray}{±0.1} & 94.3 \textcolor{gray}{±0.2} \\
\textbf{Libra FullGrad+} & \textbf{74.4} \textcolor{gray}{±0.1} (\textcolor{blue}{+5.9\%}) & \textbf{100.1} \textcolor{gray}{±0.2} (\textcolor{blue}{+4.0\%}) & \textbf{81.0} \textcolor{gray}{±0.1} (\textcolor{blue}{+4.4\%}) & \textbf{97.9} \textcolor{gray}{±0.2} (\textcolor{blue}{+3.8\%}) \\
\bottomrule
\end{tabular}
\caption{Comparison of attribution methods and their LibraGrad-enhanced versions on the ViT-S model.}
\label{tbl:LIF_m_2}
\end{table}
  \endgroup%

  \begingroup%
  \renewenvironment{table}[1][]%
    {\begin{center}}%
    {\end{center}}%
  \begin{table}[h]
\centering
\begin{tabular}{lcccc}
\toprule
Method & \multicolumn{2}{c}{SRG (GT)} & \multicolumn{2}{c}{SRG (Predicted)} \\
  & Accuracy & AOPC & Accuracy & AOPC \\
\cmidrule(r){1-1}
\cmidrule(lr){2-3}
\cmidrule(l){4-5}
Random & 49.7 \textcolor{gray}{±0.1} & 49.9 \textcolor{gray}{±0.3} & 50.2 \textcolor{gray}{±0.1} & 50.2 \textcolor{gray}{±0.2} \\
RawAtt & 63.5 \textcolor{gray}{±0.1} & 63.0 \textcolor{gray}{±0.3} & 65.8 \textcolor{gray}{±0.1} & 64.2 \textcolor{gray}{±0.3} \\
Attention Rollout & 54.9 \textcolor{gray}{±0.1} & 54.9 \textcolor{gray}{±0.3} & 56.2 \textcolor{gray}{±0.1} & 55.6 \textcolor{gray}{±0.2} \\
AliLRP & 55.7 \textcolor{gray}{±0.1} & 55.5 \textcolor{gray}{±0.3} & 56.5 \textcolor{gray}{±0.1} & 56.0 \textcolor{gray}{±0.3} \\
AttnLRP & 63.0 \textcolor{gray}{±0.1} & 62.8 \textcolor{gray}{±0.3} & 64.9 \textcolor{gray}{±0.1} & 64.0 \textcolor{gray}{±0.2} \\
DecompX & 61.3 \textcolor{gray}{±0.1} & 60.9 \textcolor{gray}{±0.3} & 62.9 \textcolor{gray}{±0.1} & 61.9 \textcolor{gray}{±0.2} \\
Integrated Gradients & 63.3 \textcolor{gray}{±0.1} & 62.3 \textcolor{gray}{±0.3} & 59.9 \textcolor{gray}{±0.1} & 59.7 \textcolor{gray}{±0.3} \\
\midrule
\IxG{}  & 56.1 \textcolor{gray}{±0.1} & 55.8 \textcolor{gray}{±0.3} & 57.0 \textcolor{gray}{±0.1} & 56.4 \textcolor{gray}{±0.3} \\
\textbf{Libra \IxG{} } & 60.7 \textcolor{gray}{±0.1} (\textcolor{blue}{+8.2\%}) & 60.1 \textcolor{gray}{±0.3} (\textcolor{blue}{+7.6\%}) & 61.8 \textcolor{gray}{±0.1} (\textcolor{blue}{+8.5\%}) & 60.9 \textcolor{gray}{±0.2} (\textcolor{blue}{+8.0\%}) \\
\midrule
AttCAT & 67.0 \textcolor{gray}{±0.1} & 65.5 \textcolor{gray}{±0.3} & 68.7 \textcolor{gray}{±0.1} & 66.9 \textcolor{gray}{±0.2} \\
\textbf{Libra AttCAT} & \underline{73.9} \textcolor{gray}{±0.1} (\textcolor{blue}{+10.3\%}) & \underline{71.9} \textcolor{gray}{±0.3} (\textcolor{blue}{+9.8\%}) & \textbf{75.7} \textcolor{gray}{±0.1} (\textcolor{blue}{+10.1\%}) & \underline{73.3} \textcolor{gray}{±0.3} (\textcolor{blue}{+9.6\%}) \\
\midrule
GenAtt & 69.6 \textcolor{gray}{±0.1} & 68.0 \textcolor{gray}{±0.3} & 72.7 \textcolor{gray}{±0.1} & 69.6 \textcolor{gray}{±0.2} \\
\textbf{Libra GenAtt} & 70.6 \textcolor{gray}{±0.1} (\textcolor{blue}{+1.4\%}) & 68.8 \textcolor{gray}{±0.3} (\textcolor{blue}{+1.3\%}) & 73.6 \textcolor{gray}{±0.1} (\textcolor{blue}{+1.3\%}) & 70.5 \textcolor{gray}{±0.3} (\textcolor{blue}{+1.2\%}) \\
\midrule
TokenTM & 68.2 \textcolor{gray}{±0.1} & 66.8 \textcolor{gray}{±0.3} & 71.2 \textcolor{gray}{±0.1} & 68.3 \textcolor{gray}{±0.2} \\
\textbf{Libra TokenTM} & 68.8 \textcolor{gray}{±0.1} (\textcolor{blue}{+1.0\%}) & 67.5 \textcolor{gray}{±0.3} (\textcolor{blue}{+1.1\%}) & 71.8 \textcolor{gray}{±0.1} (\textcolor{blue}{+0.9\%}) & 69.1 \textcolor{gray}{±0.3} (\textcolor{blue}{+1.0\%}) \\
\midrule
GradCAM+ & 62.5 \textcolor{gray}{±0.1} & 61.0 \textcolor{gray}{±0.3} & 63.7 \textcolor{gray}{±0.1} & 62.0 \textcolor{gray}{±0.3} \\
\textbf{Libra GradCAM+} & 70.4 \textcolor{gray}{±0.1} (\textcolor{blue}{+12.8\%}) & 68.6 \textcolor{gray}{±0.3} (\textcolor{blue}{+12.4\%}) & 72.2 \textcolor{gray}{±0.1} (\textcolor{blue}{+13.3\%}) & 69.9 \textcolor{gray}{±0.3} (\textcolor{blue}{+12.8\%}) \\
\midrule
HiResCAM & 47.1 \textcolor{gray}{±0.1} & 47.4 \textcolor{gray}{±0.3} & 46.2 \textcolor{gray}{±0.1} & 47.0 \textcolor{gray}{±0.3} \\
\textbf{Libra HiResCAM} & 68.0 \textcolor{gray}{±0.1} (\textcolor{blue}{+44.2\%}) & 66.4 \textcolor{gray}{±0.2} (\textcolor{blue}{+40.0\%}) & 69.8 \textcolor{gray}{±0.1} (\textcolor{blue}{+51.1\%}) & 67.7 \textcolor{gray}{±0.2} (\textcolor{blue}{+44.1\%}) \\
\midrule
XGradCAM+ & 63.3 \textcolor{gray}{±0.1} & 61.9 \textcolor{gray}{±0.3} & 64.7 \textcolor{gray}{±0.1} & 62.8 \textcolor{gray}{±0.3} \\
\textbf{Libra XGradCAM+} & 71.7 \textcolor{gray}{±0.1} (\textcolor{blue}{+13.3\%}) & 69.7 \textcolor{gray}{±0.3} (\textcolor{blue}{+12.6\%}) & 73.5 \textcolor{gray}{±0.1} (\textcolor{blue}{+13.5\%}) & 71.0 \textcolor{gray}{±0.2} (\textcolor{blue}{+12.9\%}) \\
\midrule
FullGrad+ & 65.0 \textcolor{gray}{±0.1} & 63.9 \textcolor{gray}{±0.3} & 66.7 \textcolor{gray}{±0.1} & 65.2 \textcolor{gray}{±0.3} \\
\textbf{Libra FullGrad+} & \textbf{74.0} \textcolor{gray}{±0.1} (\textcolor{blue}{+13.9\%}) & \textbf{72.0} \textcolor{gray}{±0.3} (\textcolor{blue}{+12.6\%}) & \underline{75.6} \textcolor{gray}{±0.1} (\textcolor{blue}{+13.3\%}) & \textbf{73.4} \textcolor{gray}{±0.3} (\textcolor{blue}{+12.5\%}) \\
\bottomrule
\end{tabular}
\caption{Comparison of attribution methods and their LibraGrad-enhanced versions on the ViT-S model.}
\label{tbl:SRG_m_2}
\end{table}
  \endgroup%

\end{table}

\FloatBarrier
\clearpage
\subsubsection{ViT-B}
\label{per_model:ViT-B}

\begin{table}[ht]
  \fontsize{9.5pt}{8.5pt}\selectfont

  \setlength{\tabcolsep}{2pt}
  \centering
  \begingroup%
  \renewenvironment{table}[1][]%
    {\begin{center}}%
    {\end{center}}%
  \begin{table}[h]
\centering
\begin{tabular}{lccccc}
\toprule
Method & \multicolumn{2}{c}{MIF Deletion (GT)} & \multicolumn{2}{c}{MIF Deletion (Predicted)} & Segmentation \\
  & Accuracy & AOPC & Accuracy & AOPC & AP \\
\cmidrule(r){1-1}
\cmidrule(lr){2-3}
\cmidrule(lr){4-5}
\cmidrule(l){6-6}
Random & 34.5 \textcolor{gray}{±0.1} & 12.3 \textcolor{gray}{±0.2} & 26.5 \textcolor{gray}{±0.1} & 14.2 \textcolor{gray}{±0.2} & 41.9 \textcolor{gray}{±0.4} \\
RawAtt & 50.1 \textcolor{gray}{±0.1} & 25.0 \textcolor{gray}{±0.3} & 44.6 \textcolor{gray}{±0.1} & 27.9 \textcolor{gray}{±0.3} & 46.9 \textcolor{gray}{±0.3} \\
Attention Rollout & 41.9 \textcolor{gray}{±0.1} & 18.8 \textcolor{gray}{±0.3} & 35.4 \textcolor{gray}{±0.1} & 21.2 \textcolor{gray}{±0.2} & 45.3 \textcolor{gray}{±0.3} \\
AliLRP & 39.8 \textcolor{gray}{±0.1} & 16.7 \textcolor{gray}{±0.2} & 33.3 \textcolor{gray}{±0.1} & 19.1 \textcolor{gray}{±0.2} & 43.8 \textcolor{gray}{±0.4} \\
AttnLRP & 44.5 \textcolor{gray}{±0.1} & 20.8 \textcolor{gray}{±0.3} & 38.5 \textcolor{gray}{±0.1} & 23.4 \textcolor{gray}{±0.2} & 42.0 \textcolor{gray}{±0.4} \\
DecompX & 44.0 \textcolor{gray}{±0.1} & 20.3 \textcolor{gray}{±0.3} & 37.8 \textcolor{gray}{±0.1} & 22.8 \textcolor{gray}{±0.2} & 44.3 \textcolor{gray}{±0.3} \\
Integrated Gradients & 46.9 \textcolor{gray}{±0.1} & 22.5 \textcolor{gray}{±0.2} & 35.4 \textcolor{gray}{±0.1} & 21.4 \textcolor{gray}{±0.2} & 47.5 \textcolor{gray}{±0.3} \\
\midrule
\IxG{}  & 40.4 \textcolor{gray}{±0.1} & 17.7 \textcolor{gray}{±0.2} & 34.4 \textcolor{gray}{±0.1} & 20.2 \textcolor{gray}{±0.2} & 44.8 \textcolor{gray}{±0.3} \\
\textbf{Libra \IxG{} } & 44.8 \textcolor{gray}{±0.1} (\textcolor{blue}{+10.8\%}) & 20.8 \textcolor{gray}{±0.3} (\textcolor{blue}{+17.4\%}) & 38.6 \textcolor{gray}{±0.1} (\textcolor{blue}{+12.0\%}) & 23.4 \textcolor{gray}{±0.2} (\textcolor{blue}{+15.8\%}) & 44.4 \textcolor{gray}{±0.3} (\textcolor{coral}{-0.9\%}) \\
\midrule
AttCAT & 50.4 \textcolor{gray}{±0.1} & 25.3 \textcolor{gray}{±0.2} & 46.9 \textcolor{gray}{±0.1} & 28.8 \textcolor{gray}{±0.2} & 44.5 \textcolor{gray}{±0.3} \\
\textbf{Libra AttCAT} & \underline{66.4} \textcolor{gray}{±0.1} (\textcolor{blue}{+31.7\%}) & \underline{37.5} \textcolor{gray}{±0.3} (\textcolor{blue}{+47.9\%}) & \underline{63.5} \textcolor{gray}{±0.1} (\textcolor{blue}{+35.4\%}) & \textbf{41.5} \textcolor{gray}{±0.3} (\textcolor{blue}{+44.2\%}) & 61.5 \textcolor{gray}{±0.3} (\textcolor{blue}{+38.3\%}) \\
\midrule
GenAtt & 61.9 \textcolor{gray}{±0.1} & 34.2 \textcolor{gray}{±0.3} & 58.2 \textcolor{gray}{±0.1} & 37.9 \textcolor{gray}{±0.2} & 71.0 \textcolor{gray}{±0.2} \\
\textbf{Libra GenAtt} & 65.1 \textcolor{gray}{±0.1} (\textcolor{blue}{+5.1\%}) & 36.6 \textcolor{gray}{±0.3} (\textcolor{blue}{+6.8\%}) & 61.6 \textcolor{gray}{±0.1} (\textcolor{blue}{+5.8\%}) & 40.4 \textcolor{gray}{±0.3} (\textcolor{blue}{+6.6\%}) & \textbf{77.5} \textcolor{gray}{±0.2} (\textcolor{blue}{+9.2\%}) \\
\midrule
TokenTM & 60.6 \textcolor{gray}{±0.1} & 33.8 \textcolor{gray}{±0.3} & 56.8 \textcolor{gray}{±0.1} & 37.4 \textcolor{gray}{±0.3} & 70.2 \textcolor{gray}{±0.2} \\
\textbf{Libra TokenTM} & 62.8 \textcolor{gray}{±0.1} (\textcolor{blue}{+3.6\%}) & 35.1 \textcolor{gray}{±0.3} (\textcolor{blue}{+4.0\%}) & 59.1 \textcolor{gray}{±0.1} (\textcolor{blue}{+4.1\%}) & 38.9 \textcolor{gray}{±0.3} (\textcolor{blue}{+3.8\%}) & 73.9 \textcolor{gray}{±0.2} (\textcolor{blue}{+5.2\%}) \\
\midrule
GradCAM+ & 50.5 \textcolor{gray}{±0.1} & 24.8 \textcolor{gray}{±0.2} & 45.6 \textcolor{gray}{±0.1} & 27.6 \textcolor{gray}{±0.2} & 50.2 \textcolor{gray}{±0.4} \\
\textbf{Libra GradCAM+} & 65.3 \textcolor{gray}{±0.1} (\textcolor{blue}{+29.3\%}) & 35.9 \textcolor{gray}{±0.2} (\textcolor{blue}{+44.8\%}) & 61.4 \textcolor{gray}{±0.1} (\textcolor{blue}{+34.8\%}) & 39.6 \textcolor{gray}{±0.2} (\textcolor{blue}{+43.5\%}) & 72.1 \textcolor{gray}{±0.3} (\textcolor{blue}{+43.6\%}) \\
\midrule
HiResCAM & 50.4 \textcolor{gray}{±0.1} & 25.4 \textcolor{gray}{±0.3} & 45.4 \textcolor{gray}{±0.1} & 28.5 \textcolor{gray}{±0.2} & 59.0 \textcolor{gray}{±0.3} \\
\textbf{Libra HiResCAM} & 60.8 \textcolor{gray}{±0.1} (\textcolor{blue}{+20.6\%}) & 33.4 \textcolor{gray}{±0.3} (\textcolor{blue}{+31.7\%}) & 56.7 \textcolor{gray}{±0.1} (\textcolor{blue}{+24.8\%}) & 37.0 \textcolor{gray}{±0.2} (\textcolor{blue}{+29.6\%}) & 72.6 \textcolor{gray}{±0.3} (\textcolor{blue}{+23.1\%}) \\
\midrule
XGradCAM+ & 44.0 \textcolor{gray}{±0.1} & 19.0 \textcolor{gray}{±0.2} & 38.6 \textcolor{gray}{±0.1} & 21.5 \textcolor{gray}{±0.2} & 41.0 \textcolor{gray}{±0.4} \\
\textbf{Libra XGradCAM+} & \textbf{67.4} \textcolor{gray}{±0.1} (\textcolor{blue}{+53.0\%}) & \textbf{37.7} \textcolor{gray}{±0.2} (\textcolor{blue}{+98.6\%}) & \textbf{63.9} \textcolor{gray}{±0.1} (\textcolor{blue}{+65.6\%}) & \textbf{41.5} \textcolor{gray}{±0.2} (\textcolor{blue}{+92.8\%}) & \underline{75.0} \textcolor{gray}{±0.3} (\textcolor{blue}{+82.8\%}) \\
\midrule
FullGrad+ & 48.2 \textcolor{gray}{±0.1} & 23.1 \textcolor{gray}{±0.3} & 44.2 \textcolor{gray}{±0.1} & 26.3 \textcolor{gray}{±0.2} & 45.2 \textcolor{gray}{±0.3} \\
\textbf{Libra FullGrad+} & 66.1 \textcolor{gray}{±0.1} (\textcolor{blue}{+37.1\%}) & 37.2 \textcolor{gray}{±0.3} (\textcolor{blue}{+60.9\%}) & 63.1 \textcolor{gray}{±0.1} (\textcolor{blue}{+42.9\%}) & 41.2 \textcolor{gray}{±0.3} (\textcolor{blue}{+56.7\%}) & 65.5 \textcolor{gray}{±0.3} (\textcolor{blue}{+44.8\%}) \\
\bottomrule
\end{tabular}
\caption{Comparison of attribution methods and their LibraGrad-enhanced versions on the ViT-B model. We report faithfulness metrics using Most-Influential-First Deletion, MIF with ground-truth (GT) and predicted labels, including Accuracy and Area Over Perturbation Curve (AOPC) and Segmentation Average Precision (AP). The results demonstrate that composing existing methods with LibraGrad significantly enhances their performance across all metrics.}
\label{tbl:MIF_m_2}
\end{table}
  \endgroup%

\end{table}

\begin{table}[ht]
  \fontsize{11pt}{8.5pt}\selectfont

  \setlength{\tabcolsep}{4pt}
  \centering

  \begingroup%
  \renewenvironment{table}[1][]%
    {\begin{center}}%
    {\end{center}}%
  \begin{table}[h]
\centering
\begin{tabular}{lcccc}
\toprule
Method & \multicolumn{2}{c}{LIF Deletion (GT)} & \multicolumn{2}{c}{LIF Deletion (Predicted)} \\
  & Accuracy & AOPC & Accuracy & AOPC \\
\cmidrule(r){1-1}
\cmidrule(lr){2-3}
\cmidrule(l){4-5}
Random & 65.2 \textcolor{gray}{±0.1} & 87.5 \textcolor{gray}{±0.2} & 73.3 \textcolor{gray}{±0.1} & 85.8 \textcolor{gray}{±0.2} \\
RawAtt & 67.5 \textcolor{gray}{±0.1} & 89.2 \textcolor{gray}{±0.1} & 76.2 \textcolor{gray}{±0.1} & 87.6 \textcolor{gray}{±0.1} \\
Attention Rollout & 65.9 \textcolor{gray}{±0.1} & 87.9 \textcolor{gray}{±0.2} & 73.8 \textcolor{gray}{±0.1} & 86.0 \textcolor{gray}{±0.2} \\
AliLRP & 69.9 \textcolor{gray}{±0.1} & 90.9 \textcolor{gray}{±0.2} & 77.8 \textcolor{gray}{±0.1} & 89.3 \textcolor{gray}{±0.2} \\
AttnLRP & 71.0 \textcolor{gray}{±0.1} & 92.4 \textcolor{gray}{±0.2} & 78.7 \textcolor{gray}{±0.1} & 90.8 \textcolor{gray}{±0.1} \\
DecompX & 71.1 \textcolor{gray}{±0.1} & 92.2 \textcolor{gray}{±0.2} & 79.1 \textcolor{gray}{±0.1} & 90.6 \textcolor{gray}{±0.1} \\
\bottomrule
\end{tabular}
\caption{Comparison of attribution methods and their LibraGrad-enhanced versions on the ViT-B model.}
\label{tbl:LIF_m_2}
\end{table}
  \endgroup%

  \begingroup%
  \renewenvironment{table}[1][]%
    {\begin{center}}%
    {\end{center}}%
  \begin{table}[h]
\centering
\begin{tabular}{lcccc}
\toprule
Method & \multicolumn{2}{c}{SRG (GT)} & \multicolumn{2}{c}{SRG (Predicted)} \\
  & Accuracy & AOPC & Accuracy & AOPC \\
\cmidrule(r){1-1}
\cmidrule(lr){2-3}
\cmidrule(l){4-5}
Random & 49.9 \textcolor{gray}{±0.1} & 49.9 \textcolor{gray}{±0.2} & 49.9 \textcolor{gray}{±0.1} & 50.0 \textcolor{gray}{±0.2} \\
RawAtt & 58.8 \textcolor{gray}{±0.1} & 57.1 \textcolor{gray}{±0.2} & 60.4 \textcolor{gray}{±0.1} & 57.8 \textcolor{gray}{±0.2} \\
Attention Rollout & 53.9 \textcolor{gray}{±0.1} & 53.3 \textcolor{gray}{±0.2} & 54.6 \textcolor{gray}{±0.1} & 53.6 \textcolor{gray}{±0.2} \\
AliLRP & 54.8 \textcolor{gray}{±0.1} & 53.8 \textcolor{gray}{±0.2} & 55.5 \textcolor{gray}{±0.1} & 54.2 \textcolor{gray}{±0.2} \\
AttnLRP & 57.8 \textcolor{gray}{±0.1} & 56.6 \textcolor{gray}{±0.2} & 58.6 \textcolor{gray}{±0.1} & 57.1 \textcolor{gray}{±0.2} \\
DecompX & 57.6 \textcolor{gray}{±0.1} & 56.3 \textcolor{gray}{±0.2} & 58.5 \textcolor{gray}{±0.1} & 56.7 \textcolor{gray}{±0.2} \\
\bottomrule
\end{tabular}
\caption{Comparison of attribution methods and their LibraGrad-enhanced versions on the ViT-B model.}
\label{tbl:SRG_m_2}
\end{table}
  \endgroup%

\end{table}

\FloatBarrier
\clearpage
\subsubsection{ImageNet-Hard ViT-B}
\label{per_model:ImageNet-Hard_ViT-B}

\begin{table}[ht]
  \fontsize{9.5pt}{8.5pt}\selectfont

  \setlength{\tabcolsep}{2pt}
  \centering
  \begingroup%
  \renewenvironment{table}[1][]%
    {\begin{center}}%
    {\end{center}}%
  \begin{table}[h]
\centering
\begin{tabular}{lcccc}
\toprule
Method & \multicolumn{2}{c}{MIF Deletion (GT)} & \multicolumn{2}{c}{MIF Deletion (Predicted)} \\
  & Accuracy & AOPC & Accuracy & AOPC \\
\cmidrule(r){1-1}
\cmidrule(lr){2-3}
\cmidrule(l){4-5}
Random & 81.7 \textcolor{gray}{±0.1} & 6.6 \textcolor{gray}{±0.1} & 52.4 \textcolor{gray}{±0.1} & 16.4 \textcolor{gray}{±0.2} \\
RawAtt & 85.9 \textcolor{gray}{±0.1} & 9.5 \textcolor{gray}{±0.1} & 65.9 \textcolor{gray}{±0.1} & 23.8 \textcolor{gray}{±0.2} \\
Attention Rollout & 84.7 \textcolor{gray}{±0.1} & 8.7 \textcolor{gray}{±0.1} & 62.2 \textcolor{gray}{±0.1} & 21.7 \textcolor{gray}{±0.2} \\
AliLRP & 85.7 \textcolor{gray}{±0.1} & 9.1 \textcolor{gray}{±0.1} & 64.1 \textcolor{gray}{±0.1} & 22.6 \textcolor{gray}{±0.2} \\
AttnLRP & 88.2 \textcolor{gray}{±0.1} & 10.8 \textcolor{gray}{±0.2} & 70.8 \textcolor{gray}{±0.1} & 26.3 \textcolor{gray}{±0.2} \\
DecompX & 87.1 \textcolor{gray}{±0.1} & 10.2 \textcolor{gray}{±0.1} & 67.7 \textcolor{gray}{±0.1} & 25.0 \textcolor{gray}{±0.2} \\
\bottomrule
\end{tabular}
\caption{Comparison of attribution methods and their LibraGrad-enhanced versions on the ImageNet-Hard ViT-B model.}
\label{tbl:MIF_m_2}
\end{table}
  \endgroup%

\end{table}

\begin{table}[ht]
  \fontsize{11pt}{8.5pt}\selectfont

  \setlength{\tabcolsep}{4pt}
  \centering

  \begingroup%
  \renewenvironment{table}[1][]%
    {\begin{center}}%
    {\end{center}}%
  \begin{table}[h]
\centering
\begin{tabular}{lcccc}
\toprule
Method & \multicolumn{2}{c}{LIF Deletion (GT)} & \multicolumn{2}{c}{LIF Deletion (Predicted)} \\
  & Accuracy & AOPC & Accuracy & AOPC \\
\cmidrule(r){1-1}
\cmidrule(lr){2-3}
\cmidrule(l){4-5}
Random & 18.6 \textcolor{gray}{±0.1} & 93.5 \textcolor{gray}{±0.1} & 47.0 \textcolor{gray}{±0.1} & 83.1 \textcolor{gray}{±0.2} \\
RawAtt & 19.5 \textcolor{gray}{±0.1} & 94.2 \textcolor{gray}{±0.1} & 52.0 \textcolor{gray}{±0.1} & 85.5 \textcolor{gray}{±0.2} \\
Attention Rollout & 19.8 \textcolor{gray}{±0.1} & 94.0 \textcolor{gray}{±0.1} & 49.8 \textcolor{gray}{±0.1} & 84.1 \textcolor{gray}{±0.2} \\
AliLRP & 21.5 \textcolor{gray}{±0.1} & 95.2 \textcolor{gray}{±0.1} & 53.5 \textcolor{gray}{±0.1} & 87.3 \textcolor{gray}{±0.2} \\
AttnLRP & 26.4 \textcolor{gray}{±0.1} & 98.2 \textcolor{gray}{±0.1} & 61.6 \textcolor{gray}{±0.1} & 93.4 \textcolor{gray}{±0.2} \\
DecompX & 22.6 \textcolor{gray}{±0.1} & 96.2 \textcolor{gray}{±0.1} & 56.8 \textcolor{gray}{±0.1} & 89.4 \textcolor{gray}{±0.2} \\
\bottomrule
\end{tabular}
\caption{Comparison of attribution methods and their LibraGrad-enhanced versions on the ImageNet-Hard ViT-B model.}
\label{tbl:LIF_m_2}
\end{table}
  \endgroup%

  \begingroup%
  \renewenvironment{table}[1][]%
    {\begin{center}}%
    {\end{center}}%
  \begin{table}[h]
\centering
\begin{tabular}{lcccc}
\toprule
Method & \multicolumn{2}{c}{SRG (GT)} & \multicolumn{2}{c}{SRG (Predicted)} \\
  & Accuracy & AOPC & Accuracy & AOPC \\
\cmidrule(r){1-1}
\cmidrule(lr){2-3}
\cmidrule(l){4-5}
Random & 50.1 \textcolor{gray}{±0.1} & 50.0 \textcolor{gray}{±0.1} & 49.7 \textcolor{gray}{±0.1} & 49.8 \textcolor{gray}{±0.2} \\
RawAtt & 52.7 \textcolor{gray}{±0.1} & 51.8 \textcolor{gray}{±0.1} & 58.9 \textcolor{gray}{±0.1} & 54.6 \textcolor{gray}{±0.2} \\
Attention Rollout & 52.3 \textcolor{gray}{±0.1} & 51.3 \textcolor{gray}{±0.1} & 56.0 \textcolor{gray}{±0.1} & 52.9 \textcolor{gray}{±0.2} \\
AliLRP & 53.6 \textcolor{gray}{±0.1} & 52.1 \textcolor{gray}{±0.1} & 58.8 \textcolor{gray}{±0.1} & 55.0 \textcolor{gray}{±0.2} \\
AttnLRP & 57.3 \textcolor{gray}{±0.1} & 54.5 \textcolor{gray}{±0.1} & 66.2 \textcolor{gray}{±0.1} & 59.9 \textcolor{gray}{±0.2} \\
DecompX & 54.8 \textcolor{gray}{±0.1} & 53.2 \textcolor{gray}{±0.1} & 62.2 \textcolor{gray}{±0.1} & 57.2 \textcolor{gray}{±0.2} \\
\bottomrule
\end{tabular}
\caption{Comparison of attribution methods and their LibraGrad-enhanced versions on the ImageNet-Hard ViT-B model.}
\label{tbl:SRG_m_2}
\end{table}
  \endgroup%

\end{table}

\FloatBarrier
\clearpage
\subsubsection{MURA ViT-B}
\label{per_model:MURA_ViT-B}

\begin{table}[ht]
  \fontsize{9.5pt}{8.5pt}\selectfont

  \setlength{\tabcolsep}{2pt}
  \centering
  \begingroup%
  \renewenvironment{table}[1][]%
    {\begin{center}}%
    {\end{center}}%
  \begin{table}[h]
\centering
\begin{tabular}{lcccc}
\toprule
Method & \multicolumn{2}{c}{MIF Deletion (GT)} & \multicolumn{2}{c}{MIF Deletion (Predicted)} \\
  & Accuracy & AOPC & Accuracy & AOPC \\
\cmidrule(r){1-1}
\cmidrule(lr){2-3}
\cmidrule(l){4-5}
Random & 25.4 \textcolor{gray}{±0.1} & 3.3 \textcolor{gray}{±0.1} & 15.1 \textcolor{gray}{±0.1} & 4.2 \textcolor{gray}{±0.1} \\
RawAtt & 33.4 \textcolor{gray}{±0.1} & 12.6 \textcolor{gray}{±0.3} & 24.8 \textcolor{gray}{±0.1} & 14.9 \textcolor{gray}{±0.2} \\
Attention Rollout & 29.9 \textcolor{gray}{±0.1} & 8.6 \textcolor{gray}{±0.2} & 21.5 \textcolor{gray}{±0.1} & 10.8 \textcolor{gray}{±0.2} \\
AliLRP & 28.4 \textcolor{gray}{±0.1} & 7.1 \textcolor{gray}{±0.2} & 19.2 \textcolor{gray}{±0.1} & 8.5 \textcolor{gray}{±0.2} \\
AttnLRP & 31.6 \textcolor{gray}{±0.1} & 10.4 \textcolor{gray}{±0.2} & 22.8 \textcolor{gray}{±0.1} & 12.1 \textcolor{gray}{±0.2} \\
DecompX & 30.7 \textcolor{gray}{±0.1} & 9.5 \textcolor{gray}{±0.2} & 21.6 \textcolor{gray}{±0.1} & 10.9 \textcolor{gray}{±0.2} \\
\bottomrule
\end{tabular}
\caption{Comparison of attribution methods and their LibraGrad-enhanced versions on the MURA ViT-B model.}
\label{tbl:MIF_m_2}
\end{table}
  \endgroup%

\end{table}

\begin{table}[ht]
  \fontsize{11pt}{8.5pt}\selectfont

  \setlength{\tabcolsep}{4pt}
  \centering

  \begingroup%
  \renewenvironment{table}[1][]%
    {\begin{center}}%
    {\end{center}}%
  \begin{table}[h]
\centering
\begin{tabular}{lcccc}
\toprule
Method & \multicolumn{2}{c}{LIF Deletion (GT)} & \multicolumn{2}{c}{LIF Deletion (Predicted)} \\
  & Accuracy & AOPC & Accuracy & AOPC \\
\cmidrule(r){1-1}
\cmidrule(lr){2-3}
\cmidrule(l){4-5}
Random & 75.2 \textcolor{gray}{±0.1} & 97.0 \textcolor{gray}{±0.1} & 85.8 \textcolor{gray}{±0.1} & 96.2 \textcolor{gray}{±0.1} \\
RawAtt & 75.9 \textcolor{gray}{±0.1} & 97.7 \textcolor{gray}{±0.1} & 85.7 \textcolor{gray}{±0.1} & 96.6 \textcolor{gray}{±0.1} \\
Attention Rollout & 75.0 \textcolor{gray}{±0.1} & 96.4 \textcolor{gray}{±0.1} & 84.1 \textcolor{gray}{±0.1} & 94.9 \textcolor{gray}{±0.1} \\
AliLRP & 77.8 \textcolor{gray}{±0.1} & 99.9 \textcolor{gray}{±0.1} & 87.0 \textcolor{gray}{±0.1} & 98.5 \textcolor{gray}{±0.1} \\
AttnLRP & 78.2 \textcolor{gray}{±0.1} & 100.4 \textcolor{gray}{±0.1} & 87.0 \textcolor{gray}{±0.1} & 98.7 \textcolor{gray}{±0.1} \\
DecompX & 78.7 \textcolor{gray}{±0.1} & 100.9 \textcolor{gray}{±0.1} & 87.7 \textcolor{gray}{±0.0} & 99.4 \textcolor{gray}{±0.1} \\
\bottomrule
\end{tabular}
\caption{Comparison of attribution methods and their LibraGrad-enhanced versions on the MURA ViT-B model.}
\label{tbl:LIF_m_2}
\end{table}
  \endgroup%

  \begingroup%
  \renewenvironment{table}[1][]%
    {\begin{center}}%
    {\end{center}}%
  \begin{table}[h]
\centering
\begin{tabular}{lcccc}
\toprule
Method & \multicolumn{2}{c}{SRG (GT)} & \multicolumn{2}{c}{SRG (Predicted)} \\
  & Accuracy & AOPC & Accuracy & AOPC \\
\cmidrule(r){1-1}
\cmidrule(lr){2-3}
\cmidrule(l){4-5}
Random & 50.3 \textcolor{gray}{±0.1} & 50.2 \textcolor{gray}{±0.1} & 50.4 \textcolor{gray}{±0.1} & 50.2 \textcolor{gray}{±0.1} \\
RawAtt & 54.7 \textcolor{gray}{±0.1} & 55.2 \textcolor{gray}{±0.2} & 55.3 \textcolor{gray}{±0.1} & 55.8 \textcolor{gray}{±0.2} \\
Attention Rollout & 52.5 \textcolor{gray}{±0.1} & 52.5 \textcolor{gray}{±0.2} & 52.8 \textcolor{gray}{±0.1} & 52.9 \textcolor{gray}{±0.2} \\
AliLRP & 53.1 \textcolor{gray}{±0.1} & 53.5 \textcolor{gray}{±0.2} & 53.1 \textcolor{gray}{±0.1} & 53.5 \textcolor{gray}{±0.1} \\
AttnLRP & 54.9 \textcolor{gray}{±0.1} & 55.4 \textcolor{gray}{±0.2} & 54.9 \textcolor{gray}{±0.1} & 55.4 \textcolor{gray}{±0.2} \\
DecompX & 54.7 \textcolor{gray}{±0.1} & 55.2 \textcolor{gray}{±0.2} & 54.7 \textcolor{gray}{±0.1} & 55.1 \textcolor{gray}{±0.1} \\
\bottomrule
\end{tabular}
\caption{Comparison of attribution methods and their LibraGrad-enhanced versions on the MURA ViT-B model.}
\label{tbl:SRG_m_2}
\end{table}
  \endgroup%

\end{table}

\FloatBarrier
\clearpage
\subsubsection{Oxford Pet ViT-B}
\label{per_model:Oxford_Pet_ViT-B}

\begin{table}[ht]
  \fontsize{9.5pt}{8.5pt}\selectfont

  \setlength{\tabcolsep}{2pt}
  \centering
  \begingroup%
  \renewenvironment{table}[1][]%
    {\begin{center}}%
    {\end{center}}%
  \begin{table}[h]
\centering
\begin{tabular}{lcccc}
\toprule
Method & \multicolumn{2}{c}{MIF Deletion (GT)} & \multicolumn{2}{c}{MIF Deletion (Predicted)} \\
  & Accuracy & AOPC & Accuracy & AOPC \\
\cmidrule(r){1-1}
\cmidrule(lr){2-3}
\cmidrule(l){4-5}
Random & 14.6 \textcolor{gray}{±0.1} & 4.1 \textcolor{gray}{±0.1} & 13.7 \textcolor{gray}{±0.1} & 4.3 \textcolor{gray}{±0.1} \\
RawAtt & 37.7 \textcolor{gray}{±0.1} & 29.1 \textcolor{gray}{±0.3} & 37.2 \textcolor{gray}{±0.1} & 29.6 \textcolor{gray}{±0.3} \\
Attention Rollout & 22.2 \textcolor{gray}{±0.1} & 12.2 \textcolor{gray}{±0.2} & 21.2 \textcolor{gray}{±0.1} & 12.4 \textcolor{gray}{±0.2} \\
AliLRP & 19.7 \textcolor{gray}{±0.1} & 9.7 \textcolor{gray}{±0.2} & 19.0 \textcolor{gray}{±0.1} & 10.0 \textcolor{gray}{±0.2} \\
AttnLRP & 30.9 \textcolor{gray}{±0.1} & 22.1 \textcolor{gray}{±0.2} & 30.3 \textcolor{gray}{±0.1} & 22.5 \textcolor{gray}{±0.2} \\
DecompX & 23.2 \textcolor{gray}{±0.1} & 13.8 \textcolor{gray}{±0.2} & 22.5 \textcolor{gray}{±0.1} & 14.1 \textcolor{gray}{±0.2} \\
\bottomrule
\end{tabular}
\caption{Comparison of attribution methods and their LibraGrad-enhanced versions on the Oxford Pet ViT-B model.}
\label{tbl:MIF_m_2}
\end{table}
  \endgroup%

\end{table}

\begin{table}[ht]
  \fontsize{11pt}{8.5pt}\selectfont

  \setlength{\tabcolsep}{4pt}
  \centering

  \begingroup%
  \renewenvironment{table}[1][]%
    {\begin{center}}%
    {\end{center}}%
  \begin{table}[h]
\centering
\begin{tabular}{lcccc}
\toprule
Method & \multicolumn{2}{c}{LIF Deletion (GT)} & \multicolumn{2}{c}{LIF Deletion (Predicted)} \\
  & Accuracy & AOPC & Accuracy & AOPC \\
\cmidrule(r){1-1}
\cmidrule(lr){2-3}
\cmidrule(l){4-5}
Random & 84.9 \textcolor{gray}{±0.1} & 95.6 \textcolor{gray}{±0.1} & 85.8 \textcolor{gray}{±0.1} & 95.4 \textcolor{gray}{±0.1} \\
RawAtt & 85.0 \textcolor{gray}{±0.1} & 96.3 \textcolor{gray}{±0.2} & 86.0 \textcolor{gray}{±0.1} & 96.1 \textcolor{gray}{±0.1} \\
Attention Rollout & 81.6 \textcolor{gray}{±0.1} & 92.0 \textcolor{gray}{±0.2} & 82.4 \textcolor{gray}{±0.1} & 91.7 \textcolor{gray}{±0.2} \\
AliLRP & 86.9 \textcolor{gray}{±0.1} & 98.3 \textcolor{gray}{±0.1} & 87.7 \textcolor{gray}{±0.0} & 98.0 \textcolor{gray}{±0.1} \\
AttnLRP & 88.0 \textcolor{gray}{±0.0} & 99.7 \textcolor{gray}{±0.1} & 88.7 \textcolor{gray}{±0.0} & 99.4 \textcolor{gray}{±0.1} \\
DecompX & 87.1 \textcolor{gray}{±0.1} & 98.4 \textcolor{gray}{±0.1} & 88.1 \textcolor{gray}{±0.0} & 98.3 \textcolor{gray}{±0.1} \\
\bottomrule
\end{tabular}
\caption{Comparison of attribution methods and their LibraGrad-enhanced versions on the Oxford Pet ViT-B model.}
\label{tbl:LIF_m_2}
\end{table}
  \endgroup%

  \begingroup%
  \renewenvironment{table}[1][]%
    {\begin{center}}%
    {\end{center}}%
  \begin{table}[h]
\centering
\begin{tabular}{lcccc}
\toprule
Method & \multicolumn{2}{c}{SRG (GT)} & \multicolumn{2}{c}{SRG (Predicted)} \\
  & Accuracy & AOPC & Accuracy & AOPC \\
\cmidrule(r){1-1}
\cmidrule(lr){2-3}
\cmidrule(l){4-5}
Random & 49.7 \textcolor{gray}{±0.1} & 49.9 \textcolor{gray}{±0.1} & 49.7 \textcolor{gray}{±0.1} & 49.8 \textcolor{gray}{±0.1} \\
RawAtt & 61.3 \textcolor{gray}{±0.1} & 62.7 \textcolor{gray}{±0.2} & 61.6 \textcolor{gray}{±0.1} & 62.8 \textcolor{gray}{±0.2} \\
Attention Rollout & 51.9 \textcolor{gray}{±0.1} & 52.1 \textcolor{gray}{±0.2} & 51.8 \textcolor{gray}{±0.1} & 52.0 \textcolor{gray}{±0.2} \\
AliLRP & 53.3 \textcolor{gray}{±0.1} & 54.0 \textcolor{gray}{±0.2} & 53.3 \textcolor{gray}{±0.1} & 54.0 \textcolor{gray}{±0.2} \\
AttnLRP & 59.4 \textcolor{gray}{±0.1} & 60.9 \textcolor{gray}{±0.2} & 59.5 \textcolor{gray}{±0.1} & 60.9 \textcolor{gray}{±0.2} \\
DecompX & 55.1 \textcolor{gray}{±0.1} & 56.1 \textcolor{gray}{±0.2} & 55.3 \textcolor{gray}{±0.1} & 56.2 \textcolor{gray}{±0.2} \\
\bottomrule
\end{tabular}
\caption{Comparison of attribution methods and their LibraGrad-enhanced versions on the Oxford Pet ViT-B model.}
\label{tbl:SRG_m_2}
\end{table}
  \endgroup%

\end{table}

\FloatBarrier
\clearpage
\subsubsection{ViT-L}
\label{per_model:ViT-L}

\begin{table}[ht]
  \fontsize{9.5pt}{8.5pt}\selectfont

  \setlength{\tabcolsep}{2pt}
  \centering
  \begingroup%
  \renewenvironment{table}[1][]%
    {\begin{center}}%
    {\end{center}}%
  \begin{table}[h]
\centering
\begin{tabular}{lccccc}
\toprule
Method & \multicolumn{2}{c}{MIF Deletion (GT)} & \multicolumn{2}{c}{MIF Deletion (Predicted)} & Segmentation \\
  & Accuracy & AOPC & Accuracy & AOPC & AP \\
\cmidrule(r){1-1}
\cmidrule(lr){2-3}
\cmidrule(lr){4-5}
\cmidrule(l){6-6}
Random & 36.9 \textcolor{gray}{±0.1} & 14.1 \textcolor{gray}{±0.2} & 29.5 \textcolor{gray}{±0.1} & 15.8 \textcolor{gray}{±0.2} & 42.0 \textcolor{gray}{±0.4} \\
RawAtt & 45.4 \textcolor{gray}{±0.1} & 22.9 \textcolor{gray}{±0.3} & 39.1 \textcolor{gray}{±0.1} & 25.3 \textcolor{gray}{±0.2} & 40.2 \textcolor{gray}{±0.4} \\
Attention Rollout & 39.0 \textcolor{gray}{±0.1} & 16.5 \textcolor{gray}{±0.3} & 31.4 \textcolor{gray}{±0.1} & 18.3 \textcolor{gray}{±0.3} & 39.9 \textcolor{gray}{±0.3} \\
AliLRP & 39.8 \textcolor{gray}{±0.1} & 17.2 \textcolor{gray}{±0.3} & 33.2 \textcolor{gray}{±0.1} & 19.2 \textcolor{gray}{±0.2} & 42.7 \textcolor{gray}{±0.4} \\
AttnLRP & 47.1 \textcolor{gray}{±0.1} & 24.8 \textcolor{gray}{±0.3} & 41.8 \textcolor{gray}{±0.1} & 27.6 \textcolor{gray}{±0.3} & 47.2 \textcolor{gray}{±0.3} \\
DecompX & 44.4 \textcolor{gray}{±0.1} & 22.6 \textcolor{gray}{±0.3} & 38.9 \textcolor{gray}{±0.1} & 25.3 \textcolor{gray}{±0.3} & 54.2 \textcolor{gray}{±0.3} \\
\bottomrule
\end{tabular}
\caption{Comparison of attribution methods and their LibraGrad-enhanced versions on the ViT-L model. We report faithfulness metrics using Most-Influential-First Deletion, MIF with ground-truth (GT) and predicted labels, including Accuracy and Area Over Perturbation Curve (AOPC) and Segmentation Average Precision (AP). The results demonstrate that composing existing methods with LibraGrad significantly enhances their performance across all metrics.}
\label{tbl:MIF_m_2}
\end{table}
  \endgroup%

\end{table}

\begin{table}[ht]
  \fontsize{11pt}{8.5pt}\selectfont

  \setlength{\tabcolsep}{4pt}
  \centering

  \begingroup%
  \renewenvironment{table}[1][]%
    {\begin{center}}%
    {\end{center}}%
  \begin{table}[h]
\centering
\begin{tabular}{lcccc}
\toprule
Method & \multicolumn{2}{c}{LIF Deletion (GT)} & \multicolumn{2}{c}{LIF Deletion (Predicted)} \\
  & Accuracy & AOPC & Accuracy & AOPC \\
\cmidrule(r){1-1}
\cmidrule(lr){2-3}
\cmidrule(l){4-5}
Random & 62.9 \textcolor{gray}{±0.1} & 85.4 \textcolor{gray}{±0.2} & 70.2 \textcolor{gray}{±0.1} & 83.7 \textcolor{gray}{±0.2} \\
RawAtt & 60.3 \textcolor{gray}{±0.1} & 83.3 \textcolor{gray}{±0.2} & 67.6 \textcolor{gray}{±0.1} & 81.5 \textcolor{gray}{±0.1} \\
Attention Rollout & 61.9 \textcolor{gray}{±0.1} & 84.1 \textcolor{gray}{±0.2} & 68.3 \textcolor{gray}{±0.1} & 81.9 \textcolor{gray}{±0.2} \\
AliLRP & 65.4 \textcolor{gray}{±0.1} & 87.7 \textcolor{gray}{±0.2} & 72.5 \textcolor{gray}{±0.1} & 85.9 \textcolor{gray}{±0.2} \\
AttnLRP & 70.3 \textcolor{gray}{±0.1} & 92.9 \textcolor{gray}{±0.2} & 77.6 \textcolor{gray}{±0.1} & 91.3 \textcolor{gray}{±0.2} \\
DecompX & 68.8 \textcolor{gray}{±0.1} & 91.0 \textcolor{gray}{±0.2} & 75.8 \textcolor{gray}{±0.1} & 89.3 \textcolor{gray}{±0.2} \\
Integrated Gradients & 71.1 \textcolor{gray}{±0.1} & 93.3 \textcolor{gray}{±0.2} & 73.5 \textcolor{gray}{±0.1} & 88.4 \textcolor{gray}{±0.2} \\
\midrule
\IxG{}  & 65.8 \textcolor{gray}{±0.1} & 88.4 \textcolor{gray}{±0.2} & 72.8 \textcolor{gray}{±0.1} & 86.7 \textcolor{gray}{±0.1} \\
\textbf{Libra \IxG{} } & 70.1 \textcolor{gray}{±0.1} (\textcolor{blue}{+6.6\%}) & 92.0 \textcolor{gray}{±0.2} (\textcolor{blue}{+4.0\%}) & 76.7 \textcolor{gray}{±0.1} (\textcolor{blue}{+5.4\%}) & 90.2 \textcolor{gray}{±0.2} (\textcolor{blue}{+4.0\%}) \\
\midrule
AttCAT & 71.8 \textcolor{gray}{±0.1} & 94.3 \textcolor{gray}{±0.2} & 77.5 \textcolor{gray}{±0.1} & 92.6 \textcolor{gray}{±0.2} \\
\textbf{Libra AttCAT} & \underline{76.3} \textcolor{gray}{±0.1} (\textcolor{blue}{+6.2\%}) & \underline{98.5} \textcolor{gray}{±0.2} (\textcolor{blue}{+4.5\%}) & \underline{82.2} \textcolor{gray}{±0.1} (\textcolor{blue}{+6.1\%}) & \underline{97.1} \textcolor{gray}{±0.2} (\textcolor{blue}{+4.8\%}) \\
\midrule
GenAtt & 70.0 \textcolor{gray}{±0.1} & 92.8 \textcolor{gray}{±0.2} & 78.2 \textcolor{gray}{±0.1} & 91.5 \textcolor{gray}{±0.2} \\
\textbf{Libra GenAtt} & 70.9 \textcolor{gray}{±0.1} (\textcolor{blue}{+1.3\%}) & 93.2 \textcolor{gray}{±0.2} (\textcolor{blue}{+0.5\%}) & 78.8 \textcolor{gray}{±0.1} (\textcolor{blue}{+0.7\%}) & 92.0 \textcolor{gray}{±0.2} (\textcolor{blue}{+0.5\%}) \\
\midrule
TokenTM & 68.9 \textcolor{gray}{±0.1} & 91.6 \textcolor{gray}{±0.2} & 77.3 \textcolor{gray}{±0.1} & 90.3 \textcolor{gray}{±0.2} \\
\textbf{Libra TokenTM} & 69.4 \textcolor{gray}{±0.1} (\textcolor{blue}{+0.8\%}) & 92.1 \textcolor{gray}{±0.2} (\textcolor{blue}{+0.5\%}) & 77.8 \textcolor{gray}{±0.1} (\textcolor{blue}{+0.7\%}) & 90.8 \textcolor{gray}{±0.2} (\textcolor{blue}{+0.6\%}) \\
\midrule
GradCAM+ & 70.5 \textcolor{gray}{±0.1} & 92.9 \textcolor{gray}{±0.2} & 76.8 \textcolor{gray}{±0.1} & 91.0 \textcolor{gray}{±0.2} \\
\textbf{Libra GradCAM+} & 72.6 \textcolor{gray}{±0.1} (\textcolor{blue}{+2.9\%}) & 94.4 \textcolor{gray}{±0.2} (\textcolor{blue}{+1.6\%}) & 79.1 \textcolor{gray}{±0.1} (\textcolor{blue}{+3.0\%}) & 92.7 \textcolor{gray}{±0.2} (\textcolor{blue}{+1.8\%}) \\
\midrule
HiResCAM & 53.6 \textcolor{gray}{±0.1} & 76.7 \textcolor{gray}{±0.2} & 59.3 \textcolor{gray}{±0.1} & 74.2 \textcolor{gray}{±0.3} \\
\textbf{Libra HiResCAM} & 67.4 \textcolor{gray}{±0.1} (\textcolor{blue}{+25.7\%}) & 90.0 \textcolor{gray}{±0.2} (\textcolor{blue}{+17.3\%}) & 73.8 \textcolor{gray}{±0.1} (\textcolor{blue}{+24.4\%}) & 88.0 \textcolor{gray}{±0.2} (\textcolor{blue}{+18.6\%}) \\
\midrule
XGradCAM+ & 69.5 \textcolor{gray}{±0.1} & 92.1 \textcolor{gray}{±0.2} & 75.7 \textcolor{gray}{±0.1} & 90.1 \textcolor{gray}{±0.2} \\
\textbf{Libra XGradCAM+} & 73.5 \textcolor{gray}{±0.1} (\textcolor{blue}{+5.7\%}) & 95.3 \textcolor{gray}{±0.2} (\textcolor{blue}{+3.5\%}) & 80.0 \textcolor{gray}{±0.1} (\textcolor{blue}{+5.6\%}) & 93.7 \textcolor{gray}{±0.2} (\textcolor{blue}{+3.9\%}) \\
\midrule
FullGrad+ & 71.5 \textcolor{gray}{±0.1} & 93.8 \textcolor{gray}{±0.2} & 76.8 \textcolor{gray}{±0.1} & 91.8 \textcolor{gray}{±0.2} \\
\textbf{Libra FullGrad+} & \textbf{76.8} \textcolor{gray}{±0.1} (\textcolor{blue}{+7.5\%}) & \textbf{98.9} \textcolor{gray}{±0.2} (\textcolor{blue}{+5.4\%}) & \textbf{82.6} \textcolor{gray}{±0.1} (\textcolor{blue}{+7.6\%}) & \textbf{97.4} \textcolor{gray}{±0.2} (\textcolor{blue}{+6.0\%}) \\
\bottomrule
\end{tabular}
\caption{Comparison of attribution methods and their LibraGrad-enhanced versions on the ViT-L model.}
\label{tbl:LIF_m_2}
\end{table}
  \endgroup%

  \begingroup%
  \renewenvironment{table}[1][]%
    {\begin{center}}%
    {\end{center}}%
  \begin{table}[h]
\centering
\begin{tabular}{lcccc}
\toprule
Method & \multicolumn{2}{c}{SRG (GT)} & \multicolumn{2}{c}{SRG (Predicted)} \\
  & Accuracy & AOPC & Accuracy & AOPC \\
\cmidrule(r){1-1}
\cmidrule(lr){2-3}
\cmidrule(l){4-5}
Random & 49.9 \textcolor{gray}{±0.1} & 49.7 \textcolor{gray}{±0.2} & 49.8 \textcolor{gray}{±0.1} & 49.8 \textcolor{gray}{±0.2} \\
RawAtt & 52.9 \textcolor{gray}{±0.1} & 53.1 \textcolor{gray}{±0.2} & 53.3 \textcolor{gray}{±0.1} & 53.4 \textcolor{gray}{±0.2} \\
Attention Rollout & 50.4 \textcolor{gray}{±0.1} & 50.3 \textcolor{gray}{±0.3} & 49.9 \textcolor{gray}{±0.1} & 50.1 \textcolor{gray}{±0.2} \\
AliLRP & 52.6 \textcolor{gray}{±0.1} & 52.4 \textcolor{gray}{±0.2} & 52.8 \textcolor{gray}{±0.1} & 52.5 \textcolor{gray}{±0.2} \\
AttnLRP & 58.7 \textcolor{gray}{±0.1} & 58.8 \textcolor{gray}{±0.3} & 59.7 \textcolor{gray}{±0.1} & 59.5 \textcolor{gray}{±0.2} \\
DecompX & 56.6 \textcolor{gray}{±0.1} & 56.8 \textcolor{gray}{±0.3} & 57.4 \textcolor{gray}{±0.1} & 57.3 \textcolor{gray}{±0.2} \\
Integrated Gradients & 58.7 \textcolor{gray}{±0.1} & 58.2 \textcolor{gray}{±0.3} & 54.7 \textcolor{gray}{±0.1} & 55.1 \textcolor{gray}{±0.2} \\
\midrule
\IxG{}  & 53.0 \textcolor{gray}{±0.1} & 53.0 \textcolor{gray}{±0.2} & 53.3 \textcolor{gray}{±0.1} & 53.2 \textcolor{gray}{±0.2} \\
\textbf{Libra \IxG{} } & 58.0 \textcolor{gray}{±0.1} (\textcolor{blue}{+9.5\%}) & 57.7 \textcolor{gray}{±0.3} (\textcolor{blue}{+8.9\%}) & 58.6 \textcolor{gray}{±0.1} (\textcolor{blue}{+9.9\%}) & 58.2 \textcolor{gray}{±0.2} (\textcolor{blue}{+9.4\%}) \\
\midrule
AttCAT & 60.2 \textcolor{gray}{±0.1} & 60.0 \textcolor{gray}{±0.2} & 61.2 \textcolor{gray}{±0.1} & 60.8 \textcolor{gray}{±0.2} \\
\textbf{Libra AttCAT} & \underline{70.5} \textcolor{gray}{±0.1} (\textcolor{blue}{+17.0\%}) & \underline{69.5} \textcolor{gray}{±0.3} (\textcolor{blue}{+15.8\%}) & \underline{71.8} \textcolor{gray}{±0.1} (\textcolor{blue}{+17.4\%}) & \underline{70.8} \textcolor{gray}{±0.2} (\textcolor{blue}{+16.4\%}) \\
\midrule
GenAtt & 63.2 \textcolor{gray}{±0.1} & 63.0 \textcolor{gray}{±0.2} & 65.0 \textcolor{gray}{±0.1} & 64.0 \textcolor{gray}{±0.2} \\
\textbf{Libra GenAtt} & 65.3 \textcolor{gray}{±0.1} (\textcolor{blue}{+3.3\%}) & 64.7 \textcolor{gray}{±0.3} (\textcolor{blue}{+2.7\%}) & 67.1 \textcolor{gray}{±0.1} (\textcolor{blue}{+3.2\%}) & 65.8 \textcolor{gray}{±0.3} (\textcolor{blue}{+2.8\%}) \\
\midrule
TokenTM & 61.9 \textcolor{gray}{±0.1} & 61.7 \textcolor{gray}{±0.3} & 63.6 \textcolor{gray}{±0.1} & 62.6 \textcolor{gray}{±0.2} \\
\textbf{Libra TokenTM} & 63.4 \textcolor{gray}{±0.1} (\textcolor{blue}{+2.4\%}) & 63.1 \textcolor{gray}{±0.3} (\textcolor{blue}{+2.3\%}) & 65.2 \textcolor{gray}{±0.1} (\textcolor{blue}{+2.4\%}) & 64.1 \textcolor{gray}{±0.3} (\textcolor{blue}{+2.4\%}) \\
\midrule
GradCAM+ & 62.0 \textcolor{gray}{±0.1} & 61.5 \textcolor{gray}{±0.3} & 62.7 \textcolor{gray}{±0.1} & 62.0 \textcolor{gray}{±0.2} \\
\textbf{Libra GradCAM+} & 66.7 \textcolor{gray}{±0.1} (\textcolor{blue}{+7.7\%}) & 65.5 \textcolor{gray}{±0.3} (\textcolor{blue}{+6.6\%}) & 67.8 \textcolor{gray}{±0.1} (\textcolor{blue}{+8.1\%}) & 66.4 \textcolor{gray}{±0.2} (\textcolor{blue}{+7.2\%}) \\
\midrule
HiResCAM & 43.2 \textcolor{gray}{±0.1} & 43.6 \textcolor{gray}{±0.2} & 42.5 \textcolor{gray}{±0.1} & 43.2 \textcolor{gray}{±0.2} \\
\textbf{Libra HiResCAM} & 60.7 \textcolor{gray}{±0.1} (\textcolor{blue}{+40.7\%}) & 60.1 \textcolor{gray}{±0.2} (\textcolor{blue}{+37.7\%}) & 61.4 \textcolor{gray}{±0.1} (\textcolor{blue}{+44.4\%}) & 60.6 \textcolor{gray}{±0.2} (\textcolor{blue}{+40.3\%}) \\
\midrule
XGradCAM+ & 60.2 \textcolor{gray}{±0.1} & 59.9 \textcolor{gray}{±0.3} & 60.8 \textcolor{gray}{±0.1} & 60.3 \textcolor{gray}{±0.2} \\
\textbf{Libra XGradCAM+} & 68.2 \textcolor{gray}{±0.1} (\textcolor{blue}{+13.3\%}) & 66.9 \textcolor{gray}{±0.3} (\textcolor{blue}{+11.8\%}) & 69.4 \textcolor{gray}{±0.1} (\textcolor{blue}{+14.1\%}) & 68.0 \textcolor{gray}{±0.3} (\textcolor{blue}{+12.6\%}) \\
\midrule
FullGrad+ & 60.3 \textcolor{gray}{±0.1} & 59.8 \textcolor{gray}{±0.2} & 60.9 \textcolor{gray}{±0.1} & 60.4 \textcolor{gray}{±0.2} \\
\textbf{Libra FullGrad+} & \textbf{71.2} \textcolor{gray}{±0.1} (\textcolor{blue}{+18.1\%}) & \textbf{70.0} \textcolor{gray}{±0.3} (\textcolor{blue}{+17.1\%}) & \textbf{72.5} \textcolor{gray}{±0.1} (\textcolor{blue}{+19.0\%}) & \textbf{71.3} \textcolor{gray}{±0.2} (\textcolor{blue}{+18.1\%}) \\
\bottomrule
\end{tabular}
\caption{Comparison of attribution methods and their LibraGrad-enhanced versions on the ViT-L model.}
\label{tbl:SRG_m_2}
\end{table}
  \endgroup%

\end{table}

\FloatBarrier
\clearpage
\subsubsection{EVA2-S}
\label{per_model:EVA2-S}

\begin{table}[ht]
  \fontsize{9.5pt}{8.5pt}\selectfont

  \setlength{\tabcolsep}{2pt}
  \centering
  \begingroup%
  \renewenvironment{table}[1][]%
    {\begin{center}}%
    {\end{center}}%
  \begin{table}[h]
\centering
\begin{tabular}{lccccc}
\toprule
Method & \multicolumn{2}{c}{MIF Deletion (GT)} & \multicolumn{2}{c}{MIF Deletion (Predicted)} & Segmentation \\
  & Accuracy & AOPC & Accuracy & AOPC & AP \\
\cmidrule(r){1-1}
\cmidrule(lr){2-3}
\cmidrule(lr){4-5}
\cmidrule(l){6-6}
Random & 29.9 \textcolor{gray}{±0.1} & 6.6 \textcolor{gray}{±0.2} & 21.2 \textcolor{gray}{±0.1} & 8.2 \textcolor{gray}{±0.2} & 37.7 \textcolor{gray}{±0.3} \\
RawAtt & 55.4 \textcolor{gray}{±0.1} & 30.3 \textcolor{gray}{±0.3} & 50.8 \textcolor{gray}{±0.1} & 33.9 \textcolor{gray}{±0.3} & 59.0 \textcolor{gray}{±0.3} \\
Attention Rollout & 47.0 \textcolor{gray}{±0.1} & 22.0 \textcolor{gray}{±0.4} & 41.1 \textcolor{gray}{±0.1} & 24.9 \textcolor{gray}{±0.3} & 45.3 \textcolor{gray}{±0.3} \\
AliLRP & 52.8 \textcolor{gray}{±0.1} & 27.7 \textcolor{gray}{±0.4} & 48.0 \textcolor{gray}{±0.1} & 31.3 \textcolor{gray}{±0.3} & 58.7 \textcolor{gray}{±0.3} \\
AttnLRP & 66.6 \textcolor{gray}{±0.1} & 39.6 \textcolor{gray}{±0.3} & 63.5 \textcolor{gray}{±0.1} & 44.2 \textcolor{gray}{±0.2} & 73.1 \textcolor{gray}{±0.2} \\
DecompX & 51.6 \textcolor{gray}{±0.1} & 27.0 \textcolor{gray}{±0.4} & 46.8 \textcolor{gray}{±0.1} & 30.7 \textcolor{gray}{±0.3} & 60.0 \textcolor{gray}{±0.3} \\
Integrated Gradients & 46.2 \textcolor{gray}{±0.1} & 21.0 \textcolor{gray}{±0.3} & 34.8 \textcolor{gray}{±0.1} & 19.3 \textcolor{gray}{±0.2} & 51.2 \textcolor{gray}{±0.3} \\
\midrule
\IxG{}  & 37.9 \textcolor{gray}{±0.1} & 14.1 \textcolor{gray}{±0.2} & 32.3 \textcolor{gray}{±0.1} & 17.0 \textcolor{gray}{±0.2} & 42.5 \textcolor{gray}{±0.3} \\
\textbf{Libra \IxG{} } & 67.0 \textcolor{gray}{±0.1} (\textcolor{blue}{+76.8\%}) & 39.6 \textcolor{gray}{±0.3} (\textcolor{blue}{+180.9\%}) & 64.1 \textcolor{gray}{±0.1} (\textcolor{blue}{+98.5\%}) & 44.4 \textcolor{gray}{±0.3} (\textcolor{blue}{+161.4\%}) & 72.1 \textcolor{gray}{±0.3} (\textcolor{blue}{+69.5\%}) \\
\midrule
AttCAT & 56.9 \textcolor{gray}{±0.1} & 30.9 \textcolor{gray}{±0.2} & 54.1 \textcolor{gray}{±0.1} & 35.3 \textcolor{gray}{±0.3} & 58.9 \textcolor{gray}{±0.3} \\
\textbf{Libra AttCAT} & \underline{72.1} \textcolor{gray}{±0.1} (\textcolor{blue}{+26.8\%}) & \underline{43.8} \textcolor{gray}{±0.3} (\textcolor{blue}{+41.8\%}) & \underline{69.5} \textcolor{gray}{±0.1} (\textcolor{blue}{+28.4\%}) & \underline{48.7} \textcolor{gray}{±0.2} (\textcolor{blue}{+38.1\%}) & 75.1 \textcolor{gray}{±0.3} (\textcolor{blue}{+27.6\%}) \\
\midrule
GenAtt & 46.3 \textcolor{gray}{±0.1} & 21.2 \textcolor{gray}{±0.2} & 40.7 \textcolor{gray}{±0.1} & 24.3 \textcolor{gray}{±0.2} & 42.3 \textcolor{gray}{±0.3} \\
\textbf{Libra GenAtt} & 47.7 \textcolor{gray}{±0.1} (\textcolor{blue}{+3.1\%}) & 22.5 \textcolor{gray}{±0.3} (\textcolor{blue}{+6.5\%}) & 42.1 \textcolor{gray}{±0.1} (\textcolor{blue}{+3.6\%}) & 25.6 \textcolor{gray}{±0.2} (\textcolor{blue}{+5.4\%}) & 44.3 \textcolor{gray}{±0.3} (\textcolor{blue}{+4.7\%}) \\
\midrule
TokenTM & 50.4 \textcolor{gray}{±0.1} & 25.1 \textcolor{gray}{±0.3} & 44.7 \textcolor{gray}{±0.1} & 28.3 \textcolor{gray}{±0.3} & 45.5 \textcolor{gray}{±0.3} \\
\textbf{Libra TokenTM} & 51.6 \textcolor{gray}{±0.1} (\textcolor{blue}{+2.4\%}) & 25.6 \textcolor{gray}{±0.3} (\textcolor{blue}{+1.9\%}) & 46.0 \textcolor{gray}{±0.1} (\textcolor{blue}{+2.8\%}) & 28.8 \textcolor{gray}{±0.3} (\textcolor{blue}{+1.6\%}) & 46.7 \textcolor{gray}{±0.3} (\textcolor{blue}{+2.7\%}) \\
\midrule
GradCAM+ & 50.6 \textcolor{gray}{±0.1} & 25.1 \textcolor{gray}{±0.3} & 47.1 \textcolor{gray}{±0.1} & 29.0 \textcolor{gray}{±0.3} & 49.3 \textcolor{gray}{±0.4} \\
\textbf{Libra GradCAM+} & 69.9 \textcolor{gray}{±0.1} (\textcolor{blue}{+38.0\%}) & 41.4 \textcolor{gray}{±0.3} (\textcolor{blue}{+65.2\%}) & 67.0 \textcolor{gray}{±0.1} (\textcolor{blue}{+42.1\%}) & 46.1 \textcolor{gray}{±0.2} (\textcolor{blue}{+58.9\%}) & \underline{79.8} \textcolor{gray}{±0.3} (\textcolor{blue}{+62.1\%}) \\
\midrule
HiResCAM & 63.1 \textcolor{gray}{±0.1} & 36.1 \textcolor{gray}{±0.2} & 59.1 \textcolor{gray}{±0.1} & 40.1 \textcolor{gray}{±0.2} & 73.2 \textcolor{gray}{±0.3} \\
\textbf{Libra HiResCAM} & 65.9 \textcolor{gray}{±0.1} (\textcolor{blue}{+4.4\%}) & 38.6 \textcolor{gray}{±0.3} (\textcolor{blue}{+6.8\%}) & 62.6 \textcolor{gray}{±0.1} (\textcolor{blue}{+6.0\%}) & 42.9 \textcolor{gray}{±0.2} (\textcolor{blue}{+7.1\%}) & 76.5 \textcolor{gray}{±0.3} (\textcolor{blue}{+4.5\%}) \\
\midrule
XGradCAM+ & 53.7 \textcolor{gray}{±0.1} & 27.9 \textcolor{gray}{±0.2} & 50.2 \textcolor{gray}{±0.1} & 31.9 \textcolor{gray}{±0.2} & 55.2 \textcolor{gray}{±0.4} \\
\textbf{Libra XGradCAM+} & 71.9 \textcolor{gray}{±0.1} (\textcolor{blue}{+34.1\%}) & 43.3 \textcolor{gray}{±0.3} (\textcolor{blue}{+54.9\%}) & 69.3 \textcolor{gray}{±0.1} (\textcolor{blue}{+38.0\%}) & 48.1 \textcolor{gray}{±0.2} (\textcolor{blue}{+50.8\%}) & \textbf{82.7} \textcolor{gray}{±0.3} (\textcolor{blue}{+49.9\%}) \\
\midrule
FullGrad+ & 50.9 \textcolor{gray}{±0.1} & 25.7 \textcolor{gray}{±0.2} & 48.0 \textcolor{gray}{±0.1} & 30.0 \textcolor{gray}{±0.2} & 51.5 \textcolor{gray}{±0.3} \\
\textbf{Libra FullGrad+} & \textbf{74.1} \textcolor{gray}{±0.1} (\textcolor{blue}{+45.5\%}) & \textbf{45.5} \textcolor{gray}{±0.3} (\textcolor{blue}{+77.0\%}) & \textbf{71.7} \textcolor{gray}{±0.1} (\textcolor{blue}{+49.4\%}) & \textbf{50.5} \textcolor{gray}{±0.2} (\textcolor{blue}{+68.5\%}) & 79.4 \textcolor{gray}{±0.3} (\textcolor{blue}{+54.2\%}) \\
\bottomrule
\end{tabular}
\caption{Comparison of attribution methods and their LibraGrad-enhanced versions on the EVA2-S model. We report faithfulness metrics using Most-Influential-First Deletion, MIF with ground-truth (GT) and predicted labels, including Accuracy and Area Over Perturbation Curve (AOPC) and Segmentation Average Precision (AP). The results demonstrate that composing existing methods with LibraGrad significantly enhances their performance across all metrics.}
\label{tbl:MIF_m_2}
\end{table}
  \endgroup%

\end{table}

\begin{table}[ht]
  \fontsize{11pt}{8.5pt}\selectfont

  \setlength{\tabcolsep}{4pt}
  \centering

  \begingroup%
  \renewenvironment{table}[1][]%
    {\begin{center}}%
    {\end{center}}%
  \begin{table}[h]
\centering
\begin{tabular}{lcccc}
\toprule
Method & \multicolumn{2}{c}{LIF Deletion (GT)} & \multicolumn{2}{c}{LIF Deletion (Predicted)} \\
  & Accuracy & AOPC & Accuracy & AOPC \\
\cmidrule(r){1-1}
\cmidrule(lr){2-3}
\cmidrule(l){4-5}
Random & 70.0 \textcolor{gray}{±0.1} & 93.5 \textcolor{gray}{±0.2} & 79.0 \textcolor{gray}{±0.1} & 92.3 \textcolor{gray}{±0.2} \\
RawAtt & 73.3 \textcolor{gray}{±0.1} & 96.9 \textcolor{gray}{±0.1} & 82.7 \textcolor{gray}{±0.1} & 95.7 \textcolor{gray}{±0.1} \\
Attention Rollout & 70.1 \textcolor{gray}{±0.1} & 93.6 \textcolor{gray}{±0.2} & 78.8 \textcolor{gray}{±0.1} & 91.9 \textcolor{gray}{±0.2} \\
AliLRP & 79.7 \textcolor{gray}{±0.1} & 102.6 \textcolor{gray}{±0.2} & 87.2 \textcolor{gray}{±0.1} & 100.9 \textcolor{gray}{±0.1} \\
AttnLRP & 78.8 \textcolor{gray}{±0.1} & 103.0 \textcolor{gray}{±0.1} & 87.5 \textcolor{gray}{±0.0} & 101.9 \textcolor{gray}{±0.1} \\
DecompX & 76.3 \textcolor{gray}{±0.1} & 100.4 \textcolor{gray}{±0.1} & 85.8 \textcolor{gray}{±0.1} & 99.5 \textcolor{gray}{±0.1} \\
Integrated Gradients & 82.0 \textcolor{gray}{±0.1} & 105.3 \textcolor{gray}{±0.2} & 83.5 \textcolor{gray}{±0.1} & 99.8 \textcolor{gray}{±0.2} \\
\midrule
\IxG{}  & 76.5 \textcolor{gray}{±0.1} & 100.0 \textcolor{gray}{±0.2} & 84.0 \textcolor{gray}{±0.1} & 98.9 \textcolor{gray}{±0.2} \\
\textbf{Libra \IxG{} } & 82.0 \textcolor{gray}{±0.1} (\textcolor{blue}{+7.2\%}) & 104.7 \textcolor{gray}{±0.1} (\textcolor{blue}{+4.7\%}) & \underline{88.3} \textcolor{gray}{±0.0} (\textcolor{blue}{+5.1\%}) & 102.5 \textcolor{gray}{±0.1} (\textcolor{blue}{+3.7\%}) \\
\midrule
AttCAT & \textbf{82.7} \textcolor{gray}{±0.1} & \textbf{107.2} \textcolor{gray}{±0.2} & 87.8 \textcolor{gray}{±0.0} & \textbf{105.3} \textcolor{gray}{±0.1} \\
\textbf{Libra AttCAT} & 82.2 \textcolor{gray}{±0.1} (\textcolor{coral}{-0.6\%}) & 105.0 \textcolor{gray}{±0.1} (\textcolor{coral}{-2.0\%}) & \underline{88.3} \textcolor{gray}{±0.0} (\textcolor{blue}{+0.5\%}) & 102.8 \textcolor{gray}{±0.1} (\textcolor{coral}{-2.4\%}) \\
\midrule
GenAtt & 71.9 \textcolor{gray}{±0.1} & 95.3 \textcolor{gray}{±0.2} & 80.7 \textcolor{gray}{±0.1} & 94.0 \textcolor{gray}{±0.2} \\
\textbf{Libra GenAtt} & 72.7 \textcolor{gray}{±0.1} (\textcolor{blue}{+1.1\%}) & 95.9 \textcolor{gray}{±0.2} (\textcolor{blue}{+0.6\%}) & 81.6 \textcolor{gray}{±0.1} (\textcolor{blue}{+1.1\%}) & 94.5 \textcolor{gray}{±0.2} (\textcolor{blue}{+0.6\%}) \\
\midrule
TokenTM & 73.3 \textcolor{gray}{±0.1} & 96.6 \textcolor{gray}{±0.2} & 82.1 \textcolor{gray}{±0.1} & 95.2 \textcolor{gray}{±0.1} \\
\textbf{Libra TokenTM} & 72.9 \textcolor{gray}{±0.1} (\textcolor{coral}{-0.6\%}) & 96.3 \textcolor{gray}{±0.2} (\textcolor{coral}{-0.3\%}) & 81.9 \textcolor{gray}{±0.1} (\textcolor{coral}{-0.2\%}) & 94.8 \textcolor{gray}{±0.2} (\textcolor{coral}{-0.4\%}) \\
\midrule
GradCAM+ & 77.3 \textcolor{gray}{±0.1} & 100.8 \textcolor{gray}{±0.3} & 82.8 \textcolor{gray}{±0.1} & 98.6 \textcolor{gray}{±0.2} \\
\textbf{Libra GradCAM+} & 80.1 \textcolor{gray}{±0.1} (\textcolor{blue}{+3.6\%}) & 102.6 \textcolor{gray}{±0.1} (\textcolor{blue}{+1.8\%}) & 86.4 \textcolor{gray}{±0.1} (\textcolor{blue}{+4.4\%}) & 100.3 \textcolor{gray}{±0.1} (\textcolor{blue}{+1.8\%}) \\
\midrule
HiResCAM & 79.3 \textcolor{gray}{±0.1} & 103.1 \textcolor{gray}{±0.2} & 86.1 \textcolor{gray}{±0.1} & 101.0 \textcolor{gray}{±0.1} \\
\textbf{Libra HiResCAM} & 79.4 \textcolor{gray}{±0.1} (\textcolor{blue}{+0.1\%}) & 102.4 \textcolor{gray}{±0.2} (\textcolor{coral}{-0.7\%}) & 86.3 \textcolor{gray}{±0.1} (\textcolor{blue}{+0.3\%}) & 100.5 \textcolor{gray}{±0.1} (\textcolor{coral}{-0.6\%}) \\
\midrule
XGradCAM+ & 78.3 \textcolor{gray}{±0.1} & 101.9 \textcolor{gray}{±0.3} & 83.8 \textcolor{gray}{±0.1} & 99.7 \textcolor{gray}{±0.2} \\
\textbf{Libra XGradCAM+} & 80.1 \textcolor{gray}{±0.1} (\textcolor{blue}{+2.3\%}) & 102.6 \textcolor{gray}{±0.1} (\textcolor{blue}{+0.7\%}) & 86.6 \textcolor{gray}{±0.1} (\textcolor{blue}{+3.4\%}) & 100.3 \textcolor{gray}{±0.1} (\textcolor{blue}{+0.6\%}) \\
\midrule
FullGrad+ & 82.1 \textcolor{gray}{±0.1} & \underline{106.6} \textcolor{gray}{±0.3} & 86.8 \textcolor{gray}{±0.1} & \underline{104.5} \textcolor{gray}{±0.2} \\
\textbf{Libra FullGrad+} & \underline{82.6} \textcolor{gray}{±0.1} (\textcolor{blue}{+0.7\%}) & 105.3 \textcolor{gray}{±0.2} (\textcolor{coral}{-1.2\%}) & \textbf{88.5} \textcolor{gray}{±0.0} (\textcolor{blue}{+1.9\%}) & 103.0 \textcolor{gray}{±0.1} (\textcolor{coral}{-1.4\%}) \\
\bottomrule
\end{tabular}
\caption{Comparison of attribution methods and their LibraGrad-enhanced versions on the EVA2-S model.}
\label{tbl:LIF_m_2}
\end{table}
  \endgroup%

  \begingroup%
  \renewenvironment{table}[1][]%
    {\begin{center}}%
    {\end{center}}%
  \begin{table}[h]
\centering
\begin{tabular}{lcccc}
\toprule
Method & \multicolumn{2}{c}{SRG (GT)} & \multicolumn{2}{c}{SRG (Predicted)} \\
  & Accuracy & AOPC & Accuracy & AOPC \\
\cmidrule(r){1-1}
\cmidrule(lr){2-3}
\cmidrule(l){4-5}
Random & 50.0 \textcolor{gray}{±0.1} & 50.0 \textcolor{gray}{±0.2} & 50.1 \textcolor{gray}{±0.1} & 50.2 \textcolor{gray}{±0.2} \\
RawAtt & 64.3 \textcolor{gray}{±0.1} & 63.6 \textcolor{gray}{±0.2} & 66.8 \textcolor{gray}{±0.1} & 64.8 \textcolor{gray}{±0.2} \\
Attention Rollout & 58.5 \textcolor{gray}{±0.1} & 57.8 \textcolor{gray}{±0.3} & 59.9 \textcolor{gray}{±0.1} & 58.4 \textcolor{gray}{±0.3} \\
AliLRP & 66.2 \textcolor{gray}{±0.1} & 65.1 \textcolor{gray}{±0.3} & 67.6 \textcolor{gray}{±0.1} & 66.1 \textcolor{gray}{±0.2} \\
AttnLRP & 72.7 \textcolor{gray}{±0.1} & 71.3 \textcolor{gray}{±0.2} & 75.5 \textcolor{gray}{±0.1} & 73.1 \textcolor{gray}{±0.2} \\
DecompX & 64.0 \textcolor{gray}{±0.1} & 63.7 \textcolor{gray}{±0.3} & 66.3 \textcolor{gray}{±0.1} & 65.1 \textcolor{gray}{±0.2} \\
Integrated Gradients & 64.1 \textcolor{gray}{±0.1} & 63.1 \textcolor{gray}{±0.2} & 59.2 \textcolor{gray}{±0.1} & 59.6 \textcolor{gray}{±0.2} \\
\midrule
\IxG{}  & 57.2 \textcolor{gray}{±0.1} & 57.1 \textcolor{gray}{±0.2} & 58.2 \textcolor{gray}{±0.1} & 57.9 \textcolor{gray}{±0.2} \\
\textbf{Libra \IxG{} } & 74.5 \textcolor{gray}{±0.1} (\textcolor{blue}{+30.3\%}) & 72.2 \textcolor{gray}{±0.3} (\textcolor{blue}{+26.5\%}) & 76.2 \textcolor{gray}{±0.1} (\textcolor{blue}{+31.0\%}) & 73.4 \textcolor{gray}{±0.2} (\textcolor{blue}{+26.8\%}) \\
\midrule
AttCAT & 69.8 \textcolor{gray}{±0.1} & 69.0 \textcolor{gray}{±0.2} & 71.0 \textcolor{gray}{±0.1} & 70.3 \textcolor{gray}{±0.2} \\
\textbf{Libra AttCAT} & \underline{77.2} \textcolor{gray}{±0.1} (\textcolor{blue}{+10.6\%}) & \underline{74.4} \textcolor{gray}{±0.2} (\textcolor{blue}{+7.8\%}) & \underline{78.9} \textcolor{gray}{±0.1} (\textcolor{blue}{+11.2\%}) & \underline{75.7} \textcolor{gray}{±0.2} (\textcolor{blue}{+7.8\%}) \\
\midrule
GenAtt & 59.1 \textcolor{gray}{±0.1} & 58.2 \textcolor{gray}{±0.2} & 60.7 \textcolor{gray}{±0.1} & 59.1 \textcolor{gray}{±0.2} \\
\textbf{Libra GenAtt} & 60.2 \textcolor{gray}{±0.1} (\textcolor{blue}{+1.9\%}) & 59.2 \textcolor{gray}{±0.2} (\textcolor{blue}{+1.7\%}) & 61.9 \textcolor{gray}{±0.1} (\textcolor{blue}{+1.9\%}) & 60.0 \textcolor{gray}{±0.2} (\textcolor{blue}{+1.6\%}) \\
\midrule
TokenTM & 61.8 \textcolor{gray}{±0.1} & 60.9 \textcolor{gray}{±0.2} & 63.4 \textcolor{gray}{±0.1} & 61.7 \textcolor{gray}{±0.2} \\
\textbf{Libra TokenTM} & 62.2 \textcolor{gray}{±0.1} (\textcolor{blue}{+0.6\%}) & 61.0 \textcolor{gray}{±0.3} (\textcolor{blue}{+0.1\%}) & 63.9 \textcolor{gray}{±0.1} (\textcolor{blue}{+0.8\%}) & 61.8 \textcolor{gray}{±0.2} (\textcolor{blue}{+0.1\%}) \\
\midrule
GradCAM+ & 64.0 \textcolor{gray}{±0.1} & 62.9 \textcolor{gray}{±0.3} & 65.0 \textcolor{gray}{±0.1} & 63.8 \textcolor{gray}{±0.2} \\
\textbf{Libra GradCAM+} & 75.0 \textcolor{gray}{±0.1} (\textcolor{blue}{+17.2\%}) & 72.0 \textcolor{gray}{±0.2} (\textcolor{blue}{+14.4\%}) & 76.7 \textcolor{gray}{±0.1} (\textcolor{blue}{+18.1\%}) & 73.2 \textcolor{gray}{±0.2} (\textcolor{blue}{+14.7\%}) \\
\midrule
HiResCAM & 71.2 \textcolor{gray}{±0.1} & 69.6 \textcolor{gray}{±0.2} & 72.6 \textcolor{gray}{±0.1} & 70.6 \textcolor{gray}{±0.2} \\
\textbf{Libra HiResCAM} & 72.6 \textcolor{gray}{±0.1} (\textcolor{blue}{+2.0\%}) & 70.5 \textcolor{gray}{±0.2} (\textcolor{blue}{+1.3\%}) & 74.5 \textcolor{gray}{±0.1} (\textcolor{blue}{+2.6\%}) & 71.7 \textcolor{gray}{±0.2} (\textcolor{blue}{+1.6\%}) \\
\midrule
XGradCAM+ & 66.0 \textcolor{gray}{±0.1} & 64.9 \textcolor{gray}{±0.3} & 67.0 \textcolor{gray}{±0.1} & 65.8 \textcolor{gray}{±0.2} \\
\textbf{Libra XGradCAM+} & 76.0 \textcolor{gray}{±0.1} (\textcolor{blue}{+15.2\%}) & 72.9 \textcolor{gray}{±0.2} (\textcolor{blue}{+12.3\%}) & 78.0 \textcolor{gray}{±0.1} (\textcolor{blue}{+16.4\%}) & 74.2 \textcolor{gray}{±0.2} (\textcolor{blue}{+12.8\%}) \\
\midrule
FullGrad+ & 66.5 \textcolor{gray}{±0.1} & 66.2 \textcolor{gray}{±0.3} & 67.4 \textcolor{gray}{±0.1} & 67.2 \textcolor{gray}{±0.2} \\
\textbf{Libra FullGrad+} & \textbf{78.3} \textcolor{gray}{±0.1} (\textcolor{blue}{+17.8\%}) & \textbf{75.4} \textcolor{gray}{±0.3} (\textcolor{blue}{+14.0\%}) & \textbf{80.1} \textcolor{gray}{±0.1} (\textcolor{blue}{+18.8\%}) & \textbf{76.8} \textcolor{gray}{±0.2} (\textcolor{blue}{+14.2\%}) \\
\bottomrule
\end{tabular}
\caption{Comparison of attribution methods and their LibraGrad-enhanced versions on the EVA2-S model.}
\label{tbl:SRG_m_2}
\end{table}
  \endgroup%

\end{table}

\FloatBarrier
\clearpage
\subsubsection{FlexiViT-L}
\label{per_model:FlexiViT-L}

\begin{table}[ht]
  \fontsize{9.5pt}{8.5pt}\selectfont

  \setlength{\tabcolsep}{2pt}
  \centering
  \begingroup%
  \renewenvironment{table}[1][]%
    {\begin{center}}%
    {\end{center}}%
  \begin{table}[h]
\centering
\begin{tabular}{lccccc}
\toprule
Method & \multicolumn{2}{c}{MIF Deletion (GT)} & \multicolumn{2}{c}{MIF Deletion (Predicted)} & Segmentation \\
  & Accuracy & AOPC & Accuracy & AOPC & AP \\
\cmidrule(r){1-1}
\cmidrule(lr){2-3}
\cmidrule(lr){4-5}
\cmidrule(l){6-6}
Random & 28.8 \textcolor{gray}{±0.1} & 5.2 \textcolor{gray}{±0.2} & 19.2 \textcolor{gray}{±0.1} & 6.4 \textcolor{gray}{±0.2} & 39.8 \textcolor{gray}{±0.4} \\
RawAtt & 47.3 \textcolor{gray}{±0.1} & 23.6 \textcolor{gray}{±0.3} & 41.7 \textcolor{gray}{±0.1} & 26.5 \textcolor{gray}{±0.3} & 49.8 \textcolor{gray}{±0.3} \\
Attention Rollout & 31.7 \textcolor{gray}{±0.1} & 8.2 \textcolor{gray}{±0.2} & 23.2 \textcolor{gray}{±0.1} & 9.7 \textcolor{gray}{±0.2} & 42.2 \textcolor{gray}{±0.3} \\
AliLRP & 32.5 \textcolor{gray}{±0.1} & 8.8 \textcolor{gray}{±0.2} & 24.9 \textcolor{gray}{±0.1} & 10.5 \textcolor{gray}{±0.2} & 49.6 \textcolor{gray}{±0.3} \\
AttnLRP & 30.3 \textcolor{gray}{±0.1} & 6.6 \textcolor{gray}{±0.2} & 21.8 \textcolor{gray}{±0.1} & 8.3 \textcolor{gray}{±0.2} & 43.4 \textcolor{gray}{±0.4} \\
DecompX & 42.0 \textcolor{gray}{±0.1} & 18.1 \textcolor{gray}{±0.2} & 35.5 \textcolor{gray}{±0.1} & 20.7 \textcolor{gray}{±0.2} & 59.2 \textcolor{gray}{±0.3} \\
Integrated Gradients & 31.4 \textcolor{gray}{±0.1} & 8.3 \textcolor{gray}{±0.2} & 22.3 \textcolor{gray}{±0.1} & 9.4 \textcolor{gray}{±0.2} & 41.3 \textcolor{gray}{±0.4} \\
\midrule
\IxG{}  & 28.5 \textcolor{gray}{±0.1} & 5.1 \textcolor{gray}{±0.2} & 19.9 \textcolor{gray}{±0.1} & 6.5 \textcolor{gray}{±0.2} & 41.4 \textcolor{gray}{±0.4} \\
\textbf{Libra \IxG{} } & 42.6 \textcolor{gray}{±0.1} (\textcolor{blue}{+49.6\%}) & 18.6 \textcolor{gray}{±0.2} (\textcolor{blue}{+263.5\%}) & 36.4 \textcolor{gray}{±0.1} (\textcolor{blue}{+82.8\%}) & 21.3 \textcolor{gray}{±0.2} (\textcolor{blue}{+227.8\%}) & 60.4 \textcolor{gray}{±0.3} (\textcolor{blue}{+45.9\%}) \\
\midrule
AttCAT & 45.3 \textcolor{gray}{±0.1} & 18.9 \textcolor{gray}{±0.3} & 41.9 \textcolor{gray}{±0.1} & 22.6 \textcolor{gray}{±0.3} & 45.1 \textcolor{gray}{±0.3} \\
\textbf{Libra AttCAT} & \underline{61.8} \textcolor{gray}{±0.1} (\textcolor{blue}{+36.5\%}) & \underline{35.5} \textcolor{gray}{±0.3} (\textcolor{blue}{+87.9\%}) & \underline{58.4} \textcolor{gray}{±0.1} (\textcolor{blue}{+39.3\%}) & \underline{39.6} \textcolor{gray}{±0.3} (\textcolor{blue}{+75.3\%}) & 74.4 \textcolor{gray}{±0.3} (\textcolor{blue}{+65.1\%}) \\
\midrule
GenAtt & 57.2 \textcolor{gray}{±0.1} & 31.4 \textcolor{gray}{±0.3} & 53.0 \textcolor{gray}{±0.1} & 35.1 \textcolor{gray}{±0.3} & 75.1 \textcolor{gray}{±0.2} \\
\textbf{Libra GenAtt} & 58.3 \textcolor{gray}{±0.1} (\textcolor{blue}{+1.9\%}) & 32.9 \textcolor{gray}{±0.3} (\textcolor{blue}{+4.8\%}) & 54.1 \textcolor{gray}{±0.1} (\textcolor{blue}{+2.0\%}) & 36.7 \textcolor{gray}{±0.3} (\textcolor{blue}{+4.5\%}) & \underline{79.4} \textcolor{gray}{±0.2} (\textcolor{blue}{+5.7\%}) \\
\midrule
TokenTM & 54.3 \textcolor{gray}{±0.1} & 29.3 \textcolor{gray}{±0.3} & 49.3 \textcolor{gray}{±0.1} & 32.7 \textcolor{gray}{±0.3} & 72.2 \textcolor{gray}{±0.2} \\
\textbf{Libra TokenTM} & 55.7 \textcolor{gray}{±0.1} (\textcolor{blue}{+2.5\%}) & 30.9 \textcolor{gray}{±0.3} (\textcolor{blue}{+5.4\%}) & 51.0 \textcolor{gray}{±0.1} (\textcolor{blue}{+3.4\%}) & 34.4 \textcolor{gray}{±0.3} (\textcolor{blue}{+5.4\%}) & 76.2 \textcolor{gray}{±0.2} (\textcolor{blue}{+5.5\%}) \\
\midrule
GradCAM+ & 35.8 \textcolor{gray}{±0.1} & 10.9 \textcolor{gray}{±0.2} & 28.7 \textcolor{gray}{±0.1} & 13.1 \textcolor{gray}{±0.2} & 40.5 \textcolor{gray}{±0.4} \\
\textbf{Libra GradCAM+} & 40.2 \textcolor{gray}{±0.1} (\textcolor{blue}{+12.6\%}) & 15.7 \textcolor{gray}{±0.2} (\textcolor{blue}{+44.3\%}) & 33.7 \textcolor{gray}{±0.1} (\textcolor{blue}{+17.3\%}) & 18.4 \textcolor{gray}{±0.3} (\textcolor{blue}{+40.6\%}) & 50.2 \textcolor{gray}{±0.4} (\textcolor{blue}{+23.7\%}) \\
\midrule
HiResCAM & 31.2 \textcolor{gray}{±0.1} & 7.2 \textcolor{gray}{±0.2} & 23.8 \textcolor{gray}{±0.1} & 9.0 \textcolor{gray}{±0.2} & 43.7 \textcolor{gray}{±0.3} \\
\textbf{Libra HiResCAM} & 60.1 \textcolor{gray}{±0.1} (\textcolor{blue}{+92.8\%}) & 34.2 \textcolor{gray}{±0.3} (\textcolor{blue}{+372.2\%}) & 56.5 \textcolor{gray}{±0.1} (\textcolor{blue}{+137.7\%}) & 38.1 \textcolor{gray}{±0.3} (\textcolor{blue}{+322.1\%}) & \textbf{81.6} \textcolor{gray}{±0.3} (\textcolor{blue}{+86.6\%}) \\
\midrule
XGradCAM+ & 33.4 \textcolor{gray}{±0.1} & 7.8 \textcolor{gray}{±0.2} & 26.6 \textcolor{gray}{±0.1} & 9.9 \textcolor{gray}{±0.2} & 38.5 \textcolor{gray}{±0.4} \\
\textbf{Libra XGradCAM+} & 49.7 \textcolor{gray}{±0.1} (\textcolor{blue}{+48.9\%}) & 24.1 \textcolor{gray}{±0.3} (\textcolor{blue}{+207.6\%}) & 44.3 \textcolor{gray}{±0.1} (\textcolor{blue}{+66.5\%}) & 27.2 \textcolor{gray}{±0.3} (\textcolor{blue}{+174.3\%}) & 63.3 \textcolor{gray}{±0.4} (\textcolor{blue}{+64.4\%}) \\
\midrule
FullGrad+ & 43.0 \textcolor{gray}{±0.1} & 17.5 \textcolor{gray}{±0.3} & 38.9 \textcolor{gray}{±0.1} & 20.8 \textcolor{gray}{±0.3} & 44.1 \textcolor{gray}{±0.3} \\
\textbf{Libra FullGrad+} & \textbf{62.4} \textcolor{gray}{±0.1} (\textcolor{blue}{+45.2\%}) & \textbf{35.8} \textcolor{gray}{±0.3} (\textcolor{blue}{+104.2\%}) & \textbf{59.1} \textcolor{gray}{±0.1} (\textcolor{blue}{+51.9\%}) & \textbf{39.8} \textcolor{gray}{±0.3} (\textcolor{blue}{+91.6\%}) & 75.1 \textcolor{gray}{±0.3} (\textcolor{blue}{+70.3\%}) \\
\bottomrule
\end{tabular}
\caption{Comparison of attribution methods and their LibraGrad-enhanced versions on the FlexiViT-L model. We report faithfulness metrics using Most-Influential-First Deletion, MIF with ground-truth (GT) and predicted labels, including Accuracy and Area Over Perturbation Curve (AOPC) and Segmentation Average Precision (AP). The results demonstrate that composing existing methods with LibraGrad significantly enhances their performance across all metrics.}
\label{tbl:MIF_m_2}
\end{table}
  \endgroup%

\end{table}

\begin{table}[ht]
  \fontsize{11pt}{8.5pt}\selectfont

  \setlength{\tabcolsep}{4pt}
  \centering

  \begingroup%
  \renewenvironment{table}[1][]%
    {\begin{center}}%
    {\end{center}}%
  \begin{table}[h]
\centering
\begin{tabular}{lcccc}
\toprule
Method & \multicolumn{2}{c}{LIF Deletion (GT)} & \multicolumn{2}{c}{LIF Deletion (Predicted)} \\
  & Accuracy & AOPC & Accuracy & AOPC \\
\cmidrule(r){1-1}
\cmidrule(lr){2-3}
\cmidrule(l){4-5}
Random & 70.7 \textcolor{gray}{±0.1} & 94.7 \textcolor{gray}{±0.2} & 80.7 \textcolor{gray}{±0.1} & 93.7 \textcolor{gray}{±0.1} \\
RawAtt & 72.8 \textcolor{gray}{±0.1} & 96.6 \textcolor{gray}{±0.1} & 82.6 \textcolor{gray}{±0.1} & 95.4 \textcolor{gray}{±0.1} \\
Attention Rollout & 65.0 \textcolor{gray}{±0.1} & 88.2 \textcolor{gray}{±0.2} & 72.7 \textcolor{gray}{±0.1} & 86.2 \textcolor{gray}{±0.2} \\
AliLRP & 75.8 \textcolor{gray}{±0.1} & 98.9 \textcolor{gray}{±0.1} & 84.7 \textcolor{gray}{±0.1} & 97.8 \textcolor{gray}{±0.1} \\
AttnLRP & 68.9 \textcolor{gray}{±0.1} & 92.2 \textcolor{gray}{±0.2} & 77.9 \textcolor{gray}{±0.1} & 90.9 \textcolor{gray}{±0.2} \\
DecompX & 76.7 \textcolor{gray}{±0.1} & 100.6 \textcolor{gray}{±0.1} & 86.2 \textcolor{gray}{±0.1} & 99.6 \textcolor{gray}{±0.1} \\
Integrated Gradients & 70.2 \textcolor{gray}{±0.1} & 93.8 \textcolor{gray}{±0.2} & 77.7 \textcolor{gray}{±0.1} & 91.9 \textcolor{gray}{±0.2} \\
\midrule
\IxG{}  & 69.6 \textcolor{gray}{±0.1} & 92.8 \textcolor{gray}{±0.2} & 78.3 \textcolor{gray}{±0.1} & 91.4 \textcolor{gray}{±0.2} \\
\textbf{Libra \IxG{} } & 78.2 \textcolor{gray}{±0.1} (\textcolor{blue}{+12.4\%}) & 101.6 \textcolor{gray}{±0.1} (\textcolor{blue}{+9.5\%}) & 86.9 \textcolor{gray}{±0.1} (\textcolor{blue}{+10.9\%}) & 100.4 \textcolor{gray}{±0.1} (\textcolor{blue}{+9.9\%}) \\
\midrule
AttCAT & \textbf{83.1} \textcolor{gray}{±0.1} & \textbf{106.1} \textcolor{gray}{±0.2} & \underline{88.3} \textcolor{gray}{±0.0} & \textbf{104.5} \textcolor{gray}{±0.2} \\
\textbf{Libra AttCAT} & 81.4 \textcolor{gray}{±0.1} (\textcolor{coral}{-2.0\%}) & 104.4 \textcolor{gray}{±0.1} (\textcolor{coral}{-1.5\%}) & \textbf{88.5} \textcolor{gray}{±0.0} (\textcolor{blue}{+0.2\%}) & 103.0 \textcolor{gray}{±0.1} (\textcolor{coral}{-1.4\%}) \\
\midrule
GenAtt & 77.3 \textcolor{gray}{±0.1} & 100.8 \textcolor{gray}{±0.1} & 87.0 \textcolor{gray}{±0.1} & 99.7 \textcolor{gray}{±0.1} \\
\textbf{Libra GenAtt} & 77.0 \textcolor{gray}{±0.1} (\textcolor{coral}{-0.4\%}) & 100.5 \textcolor{gray}{±0.1} (\textcolor{coral}{-0.3\%}) & 86.6 \textcolor{gray}{±0.1} (\textcolor{coral}{-0.4\%}) & 99.4 \textcolor{gray}{±0.1} (\textcolor{coral}{-0.3\%}) \\
\midrule
TokenTM & 76.0 \textcolor{gray}{±0.1} & 99.6 \textcolor{gray}{±0.1} & 86.0 \textcolor{gray}{±0.1} & 98.5 \textcolor{gray}{±0.1} \\
\textbf{Libra TokenTM} & 75.7 \textcolor{gray}{±0.1} (\textcolor{coral}{-0.4\%}) & 99.6 \textcolor{gray}{±0.1} (+0.0\%) & 85.8 \textcolor{gray}{±0.1} (\textcolor{coral}{-0.2\%}) & 98.6 \textcolor{gray}{±0.1} (\textcolor{blue}{+0.1\%}) \\
\midrule
GradCAM+ & 64.7 \textcolor{gray}{±0.1} & 87.5 \textcolor{gray}{±0.2} & 72.3 \textcolor{gray}{±0.1} & 85.7 \textcolor{gray}{±0.2} \\
\textbf{Libra GradCAM+} & 72.9 \textcolor{gray}{±0.1} (\textcolor{blue}{+12.8\%}) & 95.5 \textcolor{gray}{±0.1} (\textcolor{blue}{+9.1\%}) & 80.6 \textcolor{gray}{±0.1} (\textcolor{blue}{+11.5\%}) & 93.8 \textcolor{gray}{±0.1} (\textcolor{blue}{+9.5\%}) \\
\midrule
HiResCAM & 70.0 \textcolor{gray}{±0.1} & 92.6 \textcolor{gray}{±0.2} & 78.7 \textcolor{gray}{±0.1} & 91.2 \textcolor{gray}{±0.2} \\
\textbf{Libra HiResCAM} & 80.7 \textcolor{gray}{±0.1} (\textcolor{blue}{+15.3\%}) & 103.4 \textcolor{gray}{±0.1} (\textcolor{blue}{+11.6\%}) & 87.3 \textcolor{gray}{±0.0} (\textcolor{blue}{+11.0\%}) & 101.6 \textcolor{gray}{±0.1} (\textcolor{blue}{+11.3\%}) \\
\midrule
XGradCAM+ & 65.0 \textcolor{gray}{±0.1} & 86.6 \textcolor{gray}{±0.3} & 72.3 \textcolor{gray}{±0.1} & 84.7 \textcolor{gray}{±0.3} \\
\textbf{Libra XGradCAM+} & 77.5 \textcolor{gray}{±0.1} (\textcolor{blue}{+19.3\%}) & 100.4 \textcolor{gray}{±0.1} (\textcolor{blue}{+15.9\%}) & 85.3 \textcolor{gray}{±0.1} (\textcolor{blue}{+18.1\%}) & 99.0 \textcolor{gray}{±0.1} (\textcolor{blue}{+16.8\%}) \\
\midrule
FullGrad+ & 81.4 \textcolor{gray}{±0.1} & 104.3 \textcolor{gray}{±0.2} & 87.8 \textcolor{gray}{±0.0} & \underline{103.2} \textcolor{gray}{±0.2} \\
\textbf{Libra FullGrad+} & \underline{81.5} \textcolor{gray}{±0.1} (\textcolor{blue}{+0.1\%}) & \underline{104.5} \textcolor{gray}{±0.1} (\textcolor{blue}{+0.2\%}) & \underline{88.3} \textcolor{gray}{±0.0} (\textcolor{blue}{+0.6\%}) & 103.0 \textcolor{gray}{±0.1} (\textcolor{coral}{-0.2\%}) \\
\bottomrule
\end{tabular}
\caption{Comparison of attribution methods and their LibraGrad-enhanced versions on the FlexiViT-L model.}
\label{tbl:LIF_m_2}
\end{table}
  \endgroup%

  \begingroup%
  \renewenvironment{table}[1][]%
    {\begin{center}}%
    {\end{center}}%
  \begin{table}[h]
\centering
\begin{tabular}{lcccc}
\toprule
Method & \multicolumn{2}{c}{SRG (GT)} & \multicolumn{2}{c}{SRG (Predicted)} \\
  & Accuracy & AOPC & Accuracy & AOPC \\
\cmidrule(r){1-1}
\cmidrule(lr){2-3}
\cmidrule(l){4-5}
Random & 49.8 \textcolor{gray}{±0.1} & 50.0 \textcolor{gray}{±0.2} & 49.9 \textcolor{gray}{±0.1} & 50.0 \textcolor{gray}{±0.2} \\
RawAtt & 60.1 \textcolor{gray}{±0.1} & 60.1 \textcolor{gray}{±0.2} & 62.1 \textcolor{gray}{±0.1} & 60.9 \textcolor{gray}{±0.2} \\
Attention Rollout & 48.3 \textcolor{gray}{±0.1} & 48.2 \textcolor{gray}{±0.2} & 48.0 \textcolor{gray}{±0.1} & 48.0 \textcolor{gray}{±0.2} \\
AliLRP & 54.1 \textcolor{gray}{±0.1} & 53.9 \textcolor{gray}{±0.1} & 54.8 \textcolor{gray}{±0.1} & 54.2 \textcolor{gray}{±0.1} \\
AttnLRP & 49.6 \textcolor{gray}{±0.1} & 49.4 \textcolor{gray}{±0.2} & 49.9 \textcolor{gray}{±0.1} & 49.6 \textcolor{gray}{±0.2} \\
DecompX & 59.3 \textcolor{gray}{±0.1} & 59.3 \textcolor{gray}{±0.2} & 60.9 \textcolor{gray}{±0.1} & 60.1 \textcolor{gray}{±0.1} \\
Integrated Gradients & 50.8 \textcolor{gray}{±0.1} & 51.1 \textcolor{gray}{±0.2} & 50.0 \textcolor{gray}{±0.1} & 50.7 \textcolor{gray}{±0.2} \\
\midrule
\IxG{}  & 49.0 \textcolor{gray}{±0.1} & 49.0 \textcolor{gray}{±0.2} & 49.1 \textcolor{gray}{±0.1} & 48.9 \textcolor{gray}{±0.2} \\
\textbf{Libra \IxG{} } & 60.4 \textcolor{gray}{±0.1} (\textcolor{blue}{+23.2\%}) & 60.1 \textcolor{gray}{±0.2} (\textcolor{blue}{+22.8\%}) & 61.6 \textcolor{gray}{±0.1} (\textcolor{blue}{+25.5\%}) & 60.8 \textcolor{gray}{±0.1} (\textcolor{blue}{+24.3\%}) \\
\midrule
AttCAT & 64.2 \textcolor{gray}{±0.1} & 62.5 \textcolor{gray}{±0.3} & 65.1 \textcolor{gray}{±0.1} & 63.5 \textcolor{gray}{±0.3} \\
\textbf{Libra AttCAT} & \underline{71.6} \textcolor{gray}{±0.1} (\textcolor{blue}{+11.6\%}) & \underline{70.0} \textcolor{gray}{±0.2} (\textcolor{blue}{+12.0\%}) & \underline{73.4} \textcolor{gray}{±0.1} (\textcolor{blue}{+12.8\%}) & \underline{71.3} \textcolor{gray}{±0.2} (\textcolor{blue}{+12.2\%}) \\
\midrule
GenAtt & 67.3 \textcolor{gray}{±0.1} & 66.1 \textcolor{gray}{±0.2} & 70.0 \textcolor{gray}{±0.1} & 67.4 \textcolor{gray}{±0.2} \\
\textbf{Libra GenAtt} & 67.6 \textcolor{gray}{±0.1} (\textcolor{blue}{+0.5\%}) & 66.7 \textcolor{gray}{±0.2} (\textcolor{blue}{+0.9\%}) & 70.4 \textcolor{gray}{±0.1} (\textcolor{blue}{+0.5\%}) & 68.0 \textcolor{gray}{±0.2} (\textcolor{blue}{+1.0\%}) \\
\midrule
TokenTM & 65.2 \textcolor{gray}{±0.1} & 64.4 \textcolor{gray}{±0.2} & 67.6 \textcolor{gray}{±0.1} & 65.6 \textcolor{gray}{±0.2} \\
\textbf{Libra TokenTM} & 65.7 \textcolor{gray}{±0.1} (\textcolor{blue}{+0.8\%}) & 65.3 \textcolor{gray}{±0.2} (\textcolor{blue}{+1.3\%}) & 68.4 \textcolor{gray}{±0.1} (\textcolor{blue}{+1.1\%}) & 66.5 \textcolor{gray}{±0.2} (\textcolor{blue}{+1.4\%}) \\
\midrule
GradCAM+ & 50.2 \textcolor{gray}{±0.1} & 49.2 \textcolor{gray}{±0.2} & 50.5 \textcolor{gray}{±0.1} & 49.4 \textcolor{gray}{±0.2} \\
\textbf{Libra GradCAM+} & 56.6 \textcolor{gray}{±0.1} (\textcolor{blue}{+12.7\%}) & 55.6 \textcolor{gray}{±0.2} (\textcolor{blue}{+13.0\%}) & 57.2 \textcolor{gray}{±0.1} (\textcolor{blue}{+13.2\%}) & 56.1 \textcolor{gray}{±0.2} (\textcolor{blue}{+13.6\%}) \\
\midrule
HiResCAM & 50.6 \textcolor{gray}{±0.1} & 49.9 \textcolor{gray}{±0.2} & 51.2 \textcolor{gray}{±0.1} & 50.1 \textcolor{gray}{±0.2} \\
\textbf{Libra HiResCAM} & 70.4 \textcolor{gray}{±0.1} (\textcolor{blue}{+39.2\%}) & 68.8 \textcolor{gray}{±0.3} (\textcolor{blue}{+37.8\%}) & 71.9 \textcolor{gray}{±0.1} (\textcolor{blue}{+40.4\%}) & 69.8 \textcolor{gray}{±0.2} (\textcolor{blue}{+39.3\%}) \\
\midrule
XGradCAM+ & 49.2 \textcolor{gray}{±0.1} & 47.2 \textcolor{gray}{±0.3} & 49.4 \textcolor{gray}{±0.1} & 47.3 \textcolor{gray}{±0.3} \\
\textbf{Libra XGradCAM+} & 63.6 \textcolor{gray}{±0.1} (\textcolor{blue}{+29.3\%}) & 62.3 \textcolor{gray}{±0.2} (\textcolor{blue}{+31.8\%}) & 64.8 \textcolor{gray}{±0.1} (\textcolor{blue}{+31.1\%}) & 63.1 \textcolor{gray}{±0.2} (\textcolor{blue}{+33.3\%}) \\
\midrule
FullGrad+ & 62.2 \textcolor{gray}{±0.1} & 60.9 \textcolor{gray}{±0.3} & 63.3 \textcolor{gray}{±0.1} & 62.0 \textcolor{gray}{±0.2} \\
\textbf{Libra FullGrad+} & \textbf{71.9} \textcolor{gray}{±0.1} (\textcolor{blue}{+15.7\%}) & \textbf{70.1} \textcolor{gray}{±0.2} (\textcolor{blue}{+15.1\%}) & \textbf{73.7} \textcolor{gray}{±0.1} (\textcolor{blue}{+16.3\%}) & \textbf{71.4} \textcolor{gray}{±0.2} (\textcolor{blue}{+15.2\%}) \\
\bottomrule
\end{tabular}
\caption{Comparison of attribution methods and their LibraGrad-enhanced versions on the FlexiViT-L model.}
\label{tbl:SRG_m_2}
\end{table}
  \endgroup%

\end{table}

\FloatBarrier
\clearpage
\subsubsection{BEiT2-L}
\label{per_model:BEiT2-L}

\begin{table}[ht]
  \fontsize{9.5pt}{8.5pt}\selectfont

  \setlength{\tabcolsep}{2pt}
  \centering
  \begingroup%
  \renewenvironment{table}[1][]%
    {\begin{center}}%
    {\end{center}}%
  \begin{table}[h]
\centering
\begin{tabular}{lccccc}
\toprule
Method & \multicolumn{2}{c}{MIF Deletion (GT)} & \multicolumn{2}{c}{MIF Deletion (Predicted)} & Segmentation \\
  & Accuracy & AOPC & Accuracy & AOPC & AP \\
\cmidrule(r){1-1}
\cmidrule(lr){2-3}
\cmidrule(lr){4-5}
\cmidrule(l){6-6}
Random & 25.1 \textcolor{gray}{±0.1} & 5.6 \textcolor{gray}{±0.2} & 18.3 \textcolor{gray}{±0.1} & 6.8 \textcolor{gray}{±0.1} & 39.8 \textcolor{gray}{±0.4} \\
RawAtt & 34.2 \textcolor{gray}{±0.1} & 15.3 \textcolor{gray}{±0.2} & 29.5 \textcolor{gray}{±0.1} & 17.5 \textcolor{gray}{±0.2} & 47.6 \textcolor{gray}{±0.3} \\
Attention Rollout & 26.0 \textcolor{gray}{±0.1} & 7.2 \textcolor{gray}{±0.1} & 19.7 \textcolor{gray}{±0.1} & 8.6 \textcolor{gray}{±0.1} & 42.2 \textcolor{gray}{±0.3} \\
AliLRP & 31.9 \textcolor{gray}{±0.1} & 12.4 \textcolor{gray}{±0.2} & 26.2 \textcolor{gray}{±0.1} & 13.9 \textcolor{gray}{±0.2} & 43.9 \textcolor{gray}{±0.3} \\
AttnLRP & 42.1 \textcolor{gray}{±0.1} & 22.6 \textcolor{gray}{±0.3} & 37.7 \textcolor{gray}{±0.1} & 25.0 \textcolor{gray}{±0.2} & 66.0 \textcolor{gray}{±0.3} \\
DecompX & 36.5 \textcolor{gray}{±0.1} & 17.3 \textcolor{gray}{±0.3} & 31.7 \textcolor{gray}{±0.1} & 19.4 \textcolor{gray}{±0.2} & 55.6 \textcolor{gray}{±0.3} \\
Integrated Gradients & 31.7 \textcolor{gray}{±0.1} & 12.5 \textcolor{gray}{±0.2} & 23.2 \textcolor{gray}{±0.1} & 11.9 \textcolor{gray}{±0.1} & 46.7 \textcolor{gray}{±0.3} \\
\midrule
\IxG{}  & 28.2 \textcolor{gray}{±0.1} & 9.0 \textcolor{gray}{±0.1} & 21.8 \textcolor{gray}{±0.1} & 10.3 \textcolor{gray}{±0.1} & 39.6 \textcolor{gray}{±0.4} \\
\textbf{Libra \IxG{} } & 37.7 \textcolor{gray}{±0.1} (\textcolor{blue}{+33.6\%}) & 18.0 \textcolor{gray}{±0.2} (\textcolor{blue}{+100.2\%}) & 33.0 \textcolor{gray}{±0.1} (\textcolor{blue}{+51.4\%}) & 20.2 \textcolor{gray}{±0.2} (\textcolor{blue}{+96.6\%}) & 54.8 \textcolor{gray}{±0.3} (\textcolor{blue}{+38.4\%}) \\
\midrule
AttCAT & 38.4 \textcolor{gray}{±0.1} & 18.9 \textcolor{gray}{±0.2} & 33.9 \textcolor{gray}{±0.1} & 21.0 \textcolor{gray}{±0.2} & 52.2 \textcolor{gray}{±0.3} \\
\textbf{Libra AttCAT} & \underline{52.5} \textcolor{gray}{±0.1} (\textcolor{blue}{+36.6\%}) & \underline{31.6} \textcolor{gray}{±0.3} (\textcolor{blue}{+66.8\%}) & \underline{48.9} \textcolor{gray}{±0.1} (\textcolor{blue}{+44.4\%}) & \underline{34.6} \textcolor{gray}{±0.2} (\textcolor{blue}{+64.9\%}) & 65.5 \textcolor{gray}{±0.3} (\textcolor{blue}{+25.4\%}) \\
\midrule
GenAtt & 35.6 \textcolor{gray}{±0.1} & 17.0 \textcolor{gray}{±0.3} & 30.8 \textcolor{gray}{±0.1} & 19.2 \textcolor{gray}{±0.2} & 47.9 \textcolor{gray}{±0.3} \\
\textbf{Libra GenAtt} & 37.6 \textcolor{gray}{±0.1} (\textcolor{blue}{+5.6\%}) & 18.4 \textcolor{gray}{±0.3} (\textcolor{blue}{+8.4\%}) & 32.9 \textcolor{gray}{±0.1} (\textcolor{blue}{+6.8\%}) & 20.7 \textcolor{gray}{±0.3} (\textcolor{blue}{+7.9\%}) & 48.8 \textcolor{gray}{±0.3} (\textcolor{blue}{+1.8\%}) \\
\midrule
TokenTM & 43.9 \textcolor{gray}{±0.1} & 24.3 \textcolor{gray}{±0.3} & 39.6 \textcolor{gray}{±0.1} & 26.8 \textcolor{gray}{±0.3} & 56.0 \textcolor{gray}{±0.3} \\
\textbf{Libra TokenTM} & 42.6 \textcolor{gray}{±0.1} (\textcolor{coral}{-2.8\%}) & 23.1 \textcolor{gray}{±0.3} (\textcolor{coral}{-4.8\%}) & 38.3 \textcolor{gray}{±0.1} (\textcolor{coral}{-3.4\%}) & 25.5 \textcolor{gray}{±0.3} (\textcolor{coral}{-5.0\%}) & 54.2 \textcolor{gray}{±0.3} (\textcolor{coral}{-3.3\%}) \\
\midrule
GradCAM+ & 38.4 \textcolor{gray}{±0.1} & 18.2 \textcolor{gray}{±0.2} & 33.4 \textcolor{gray}{±0.1} & 20.1 \textcolor{gray}{±0.2} & 53.5 \textcolor{gray}{±0.4} \\
\textbf{Libra GradCAM+} & 42.3 \textcolor{gray}{±0.1} (\textcolor{blue}{+10.2\%}) & 22.0 \textcolor{gray}{±0.2} (\textcolor{blue}{+21.0\%}) & 37.5 \textcolor{gray}{±0.1} (\textcolor{blue}{+12.4\%}) & 24.3 \textcolor{gray}{±0.2} (\textcolor{blue}{+20.5\%}) & \underline{69.4} \textcolor{gray}{±0.4} (\textcolor{blue}{+29.9\%}) \\
\midrule
HiResCAM & 40.3 \textcolor{gray}{±0.1} & 20.1 \textcolor{gray}{±0.2} & 35.8 \textcolor{gray}{±0.1} & 22.3 \textcolor{gray}{±0.2} & 60.8 \textcolor{gray}{±0.3} \\
\textbf{Libra HiResCAM} & 41.5 \textcolor{gray}{±0.1} (\textcolor{blue}{+2.8\%}) & 21.2 \textcolor{gray}{±0.2} (\textcolor{blue}{+5.7\%}) & 37.2 \textcolor{gray}{±0.1} (\textcolor{blue}{+4.1\%}) & 23.6 \textcolor{gray}{±0.2} (\textcolor{blue}{+5.9\%}) & 69.0 \textcolor{gray}{±0.3} (\textcolor{blue}{+13.4\%}) \\
\midrule
XGradCAM+ & 35.6 \textcolor{gray}{±0.1} & 16.0 \textcolor{gray}{±0.2} & 30.6 \textcolor{gray}{±0.1} & 17.9 \textcolor{gray}{±0.2} & 49.0 \textcolor{gray}{±0.4} \\
\textbf{Libra XGradCAM+} & 49.5 \textcolor{gray}{±0.1} (\textcolor{blue}{+39.0\%}) & 28.6 \textcolor{gray}{±0.3} (\textcolor{blue}{+78.5\%}) & 45.6 \textcolor{gray}{±0.1} (\textcolor{blue}{+49.2\%}) & 31.4 \textcolor{gray}{±0.3} (\textcolor{blue}{+75.3\%}) & \textbf{71.4} \textcolor{gray}{±0.3} (\textcolor{blue}{+45.7\%}) \\
\midrule
FullGrad+ & 34.4 \textcolor{gray}{±0.1} & 14.9 \textcolor{gray}{±0.2} & 29.0 \textcolor{gray}{±0.1} & 16.6 \textcolor{gray}{±0.2} & 47.4 \textcolor{gray}{±0.3} \\
\textbf{Libra FullGrad+} & \textbf{53.4} \textcolor{gray}{±0.1} (\textcolor{blue}{+55.5\%}) & \textbf{32.4} \textcolor{gray}{±0.3} (\textcolor{blue}{+118.0\%}) & \textbf{50.0} \textcolor{gray}{±0.1} (\textcolor{blue}{+72.3\%}) & \textbf{35.5} \textcolor{gray}{±0.3} (\textcolor{blue}{+113.7\%}) & 67.9 \textcolor{gray}{±0.3} (\textcolor{blue}{+43.2\%}) \\
\bottomrule
\end{tabular}
\caption{Comparison of attribution methods and their LibraGrad-enhanced versions on the BEiT2-L model. We report faithfulness metrics using Most-Influential-First Deletion, MIF with ground-truth (GT) and predicted labels, including Accuracy and Area Over Perturbation Curve (AOPC) and Segmentation Average Precision (AP). The results demonstrate that composing existing methods with LibraGrad significantly enhances their performance across all metrics.}
\label{tbl:MIF_m_2}
\end{table}
  \endgroup%

\end{table}

\begin{table}[ht]
  \fontsize{11pt}{8.5pt}\selectfont

  \setlength{\tabcolsep}{4pt}
  \centering

  \begingroup%
  \renewenvironment{table}[1][]%
    {\begin{center}}%
    {\end{center}}%
  \begin{table}[h]
\centering
\begin{tabular}{lcccc}
\toprule
Method & \multicolumn{2}{c}{LIF Deletion (GT)} & \multicolumn{2}{c}{LIF Deletion (Predicted)} \\
  & Accuracy & AOPC & Accuracy & AOPC \\
\cmidrule(r){1-1}
\cmidrule(lr){2-3}
\cmidrule(l){4-5}
Random & 74.6 \textcolor{gray}{±0.1} & 94.1 \textcolor{gray}{±0.1} & 81.7 \textcolor{gray}{±0.1} & 93.2 \textcolor{gray}{±0.1} \\
RawAtt & 76.6 \textcolor{gray}{±0.1} & 95.8 \textcolor{gray}{±0.1} & 83.7 \textcolor{gray}{±0.1} & 94.9 \textcolor{gray}{±0.1} \\
Attention Rollout & 69.9 \textcolor{gray}{±0.1} & 89.1 \textcolor{gray}{±0.2} & 75.6 \textcolor{gray}{±0.1} & 87.4 \textcolor{gray}{±0.2} \\
AliLRP & 78.0 \textcolor{gray}{±0.1} & 96.7 \textcolor{gray}{±0.1} & 84.5 \textcolor{gray}{±0.1} & 95.5 \textcolor{gray}{±0.1} \\
AttnLRP & 78.4 \textcolor{gray}{±0.1} & 97.8 \textcolor{gray}{±0.1} & 85.7 \textcolor{gray}{±0.1} & 96.8 \textcolor{gray}{±0.1} \\
DecompX & 77.6 \textcolor{gray}{±0.1} & 97.2 \textcolor{gray}{±0.1} & 84.9 \textcolor{gray}{±0.1} & 96.3 \textcolor{gray}{±0.1} \\
Integrated Gradients & 79.4 \textcolor{gray}{±0.1} & 98.8 \textcolor{gray}{±0.1} & 84.2 \textcolor{gray}{±0.1} & 96.5 \textcolor{gray}{±0.2} \\
\midrule
\IxG{}  & 75.5 \textcolor{gray}{±0.1} & 94.5 \textcolor{gray}{±0.1} & 82.0 \textcolor{gray}{±0.1} & 93.5 \textcolor{gray}{±0.1} \\
\textbf{Libra \IxG{} } & 79.2 \textcolor{gray}{±0.1} (\textcolor{blue}{+4.9\%}) & 98.1 \textcolor{gray}{±0.1} (\textcolor{blue}{+3.9\%}) & 85.7 \textcolor{gray}{±0.1} (\textcolor{blue}{+4.5\%}) & 96.9 \textcolor{gray}{±0.1} (\textcolor{blue}{+3.6\%}) \\
\midrule
AttCAT & \textbf{81.8} \textcolor{gray}{±0.1} & \textbf{101.1} \textcolor{gray}{±0.1} & \textbf{87.5} \textcolor{gray}{±0.0} & \textbf{100.0} \textcolor{gray}{±0.1} \\
\textbf{Libra AttCAT} & \underline{80.8} \textcolor{gray}{±0.1} (\textcolor{coral}{-1.2\%}) & 99.2 \textcolor{gray}{±0.1} (\textcolor{coral}{-1.8\%}) & \underline{87.0} \textcolor{gray}{±0.1} (\textcolor{coral}{-0.6\%}) & 97.9 \textcolor{gray}{±0.1} (\textcolor{coral}{-2.0\%}) \\
\midrule
GenAtt & 75.6 \textcolor{gray}{±0.1} & 95.2 \textcolor{gray}{±0.1} & 83.2 \textcolor{gray}{±0.1} & 94.4 \textcolor{gray}{±0.1} \\
\textbf{Libra GenAtt} & 75.5 \textcolor{gray}{±0.1} (\textcolor{coral}{-0.2\%}) & 95.2 \textcolor{gray}{±0.2} (+0.0\%) & 83.2 \textcolor{gray}{±0.1} (+0.0\%) & 94.3 \textcolor{gray}{±0.2} (\textcolor{coral}{-0.1\%}) \\
\midrule
TokenTM & 76.8 \textcolor{gray}{±0.1} & 96.2 \textcolor{gray}{±0.1} & 84.6 \textcolor{gray}{±0.1} & 95.5 \textcolor{gray}{±0.1} \\
\textbf{Libra TokenTM} & 76.2 \textcolor{gray}{±0.1} (\textcolor{coral}{-0.8\%}) & 95.5 \textcolor{gray}{±0.1} (\textcolor{coral}{-0.8\%}) & 83.8 \textcolor{gray}{±0.1} (\textcolor{coral}{-1.0\%}) & 94.6 \textcolor{gray}{±0.1} (\textcolor{coral}{-0.9\%}) \\
\midrule
GradCAM+ & 79.2 \textcolor{gray}{±0.1} & 98.5 \textcolor{gray}{±0.2} & 85.1 \textcolor{gray}{±0.1} & 97.1 \textcolor{gray}{±0.1} \\
\textbf{Libra GradCAM+} & 78.4 \textcolor{gray}{±0.1} (\textcolor{coral}{-1.1\%}) & 97.1 \textcolor{gray}{±0.1} (\textcolor{coral}{-1.3\%}) & 84.2 \textcolor{gray}{±0.1} (\textcolor{coral}{-0.9\%}) & 95.6 \textcolor{gray}{±0.1} (\textcolor{coral}{-1.5\%}) \\
\midrule
HiResCAM & 79.4 \textcolor{gray}{±0.1} & 98.3 \textcolor{gray}{±0.1} & 85.5 \textcolor{gray}{±0.1} & 97.0 \textcolor{gray}{±0.1} \\
\textbf{Libra HiResCAM} & 80.0 \textcolor{gray}{±0.1} (\textcolor{blue}{+0.8\%}) & 98.4 \textcolor{gray}{±0.1} (\textcolor{blue}{+0.2\%}) & 86.0 \textcolor{gray}{±0.1} (\textcolor{blue}{+0.6\%}) & 97.1 \textcolor{gray}{±0.1} (\textcolor{blue}{+0.1\%}) \\
\midrule
XGradCAM+ & 78.9 \textcolor{gray}{±0.1} & 97.9 \textcolor{gray}{±0.2} & 84.3 \textcolor{gray}{±0.1} & 96.4 \textcolor{gray}{±0.1} \\
\textbf{Libra XGradCAM+} & 79.5 \textcolor{gray}{±0.1} (\textcolor{blue}{+0.7\%}) & 98.0 \textcolor{gray}{±0.1} (\textcolor{blue}{+0.1\%}) & 85.6 \textcolor{gray}{±0.1} (\textcolor{blue}{+1.6\%}) & 96.6 \textcolor{gray}{±0.1} (\textcolor{blue}{+0.2\%}) \\
\midrule
FullGrad+ & 79.9 \textcolor{gray}{±0.1} & 98.9 \textcolor{gray}{±0.1} & 86.0 \textcolor{gray}{±0.1} & \underline{98.0} \textcolor{gray}{±0.1} \\
\textbf{Libra FullGrad+} & \underline{80.8} \textcolor{gray}{±0.1} (\textcolor{blue}{+1.2\%}) & \underline{99.3} \textcolor{gray}{±0.1} (\textcolor{blue}{+0.3\%}) & 86.9 \textcolor{gray}{±0.1} (\textcolor{blue}{+1.0\%}) & \underline{98.0} \textcolor{gray}{±0.1} (+0.0\%) \\
\bottomrule
\end{tabular}
\caption{Comparison of attribution methods and their LibraGrad-enhanced versions on the BEiT2-L model.}
\label{tbl:LIF_m_2}
\end{table}
  \endgroup%

  \begingroup%
  \renewenvironment{table}[1][]%
    {\begin{center}}%
    {\end{center}}%
  \begin{table}[h]
\centering
\begin{tabular}{lcccc}
\toprule
Method & \multicolumn{2}{c}{SRG (GT)} & \multicolumn{2}{c}{SRG (Predicted)} \\
  & Accuracy & AOPC & Accuracy & AOPC \\
\cmidrule(r){1-1}
\cmidrule(lr){2-3}
\cmidrule(l){4-5}
Random & 49.8 \textcolor{gray}{±0.1} & 49.8 \textcolor{gray}{±0.1} & 50.0 \textcolor{gray}{±0.1} & 50.0 \textcolor{gray}{±0.1} \\
RawAtt & 55.4 \textcolor{gray}{±0.1} & 55.6 \textcolor{gray}{±0.2} & 56.6 \textcolor{gray}{±0.1} & 56.2 \textcolor{gray}{±0.2} \\
Attention Rollout & 47.9 \textcolor{gray}{±0.1} & 48.1 \textcolor{gray}{±0.2} & 47.7 \textcolor{gray}{±0.1} & 48.0 \textcolor{gray}{±0.2} \\
AliLRP & 55.0 \textcolor{gray}{±0.1} & 54.6 \textcolor{gray}{±0.2} & 55.3 \textcolor{gray}{±0.1} & 54.7 \textcolor{gray}{±0.1} \\
AttnLRP & 60.3 \textcolor{gray}{±0.1} & 60.2 \textcolor{gray}{±0.2} & 61.7 \textcolor{gray}{±0.1} & 60.9 \textcolor{gray}{±0.2} \\
DecompX & 57.0 \textcolor{gray}{±0.1} & 57.3 \textcolor{gray}{±0.2} & 58.3 \textcolor{gray}{±0.1} & 57.8 \textcolor{gray}{±0.2} \\
Integrated Gradients & 55.6 \textcolor{gray}{±0.1} & 55.7 \textcolor{gray}{±0.2} & 53.7 \textcolor{gray}{±0.1} & 54.2 \textcolor{gray}{±0.2} \\
\midrule
\IxG{}  & 51.9 \textcolor{gray}{±0.1} & 51.7 \textcolor{gray}{±0.1} & 51.9 \textcolor{gray}{±0.1} & 51.9 \textcolor{gray}{±0.1} \\
\textbf{Libra \IxG{} } & 58.4 \textcolor{gray}{±0.1} (\textcolor{blue}{+12.7\%}) & 58.1 \textcolor{gray}{±0.2} (\textcolor{blue}{+12.3\%}) & 59.3 \textcolor{gray}{±0.1} (\textcolor{blue}{+14.4\%}) & 58.5 \textcolor{gray}{±0.2} (\textcolor{blue}{+12.8\%}) \\
\midrule
AttCAT & 60.1 \textcolor{gray}{±0.1} & 60.0 \textcolor{gray}{±0.2} & 60.7 \textcolor{gray}{±0.1} & 60.5 \textcolor{gray}{±0.1} \\
\textbf{Libra AttCAT} & \underline{66.6} \textcolor{gray}{±0.1} (\textcolor{blue}{+10.9\%}) & \underline{65.4} \textcolor{gray}{±0.2} (\textcolor{blue}{+9.0\%}) & \underline{67.9} \textcolor{gray}{±0.1} (\textcolor{blue}{+12.0\%}) & \underline{66.3} \textcolor{gray}{±0.2} (\textcolor{blue}{+9.6\%}) \\
\midrule
GenAtt & 55.6 \textcolor{gray}{±0.1} & 56.1 \textcolor{gray}{±0.2} & 57.0 \textcolor{gray}{±0.1} & 56.8 \textcolor{gray}{±0.2} \\
\textbf{Libra GenAtt} & 56.6 \textcolor{gray}{±0.1} (\textcolor{blue}{+1.7\%}) & 56.8 \textcolor{gray}{±0.2} (\textcolor{blue}{+1.3\%}) & 58.1 \textcolor{gray}{±0.1} (\textcolor{blue}{+1.9\%}) & 57.5 \textcolor{gray}{±0.2} (\textcolor{blue}{+1.3\%}) \\
\midrule
TokenTM & 60.3 \textcolor{gray}{±0.1} & 60.3 \textcolor{gray}{±0.2} & 62.1 \textcolor{gray}{±0.1} & 61.2 \textcolor{gray}{±0.2} \\
\textbf{Libra TokenTM} & 59.4 \textcolor{gray}{±0.1} (\textcolor{coral}{-1.6\%}) & 59.3 \textcolor{gray}{±0.3} (\textcolor{coral}{-1.6\%}) & 61.0 \textcolor{gray}{±0.1} (\textcolor{coral}{-1.7\%}) & 60.0 \textcolor{gray}{±0.2} (\textcolor{coral}{-1.8\%}) \\
\midrule
GradCAM+ & 58.8 \textcolor{gray}{±0.1} & 58.3 \textcolor{gray}{±0.2} & 59.2 \textcolor{gray}{±0.1} & 58.6 \textcolor{gray}{±0.2} \\
\textbf{Libra GradCAM+} & 60.3 \textcolor{gray}{±0.1} (\textcolor{blue}{+2.6\%}) & 59.6 \textcolor{gray}{±0.2} (\textcolor{blue}{+2.1\%}) & 60.9 \textcolor{gray}{±0.1} (\textcolor{blue}{+2.8\%}) & 59.9 \textcolor{gray}{±0.2} (\textcolor{blue}{+2.3\%}) \\
\midrule
HiResCAM & 59.9 \textcolor{gray}{±0.1} & 59.2 \textcolor{gray}{±0.2} & 60.6 \textcolor{gray}{±0.1} & 59.6 \textcolor{gray}{±0.2} \\
\textbf{Libra HiResCAM} & 60.8 \textcolor{gray}{±0.1} (\textcolor{blue}{+1.5\%}) & 59.8 \textcolor{gray}{±0.2} (\textcolor{blue}{+1.1\%}) & 61.6 \textcolor{gray}{±0.1} (\textcolor{blue}{+1.6\%}) & 60.4 \textcolor{gray}{±0.2} (\textcolor{blue}{+1.2\%}) \\
\midrule
XGradCAM+ & 57.3 \textcolor{gray}{±0.1} & 56.9 \textcolor{gray}{±0.2} & 57.4 \textcolor{gray}{±0.1} & 57.2 \textcolor{gray}{±0.1} \\
\textbf{Libra XGradCAM+} & 64.5 \textcolor{gray}{±0.1} (\textcolor{blue}{+12.6\%}) & 63.3 \textcolor{gray}{±0.2} (\textcolor{blue}{+11.1\%}) & 65.6 \textcolor{gray}{±0.1} (\textcolor{blue}{+14.2\%}) & 64.0 \textcolor{gray}{±0.2} (\textcolor{blue}{+12.0\%}) \\
\midrule
FullGrad+ & 57.1 \textcolor{gray}{±0.1} & 56.9 \textcolor{gray}{±0.2} & 57.5 \textcolor{gray}{±0.1} & 57.3 \textcolor{gray}{±0.2} \\
\textbf{Libra FullGrad+} & \textbf{67.1} \textcolor{gray}{±0.1} (\textcolor{blue}{+17.5\%}) & \textbf{65.8} \textcolor{gray}{±0.2} (\textcolor{blue}{+15.7\%}) & \textbf{68.5} \textcolor{gray}{±0.1} (\textcolor{blue}{+19.0\%}) & \textbf{66.8} \textcolor{gray}{±0.2} (\textcolor{blue}{+16.5\%}) \\
\bottomrule
\end{tabular}
\caption{Comparison of attribution methods and their LibraGrad-enhanced versions on the BEiT2-L model.}
\label{tbl:SRG_m_2}
\end{table}
  \endgroup%

\end{table}

\FloatBarrier
\clearpage
\subsubsection{SigLIP-L}
\label{per_model:SigLIP-L}
Since SigLIP does not have a CLS token, certain attribution methods couldn't be applied and were omitted.
\begin{table}[ht]
  \fontsize{9.5pt}{8.5pt}\selectfont

  \setlength{\tabcolsep}{2pt}
  \centering
  \begingroup%
  \renewenvironment{table}[1][]%
    {\begin{center}}%
    {\end{center}}%
  \begin{table}[!t]
\centering
\begin{tabular}{lccccc}
\toprule
Method & \multicolumn{2}{c}{MIF Deletion (GT)} & \multicolumn{2}{c}{MIF Deletion (Predicted)} & Segmentation \\
  & Accuracy & AOPC & Accuracy & AOPC & AP \\
\cmidrule(r){1-1}
\cmidrule(lr){2-3}
\cmidrule(lr){4-5}
\cmidrule(l){6-6}
Random & 39.0 \textcolor{gray}{±0.1} & 17.3 \textcolor{gray}{±0.2} & 32.8 \textcolor{gray}{±0.1} & 19.1 \textcolor{gray}{±0.2} & 33.0 \textcolor{gray}{±0.3} \\
AliLRP & 58.8 \textcolor{gray}{±0.1} & 36.6 \textcolor{gray}{±0.3} & 55.4 \textcolor{gray}{±0.1} & 40.0 \textcolor{gray}{±0.3} & 33.5 \textcolor{gray}{±0.3} \\
AttnLRP & 64.7 \textcolor{gray}{±0.1} & 42.4 \textcolor{gray}{±0.3} & 62.2 \textcolor{gray}{±0.1} & 46.2 \textcolor{gray}{±0.3} & 36.0 \textcolor{gray}{±0.3} \\
DecompX & 54.5 \textcolor{gray}{±0.1} & 32.6 \textcolor{gray}{±0.2} & 51.1 \textcolor{gray}{±0.1} & 35.7 \textcolor{gray}{±0.2} & 40.5 \textcolor{gray}{±0.3} \\
Integrated Gradients & 52.7 \textcolor{gray}{±0.1} & 30.0 \textcolor{gray}{±0.2} & 44.0 \textcolor{gray}{±0.1} & 28.8 \textcolor{gray}{±0.2} & 41.6 \textcolor{gray}{±0.3} \\
\midrule
\IxG{}  & 44.4 \textcolor{gray}{±0.1} & 23.2 \textcolor{gray}{±0.2} & 40.8 \textcolor{gray}{±0.1} & 26.0 \textcolor{gray}{±0.2} & 35.5 \textcolor{gray}{±0.3} \\
\textbf{Libra \IxG{} } & 54.7 \textcolor{gray}{±0.1} (\textcolor{blue}{+23.4\%}) & 32.4 \textcolor{gray}{±0.2} (\textcolor{blue}{+40.0\%}) & 51.1 \textcolor{gray}{±0.1} (\textcolor{blue}{+25.4\%}) & 35.6 \textcolor{gray}{±0.2} (\textcolor{blue}{+36.9\%}) & 39.9 \textcolor{gray}{±0.3} (\textcolor{blue}{+12.3\%}) \\
\midrule
AttCAT & 48.3 \textcolor{gray}{±0.1} & 27.4 \textcolor{gray}{±0.3} & 45.9 \textcolor{gray}{±0.1} & 30.9 \textcolor{gray}{±0.2} & 37.6 \textcolor{gray}{±0.3} \\
\textbf{Libra AttCAT} & \textbf{79.0} \textcolor{gray}{±0.1} (\textcolor{blue}{+63.4\%}) & \textbf{55.0} \textcolor{gray}{±0.3} (\textcolor{blue}{+100.5\%}) & \textbf{77.4} \textcolor{gray}{±0.1} (\textcolor{blue}{+68.6\%}) & \textbf{59.7} \textcolor{gray}{±0.2} (\textcolor{blue}{+93.1\%}) & 46.8 \textcolor{gray}{±0.3} (\textcolor{blue}{+24.2\%}) \\
\midrule
GradCAM+ & 47.6 \textcolor{gray}{±0.1} & 25.4 \textcolor{gray}{±0.3} & 43.5 \textcolor{gray}{±0.1} & 28.1 \textcolor{gray}{±0.2} & 44.3 \textcolor{gray}{±0.4} \\
\textbf{Libra GradCAM+} & 51.0 \textcolor{gray}{±0.1} (\textcolor{blue}{+7.2\%}) & 28.8 \textcolor{gray}{±0.3} (\textcolor{blue}{+13.6\%}) & 47.4 \textcolor{gray}{±0.1} (\textcolor{blue}{+9.0\%}) & 31.9 \textcolor{gray}{±0.3} (\textcolor{blue}{+13.5\%}) & 41.7 \textcolor{gray}{±0.3} (\textcolor{coral}{-5.7\%}) \\
\midrule
HiResCAM & 37.1 \textcolor{gray}{±0.1} & 15.7 \textcolor{gray}{±0.2} & 31.4 \textcolor{gray}{±0.1} & 17.5 \textcolor{gray}{±0.2} & 36.3 \textcolor{gray}{±0.3} \\
\textbf{Libra HiResCAM} & 50.0 \textcolor{gray}{±0.1} (\textcolor{blue}{+34.8\%}) & 27.5 \textcolor{gray}{±0.3} (\textcolor{blue}{+75.7\%}) & 46.1 \textcolor{gray}{±0.1} (\textcolor{blue}{+46.7\%}) & 30.4 \textcolor{gray}{±0.2} (\textcolor{blue}{+73.7\%}) & \underline{47.5} \textcolor{gray}{±0.3} (\textcolor{blue}{+30.8\%}) \\
\midrule
XGradCAM+ & 54.8 \textcolor{gray}{±0.1} & 34.5 \textcolor{gray}{±0.3} & 51.4 \textcolor{gray}{±0.1} & 37.8 \textcolor{gray}{±0.2} & 43.0 \textcolor{gray}{±0.3} \\
\textbf{Libra XGradCAM+} & 66.3 \textcolor{gray}{±0.1} (\textcolor{blue}{+21.0\%}) & 42.5 \textcolor{gray}{±0.3} (\textcolor{blue}{+23.2\%}) & 63.6 \textcolor{gray}{±0.1} (\textcolor{blue}{+23.7\%}) & 46.3 \textcolor{gray}{±0.3} (\textcolor{blue}{+22.6\%}) & 44.3 \textcolor{gray}{±0.4} (\textcolor{blue}{+3.1\%}) \\
\midrule
FullGrad+ & 46.6 \textcolor{gray}{±0.1} & 25.8 \textcolor{gray}{±0.3} & 43.6 \textcolor{gray}{±0.1} & 29.0 \textcolor{gray}{±0.2} & 37.7 \textcolor{gray}{±0.3} \\
\textbf{Libra FullGrad+} & \underline{75.3} \textcolor{gray}{±0.1} (\textcolor{blue}{+61.7\%}) & \underline{50.7} \textcolor{gray}{±0.3} (\textcolor{blue}{+96.6\%}) & \underline{73.5} \textcolor{gray}{±0.1} (\textcolor{blue}{+68.5\%}) & \underline{55.1} \textcolor{gray}{±0.2} (\textcolor{blue}{+89.7\%}) & \textbf{51.7} \textcolor{gray}{±0.3} (\textcolor{blue}{+37.1\%}) \\
\bottomrule
\end{tabular}
\caption{Comparison of attribution methods and their LibraGrad-enhanced versions on the SigLIP-L model. We report faithfulness metrics using Most-Influential-First Deletion, MIF with ground-truth (GT) and predicted labels, including Accuracy and Area Over Perturbation Curve (AOPC) and Segmentation Average Precision (AP). The results demonstrate that composing existing methods with LibraGrad significantly enhances their performance across all metrics.}
\label{tbl:MIF_m_2}
\end{table}
  \endgroup%

\end{table}

\begin{table}[ht]
  \fontsize{11pt}{8.5pt}\selectfont

  \setlength{\tabcolsep}{4pt}
  \centering

  \begingroup%
  \renewenvironment{table}[1][]%
    {\begin{center}}%
    {\end{center}}%
  \begin{table}[!t]
\centering
\begin{tabular}{lcccc}
\toprule
Method & \multicolumn{2}{c}{LIF Deletion (GT)} & \multicolumn{2}{c}{LIF Deletion (Predicted)} \\
  & Accuracy & AOPC & Accuracy & AOPC \\
\cmidrule(r){1-1}
\cmidrule(lr){2-3}
\cmidrule(l){4-5}
Random & 61.1 \textcolor{gray}{±0.1} & 82.7 \textcolor{gray}{±0.2} & 67.1 \textcolor{gray}{±0.1} & 81.0 \textcolor{gray}{±0.1} \\
AliLRP & 70.8 \textcolor{gray}{±0.1} & 91.2 \textcolor{gray}{±0.2} & 77.0 \textcolor{gray}{±0.1} & 89.8 \textcolor{gray}{±0.2} \\
AttnLRP & 75.0 \textcolor{gray}{±0.1} & 96.0 \textcolor{gray}{±0.2} & 82.2 \textcolor{gray}{±0.1} & 95.0 \textcolor{gray}{±0.1} \\
DecompX & 71.3 \textcolor{gray}{±0.1} & 91.8 \textcolor{gray}{±0.2} & 78.1 \textcolor{gray}{±0.1} & 90.5 \textcolor{gray}{±0.2} \\
Integrated Gradients & 75.9 \textcolor{gray}{±0.1} & 97.0 \textcolor{gray}{±0.3} & 75.6 \textcolor{gray}{±0.1} & 91.2 \textcolor{gray}{±0.2} \\
\midrule
\IxG{}  & 67.4 \textcolor{gray}{±0.1} & 89.7 \textcolor{gray}{±0.3} & 71.6 \textcolor{gray}{±0.1} & 87.6 \textcolor{gray}{±0.2} \\
\textbf{Libra \IxG{} } & 71.8 \textcolor{gray}{±0.1} (\textcolor{blue}{+6.5\%}) & 91.9 \textcolor{gray}{±0.3} (\textcolor{blue}{+2.4\%}) & 78.3 \textcolor{gray}{±0.1} (\textcolor{blue}{+9.4\%}) & 90.6 \textcolor{gray}{±0.2} (\textcolor{blue}{+3.5\%}) \\
\midrule
AttCAT & 73.8 \textcolor{gray}{±0.1} & 95.3 \textcolor{gray}{±0.3} & 76.6 \textcolor{gray}{±0.1} & 92.4 \textcolor{gray}{±0.2} \\
\textbf{Libra AttCAT} & \textbf{80.0} \textcolor{gray}{±0.1} (\textcolor{blue}{+8.4\%}) & \textbf{99.6} \textcolor{gray}{±0.2} (\textcolor{blue}{+4.5\%}) & \textbf{85.9} \textcolor{gray}{±0.1} (\textcolor{blue}{+12.2\%}) & \textbf{98.4} \textcolor{gray}{±0.1} (\textcolor{blue}{+6.4\%}) \\
\midrule
GradCAM+ & 45.8 \textcolor{gray}{±0.1} & 64.3 \textcolor{gray}{±0.4} & 49.0 \textcolor{gray}{±0.1} & 60.7 \textcolor{gray}{±0.3} \\
\textbf{Libra GradCAM+} & 62.8 \textcolor{gray}{±0.1} (\textcolor{blue}{+37.1\%}) & 83.0 \textcolor{gray}{±0.3} (\textcolor{blue}{+29.0\%}) & 67.5 \textcolor{gray}{±0.1} (\textcolor{blue}{+37.8\%}) & 80.8 \textcolor{gray}{±0.2} (\textcolor{blue}{+33.1\%}) \\
\midrule
HiResCAM & 48.1 \textcolor{gray}{±0.1} & 69.4 \textcolor{gray}{±0.4} & 51.9 \textcolor{gray}{±0.1} & 66.3 \textcolor{gray}{±0.3} \\
\textbf{Libra HiResCAM} & 63.7 \textcolor{gray}{±0.1} (\textcolor{blue}{+32.2\%}) & 84.8 \textcolor{gray}{±0.3} (\textcolor{blue}{+22.1\%}) & 68.2 \textcolor{gray}{±0.1} (\textcolor{blue}{+31.5\%}) & 82.6 \textcolor{gray}{±0.2} (\textcolor{blue}{+24.6\%}) \\
\midrule
XGradCAM+ & 57.3 \textcolor{gray}{±0.1} & 78.4 \textcolor{gray}{±0.4} & 60.6 \textcolor{gray}{±0.1} & 75.2 \textcolor{gray}{±0.3} \\
\textbf{Libra XGradCAM+} & 70.5 \textcolor{gray}{±0.1} (\textcolor{blue}{+23.2\%}) & 89.8 \textcolor{gray}{±0.3} (\textcolor{blue}{+14.6\%}) & 76.4 \textcolor{gray}{±0.1} (\textcolor{blue}{+26.1\%}) & 88.4 \textcolor{gray}{±0.2} (\textcolor{blue}{+17.5\%}) \\
\midrule
FullGrad+ & 70.4 \textcolor{gray}{±0.1} & 92.2 \textcolor{gray}{±0.3} & 73.3 \textcolor{gray}{±0.1} & 89.3 \textcolor{gray}{±0.2} \\
\textbf{Libra FullGrad+} & \underline{79.8} \textcolor{gray}{±0.1} (\textcolor{blue}{+13.4\%}) & \underline{99.4} \textcolor{gray}{±0.2} (\textcolor{blue}{+7.8\%}) & \underline{85.8} \textcolor{gray}{±0.1} (\textcolor{blue}{+17.0\%}) & \underline{98.2} \textcolor{gray}{±0.1} (\textcolor{blue}{+10.0\%}) \\
\bottomrule
\end{tabular}
\caption{Comparison of attribution methods and their LibraGrad-enhanced versions on the SigLIP-L model.}
\label{tbl:LIF_m_2}
\end{table}
  \endgroup%

  \begingroup%
  \renewenvironment{table}[1][]%
    {\begin{center}}%
    {\end{center}}%
  \begin{table}[!t]
\centering
\begin{tabular}{lcccc}
\toprule
Method & \multicolumn{2}{c}{SRG (GT)} & \multicolumn{2}{c}{SRG (Predicted)} \\
  & Accuracy & AOPC & Accuracy & AOPC \\
\cmidrule(r){1-1}
\cmidrule(lr){2-3}
\cmidrule(l){4-5}
Random & 50.0 \textcolor{gray}{±0.1} & 50.0 \textcolor{gray}{±0.2} & 50.0 \textcolor{gray}{±0.1} & 50.0 \textcolor{gray}{±0.2} \\
AliLRP & 64.8 \textcolor{gray}{±0.1} & 63.9 \textcolor{gray}{±0.3} & 66.2 \textcolor{gray}{±0.1} & 64.9 \textcolor{gray}{±0.2} \\
AttnLRP & 69.8 \textcolor{gray}{±0.1} & 69.2 \textcolor{gray}{±0.3} & 72.2 \textcolor{gray}{±0.1} & 70.6 \textcolor{gray}{±0.2} \\
DecompX & 62.9 \textcolor{gray}{±0.1} & 62.2 \textcolor{gray}{±0.2} & 64.6 \textcolor{gray}{±0.1} & 63.1 \textcolor{gray}{±0.2} \\
Integrated Gradients & 64.3 \textcolor{gray}{±0.1} & 63.5 \textcolor{gray}{±0.3} & 59.8 \textcolor{gray}{±0.1} & 60.0 \textcolor{gray}{±0.2} \\
\midrule
\IxG{}  & 55.9 \textcolor{gray}{±0.1} & 56.4 \textcolor{gray}{±0.3} & 56.2 \textcolor{gray}{±0.1} & 56.8 \textcolor{gray}{±0.2} \\
\textbf{Libra \IxG{} } & 63.3 \textcolor{gray}{±0.1} (\textcolor{blue}{+13.2\%}) & 62.2 \textcolor{gray}{±0.3} (\textcolor{blue}{+10.1\%}) & 64.7 \textcolor{gray}{±0.1} (\textcolor{blue}{+15.2\%}) & 63.1 \textcolor{gray}{±0.2} (\textcolor{blue}{+11.1\%}) \\
\midrule
AttCAT & 61.0 \textcolor{gray}{±0.1} & 61.4 \textcolor{gray}{±0.3} & 61.2 \textcolor{gray}{±0.1} & 61.7 \textcolor{gray}{±0.2} \\
\textbf{Libra AttCAT} & \textbf{79.5} \textcolor{gray}{±0.1} (\textcolor{blue}{+30.2\%}) & \textbf{77.3} \textcolor{gray}{±0.3} (\textcolor{blue}{+26.0\%}) & \textbf{81.6} \textcolor{gray}{±0.1} (\textcolor{blue}{+33.3\%}) & \textbf{79.0} \textcolor{gray}{±0.2} (\textcolor{blue}{+28.2\%}) \\
\midrule
GradCAM+ & 46.7 \textcolor{gray}{±0.1} & 44.9 \textcolor{gray}{±0.3} & 46.2 \textcolor{gray}{±0.1} & 44.4 \textcolor{gray}{±0.3} \\
\textbf{Libra GradCAM+} & 56.9 \textcolor{gray}{±0.1} (\textcolor{blue}{+21.9\%}) & 55.9 \textcolor{gray}{±0.3} (\textcolor{blue}{+24.6\%}) & 57.4 \textcolor{gray}{±0.1} (\textcolor{blue}{+24.2\%}) & 56.4 \textcolor{gray}{±0.3} (\textcolor{blue}{+26.9\%}) \\
\midrule
HiResCAM & 42.6 \textcolor{gray}{±0.1} & 42.5 \textcolor{gray}{±0.3} & 41.7 \textcolor{gray}{±0.1} & 41.9 \textcolor{gray}{±0.2} \\
\textbf{Libra HiResCAM} & 56.8 \textcolor{gray}{±0.1} (\textcolor{blue}{+33.4\%}) & 56.1 \textcolor{gray}{±0.3} (\textcolor{blue}{+32.0\%}) & 57.2 \textcolor{gray}{±0.1} (\textcolor{blue}{+37.2\%}) & 56.5 \textcolor{gray}{±0.2} (\textcolor{blue}{+34.9\%}) \\
\midrule
XGradCAM+ & 56.0 \textcolor{gray}{±0.1} & 56.4 \textcolor{gray}{±0.3} & 56.0 \textcolor{gray}{±0.1} & 56.5 \textcolor{gray}{±0.2} \\
\textbf{Libra XGradCAM+} & 68.4 \textcolor{gray}{±0.1} (\textcolor{blue}{+22.1\%}) & 66.2 \textcolor{gray}{±0.3} (\textcolor{blue}{+17.2\%}) & 70.0 \textcolor{gray}{±0.1} (\textcolor{blue}{+25.0\%}) & 67.3 \textcolor{gray}{±0.2} (\textcolor{blue}{+19.2\%}) \\
\midrule
FullGrad+ & 58.5 \textcolor{gray}{±0.1} & 59.0 \textcolor{gray}{±0.3} & 58.4 \textcolor{gray}{±0.1} & 59.2 \textcolor{gray}{±0.2} \\
\textbf{Libra FullGrad+} & \underline{77.6} \textcolor{gray}{±0.1} (\textcolor{blue}{+32.7\%}) & \underline{75.0} \textcolor{gray}{±0.3} (\textcolor{blue}{+27.2\%}) & \underline{79.6} \textcolor{gray}{±0.1} (\textcolor{blue}{+36.2\%}) & \underline{76.7} \textcolor{gray}{±0.2} (\textcolor{blue}{+29.5\%}) \\
\bottomrule
\end{tabular}
\caption{Comparison of attribution methods and their LibraGrad-enhanced versions on the SigLIP-L model.}
\label{tbl:SRG_m_2}
\end{table}
  \endgroup%

\end{table}

\FloatBarrier
\clearpage
\subsubsection{CLIP-H}
\label{per_model:CLIP-H}

\begin{table}[ht]
  \fontsize{9.5pt}{8.5pt}\selectfont

  \setlength{\tabcolsep}{2pt}
  \centering
  \begingroup%
  \renewenvironment{table}[1][]%
    {\begin{center}}%
    {\end{center}}%
  \begin{table}[h]
\centering
\begin{tabular}{lccccc}
\toprule
Method & \multicolumn{2}{c}{MIF Deletion (GT)} & \multicolumn{2}{c}{MIF Deletion (Predicted)} & Segmentation \\
  & Accuracy & AOPC & Accuracy & AOPC & AP \\
\cmidrule(r){1-1}
\cmidrule(lr){2-3}
\cmidrule(lr){4-5}
\cmidrule(l){6-6}
Random & 34.3 \textcolor{gray}{±0.1} & 11.2 \textcolor{gray}{±0.2} & 28.0 \textcolor{gray}{±0.1} & 12.7 \textcolor{gray}{±0.2} & 37.8 \textcolor{gray}{±0.3} \\
RawAtt & 46.9 \textcolor{gray}{±0.1} & 21.0 \textcolor{gray}{±0.2} & 42.5 \textcolor{gray}{±0.1} & 23.3 \textcolor{gray}{±0.2} & 41.6 \textcolor{gray}{±0.3} \\
Attention Rollout & 46.4 \textcolor{gray}{±0.1} & 20.5 \textcolor{gray}{±0.3} & 41.3 \textcolor{gray}{±0.1} & 22.5 \textcolor{gray}{±0.3} & 51.7 \textcolor{gray}{±0.4} \\
AliLRP & 40.0 \textcolor{gray}{±0.1} & 15.7 \textcolor{gray}{±0.2} & 34.4 \textcolor{gray}{±0.1} & 17.3 \textcolor{gray}{±0.2} & 38.1 \textcolor{gray}{±0.3} \\
AttnLRP & 50.8 \textcolor{gray}{±0.1} & 24.0 \textcolor{gray}{±0.3} & 46.7 \textcolor{gray}{±0.1} & 26.4 \textcolor{gray}{±0.2} & 50.9 \textcolor{gray}{±0.3} \\
DecompX & 46.7 \textcolor{gray}{±0.1} & 21.3 \textcolor{gray}{±0.2} & 42.4 \textcolor{gray}{±0.1} & 23.5 \textcolor{gray}{±0.2} & 55.0 \textcolor{gray}{±0.3} \\
Integrated Gradients & 37.1 \textcolor{gray}{±0.1} & 13.5 \textcolor{gray}{±0.2} & 31.0 \textcolor{gray}{±0.1} & 15.0 \textcolor{gray}{±0.2} & 36.9 \textcolor{gray}{±0.3} \\
\midrule
\IxG{}  & 37.5 \textcolor{gray}{±0.1} & 13.7 \textcolor{gray}{±0.2} & 31.4 \textcolor{gray}{±0.1} & 15.2 \textcolor{gray}{±0.2} & 36.8 \textcolor{gray}{±0.3} \\
\textbf{Libra \IxG{} } & 47.5 \textcolor{gray}{±0.1} (\textcolor{blue}{+26.8\%}) & 21.8 \textcolor{gray}{±0.2} (\textcolor{blue}{+59.4\%}) & 43.1 \textcolor{gray}{±0.1} (\textcolor{blue}{+37.3\%}) & 24.0 \textcolor{gray}{±0.2} (\textcolor{blue}{+57.9\%}) & 54.2 \textcolor{gray}{±0.3} (\textcolor{blue}{+47.3\%}) \\
\midrule
AttCAT & 42.5 \textcolor{gray}{±0.1} & 18.8 \textcolor{gray}{±0.2} & 39.0 \textcolor{gray}{±0.1} & 21.3 \textcolor{gray}{±0.1} & 38.9 \textcolor{gray}{±0.3} \\
\textbf{Libra AttCAT} & \underline{61.5} \textcolor{gray}{±0.1} (\textcolor{blue}{+44.8\%}) & \underline{31.7} \textcolor{gray}{±0.3} (\textcolor{blue}{+68.9\%}) & \underline{58.5} \textcolor{gray}{±0.1} (\textcolor{blue}{+49.8\%}) & \underline{34.7} \textcolor{gray}{±0.2} (\textcolor{blue}{+62.8\%}) & 61.7 \textcolor{gray}{±0.3} (\textcolor{blue}{+58.6\%}) \\
\midrule
GenAtt & 54.4 \textcolor{gray}{±0.1} & 26.8 \textcolor{gray}{±0.2} & 51.0 \textcolor{gray}{±0.1} & 29.6 \textcolor{gray}{±0.2} & 55.9 \textcolor{gray}{±0.3} \\
\textbf{Libra GenAtt} & 61.0 \textcolor{gray}{±0.1} (\textcolor{blue}{+12.2\%}) & 31.5 \textcolor{gray}{±0.3} (\textcolor{blue}{+17.5\%}) & 58.1 \textcolor{gray}{±0.1} (\textcolor{blue}{+14.0\%}) & 34.5 \textcolor{gray}{±0.2} (\textcolor{blue}{+16.7\%}) & \textbf{76.2} \textcolor{gray}{±0.2} (\textcolor{blue}{+36.1\%}) \\
\midrule
TokenTM & 55.4 \textcolor{gray}{±0.1} & 27.4 \textcolor{gray}{±0.3} & 51.9 \textcolor{gray}{±0.1} & 30.1 \textcolor{gray}{±0.2} & 58.6 \textcolor{gray}{±0.3} \\
\textbf{Libra TokenTM} & 60.6 \textcolor{gray}{±0.1} (\textcolor{blue}{+9.3\%}) & 31.2 \textcolor{gray}{±0.3} (\textcolor{blue}{+14.0\%}) & 57.4 \textcolor{gray}{±0.1} (\textcolor{blue}{+10.6\%}) & 34.1 \textcolor{gray}{±0.2} (\textcolor{blue}{+13.5\%}) & 71.5 \textcolor{gray}{±0.3} (\textcolor{blue}{+22.1\%}) \\
\midrule
GradCAM+ & 38.6 \textcolor{gray}{±0.1} & 14.5 \textcolor{gray}{±0.2} & 33.0 \textcolor{gray}{±0.1} & 16.2 \textcolor{gray}{±0.2} & 43.0 \textcolor{gray}{±0.4} \\
\textbf{Libra GradCAM+} & 41.8 \textcolor{gray}{±0.1} (\textcolor{blue}{+8.4\%}) & 16.8 \textcolor{gray}{±0.2} (\textcolor{blue}{+15.6\%}) & 36.2 \textcolor{gray}{±0.1} (\textcolor{blue}{+9.8\%}) & 18.6 \textcolor{gray}{±0.2} (\textcolor{blue}{+14.9\%}) & 47.4 \textcolor{gray}{±0.4} (\textcolor{blue}{+10.2\%}) \\
\midrule
HiResCAM & 42.3 \textcolor{gray}{±0.1} & 17.6 \textcolor{gray}{±0.2} & 37.6 \textcolor{gray}{±0.1} & 19.7 \textcolor{gray}{±0.2} & 45.9 \textcolor{gray}{±0.3} \\
\textbf{Libra HiResCAM} & 52.8 \textcolor{gray}{±0.1} (\textcolor{blue}{+24.8\%}) & 25.4 \textcolor{gray}{±0.2} (\textcolor{blue}{+44.3\%}) & 48.9 \textcolor{gray}{±0.1} (\textcolor{blue}{+29.9\%}) & 27.9 \textcolor{gray}{±0.2} (\textcolor{blue}{+41.9\%}) & 56.8 \textcolor{gray}{±0.3} (\textcolor{blue}{+23.7\%}) \\
\midrule
XGradCAM+ & 44.2 \textcolor{gray}{±0.1} & 19.2 \textcolor{gray}{±0.2} & 39.4 \textcolor{gray}{±0.1} & 21.3 \textcolor{gray}{±0.2} & 47.7 \textcolor{gray}{±0.4} \\
\textbf{Libra XGradCAM+} & 60.8 \textcolor{gray}{±0.1} (\textcolor{blue}{+37.4\%}) & 31.1 \textcolor{gray}{±0.2} (\textcolor{blue}{+62.0\%}) & 57.7 \textcolor{gray}{±0.1} (\textcolor{blue}{+46.4\%}) & 34.1 \textcolor{gray}{±0.2} (\textcolor{blue}{+59.7\%}) & \underline{73.3} \textcolor{gray}{±0.3} (\textcolor{blue}{+53.8\%}) \\
\midrule
FullGrad+ & 41.4 \textcolor{gray}{±0.1} & 18.1 \textcolor{gray}{±0.2} & 37.6 \textcolor{gray}{±0.1} & 20.5 \textcolor{gray}{±0.2} & 38.5 \textcolor{gray}{±0.3} \\
\textbf{Libra FullGrad+} & \textbf{63.8} \textcolor{gray}{±0.1} (\textcolor{blue}{+54.3\%}) & \textbf{33.6} \textcolor{gray}{±0.3} (\textcolor{blue}{+85.9\%}) & \textbf{61.1} \textcolor{gray}{±0.1} (\textcolor{blue}{+62.3\%}) & \textbf{36.8} \textcolor{gray}{±0.2} (\textcolor{blue}{+79.1\%}) & 71.5 \textcolor{gray}{±0.3} (\textcolor{blue}{+85.7\%}) \\
\bottomrule
\end{tabular}
\caption{Comparison of attribution methods and their LibraGrad-enhanced versions on the CLIP-H model. We report faithfulness metrics using Most-Influential-First Deletion, MIF with ground-truth (GT) and predicted labels, including Accuracy and Area Over Perturbation Curve (AOPC) and Segmentation Average Precision (AP). The results demonstrate that composing existing methods with LibraGrad significantly enhances their performance across all metrics.}
\label{tbl:MIF_m_2}
\end{table}
  \endgroup%

\end{table}

\begin{table}[ht]
  \fontsize{11pt}{8.5pt}\selectfont

  \setlength{\tabcolsep}{4pt}
  \centering

  \begingroup%
  \renewenvironment{table}[1][]%
    {\begin{center}}%
    {\end{center}}%
  \begin{table}[h]
\centering
\begin{tabular}{lcccc}
\toprule
Method & \multicolumn{2}{c}{LIF Deletion (GT)} & \multicolumn{2}{c}{LIF Deletion (Predicted)} \\
  & Accuracy & AOPC & Accuracy & AOPC \\
\cmidrule(r){1-1}
\cmidrule(lr){2-3}
\cmidrule(l){4-5}
Random & 65.8 \textcolor{gray}{±0.1} & 88.8 \textcolor{gray}{±0.2} & 72.4 \textcolor{gray}{±0.1} & 87.5 \textcolor{gray}{±0.2} \\
RawAtt & 68.7 \textcolor{gray}{±0.1} & 91.1 \textcolor{gray}{±0.1} & 76.0 \textcolor{gray}{±0.1} & 90.0 \textcolor{gray}{±0.2} \\
Attention Rollout & 68.1 \textcolor{gray}{±0.1} & 90.7 \textcolor{gray}{±0.2} & 74.6 \textcolor{gray}{±0.1} & 89.4 \textcolor{gray}{±0.2} \\
AliLRP & 69.1 \textcolor{gray}{±0.1} & 91.3 \textcolor{gray}{±0.1} & 75.3 \textcolor{gray}{±0.1} & 89.9 \textcolor{gray}{±0.1} \\
AttnLRP & 76.8 \textcolor{gray}{±0.1} & 97.3 \textcolor{gray}{±0.2} & 83.3 \textcolor{gray}{±0.1} & 96.1 \textcolor{gray}{±0.1} \\
DecompX & 74.8 \textcolor{gray}{±0.1} & 95.4 \textcolor{gray}{±0.2} & 81.7 \textcolor{gray}{±0.1} & 94.2 \textcolor{gray}{±0.2} \\
Integrated Gradients & 63.3 \textcolor{gray}{±0.1} & 87.2 \textcolor{gray}{±0.1} & 69.4 \textcolor{gray}{±0.1} & 85.7 \textcolor{gray}{±0.1} \\
\midrule
\IxG{}  & 62.7 \textcolor{gray}{±0.1} & 86.5 \textcolor{gray}{±0.2} & 68.8 \textcolor{gray}{±0.1} & 84.9 \textcolor{gray}{±0.1} \\
\textbf{Libra \IxG{} } & 76.0 \textcolor{gray}{±0.1} (\textcolor{blue}{+21.2\%}) & 96.3 \textcolor{gray}{±0.2} (\textcolor{blue}{+11.3\%}) & 82.2 \textcolor{gray}{±0.1} (\textcolor{blue}{+19.4\%}) & 94.7 \textcolor{gray}{±0.2} (\textcolor{blue}{+11.5\%}) \\
\midrule
AttCAT & 72.3 \textcolor{gray}{±0.1} & 96.3 \textcolor{gray}{±0.2} & 76.9 \textcolor{gray}{±0.1} & 94.8 \textcolor{gray}{±0.2} \\
\textbf{Libra AttCAT} & \underline{78.1} \textcolor{gray}{±0.1} (\textcolor{blue}{+7.9\%}) & \underline{98.1} \textcolor{gray}{±0.1} (\textcolor{blue}{+1.8\%}) & \underline{83.8} \textcolor{gray}{±0.1} (\textcolor{blue}{+9.0\%}) & \underline{96.4} \textcolor{gray}{±0.1} (\textcolor{blue}{+1.6\%}) \\
\midrule
GenAtt & 73.4 \textcolor{gray}{±0.1} & 95.3 \textcolor{gray}{±0.2} & 80.8 \textcolor{gray}{±0.1} & 94.3 \textcolor{gray}{±0.2} \\
\textbf{Libra GenAtt} & 75.0 \textcolor{gray}{±0.1} (\textcolor{blue}{+2.2\%}) & 95.5 \textcolor{gray}{±0.1} (\textcolor{blue}{+0.3\%}) & 82.5 \textcolor{gray}{±0.1} (\textcolor{blue}{+2.0\%}) & 94.5 \textcolor{gray}{±0.1} (\textcolor{blue}{+0.2\%}) \\
\midrule
TokenTM & 73.1 \textcolor{gray}{±0.1} & 94.5 \textcolor{gray}{±0.1} & 80.6 \textcolor{gray}{±0.1} & 93.4 \textcolor{gray}{±0.1} \\
\textbf{Libra TokenTM} & 74.1 \textcolor{gray}{±0.1} (\textcolor{blue}{+1.3\%}) & 94.6 \textcolor{gray}{±0.1} (\textcolor{blue}{+0.2\%}) & 81.7 \textcolor{gray}{±0.1} (\textcolor{blue}{+1.4\%}) & 93.6 \textcolor{gray}{±0.1} (\textcolor{blue}{+0.2\%}) \\
\midrule
GradCAM+ & 63.8 \textcolor{gray}{±0.1} & 87.4 \textcolor{gray}{±0.2} & 69.4 \textcolor{gray}{±0.1} & 85.6 \textcolor{gray}{±0.2} \\
\textbf{Libra GradCAM+} & 65.6 \textcolor{gray}{±0.1} (\textcolor{blue}{+2.9\%}) & 88.4 \textcolor{gray}{±0.3} (\textcolor{blue}{+1.1\%}) & 70.7 \textcolor{gray}{±0.1} (\textcolor{blue}{+1.9\%}) & 86.4 \textcolor{gray}{±0.2} (\textcolor{blue}{+0.9\%}) \\
\midrule
HiResCAM & 72.4 \textcolor{gray}{±0.1} & 94.3 \textcolor{gray}{±0.2} & 77.9 \textcolor{gray}{±0.1} & 92.7 \textcolor{gray}{±0.2} \\
\textbf{Libra HiResCAM} & 74.7 \textcolor{gray}{±0.1} (\textcolor{blue}{+3.3\%}) & 95.6 \textcolor{gray}{±0.1} (\textcolor{blue}{+1.4\%}) & 80.9 \textcolor{gray}{±0.1} (\textcolor{blue}{+3.9\%}) & 94.1 \textcolor{gray}{±0.1} (\textcolor{blue}{+1.5\%}) \\
\midrule
XGradCAM+ & 69.7 \textcolor{gray}{±0.1} & 92.1 \textcolor{gray}{±0.2} & 75.4 \textcolor{gray}{±0.1} & 90.6 \textcolor{gray}{±0.2} \\
\textbf{Libra XGradCAM+} & 75.5 \textcolor{gray}{±0.1} (\textcolor{blue}{+8.5\%}) & 95.8 \textcolor{gray}{±0.1} (\textcolor{blue}{+3.9\%}) & 81.0 \textcolor{gray}{±0.1} (\textcolor{blue}{+7.4\%}) & 93.9 \textcolor{gray}{±0.2} (\textcolor{blue}{+3.7\%}) \\
\midrule
FullGrad+ & 71.4 \textcolor{gray}{±0.1} & 95.0 \textcolor{gray}{±0.2} & 76.2 \textcolor{gray}{±0.1} & 93.4 \textcolor{gray}{±0.2} \\
\textbf{Libra FullGrad+} & \textbf{79.1} \textcolor{gray}{±0.1} (\textcolor{blue}{+10.9\%}) & \textbf{98.4} \textcolor{gray}{±0.1} (\textcolor{blue}{+3.6\%}) & \textbf{84.9} \textcolor{gray}{±0.1} (\textcolor{blue}{+11.4\%}) & \textbf{96.8} \textcolor{gray}{±0.2} (\textcolor{blue}{+3.6\%}) \\
\bottomrule
\end{tabular}
\caption{Comparison of attribution methods and their LibraGrad-enhanced versions on the CLIP-H model.}
\label{tbl:LIF_m_2}
\end{table}
  \endgroup%

  \begingroup%
  \renewenvironment{table}[1][]%
    {\begin{center}}%
    {\end{center}}%
  \begin{table}[h]
\centering
\begin{tabular}{lcccc}
\toprule
Method & \multicolumn{2}{c}{SRG (GT)} & \multicolumn{2}{c}{SRG (Predicted)} \\
  & Accuracy & AOPC & Accuracy & AOPC \\
\cmidrule(r){1-1}
\cmidrule(lr){2-3}
\cmidrule(l){4-5}
Random & 50.0 \textcolor{gray}{±0.1} & 50.0 \textcolor{gray}{±0.2} & 50.2 \textcolor{gray}{±0.1} & 50.1 \textcolor{gray}{±0.2} \\
RawAtt & 57.8 \textcolor{gray}{±0.1} & 56.1 \textcolor{gray}{±0.2} & 59.2 \textcolor{gray}{±0.1} & 56.7 \textcolor{gray}{±0.2} \\
Attention Rollout & 57.2 \textcolor{gray}{±0.1} & 55.6 \textcolor{gray}{±0.3} & 58.0 \textcolor{gray}{±0.1} & 55.9 \textcolor{gray}{±0.2} \\
AliLRP & 54.5 \textcolor{gray}{±0.1} & 53.5 \textcolor{gray}{±0.2} & 54.8 \textcolor{gray}{±0.1} & 53.6 \textcolor{gray}{±0.2} \\
AttnLRP & 63.8 \textcolor{gray}{±0.1} & 60.6 \textcolor{gray}{±0.2} & 65.0 \textcolor{gray}{±0.1} & 61.3 \textcolor{gray}{±0.2} \\
DecompX & 60.8 \textcolor{gray}{±0.1} & 58.3 \textcolor{gray}{±0.2} & 62.1 \textcolor{gray}{±0.1} & 58.9 \textcolor{gray}{±0.2} \\
Integrated Gradients & 50.2 \textcolor{gray}{±0.1} & 50.3 \textcolor{gray}{±0.2} & 50.2 \textcolor{gray}{±0.1} & 50.3 \textcolor{gray}{±0.1} \\
\midrule
\IxG{}  & 50.1 \textcolor{gray}{±0.1} & 50.1 \textcolor{gray}{±0.2} & 50.1 \textcolor{gray}{±0.1} & 50.1 \textcolor{gray}{±0.1} \\
\textbf{Libra \IxG{} } & 61.7 \textcolor{gray}{±0.1} (\textcolor{blue}{+23.3\%}) & 59.0 \textcolor{gray}{±0.2} (\textcolor{blue}{+17.8\%}) & 62.6 \textcolor{gray}{±0.1} (\textcolor{blue}{+25.0\%}) & 59.4 \textcolor{gray}{±0.2} (\textcolor{blue}{+18.6\%}) \\
\midrule
AttCAT & 57.4 \textcolor{gray}{±0.1} & 57.5 \textcolor{gray}{±0.2} & 58.0 \textcolor{gray}{±0.1} & 58.1 \textcolor{gray}{±0.2} \\
\textbf{Libra AttCAT} & \underline{69.8} \textcolor{gray}{±0.1} (\textcolor{blue}{+21.6\%}) & \underline{64.9} \textcolor{gray}{±0.2} (\textcolor{blue}{+12.8\%}) & \underline{71.2} \textcolor{gray}{±0.1} (\textcolor{blue}{+22.7\%}) & \underline{65.5} \textcolor{gray}{±0.2} (\textcolor{blue}{+12.9\%}) \\
\midrule
GenAtt & 63.9 \textcolor{gray}{±0.1} & 61.1 \textcolor{gray}{±0.2} & 65.9 \textcolor{gray}{±0.1} & 62.0 \textcolor{gray}{±0.2} \\
\textbf{Libra GenAtt} & 68.0 \textcolor{gray}{±0.1} (\textcolor{blue}{+6.5\%}) & 63.5 \textcolor{gray}{±0.2} (\textcolor{blue}{+4.1\%}) & 70.3 \textcolor{gray}{±0.1} (\textcolor{blue}{+6.7\%}) & 64.5 \textcolor{gray}{±0.2} (\textcolor{blue}{+4.1\%}) \\
\midrule
TokenTM & 64.3 \textcolor{gray}{±0.1} & 60.9 \textcolor{gray}{±0.2} & 66.2 \textcolor{gray}{±0.1} & 61.8 \textcolor{gray}{±0.2} \\
\textbf{Libra TokenTM} & 67.3 \textcolor{gray}{±0.1} (\textcolor{blue}{+4.7\%}) & 62.9 \textcolor{gray}{±0.2} (\textcolor{blue}{+3.3\%}) & 69.5 \textcolor{gray}{±0.1} (\textcolor{blue}{+5.0\%}) & 63.9 \textcolor{gray}{±0.2} (\textcolor{blue}{+3.5\%}) \\
\midrule
GradCAM+ & 51.2 \textcolor{gray}{±0.1} & 51.0 \textcolor{gray}{±0.2} & 51.2 \textcolor{gray}{±0.1} & 50.9 \textcolor{gray}{±0.2} \\
\textbf{Libra GradCAM+} & 53.7 \textcolor{gray}{±0.1} (\textcolor{blue}{+4.9\%}) & 52.6 \textcolor{gray}{±0.2} (\textcolor{blue}{+3.2\%}) & 53.5 \textcolor{gray}{±0.1} (\textcolor{blue}{+4.5\%}) & 52.5 \textcolor{gray}{±0.2} (\textcolor{blue}{+3.1\%}) \\
\midrule
HiResCAM & 57.3 \textcolor{gray}{±0.1} & 55.9 \textcolor{gray}{±0.2} & 57.8 \textcolor{gray}{±0.1} & 56.2 \textcolor{gray}{±0.2} \\
\textbf{Libra HiResCAM} & 63.8 \textcolor{gray}{±0.1} (\textcolor{blue}{+11.2\%}) & 60.5 \textcolor{gray}{±0.2} (\textcolor{blue}{+8.1\%}) & 64.9 \textcolor{gray}{±0.1} (\textcolor{blue}{+12.3\%}) & 61.0 \textcolor{gray}{±0.2} (\textcolor{blue}{+8.6\%}) \\
\midrule
XGradCAM+ & 56.9 \textcolor{gray}{±0.1} & 55.7 \textcolor{gray}{±0.2} & 57.4 \textcolor{gray}{±0.1} & 56.0 \textcolor{gray}{±0.2} \\
\textbf{Libra XGradCAM+} & 68.2 \textcolor{gray}{±0.1} (\textcolor{blue}{+19.7\%}) & 63.4 \textcolor{gray}{±0.2} (\textcolor{blue}{+13.9\%}) & 69.3 \textcolor{gray}{±0.1} (\textcolor{blue}{+20.8\%}) & 64.0 \textcolor{gray}{±0.2} (\textcolor{blue}{+14.4\%}) \\
\midrule
FullGrad+ & 56.4 \textcolor{gray}{±0.1} & 56.5 \textcolor{gray}{±0.2} & 56.9 \textcolor{gray}{±0.1} & 57.0 \textcolor{gray}{±0.2} \\
\textbf{Libra FullGrad+} & \textbf{71.5} \textcolor{gray}{±0.1} (\textcolor{blue}{+26.8\%}) & \textbf{66.0} \textcolor{gray}{±0.2} (\textcolor{blue}{+16.7\%}) & \textbf{73.0} \textcolor{gray}{±0.1} (\textcolor{blue}{+28.2\%}) & \textbf{66.8} \textcolor{gray}{±0.2} (\textcolor{blue}{+17.2\%}) \\
\bottomrule
\end{tabular}
\caption{Comparison of attribution methods and their LibraGrad-enhanced versions on the CLIP-H model.}
\label{tbl:SRG_m_2}
\end{table}
  \endgroup%

\end{table}

\FloatBarrier
\clearpage
\subsubsection{DeiT3-H}
\label{per_model:DeiT3-H}

\begin{table}[ht]
  \fontsize{9.5pt}{8.5pt}\selectfont

  \setlength{\tabcolsep}{2pt}
  \centering
  \begingroup%
  \renewenvironment{table}[1][]%
    {\begin{center}}%
    {\end{center}}%
  \begin{table}[h]
\centering
\begin{tabular}{lccccc}
\toprule
Method & \multicolumn{2}{c}{MIF Deletion (GT)} & \multicolumn{2}{c}{MIF Deletion (Predicted)} & Segmentation \\
  & Accuracy & AOPC & Accuracy & AOPC & AP \\
\cmidrule(r){1-1}
\cmidrule(lr){2-3}
\cmidrule(lr){4-5}
\cmidrule(l){6-6}
Random & 35.6 \textcolor{gray}{±0.1} & 16.6 \textcolor{gray}{±0.2} & 29.0 \textcolor{gray}{±0.1} & 19.2 \textcolor{gray}{±0.2} & 37.8 \textcolor{gray}{±0.3} \\
RawAtt & 56.1 \textcolor{gray}{±0.1} & 33.3 \textcolor{gray}{±0.3} & 52.0 \textcolor{gray}{±0.1} & 37.2 \textcolor{gray}{±0.2} & 49.7 \textcolor{gray}{±0.3} \\
Attention Rollout & 37.1 \textcolor{gray}{±0.1} & 19.0 \textcolor{gray}{±0.2} & 31.2 \textcolor{gray}{±0.1} & 21.9 \textcolor{gray}{±0.2} & 34.1 \textcolor{gray}{±0.3} \\
AliLRP & 59.6 \textcolor{gray}{±0.1} & 37.3 \textcolor{gray}{±0.3} & 56.3 \textcolor{gray}{±0.1} & 41.7 \textcolor{gray}{±0.2} & 52.2 \textcolor{gray}{±0.3} \\
AttnLRP & 45.4 \textcolor{gray}{±0.1} & 28.1 \textcolor{gray}{±0.3} & 40.7 \textcolor{gray}{±0.1} & 31.7 \textcolor{gray}{±0.2} & 36.0 \textcolor{gray}{±0.3} \\
DecompX & 51.6 \textcolor{gray}{±0.1} & 32.2 \textcolor{gray}{±0.3} & 47.2 \textcolor{gray}{±0.1} & 35.9 \textcolor{gray}{±0.2} & 49.5 \textcolor{gray}{±0.3} \\
Integrated Gradients & 43.7 \textcolor{gray}{±0.1} & 24.9 \textcolor{gray}{±0.3} & 33.2 \textcolor{gray}{±0.1} & 22.8 \textcolor{gray}{±0.2} & 38.9 \textcolor{gray}{±0.3} \\
\midrule
\IxG{}  & 40.4 \textcolor{gray}{±0.1} & 21.9 \textcolor{gray}{±0.3} & 35.1 \textcolor{gray}{±0.1} & 25.1 \textcolor{gray}{±0.2} & 39.6 \textcolor{gray}{±0.3} \\
\textbf{Libra \IxG{} } & 52.1 \textcolor{gray}{±0.1} (\textcolor{blue}{+29.2\%}) & 32.4 \textcolor{gray}{±0.3} (\textcolor{blue}{+48.1\%}) & 47.7 \textcolor{gray}{±0.1} (\textcolor{blue}{+36.0\%}) & 36.3 \textcolor{gray}{±0.2} (\textcolor{blue}{+44.7\%}) & 49.0 \textcolor{gray}{±0.3} (\textcolor{blue}{+23.8\%}) \\
\midrule
AttCAT & 48.2 \textcolor{gray}{±0.1} & 28.6 \textcolor{gray}{±0.3} & 44.0 \textcolor{gray}{±0.1} & 32.3 \textcolor{gray}{±0.3} & 41.7 \textcolor{gray}{±0.3} \\
\textbf{Libra AttCAT} & \underline{72.6} \textcolor{gray}{±0.1} (\textcolor{blue}{+50.6\%}) & \underline{47.6} \textcolor{gray}{±0.3} (\textcolor{blue}{+66.5\%}) & \underline{70.5} \textcolor{gray}{±0.1} (\textcolor{blue}{+60.2\%}) & \underline{52.8} \textcolor{gray}{±0.2} (\textcolor{blue}{+63.4\%}) & 60.1 \textcolor{gray}{±0.3} (\textcolor{blue}{+44.1\%}) \\
\midrule
GenAtt & 67.2 \textcolor{gray}{±0.1} & 43.3 \textcolor{gray}{±0.3} & 64.6 \textcolor{gray}{±0.1} & 48.1 \textcolor{gray}{±0.2} & 66.2 \textcolor{gray}{±0.2} \\
\textbf{Libra GenAtt} & 69.1 \textcolor{gray}{±0.1} (\textcolor{blue}{+2.9\%}) & 45.0 \textcolor{gray}{±0.3} (\textcolor{blue}{+3.8\%}) & 66.5 \textcolor{gray}{±0.1} (\textcolor{blue}{+3.0\%}) & 49.7 \textcolor{gray}{±0.2} (\textcolor{blue}{+3.5\%}) & \textbf{76.5} \textcolor{gray}{±0.2} (\textcolor{blue}{+15.5\%}) \\
\midrule
TokenTM & 66.2 \textcolor{gray}{±0.1} & 42.6 \textcolor{gray}{±0.3} & 63.3 \textcolor{gray}{±0.1} & 47.2 \textcolor{gray}{±0.2} & 61.7 \textcolor{gray}{±0.2} \\
\textbf{Libra TokenTM} & 68.1 \textcolor{gray}{±0.1} (\textcolor{blue}{+2.8\%}) & 44.1 \textcolor{gray}{±0.3} (\textcolor{blue}{+3.6\%}) & 65.2 \textcolor{gray}{±0.1} (\textcolor{blue}{+3.0\%}) & 48.8 \textcolor{gray}{±0.2} (\textcolor{blue}{+3.3\%}) & 70.8 \textcolor{gray}{±0.2} (\textcolor{blue}{+14.7\%}) \\
\midrule
GradCAM+ & 49.5 \textcolor{gray}{±0.1} & 28.3 \textcolor{gray}{±0.3} & 44.5 \textcolor{gray}{±0.1} & 31.8 \textcolor{gray}{±0.2} & 60.3 \textcolor{gray}{±0.4} \\
\textbf{Libra GradCAM+} & 52.6 \textcolor{gray}{±0.1} (\textcolor{blue}{+6.2\%}) & 31.4 \textcolor{gray}{±0.3} (\textcolor{blue}{+10.9\%}) & 48.7 \textcolor{gray}{±0.1} (\textcolor{blue}{+9.6\%}) & 35.5 \textcolor{gray}{±0.2} (\textcolor{blue}{+11.7\%}) & 46.7 \textcolor{gray}{±0.4} (\textcolor{coral}{-22.5\%}) \\
\midrule
HiResCAM & 32.5 \textcolor{gray}{±0.1} & 15.0 \textcolor{gray}{±0.2} & 25.8 \textcolor{gray}{±0.1} & 17.4 \textcolor{gray}{±0.2} & 41.3 \textcolor{gray}{±0.3} \\
\textbf{Libra HiResCAM} & 57.4 \textcolor{gray}{±0.1} (\textcolor{blue}{+76.7\%}) & 35.4 \textcolor{gray}{±0.3} (\textcolor{blue}{+136.8\%}) & 53.8 \textcolor{gray}{±0.1} (\textcolor{blue}{+108.5\%}) & 39.7 \textcolor{gray}{±0.2} (\textcolor{blue}{+127.5\%}) & \underline{76.3} \textcolor{gray}{±0.3} (\textcolor{blue}{+84.9\%}) \\
\midrule
XGradCAM+ & 49.1 \textcolor{gray}{±0.1} & 27.9 \textcolor{gray}{±0.3} & 45.1 \textcolor{gray}{±0.1} & 31.8 \textcolor{gray}{±0.2} & 48.9 \textcolor{gray}{±0.4} \\
\textbf{Libra XGradCAM+} & 68.8 \textcolor{gray}{±0.1} (\textcolor{blue}{+40.2\%}) & 44.2 \textcolor{gray}{±0.3} (\textcolor{blue}{+58.3\%}) & 66.1 \textcolor{gray}{±0.1} (\textcolor{blue}{+46.7\%}) & 49.0 \textcolor{gray}{±0.2} (\textcolor{blue}{+54.2\%}) & 59.4 \textcolor{gray}{±0.3} (\textcolor{blue}{+21.5\%}) \\
\midrule
FullGrad+ & 45.8 \textcolor{gray}{±0.1} & 26.2 \textcolor{gray}{±0.3} & 41.9 \textcolor{gray}{±0.1} & 30.0 \textcolor{gray}{±0.3} & 40.6 \textcolor{gray}{±0.3} \\
\textbf{Libra FullGrad+} & \textbf{73.5} \textcolor{gray}{±0.1} (\textcolor{blue}{+60.4\%}) & \textbf{48.5} \textcolor{gray}{±0.3} (\textcolor{blue}{+84.8\%}) & \textbf{71.5} \textcolor{gray}{±0.1} (\textcolor{blue}{+70.7\%}) & \textbf{53.7} \textcolor{gray}{±0.2} (\textcolor{blue}{+78.8\%}) & 65.1 \textcolor{gray}{±0.3} (\textcolor{blue}{+60.4\%}) \\
\bottomrule
\end{tabular}
\caption{Comparison of attribution methods and their LibraGrad-enhanced versions on the DeiT3-H model. We report faithfulness metrics using Most-Influential-First Deletion, MIF with ground-truth (GT) and predicted labels, including Accuracy and Area Over Perturbation Curve (AOPC) and Segmentation Average Precision (AP). The results demonstrate that composing existing methods with LibraGrad significantly enhances their performance across all metrics.}
\label{tbl:MIF_m_2}
\end{table}
  \endgroup%

\end{table}

\begin{table}[ht]
  \fontsize{11pt}{8.5pt}\selectfont

  \setlength{\tabcolsep}{4pt}
  \centering

  \begingroup%
  \renewenvironment{table}[1][]%
    {\begin{center}}%
    {\end{center}}%
  \begin{table}[h]
\centering
\begin{tabular}{lcccc}
\toprule
Method & \multicolumn{2}{c}{LIF Deletion (GT)} & \multicolumn{2}{c}{LIF Deletion (Predicted)} \\
  & Accuracy & AOPC & Accuracy & AOPC \\
\cmidrule(r){1-1}
\cmidrule(lr){2-3}
\cmidrule(l){4-5}
Random & 64.2 \textcolor{gray}{±0.1} & 83.5 \textcolor{gray}{±0.2} & 70.7 \textcolor{gray}{±0.1} & 81.1 \textcolor{gray}{±0.1} \\
RawAtt & 70.9 \textcolor{gray}{±0.1} & 86.3 \textcolor{gray}{±0.2} & 78.4 \textcolor{gray}{±0.1} & 84.3 \textcolor{gray}{±0.2} \\
Attention Rollout & 59.0 \textcolor{gray}{±0.1} & 77.9 \textcolor{gray}{±0.3} & 64.5 \textcolor{gray}{±0.1} & 74.8 \textcolor{gray}{±0.2} \\
AliLRP & 79.5 \textcolor{gray}{±0.1} & 97.5 \textcolor{gray}{±0.2} & 86.1 \textcolor{gray}{±0.1} & 96.0 \textcolor{gray}{±0.1} \\
AttnLRP & 74.1 \textcolor{gray}{±0.1} & 94.2 \textcolor{gray}{±0.2} & 80.8 \textcolor{gray}{±0.1} & 92.2 \textcolor{gray}{±0.2} \\
DecompX & 75.8 \textcolor{gray}{±0.1} & 95.2 \textcolor{gray}{±0.2} & 83.1 \textcolor{gray}{±0.1} & 93.4 \textcolor{gray}{±0.1} \\
Integrated Gradients & 71.5 \textcolor{gray}{±0.1} & 91.2 \textcolor{gray}{±0.3} & 74.6 \textcolor{gray}{±0.1} & 84.9 \textcolor{gray}{±0.2} \\
\midrule
\IxG{}  & 71.9 \textcolor{gray}{±0.1} & 90.5 \textcolor{gray}{±0.2} & 77.7 \textcolor{gray}{±0.1} & 87.8 \textcolor{gray}{±0.2} \\
\textbf{Libra \IxG{} } & 77.1 \textcolor{gray}{±0.1} (\textcolor{blue}{+7.2\%}) & 95.8 \textcolor{gray}{±0.2} (\textcolor{blue}{+5.9\%}) & 83.7 \textcolor{gray}{±0.1} (\textcolor{blue}{+7.7\%}) & 94.0 \textcolor{gray}{±0.2} (\textcolor{blue}{+7.1\%}) \\
\midrule
AttCAT & 75.3 \textcolor{gray}{±0.1} & 93.6 \textcolor{gray}{±0.2} & 80.5 \textcolor{gray}{±0.1} & 90.6 \textcolor{gray}{±0.2} \\
\textbf{Libra AttCAT} & \underline{81.7} \textcolor{gray}{±0.1} (\textcolor{blue}{+8.4\%}) & \underline{100.0} \textcolor{gray}{±0.2} (\textcolor{blue}{+6.9\%}) & \textbf{87.7} \textcolor{gray}{±0.0} (\textcolor{blue}{+8.9\%}) & \underline{98.6} \textcolor{gray}{±0.1} (\textcolor{blue}{+8.8\%}) \\
\midrule
GenAtt & 76.9 \textcolor{gray}{±0.1} & 94.9 \textcolor{gray}{±0.2} & 85.7 \textcolor{gray}{±0.1} & 93.6 \textcolor{gray}{±0.1} \\
\textbf{Libra GenAtt} & 77.3 \textcolor{gray}{±0.1} (\textcolor{blue}{+0.5\%}) & 95.3 \textcolor{gray}{±0.2} (\textcolor{blue}{+0.5\%}) & 86.0 \textcolor{gray}{±0.1} (\textcolor{blue}{+0.3\%}) & 94.0 \textcolor{gray}{±0.1} (\textcolor{blue}{+0.5\%}) \\
\midrule
TokenTM & 76.2 \textcolor{gray}{±0.1} & 94.2 \textcolor{gray}{±0.2} & 85.0 \textcolor{gray}{±0.1} & 93.0 \textcolor{gray}{±0.1} \\
\textbf{Libra TokenTM} & 76.4 \textcolor{gray}{±0.1} (\textcolor{blue}{+0.4\%}) & 94.6 \textcolor{gray}{±0.2} (\textcolor{blue}{+0.4\%}) & 85.4 \textcolor{gray}{±0.1} (\textcolor{blue}{+0.4\%}) & 93.3 \textcolor{gray}{±0.2} (\textcolor{blue}{+0.4\%}) \\
\midrule
GradCAM+ & 69.3 \textcolor{gray}{±0.1} & 86.7 \textcolor{gray}{±0.2} & 75.8 \textcolor{gray}{±0.1} & 84.4 \textcolor{gray}{±0.2} \\
\textbf{Libra GradCAM+} & 74.1 \textcolor{gray}{±0.1} (\textcolor{blue}{+6.9\%}) & 91.9 \textcolor{gray}{±0.2} (\textcolor{blue}{+6.0\%}) & 80.7 \textcolor{gray}{±0.1} (\textcolor{blue}{+6.4\%}) & 89.8 \textcolor{gray}{±0.2} (\textcolor{blue}{+6.4\%}) \\
\midrule
HiResCAM & 68.1 \textcolor{gray}{±0.1} & 86.2 \textcolor{gray}{±0.2} & 75.5 \textcolor{gray}{±0.1} & 84.2 \textcolor{gray}{±0.1} \\
\textbf{Libra HiResCAM} & 75.2 \textcolor{gray}{±0.1} (\textcolor{blue}{+10.4\%}) & 89.8 \textcolor{gray}{±0.3} (\textcolor{blue}{+4.1\%}) & 80.6 \textcolor{gray}{±0.1} (\textcolor{blue}{+6.8\%}) & 86.9 \textcolor{gray}{±0.2} (\textcolor{blue}{+3.3\%}) \\
\midrule
XGradCAM+ & 71.4 \textcolor{gray}{±0.1} & 89.9 \textcolor{gray}{±0.3} & 77.1 \textcolor{gray}{±0.1} & 87.3 \textcolor{gray}{±0.2} \\
\textbf{Libra XGradCAM+} & 79.6 \textcolor{gray}{±0.1} (\textcolor{blue}{+11.5\%}) & 97.0 \textcolor{gray}{±0.2} (\textcolor{blue}{+7.9\%}) & 86.4 \textcolor{gray}{±0.1} (\textcolor{blue}{+12.0\%}) & 95.3 \textcolor{gray}{±0.1} (\textcolor{blue}{+9.1\%}) \\
\midrule
FullGrad+ & 74.6 \textcolor{gray}{±0.1} & 92.8 \textcolor{gray}{±0.2} & 79.9 \textcolor{gray}{±0.1} & 90.1 \textcolor{gray}{±0.2} \\
\textbf{Libra FullGrad+} & \textbf{81.8} \textcolor{gray}{±0.1} (\textcolor{blue}{+9.7\%}) & \textbf{100.4} \textcolor{gray}{±0.2} (\textcolor{blue}{+8.2\%}) & \underline{87.6} \textcolor{gray}{±0.0} (\textcolor{blue}{+9.6\%}) & \textbf{98.8} \textcolor{gray}{±0.1} (\textcolor{blue}{+9.7\%}) \\
\bottomrule
\end{tabular}
\caption{Comparison of attribution methods and their LibraGrad-enhanced versions on the DeiT3-H model.}
\label{tbl:LIF_m_2}
\end{table}
  \endgroup%

  \begingroup%
  \renewenvironment{table}[1][]%
    {\begin{center}}%
    {\end{center}}%
  \begin{table}[h]
\centering
\begin{tabular}{lcccc}
\toprule
Method & \multicolumn{2}{c}{SRG (GT)} & \multicolumn{2}{c}{SRG (Predicted)} \\
  & Accuracy & AOPC & Accuracy & AOPC \\
\cmidrule(r){1-1}
\cmidrule(lr){2-3}
\cmidrule(l){4-5}
Random & 49.9 \textcolor{gray}{±0.1} & 50.0 \textcolor{gray}{±0.2} & 49.8 \textcolor{gray}{±0.1} & 50.1 \textcolor{gray}{±0.2} \\
RawAtt & 63.5 \textcolor{gray}{±0.1} & 59.8 \textcolor{gray}{±0.2} & 65.2 \textcolor{gray}{±0.1} & 60.7 \textcolor{gray}{±0.2} \\
Attention Rollout & 48.1 \textcolor{gray}{±0.1} & 48.4 \textcolor{gray}{±0.3} & 47.8 \textcolor{gray}{±0.1} & 48.3 \textcolor{gray}{±0.2} \\
AliLRP & 69.6 \textcolor{gray}{±0.1} & 67.4 \textcolor{gray}{±0.2} & 71.2 \textcolor{gray}{±0.1} & 68.8 \textcolor{gray}{±0.2} \\
AttnLRP & 59.7 \textcolor{gray}{±0.1} & 61.2 \textcolor{gray}{±0.3} & 60.8 \textcolor{gray}{±0.1} & 62.0 \textcolor{gray}{±0.2} \\
DecompX & 63.7 \textcolor{gray}{±0.1} & 63.7 \textcolor{gray}{±0.2} & 65.1 \textcolor{gray}{±0.1} & 64.7 \textcolor{gray}{±0.2} \\
Integrated Gradients & 57.6 \textcolor{gray}{±0.1} & 58.1 \textcolor{gray}{±0.3} & 53.9 \textcolor{gray}{±0.1} & 53.8 \textcolor{gray}{±0.2} \\
\midrule
\IxG{}  & 56.1 \textcolor{gray}{±0.1} & 56.2 \textcolor{gray}{±0.2} & 56.4 \textcolor{gray}{±0.1} & 56.4 \textcolor{gray}{±0.2} \\
\textbf{Libra \IxG{} } & 64.6 \textcolor{gray}{±0.1} (\textcolor{blue}{+15.1\%}) & 64.1 \textcolor{gray}{±0.2} (\textcolor{blue}{+14.1\%}) & 65.7 \textcolor{gray}{±0.1} (\textcolor{blue}{+16.5\%}) & 65.1 \textcolor{gray}{±0.2} (\textcolor{blue}{+15.4\%}) \\
\midrule
AttCAT & 61.8 \textcolor{gray}{±0.1} & 61.1 \textcolor{gray}{±0.3} & 62.3 \textcolor{gray}{±0.1} & 61.5 \textcolor{gray}{±0.2} \\
\textbf{Libra AttCAT} & \underline{77.1} \textcolor{gray}{±0.1} (\textcolor{blue}{+24.9\%}) & \underline{73.8} \textcolor{gray}{±0.2} (\textcolor{blue}{+20.8\%}) & \underline{79.1} \textcolor{gray}{±0.1} (\textcolor{blue}{+27.0\%}) & \underline{75.7} \textcolor{gray}{±0.2} (\textcolor{blue}{+23.1\%}) \\
\midrule
GenAtt & 72.1 \textcolor{gray}{±0.1} & 69.1 \textcolor{gray}{±0.2} & 75.2 \textcolor{gray}{±0.1} & 70.8 \textcolor{gray}{±0.2} \\
\textbf{Libra GenAtt} & 73.2 \textcolor{gray}{±0.1} (\textcolor{blue}{+1.6\%}) & 70.2 \textcolor{gray}{±0.2} (\textcolor{blue}{+1.5\%}) & 76.2 \textcolor{gray}{±0.1} (\textcolor{blue}{+1.4\%}) & 71.9 \textcolor{gray}{±0.2} (\textcolor{blue}{+1.5\%}) \\
\midrule
TokenTM & 71.2 \textcolor{gray}{±0.1} & 68.4 \textcolor{gray}{±0.2} & 74.2 \textcolor{gray}{±0.1} & 70.1 \textcolor{gray}{±0.2} \\
\textbf{Libra TokenTM} & 72.3 \textcolor{gray}{±0.1} (\textcolor{blue}{+1.5\%}) & 69.3 \textcolor{gray}{±0.2} (\textcolor{blue}{+1.4\%}) & 75.3 \textcolor{gray}{±0.1} (\textcolor{blue}{+1.6\%}) & 71.0 \textcolor{gray}{±0.2} (\textcolor{blue}{+1.4\%}) \\
\midrule
GradCAM+ & 59.4 \textcolor{gray}{±0.1} & 57.5 \textcolor{gray}{±0.2} & 60.1 \textcolor{gray}{±0.1} & 58.1 \textcolor{gray}{±0.2} \\
\textbf{Libra GradCAM+} & 63.4 \textcolor{gray}{±0.1} (\textcolor{blue}{+6.6\%}) & 61.6 \textcolor{gray}{±0.2} (\textcolor{blue}{+7.2\%}) & 64.7 \textcolor{gray}{±0.1} (\textcolor{blue}{+7.6\%}) & 62.6 \textcolor{gray}{±0.2} (\textcolor{blue}{+7.8\%}) \\
\midrule
HiResCAM & 50.3 \textcolor{gray}{±0.1} & 50.6 \textcolor{gray}{±0.2} & 50.7 \textcolor{gray}{±0.1} & 50.8 \textcolor{gray}{±0.2} \\
\textbf{Libra HiResCAM} & 66.3 \textcolor{gray}{±0.1} (\textcolor{blue}{+31.8\%}) & 62.6 \textcolor{gray}{±0.3} (\textcolor{blue}{+23.7\%}) & 67.2 \textcolor{gray}{±0.1} (\textcolor{blue}{+32.7\%}) & 63.3 \textcolor{gray}{±0.2} (\textcolor{blue}{+24.6\%}) \\
\midrule
XGradCAM+ & 60.2 \textcolor{gray}{±0.1} & 58.9 \textcolor{gray}{±0.3} & 61.1 \textcolor{gray}{±0.1} & 59.6 \textcolor{gray}{±0.2} \\
\textbf{Libra XGradCAM+} & 74.2 \textcolor{gray}{±0.1} (\textcolor{blue}{+23.2\%}) & 70.6 \textcolor{gray}{±0.2} (\textcolor{blue}{+19.8\%}) & 76.3 \textcolor{gray}{±0.1} (\textcolor{blue}{+24.8\%}) & 72.2 \textcolor{gray}{±0.2} (\textcolor{blue}{+21.2\%}) \\
\midrule
FullGrad+ & 60.2 \textcolor{gray}{±0.1} & 59.5 \textcolor{gray}{±0.3} & 60.9 \textcolor{gray}{±0.1} & 60.1 \textcolor{gray}{±0.3} \\
\textbf{Libra FullGrad+} & \textbf{77.6} \textcolor{gray}{±0.1} (\textcolor{blue}{+29.0\%}) & \textbf{74.4} \textcolor{gray}{±0.2} (\textcolor{blue}{+25.1\%}) & \textbf{79.6} \textcolor{gray}{±0.1} (\textcolor{blue}{+30.6\%}) & \textbf{76.3} \textcolor{gray}{±0.2} (\textcolor{blue}{+27.0\%}) \\
\bottomrule
\end{tabular}
\caption{Comparison of attribution methods and their LibraGrad-enhanced versions on the DeiT3-H model.}
\label{tbl:SRG_m_2}
\end{table}
  \endgroup%

\end{table}

\clearpage{}
\twocolumn
\section{Related Work}
\label{apn:relatedWork}

Input attribution methods are techniques designed to quantify the influence of individual input features, or groups of them, on a model's output\nightCite{Samek2021ExplainingDN, simonyan-2014-deep, Binder2016LayerWiseRP, kindermans-2016-investigating, Zhang2016TopDownNA, Springenberg2014StrivingFS, LYU2022TowardsFM, Madsen2021PosthocIF}. Input attribution methods can assist in understanding a model's decision locally for a single input considered in isolation. They also act as foundational elements for more advanced explanation techniques. For instance, in concept-based explanation methods like CRAFT\nightCite{Fel2022CRAFTCR}, attribution methods are employed for two main purposes: to quantify the impact of each activated concept and to identify the specific input features responsible for activating these concepts.

Attribution methods have a wide array of applications beyond merely explaining model outputs to humans\nightCite{Weber2022BeyondEO, Ede2022ExplainTN, Selvaraju2019TakingAH, voita-2019-analyzing}. They are useful for enhancing the robustness of models against out-of-distribution data, spurious correlations, and adversarial inputs\nightCite{Noohdani2024DecomposeandComposeAC, Chefer2022OptimizingRM, Anders2020XAIFA, Yang2019MLLOODA}. Additionally, attribution methods have been employed to improve the performance of text-to-image models\nightCite{paiss-2022-token, Chefer2023AttendandExciteAS, Kim2023DenseTG}. Furthermore, adapting forward-mode attribution methods has been explored for on-the-fly feature pruning\nightCite{Fayyaz2021AdaptiveTS, Modarressi2022AdapLeRSU} and model quantization\nightCite{Becking2021ECQxEQ}. Attribution methods have been utilized to construct more effective adversarial attacks against models\nightCite{Huang2022TransferableAA, Zhang2022ImprovingAT, Wu2020BoostingTT}.

Given a multi-output neural model, let $f: \mathbb{R}^n \rightarrow \mathbb{R}$ be a selected output function. For instance, if $\text{Model}(x) = (p_1, ..., p_k)$ represents class probabilities, we might choose $f(x) = p_i$ to analyze the model's prediction for the $i$-th class. An attribution method $A$ generates relevance scores $A(f)(x)_i$ for each feature $x_i$.

\subsection{Gradient-Based Attribution Methods}
\nightParagraph{\IxG{}} \IxGShort{}\nightCite{Shrikumar2016NotJA, Shrikumar2017LearningIF, Ancona2017TowardsBU} assigns feature relevance by
$\IxGShort{}(f)(x) = x \odot \nabla_x f(x)
$, where $\odot$ denotes element-wise multiplication.
\nightParagraph{FullGrad}
Expanding on \IxG{}, FullGrad\nightCite{Srinivas2019FullGradientRF} includes not only the input features but also the bias terms of each layer in the neural network. The FullGrad attribution map is calculated as:
\begin{equation*}
\text{FullGrad}(f)(x_{0}) = \IxGShort{}(f)(x_0) + \sum_{l=0}^{L-1} \sum_{b \in B_l} \IxGShort{}(f_{b})(b)
\end{equation*}
where $\IxGShort{}(f)(x_0)$ denotes the \IxG{} for the input $x_0$, and $\IxGShort{}(f_{b})(b)$ is the \IxG{} attribution map of the sub-network $f_{b}$ with a bias term $b$ from layer $l$ as the input.
Also, $f_b$ is the sub-network of $f$ starting from the bias term $b$ and going until the end of the model, whereas $B_l$ denotes the set of all bias terms in layer $l$.
\FullGradPLUSLong{} (\ShortenedTo{\FullGradPLUS{}})\nightCite{skipplus-cvprw24} is defined as follows:
\begin{equation*}
\begin{aligned}
  \FullGradPLUS{} & (f)(x_{0}) = \\
  & \sum_{l=0}^{L-1} \IxGShort{}(f_l)(x_l) + \sum_{l=0}^{L-1} \sum_{b \in B_l} \IxGShort{}(f_{b})(b)
\end{aligned}
\end{equation*}
where $\IxGShort{}(f_l)(x_l)$ is the \IxG{} attribution map of the sub-network $f_l$ with input $x_l$. \FullGradPLUS{} aggregates the input attribution maps of each layer along with the attribution maps of all bias terms in each layer.

\nightParagraph{\IG{}}
\IGShort{}\nightCite{pmlr-v70-sundararajan17a} computes attributions \wrt{} a baseline input $\bar{x}$ (\eg{}, zero):
\begin{equation*}
\IGShort{}(f)(x) = (x-\bar{x}) \odot \int_{\alpha=0}^1 \nabla_x f(\bar{x}+\alpha(x-\bar{x})) d\alpha
\end{equation*}

In practice, we approximate the integral using a 50-step Riemann summation.

\paragraph{GradCAM.}
GradCAM\nightCite{selvaraju-2017-gradcam} averages the gradient signal across each channel before multiplying it with the input, and operates on the last layer of the network:
\begin{itemize}
\item \(A^k\): the k-th channel of the feature map in the final layer
\item \(c\): the class w.r.t. which the attribution map is computed
\item \(y^c\): the class score (logit)
\item Gradients are averaged over the width and height dimensions (indexed by i and j respectively) to obtain the neuron (channel) importance weights \(\alpha_k^c\):
\end{itemize}
\begin{equation*}
\alpha{}_{k}^c =
\overbrace{
    \frac{1}{Z}\sum_{i}\sum_{j}
}^{\text{global average pooling}}
\hspace{-17pt}
\underbrace{
    \vphantom{\sum_{i}\sum_{j}} \frac{\partial y^c}{\partial A_{ij}^{k}}
}_{\text{gradients via backprop}}
\end{equation*}

\paragraph{XGradCAM+.}
XGradCAM weights the gradients by their corresponding activation value when computing the spatial average\nightCite{Fu2020AxiombasedGT}. XGradCAM was proposed on ReLU CNNs where the activations were always positive, hence they did not specify using the absolute value of the activations in the above computation, as is more intuitive. The variant with absolute activations is named XGradCAM+\nightCite{skipplus-cvprw24}.

\paragraph{HiResCAM.}
HiResCAM\nightCite{Draelos2020UseHI} is equivalent to \IxG{} on the last layer of the model. (Standard \IxG{} is applied on the first layer of the model.)

\paragraph{PLUS.} PLUS\nightCite{skipplus-cvprw24} is a way for attribution methods to better aggregate information across layers.

\subsubsection{Gradient-Attention Hybrids}
\paragraph{AttCAT.} \AttCAT{}\nightCite{qiang-2022-attcat} combines attention weights with \IxG{} to create a hybrid attribution method. The approach operates by first computing the input-times-gradient attribution at each layer, then weighting these attributions using the attention weights from the corresponding attention heads. The method addresses the limitations of pure attention-based or pure gradient-based approaches by leveraging both sources of information. By incorporating both attention patterns and gradient information, AttCAT can better capture the model's decision-making process, particularly in cases where either attention or gradient alone might miss important feature interactions. The final attribution map is computed by aggregating these weighted scores across all layers and attention heads.

\paragraph{TransAtt.}
TransAtt\nightCite{chefer-2021-transformer} employs the Deep Taylor Decomposition technique\nightCite{montavon-2017-explaining} to attribute local relevance and subsequently propagates these relevance scores through the entire architecture of a Transformer model.
This process effectively enables the backward propagation of information across all layers, starting from the output and extending back to the input. 
Additionally, this method incorporates gradients of attention weights. 
The method's functioning can be summarized as follows:
\[Rollout\left(\expected{H \coloneqq \text{Heads}}{\left(\relScore\odot\attngrad\right)^+}\right),\]
where \(\relScore\) stands for the relevancy scores of attention weights.
The Rollout technique is a method to aggregate the layer-wise attribution maps. We refer the reader to\nightCite{abnar-2020-quantifying} for a detailed overview.

\paragraph{GenAtt.}
The dependence of TransAtt on specific rules for the propagation of relevance scores imposes limitations on its capacity to furnish explanations for various types of Transformer architectures. 
To cope with this issue, GenAtt\nightCite{chefer-2021-generic} attempts to explain predictions for any Transformer-based architecture by using the attention weights in each block to update the relevancy maps, as demonstrated by the following expression:
\[Rollout\left(\expected{H \coloneqq \text{Heads}}{\left(\attn\odot\attngrad\right)^+}\right).\]

The notation \(()^+\) denotes a filtering through the ReLU function.
\cite{chefer-2021-generic} show that GenAtt is at least as effective as TransAtt, if not better.

\nightParagraph{\TokenTM{}} \TokenTM{}\nightCite{Wu2024TokenTM} further improves GenAtt by taking token transformations into account.

\subsection{LRP Methods}
Layer-wise Relevance Propagation (LRP) is a principled attribution method that propagates relevance scores backward through a neural network by following specific propagation rules.

\paragraph{AliLRP.} AliLRP\nightCite{pmlr-v162-ali22a} extends traditional LRP for Transformer architectures by introducing specialized propagation rules that offer better numerical stability.

\paragraph{AttnLRP.} AttnLRP\nightCite{pmlr-v235-achtibat24a} extends LRP to handle attention layers.

\subsection{\ForwardAttn{} Methods}

\paragraph{Attention×Input\_Norm (AttIN).}
\citet{kobayashi-2020-attention} multiply the attention weights by the norms of the vectors corresponding to each attention weight.
\citet{kobayashi-2021-incorporating} extends AttIN to also incorporate the residual connections.

\paragraph{GlobEnc \& ALTI.}
AttIN assumes that tokens retain their original identity. As each self-attention module mixes all the tokens, this assumption might not necessarily hold.
Using gradient-based techniques,\nightCitet{brunner-2020-identifiability} studies contextual information aggregation across the model.
Following\nightCitet{brunner-2020-identifiability} work, the global token attribution analysis method GlobEnc\nightCite{modarressi-2022-globenc} further extends AttIN by including the Transformer block's second normalization layer in its analysis.
In parallel with GlobEnc, the Aggregation of Layer-Wise Token-to-Token Interactions method ALTI\nightCite{ferrando-2022-measuring} was introduced. ALTI shares core concepts with GlobEnc, but the two differ in certain mathematical specifics.

\paragraph{DecompX.}

DecompX\nightCite{modarressi-2023-decompx} enhances GlobEnc by integrating the one element previously overlooked by GlobEnc: the MLP module in the Encoder Transformer layer. This inclusion enables DecompX to generate a set of decomposed vectors that collectively sum up to the actual output vector. Unlike GlobEnc and ALTI, which require computing and aggregating layer-wise attribution maps using techniques like Rollout, DecompX facilitates the direct propagation of these decomposed vectors across layers. This capability allows for the direct computation of attribution maps from any layer to any other layer.

\subsection{Black-Box Methods}
Black-box attribution methods treat the model as an opaque entity, (partially) disregarding its internal structure and gradients. These methods typically involve perturbing the input and observing the corresponding changes in the model's output to infer the importance of each input feature. However, this approach often comes with significant computational costs due to the need for multiple model evaluations. In contrast, white-box methods leverage the internal structure and gradients of the model, providing a more efficient and fine-grained understanding of the model's behavior.

In this paper, we focus on white-box methods for several reasons. Firstly, they offer a more computationally efficient approach compared to black-box methods. Secondly, and more importantly, black-box methods can be seen as directly optimizing the faithfulness metrics on which we evaluate the attribution methods. This raises concerns related to Goodhart's law, which states that when a measure becomes a target, it ceases to be a good measure. In other words, the faithfulness metrics we use are merely proxies for the ultimate desirable properties we seek in attribution methods. By directly optimizing these metrics, black-box methods may inadvertently introduce biases or artifacts that undermine the true faithfulness of the attributions. Therefore, to avoid this potential pitfall and maintain a more objective evaluation, we refrain from including comparisons with black-box methods in this study, acknowledging that they have different trade-offs and use cases.

\paragraph{LIME}\hspace*{-7pt}\cite{Ribeiro2016WhySI} explains the predictions of any classifier by learning a local interpretable model around the prediction.

\paragraph{RISE}\hspace*{-7pt}\cite{Petsiuk2018RISERI} is a black-box approach that generates an importance map indicating the saliency of each pixel for the model's prediction by probing the model with randomly masked versions of the input image and obtaining the corresponding outputs.

\paragraph{PAMI}\hspace*{-7pt}\cite{Shi2023PAMIPI} masks the majority of the input and uses the corresponding model output as the relative contribution of the preserved input part to the original model prediction.

\paragraph{ScoreCAM}\hspace*{-7pt}\cite{Wang2019ScoreCAMSV} is a post-hoc visual explanation method based on class activation mapping that eliminates the dependence on gradients by obtaining the weight of each activation map through its forward passing score on the target class.

\paragraph{ViT-CX}\hspace*{-7pt}\cite{Xie2022ViTCXCE} adapts ScoreCAM for ViTs.

\paragraph{AtMan}\hspace*{-7pt}\cite{deb-2023-atman} is a perturbation method that manipulates the attention mechanisms of transformers to produce relevance maps for the input with respect to the output prediction.

\paragraph{HSIC}\hspace*{-7pt}\cite{Novello2022MakingSO} is a black-box attribution method based on the Hilbert-Schmidt Independence Criterion, measuring the dependence between regions of an input image and the model's output using kernel embeddings of distributions.

\section*{Acknowledgements}
My heartfelt thanks go to my family for their steadfast support.

\end{document}